\documentclass{svmono}
\usepackage{microtype}
\usepackage{newtxmath}
\usepackage[T1]{fontenc}
\usepackage[tt=false]{libertine}

\usepackage{amsmath}
\usepackage{shorttoc}

\usepackage{xcolor}
\usepackage{multicol}
\usepackage{listings}
\PassOptionsToPackage{hyphens}{url}
\usepackage[colorlinks=false,linkbordercolor=green,linktocpage=true,pagebackref=true]{hyperref}
\usepackage{tabularx}
\usepackage{graphicx}
\usepackage[toc]{glossaries}
\usepackage{tikz}
\usetikzlibrary{positioning}
\usetikzlibrary{arrows}
\usetikzlibrary{chains}

\usepackage{bm}
\usepackage{csquotes}
\usepackage{rotating}

\makeglossaries

\usepackage[ruled,chapter]{algorithm}
\usepackage{algorithmicx}
\usepackage{algpseudocode}
\DeclareMathOperator*{\argmax}{arg\,max}
\DeclareMathOperator*{\argmin}{arg\,min}
\newcommand{\Tau}{\mathcal{T}}

\usepackage{tcolorbox}

\newcolumntype{P}[1]{>{\centering\arraybackslash}p{#1}}
\newcolumntype{M}[1]{>{\centering\arraybackslash}m{#1}}

\newcommand*\diff{\mathop{}\!\mathrm{d}}

\usepackage{makeidx}
\makeindex

\title{{\huge\bf Deep Reinforcement Learning}\\ \ \\
\centering{\includegraphics[width=7cm]{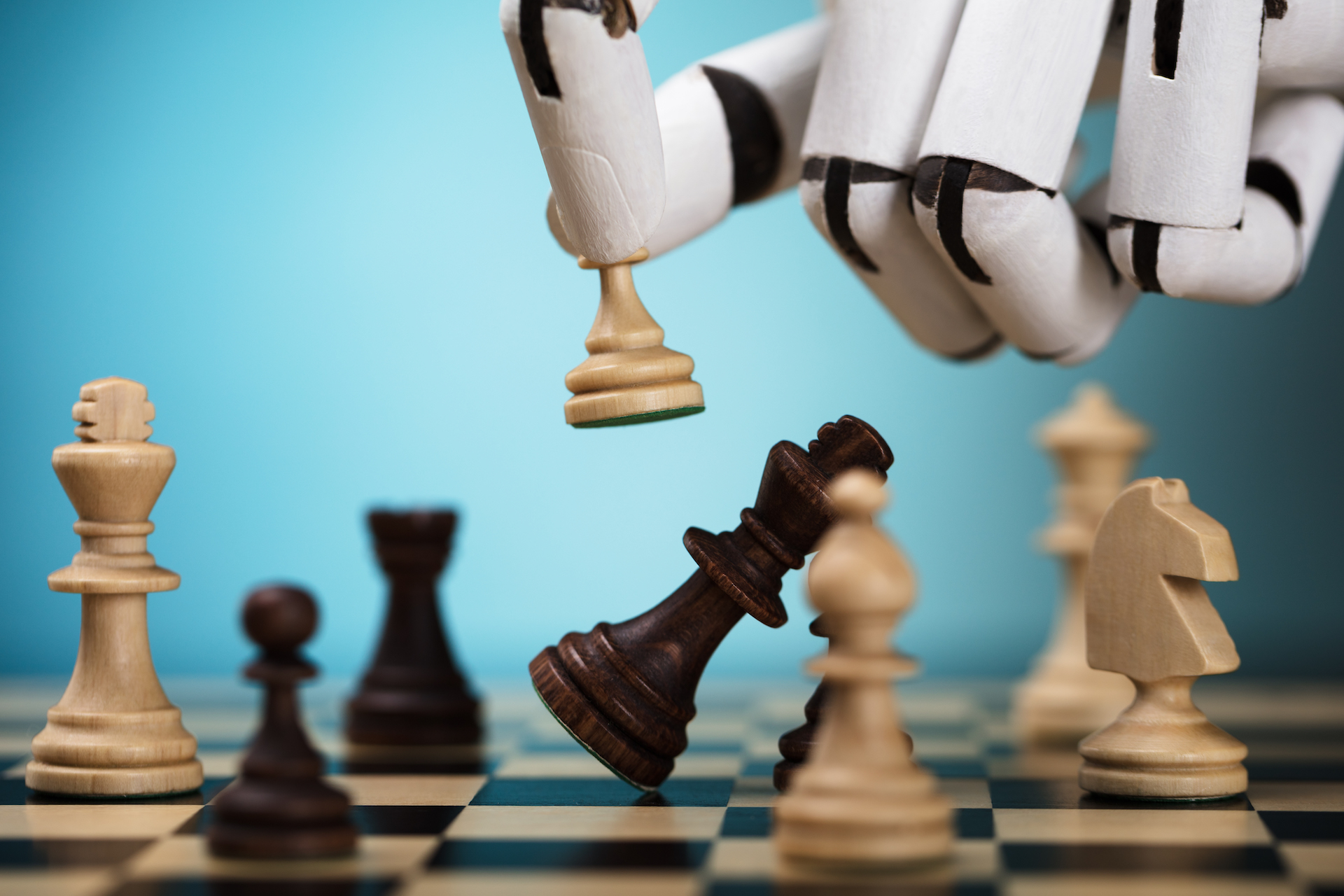}}}

\author{Aske Plaat}
\date{\today}

\begin{document}
\maketitle
\frontmatter

\newpage
\vskip 10cm
\noindent This is a preprint of the following work:\\ \emph{Aske Plaat,\\ Deep
Reinforcement Learning,\\ 2022,\\ Springer Nature,}\\ reproduced with
permission of Springer Nature Singapore Pte Ltd.\\ The final authenticated version
is available online at: \url{https://doi.org/10.1007/978-981-19-0638-1}

\newpage


\newglossaryentry{xlearning} 
{
    name={learning},
    description={agent samples feedback on irreversible actions from environment; environment has state, no undo}
}  

\newglossaryentry{reinforcement learning}
{
  name={reinforcement learning},
  description={agent learns a policy for a sequential decision problem from environment feedback on its actions}
}

\newglossaryentry{deep learning}
{
    name={deep learning},
    description={training a deep neural network to approximate a function, used for high-dimensional problems}
}

\newglossaryentry{deep reinforcement learning}
{
    name={deep reinforcement learning},
    description={approximating value, policy, and transition functions with a deep neural network}
}

\newglossaryentry{deep supervised learning}
{
    name={deep supervised learning},
    description={approximating a function with a deep neural network; often for regression of image classification}
}

\newglossaryentry{supervised learning}
{
    name={supervised learning},
    description={training a predictive model on a labeled dataset}
}

\newglossaryentry{unsupervised learning}
{
    name={unsupervised learning},
    description={clustering elements in an unlabeled  dataset based on an inherent metric}
}

\newglossaryentry{machine learning}
{
    name={machine learning},
    description={learning a function or model from data}
}

\newglossaryentry{planning}
{
    name={planning},
    description={agent performs reversible actions with local transition model; local state, backtracking, undo}
}

\newglossaryentry{state}
{
    name={\ensuremath{S}},
    description={state}
}  

\newglossaryentry{action}
{
    name={\ensuremath{A}},
    description={action in a state}
}  

\newglossaryentry{transition}
{
    name={\ensuremath{T}},
    description={transition}
}  

\newglossaryentry{reward}
{
    name={\ensuremath{R}},
    description={reward}
}

\newglossaryentry{data}
{
    name={\ensuremath{D}},
    description={dataset}
}

\newglossaryentry{options}
{
    name={\ensuremath{\Omega}},
    description={set of options $\omega$ (hierarchical reinforcement learning)}
}

\newglossaryentry{initiation}
{
    name={\ensuremath{I_\omega}},
    description={initiation set for option $\omega$}
}

\newglossaryentry{termination}
{
    name={\ensuremath{\beta_\omega(s)}},
    description={termination condition for option $\omega$ at state $s$}
}

\newglossaryentry{trace}
{
    name={\ensuremath{\tau}},
    description={trajectory, trace, episode, sequence}
}  

\newglossaryentry{policy}
{
    name={\ensuremath{\pi}},
    description={policy}
}

\newglossaryentry{task}
{
    name={\ensuremath{\mathcal{T}_i}},
    description={base-learning task $\mathcal{T}_i=(D_{i,train},\mathcal{L}_i)$, part of a meta-learning task}
}

\newglossaryentry{loss}
{
    name={\ensuremath{\mathcal{L}}},
    description={loss function}
}

\newglossaryentry{value}
{
    name={\ensuremath{V}},
    description={value}
}  

\newglossaryentry{state-action}
{
    name={\ensuremath{Q}},
    description={state-action value}
}  

\newglossaryentry{advantage}
{
    name={\ensuremath{A}},
    description={advantage function in actor critic}
}

\newglossaryentry{cp}
{
    name={\ensuremath{C_p}},
    description={exploration/exploitation constant in MCTS; high is more exploration}
}

\newglossaryentry{alpha}
{
    name={\ensuremath{\alpha}},
    description={learning rate}
}

\newglossaryentry{omega-hyp}
{
    name={\ensuremath{\omega}},
    description={hyperparameters (meta-learning)}
}

\newglossaryentry{omega-hier}
{
    name={\ensuremath{\omega}},
    description={option (hierarchical reinforcement learning)}
}

\newglossaryentry{theta}
{
    name={\ensuremath{\theta}},
    description={parameters, weights in the neural network}
}

\newglossaryentry{phi}
{
    name={\ensuremath{\phi}},
    description={parameters for the value network in actor critic (as opposed to $\theta$, the policy parameters)}
}  

\newglossaryentry{greedy}
{
    name={\ensuremath{\epsilon}-greedy},
    description={exploration/exploitation rule that selects an $\epsilon$ fraction of random exploration actions}
}

\newglossaryentry{gamma}
{
    name={\ensuremath{\gamma}},
    description={discount rate, to reduce the importance of future rewards}
}

\newglossaryentry{transfer learning}
{
    name={transfer learning},
    description={using part of a network (pretraining) to speedup
      learning (finetuning) on a new dataset}
}

\newglossaryentry{meta-learning}
{
    name={meta-learning},
    description={learning to learn hyperparameters; Use a sequence of related tasks to
      learn a new task quicker}
}  

\newglossaryentry{sequential decision problem}
{
        name={sequential decision problem},
        description={problem consisting of a sequence of decisions}
}

\newglossaryentry{Markov decision process}
{
        name={Markov decision process},
        description={stochastic decision process that has the Markov (no-memory) property: the next state depends only on the current state and the action}
}

\newglossaryentry{finetuning}
{
        name={finetuning},
        description={training the pre-trained network on the new dataset}
}

\newglossaryentry{optimization}
{
        name={optimization},
        description={find an optimal element in a space; used in many aspects in machine learning}
}

\newglossaryentry{function approximation}
{
        name={function approximation},
        description={approximation of a mathematical function, a main goal of machine learning, often performed by deep learning }
}

\newglossaryentry{pretraining}
{
        name={pretraining},
        description={parameter transfer of the old task to the new task}
}

\newglossaryentry{few-shot learning}
{
        name={few-shot learning},
        description={task with which meta learning is often evaluated, to see how well the meta-learner can learn with only a few training examples}
}

\newglossaryentry{zero-shot learning}
{
        name={zero-shot learning},
        description={an example has to be  recognized  as belonging to 
a class without ever having been trained on  an example of this
class}
}

\newglossaryentry{entropy}
{
        name={entropy},
        description={measure of the amount of uncertainty in a distribution}
}

\newglossaryentry{parameters}
{
        name={parameters},
        description={the parameters $\theta$ (weights of a neural network) connect the neurons, together they determine the functional relation between input and output}
}

\newglossaryentry{hyperparameters}
{
        name={hyperparameters},
        description={determine the behavior of a learning algorithm; Base-learning learns parameters $\theta$, meta-learning learns hyperparameters $\omega$}
}

\newglossaryentry{exploration}
{
        name={exploration},
        description={selecting other actions than those that the policy $\pi(s)$ suggests}
}

\newglossaryentry{exploitation}
{
        name={exploitation},
        description={selecting actions as suggested by the current best policy $\pi(s)$}
}

\newglossaryentry{bootstrapping}
{
        name={bootstrapping},
        description={old estimates of a value are refined with new updates}
}

\newglossaryentry{overfitting}
{
        name={overfitting},
        description={high-capacity models can overtrain, where they model the signal and the noise, instead of just the signal}
}

\newglossaryentry{accuracy}
{
        name={accuracy},
        description={the total number of true positives and negatives  divided by the
total number of predictions}
}

\newglossaryentry{DQN}
{
        name={DQN},
        description={deep Q-network}
}

\newglossaryentry{DDQN}
{
        name={DDQN},
        description={double deep Q-network}
}

\newglossaryentry{DDDQN}
{
        name={DDDQN},
        description={dueling double deep Q-network}
}

\newglossaryentry{PEX}
{
        name={PEX},
        description={prioritized experience replay}
}

\newglossaryentry{REINFORCE}
{
        name={REINFORCE},
        description={REward Increment = Non-negative Factor $\times$ Offset Reinforcement $\times$ Characteristic Eligibility}
}

\newglossaryentry{A3C}
{
        name={A3C},
        description={asynchronous advantage actor critic}
}

\newglossaryentry{A2C}
{
        name={A2C},
        description={advantage actor critic}
}

\newglossaryentry{TRPO}
{
        name={TRPO},
        description={trust region policy optimization}
}

\newglossaryentry{PPO}
{
        name={PPO},
        description={proximal policy optimization}
}

\newglossaryentry{DDPG}
{
        name={DDPG},
        description={deep deterministic policy gradient}
}

\newglossaryentry{D4PG}
{
        name={D4PG},
        description={distributional distributed deep deterministic policy gradient}
}

\newglossaryentry{SAC}
{
        name={SAC},
        description={soft actor critic}
}

\newglossaryentry{value iteration}
{
        name={VI},
        description={value iteration}
}

\newglossaryentry{SARSA}
{
        name={SARSA},
        description={state action reward state action}
}

\newglossaryentry{TD}
{
        name={TD},
        description={temporal difference}
}

\newglossaryentry{ALE}
{
        name={ALE},
        description={atari learning environment}
}

\newglossaryentry{MuJoCo}
{
        name={MuJoCo},
        description={multi Joint dynamics with Contact}
}

\newglossaryentry{VIN}
{
        name={VIN},
        description={value iteration network}
}

\newglossaryentry{VPN}
{
        name={VPN},
        description={value prediction network}
}

\newglossaryentry{MADDPG}
{
        name={MADDPG},
        description={multi agent DDPG}
}

\newglossaryentry{CFR}
{
        name={CFR},
        description={counterfactual regret minimization}
}

\newglossaryentry{PETS}
{
        name={PETS},
        description={probabilisitic ensemble with trajectory sampling}
}

\newglossaryentry{PILCO}
{
        name={PILCO},
        description={probabilistic inference for learning control}
}

\newglossaryentry{LSTM}
{
        name={LSTM},
        description={long short-term memory}
}

\newglossaryentry{CPU}
{
        name={CPU},
        description={central processing unit}
}

\newglossaryentry{GPU}
{
        name={GPU},
        description={graphical processing unit}
}

\newglossaryentry{TPU}
{
        name={TPU},
        description={tensor processing unit}
}

\newglossaryentry{PBT}
{
        name={PBT},
        description={population based training}
}

\newglossaryentry{ACO}
{
        name={ACO},
        description={ant colony optimization}
}

\newglossaryentry{BERT}
{
        name={BERT},
        description={bidirectional encoder representations from 
transformers}
}

\newglossaryentry{GPT-3}
{
        name={GPT-3},
        description={generative pretrained transformer 3}
}

\newglossaryentry{MAML}
{
        name={MAML},
        description={model-agnostic meta-learning}
}

\newglossaryentry{ZSL}
{
        name={ZSL},
        description={zero-shot learning}
}

\newglossaryentry{XAI}
{
        name={XAI},
        description={explainable artificial intelligence} 
}

\newglossaryentry{GAN}
{
        name={GAN},
        description={generative adversarial network}
}

\chapter*{Preface}

Deep reinforcement learning  has gathered much
attention recently. Impressive results  were achieved in
activities as diverse as autonomous driving, game playing,
molecular recombination, and robotics.  In all these fields, computer programs
have learned to solve difficult problems. They have   
learned to fly model helicopters and perform aerobatic manoeuvers such as loops
and rolls. In some applications they have even become better than the
best  humans, such as in Atari, Go, poker and StarCraft. 

The way in which deep reinforcement learning explores complex
environments  reminds us how  children learn, by playfully trying
out things,  getting   feedback, and trying again.
The computer seems to truly 
possess  aspects of human learning;  deep reinforcement learning
touches the dream of artificial intelligence.

The successes in research have not gone unnoticed by
educators, and universities have  started to 
offer courses on the subject.
The aim of this book is to provide a comprehensive
overview of the field of deep reinforcement learning. The book is
written for graduate students of
artificial intelligence, and for researchers and
practitioners who wish to
better understand deep reinforcement learning methods and their challenges. We assume an
undergraduate-level  of
understanding of computer science and artificial intelligence; the
programming language of this book is Python.

We describe the 
foundations, the algorithms and the applications of deep reinforcement
learning. We cover the established
model-free and model-based methods that form the basis of the
field. Developments go quickly, and  we also cover more advanced topics: deep
multi-agent reinforcement learning, deep hierarchical reinforcement
learning, and deep meta learning.

 We hope that learning about deep reinforcement learning will give you
 as much joy as the many researchers   experienced when they
 developed their algorithms, finally got them to work, and saw them learn!

\section*{Acknowledgments}
This book benefited from the help of many
friends. 
First of all, I  thank everyone at the Leiden
Institute of Advanced Computer Science, for creating such a
fun and vibrant environment to work in. 

Many people contributed to this book. Some material is based on the 
book that we  
used  in our previous reinforcement learning course
and on  lecture
notes on policy-based methods written by
Thomas Moerland. 
Thomas also provided
invaluable critique on an earlier draft of the book. Furthermore,  as this book 
was being prepared, we worked on  survey articles on
deep model-based reinforcement learning,
deep meta-learning, 
and
deep multi-agent reinforcement learning. 
I thank  Mike Preuss, Walter Kosters, Mike Huisman, Jan van Rijn, Annie Wong,
Anna Kononova, and Thomas B\"ack, the co-authors on these articles.

Thanks to reader feedback the 2023 version of this book has been updated to include the Monte Carlo
sampling and the n-step methods, and to provide a better explanation of on-policy 
and off-policy learning.

I thank all members of the Leiden reinforcement learning community for
their input and  enthusiasm. I thank especially Thomas Moerland, Mike Preuss, Matthias
M\"uller-Brockhausen, Mike Huisman, Hui Wang, and Zhao Yang, for their help with
the course for which this book is written. I thank Wojtek Kowalczyk
for insightful discussions on 
deep supervised learning, and Walter Kosters for his views on
combinatorial search, as well as for his neverending sense of humor. 

A very special thank you goes to Thomas B\"ack, for our 
many discussions on science, the universe, and everything (including, 
especially, evolution). Without 
you, this effort would not have been possible. 

This book is a result of the graduate course on reinforcement learning
that we teach in Leiden. I  thank all students of this course,
past, present, and future, for their wonderful enthusiasm, sharp questions, and many
suggestions. This book was written for you and by you!

Finally, I thank Saskia, Isabel, Rosalin, Lily, and Dahlia, for being who they
are, for giving feedback and letting me learn,  and
for their boundless love.

\vspace{\baselineskip}
\begin{flushright}\noindent
Leiden,\hfill {\it ~ }\\
December 2021\hfill {\it Aske Plaat}\\
\end{flushright}

\renewcommand*{\lstlistlistingname}{List of Listings}

\setcounter{tocdepth}{3}

\pdfbookmark[1]{Contents}{toc}
\shorttoc{Contents}{1}

\cleardoublepage
\pdfbookmark[1]{Detailed Contents}{Detailed toc}

\tableofcontents

\lstset{language=Python}
\lstset{escapeinside={(*}{*)}}
\lstset{frame=lines,captionpos=b}
\lstset{basicstyle=\footnotesize\ttfamily,breaklines=true,numbers=left}






\mainmatter


\chapter{Introduction}\label{chap:intro}

Deep reinforcement learning studies  how we learn to solve complex
  problems, problems that require us
to find a solution to a sequence of decisions in high dimensional states.
To make bread, we must use the right
flour, add some salt, yeast and sugar, prepare the dough (not too dry and
not too wet), pre-heat the oven to the right temperature, and bake the
bread (but not too long); to win a ballroom dancing contest we must
find the right partner, 
learn to dance, practice, and beat the competition; to win in chess we
must study, practice, and make all the right moves.

\section{What is Deep Reinforcement Learning?}
Deep reinforcement learning is the combination of deep learning and
reinforcement learning.

The goal of deep reinforcement learning  is to learn 
optimal actions that maximize our reward
for all states that our environment can be in (the bakery, the dance
hall, the chess board). We do this by
interacting with complex, high-dimensional environments, trying out actions, and learning
from the feedback.

The field of \emph{deep} learning is about approximating functions in
high-dimensional problems; problems that are so complex that 
tabular methods cannot find exact solutions anymore.  Deep learning uses
deep neural networks to find \emph{approximations} for large, complex, 
high-dimensional environments, such as in image and speech
recognition.
The field has made impressive progress; computers can now recognize
pedestrians in a sequence of images (to avoid running over them), and can
understand sentences such as: ``What is the weather going to be like
tomorrow?''

The field of \emph{reinforcement} learning is about learning from
feedback; it learns by trial and error. Reinforcement learning does
not need a pre-existing dataset to train on; it chooses its own
actions, and learns from the \emph{feedback} that an environment provides. 
It stands to
reason that in this process of trial and error, our agent will make  mistakes (the
fire extinguisher is essential  to survive  the process of
learning to bake bread). The field of reinforcement
learning is all about learning from 
 success as well as from
mistakes.

In recent years the two fields of \emph{deep} and \emph{reinforcement} learning have
come together, and  have yielded new  algorithms, that are able to \emph{approximate}  high-dimensional problems by \emph{feedback} on their actions. 
Deep learning has brought new
methods and new successes, 
with advances in policy-based methods, in model-based
approaches, in transfer learning, in hierarchical reinforcement
learning,  and in multi-agent learning. 
\begin{table}[t]
  \begin{center}
    \begin{tabular}{lcc}
      & {\bf Low-Dimensional States} & {\bf High-Dimensional States} \\ \hline
      \hline
      {\bf Static Dataset} & classic supervised learning & deep supervised learning
      \\ 
      {\bf Agent/Environment Interaction} & tabular reinforcement learning & {\em deep
                                                           reinforcement
                                                           learning}\\
      \hline
    \end{tabular}
    \caption{The Constituents of Deep Reinforcement Learning}\label{tab:drl}
  \end{center}
\end{table}

The  fields also exist separately, as deep supervised learning and as
tabular reinforcement learning (see Table~\ref{tab:drl}). The aim of
deep supervised 
learning is to generalize and approximate complex, high-dimensional,
functions from pre-existing datasets, 
without interaction; Appendix~\ref{app:deep} discusses deep supervised
learning. The aim of tabular reinforcement learning is to
learn by interaction in simpler, low-dimensional,
environments such as Grid worlds; Chap.~\ref{ch:tab} discusses tabular reinforcement
learning.

Let us have a closer look at the two fields.

\subsection{Deep Learning}
Classic machine learning algorithms learn a predictive model
on data, using methods such as linear regression, decision trees,
random forests, support vector machines, and artificial neural networks. The models aim to
generalize, to make predictions. Mathematically speaking, machine
learning aims to approximate a  function from data. 

In the past, when computers were slow, the neural networks that were used
consisted of a few layers of fully connected neurons, and did not
perform exceptionally well on difficult problems.  This changed with the advent of deep
learning and faster computers. Deep neural
networks now consist of many layers of neurons and use
different types of connections.\footnote{Where \emph{many} means \emph{more than one} hidden layer
  in between the input and output layer.} Deep networks and \gls{deep learning}
have
taken the accuracy of certain important machine learning tasks to a
new level, and have allowed machine learning to be applied to complex,
high-dimensional, problems, such as recognizing cats and dogs in high-resolution
(mega-pixel) images.

Deep learning allows
high-dimensional problems to be solved in real-time;
it has allowed machine learning to be applied to day-to-day
tasks such as the face-recognition and speech-recognition that we use
in our smartphones.

\subsection{Reinforcement Learning}
Let us look more deeply at reinforcement learning, to see what it means to learn from our own
actions.

Reinforcement learning is a field  in which an agent learns
by interacting with an environment. In \gls{supervised learning}
  we need
pre-existing datasets of labeled 
examples to approximate a function;  \gls{reinforcement learning}
only needs
an environment that  provides   feedback
signals for actions that the agent is trying out. This requirement is 
easier to fulfill,  allowing reinforcement learning to be applicable to
more situations than supervised learning. 

Reinforcement   
learning agents generate, by their actions, their own on-the-fly
data, through the environment's rewards.  Agents can choose which
actions to learn from; reinforcement learning is a form of active learning. In this
sense, our agents are like children, that, 
through playing and exploring, 
teach themselves a certain task. This level of autonomy  is
one of the aspects that attracts researchers to the field. The reinforcement
learning agent chooses 
which action to perform---which hypothesis to test---and adjusts its
knowledge of what works, building up a 
policy of actions that are to be performed in the different states of
the world
that it has encountered.\index{hypothesis testing}\index{active
  learning}
(This freedom is also what makes  reinforcement learning hard,
because when you are  allowed to choose your own examples, it is all
too  easy to stay in your comfort zone, stuck in a  positive reinforcement
bubble, believing you are doing great, but  learning very little of the
 world around you.)

\subsection{Deep Reinforcement Learning}
{\emph{Deep} reinforcement learning combines  methods for learning
  high-dimensional problems with reinforcement learning, allowing
  \emph{high-dimensional, interactive} learning.
A major reason for the interest in \gls{deep reinforcement learning}
is that it  works well on current computers, 
and does  so  in seemingly different 
applications. For example, in Chap.~\ref{ch:play} we will see how deep reinforcement learning can learn eye-hand
coordination tasks to play 1980s video games, in Chap.~\ref{ch:pol}
we see how a simulated robot cheetah learns to jump, and in Chap.~\ref{chap:given}
we see how it can teach itself to play complex  games of
strategy to the extent that world champions are beaten.

Let us  have a closer
look at the kinds of applications on which deep reinforcement learning
does so well.

\subsection{Applications}\index{trial and error}
In its most basic form, reinforcement learning is a way to teach an
agent to operate in the world. As a child learns to walk from actions
and feedback, so do reinforcement learning agents learn from actions and
feedback. 
Deep reinforcement learning can learn to
solve large and complex decision problems---problems whose solution is
not yet known, but for  which an approximating trial-and-error mechanism exists that can
learn a solution out of  repeated
interactions with the problem. This may sound a bit cryptical and convoluted,
but approximation and trial and error are something that we do in
real life all the time. Generalization and approximation allow us to infer 
patterns or rules from examples.
Trial and error is a
method by which humans  learn how to deal with things that are
unfamiliar to them (``What happens if I press this button?
Oh. Oops.'' Or: ``What happens if I do not put my  leg before my
other leg while moving forward? Oh. Ouch.'').

\subsubsection*{Sequential Decision  Problems}
Learning to operate  in the world  is a high level goal; we can be
more specific. Reinforcement learning is about the agent's behavior.
Reinforcement learning can find solutions for \emph{sequential decision 
 problems}, or optimal control problems, as they are known in engineering.
There are many situations in the real world where, in order to reach a goal, a
sequence of decisions must be made. 
Whether it is baking a cake, building a house, or playing a card game; a
sequence of decisions has to be made.
Reinforcement
learning provides efficient ways to learn solutions to sequential decision
 problems.
\index{sequential decision problem}

Many real world problems can be modeled as a sequence of decisions~\cite{neptune2021}.
For example, in autonomous driving, an agent is faced with questions of speed
control, finding drivable areas, and, most importantly, avoiding
collisions. 
In healthcare,  treatment plans contain many sequential decisions, and
factoring the effects of delayed treatment can be studied. In  customer centers, natural language processing can
help improve chatbot dialogue,  question answering, and even 
machine translation.
In marketing and communication, recommender systems recommend news,
personalize suggestions, deliver notifications to user, or otherwise
optimize the product experience. 
In trading and finance, systems decide to  hold, buy or sell financial
titles,
in order to
optimize future reward.
In politics and governance, the effects of policies can be simulated
as a sequence of decisions
before they are implemented.
In mathematics and entertainment, playing board games, card games, and strategy
games consists of a sequence of decisions. In computational
creativity, making a painting requires a sequence of esthetic decisions.
In industrial robotics and engineering,  the grasping of items and the manipulation of materials consists of a sequence of decisions.
In chemical manufacturing, the optimization of  production
processes consists of many decision steps, that influence the yield
and quality of the product. 
Finally, in energy grids, the efficient and safe distribution of energy can be
modeled as  a \gls{sequential decision problem}.

In all these situations, we must make a sequence of decisions. In all
these situations, taking the wrong decision can be very
costly.

The algorithmic research on sequential decision making has focused on two types of
applications: (1) robotic problems and (2) games. Let us have a closer look at these two 
domains, starting with robotics.

\subsubsection*{Robotics}

In principle, all actions that a robot should take can be
pre-programmed step-by-step by a  programmer in meticulous detail. In highly
controlled environments, such as  a welding robot in a car factory,
this can conceivably work, although any small change
or any new task 
requires reprogramming the robot.

It is surprisingly  hard to  manually
program a robot to perform a complex 
task. Humans are not aware of their own operational knowledge, such as
what ``voltages''  we put on which muscles when we pick up a cup.  
It is much easier to define a desired goal state, and let the system
find the complicated solution by itself. Furthermore, in environments that are only
slightly  challenging, when the robot must
be able to respond more flexibly to different conditions, an adaptive
program is needed.

It will be no surprise that the application
area of robotics is an important driver for machine
learning research, and 
robotics researchers
turned early on to finding methods by which the robots could teach
themselves certain behavior.

\begin{figure}[t]
\begin{center}
\includegraphics[width=7cm]{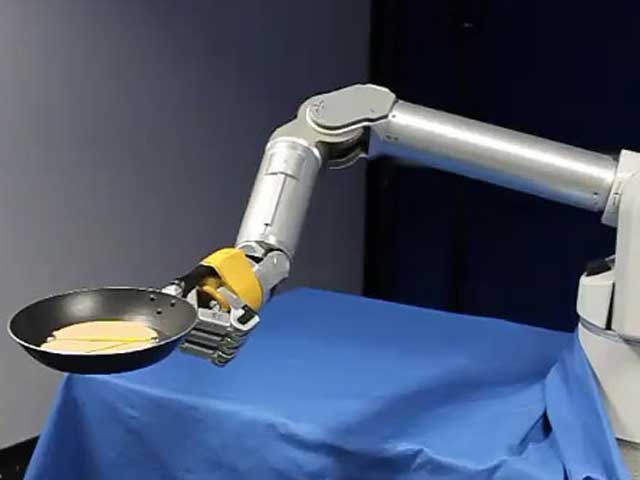}
\caption{Robot Flipping Pancakes~\cite{kormushev2010robot}}\label{fig:robot}
\end{center}
\end{figure}

\begin{figure}[t]
\begin{center}
\includegraphics[width=9cm]{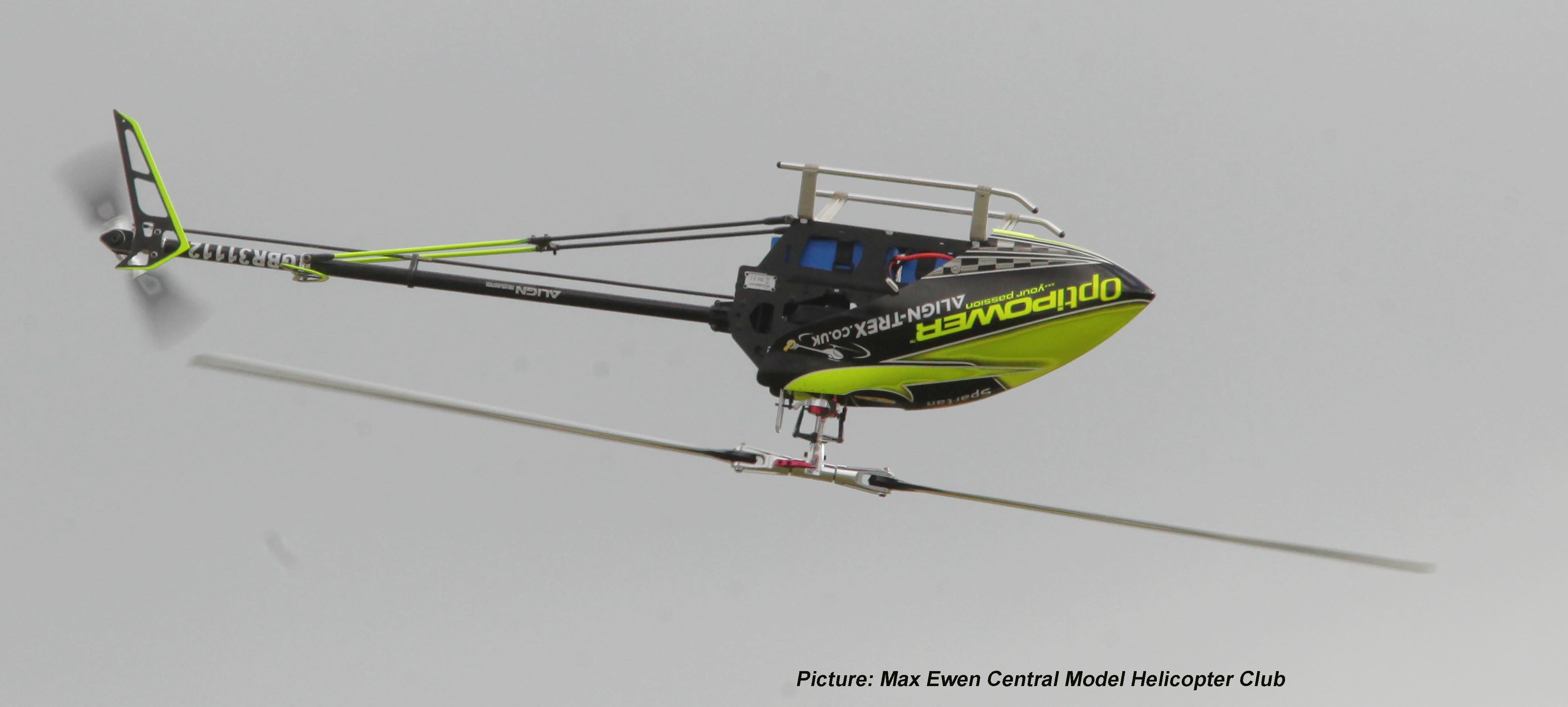}
\caption{Aerobatic Model Helicopter~\cite{abbeel2007application}}\label{fig:helicopter}
\end{center}
\end{figure}

The literature on robotics experiments  is varied and rich.
A robot  can teach itself how to navigate a maze, how to perform manipulation
tasks, and how to learn locomotion tasks.

Research into adaptive
robotics has made quite some progress. For example, one of the  recent
achievements involves flipping 
pancakes~\cite{kormushev2010robot} and flying an aerobatic 
model helicopter~\cite{abbeel2010autonomous,abbeel2007application}; see
Figs.~\ref{fig:robot} and~\ref{fig:helicopter}.
Frequently, learning tasks are combined with computer
vision, where a robot has to learn by visually interpreting the consequences of
its own actions.

\begin{figure}[t]
\begin{center}
\includegraphics[width=7cm]{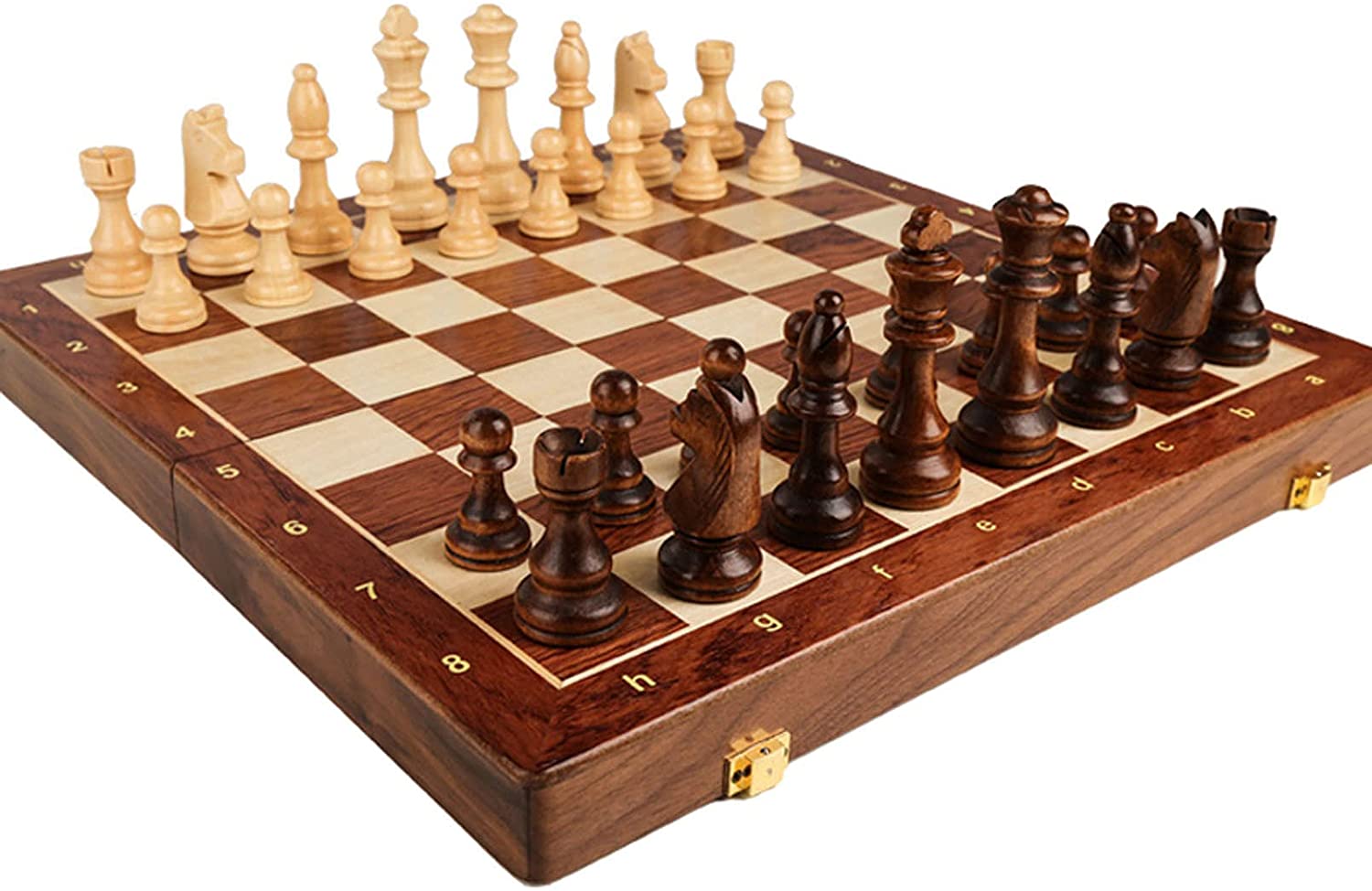}
\caption{Chess}\label{fig:chess}
\end{center}
\end{figure}

\begin{figure}[t]
\begin{center}
\includegraphics[width=7cm]{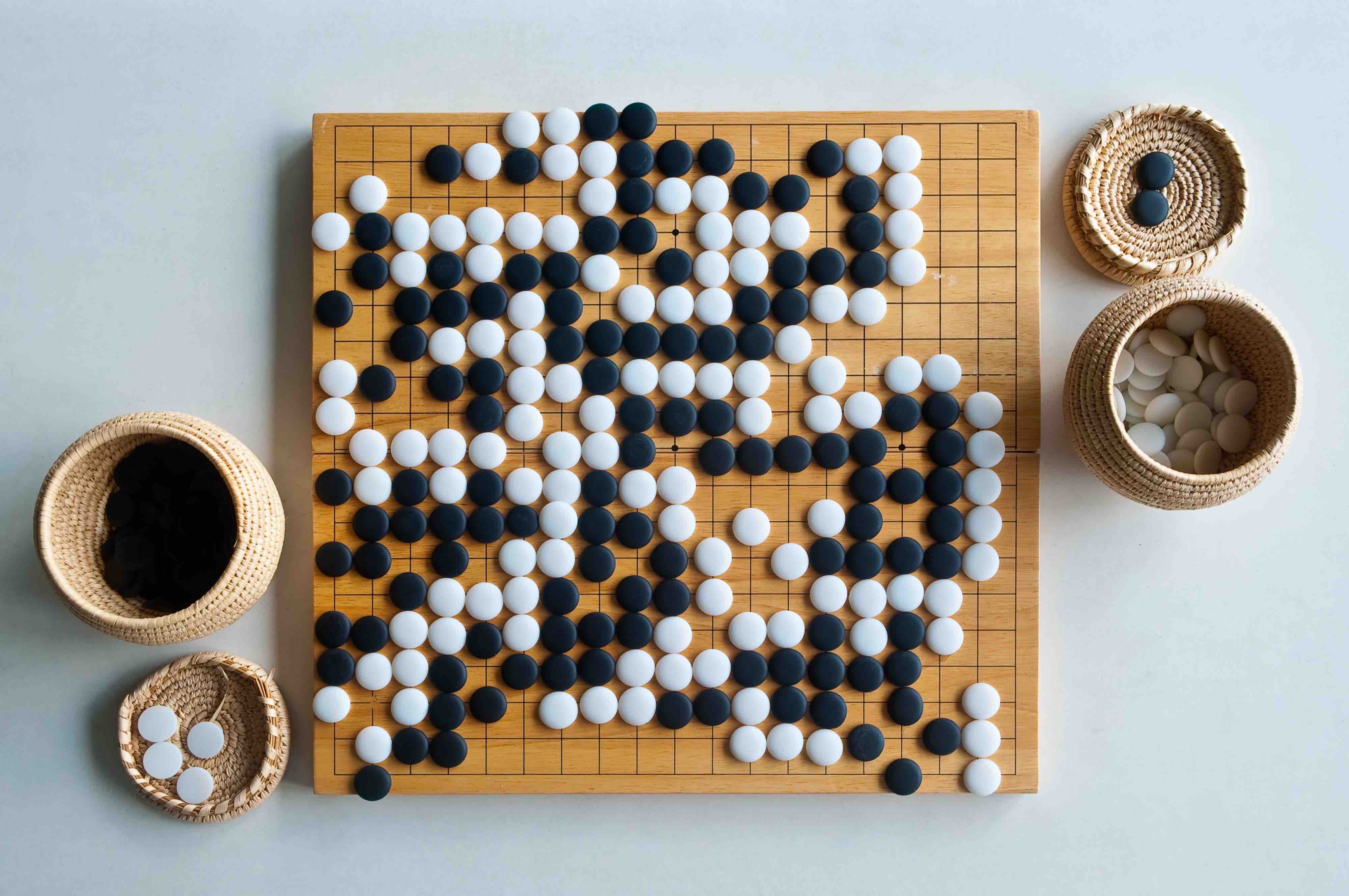}
\caption{Go}\label{fig:go}
\end{center}
\end{figure}

\begin{figure}[t]
\begin{center}
\includegraphics[width=7cm]{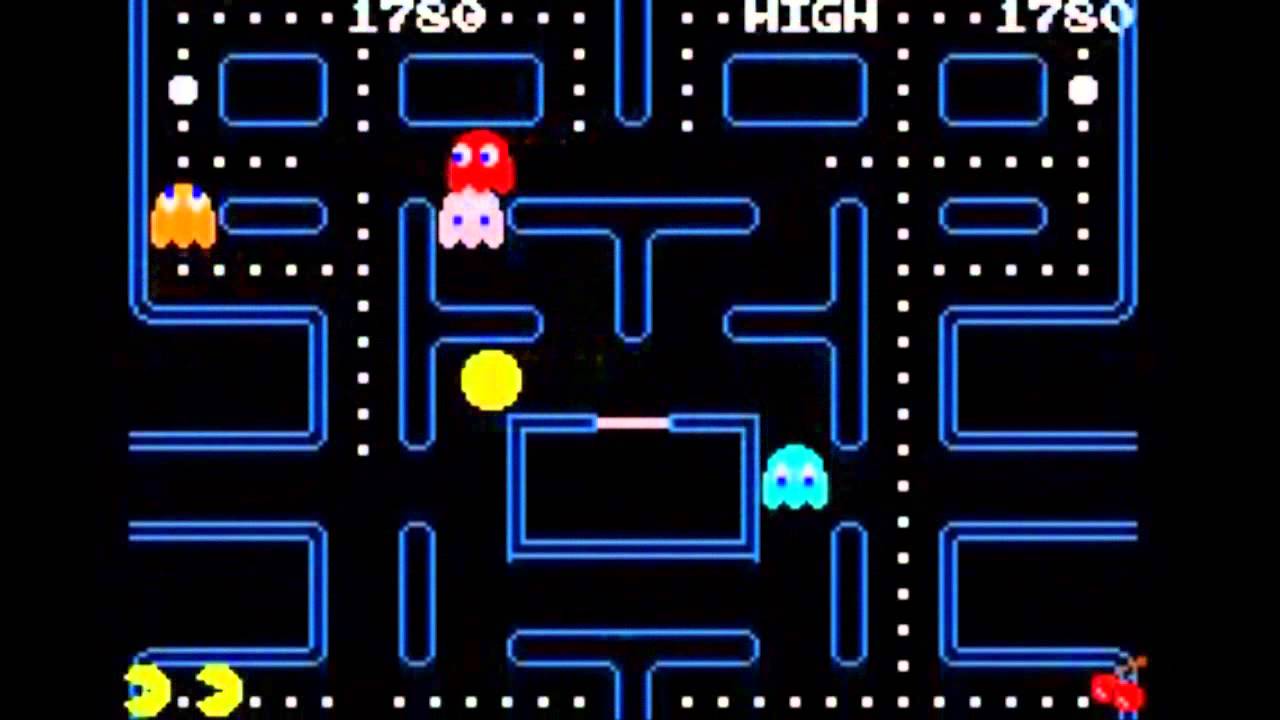}
\caption{Pac-Man \cite{bellemare2013arcade}}\label{fig:pacman}
\end{center}
\end{figure}

\begin{figure}[t]
\begin{center}
  \includegraphics[width=7cm]{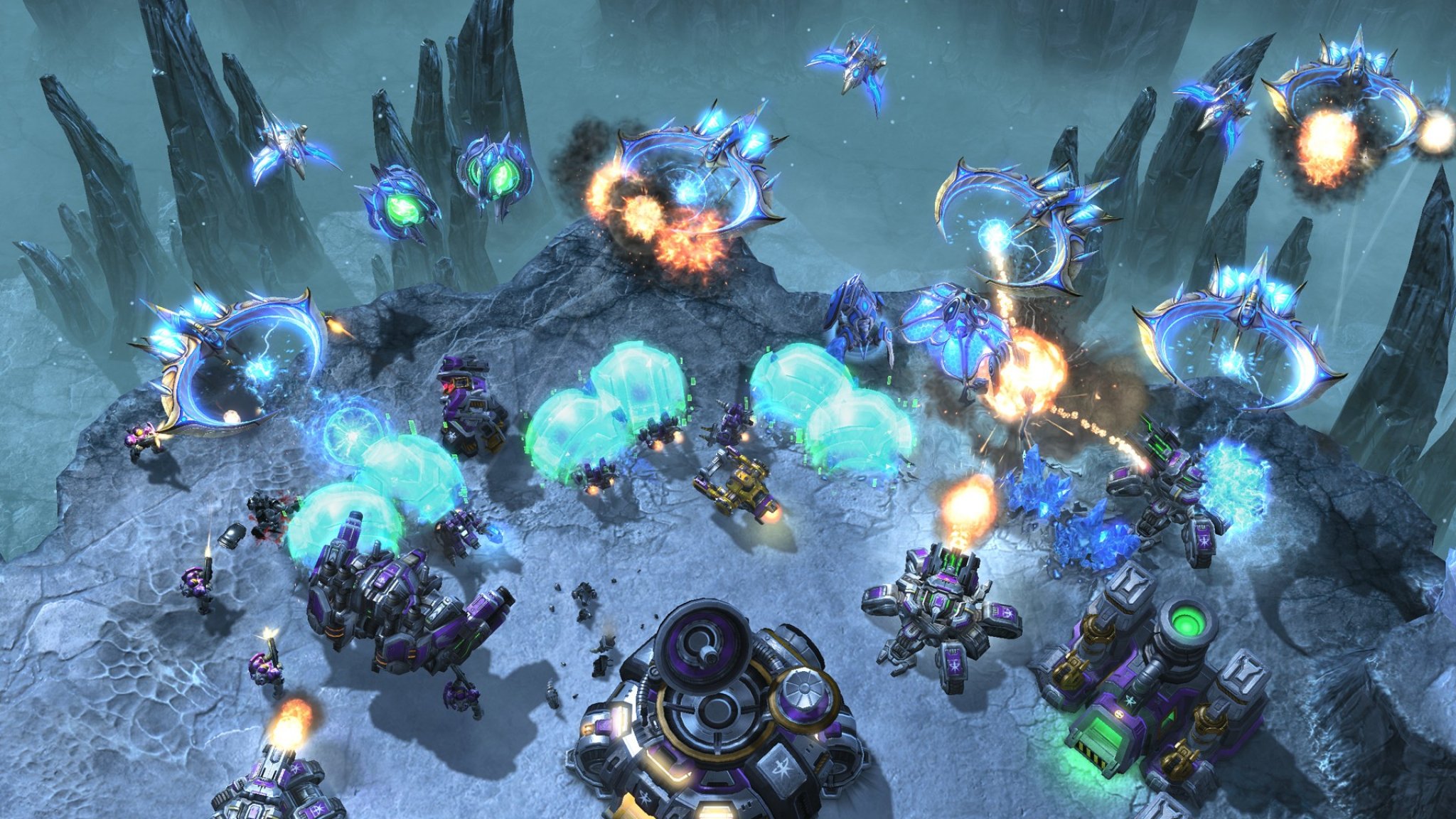}
\caption{StarCraft \cite{vinyals2019grandmaster}}\label{fig:starcraft}
\end{center}
\end{figure}

\subsubsection*{Games}
Let us now turn to games. Puzzles and games have been used from the earliest days  to study aspects of 
intelligent behavior. Indeed, before computers were powerful enough
to execute chess programs, in the days of Shannon and Turing, paper designs were made, in the hope that
understanding chess would teach us something about the nature of
intelligence~\cite{shannon1988programming,turing1953digital}. 

Games allow researchers to limit the scope of their studies, to focus
on intelligent decision making in a limited 
environment, without  having to master the full complexity of the real
world.  In addition to board games such as chess and Go, video
games are  being used extensively to test  intelligent methods in computers.
Examples are Arcade-style games such as Pac-Man~\cite{mnih2015human}
and multi-player strategy games such as StarCraft~\cite{vinyals2019grandmaster}.
See Figs.~\ref{fig:chess}--\ref{fig:starcraft}.
\index{chess}\index{Go}\index{StarCraft}

\subsection{Four Related Fields}
Reinforcement learning is a rich  field, that has
existed in some form long before the artificial intelligence endeavour had
started, as a part of biology, psychology, and
education~\cite{bertsekas1996neuro,kaelbling1996reinforcement,sutton2018introduction}. In    
artificial intelligence it has become one of the three main categories
of machine learning, the other two being supervised and unsupervised
learning~\cite{bishop2006pattern}. This book is a book of algorithms that
are inspired by topics from the natural and social sciences.  Although the rest of
the book will be about these algorithms, it is interesting to briefly discuss
the  links of deep reinforcement learning to human and 
animal learning.
We will  introduce the four scientific disciplines that have a
profound influence on deep reinforcement learning.

\subsubsection{Psychology}\index{Skinner}\index{Pavlov}
In psychology, reinforcement learning is also known  as \emph{learning by
conditioning} or as \emph{operant
conditioning}. Figure~\ref{fig:dog}
illustrates the folk psychological idea of how a dog can be conditioned. A natural reaction to food is that a dog salivates. By ringing
a bell whenever the dog is given food, the dog learns to associate the
sound with food, and after enough trials, the dog starts salivating as soon as
it hears the bell, presumably in 
anticipation of the food, whether it is there  or not.

\begin{figure}[t]
\begin{center}
\includegraphics[width=\textwidth]{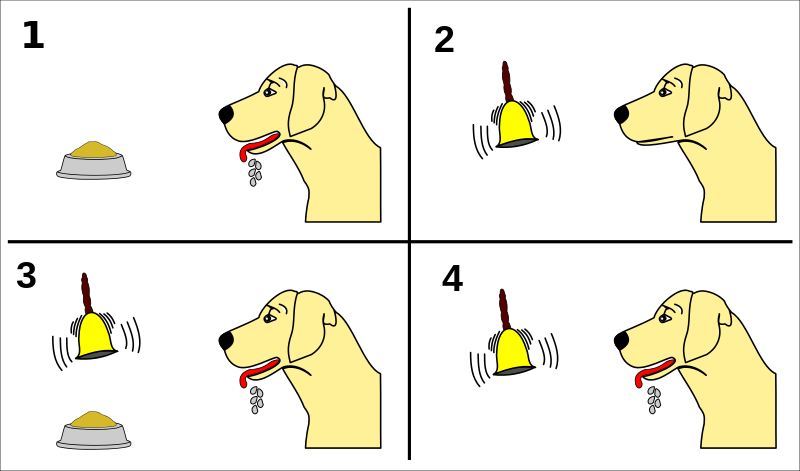}
\caption{Classical Conditioning: (1) a dog salivates when seeing food,
  (2) but initially not when hearing a bell, (3) when the sound rings often
  enough together when food is served, the dog starts to associate the bell with
  food, and (4) also salivates when only the bell rings}\label{fig:dog}
\end{center}
\end{figure}

The behavioral scientists Pavlov (1849--1936) and  Skinner (1904--1990)  are 
well-known for their work on conditioning.  Phrases such as 
\emph{Pavlov-reaction} have entered our everyday language, and various
jokes about  conditioning exist (see, for example,
Fig.~\ref{fig:pavlov}). 
Psychological research into learning is one of the main influences on
reinforcement learning as we know it in artificial intelligence.

\begin{figure}[t]
\begin{center}
\includegraphics[width=5cm]{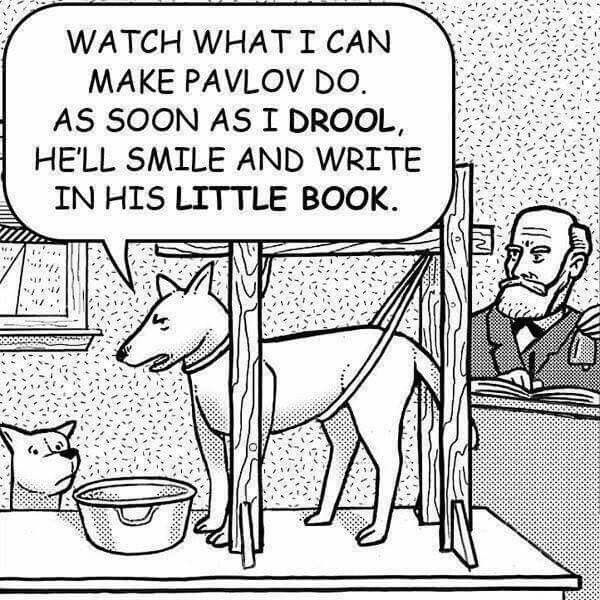}
\caption{Who is Conditioning Whom?}\label{fig:pavlov}
\end{center}
\end{figure}

\subsubsection{Mathematics}
Mathematical logic is another foundation of deep reinforcement
learning. Discrete \gls{optimization} and graph theory are of great
importance for the 
formalization of reinforcement learning, as we will see in
Sect.~\ref{sec:mdp} on Markov decision processes. 
Mathematical formalizations have enabled 
the development of efficient planning and optimization algorithms, that
are at the core of current progress.

Planning and optimization are an important part of deep reinforcement
learning. They are also related to the field of operations
research, although there the emphasis is on  (non-sequential)
combinatorial optimization problems. In AI, planning and 
optimization are used as building blocks for creating learning systems
for sequential, high-dimensional,  problems that can include visual, textual or
auditory input.

The field of symbolic reasoning is based on logic, it is one
of the earliest success stories in artificial intelligence. Out of
work in symbolic reasoning came heuristic
search~\cite{pearl1984heuristics}, expert 
systems,  and theorem proving systems. Well-known systems 
are the  STRIPS planner~\cite{fikes1971strips}, the Mathematica computer algebra system~\cite{buchberger1982computer}, 
the logic programming language PROLOG~\cite{clocksin2012programming}, and
also systems such as SPARQL for semantic (web)
reasoning~\cite{antoniou2004semantic,berners2001semantic}.

Symbolic AI focuses on reasoning in discrete domains, such as decision trees,
planning, and
games of strategy, such as chess and checkers. Symbolic AI 
has driven success in methods to search the web, to power online social
networks, and to power online commerce. These
highly successful technologies are the basis of much of our modern
society and  economy. In 2011 the highest recognition in computer
science, the Turing award, was  awarded to  Judea Pearl for work in
causal reasoning (Fig.~\ref{fig:pearl}).\footnote{Joining a long list of
   AI researchers that have been honored earlier with a Turing award:  Minsky,
  McCarthy, Newell, Simon,  Feigenbaum and Reddy.} Pearl later
published an influential  book to popularize the field~\cite{pearl2018book}. 

\begin{figure}[t]
\begin{center}
\includegraphics[width=5cm]{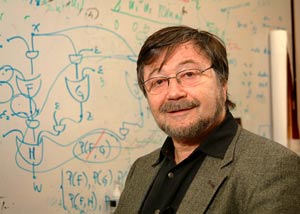}
\caption{Turing-award winner Judea Pearl}\label{fig:pearl}
\end{center}
\end{figure}

Another area of mathematics that  has played a large role in deep reinforcement
learning is the field of continuous (numerical) optimization. Continuous methods
are important, for example, in  efficient gradient
descent and backpropagation methods that are at the heart of current
deep learning algorithms.  

\subsubsection{Engineering}
In engineering, the field of reinforcement learning is better known as
\emph{optimal control}.
The theory of optimal control of  dynamical systems was developed by Richard
Bellman and Lev Pontryagin~\cite{bertsekas1995dynamic}. Optimal control theory originally focused
on  dynamical systems, and the technology and methods relate to
continuous optimization methods such as used in robotics (see
Fig.~\ref{fig:optctrl} for an illustration of optimal control at work in docking two
space vehicles). Optimal control theory is of central importance to
many problems in engineering.

\begin{figure}[t]
\begin{center}
\includegraphics[width=8cm]{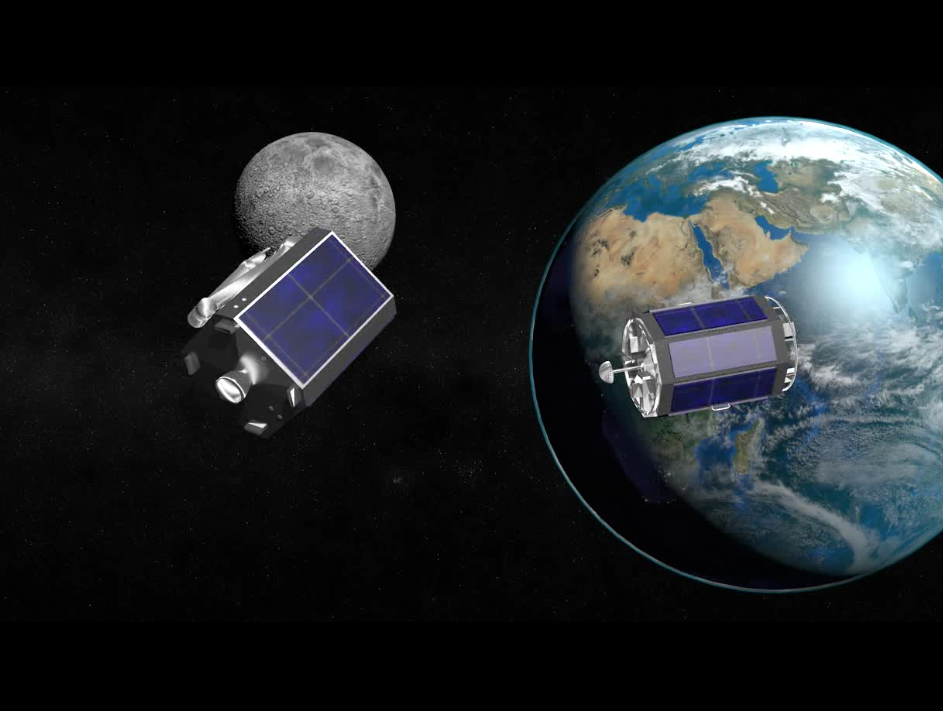}
\caption{Optimal Control of Dynamical Systems at Work}\label{fig:optctrl}
\end{center}
\end{figure}

To this day reinforcement learning and optimal control  use a
different terminology and notation. States and 
actions are denoted as $s$ and $a$ in state-oriented
reinforcement learning, where the engineering world of optimal control
uses $x$ and $u$. In this book the former notation is used.

\subsubsection{Biology}
Biology has a profound influence on computer science. Many
nature-inspired optimization algorithms have been developed in
artificial intelligence. An important nature-inspired school of thought is  connectionist
AI.

Mathematical logic and engineering approach
intelligence as  a top-down deductive  process;
observable effects in the real world follow from the application of 
theories and the laws of nature, and   intelligence follows deductively from theory. In contrast, 
connectionism approaches intelligence in a bottom-up
fashion. Connectionist intelligence emerges  out of many low level 
interactions.  Intelligence follows inductively from practice. Intelligence is
embodied: the bees in bee hives, the ants in ant colonies, and
the neurons in the brain all interact, and out of the connections and
interactions arises  behavior that  we recognize as
intelligent~\cite{bonabeau1999swarm}. 

Examples of the connectionist approach to  intelligence are nature-inspired algorithms 
such as Ant colony optimization~\cite{dorigo1997ant}, swarm
intelligence~\cite{kennedy2006swarm,bonabeau1999swarm},   evolutionary
algorithms~\cite{back1993overview,fogel1994introduction,holland1992genetic},
robotic intelligence~\cite{brooks1991intelligence}, and, last but not
least,  artificial neural
networks and deep learning~\cite{haykin1994neural,lecun2015deep,goodfellow2016deep}.

\begin{figure}[t]
\begin{center}
\includegraphics[width=8cm]{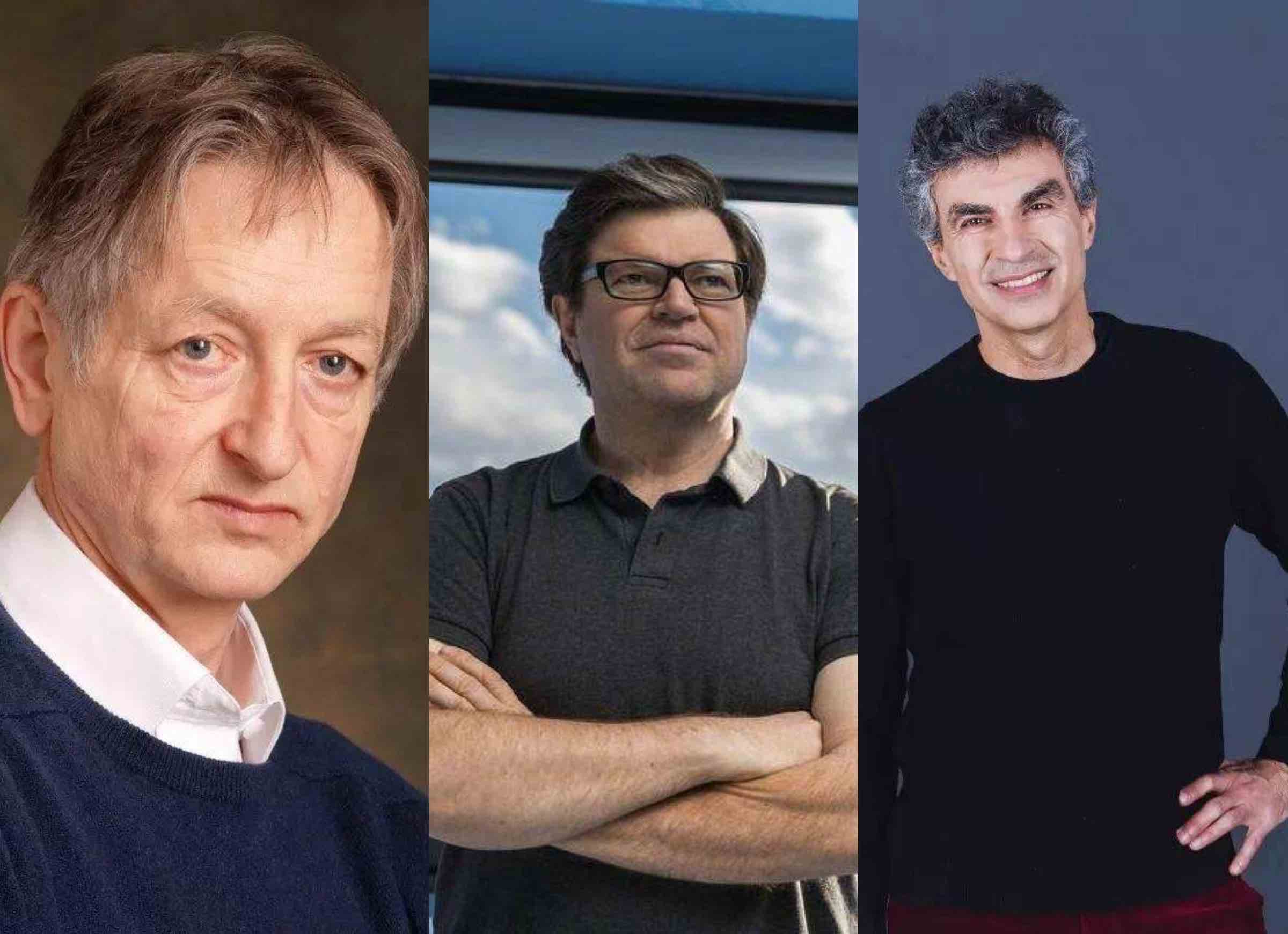}
\caption[Hinton, LeCun, Bengio]{Turing-award winners Geoffrey Hinton,  Yann LeCun, and Yoshua Bengio}\label{fig:bhl}
\end{center}
\end{figure}

It should be noted that both the symbolic and the connectionist school
of AI have been very successful.
After the enormous economic impact of search and symbolic
AI (Google, Facebook, Amazon, Netflix), much of the interest in AI in the last
two decades has been inspired by the 
success of the connectionist approach in computer language and vision.
In 2018  the Turing award
was awarded to three key researchers in deep learning: Bengio, Hinton,
and LeCun (Fig.~\ref{fig:bhl}).\index{Bengio, Yoshua}\index{LeCun, Yann}\index{Hinton, Geoffrey} Their most famous paper on deep
learning may well be~\cite{lecun2015deep}.

\section{Three Machine Learning Paradigms}

Now that we have introduced the general context and origins of deep
reinforcement learning, let us switch gears, and talk about
\gls{machine learning}. Let us see how deep reinforcement learning fits in the general
picture of the field. At the same time, we will take the opportunity
to introduce some notation and basic concepts.

In the next section we will then provide an outline of the book. But
first it is time for machine learning.
We start at the beginning,  with \gls{function approximation}.

\subsubsection*{Representing a Function}
Functions are a central part in artificial intelligence. A function $f$
transforms input $x$
to output $y$ according to some method, and we write   $f(x)\rightarrow y$.
In order to perform calculations with  function $f$, the function  must be
represented as a computer program in some form in  memory.
We also write  function 
$$f: X \to Y, $$
where the domain $X$ and range $Y$ can be discrete or continuous; the
dimensionality (number of attributes in $X$) can be arbitrary.

Often, in the real world, the same input may yield a range of different
outputs, and  we
would like our function to provide a conditional probability
distribution, a function that maps 
$$f: X \to p(Y). $$
Here the function  maps the domain to a probability distribution $p$ over the
range. Representing a conditional probability allows us to model
functions for which the input does not always give the same
output. (Appendix~\ref{app:math} provides more mathematical background.)

\subsubsection*{Given versus Learned Function} 

Sometimes the function that we
are interested in is given, and we can  represent
the function  by a specific algorithm that computes  an  analytical
expression that is known exactly. This is, for example,  the case for  laws of physics, or when
we make explicit assumptions for a particular system.  

\begin{tcolorbox}
{\bf Example}: Newton's second Law of Motion states that for objects
with constant mass 
$$ F = m \cdot a, $$
where $F$ denotes the net force on the object, $m$ denotes its mass,
and $a$ denotes its acceleration.  In this case, the analytical
expression defines the entire function, for every possible combination
of the inputs. 
\end{tcolorbox}
However, for many functions in the real
world, we do not have an analytical expression. Here, we enter the
realm of machine learning, in particular of \emph{supervised
  learning}. When we do not know an analytical expression for a
function, our best approach is to collect data---examples of $(x,y)$
pairs---and reverse engineer or \emph{learn} the function from this data. See Fig.~\ref{fig_linear_regression}.

\begin{figure}[t]
  \centering
      \includegraphics[width = 0.6\textwidth]{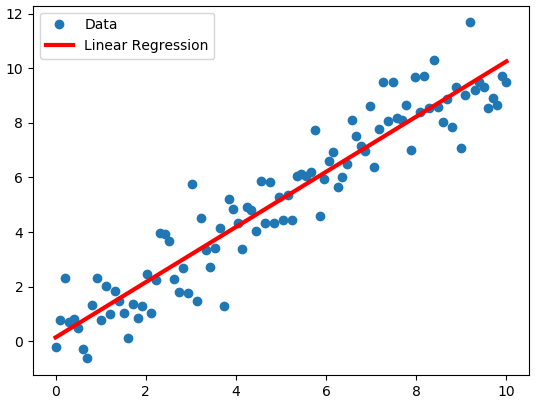}
  \caption[Linear Regression]{Example of learning a function; data
    points are in blue, a possible learned linear
    function is the red line, which allows us to make
    predictions $\hat{y}$ for any new input $x$} 
    \label{fig_linear_regression}
\end{figure}

\begin{tcolorbox}
{\bf Example}: A company wants to predict the chance that you buy a
shampoo to color your hair, based on your age. They collect many data points of $x \in
\mathbb{N}$, your age (a natural number), that map to $y \in
\{0,1\}$, a binary indicator whether you bought their shampoo. They
then want to {\it learn} the mapping 
$$ \hat{y} = f(x)$$
where $f$ is the desired function that tells the company who will buy
the product and $\hat{y}$ is the predicted $y$ (admittedly overly simplistic in this example).
\end{tcolorbox}
Let us see which methods exist in machine learning to find function
approximations. 

\subsubsection*{Three Paradigms}
There are three main paradigms for how the observations can be
provided in machine learning: 
(1)  supervised learning, (2) reinforcement learning, and (3)
unsupervised learning.

\subsection{Supervised Learning}\index{supervised learning}
The first and most basic paradigm for machine learning is supervised learning. In
supervised learning, the  data to learn the function $f(x)$   is
provided to the learning algorithm in $(x,y)$ example-pairs. Here $x$ is the input, and $y$
the observed output value to be learned for that particular input value $x$. The $y$
values can be thought of as supervising the learning process, they
teach the learning process the right answers for each input value
$x$, hence the name \emph{supervised} learning.

The data pairs to be learned from  are organized
in a dataset, which must be present in its entirety before the
algorithm can start. 
During the learning process, an estimate of the real function that
generated the data is created,
$\hat{f}$. The $x$ values of the pair are also called the \emph{input}, and
the $y$ values are the \emph{label} to be learned.

Two well-known problems in
supervised learning are  regression  and  classification. Regression
predicts a continuous number, classification a 
dicrete category.
The best known regression relation is the linear relation: the
familiar straight line through a cloud of observation points that we
all know from our introductory statistics course.
Figure~\ref{fig_linear_regression} shows such a linear relationship $\hat{y}=a\cdot x +
b$. The linear function can be characterized with two parameters $a$
and $b$. Of course, more complex functions are possible, such as
quadratic regression, non-linear regression, or regression with higher-order
polynomials~\cite{draper1998applied}.

The supervisory signal is computed for each data item $i$ as the
difference between the current estimate and the given label, for
example by  $(\hat{f}(x_i) - y_i)^2$. 
Such an error function $(\hat{f}(x) - y)^2$ is also known as a loss
function; it  measures the quality of our prediction. 
The closer our prediction is to the true label, the lower the loss.
There are many ways to compute this closeness, such as the 
mean squared error loss  $\mathcal{L}=\frac{1}{N}\sum_1^N(\hat{f}(x_i)-y_i)^2$, which
is used often for regression over $N$ observations. This loss function can be used by a
supervised learning algorithm to adjust model parameters $a$ and $b$
to fit the function $\hat{f}$ to the data. Some of the many possible learning
algorithms are  linear regression and support vector machines~\cite{bishop2006pattern,russell2016artificial}.

In classification, a relation between an input value and a class label
is learned. A well-studied classification problem
is image recognition, where  two-dimensional images
are to be categorized. Table~\ref{tab:list-class} shows a tiny dataset
of labeled images of the proverbial cats and dogs. A popular loss
function for classification is the
cross-entropy loss $\mathcal{L}=-\sum_1^Ny_i\log(\hat{f}(x_i))$, see also Sect.~\ref{sec:cem}.
Again, such a loss function can be used to adjust the model parameters
to fit the function to the data. The model can be small and linear,
with few parameters, or it can be
large, with many parameters, such as a neural network, which is often used for image classification.

\begin{table}[t]
  \begin{center}
    \begin{tabular}{c|c|c|c|c|c}

      \includegraphics[height=1cm]{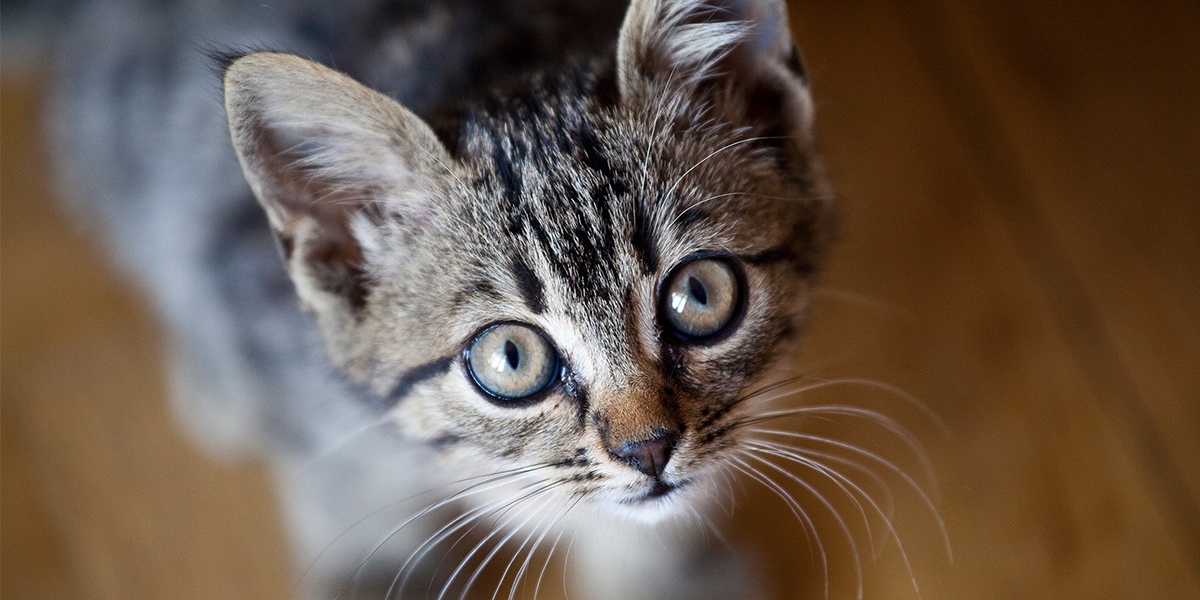}
                 &\includegraphics[height=1cm]{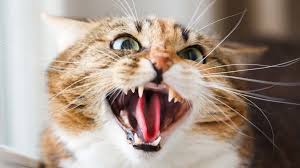}
                          &\includegraphics[height=1cm]{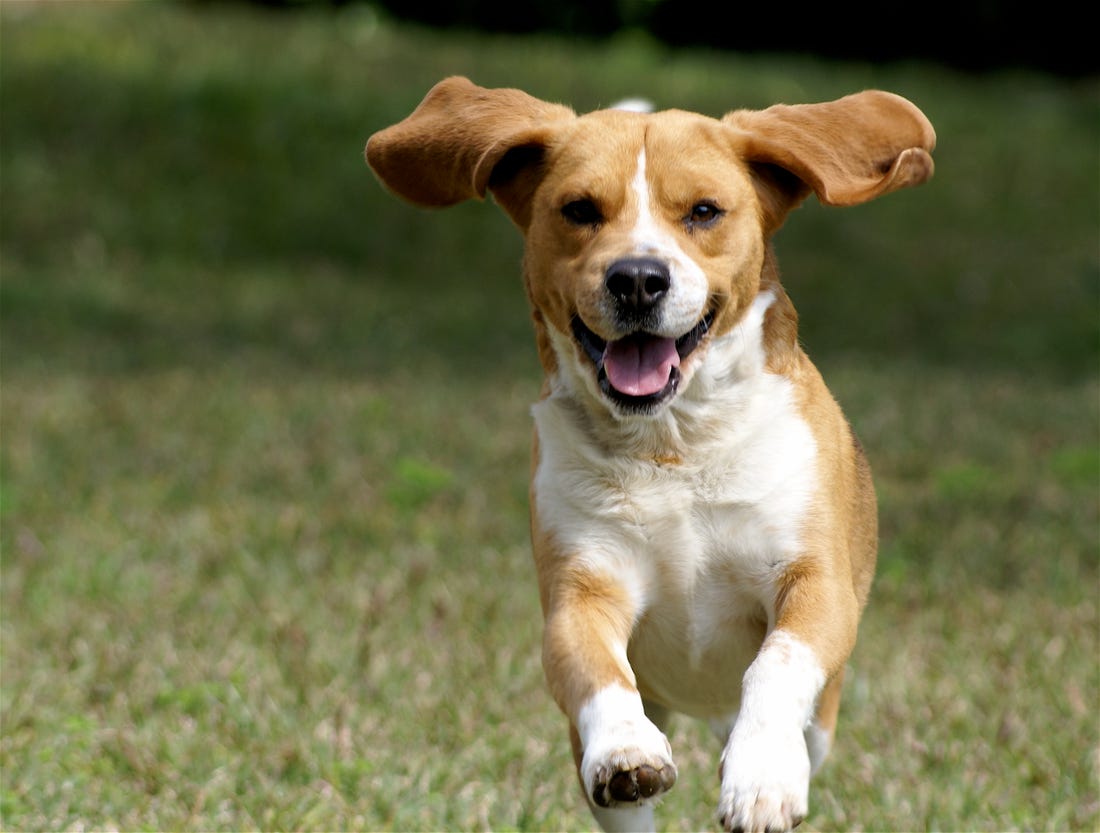}
                                   &\includegraphics[height=1cm]{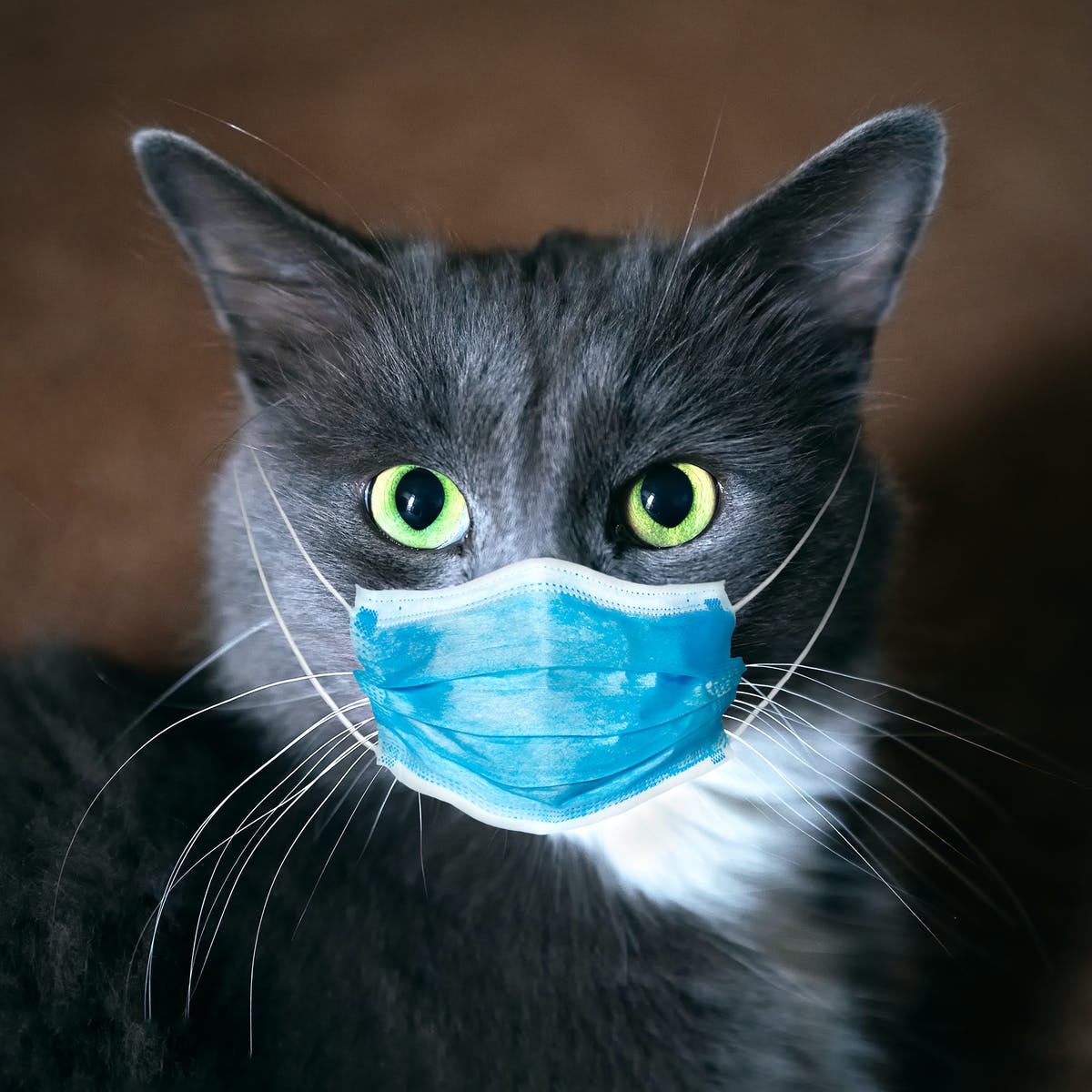}
                                            &\includegraphics[height=1cm]{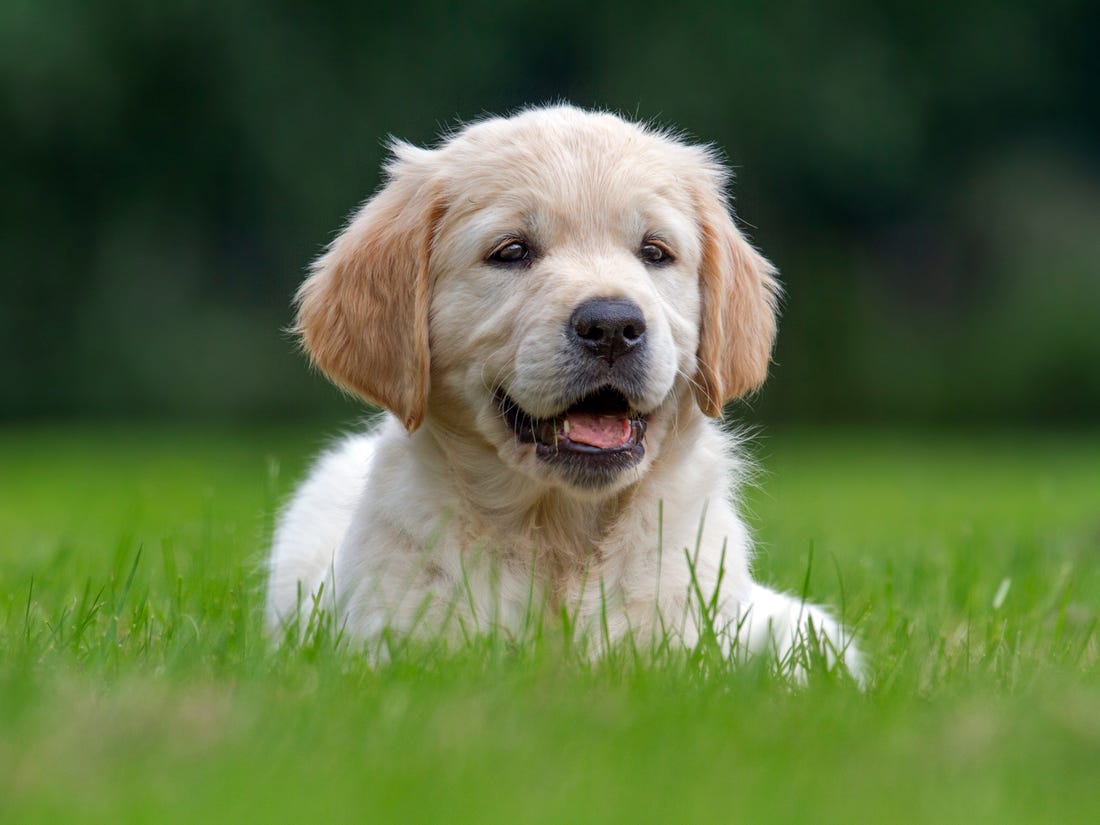}
                                              &\includegraphics[height=1cm]{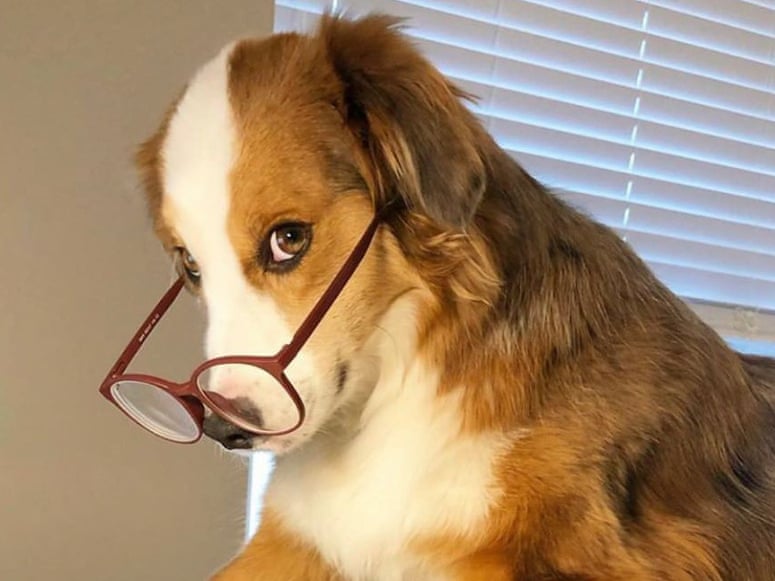}\\ \hline
      ``Cat''& ``Cat''& ``Dog''& ``Cat''& ``Dog''& ``Dog''
    \end{tabular}
    \caption{(Input/output)-Pairs for a Supervised Classification Problem}\label{tab:list-class}
  \end{center}
\end{table}

In supervised learning a large dataset exists where all input items
have an associated training label.
Reinforcement learning is different, it does not assume the pre-existence
of a  large labeled training set. Unsupervised learning does require a
large dataset, but no user-supplied output labels; all it needs are
the inputs.

Deep learning function approximation was first developed in a
supervised setting.
Although this book is about deep reinforcement learning, we will encounter
supervised learning concepts frequently, whenever we discuss the deep
learning aspect of deep reinforcement learning. 

\subsection{Unsupervised Learning}\index{unsupervised learning}\label{sec:unsup}
When there are no labels  in the dataset, different learning
algorithms must be used. Learning without labels is called
\emph{unsupervised learning}. In \gls{unsupervised learning} an inherent metric of
the data items is used, such as distance. A typical problem in unsupervised
learning is to find
patterns in the data, such as clusters or
subgroups~\cite{vreeken2011krimp,van2012diverse}.

Popular unsupervised learning algorithms are 
$k$-means algorithms, and principal component
analysis~\cite{scholkopf1997kernel,jolliffe2016principal}. Other
popular unsupervised methods  are
dimensionality reduction techniques from visualization, 
such as t-SNE~\cite{maaten2008visualizing}, minimum description
length~\cite{grunwald2007minimum} and data
compression~\cite{barron1998minimum}. A popular application of
unsupervised learning are autoencoders, see
Sect.~\ref{sec:vae}~\cite{kingma2013auto,kingma2019introduction}. 

The relation between supervised and unsupervised learning is
sometimes characterized as follows: supervised learning aims to learn
the conditional probability distribution $p(x|y)$ of input data
conditioned on a label $y$, whereas unsupervised learning aims to
learn the \emph{a priori} probability distribution
$p(x)$~\cite{hinton1999unsupervised}.

We will encounter unsupervised methods in this book in a few places,
specifically, when autoencoders and dimension reduction are discussed,
for example, in Chap.~\ref{chap:model}. At the end of this book
explainable artificial intelligence is discussed, where interpretable
models play an important role, in Chap.~\ref{chap:conc}.

\subsection{Reinforcement Learning}
The last machine learning paradigm is, indeed,  reinforcement
learning.  
There are three differences between reinforcement learning and the
previous paradigms.
First, reinforcement learning learns by \emph{interaction}; in contrast to supervised and unsupervised learning, in
reinforcement learning data items come one by one. The dataset is
produced dynamically, as it were. The objective in reinforcement
learning is to find the policy: a function that gives us the best
action  in each state that the world can be in.

\begin{figure}[t]
\begin{center}
    \includegraphics[width=8cm]{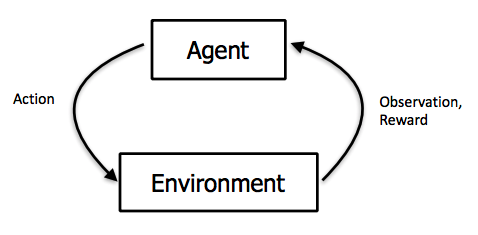}
\caption{Agent and Environment}\label{fig:agent1}
\end{center}
\end{figure}
\index{reinforcement learning}\index{agent}\index{environment}
The approach of reinforcement learning is to learn the policy for the world by
interacting with it. In reinforcement learning we recognize  an \emph{agent}, that does the
learning of the policy, and an \emph{environment}, that provides
feedback to the agent's actions (and that
performs state changes, see Fig.~\ref{fig:agent1}). In reinforcement learning, the agent stands for the human,
and the environment for the world. The
goal of  reinforcement learning  is  to find the 
actions for each state that maximize the long term
accumulated expected reward. This optimal function of states to actions is called the
\emph{optimal policy}.

In reinforcement learning there is no teacher or supervisor, and there
is no static 
dataset.  There is, however, 
the environment,  that will tell us how good
the state is in which we find ourselves. This brings us to the second
difference: the \emph{reward value}. Reinforcement learning gives
us partial information, a number indicating the quality of the action
that brought us to our state, where supervised learning gives full
information: a label that provides the
correct answer  in that state (Table~\ref{tab:suprl1}).
In this sense, reinforcement 
learning is in between supervised learning, in which
all data items have a label, and unsupervised learning,
where no data  has a label.

The third difference is  that reinforcement learning is 
used to solve \emph{sequential decision problems}. Supervised and
unsupervised learning learn single-step relations between items;
reinforcement learning learns a policy, which is the answer to
a multi-step  problem. Supervised learning can classify a set of
images for you; unsupervised learning can tell you which items belong
together; reinforcement learning can tell you  the winning sequence of
moves in a game of chess, or the action-sequence that robot-legs
need to take in order to walk.

\begin{table}[t]
  \begin{center}
  \begin{tabular}{lcc}
    {\bf Concept}&{\bf Supervised Learning} \qquad& {\bf Reinforcement Learning} \\
    \hline\hline
    Inputs $x$\qquad\qquad&Full  dataset of states   &  Partial (One
                                                       state at a time) \\ 
    Labels $y$&Full (correct action) & Partial (Numeric action reward) \\
    \hline
  \end{tabular}
  \caption{Supervised vs.\ Reinforcement Learning}\label{tab:suprl1}
\end{center}
\end{table}

These three differences have consequences.
Reinforcement learning provides the data to the learning algorithm
step by step, action by
action; whereas in supervised learning the data is provided all at
once in one
large dataset. The step-by-step approach is well suited 
to sequential decision problems.
On the other hand, many  deep learning methods were developed for
supervised learning and may work differently when  data items are
generated one-by-one. Furthermore, since  actions are selected using
the policy function, and action rewards are used to update this same
policy function, there is a possibility of circular feedback and local
minima. Care must be taken to ensure  convergence to global optima in
our methods.
Human learning also suffers from this
problem, when a stubborn child refuses to explore outside of its comfort
zone. 
This topic is is discussed in Sect.~\ref{sec:multiarmed}.

Another difference  is that in  supervised learning the 
pupil learns from a finite-sized teacher (the dataset), and at some
point may have learned all there is to learn. The  reinforcement
learning paradigm allows a learning setup where the
agent can continue to sample  the environment indefinitely, and will continue
to become smarter as long as the environment remains challenging
(which can be a long time, for example in games such as chess and Go).\footnote{In fact, some argue that reward is
  enough for artificial general intelligence, see Silver,
  Singh, Precup, and Sutton~\cite{silver2021reward}.}

For these reasons there is great interest in reinforcement learning,
although getting the methods to work is often harder than for supervised learning.

Most classical reinforcement  learning use tabular
methods that work for low-dimensional problems with small state
spaces.  Many real world problems are complex and high-dimensional, with
large state spaces.
Due to steady improvements in learning algorithms, datasets, and
compute power, deep learning methods have become quite
powerful. 
Deep reinforcement
learning methods have emerged that successfully combine step-by-step
sampling in high-dimensional problems with large state spaces. We will
discuss these methods in the subsequent chapters of this book.

\section{Overview of the Book}

The aim of this book is to present
the latest insights in deep reinforcement learning in a single
comprehensive volume, suitable for teaching a graduate level
one-semester course. 

In addition to covering state of the art 
algorithms, we cover necessary background in classic reinforcement
learning and in deep learning. We also cover   advanced,
forward looking developments in self-play, and in multi-agent, hierarchical, and meta-learning.

\subsection{Prerequisite Knowledge}

In an effort to be comprehensive, we make modest assumptions about  previous
knowledge. We assume a  bachelor level of computer science or
artificial intelligence, and an interest in artificial
intelligence and machine learning.
A good introductory textbook 
on   artificial intelligence is  Russell and Norvig: \emph{Artificial  
  Intelligence, A Modern Approach}~\cite{russell2016artificial}.

\begin{figure}[t]
\begin{center}
\begin{tikzpicture}[>=triangle 45,
  desc/.style={
		scale=1.0,
		rectangle,
		rounded corners,
		draw=black, 
		}]
  \node [desc,dashed] (dl) at    (6,2) {B. Deep Supervised  Learning};
  \node [desc] (tab) at           (0,2) {2. Tabular Value-Based
    Reinforcement Learning};
  \node [desc] (deep)  at       (3,3) {3. Deep Value-Based
    Reinforcement Learning};
  \node [desc] (pol) at        (3,4) {4. Policy-Based Reinforcement Learning};
  \node [desc] (model) at       (3,5) {5. Model-Based Reinforcement Learning};
  \node [desc] (two) at        (3,6) {6. Two-Agent Self-Play};
  \node [desc] (multi) at      (3,7) {7. Multi-Agent Reinforcement Learning};
  \node [desc] (hier) at      (3,8) {8. Hierarchical Reinforcement Learning};
  \node [desc] (meta) at      (3,9) {9. Meta-Learning};
  \node [desc] (conc) at      (3,10) {10. Further Developments};
  \draw [semithick,dashed,->] (dl) -- (deep);
  \draw [semithick,->] (tab) -- (deep);
  \draw [semithick,->] (deep) -- (pol);
  \draw [semithick,->] (pol) -- (model);
  \draw [semithick,->] (model) -- (two);
  \draw [semithick,->] (two) -- (multi);
  \draw [semithick,->] (multi) -- (hier);
  \draw [semithick,->] (hier) -- (meta);
  \draw [semithick,->] (meta) -- (conc);

\end{tikzpicture}
\caption[Chapter overview]{Deep Reinforcement Learning is built on
  Deep Supervised Learning and Tabular Reinforcement Learning}\label{fig:ov}
  \end{center}
\end{figure}
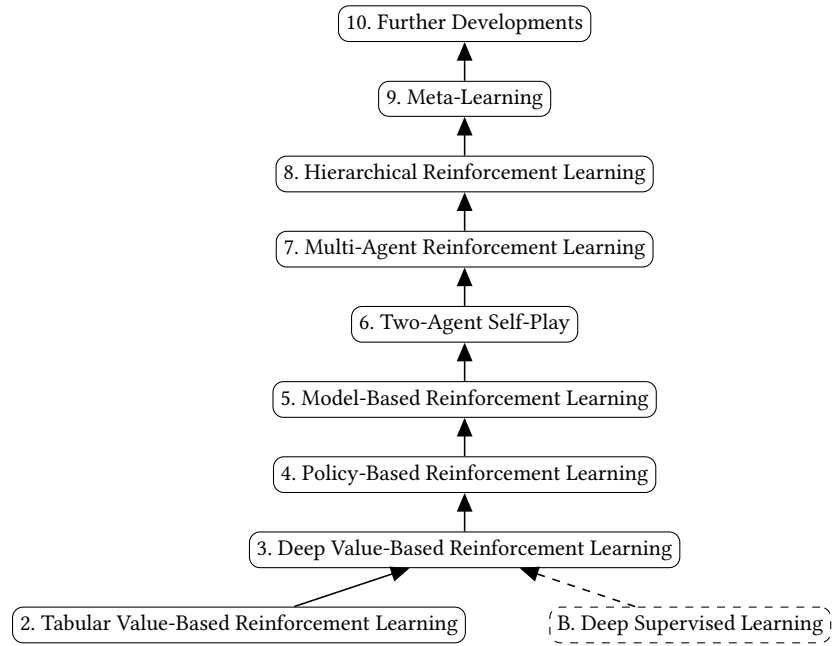

Figure~\ref{fig:ov} shows an overview of the structure of the
book. Deep reinforcement learning combines deep supervised learning
and classical (tabular) reinforcement learning. The figure shows how the
chapters are built on this dual foundation. For  deep
reinforcement learning, the field of deep supervised learning is of
great importance. It is a large field; deep, and
rich. Many students may have followed a
course on deep learning, if not, Appendix~\ref{app:deep} provides you
with the necessary background (dashed). Tabular reinforcement
learning, on the other hand, 
may be new to you, and we start our story with this topic in
Chap.~\ref{chap:drl}.

We also assume undergraduate level familiarity with the Python programming 
language. Python   has become the
programming language of choice for
machine learning research, and the host-language of most machine
learning packages. All example code in this book is in Python, and
major machine learning environments such as scikit-learn, TensorFlow,
Keras and PyTorch work 
best from Python. See \url{https://www.python.org} for pointers on
how to get started in Python. Use the latest stable version, unless the text
mentions otherwise.

We  assume an undergraduate level of familiarity with
mathematics---a basic 
understanding of set theory, graph theory, probability theory and
information theory is
necessary, although this is not a book of
mathematics. Appendix~\ref{app:math} contains a summary to refresh
your mathematical knowledge, and to provide an  introduction to the notation
that is used in the book.

\subsubsection*{Course}
There is a lot of material in the chapters, both basic and advanced,
with many pointers to the literature. One option is to teach a single
course about all topics in the book. Another option is to go slower
and deeper, to spend sufficient time on the basics,
and create a course about Chaps.~\ref{chap:feedback}--\ref{chap:model}
to cover the basic topics (value-based, policy-based, and model-based learning),
and to create a separate course about
Chaps.~\ref{chap:self}--\ref{chap:meta} to cover the more advanced topics of
multi-agent, hierarchical, and meta-learning.

\subsubsection*{Blogs and GitHub}
The field of deep reinforcement learning is a highly active field, in
which theory and practice go hand in hand. The culture of the field is
open, and you will easily find many blog posts about interesting
topics, some quite good. Theory drives experimentation, and experimental results drive
theoretical insights. Many researchers publish their papers on arXiv
and  their
algorithms, hyperparameter settings  and  environments on
GitHub.

In this book we aim for the same atmosphere. 
Throughout the text we provide links to
code, and we challenge you with hands-on sections to get your hands
dirty to perform your own experiments. 
All links to web pages that we use have
been stable for some time.
\begin{tcolorbox}
{\bf Website:}   \url{https://deep-reinforcement-learning.net}  is the
companion website for this book. It
contains  updates, slides, and other course material that you are
welcome to explore and use.
\end{tcolorbox}

\subsection{Structure of the Book}

The field of deep reinforcement learning consists of two main areas:
model-free reinforcement learning  and
model-based reinforcement learning. Both areas have two subareas.
The chapters of this book are organized according to this structure.
\begin{itemize}
\item Model-free methods
  \begin{itemize}
  \item Value-based methods: Chap.~\ref{chap:drl} (tabular) and \ref{ch:play} (deep)
  \item Policy-based methods: Chap.~\ref{ch:pol}
  \end{itemize}
\item Model-based methods
  \begin{itemize}
  \item Learned model: Chap.~\ref{chap:learned}
  \item Given model: Chap.~\ref{chap:given}
  \end{itemize}
\end{itemize}
Then, we have three chapters on more specialized topics.
\begin{itemize}
\item Multi-agent reinforcement learning: Chap.~\ref{chap:team}
\item Hierarchical reinforcement learning: Chap.~\ref{chap:hier}
\item Transfer and Meta-learning: Chap.~\ref{chap:txl}
\end{itemize}
Appendix~\ref{app:deep} provides a necessary review of
deep supervised learning.

The  style of
each chapter is to first provide the main idea of the chapter in an intuitive
\emph{example}, to then explain the kind of \emph{problem} 
to be solved, and then to discuss
algorithmic concepts  that \emph{agents} use, and the
\emph{environments} that have been solved in practice with these algorithms. The
sections of the chapters are named accordingly: their names end in problem-agent-environment. At the end of each chapter we
provide questions for quizzes to check your understanding of 
the concepts, and we provide exercises for larger programming
assignments (some doable, some quite challenging). We also end each chapter
with a summary and references to further reading.

Let us now  look in more detail at what topics the chapters cover.

\subsubsection*{Chapters}
After this introductory chapter,  we continue with \emph{Chap.~\ref{chap:drl}}, in
which we discuss in detail the basic concepts of tabular (non-deep) reinforcement
learning. We start with Markov decision processes and discuss them at
length. We will introduce tabular planning and learning, and important
concepts such as state, action, reward, value, and policy. We will
encounter the first, tabular, value-based model-free learning
algorithms (for an overview, see
Table~\ref{tab:val}). Chapter~\ref{chap:drl} is the only non-deep
chapter of the book. All other chapters cover deep methods.

\emph{Chapter~\ref{ch:play}}   explains deep value-based reinforcement
learning. 
The chapter covers the first deep algorithms that have been
devised to find the optimal policy. We will still be working in the value-based,
model-free, paradigm. At the end of the chapter we will analyze a player that
teaches itself how to play 1980s Atari 
video games. Table~\ref{tab:rainbow} lists some of the many stable deep value-based
model-free algorithms.

Value-based reinforcement learning works well with applications such
as games, with discrete action spaces. 
The next chapter, \emph{Chap.~\ref{ch:pol}}, discusses a different
approach: deep policy-based reinforcement learning
(Table~\ref{tab:pol}). In addition to discrete spaces, this approach is also
suited for  continuous actions spaces, such as robot arm
movement, and simulated articulated locomotion. We see how a simulated
half-cheetah teaches itself  to run.

The next chapter, \emph{Chap.~\ref{chap:learned}}, introduces deep model-based
reinforcement learning with a learned model, a method that first builds up a transition
model of the environment before it builds the policy. Model-based
reinforcement learning holds the promise of higher sample efficiency,
and thus faster learning. New developments, such as latent models, are
discussed. Applications are both in robotics and in games 
(Table~\ref{tab:overview}).

The next chapter, \emph{Chap.~\ref{chap:given}},  studies how a self-play system can be
created for applications where the transition model is 
given by the problem description. This is the case in two-agent games, where the rules for  moving in the game
determine the transition function. We study how TD-Gammon and
AlphaZero achieve \emph{tabula rasa} learning: teaching themselves from zero knowledge to world
champion level play through playing against a copy of itself
(Table~\ref{tab:az}). In this chapter deep residual networks and Monte 
Carlo Tree Search result in curriculum learning.

\emph{Chapter~\ref{chap:team}} introduces  recent
developments in deep multi-agent and team learning. The chapter covers
competition and collaboration, 
population-based methods, and playing in teams. Applications of
these methods are found in games such as poker and StarCraft (Table~\ref{tab:multigame}).

\emph{Chapter~\ref{chap:hier}} covers deep hierarchical reinforcement
learning. Many tasks exhibit an inherent hierarchical structure, in
which clear subgoals can be identified. The options framework is
discussed, and methods that can identify subgoals, subpolicies, and
meta policies. Different approaches for tabular and deep  hierarchical methods are
discussed (Table~\ref{tab:hrl}).

The final technical chapter, \emph{Chap.~\ref{chap:txl}}, covers deep meta-learning, or learning to learn. One of the major hurdles in
machine learning is the long time it takes to learn to solve a new
task. Meta-learning and transfer learning aim to speed up learning of
new tasks by using information that has been learned previously for
related tasks; algorithms are listed in Table~\ref{tab:meta}. At the
end of the chapter we will experiment with 
few-shot learning, where a task has to be learned without having
seen more than a few training examples.

\emph{Chapter~\ref{chap:conc}} concludes the book by reviewing what we have
learned, and by looking ahead into what the future may bring.

 \emph{Appendix~\ref{app:math}} provides
mathematical background information and notation.
\emph{Appendix~\ref{ch:deep}} provides a chapter-length overview of machine learning and deep supervised learning. If you wish
to refresh your knowledge of deep learning, please go to this appendix before you read
Chap.~\ref{ch:play}. \emph{Appendix~\ref{ch:env}} provides lists of useful
software environments and software packages for deep reinforcement learning.


\chapter{Tabular Value-Based Reinforcement Learning}\label{chap:drl}\label{chap:feedback}\label{ch:tab}

This chapter will introduce 
the classic, tabular, field of reinforcement learning, to build a
foundation for the next chapters. 
First,  we will introduce the concepts of agent and environment. Next come Markov
decision processes, the formalism that is used to reason mathematically
about reinforcement learning. We  discuss at some
length the elements of reinforcement learning: states,
actions, values, policies. 

We learn about transition functions, and solution
methods that are based on  dynamic
programming using the transition model. There are many situations where agents do
not have access to the transition model, and state and reward information must
be acquired from  the environment.
Fortunately, 
methods  exist to find the optimal policy without a model, by querying
the environment. These methods, appropriately named 
model-free methods, will be introduced in this chapter.
 Value-based model-free methods are the most basic learning approach of
reinforcement learning. They work well in problems with
deterministic environments and discrete action spaces, such as mazes
and games.
Model-free learning makes few demands on the environment, building up
the policy function $\pi(s)\rightarrow a$ by sampling  the environment.

After  we have discussed
these concepts, it is time to apply them, 
and to understand the kinds of sequential decision problems that we  can 
solve.  We will look at Gym, a collection of reinforcement learning
environments. We will also look at
simple Grid world puzzles, and see how to navigate those.

This is a non-deep chapter: in this chapter functions are 
 exact,  states are stored in tables, an approach that works as long
 as problems are small enough to fit in memory. The
next chapter shows how function approximation with neural networks
 works when there are more states than  fit in memory.

The chapter is concluded with  exercises, a summary, and pointers
to further reading.

\section*{Core Concepts}
\begin{itemize}
\item Agent, environment
\item MDP: state, action, reward, value, policy
\item Planning and learning
\item Exploration and exploitation
\item Gym, baselines
\end{itemize}

\section*{Core Problem}
\begin{itemize}
\item Learn a policy from interaction with the environment
\end{itemize}

\section*{Core Algorithms}
\begin{itemize}
\item Value iteration (Listing~\ref{lst:val-it})
\item Temporal difference learning (Sect.~\ref{sec:td})
\item Q-learning (Listing~\ref{lst:q})
\end{itemize}

\section*{Finding a Supermarket}

Imagine that you have just moved to a new city, you are hungry, and you  want to
buy some  groceries. There is a somewhat unrealistic catch: you do not
 have a map of the city and  you forgot to charge your  smartphone. It is a sunny day, you
put on your hiking 
shoes, and after some  random exploration you have found a way
to a supermarket and have bought your groceries. You have carefully
noted your route in a notebook, and you retrace
your steps, finding your way back to your new home.

What will
you do the next time that you need groceries? One option is to follow
exactly the same route, exploiting your current knowledge. This option is
guaranteed to bring you to the 
store, at no additional cost for exploring possible alternative routes. Or
you could be adventurous, and explore, trying to find a new route
that may actually be quicker than the old route. Clearly, there is a
trade-off: you should not spend so much time  exploring that you can
not recoup the gains of a potential shorter 
route before you move elsewhere.

Reinforcement learning is a natural way of learning the optimal route
as we go, by trial and error, from the effects of the actions that we
take in our environment. 

This little story contained many of the elements of a reinforcement learning
problem, and how to solve it. There is  an \emph{agent}
(you), an \emph{environment} (the city), there are \emph{states} (your location at different
points in time), \emph{actions} (assuming a Manhattan-style grid, moving a block left, right, 
forward, or back), there are  \emph{trajectories} (the routes to the
supermarket that you tried), there is a \emph{policy} (that tells which action you will
take at a particular location), there is a concept of \emph{cost/reward} (the length of your current
path), we see
\emph{exploration} of new routes, \emph{exploitation} of old routes,  a
trade-off between them, and your notebook in which you have been sketching a map of
the city  (your local \emph{transition
model}).

By the end of this chapter you will have learned which role   all
these topics play in reinforcement learning.

\section{Sequential Decision Problems}

Reinforcement learning is used to solve sequential decision problems~\cite{arulkumaran2017deep,franccois2018introduction}. 
Before we dive into the algorithms, let us  have a closer look at these  problems, to better
understand the challenges that the agents must solve.

\begin{figure}[t]
  \begin{center}
    \includegraphics[width=4cm]{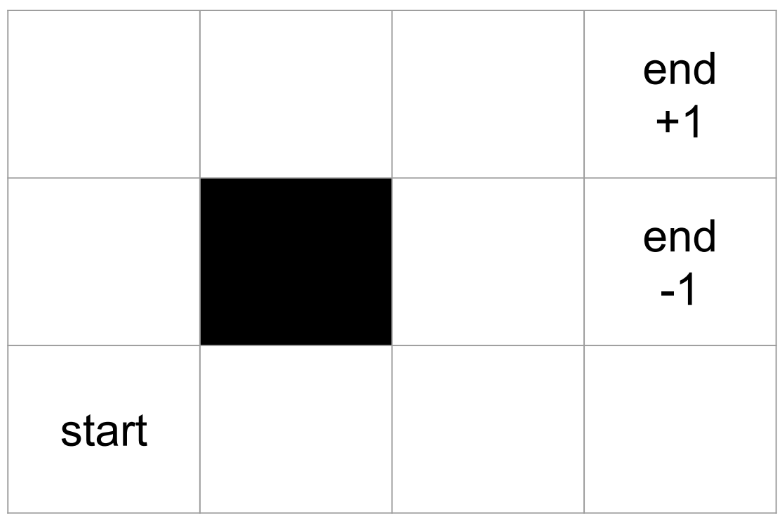}
  \end{center}
  \caption[Grid World]{Grid World with a goal, an ``un-goal,'' and a wall}\label{fig:gridworld}
\end{figure}

In a sequential decision problem
the agent has to make a sequence of decisions in order to solve a
problem. Solving implies to find the
sequence with the highest (expected cumulative future) reward. The solver is called the \emph{agent},
and the problem is called \emph{environment} (or sometimes the
\emph{world}).

We will now discuss basic examples of sequential decision problems.

\subsubsection*{Grid Worlds}
Some of the first environments that we encounter in reinforcement learning
are \emph{Grid worlds} (Fig.~\ref{fig:gridworld}).  These environments 
consist of a rectangular grid of squares, with a start square, and a
goal square. The aim is for the agent to find the sequence of actions
that it must
take (up, down, left, right) to arrive at the goal square. In fancy
versions a ``loss'' square is added, that scores minus points, or a
``wall'' square, that is impenetrable for the agent.
By exploring the grid, taking different actions, and recording the
reward (whether it reached the goal square), the agent can find a
route---and when it has a route, it can try to improve that route, to
find a shorter route to the goal.

\begin{figure}[t]
  \begin{center}
    \includegraphics[width=8cm]{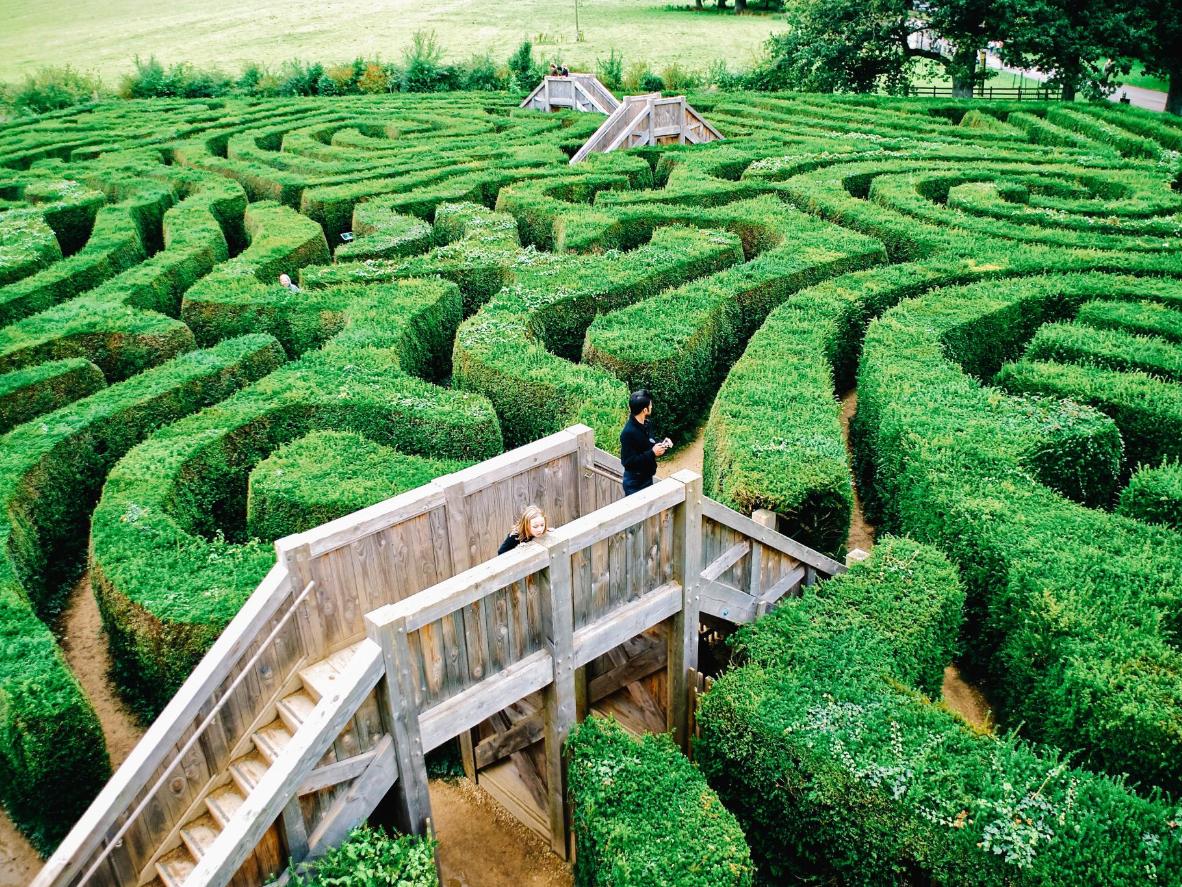}
  \end{center}
  \caption{The Longleat Hedge Maze in Wiltshire, England}\label{fig:maze}
\end{figure}

Grid world is a simple environment that is well-suited for
manually playing around with reinforcement learning algorithms, to
build up intuition of what the algorithms do.
In this chapter we will model reinforcement learning problems formally,
and  encounter algorithms that find 
optimal routes in Grid world.

\begin{figure}[t]
  \begin{center}
    \includegraphics[width=8cm]{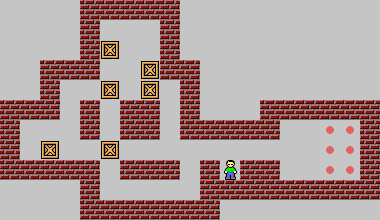}
  \end{center}
  \caption{Sokoban Puzzle~\cite{chao2013}}\label{fig:sokoban}
\end{figure}

\subsection*{Mazes and Box Puzzles}\label{sec:mazes}
After Grid world problems, there are more complicated problems, with
extensive wall structures to make navigation more difficult (see Fig.~\ref{fig:maze}). 
Trajectory planning algorithms play a central role in
robotics~\cite{latombe2012robot,gasparetto2015path}; there is a long
tradition of using 2D and 3D mazes for
path-finding problems in reinforcement learning.   The Taxi domain
was introduced by Dietterich~\cite{dietterich2000hierarchical}, and box-pushing
problems such as Sokoban have also been used frequently~\cite{junghanns2001sokoban,dor1999sokoban,murase1996automatic,zhou2013tabled},
see Fig.~\ref{fig:sokoban}. The challenge in Sokoban is that boxes can
only be pushed, not pulled. Actions can have the effect of creating an
inadvertent dead-end for into the future, making Sokoban a difficult
puzzle to play. The
action space of these puzzles and mazes is discrete.

Small versions of the mazes can be solved exactly by planning, 
larger instances are only suitable for approximate planning or
learning methods.  Solving these planning problems exactly is   NP-hard or
PSPACE-hard~\cite{culberson1997sokoban,hearn2009games}, as a
consequence the
computational time required to solve  problem instances exactly grows
exponentially with the problem size, and becomes quickly infeasible
for all but the smallest problems.

Let us see how we can model agents to act in these types of environments.

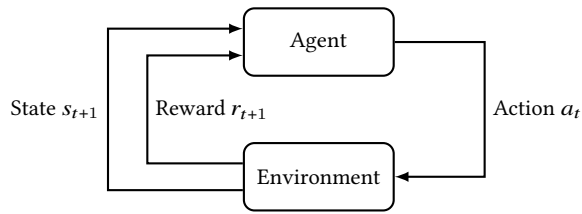
\begin{figure}[t]
\begin{center}

\tikzstyle{block} = [rectangle, draw, 
    text width=6em, text centered, rounded corners, minimum height=3em]
    
\tikzstyle{line} = [draw, -latex]

\begin{tikzpicture}[node distance = 6em, auto, thick]
    \node [block] (Agent) {Agent};
    \node [block, below of=Agent] (Environment) {Environment};
    
     \path [line] (Agent.0) --++ (4em,0em) |- node [near start]{Action $a_t$} (Environment.0);
     \path [line] (Environment.190) --++ (-6em,0em) |- node [near start] {State  $s_{t+1}$} (Agent.170);
     \path [line] (Environment.170) --++ (-4.25em,0em) |- node [near start, right] {Reward $r_{t+1}$} (Agent.190);
\end{tikzpicture}
\caption{Agent and environment~\cite{sutton2018introduction}}\label{fig:agent}
\end{center}
\end{figure}
\index{reinforcement learning}\index{agent}\index{environment}

\section{Tabular Value-Based Agents}

Reinforcement learning  finds the best policy to operate in
the environment by interacting with it. 
The reinforcement learning paradigm consists of an agent (you, the learner) and an 
environment (the world, which is in a certain state, and gives you
feedback on your actions).

\subsection{Agent and Environment}
In Fig.~\ref{fig:agent} the  agent and environment are shown, together
with action $a_t$, next state $s_{t+1}$, and its reward $r_{t+1}$.  Let us have a closer
look at the figure.

The environment is in a certain  state $s_t$ at time $t$. Then, the
agent performs  action $a_t$,
resulting in a transition  in the environment from state $s_t$ to  
$s_{t+1}$ at the next time step, also denoted as $s\to s'$. Along with this new state
comes a reward value $r_{t+1}$ (which may be a positive or a negative value).
The goal of reinforcement learning is to find the sequence of  actions that
gives the best reward. More formally, the goal is to find the optimal policy function
$\pi^\star$ that gives in each state the best action to take in that
state.
By trying different actions, and accumulating the rewards, the agent
can find the best action
for each state. In this way, with the reinforcing reward values, the
optimal policy is  learned   from repeated interaction with the
environment, and the problem is ``solved.''

In reinforcement
learning the environment gives us only a number as an indication of
the quality of an action that we performed,
and we are left to derive the correct action policy from that, as we
can see in Fig.~\ref{fig:agent}. 
On the other hand, reinforcement learning allows us to generate as
many action-reward pairs as we need, without   a large
hand-labeled dataset, and we can choose ourselves which actions to try.

\subsection{Markov Decision Process }
\label{sec:mdp}\index{Markov decision process}

Sequential decision problems can be 
modelled as Markov decision processes
(MDPs)~\cite{littman1994markov}. 
Markov decision problems  have  the Markov property: the next
state  depends only on the current state and the actions available in
it (no historical
memory of previous states or information from elsewhere influences the
next state)~\cite{howard1964dynamic}.
The no-memory property is important
because it makes reasoning about future states possible using only the
information present in the current state. If previous
histories would  influence the current state, and these would all have
to be taken into account, then reasoning about
the current state would be much harder or even infeasible.

Markov processes are named after Russian
mathematician Andrey Markov (1856--1922) who is best known for his
work on these stochastic processes. See~\cite{arulkumaran2017deep,franccois2018introduction}
for an introduction into MDPs. The MDP formalism is the mathematical basis
under reinforcement learning, and we will introduce the relevant
elements in this chapter. We follow
Moerland~\cite{moerland2021lecture} and Fran\c cois-Lavet et al.~\cite{franccois2018introduction} for some of the notation and examples in
this section.

\subsubsection*{Formalism}

We define a \gls{Markov decision process} for reinforcement learning   as a 5-tuple $(S,A,T_a,R_a,\gamma)$:
\begin{itemize}
\item $S$ is a finite set of legal {\em states} of the environment; the initial state is denoted as $s_0$
\item $A$ is a finite set of {\em actions} (if the set of actions differs
  per state, then $A_s$ is the finite set of
  actions in state $s$)
\item $T_a(s, s^\prime) = \Pr(s_{t+1} = s^\prime | s_t = s, a_t = a)$
  is the  probability that action $a$ in state $s$ at time $t$ will
  {\em transition} to state $s^\prime$ at time $t+1$ in the
  environment 
\item $R_a(s,s^\prime)$ is the {\em reward} received after action $a$
  transitions state $s$ to state $s^\prime$
\item $\gamma \in[0,1]$ is the {\em discount factor} representing the
  difference between future and present rewards.
\end{itemize}

\subsubsection{State $S$}\index{state}
Let us have a deeper look at the  Markov-tuple \gls{state}, $A, T_a, R_a, \gamma$, to see their role in
the reinforcement learning paradigm, and how, together, they can model
and describe reward-based learning processes.

At the basis of every Markov decision process is a description of the
state $s_t$ of the system
at a certain time $t$.

\subsubsection*{State Representation}
The state $s$  contains the information to uniquely represent the configuration
of the environment.

Often there is a straightforward way to
uniquely represent the state in a computer memory. For the
supermarket example, each identifying location  is a state (such as: I am at the corner of
8th Av and 27nd St).   For chess, this
can be the location of all pieces on the board (plus  information for
the 50 move repetition rule, castling rights, and en-passant state). For robotics this can be the orientation of
all joints of the robot, and the location of the limbs of the
robot. For Atari, the state comprises the values of all screen
pixels.

Using its current behavior policy, the agent 
chooses an action $a$, which is performed in the environment.  How
the environment reacts to the action is defined by the  transition
model $T_a(s,s')$ that is internal to the environment, which the agent does not  know. The
environment returns the new state $s'$, as well as  a reward value
$r'$ for the new state.

\subsubsection*{Deterministic and Stochastic Environment}\index{stochastic
  environment}\index{deterministic environment}

In discrete deterministic environments the transition function
defines a one-step transition, as each action (from a certain old
state) deterministically leads
to a single new state. This is the case in Grid worlds, Sokoban, and
in games such as chess and
checkers, where a move action deterministically leads to one new board
position.

An example of a non-deterministic situation is a robot movement in an
environment. In a certain state, a
robot arm is holding a bottle. An agent-action can be turning the bottle in a
certain orientation (presumably to pour a drink in a cup). The next
state may be a full cup, or it may be a 
mess, if the bottle was not poured in the correct orientation, or
location, or if something happened in the environment such as someone
bumping  the table. The
outcome of the action is unknown beforehand by the agent, and depends
on elements in the environment, that are not known to the agent.

\subsubsection{Action $A$}
Now that we have looked at the state, it is time to look at the second
item that defines an MDP, the \emph{action}. 

\subsubsection*{Irreversible Environment Action}
\index{irreversible actions}
When the agent is in state $s$, it chooses an action \gls{action}  to perform,
based on its current behavior 
policy $\pi(a|s)$
(policies are explained soon).
The agent communicates
the selected action $a$ to the environment (Fig.~\ref{fig:agent}).   For
the supermarket example, an example of an action could be 
walking along a block in a certain  direction (such as:
\emph{East}). For Sokoban, an action 
can be pushing  a box to a new location in the
warehouse. Note that in different states the possible actions may  differ. For
the supermarket example, walking East may 
not be possible at each street corner, and in Sokoban pushing a box 
in a certain direction will only be possible in states where this
direction is not blocked by a wall. 

An action changes the state of the environment
irreversibly. In the reinforcement learning paradigm, there is no \emph{undo}
operator for the environment (nor is there in the real world). When the
environment has performed a state transition, it is final. The new
state is communicated 
to the agent, together with a reward 
value. The actions that the agent performs in the environment are also
known as its \emph{behavior}, just as the actions of a human in the world
constitute the human's behavior.

\subsubsection*{Discrete or Continuous Action Space}
\index{continuous action space}\index{discrete action space}\label{sec:discrete}
The actions are discrete in some applications, continuous in
others. For example, the actions in board games, and choosing a direction in a
navigation task in a grid, are  discrete.

In contrast, arm and joint movements of robots, and bet sizes in certain games, are
 continuous (or span a very large range of values). Applying algorithms
to continuous or very large action spaces
either requires discretization of the continuous space (into buckets) or the
development of a different kind of algorithm. As we will see in
Chaps.~\ref{chap:value} and \ref{chap:policy}, value-based methods 
work well for discrete action spaces, and policy-based methods
 work well for both action spaces.

For the supermarket example we can actually choose between modeling our
actions discrete or continuous. From every state, we can  move
any number of steps, small or large, integer or fractional, in any
direction. We can even walk a curvy 
path. So,  strictly speaking, the action space is continuous. However,
if, as in some cities, the streets are organized in a 
rectangular Manhattan-pattern, then it makes sense to
discretize the continuous space, and to only consider discrete actions
that take us to the next street corner. Then, our action space has
become discrete, by using extra knowledge of the problem
structure.\footnote{If we assume that supermarkets are large, block-sized, items that
typically can be found on street corners, then we can discretize the
action space. Note that we may miss small
sub-block-sized supermarkets, because of this simplification. Another,
better, simplification, would be to discretize the action space into
walking distances of the size of the smallest supermarket that we
expect to ever encounter.}

\subsubsection{Transition $T_a$}

After having discussed state and  action,  it is time to look at
the transition function $T_a(s,s')$.   
 The transition function \gls{transition}$_a$  determines how the state
changes  after an action has been selected.  In model-free
reinforcement learning the transition function is implicit to the
solution algorithm: the environment has access to the transition
function, and uses it to compute the next state $s'$, but the agent has
\emph{not}. (In
Chap.~\ref{chap:learned} we will discuss model-based reinforcement
learning. There the agent has its own transition function, an
approximation of the environment's transition function, which is
learned from the environment feedback.)

\begin{figure}[t]
  \begin{center}
    \begin{tabular}{c|c|c}
      \tikzset{
  treenode/.style = {align=center, inner sep=0pt, text centered,
    font=\sffamily},
  arn_n/.style = {treenode, circle, white, draw=black,
    fill=black, text width=2mm},
  arn_r/.style = {treenode, circle, black, draw=black, 
    text width=3mm, thick}
}

\begin{tikzpicture}[->,>=stealth',level/.style={sibling distance = 1.2cm/#1,
  level distance = 1cm}] 
\node [arn_r,label=above:{$s$}] {}
    child{ node [arn_n] {} 
            child{ node [arn_r] {} 
            }
            child{ node [arn_r] {}
            }                            
    }
    child{ node [arn_n] {} 
            child{ node [arn_r] {}
            }            
            child{ node [arn_r] {}
            }
            edge from parent node[right] {$\pi$} 
    }
    child{ node [arn_n,label=above:{$a$}] {}
            child{ node [arn_r] {} 
            }
            child{ node [arn_r,label=right:{$s'$}] {}  edge from
              parent node[right] {$t_a, r_a$} 
            }
    }
;  
\end{tikzpicture}
      &
      \tikzset{
  treenode/.style = {align=center, inner sep=0pt, text centered,
    font=\sffamily},
  arn_n/.style = {treenode, circle, white, draw=black,
    fill=black, text width=2mm},
  arn_r/.style = {treenode, circle, black, draw=black, 
    text width=3mm, thick}
}

\begin{tikzpicture}[->,>=stealth',level/.style={sibling distance = 1.2cm/#1,
  level distance = 1cm}] 
\node [arn_r,label=above:{$s$}] {}
    child{ node [arn_n] {} 
            child{ node [arn_r] {}
            }                            
    }
    child{ node [arn_n] {} 
            child{ node [arn_r] {}
            }
            edge from parent node[right] {$\pi$} 
    }
    child{ node [arn_n,label=above:{$a$}] {}
            child{ node [arn_r,label=right:{$s'$}] {}  edge from
              parent node[right] {$t_a, r_a$} 
            }
    }
;  
\end{tikzpicture}
      &
        \tikzset{
  treenode/.style = {align=center, inner sep=0pt, text centered,
    font=\sffamily},
  arn_n/.style = {treenode, circle, white, draw=black,
    fill=black, text width=2mm},
  arn_r/.style = {treenode, circle, black, draw=black, 
    text width=3mm, thick}
}

\begin{tikzpicture}[->,>=stealth',level/.style={sibling distance = 1.2cm/#1,
  level distance = 1cm}] 
\node [arn_r,label=above:{$s$}] {}
    child{ node [arn_r] {} 
    }
    child{ node [arn_r] {} 
            edge from parent node[right] {$\pi$} 
    }
    child{ node [arn_r,label=right:{$a,s'$}] {}
            edge from parent node[right] {$t_a, r_a$} 
    }
;  
\end{tikzpicture}
        \end{tabular}
\caption[Backup Diagrams]{Backup Diagrams for MDP Transitions: Stochastic (left) and
  Deterministic (middle and right)~\cite{sutton2018introduction}}\label{fig:mm}\label{fig:rltree}
\end{center}
\end{figure}
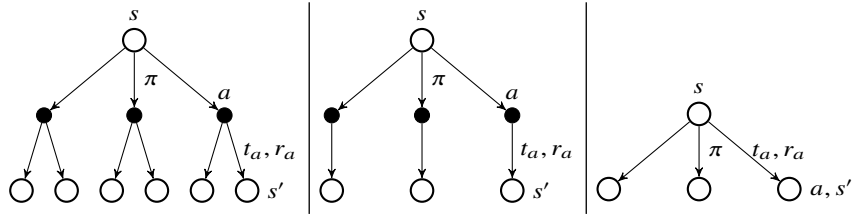

\subsubsection*{Graph View of the State Space}\index{graph view of
  state space}
We have  discussed states, actions and transitions.
The dynamics of
the MDP are modelled by transition
function $T_a(\cdot)$ and reward function $R_a(\cdot)$.
The imaginary space of all possible states is called the \emph{state space}. The state space
is typically large.
The two functions define  a two-step transition from state $s$ to $s'$,
via action 
$a$: $s\to a \to s'$.

To help our understanding of the transitions
between states we can use a 
graphical depiction, 
as in Fig.~\ref{fig:rltree}.

In the figure, states and actions are depicted as nodes (vertices), and 
transitions are links (edges) between the nodes. States are drawn as open
circles, and actions as smaller black circles. In a certain state $s$,
the agent can choose which action $a$ to perform, that is then acted out in
the environment. The environment  returns the new state $s'$ and
the reward $r'$.

Figure~\ref{fig:rltree} shows a transition graph of the elements of the MDP tuple $s, a, t_a, r_a$ as
well as $s^\prime$,  and policy $\pi$, and how the value can be calculated. The root node
at the top is state $s$, where policy $\pi$ allows the agent to choose between three
actions $a$, that, following distribution Pr,  each can transition to
two possible
states $s^\prime$, with their reward $r'$. In the figure, a single
transition is shown. Please use your imagination to
picture the other transitions as the graph extends down.

In the left panel of the figure the environment can
choose which new state it returns in response to the action
(stochastic environment), in the
middle panel there is only one state for each action (deterministic
environment); the tree can then be simplified, showing only the
states, as in the right panel.

To calculate  the value of the
root of the tree a backup procedure can be followed. Such a
procedure calculates the value of a parent from the values of the
children, recursively, in a bottom-up
fashion, summing or maxing their values from the leaves to the root of
the tree. This calculation uses discrete 
time steps, indicated by subscripts to the state and action, as in
$s_t, s_{t+1}, s_{t+2},\ldots$. For brevity,
$s_{t+1}$ is sometimes written as $s'$. 
The figure shows a single transition step; an episode in reinforcement
learning typically consists of a sequence of many time steps.

\subsubsection*{Trial and Error, Down and Up}
A graph such as the one in  the center and right panel of
Fig.~\ref{fig:rltree}, where child nodes 
have only one parent node and without cycles, is  known as a
\emph{tree}. In computer science the root of a tree is at
the top, and  branches 
grow downward to the leaves.

As actions are performed and states and rewards are returned backup
the tree, a learning process
is taking place in the agent.
We can use Fig.~\ref{fig:rltree}  to better understand
the learning process that is unfolding.

The rewards of actions are learned by the agent by interacting with the
environment, performing the actions. In the tree  of
Fig.~\ref{fig:rltree} an action  selection moves \emph{downward}, towards the leaves. At the
deeper states, we find the rewards, which we  propagate to the parent
states \emph{upwards}. Reward learning is learning by backpropagation: in
Fig.~\ref{fig:rltree} the reward information flows upward in the
diagram from the leaves to the root. Action selection moves down,
reward learning flows up.\label{sec:backup}

Reinforcement learning is learning by trial and
error.
\emph{Trial} is selecting an action down (using the
behavior policy) to perform in the
environment. \emph{Error} is moving up the
tree, receiving a feedback reward from the environment, and reporting
that back up the tree to the state to update the current behavior
policy.\index{trial and error}
The downward selection policy chooses which actions
to explore, and the upward propagation of the error signal performs the
learning of the policy.\index{downward selection}\index{upward
  learning}\index{trial and error}

Figures such as the one in Fig.~\ref{fig:rltree} are useful for seeing
how values are calculated. The  basic notions are trial, and error,
or down, and up.

\subsubsection{Reward $R_a$}
The reward function \gls{reward}$_a$  is of central
importance in reinforcement learning. It indicates the measure of
quality of that state, 
such \emph{solved}, or \emph{distance}. Rewards are associated with
single states, indicating their quality. However, we are  most often interested
in the quality of a full decision making sequence from root to leaves
(this sequence of decisions would be one possible  answer to our
sequential decision problem). 

The
reward of such a full sequence is called the \emph{return}, sometimes
denoted confusingly as $R$, just as the reward. The
expected cumulative discounted future reward of a state is called the \emph{value}
function $V^\pi(s)$. The value function $V^\pi(s)$ is the
expected cumulative reward of $s$ where  actions are chosen
according to policy $\pi$. The value function plays a central role in
 reinforcement learning algorithms; in a few moments we will look deeper into
return and value.
\index{value function}

\subsubsection{Discount Factor $\gamma$}\label{sec:episodic}
\index{episodic task}\index{continuous task}

We distinguish between two types of tasks: (1) continuous time, long
running, tasks, and (2) episodic tasks---tasks that end.
In continuous and long running tasks it makes sense to discount
rewards from far in the future
in order to more strongly value current information at the present time.
To achieve this a discount factor \gls{gamma} is used in our MDP that reduces the
impact of far away
rewards. Many continuous tasks use discounting,  $\gamma \neq 1$.

However, in this book we will often discuss episodic problems, where $\gamma$
is irrelevant.
Both the supermarket example and the game of chess are episodic, and
discounting does not make sense in these problems, $\gamma = 1$.

\subsubsection{Policy   $\pi$} \label{sec_policy_definition}
\index{policy}\index{action}\label{sec:policy}\index{policy function}
Of central importance in reinforcement learning is the policy function
\gls{policy}. The policy function $\pi$ answers the question how the different
actions $a$ at state $s$ should be chosen.
Actions are anchored in states.
The central question of MDP optimization is  how to choose our
actions. The policy $\pi$
is a {\it conditional probability distribution} that for each
possible state specifies the probability of each possible
action. The function $\pi$ is a mapping from the state space to a probability
distribution over the action space: 
$$\pi: S \to p(A)$$
where $p(A)$ can be a discrete or continuous probability
distribution. For a particular probability (density) from this
distribution we write  
$$\pi(a|s)$$

\begin{tcolorbox}
{\bf Example}: For a discrete state space and discrete action space, we may store an explicit policy as a table, e.g.: 

\begin{center}
\begin{tabular}{l|cccc}
{$s$} & $\pi(a\text{=up}|s)$ & $\pi(a\text{=down}|s)$ & $\pi(a\text{=left}|s)$ & $\pi(a\text{=right}|s)$ \\
\hline
1 & 0.2 & 0.8 & 0.0 & 0.0 \\
2 & 0.0 & 0.0 & 0.0 & 1.0 \\
3 & 0.7 & 0.0 & 0.3 & 0.0 \\
etc. & . & . & . & .
\end{tabular}
\end{center}
\end{tcolorbox}
%
A special case of a policy is a {\it deterministic policy}, denoted by $$\pi(s)$$
where

$$\pi: S \to A $$

A deterministic policy selects  a single action in every state. Of
course the deterministic action may differ between states, as in the
example below: 

\vspace{0.3cm} 

\begin{tcolorbox}
{\bf Example}: An example of a deterministic discrete policy is

\begin{center}
\begin{tabular}{l|cccc}
{$s$} & $\pi(a\text{=up}|s)$ & $\pi(a\text{=down}|s)$ & $\pi(a\text{=left}|s)$ & $\pi(a\text{=right}|s)$ \\
\hline
1 & 0.0 & 1.0 & 0.0 & 0.0 \\
2 & 0.0 & 0.0 & 0.0 & 1.0 \\
3 & 1.0 & 0.0 & 0.0 & 0.0 \\
etc. & . & . & . & .
\end{tabular}

We would write $\pi(s=1) = \text{down}$, $\pi(s=2) = \text{right}$, etc. 
\end{center}
\end{tcolorbox}

\subsection{MDP Objective}
Finding the optimal policy function is the goal of the reinforcement
learning problem, and the remainder of this book will discuss many
different algorithms to achieve this goal under different
circumstances. Let us have a closer look at the objective of
reinforcement learning.
Before we can do so, we will look at traces, their return, and  value functions.

\subsubsection{Trace $\tau$}\index{trajectory}\index{trace}\index{episode}\index{sequence}\label{sec:trace}
As we  start interacting with the MDP, at each timestep $t$, we
observe $s_t$, take an action $a_t$  and then observe the next state
$s_{t+1} \sim T_{a_t}(s)$ and reward
$r_t=R_{a_t}(s_t,s_{t+1})$. Repeating this process leads to a sequence
or {\it
  trace}  in the environment, which we denote by $\tau_t^n$:  
$$ \tau_t^n =\{s_t,a_t,r_t,s_{t+1},..,a_{t+n},r_{t+n},s_{t+n+1}\}$$
Here, $n$ denotes the length of the \gls{trace}. In practice, we often
assume $n=\infty$, which means that we run the trace until the domain
terminates. In those cases, we will simply write $\tau_t =
\tau_t^\infty$. Traces are one of the basic building blocks of
reinforcement learning algorithms. They are a single full rollout of a sequence from
the sequential decision problem. They are also called trajectory,
episode, or simply sequence (Fig.~\ref{fig:trace} shows a single
transition step, and an example of a three-step trace).

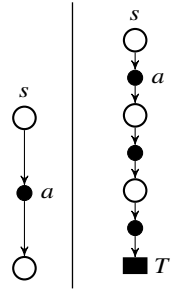
\begin{figure}[t]
\begin{center}
\tikzset{
  treenode/.style = {align=center, inner sep=0pt, text centered,
    font=\sffamily},
  arn_n/.style = {treenode, circle, white, draw=black, fill=black, text width=2mm},
  arn_r/.style = {treenode, circle, black, draw=black, text width=3mm, thick},
  arn_t/.style = {treenode, rectangle, fill, black, draw=black, text width=3mm, thick}
}

\begin{tabular}{c|c}
\begin{tikzpicture}[->,>=stealth',level/.style={sibling distance = 1.2cm/#1,
  level distance = 1cm}] 
\node [arn_r,label=above:{$s$}] {}
    child{ node [arn_n, label=right:{$a$}] {} 
            child{ node [arn_r] {} 
            }
    }
;  
\end{tikzpicture}
  
  \qquad
  
  &

    \qquad
    
\begin{tikzpicture}[->,>=stealth',level/.style={sibling distance = 1.2cm/#1,
  level distance = 5mm}] 
\node [arn_r,label=above:{$s$}] {}
    child{ node [arn_n, label=right:{$a$}] {} 
      child{ node [arn_r] {}
        child{node[arn_n] {}
          child{node[arn_r] {}
            child{node[arn_n] {}
              child{node[arn_t, label=right:{$T$}] {$T$}
              }
            }
          }
        }
      }
    }
;  
\end{tikzpicture}
\end{tabular}   
\caption{Single Transition Step versus Full 3-Step Trace/Episode/Trajectory}\label{fig:trace}
\end{center}
\end{figure}

\begin{tcolorbox}
{\bf Example}:
A short trace with three actions could look like: $$\tau_0^2=\{s_0\text{=1},a_0\text{=up},r_0\text{=$-1$},s_1\text{=2},a_1\text{=up},r_1\text{=$-1$},s_2\text{=3},a_2\text{=left},r_2\text{=20},s_3\text{=5}\}$$
\end{tcolorbox}
%
Since both the policy and the transition dynamics can be stochastic,
we will not always get the same trace from the start state. Instead,
we will get a {\it distribution} over traces. The distribution of
traces from the start state (distribution) is denoted by
$p(\tau_0)$. 
The probability of each possible trace from the start is
actually given by the product of the probability of each specific
transition in the trace: 

\begin{align} 
p(\tau_0) &= p_0(s_0) \cdot \pi(a_0|s_0) \cdot T_{a_0}(s_0,s_1) \cdot \pi(a_1|s_1) ... \nonumber \\
&=p_0(s_0) \cdot \prod_{t=0}^\infty \pi(a_{t}|s_{t}) \cdot T_{a_t}(s_t,s_{t+1}) \label{eq_h0} 
\end{align}

Policy-based
reinforcement learning depends heavily  on  traces, and we will
discuss traces more deeply in Chap.~\ref{chap:policy}. Value-based 
reinforcement learning (this chapter) uses  single transition steps.

\subsubsection*{Return $R$}
We have not yet formally defined what we actually want to achieve in the
sequential decision-making task---which is, informally, the best
policy.  The  sum of  the future reward of a trace is  known as the
\emph{return}. The return of trace $\tau_t$ is:
\begin{align} 
R(\tau_t) &= r_t + \gamma \cdot r_{t+1} + \gamma^2 \cdot r_{t+2} + ...  \nonumber \\
 &= r_t + \sum_{i=1}^\infty \gamma^i r_{t+i} \label{eq_infinite_return}
\end{align}
where $\gamma \in [0,1]$ is the discount factor. Two
extreme cases are:   
\begin{itemize}
\item $\gamma=0$: A myopic agent, which only considers the immediate reward, $R(\tau_t) = r_t$
\item $\gamma=1$: A far-sighted agent, which treats all future rewards as equal, $R(\tau_t) = r_t + r_{t+1} + r_{t+2} + \ldots$
\end{itemize}
Note that if we would use an {\it infinite-horizon} return
(Eq. \ref{eq_infinite_return}) and $\gamma=1.0$,  then the
cumulative reward may become unbounded. Therefore, in continuous problems, we  use a
discount factor close to 1.0, such as $\gamma=0.99$.

\begin{tcolorbox}
{\bf Example}: For the previous trace example we assume $\gamma =
0.9$. The return (cumulative reward) is equal to:

$$ R(\tau_0^2) = -1 + 0.9 \cdot -1 + 0.9^2 \cdot 20 = 16.2 - 1.9 = 14.3 $$
\end{tcolorbox}

\subsubsection{State Value  $V$}

The real measure of optimality that we are interested in is not the
return of just one trace. The environment can be stochastic, and so
can our policy, and for a given policy we do not always get the same
trace. Therefore, we are actually interested in the  
\emph{expected} cumulative reward that a certain policy achieves. The
expected cumulative discounted future reward of a state is better known as the {\em
  value} of that state.  

We define the  state value \gls{value}$^\pi(s)$ as the   
return we expect to achieve when an agent starts in state $s$ and then
follows policy $\pi$, as: 
\begin{equation} 
V^\pi(s) = \mathbb{E}_{\tau_t \sim p(\tau_t)}\big[ \sum_{i=0}^\infty \gamma^i \cdot r_{t+i} | s_t = s \big]
\end{equation} 

\begin{tcolorbox}
{\bf Example}: Imagine that we have a policy $\pi$, which from state $s$
can result in two traces. The first trace has a cumulative reward of
20, and occurs in 60\% of the times. The other trace has a cumulative
reward of 10, and occurs 40\% of the times.  What is the value of
  state $s$? 

$$ V^\pi(s) = 0.6 \cdot 20 + 0.4 \cdot 10 = 16. $$

The average return (cumulative reward) that we expect
to get from state $s$ under this policy is $16$.   
\end{tcolorbox}
Every policy $\pi$ has  one unique associated value function
$V^\pi(s)$. We often omit $\pi$ to simplify notation, simply
writing $V(s)$, knowing a state value is always conditioned on a
certain policy. 

The state value is defined for every possible state $s \in
S$. $V(s)$  maps every
state to a real number (the expected return): 

$$V: S \to \mathbb{R}$$ 

\begin{tcolorbox}
{\bf Example}:
In a discrete state space, the value function can be represented as a table of size $|S|$. 

\begin{center}
\begin{tabular}{l|l}
$s$ & $V^\pi(s)$ \\
\hline
1 & 2.0 \\
2 & 4.0 \\
3 & 1.0 \\
etc. & . \\
\end{tabular}
\end{center}

\end{tcolorbox}

Finally,  the state value of a terminal state is by definition zero: 
$$ s=\text{terminal} \quad \Rightarrow \quad V(s) := 0.$$

\subsubsection{State-Action Value $Q$}\label{sec:q}
In addition to  state values $V^\pi(s)$, we also define state-action
value \gls{state-action}$^\pi(s,a)$.\footnote{The reason for the choice for letter Q is lost in the
  mists of time. Perhaps it is meant to indicate quality.}   The only difference is that we now condition on a
state {\it and action}. We estimate the average return we expect
to achieve when taking action $a$ in state $s$, and  follow
policy $\pi$ afterwards:  
\begin{equation}
Q^\pi(s,a) = \mathbb{E}_{\tau_t \sim p(\tau_t)}\big[ \sum_{i=0}^\infty \gamma^i \cdot r_{t+i}|s_t=s,a_t=a \big]
\end{equation}
Every policy $\pi$ has only one unique associated state-action value
function $Q^\pi(s,a)$. We often omit $\pi$ to simplify
notation. Again, the state-action value is a  function  
$$Q: S \times A \to \mathbb{R} $$ 
which maps every state-action pair to a real number.

\begin{tcolorbox} 
{\bf Example}:
For a discrete state and action space, $Q(s,a)$ can be represented as
a table of size $|S| \times |A|$. Each table entry
stores a $Q(s,a)$ estimate for the specific $s,a$ combination: 

\begin{center}
\begin{tabular}{l|cccc}
 & $a$=up & $a$=down & $a$=left & $a$=right \\
\hline
$s$=1 & 4.0 & 3.0 & 7.0 & 1.0 \\
$s$=2 & 2.0 & -4.0 & 0.3 & 1.0 \\
$s$=3 & 3.5 & 0.8 & 3.6 & 6.2 \\
etc. & . & . & . & .
\end{tabular}
\end{center}

\end{tcolorbox}

The state-action value of a terminal state is by definition zero: 
$$ s=\text{terminal} \quad \Rightarrow \quad Q(s,a) := 0, \quad \forall a $$

\subsubsection{Reinforcement Learning Objective}
\index{objective}

We now have the ingredients to formally state the objective $J(\cdot)$  of reinforcement
learning. The objective is to achieve the highest possible average return from the start state:

\begin{equation} 
J(\pi) = V^{\pi}(s_0) = \mathbb{E}_{\tau_0 \sim p(\tau_0|\pi)} \Big[ R(\tau_0) \Big].\label{eq_rl_objective}
\end{equation}
for $p(\tau_0)$ given in Eq.~\ref{eq_h0}. There is
one optimal value function, which achieves higher or equal value
than all other value functions. We search for a policy that achieves
this optimal value function, which we call the optimal policy
$\pi^\star$: 

\begin{equation} 
\pi^\star(a|s) = \argmax_\pi V^\pi(s_0)
\end{equation}
This function $\pi^\star$ is the optimal policy, it uses the $\argmax$
function to select the policy with the optimal value.
The goal in reinforcement learning is to find  this optimal policy
for start
state $s_0$.

A potential benefit of state-action values $Q$ over state values
$V$ is that state-action values directly tell what every action is
worth. This   may be useful for action selection, since, for
discrete action spaces,
$$a^\star = \argmax_{a\in A} Q^\star(s,a)$$ the Q function directly identifies
the best action. Equivalently, the optimal policy can be obtained
directly from the optimal Q function: $$\pi^\star(s)=\argmax_{a\in A}Q^\star(s,a).$$  

We will now turn to construct
algorithms to compute the value function and the policy function.

\subsubsection{Bellman Equation}\index{Bellman equation}

To calculate the value function,
let us look  again at the tree in Fig.~\ref{fig:rltree} on
page~\pageref{fig:rltree}, and imagine that  it is many
times larger, with subtrees that extend to fully cover the
state space. Our task is to
compute the value of the root, based on the reward values at the real
leaves, using the transition function $T_a$. One way
to calculate the value $V(s)$ is to traverse this full state space tree,  computing the
value of a parent node by taking the reward value and the sum of the
children, discounting this value  by  $\gamma$.

\index{Bellman, Richard}\index{dynamic programming}\index{Bellman equation}
This intuitive approach was first formalized by Richard Bellman in
1957. Bellman showed that
discrete optimization problems can be described as a recursive backward
induction problem~\cite{bellman1957dynamic}. He  introduced the term 
dynamic programming  to recursively traverse the states
and actions. The so-called \emph{Bellman equation} shows the
relationship between the 
value function in  state $s$ and the future child state $s'$, when we
follow the transition function.

The discrete Bellman equation of the  value of   state $s$ after
following  policy $\pi$ is:\footnote{State-action value and continuous Bellman equations can be
  found in  Appendix~\ref{sec:bellman}.}
\begin{equation}
  V^\pi(s) = \sum_{a \in A} \pi(a|s) \Big[
  \sum_{s'\in S} T_a(s,s') \big[ R_a(s,s') + \gamma \cdot
  V^\pi(s') \big] \Big] \label{eq:bellman}
\end{equation}
where $\pi$ is the probability of action $a$ in state $s$, $T$ is the stochastic transition function, $R$ is the reward function and
$\gamma$ is the discount rate.
Note the recursion on the value function, and that for the Bellman
equation the transition and reward functions  must be
known for all states by the agent.

Together, the transition and reward model are referred to as the
\emph{dynamics model} of the environment.\index{dynamics model} The
dynamics model is often not known by the agent, and model-free methods have been developed to
compute the  value function and policy function without them.

The recursive Bellman equation is the basis of  algorithms to
compute the value function, and other relevant functions to solve
reinforcement learning problems. In the next section we will study
these solution methods.

\subsection{MDP Solution Methods}
\index{planning}\index{dynamic programming}
\begin{figure}[t]
\begin{center}
\includegraphics[width=6cm]{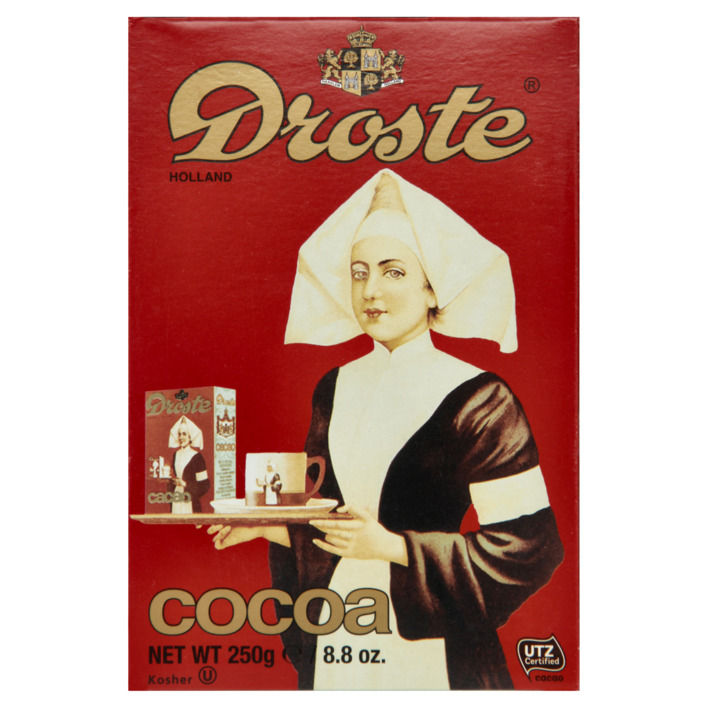} 
\caption{Recursion: Droste effect}\label{fig:dragon}
\end{center}
\end{figure}
The Bellman equation is a recursive equation: it shows how to calculate the value
of a state, out of the  values of applying the function
specification again on the successor states. Figure~\ref{fig:dragon} shows a
recursive picture, of a picture in a 
picture, in a picture,  etc. In algorithmic form, dynamic programming
calls its own 
code on states that are closer and closer to the leaves, until the
leaves are reached, and the recursion can not go further.

Dynamic programming  uses the principle of divide and conquer: it
begins with a  start state  whose value is to be determined by
searching a large subtree,
which it does by going down into the recursion,
finding the value of sub-states that are closer to terminals. At
terminals the reward values are known, and these are then used in the
construction of the parent values, as it goes up, back out of the
recursion, 
and ultimately arrives  at the root value itself.
\index{recursion}\index{divide and conquer}

A simple dynamic programming method to iteratively
traverse the state space to calculate Bellman's equation is
\emph{value iteration} (VI).
\index{value iteration}\label{sec:vi}
Pseudocode for a basic
version of \gls{value iteration} is 
shown in Listing~\ref{lst:val-it}, based
on~\cite{alpaydin2009introduction}. 
Value iteration  converges to the optimal  value function by
iteratively improving the estimate of $V(s)$. The value function $V(s)$ is first
initialized to random values. Value iteration repeatedly updates $Q(s,a)$ and 
$V(s)$ values, looping over the states and their actions,  until convergence occurs (when the values of $V(s)$ stop
changing much).

\lstset{label={lst:val-it}}
\lstset{caption={Value Iteration pseudocode}}
\begin{lstlisting}[language=Python,float]
def value_iteration():
    initialize(V) 
    while not convergence(V):
        for s in range(S):
            for a in range(A):
                Q[s,a] = (*$\sum_{s'\in S} T_a(s,s')(R_a(s,s') + \gamma V[s'])$*)
            V[s] = max_a(Q[s,a])
    return V

\end{lstlisting}

Value iteration  
works with a finite set of actions.   It has been proven  to
converge to the optimal values, but, as we can see in the pseudocode
in Listing~\ref{lst:val-it},
it does so quite inefficiently by essentially repeatedly enumerating the entire
state space in a triply nested loop, traversing the state space many
times. Soon we will see more efficient methods.

\subsubsection{\em Hands On: Value Iteration in Gym}\label{sec:valit-taxi}
We have discussed in detail how to model a reinforcement learning
problem with an MDP. We have talked in depth and at length about
states, actions, and policies. It is now time for some hands-on work, to
experiment with the theoretical concepts. 
We will start
with the environment.

\subsubsection*{OpenAI Gym}\index{Gym}\index{OpenAI Gym}\label{sec:gym}
OpenAI has created the Gym suite of \emph{environments} for Python, which
has become the de facto standard in the field~\cite{brockman2016openai}.  The Gym suite
can be found at OpenAI\footnote{\url{https://gym.openai.com}} and on
GitHub.\footnote{\url{https://github.com/openai/gym}} Gym works on
Linux, macOS and Windows. An active
community exists and new environments are  created continuously and
uploaded to the Gym website. Many interesting
environments are available for experimentation, to create your own
agent algorithm for, and test it.

If you browse Gym on GitHub, you will see different sets of environments, from easy to
advanced. There are the classics, such as Cartpole and Mountain
car. There are also small text environments. Taxi is there, and 
the Arcade Learning Environment~\cite{bellemare2013arcade}, which was
used in the paper that introduced DQN~\cite{mnih2013playing}, as we
will discuss at length in the next chapter. 
MuJoCo\footnote{\url{http://www.mujoco.org}} is also available, an
environment for experimentation with simulated
robotics~\cite{todorov2012mujoco}, or you can use pybullet.\footnote{\url{https://pybullet.org/wordpress/}}

You should now install Gym. 
Go to the Gym page on
\url{https://gym.openai.com} and read the documentation. Make sure
Python is installed on your system (does typing python at the command
prompt work?), and that your Python version is up to
date (version 3.10 at the time of this writing). Then type
\begin{tcolorbox}
  \verb|pip install gym|
\end{tcolorbox}
\noindent to install Gym with the Python package
manager. Soon, you will also be needing deep learning suites, such as
TensorFlow or PyTorch. It is recommended to install Gym in the same virtual
environment as your upcoming PyTorch and TensorFlow installation, so that you can use both at
the same time (see Sect.~\ref{sec:keras}).
You may have to install or update other packages, such as numpy, scipy and
pyglet, to get Gym to work, depending on your system installation.

You can check if the installation works by trying if the CartPole
environment works, see Listing~\ref{lst:gym}. A window should appear
on your screen in which a Cartpole is making random movements (your
window system should support OpenGL, and you may need a version of
pyglet newer than version 1.5.11 on some operating systems).

\lstset{label={lst:gym}}
\lstset{caption={Running the Gym CartPole Environment from Gym}}
\begin{lstlisting}[language=Python,float]
import gym

env = gym.make('CartPole-v0')
env.reset()
for _ in range(1000):
    env.render()
    env.step(env.action_space.sample()) # take a random action
env.close()
\end{lstlisting}

\lstset{label={lst:evaltaxivi}}
\lstset{caption={Value Iteration for Gym Taxi}}
\begin{lstlisting}[language=Python,float]
import gym
import numpy as np

def iterate_value_function(v_inp, gamma, env):
    ret = np.zeros(env.nS)
    for sid in range(env.nS):
        temp_v = np.zeros(env.nA)
        for action in range(env.nA):
            for (prob, dst_state, reward, is_final) in env.P[sid][action]:
                temp_v[action] += prob*(reward + gamma*v_inp[dst_state]*(not is_final))
        ret[sid] = max(temp_v)
    return ret

def build_greedy_policy(v_inp, gamma, env):
    new_policy = np.zeros(env.nS)
    for state_id in range(env.nS):
        profits = np.zeros(env.nA)
        for action in range(env.nA):
            for (prob, dst_state, reward, is_final) in env.P[state_id][action]:
                profits[action] += prob*(reward + gamma*v[dst_state])
        new_policy[state_id] = np.argmax(profits)
    return new_policy


env = gym.make('Taxi-v3')
gamma = 0.9
cum_reward = 0
n_rounds = 500
env.reset()
for t_rounds in range(n_rounds):
    # init env and value function
    observation = env.reset()
    v = np.zeros(env.nS)

    # solve MDP
    for _ in range(100):
        v_old = v.copy()
        v = iterate_value_function(v, gamma, env)
        if np.all(v == v_old):
            break
    policy = build_greedy_policy(v, gamma, env).astype(np.int)

    # apply policy
    for t in range(1000):
        action = policy[observation]
        observation, reward, done, info = env.step(action)
        cum_reward += reward
        if done:
            break
    if t_rounds % 50 == 0 and t_rounds > 0:
        print(cum_reward * 1.0 / (t_rounds + 1))
env.close()
\end{lstlisting}

\subsubsection*{Taxi Example with Value Iteration}\label{sec:taxi-vi}

The Taxi example 
(Fig.~\ref{fig:taxi}) is 
an environment where taxis move up, down, left, and right, and 
pickup and drop off passengers. Let us see how we can use value
iteration to solve the Taxi problem.
\begin{figure}[t]
\begin{center}
\includegraphics[width=5cm]{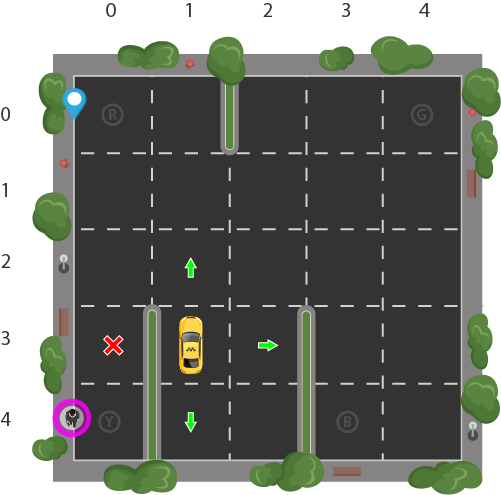}
\caption{Taxi world~\cite{learn}}\label{fig:taxi}
\end{center}
\end{figure}
The Gym documentation describes the Taxi world as follows.
There are four designated locations in the Grid world indicated by 
R(ed), B(lue), G(reen), and Y(ellow). When the episode starts, the
taxi starts off at a random square and the passenger is at a 
random location. The taxi drives to the passenger's location, picks
up the passenger, drives to the passenger's destination (another
one of the four specified locations), and then drops off the
passenger. Once the passenger is dropped off, the episode ends.

The Taxi problem has  500 discrete states: there are 25 taxi positions, five
possible locations of the passenger (including the case when the
passenger is in the taxi), and 4 destination locations $(25\times
5\times 4)$.  

The environment returns  a new result tuple at each step.  There are six
discrete deterministic actions for 
the Taxi driver:
\begin{itemize}
\item[0:] Move south
\item[1:] Move north
\item[2:] Move east 
\item[3:] Move west 
\item[4:] Pick up passenger
\item[5:] Drop off passenger
\end{itemize}    

There is a reward of $-1$ for each action and an additional reward of
$+20$ for delivering the passenger, and a reward of $-10$ for
executing actions {\em pickup\/} and {\em dropoff\/} illegally.

The Taxi environment has a simple transition
function, which is used by the agent in the value iteration
code.\footnote{Note that the code uses the environment to compute the
  next state, so that we do not have to implement a version of the
  transition function for the
  agent.}
Listing~\ref{lst:evaltaxivi} shows an implementation of  
value iteration that uses the Taxi environment to find a
solution. This code is written by Mikhail Trofimov, and
illustrates clearly how  value iteration first creates the value
function for the states, and then that a policy is formed by finding
the best action in each state, in the build-greedy-policy function.\footnote{\url{https://gist.github.com/geffy/b2d16d01cbca1ae9e13f11f678fa96fd\#file-taxi-vi-py}}

To get a feeling for how the algorithms work, please use the value iteration code with the Gym Taxi environment,  see to
Listing~\ref{lst:evaltaxivi}. Run the code, and play around with some
of the hyperparameters to familiarize yourself a bit with Gym and with
planning by value iteration. Try to visualize for yourself what the
algorithm is doing. This will prepare you for the more
complex algorithms that we will look into next.


\subsubsection{Model-Free Learning}
\label{sec:tab}\index{tabular method}
The value iteration algorithm can compute the policy function. It
uses   the transition model in its computation. Frequently, we are in
a situation when the  transition probabilities are not known to
the agent, and we need other methods to compute  the policy function.
For this situation,  model-free algorithms have
been developed.

The development of these model-free methods is a major milestone of
reinforcement learning, and we will spend some time to understand how
they work. We will start with value-based model-free algorithms. 
%
%
%
\begin{table}[t]
  \begin{center}
  \begin{tabular}{lll}
    {\bf Name} & {\bf Approach}  &  {\bf Ref}  \\
    \hline\hline
    Value Iteration & Model-based enumeration& \cite{bellman1957dynamic,alpaydin2009introduction}\\
    SARSA & On-policy temporal difference model-free &  \cite{rummery1994line}\\
    Q-learning & Off-policy temporal difference model-free  & \cite{watkins1989learning}\\
    \hline
  \end{tabular}
  \caption{Tabular Value-Based Approaches}\label{tab:val}
\end{center}
\end{table}
%
We will see how, when the agent  does not
 know the transition function, an optimal policy can be learned by sampling rewards
 from the environment.
Table~\ref{tab:val} lists value iteration in
conjunction with the value-based model-free algorithms
that we cover in this chapter. (Policy-based model-free algorithms
will be
covered in Chap.~\ref{ch:pol}.)

These algorithms are based on a few
principles. 
First we will discuss how the principle of \emph{sampling} can be used
to construct a value function. We discuss both
full-episode Monte Carlo sampling and single-step temporal difference
learning;  we encounter the principle of \gls{bootstrapping} and
the \emph{bias-variance} trade-off; and
we will see how the
value function can be use to find the best actions, to form the
policy.

Second, we will discuss which mechanisms for action selection
exist, where  we will encounter the \emph{exploration/exploitation} 
trade-off. Third, we will discuss how to learn from the rewards of the
selected actions. We will encounter \emph{on-policy} learning and
\emph{off-policy} 
learning. 
%
Finally, we wil discuss two  full algorithms in which all these
concepts come together: SARSA and Q-learning.
Let us now start by having a closer look at sampling actions with
Monte Carlo sampling and temporal difference learning.

\subsubsection*{Monte Carlo Sampling}\index{Monte Carlo Sampling}
A straightforward way to sample rewards  is to generate a random
episode, and use its return to update the value function at the visited
states. This approach consists of two loops: a simple loop over the time steps of
the episode, embedded in a loop to sample long enough for the value
function to convergence. This approach, of randomly sampling  full
episodes, has become known as the Monte Carlo approach (after the
famous casino, because of the
random action selection).

\lstset{label={lst:mc}}
\lstset{caption={Monte Carlo Sampling code}}
\begin{lstlisting}[language=Python,float]
def monte_carlo(n_samples, ep_length, alpha, gamma):
    # 0: initialize
    t = 0; total_t = 0
    Qsa = []

    # sample n_times 
    while total_t < n_samples:

        # 1: generate a full episode
        s = env.reset()
        s_ep = []
        a_ep = []
        r_ep = []
        for t in range(ep_length):
            a = select_action(s, Qsa)
            s_next, r, done = env.step(a)
            s_ep.append(s)
            a_ep.append(a)
            r_ep.append(r)

            total_t += 1
            if done or total_t >= n_times:
                break;
            s = s_next

        # 2: update Q function with a full episode (incremental
        # implementation)
        g = 0.0
        for t in reversed(range(len(a_ep))):
            s = s_ep[t]; a = a_ep[t]
            g = r_ep[t] + gamma * g
            Qsa[s,a] = Qsa[s,a] + alpha * (g - Qsa[s,a])
     
    return Qsa

def select_action(s, Qsa):
        
    # policy is egreedy
    epsilon = 0.1
    if np.random.rand() < epsilon:
        a = np.random.randint(low=0,high=env.n_actions)
    else:
        a = argmax(Qsa[s])
    return a
        
env = gym.make('Taxi-v3')
monte_carlo(n_samples=10000, ep_length=100, alpha=0.1, gamma=0.99)
\end{lstlisting}

Listing~\ref{lst:mc} shows code for the Monte Carlo
approach.
We see three elements in the code. First, the main variables are
initialized. Then a loop for the desired number of total samples
performs the unrolling of the episodes.
For each episode the state, action and reward lists are initialized,
and then filled with the samples from the environment until we hit the
terminal state of the episode.\footnote{With epsilon-greedy action
selection, see next subsection.}
Then, at the end of the episode, the return is calcuated in variable
$g$ (the return of a state is the sum of its discounted future rewards). The
learning rate $\alpha$ is then used  to update the $Q$ function in an
incremental implementation.\footnote{The incremental implementation works
for nonstationary situations, where the transition probabilities may
change, hence the previous $Q$ values are subtracted.} The main
purpose of the code is to illustate full episode learning. Since it is
a complete working algorithms, the code also
uses on-policy learning with $\epsilon$-greedy 
selection, topics that we will discuss in the next subsection.

The Monte Carlo approach is a basic building block of value based
reinforcement learning. An advantage of the approach is its simplicity. A
disadvantage is that a full episode has to be sampled before the reward
values are used, and sample efficiency may be low. For this reason (and others, as we will soon see)
another approach was developed, inspired by the way the Bellman
equation  bootstraps on intermediate values.

\subsubsection*{Temporal Difference Learning}\label{sec:td}
\index{temporal difference}\index{TD}\index{bootstrapping}

Recall that in value iteration  the value function was calculated 
recursively  using the values of successor states,
following Bellman's equation (Eq.~\ref{eq:bellman}). 


\emph{Bootstrapping} is the process of subsequent refinement by which old estimates
of a value are refined with new updates. It means literally: pull
yourself up (out of the swamp) by your boot straps. 
Bootstrapping solves the problem of computing a final value when we 
only know how to compute step-by-step intermediate values.
Bellman's recursive computation is a form of \gls{bootstrapping}. 
In model-free learning, we can use a similar approach, when the role of the transition function is
replaced by a sequence of environment samples. 

A bootstrapping method that can be used to process the samples, and
to refine them to approximate the final state values, is \emph{temporal
difference learning}. Temporal difference learning, \gls{TD} for short, was
introduced by Sutton~\cite{sutton1988learning} in 1988.
The temporal difference in the
name refers to the difference in values between two time
steps, which are used to calculate the  value at the new time step.

Temporal difference learning  works by updating the  current estimate
of the state value $V(s)$ (the bootstrap-value) with an 
error value (new minus current) based on the estimate of the next state that it has gotten
through sampling the environment: 
\begin{equation}
V(s)\leftarrow V(s)+\alpha[r^\prime+\gamma
V(s^\prime)-V(s)] \label{eq:td}
\end{equation}
Here $s$ is the current state, $s^\prime$ the new state, and
$r^\prime$ the reward of the new state. Note the introduction of
\gls{alpha}, the learning rate, which controls how fast the algorithm
learns (bootstraps). It is an important parameter; setting the value
too high can be detrimental since the last value then dominates the 
bootstrap process too much. Finding the optimal value will require
experimentation.    The $\gamma$ 
parameter is the discount rate. The last term $-V(s)$
subtracts the value of the current state, to compute the
temporal difference. Another way to write this update rule is
$$V(s)\leftarrow \alpha[r^\prime+\gamma V(s^\prime)] + (1-\alpha)V(s)$$
as
the difference between the new temporal difference target and the old value. Note the
absence of transition model $T$ in the formula; temporal difference is
a model-free update formula. 
Listing~\ref{lst:td} shows code for the TD
approach, for the state-action value function. (This code is
off-policy, and  uses the same $\epsilon$-greedy selection function as
Monte Carlo sampling.)

\lstset{label={lst:td}}
\lstset{caption={Temporal Difference Q-learning code}}
\begin{lstlisting}[language=Python,float]
def temporal_difference(n_samples, alpha, gamma):
    # 0: initialize
    Qsa = []
    s = env.reset()

    for t in range(n_samples):
        a = select_action(s, Qsa)
        s_next, r, done = env.step(a)

        # update Q function each time step with max of action values
        Qsa[s,a] = Qsa[s,a] + alpha * (r + gamma * np.max(Qsa[s_next]) - Qsa[s,a])
     
        if done:
            s = env.reset()
        else:
            s = s_next

    return Qsa

def select_action(s, Qsa):
        
    # policy is egreedy
    epsilon = 0.1
    if np.random.rand() < epsilon:
        a = np.random.randint(low=0,high=env.n_actions)
    else:
        a = argmax(Qsa[s])
    return a
        
env = gym.make('Taxi-v3')
temporal_difference(n_samples=10000, alpha=0.1, gamma=0.99)
\end{lstlisting}


The introduction of the temporal difference method has allowed
model-free methods to be used successfully in various  reinforcement
learning settings. Most notably, it was the basis of the program
TD-Gammon, that beat human world-champions in the game of Backgammon
in the early 1990s~\cite{tesauro1995td}.\index{backgammon}

\subsubsection*{Bias-Variance Trade-off}\index{bias-variance trade-off}\index{n-step}
A crucial difference between the  Monte Carlo method and
the 
temporal difference method is the use of bootstrapping to calculate
the value function. The use of bootstrapping has an important
consequence: it trades off bias and variance (see Fig.~\ref{fig:biasvariance2}). Monte Carlo does not use
bootstrapping. It performs a full episode with many random action choices
before it uses the reward. As such, its action choices are unbiased
(they are fully random), they are not influenced by previous reward
values. However, the fully random choices also cause a
high variance of returns between episodes. We say that Monte Carlo is a
low-bias/high-variance algorithm.

\begin{figure}[t]
\begin{center}
    \includegraphics[width=8cm]{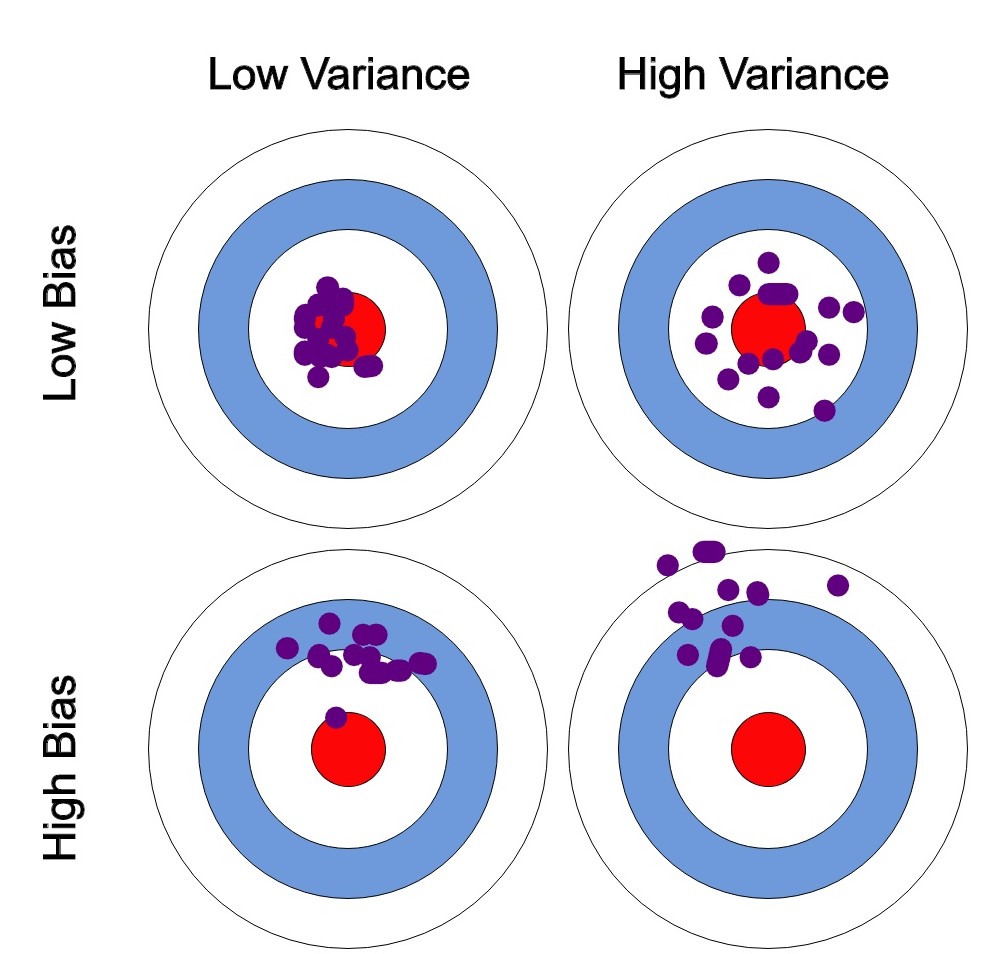}
\caption{High and Low Bias, and High and Low Variance}\label{fig:biasvariance2}
\end{center}
\end{figure}

In contrast, temporal difference bootstraps the $Q$-function with the
values of the previous steps, refining the function values with the
rewards after each single step. It learns more quickly, at each step, 
but once a step has been taken, old reward values linger around in the
bootstrapped function value for a long time, biasing the function
value. On the other hand, because these old values are part of the new
bootstrapped value, the variance  is lower. Thus, because
of bootstrapping, TD is a high-bias/low variance
method. Figure~\ref{fig:biasvariance2} illustrates the concepts of
bias and variance with  pictures of  dart
boards.

Both approaches have their uses in different circumstances. In fact,
we can think of situations where a middle ground (of medium
bias/medium variance) might be useful. This is the idea behind the
so-called \emph{n-step} approach: do not sample a full episode, and also not a single
step, but sample a few steps at a time before using the reward
values. The n-step algorithm has medium
bias and medium variance. Figure~\ref{fig:nstep} from~\cite{sutton2018introduction}  illustrates the
relation between Monte Carlo sampling, n-step, and temporal difference learning.

\begin{figure}[t]
\begin{center}
    \includegraphics[width=10cm]{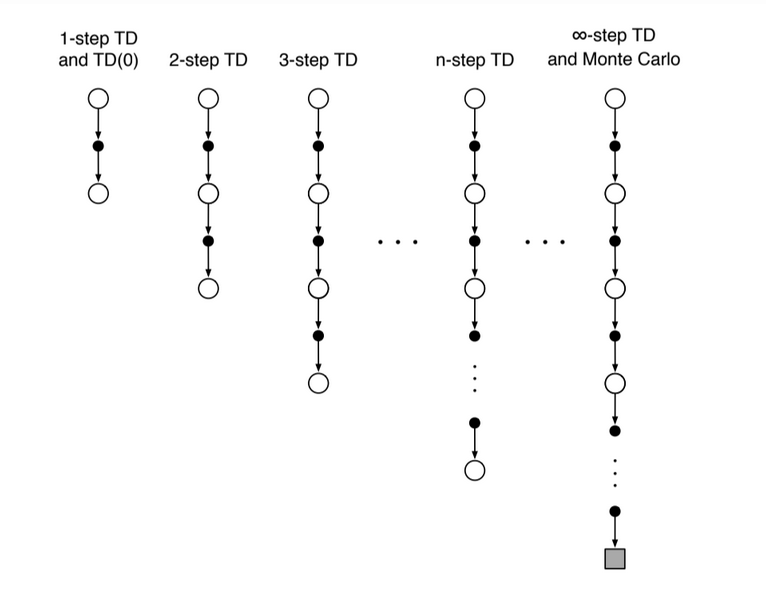}
\caption{Single Step  Temporal Difference Learning, N-Step, and Monte Carlo Sampling \cite{sutton2018introduction}}\label{fig:nstep}
\end{center}
\end{figure}

\subsubsection*{Find Policy by Value-based Learning}

The goal of reinforcement learning is to construct the policy with the
highest cumulative reward. Thus, we must find the best action $a$ in
each state $s$. In the value-based approach we know the value functions $V(s)$ or $Q(s,a)$.
How can that help us to find action $a$?
In a discrete action space,  there is at least one discrete action with the
highest value. 
Thus, if we have the optimal state-value  $V^\star$, then the
optimal policy can be found by finding the action with that value. This relationship is given
by $$\pi^\star= \max\limits_{\pi} V^{\pi}(s) = \max\limits_{a,\pi}
Q^{\pi}(s,a)$$ and the $\argmax$ function finds the best action for
us $$a^\star = \argmax_{a\in A} Q^\star(s,a) .$$
In this way the optimal policy sequence of best actions $\pi^\star(s)$ can
be recovered from the values, hence the name \emph{value-based method}~\cite{wong2021survey}.

A full reinforcement learning algorithm consists of a rule for the
selection part
(downward) and a rule for the learning part (upward).
Now that we know how to calculate the value function (the up-motion in
the tree diagram), let us see how
we can select the action in our model-free algorithm (the down-motion
in the tree diagram).

\subsubsection{Exploration}
\index{exploration}\index{exploitation}\label{sec:epsilon}\label{sec:multiarmed}\index{variance}\index{confidence}\index{behavior policy}

Since there is no local transition function,   model-free methods 
perform their state changes directly in
the  environment.  This may be an expensive operation, for example,
when a real-world robot arm has to perform a movement.  
The sampling policy should choose promising actions to reduce the number of
samples as much as possible, and not waste any actions. What behavior policy should we use? It is
tempting to favor at each state the actions with the highest 
Q-value, since then we would be following what is currently thought to
be the best policy.

This approach is called the
\emph{greedy} approach. It appears attractive, but is short-sighted
and risks settling for 
local maxima. Following the  trodden path based on only a few early
samples risks missing a
potential better path. Indeed, the greedy approach  is high bias,
using values based on few samples. We run
the risk of circular reinforcement, if  we update the same behavior policy that
we use to choose our samples from. 
In addition to exploiting known good actions, a certain amount 
of \gls{exploration} of unknown actions is necessary. Smart sampling
strategies use a mix of the current behavior policy (\emph{exploitation}) and
randomness (\emph{exploration}) to select which action to perform in
the environment.

\subsubsection*{Bandit Theory}\index{multi-armed bandit}
The exploration/exploitation trade-off, the question of how to get the
most reliable information at the least cost, has been studied extensively in
the literature for single step decision problems~\cite{holland1975adaptation,witten1976apparent}.
The field has the colorful name of multi-armed bandit
theory~\cite{auer2002using,lai1985asymptotically,gittins1979bandit,robbins1952some}.\index{bandit theory}
A \emph{bandit} in this context refers to a casino slot machine, with not
one arm, but many arms, each with a different and unknown  payout
probability. Each trial costs a coin. The
multi-armed bandit problem is then to find a strategy that finds the
arm with the highest payout at the least cost.

A multi-armed bandit is a single-state single-decision
reinforcement learning problem, a one-step non-sequential decision making
problem, with the arms  representing the possible actions. This
simplified model of stochastic decision making allows the in-depth 
study of  exploration/exploitation strategies.

Single-step exploration/exploitation questions arise for example in
clinical trials, where new drugs are tested on test-subjects (real
people). The bandit is the trial, and the arms are the choice how many
of the test subjects are given the real experimental drug, and how
many are given the placebo. This is a  serious setting, since the
cost may be measured in the quality of human lives.

In a conventional fixed randomized controlled trial (supervised setup)
 the sizes of the groups that get the experimental drugs and the
control group would be fixed, and the confidence interval and the duration of the
test would also be fixed. 
In an adaptive trial (bandit setup)
the sizes would adapt during the trial depending on the 
outcomes, with more people getting the 
drug if it appears to work, and fewer if it does not.

\begin{figure}[t]
\begin{center}
    \includegraphics[width=8cm]{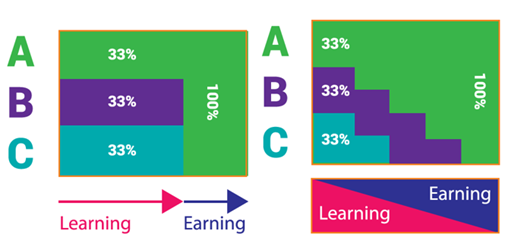}
\caption{Adaptive Trial \cite{brillio2019}}\label{fig:earning}
\end{center}
\end{figure}
Let us have a look at Fig.~\ref{fig:earning}. Assume that the learning process is a
clinical trial in which three new compounds are tested for their medical
effect on test subjects. In the fixed trial (left panel) all test subjects
receive the medicine of their group to the end of the test period, after which the data
set is complete and we can determine which of the compounds has the
best effect. At that point we know which group has had the best
medicine, and which two thirds of the subjects did not, with possibly
harmful effect. Clearly, this is not a satisfactory situation. It
would be  better if we could gradually adjust the proportion of the
subjects that receive the medicine that currently looks best, as our
confidence in our test results increases as the trial progresses. Indeed, this is what
reinforcement learning does (Fig.~\ref{fig:earning}, right panel). It
uses a mix of exploration and \gls{exploitation}, adapting the
treatment, giving more subjects the promising medicine, while
achieving the same confidence as the static
trial at the end~\cite{lai1985asymptotically,lai1987adaptive}.

\subsubsection*{$\epsilon$-greedy Exploration}
A popular pragmatic exploration/exploitation approach is to use a fixed ratio of exploration
versus exploitation. 
This approach is known as the \gls{greedy} approach, which  is to mostly try  the (greedy) action that
currently has the highest policy value  
except to explore an $\epsilon$ fraction of times a randomly selected
other action. If $\epsilon = 0.1$ then 90\%
of the times the currently-best action is taken, and 10\% of the times a random
other action.  

The algorithmic choice between greedily exploiting known information and
exploring unknown actions
to gain new information is called the exploration/exploitation
trade-off. It is a central concept in reinforcement
learning; it determines how much confidence we have in
our outcome, and how quickly the confidence can be increased and the
variance reduced.
A second approach is to use an adapative $\epsilon$-ratio, that changes
over time,  or over other statistics of the learning process.

Other popular  approaches to add exploration are to add Dirichlet-noise~\cite{kotz2004continuous} or
to use Thompson sampling~\cite{thompson1933likelihood,russo2017tutorial}.

\subsubsection{Off-Policy Learning}
\index{off-policy learning}\index{on-policy
  learning}\label{sec:off-policy}\index{bootstrapping}\index{GLIE convergence}

In addition to the selection question, another main
theme in the design of full reinforcement learning algorithms is which
learning method to use. 
Reinforcement learning is concerned with learning an action-policy from
the rewards. The
agent selects an action to perform, and learns from the reward that it gets back
from the environment.
The question is whether  the agent should perform  updates strictly on-policy---only learning
from its most recent action---or allow
off-policy updates,
learning from all available information.

In \emph{on-policy} learning, the learning  takes place by  using
the value of the action that was selected  by the  
policy. The policy determines the action to
take, and the value of that action is used to update the value of the policy
function: the
learning is {\em on-policy}.

There is, however, an alternative to this
straightforward method. In {\emph{off-policy} methods, the learning takes
place by backing up values of \emph{another} action, not necessarily the one selected by
the behavior policy. This method makes
sense when the agent explores. When the behavior policy explores, it selects a \emph{non-optimal} 
action. The policy does not perform the greedy exploitation action; of
course, this usually results in an inferior reward value. On-policy
learning would then blindly  backup the value of the 
non-optimal exploration action. Off-policy learning,
however, is free to backup another value instead. It makes sense to choose the value of the best action, and
not the inferior one
selected by the exploration policy. The advantage of this off-policy
approach is that it does not  pollute the behavior policy  with a
value that is most likely  inferior.

The difference between on-policy and off-policy is only in how they
act when exploring the non-greedy action. In the case of exploration,
off-policy learning can be more 
efficient, by not stubbornly backing up the value of the action
selected by the behavior policy, but the value of an older, better,
action.

An important point is that the convergence behavior of on-policy and off-policy learning is
different.  In general,  tabular
reinforcement learning have been proven to converge when the policy is  greedy in the limit with
infinite exploration
(GLIE)~\cite{sutton2018introduction}. This means that (1) if a state is
visited infinitely often, that each action is also chosen infinitely
often, and that (2) in the limit the policy is greedy with respect to
the learned $Q$ function. 
Off-policy
methods learn from the greedy rewards and thus converge to the optimal policy, after having sampled enough
states. However,  on-policy
methods with a fixed $\epsilon$ do not converge to the optimal policy, since they keep
selecting explorative actions. When we use a
variable-$\epsilon$-policy in which the value of $\epsilon$ goes to
zero, then on-policy methods do converge, since then they choose, in the limit, the greedy
action.\footnote{However, in the next chapter, deep learning is
  introduced, and a complication arises. In deep learning states and values  are no
  longer exact but are approximated. Now, off-policy methods become less stable than
  on-policy methods. In a neural network, states are ``connected'' via joint
  features. The max-operator in off-policy methods pushes up training
  targets of these connected states. As a consequence deep off-policy
  methods may not converge.  A so-called \emph{deadly triad} of function approximation, 
  bootstrapping and off-policy learning occurs that causes unstable
  convergence. Because of this, with function approximation, on-policy
  methods are sometimes favored.}


A well-known tabular on-policy algorithm is  SARSA.\footnote{The name of the SARSA algorithm is a play on the MDP symbols as
  they occur in the action value update formula: $s, a, r, s, a$.}  An
even more well-known  
off-policy algorithm is Q-learning.


\subsubsection*{On-Policy SARSA}\index{SARSA}\label{sec:sarsa}
\gls{SARSA} is  an on-policy algorithm~\cite{rummery1994line}.
On-policy learning updates the policy  with the action values of the  
policy.
%
The SARSA update formula is
\begin{equation}
Q(s_t,a_t)\leftarrow Q(s_t,a_t)+\alpha[r_{t+1}+\gamma
Q(s_{t+1},a_{t+1})-Q(s_t,a_t)]. \label{eq:sarsa}
\end{equation}
Going back to temporal difference (Eq.~\ref{eq:td}), we see that the
SARSA formula looks very much like TD, 
although now we deal with  state-action values.

On-policy learning selects an action, evaluates it in the environment,
and follows the  actions, guided by the behavior policy. The behavior
policy is not specified
by the formula, but might be $\epsilon$-greedy, or an other policy
that trades off exploration and exploitation.  On-policy
learning 
samples the state space following the behavior policy, and improves the
policy by backing up values of the selected actions. Note that the
term $Q(s_{t+1}, a_{t+1})$ can also be written as
$Q(s_{t+1},\pi(s_{t+1}))$, to highlight the difference with off-policy
learning.
SARSA updates its Q-values
using the Q-value of the next state $s$ and the current policy's
action. 
The primary advantage of on-policy learning is  its  predictive behavior.

\subsubsection*{Off-Policy Q-Learning}\index{Q-learning}
Off-policy learning is more complicated; it may learn its policy from
actions that are different from  the one just taken.

The best-known off-policy
algorithm is Q-learning~\cite{watkins1989learning}.\label{qlearning}
It performs exploiting and exploring selection actions as before, but it evaluates states as if a
greedy policy is used always, 
even when the actual behavior performed an exploration step.

The Q-learning update formula is
\begin{equation}
  Q(s_t,a_t)\leftarrow Q(s_t,a_t)+\alpha[r_{t+1}+\gamma \max_a
  Q(s_{t+1},a)-Q(s_t,a_t)]. \label{eq:qlearning}
  \end{equation}
The only difference from on-policy learning is that the $\gamma
Q(s_{t+1},a_{t+1})$ term from Eq.~\ref{eq:sarsa} has been replaced by $\gamma \max_a
Q(s_{t+1},a)$. We now learn  from backup values of the best action,
not the one that was actually evaluated.
Listing~\ref{lst:td} showed the  pseudocode for Q-learning. Indeed,
the term temporal difference learning is sometimes used for the
Q-learning algorithm.


The reason that Q-learning is called off-policy is that it updates its
Q-values using the Q-value of the next state $s_{t+1}$, {\em and the greedy action}
(not necessarily the behavior policy's action---it is learning off the
behavior policy).
Off-policy learning collects all available  information and uses it to construct the best target policy.


\subsubsection*{Sparse Rewards and Reward Shaping}
\index{reward}\index{sparse reward}\index{reward shaping}
Before we conclude this section, we should discuss sparsity.
For some environments a reward exists for each state. For
the supermarket example a reward can be
calculated for each state that the agent has walked to.  (The reward is
 the opposite of the cost expended in walking.) Environments in which a reward
exists in each state are said to have a dense reward structure.

For other environments rewards may exist for only some of the
states. For example, in chess, rewards only exist at terminal board positions
where there is a win or a draw. In all other states the
return depends on the future states and must be calculated by the
agent by propagating reward values from future states up towards the
root state $s_0$.
Such an environment is said to have a sparse reward structure.

Finding a good policy is more complicated when the  reward
structure is sparse. A graph of the landscape of such a sparse reward
function would show a flat landscape with a few sharp mountain
peaks. Reinforcement learning algorithms use the reward-gradient to
find good returns. Finding the optimum in a flat landscape where the
gradient is 
zero, is hard.  In some applications it is possible to change the reward
function to have a shape more amenable to gradient-based optimization
algorithms such as we use in deep learning.  Reward shaping can make
all the difference when no solution can be found with a naive reward function. It is a way of
incorporating heuristic knowledge into the MDP.  A
large literature on reward shaping and heuristic
information exists~\cite{ng1999policy}. The use of heuristics on board games
such as chess and checkers can also be regarded as reward
shaping.

\subsubsection{\em Hands On: Q-learning on Taxi }\label{sec:taxi-q}

To get a feeling for how these  algorithms work in
practice, let us see how Q-learning solves the Taxi problem.

In Sect.~\ref{sec:taxi-vi} we discussed how value iteration can be
used for the Taxi
problem, provided that the agent has access to the transition model. We will now see how we
 solve this problem if we do not have the transition model. Q-learning samples actions, and
records the reward values in a Q-table, converging to the state-action value
function. When in all states the best values of the best actions are
known, then these can be used to sequence the optimal policy. 

Let us see how a value-based model-free algorithm  solves a simple
$5\times 5$ Taxi problem. Refer to Fig.~\ref{fig:taxi} on
page~\pageref{fig:taxi} for an illustration of  Taxi world.

Please recall that in Taxi world, the taxi can be in one of 25
locations
and there are
$25 \times (4 + 1) \times 4 = 500$ different states that the
environment can be in.

We  follow the reward model as it is used in the Gym Taxi
environment.
Recall that our goal is to find a policy (actions in
each state) that
leads to the highest cumulative reward.
Q-learning learns the best policy through guided sampling. The agent records
the rewards that it gets from actions that it performs in the environment. The
Q-values are the expected rewards of the actions in the states. The agent
uses the Q-values to guide which actions it will sample. Q-values
$Q(s,a)$ are
stored in an array that is indexed by state and action. The Q-values
guide the exploration, higher values indicate better actions.

Listing~\ref{lst:q} shows the full Q-learning algorithm, in Python, after~\cite{learn}. It
uses an $\epsilon$-greedy behavior policy: mostly the best action is
followed, but in a certain fraction a random action is chosen, for exploration.
Recall that the Q-values are updated according to the Q-learning formula:
$$Q(s_t,a_t)\leftarrow Q(s_t,a_t)+\alpha[r_{t+1}+\gamma \max_a
Q(s_{t+1},a)-Q(s_t,a_t)] $$
where $0\leq\gamma\leq 1$ is the discount factor and $0<\alpha\leq 1$
the learning rate. Note that Q-learning uses bootstrapping, and the
initial Q-values are set to a random value (their value will disappear
slowly due to the learning rate).

Q-learning  learns the best action  in the current state by
looking at the reward for the current state-action combination, plus the maximum
rewards for the next state. Eventually the best policy is found in
this way, and the taxi will consider the route consisting of a
sequence of the best rewards.

To summarize informally:
\begin{enumerate}
\item Initialize the Q-table to random values
\item Select a state $s$
\item For all possible actions from $s$ select the one with the
  highest Q-value and travel to this state, which becomes the
  new $s$, or, with $\epsilon$ greedy, explore
\item Update the values in the Q-array using the equation
\item Repeat until the goal is reached; when the goal state is
  reached, repeat the process until the Q-values 
  stop changing (much), then stop.
\end{enumerate}
\lstset{label={lst:q}}
\lstset{caption={Q-learning Taxi example, after~\cite{learn}}}
\begin{lstlisting}[language=Python,float]
# Q learning for OpenAI Gym Taxi environment
import gym
import numpy as np
import random
#Environment Setup
env = gym.make("Taxi-v2")
env.reset()
env.render()
# Q[state,action] table implementation
Q = np.zeros([env.observation_space.n, env.action_space.n])
gamma = 0.7   # discount factor
alpha = 0.2   # learning rate
epsilon = 0.1 # epsilon greedy
for episode in range(1000):
    done = False
    total_reward = 0
    state = env.reset()
    while not done:
        if random.uniform(0, 1) < epsilon:
            action = env.action_space.sample() # Explore state space
        else:
            action = np.argmax(Q[state]) # Exploit learned values        
        next_state, reward, done, info = env.step(action) # invoke Gym
        next_max = np.max(Q[next_state])
        old_value = Q[state,action]
       
        new_value = old_value + alpha * (reward + gamma * next_max - old_value) 

        Q[state,action] = new_value
        total_reward += reward
        state = next_state    
    if episode % 100 == 0:
        print("Episode {} Total Reward: {}".format(episode,total_reward))

        
\end{lstlisting}
Listing~\ref{lst:q} shows Q-learning code for finding the
policy in  Taxi world.

\lstset{label={lst:evaltaxi}}
\lstset{caption={Evaluate the optimal Taxi result, after~\cite{learn}}}
\begin{lstlisting}[language=Python,float]
total_epochs, total_penalties = 0, 0
ep = 100
for _ in range(ep):
    state = env.reset()
    epochs, penalties, reward = 0, 0, 0
    done = False
    while not done:
        action = np.argmax(Q[state])
        state, reward, done, info = env.step(action)
        if reward == -10:
            penalties += 1
        epochs += 1
    total_penalties += penalties
    total_epochs += epochs
print(f"Results after {ep} episodes:")
print(f"Average timesteps per episode: {total_epochs / ep}")
print(f"Average penalties per episode: {total_penalties / ep}")
\end{lstlisting}

The optimal policy can be found by sequencing together the actions
with the highest Q-value in each state.
Listing~\ref{lst:evaltaxi} shows the  code for this. The number of
illegal pickups/drop-offs is shown as penalty. 

This example shows how the optimal policy can be found by the
introduction of a Q-table that records the quality of irreversible actions in
each state, and uses that table to converge the rewards to the value
function. In this way the optimal policy can be found model-free.

\subsubsection*{Tuning your Learning Rate}\index{learning rate tuning}
Go ahead, implement and run this code, and play around to become
familiar with the algorithm. Q-learning is an excellent algorithm to
learn the essence of how reinforcement learning works. Try out
different values for hyperparameters, such as the exploration
parameter $\epsilon$, the discount factor $\gamma$ and
the learning rate $\alpha$. To be successful in this field, it helps to have a
feeling for these hyperparameters. A choice close to 1 for the
discount parameter is usually a good start, and a choice close to 0
for the learning rate is a good start. You may feel a tendency to do
the opposite, to choose the learning rate as high as possible (close
to 1) to learn as quickly as possible. Please go ahead and see which
works best in Q-learning (you can have a look at
\cite{even2003learning}).  In many deep learning environments a high
learning rate  is a
recipe for disaster, your algorithm may not converge at all, and
Q-values can become unbounded. Play around with tabular Q-learning,
and approach your deep learning slowly, with
gentle steps!

The Taxi  example is small, and you will
get results quickly. It is well suited to build up useful
intuition. In later chapters, we will do experiments with 
deep learning that take longer to converge, and acquiring intuition
for tuning hyperparameter values will be more expensive.

\subsubsection*{Conclusion}
We have now seen how a value function can be learned by an agent
without having the 
transition function, by sampling  the environment. Model-free
methods use actions that are irreversible for the agent. The agent samples
states and rewards from the environment, using a behavior policy with
the current best action, and following an
exploration/exploitation trade-off. The backup rule for learning is
based on bootstrapping, and can follow the rewards of the actions
on-policy, including the value of the occasional explorative action,
or off-policy, always using the value of the best action. We have seen
two model-free tabular algorithms, SARSA and Q-learning, where the
value function is assumed to be stored in an exact table  structure.

In the next chapter we will move to network-based algorithms for
high-dimensional state spaces, 
based on function approximation with a deep neural network. 

\section{Classic Gym Environments}
\label{sec:bench}
Now that we have discussed at length the tabular agent algorithms, it is time
to have a  look at the environments,
the other part of the reinforcement
learning model.
%
Without them, progress cannot be measured, and results
cannot be compared in a meaningful way. In a real sense, environments 
define the kind of intelligence that our artificial methods can be
trained to perform.

In this chapter we will start with a few  smaller environments, that are suited for the
tabular algorithms that we have discussed.
Two environments that have been around since the early days of
reinforcement learning are Mountain car and Cartpole 
(see Fig.~\ref{fig:cartpole}). 

\subsection{Mountain Car and Cartpole}\index{mountain car}\index{cartpole}
Mountain car is a physics puzzle in which a car on a one-dimensional
road is in a valley between two mountains. The goal for the car is to
drive up the mountain and reach the flag on the right. The car's
engine can go forward and backward. The problem is
that the car's engine is not strong enough to climb the mountain by
itself in a single pass~\cite{moore1990efficient}, but it can do so with the help of gravity: by repeatedly going
back and forth the car can build up momentum. The challenge for the
reinforcement learning agent is to apply alternating backward and forward
forces at the right moment.

Cartpole is a pole-balancing problem. A pole is attached by a joint to a 
movable cart, which can  be pushed forward or backward.  The
pendulum starts upright, and must be kept upright by applying either a
force of $+1$ or $-1$ to the cart. The puzzle ends when the pole falls
over, or when the cart runs too far left or
right~\cite{barto1983neuronlike}. Again the challenge is to apply the
right force at the right moment, solely by feedback of the pole being
upright or too far down.

\begin{figure}[t]
  \begin{center}
    \begin{tabular}{cc}
      \includegraphics[width=5cm]{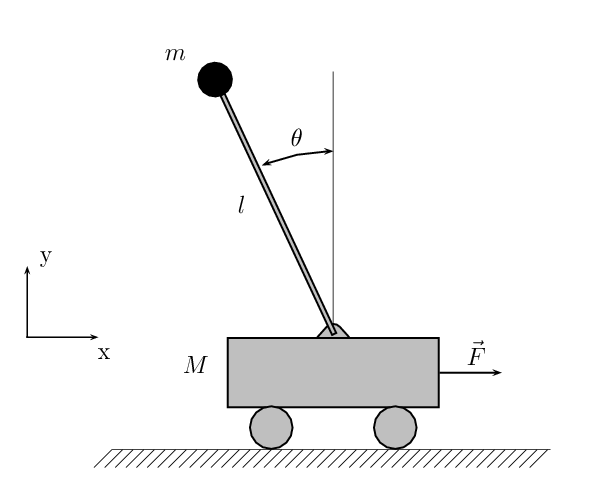}
      &
      \includegraphics[width=5cm]{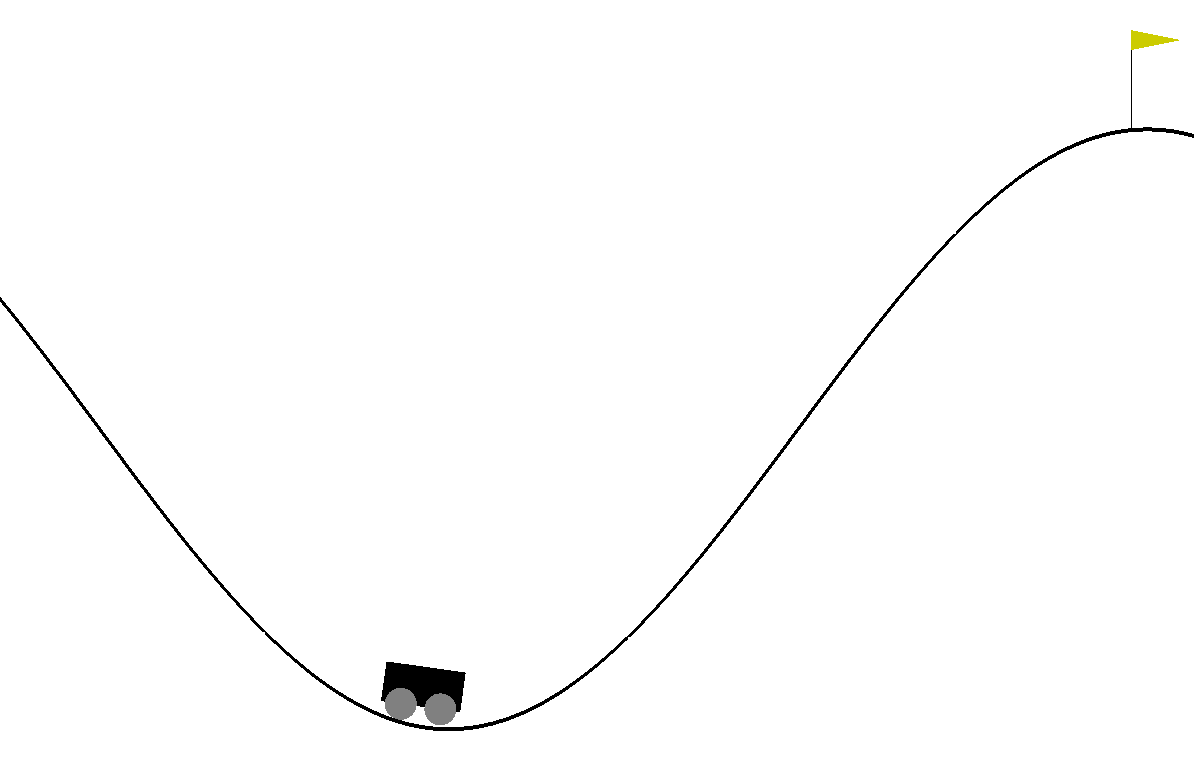}
    \end{tabular}
 \caption{Cartpole and Mountain Car}\label{fig:cartpole}
 \end{center}
\end{figure}

\subsection{Path Planning and Board Games}



Navigation tasks and board games provide environments for reinforcement
learning that are simple to understand. They are well suited  to
reason about new agent algorithms. Navigation problems, and the
heuristic search trees built for 
board games, can be of moderate size, and are then suited for determining the
best action by dynamic programming  methods, such as
tabular Q-learning, A*, branch and bound, and
alpha-beta~\cite{russell2016artificial}. These are 
straightforward search  methods that do not attempt to
generalize to new, unseen, states. They  find the best action in a
 space of states, all of which are present at training
time---the optimization methods do not perform generalization from training to test time.

\subsubsection*{Path Planning}
Path planning  (Fig~\ref{fig:gridworld}) is a classic problem that is related to
robotics~\cite{latombe2012robot,gasparetto2015path}.
Popular versions are mazes, as we have seen earlier
(Fig.~\ref{fig:maze}). The Taxi domain (Fig.~\ref{fig:taxi})
was originally introduced
in the context of
hierarchical problem solving~\cite{dietterich2000hierarchical}. Box-pushing
problems such as Sokoban are frequently used as well~\cite{junghanns2001sokoban,dor1999sokoban,murase1996automatic,zhou2013tabled},
see Fig.~\ref{fig:sokoban}. The
action space of these puzzles and mazes is discrete. Basic path and
motion planning can  enumerate possible solutions~\cite{culberson1997sokoban,hearn2009games}.

Small versions of  mazes can be solved exactly by enumeration, 
larger instances are only suitable for approximation  methods.
Mazes can be used to test  algorithms for 
path finding problems and are frequently used to do so.
Navigation tasks and box-pushing games such as Sokoban can  feature rooms or subgoals, that may
then be used  to test algorithms for hierarchically
structured
problems~\cite{farquhar2018treeqn,guez2019investigation,racaniere2017imagination,feng2020solving}
(Chap.~\ref{chap:hier}). 
The problems can be made more difficult  
by enlarging the grid and by inserting more obstacles. 

\subsubsection*{Board Games}
Board games are a classic group of benchmarks for planning and
learning  since the earliest days of artificial intelligence.
Two-person zero-sum perfect information board games such as tic tac
toe, chess,
checkers, Go, and shogi have been used to test algorithms since the 1950s. The action
space of these games is discrete.  Notable achievements were in
checkers, chess, and Go, where  human world champions were defeated in
1994, 1997, and 2016,
respectively~\cite{schaeffer1996chinook,campbell2002deep,silver2016mastering}.

The board games are typically used ``as is'' and are not changed
for different experiments (in contrast to mazes, that are often
adapted in size or complexity for specific purposes of the
experiment). Board games are  used for the difficulty of the
challenge. The ultimate goals is  to beat human grandmasters or
even the world champion.
%
Board games have been traditional mainstays of artificial
intelligence, mostly associated 
with the  search-based symbolic reasoning approach to
artificial intelligence~\cite{russell2016artificial}. In contrast, the
benchmarks in the next chapter are 
associated with connectionist artificial intelligence.

\section*{Summary and Further Reading}
\addcontentsline{toc}{section}{\protect\numberline{}Summary and Further Reading}
This has been a long chapter, to provide a solid basis for the rest of
the book. We will summarize the chapter, and
provide references 
for further reading.
\subsection*{Summary}
Reinforcement learning can learn behavior
that achieves  high rewards, using feedback from the environment. Reinforcement learning has no 
supervisory labels, it can learn beyond a teacher, as long as there
is an environment that provides feedback. 

Reinforcement learning problems are  modeled as a Markov
decision problem, consisting of a 5-tuple  $(S,A,T_a,R_a,\gamma)$ for
states, actions, transition, reward, and discount factor.  The agent
performs an action, and the environment returns the new 
state and the reward value to be associated with the new state.

Games and robotics are two important fields of application. Fields of
application can be episodic (they end---such as a game of chess) or
continuous (they do not end---a robot remains in the world). In
continuous problems it often makes sense to discount behavior that is
far from the present, episodic problems typically do not bother with a
discount factor---a win is a win.

Environments can be deterministic (many board games are
deter\-minis\-tic---boards don't move) or
stochastic (many robotic worlds are stochastic---the world around a
robot moves).
The action space can be discrete (a piece either moves to a square or it does not) or continuous
(typical robot joints move continuously over an angle).

The goal in reinforcement learning is to find the optimal policy that gives for all states the best 
actions, maximizing the cumulative future
reward. The policy function is used in two different ways. In a
discrete environment  the policy function $a = \pi(s)$ 
returns for each state the best action in that sate. (Alternatively the
value function  returns the value of each action in each state,
out of which the argmax function can be used to find the actions with the highest
value.)

The optimal policy can be found by finding the maximal value of a
state. The value function $V(s)$ returns the expected reward for a
state. When the transition function $T_a(s,s')$ is present, the agent
can use Bellman's equation, or a dynamic programming method to
recursively traverse the behavior space. Value iteration is one such
dynamic programming method. Value  iteration traverses all actions of all states,
backing up reward values, until the value function stops
changing. 
The state-action value
$Q(s,a)$ determines the value of an action of a state. 

Bellman's equation calculates the value of a state by calculating
the value of successor states. Accessing successor states (by
following the action and transition) is also called expanding a
successor state. In a tree diagram successor states are called child nodes,
and expanding is a downward action. Backpropagating the reward values
to the parent node is a movement upward in the tree.

Methods where the agent makes use of the transition model are called
model-based methods. When the agent does not use the transition model,
they are model-free methods.
In many situations the learning agent does not have access to the
transition model of the environment, and planning methods cannot be
used by the agent.  Value-based model-free methods can find an
optimal policy by using only irreversible actions, sampling the
environment to find the value of the actions. 

A major determinant in model-free reinforcement learning is the
exploration/exploitation trade-off, or how much of the information
that has been  learned from the environment is used in choosing actions
to sample. We discussed the advantages of
exploiting the latest knowledge in settings where environment actions
are very costly, such as clinial trials.
A well-known exploration/exploitation method is $\epsilon$-greedy,
where  the greedy (best) action is followed from the behavior
policy, except in $\epsilon$ times, when random exploration is
performed. Always following the policy's best action runs the risk of
getting stuck in a cycle. Exploring random nodes allows  breaking free of
such cycles.

So far we have discussed the  action selection
operation.  How should we process the rewards that
are found at  nodes? Here we introduced another fundamental element
of reinforcement learning: bootstrapping, or finding a value by
refining  a previous value. Temporal difference learning uses
the principle of bootstrapping to find the value of a state by adding
appropriately discounted future reward values to the state value
function.

We have now discussed both up and down motions, and can construct full
model-free algorithms. The best-known algorithm may well be
Q-learning, which learns the action-value function of each action in
each state through off-policy temporal difference learning. Off-policy
algorithms improve the  policy function with the value of
the best action, even if the (exploring) behavior action was different. 

In the next chapters we will look at  value-based and policy-based
model-free methods for large, complex problems, that make use of
function approximation (deep learning).

\subsection*{Further Reading}
There is a rich literature on tabular reinforcement learning. 
A standard work for tabular value-based reinforcement learning is
Sutton and Barto~\cite{sutton2018introduction}. Two condensed
introductions to reinforcement learning are~\cite{arulkumaran2017deep,franccois2018introduction}.
Another major work on
reinforcement learning is Bertsekas and
Tsitsiklis~\cite{bertsekas1996neuro}. Kaelbling has written an
important survey article on the field~\cite{kaelbling1996reinforcement}.
The early works of Richard Bellman on dynamic programming, and
planning algorithms are~\cite{bellman1957dynamic,bellman1965application}.
For a recent treatment of games and reinforcement learning, with a
focus on heuristic search methods and the methods behind AlphaZero,
see~\cite{plaat2020learning}. 

The methods of this chapter are based on bootstrapping~\cite{bellman1957dynamic} and temporal
difference learning~\cite{sutton1988learning}. The on-policy algorithm
SARSA~\cite{rummery1994line} and the off-policy algorithm
Q-Learning~\cite{watkins1989learning} are among the best known exact,
tabular, value-based model-free algorithms.

Mazes and Sokoban
grids are sometimes procedurally
generated~\cite{shaker2016procedural,hendrikx2013procedural,togelius2013procedural}. The
goal for the 
algorithms is typically to find a solution for a grid of a certain
difficulty class, to find a shortest path solution, or, in transfer
learning,  to learn to solve a class of grids by training on a
different class  of grids~\cite{yang2021transfer}.

For general reference, one of the
major  textbooks on artificial intelligence is written by
Russell and Norvig~\cite{russell2016artificial}. A more specific
textbook on machine learning is by Bishop~\cite{bishop2006pattern}.


\section*{Exercises}
\addcontentsline{toc}{section}{\protect\numberline{}Exercises}
We will end  with questions
on key concepts, with programming exercises to
build up more experience,

\subsubsection*{Questions}
The questions below are meant to refresh your memory, and should be
answered with yes, no, or short answers of one or two sentences.
\begin{enumerate}

\item In reinforcement learning the agent can choose which
  training examples are generated. Why is
  this beneficial? What is a potential problem?
\item What is Grid world?
\item Which five elements does an MDP have to model reinforcement
  learning problems?
\item In a tree diagram, is successor selection of behavior up or down?
\item In a tree diagram, is learning  values through backpropagation
  up or down?
\item What is $\tau$?
\item What is  $\pi(s)$?
\item What is  $V(s)$?
\item What is  $Q(s,a)$?
\item What is dynamic programming?
\item What is recursion?
\item Do you know a dynamic programming method to determine the
  value of a state?
\item Is an action in an environment reversible for the agent?
\item Mention two typical application areas of reinforcement
  learning.
\item Is the action space of games typically discrete or continuous?  
\item Is the action space of robots typically discrete or
  continuous?
\item Is the environment of games typically deterministic or stochastic?
\item Is the environment of robots typically deterministic or
  stochastic?
\item What is the goal of reinforcement learning?
\item Which of the five MDP elements is not used in episodic
  problems?
\item Which model or function is meant when we say ``model-free''
  or ``model-based''?
      \item What type of action space and what type of environment are
    suited for value-based methods?
    \item Why are value-based methods used for games and not for robotics?
\item Name two basic Gym environments.
\end{enumerate}

 \lstset{label={lst:sarsa}}
 \lstset{caption={SARSA Taxi example, after~\cite{learn}}}
\begin{lstlisting}[language=Python,float]
# SARSA for OpenAI Gym Taxi environment
import gym
import numpy as np
import random
#Environment Setup
env = gym.make("Taxi-v2")
env.reset()
env.render()
# Q[state,action] table implementation
Q = np.zeros([env.observation_space.n, env.action_space.n])
gamma = 0.7   # discount factor
alpha = 0.2   # learning rate
epsilon = 0.1 # epsilon greedy
for episode in range(1000):
    done = False
    total_reward = 0
    current_state = env.reset()
    if random.uniform(0, 1) < epsilon:
        current_action = env.action_space.sample() # Explore state space
    else:
        current_action = np.argmax(Q[current_state]) # Exploit learned values        
    while not done:
        next_state, reward, done, info = env.step(current_action) # invoke Gym
        if random.uniform(0, 1) < epsilon:
            next_action = env.action_space.sample() # Explore state space
        else:
            next_action = np.argmax(Q[next_state]) # Exploit learned values        
        sarsa_value = Q[next_state,next_action]
        old_value = Q[current_state,current_action]
       
        new_value = old_value + alpha * (reward + gamma * sarsa_value - old_value) 

        Q[current_state,current_action] = new_value
        total_reward += reward
        current_state = next_state    
        current_action = next_action    
    if episode % 100 == 0:
        print("Episode {} Total Reward: {}".format(episode,total_reward))
\end{lstlisting}
    
\subsubsection*{Exercises}
There is an even  better way to learn about deep reinforcement learning
then reading about it, and that is to perform experiments yourself, to see the
learning processes unfold before your own eyes. The following
exercises are meant as starting points for your own discoveries in the
world of deep reinforcement learning.

Consider using  Gym to implement these
exercises. Section~\ref{sec:gym} explains how to install Gym. 
\begin{enumerate}
\item \emph{Q-learning} Implement Q-learning for Taxi, including the procedure to derive
  the best policy for the Q-table. Go to Sect.~\ref{sec:taxi-q} and implement it.
 Print the Q-table, to see the values on the squares. You could print a
 live policy as the search progresses. 
 Try different values for $\epsilon$, the exploration rate. Does it learn faster?
Does it keep finding the optimal solution?
  Try different values for $\alpha$, the learning rate. Is it faster?
\item \emph{SARSA}  Implement SARSA, the code is in
  Listing~\ref{lst:sarsa}. Compare your results to Q-learning, can you
  see how SARSA chooses different paths? Try different $\epsilon$
  and $\alpha$.
\item \emph{Problem size} How large can problems be before converging starts taking
  too long?
\item \emph{Cartpole} Run Cartpole with the greedy policy computed by value
  iteration. Can you make it work? Is value iteration a suitable
  algorithm for Cartpole? If not, why do you think it is not?
\end{enumerate}


\chapter{Deep Value-Based Reinforcement Learning}\label{ch:play}\label{chap:value}
The previous chapter introduced the field of classic  reinforcement
learning. We learned about  
agents and environments, and about states, 
actions, values, and policy functions. We also saw our first planning
and learning
algorithms: value iteration, SARSA and
Q-learning. 
The methods in the previous chapter were exact, tabular, methods, that
work  for problems of 
moderate size that fit in memory.

In this chapter we move to
high-dimensional problems with large state spaces that no longer fit
in memory.  We will go beyond tabular methods and use methods to
  \emph{approximate} the value  function and to \emph{generalize} 
 beyond  trained behavior. We will do so with \emph{deep learning}.

The
methods in this chapter are deep, model-free, value-based methods,
related to 
Q-learning.
We will start by having a closer look at the new, larger, environments
that our agents must now be able to solve (or rather, approximate).
%
Next, we will look at deep reinforcement learning algorithms. In reinforcement
learning the  current behavior policy determines which
action is selected next, a process that can be
self-reinforcing. There is no ground truth, as in supervised
learning. The targets of loss functions are no longer
static, or even stable.  In deep reinforcement learning convergence to
$V$ and $Q$ values is based on a
bootstrapping process, and the first challenge  is to find training methods that converge to 
stable  function values.
Furthermore, since with neural networks the states  are approximated based on their features, 
convergence proofs can no longer count on identifying  states
individually. For many years it was assumed that  deep reinforcement
learning is inherently unstable due to a so-called \emph{deadly triad} of
bootstrapping, function approximation and off-policy learning.

However, surprisngly, solutions have been found for many of the challenges.
By combining a number of approaches (such as
the replay buffer and increased exploration) the Deep
Q-Networks algorithm (DQN)  was able to achieve stable learning in a
high-dimensional environment.  The success of
DQN spawned a large research effort to 
improve training further. We will discuss some of these new methods.

The chapter is concluded with  exercises, a summary, and pointers
to further reading.

\begin{tcolorbox}
{\bf Deep Learning:} Deep reinforcement learning builds on deep supervised learning, and
this chapter and the rest of the book assume a basic understanding
of deep learning.
When your knowledge of parameterized neural networks and function
approximation is rusty, this is the time to  go to 
Appendix~\ref{ch:deep} and take  an in-depth refresher.
The appendix also reviews essential concepts such as training, testing,
accuracy,  overfitting and the bias-variance trade-off. When in doubt,
try to answer the questions on page~\pageref{sec:dlq}. 
\end{tcolorbox}

\section*{Core Concepts}
\begin{itemize}
\item Stable convergence
\item Replay buffer
\end{itemize}

\section*{Core Problem}
\begin{itemize}
\item  Achieve stable deep reinforcement learning in large problems
\end{itemize}

\section*{Core Algorithm}
\begin{itemize}
\item Deep Q-network  (Listing~\ref{lst:dqn-pseudo})
\end{itemize}

\section*{End-to-end Learning}

\begin{figure}[t]
\begin{center}
\includegraphics[width=7cm]{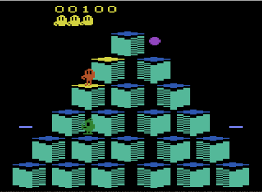}
\caption{Example Game from the Arcade Learning Environment \cite{bellemare2013arcade}}\label{fig:ale}
\end{center}
\end{figure}

\begin{figure}[t]
\begin{center}
\includegraphics[width=5cm]{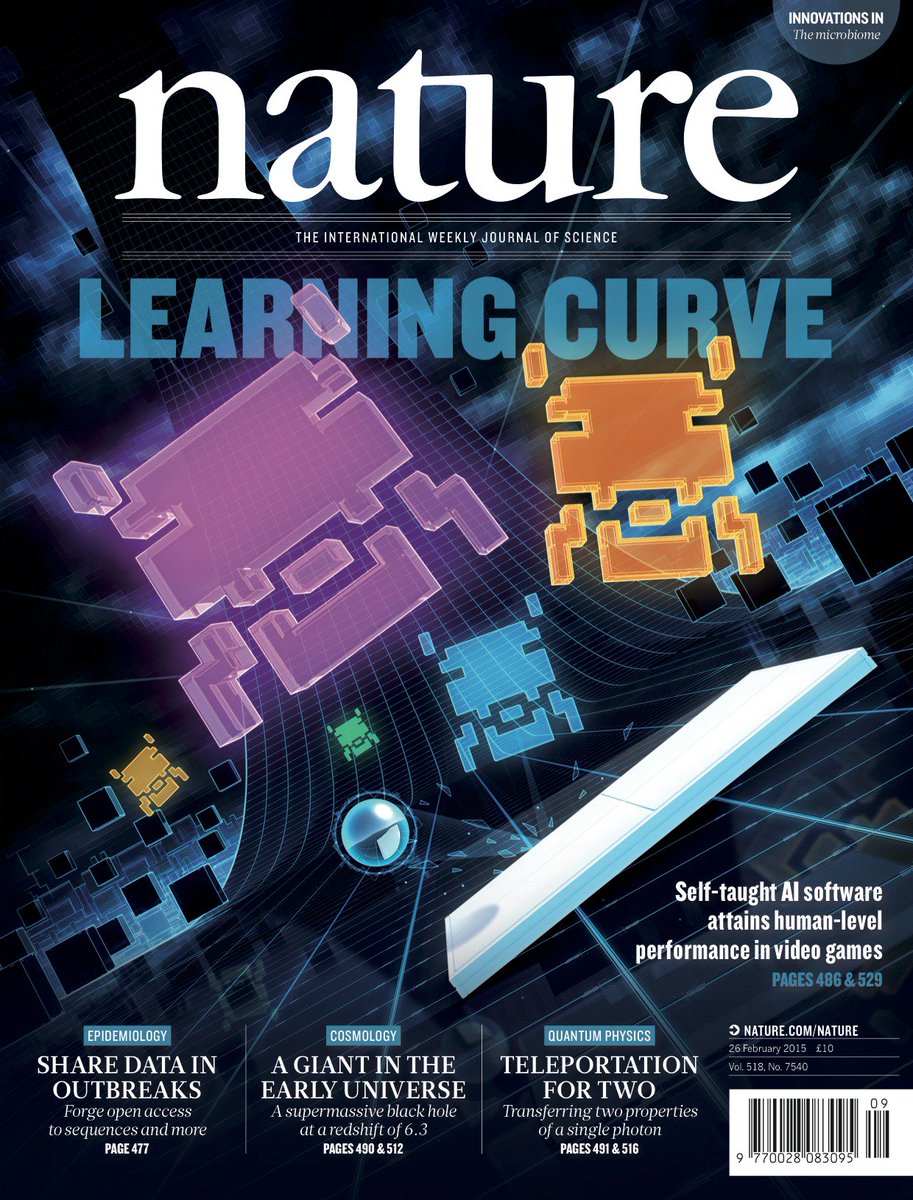}
\caption{Atari Experiments on the Cover of Nature}\label{fig:atarinature}
\end{center}
\end{figure}

Before the advent of deep learning, traditional
reinforcement learning had been used mostly on smaller problems such as puzzles,
or the supermarket example.
Their state space fits in the memories of our
computers. Reward shaping, in the form of domain-specific heuristics,
can be used  to 
shoehorn the problem into a computer, for example,  in chess and
checkers~\cite{campbell2002deep,hsu2004behind,schaeffer2008one}. Impressive
results are achieved, but at the cost of extensive problem-specific
reward shaping and heuristic engineering~\cite{plaat2020learning}. Deep learning changed
this situation, and reinforcement learning is now used on 
high-dimensional  problems that are too large to fit into memory. 

\index{ImageNet} 
In the field of supervised  learning, a yearly  competition had created 
years of steady  progress in which the accuracy of image classification had
steadily improved. Progress was driven by the availability of
ImageNet, a large database of
labeled images~\cite{fei2009imagenet,deng2009imagenet}, by increases in
computation power  through GPUs, and by steady
improvement of machine learning algorithms,  especially in deep
convolutional neural networks. In 2012, a paper by Krizhevsky,
Sutskever and Hinton presented a method that out-performed other
algorithms by a large margin, and approached the performance of human
image recognition~\cite{krizhevsky2012imagenet}. The paper
introduced the AlexNet architecture (after the first name of the first
author) and 2012 is often regarded as the year of the breakthrough of
deep learning. (See Appendix~\ref{sec:alexnet} for details.) This
breakthrough raised the 
question whether something similar in deep reinforcement learning could be achieved.

We did not have to wait long, only a year later, in 2013, at the deep learning workshop of one of the main AI
conferences, a paper was presented 
on an algorithm  that could play 1980s
Atari video games  just by training on the pixel input of the video
screen (Fig.~\ref{fig:ale}). The 
algorithm used a combination of deep learning and Q-learning, and was
named Deep Q-Network, or DQN~\cite{mnih2013playing,mnih2015human}. An
illuminating video of how it learned to play  the game Breakout is
\href{https://www.youtube.com/watch?v=TmPfTpjtdgg}{here}.\footnote{\url{https://www.youtube.com/watch?v=TmPfTpjtdgg}}
This  was a breakthrough for  reinforcement learning. Many researchers at the workshop could  relate to this
achievement, perhaps because they had spent  hours playing 
Space Invaders, Pac-Man and Pong themselves when they  
were younger. Two years after the presentation at the deep learning
workshop a longer article appeared in the journal 
Nature in which a refined and expanded version of DQN was
presented 
(see Fig.~\ref{fig:atarinature} for the journal cover). 

Why was this such a momentous achievement? Besides the fact that the
problem that was solved was easily understood, true eye-hand
coordination of this complexity had not been achieved by a computer
before; furthermore, the end-to-end
learning from pixel to joystick  implied artificial behavior that was  close
to how humans play games.  DQN essentially launched the field of deep
reinforcement learning. For the first time the power of deep learning
had been successfully combined with behavior learning, for  an imaginative
problem.

A major technical challenge that was overcome by DQN was the instability of
the deep reinforcement learning process. In fact, there were convincing
theoretical analyses at the time that this instability was fundamental, and
it was generally assumed  that it would be next to impossible to
overcome~\cite{baird1995residual,gordon1999approximate,tsitsiklis1997analysis,sutton2018introduction},
since the target of the loss-function depended on the convergence of
the reinforcement
learning process itself. 
%
By the end of this chapter we will have covered the problems of
convergence and 
stability in reinforcement learning. We will have seen how DQN
addresses these problems, and we will also have discussed some of the
many further
solutions that were devised after DQN.

But let us first have a look at the kind of new, high-dimensional,
environments that were the cause of these developments.


\section{Large, High-Dimensional, Problems}

In the previous chapter, Grid worlds and mazes were introduced
as basic sequential decision making problems in which exact, tabular, reinforcement
learning methods work well. These are problems of moderate
complexity. The complexity of a problem is related to
the number of  unique states that a problem has, or how large the state space 
is.  Tabular methods work for small problems,
where the entire state space fits in memory. This is for example the
case with linear regression, which has only one variable $x$ and two parameters $a$ and $b$, or the
Taxi problem, which has a state space of size $500$.
In this chapter we will be more ambitious and introduce various games,
most notably 
 Atari arcade games. The state space of a single frame of Atari video input is $210\times
 160$ pixels of $256$ RGB color values $=256^{33600}$.


There is a qualitative difference between small ($500$) and large ($256^{33600}$)
problems.\footnote{See Sect.~\ref{sec:curse}, where we discuss
 the curse of dimensionality.} For small problems the policy can be learned by loading all
states of a problem in memory. States are identified individually, and
each has its own  best action, that we can try to find. Large
problems, in contrast, do not fit in memory, the policy cannot be
memorized, and states  are
grouped together based on  their features (see 
Sect.~\ref{sec:hi-dim}, where we discuss feature learning). A parameterized network maps states  to  actions and
values; states are no longer individually identifiable in a lookup table.

When deep learning methods were
introduced in reinforcement learning,  larger
problems than before could be solved. Let us have a look at   those problems.

\subsection{Atari Arcade Games}\index{Atari
  2600}\index{Atari}\index{ALE}\index{Arcade Learning Environment}\label{sec:ale}




Learning actions
directly from high-dimensional  sound and vision inputs
is one of the long-standing challenges of artificial intelligence. To
stimulate this research, in 2012 a test-bed was created  designed to
provide challenging reinforcement learning tasks. It was called the
Arcade learning environment, or \gls{ALE}~\cite{bellemare2013arcade}, and it
was based on a simulator for 1980s Atari 2600 video games.  Figure~\ref{fig:console} shows 
a picture of a distinctly retro  Atari 2600 gaming console.

\begin{figure}[t]
\begin{center}
\includegraphics[width=8cm]{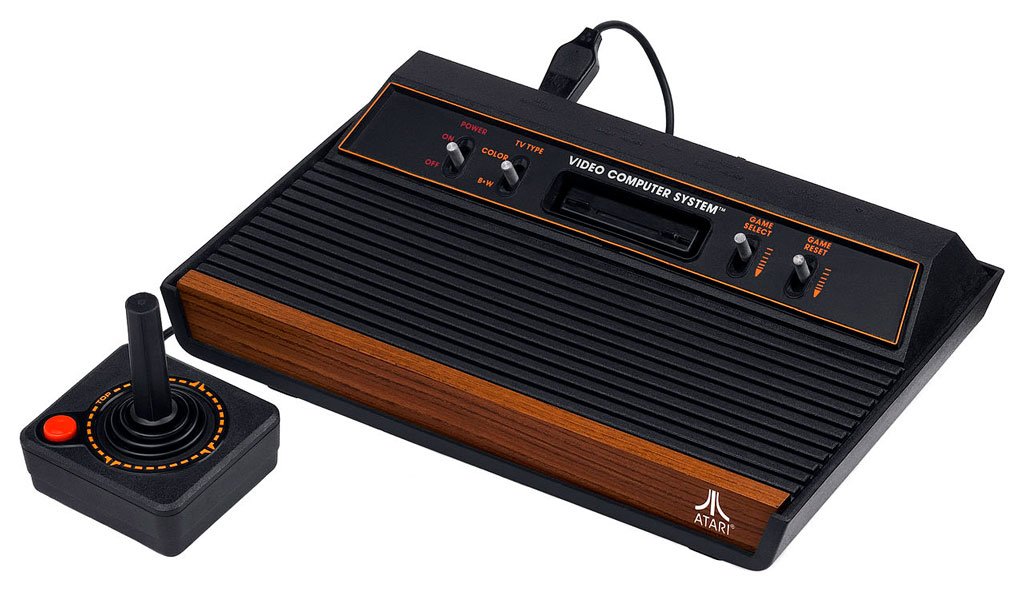}
\caption{Atari 2600 console}\label{fig:console}
\end{center}
\end{figure}

Among other things ALE contains an emulator of the Atari 2600
console. ALE presents agents with a high-dimensional\footnote{That is,
  high dimensional for machine 
  learning. $210\times 160$ pixels is not exactly high-definition
  video quality.} visual input
($210\times 160$ RGB video at $60$ Hz, or 60 images per second) of tasks that were designed to
be interesting and challenging for human players (Fig.~\ref{fig:ale}
showed an example of such a game and Fig.~\ref{fig:atarigames} shows a few more). The game cartridge ROM  holds 2-4 kB of game code, while the
console random-access memory is small, just 128
bytes (really, just 128 bytes, although the video memory is larger, of
course).
The actions can be selected via a joystick (9 directions), which has a fire button (fire on/off), giving 18 actions in total.

The Atari games provide challenging eye-hand coordination and
reasoning tasks, that are both familiar and challenging to humans,
providing a good  test-bed for learning sequential decision making.

Atari games, with high-resolution video input at high frame rates, are
an entirely different kind of challenge than Grid worlds or
board games. Atari is a  step closer to a human environment in which
visual inputs should quickly be followed by correct actions. Indeed, the
Atari benchmark called for very different agent algorithms, prompting
the move from tabular algorithms to algorithms based on function
approximation and deep learning.
ALE has become a  standard benchmark in deep reinforcement
learning research. 

\begin{figure}[t]
\begin{center}
\includegraphics[width=10cm]{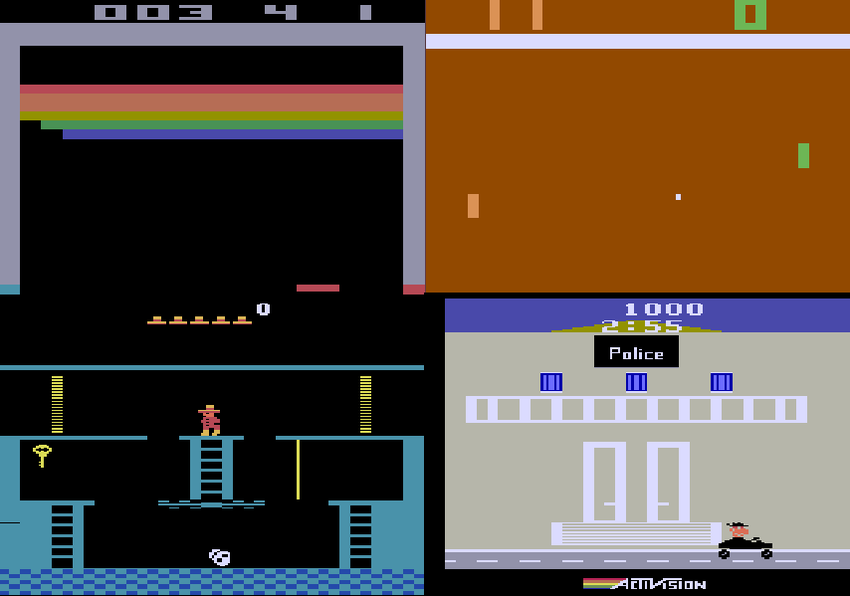}
\caption[Atari games]{Screenshots of 4 Atari Games (Breakout, Pong,
  Montezuma's Revenge, and Private Eye)}\label{fig:atarigames}
\end{center}
\end{figure}

\subsection{Real-Time Strategy and Video Games}\label{sec:rts}
Real-time strategy   games  provide an even greater  challenge
than simulated 1980s Atari consoles. Games  such as
StarCraft (Fig.~\ref{fig:starcraft})~\cite{ontanon2013survey},  and Capture the Flag~\cite{jaderberg2019human}
have very large state spaces. These are games with large maps,
many players, many pieces, and many types of actions.
The state space of StarCraft is estimated at $10^{1685}$~\cite{ontanon2013survey}, more than $1500$
orders of magnitude  larger than 
Go ($10^{170}$) \cite{muller2002computer,tromp2016} and more than $1635$ orders of magnitude large than
chess ($10^{47}$)~\cite{hsu1990grandmaster}.  Most real time strategy
games are multi-player, non-zero-sum, 
imperfect information 
games that also feature high-dimensional pixel input, reasoning, and team
collaboration.
The action space
is stochastic and is a mix of discrete and continuous actions.

Despite the challenging nature, impressive achievements have been
reported recently in  three 
games where human performance was matched or even
exceeded~\cite{vinyals2019grandmaster,berner2019dota,jaderberg2019human},
see also
Chap.~\ref{chap:multi}.

Let us have a look at the methods that can solve these very different
types of problems.

%
%
\section{Deep  Value-Based Agents}
We will now turn to agent algorithms for solving large sequential decision problems. The
main challenge of this section is to create  an agent algorithm that can
learn a good policy by interacting with the world---with a large problem, not  a
toy problem. From now on, our agents will be deep learning agents. 


The questions that we are faced with, are the following.
How  can we use deep learning for  high-dimensional and large
sequential decision making 
environments? How can  tabular value and policy functions $V$, $Q$,
and $\pi$ be transformed   into \gls{theta}
parameterized functions $V_\theta$, $Q_\theta$, and $\pi_\theta$?




\subsection{Generalization of Large Problems with Deep Learning}

Recall from Appendix~\ref{ch:deep} that deep supervised learning
uses a static dataset to approximate a 
function, and that the labels are static  targets in an
optimization process where the loss-function is minimized.  

Deep reinforcement learning is based on the observation that
bootstrapping is also a kind of minimization process in which an error
(or difference) is minimized.  In reinforcement learning this
bootstrapping process converges on the true state value and
state-action value functions.
However,  the
Q-learning bootstrapping process lacks   static 
 ground truths; our data items are generated
dynamically, and our loss-function targets move. The movement of the
loss-function targets is
influenced by the same policy function that the convergence process is trying to
learn.

It has taken  quite
some effort to find  deep learning 
algorithms that converge to stable functions on these moving targets. Let us try to
understand in more detail how the supervised methods have to be
adapted in order to work in reinforcement learning. We do this  by comparing three
algorithmic structures: supervised minimization, tabular Q-learning,
and deep Q-learning.

\subsubsection{Minimizing Supervised Target Loss}
Listing~\ref{lst:train_sl}
shows pseudocode for a typical 
supervised deep learning training algorithm, consisting of an input dataset,
a forward pass that calculates the network output, a loss computation,
and a backward pass. See Appendix~\ref{ch:deep}
or~\cite{goodfellow2016deep} for more details.

We see that the 
code consists of a double loop: the outer loop controls the training
epochs. Epochs consist of forward approximation of the target value
using the parameters, computation of the gradient, and backward adjusting
of the parameters with the gradient. In each epoch the inner loop
serves all examples of the static 
dataset to the forward computation of the output value, the loss and
the gradient computation, so that the parameters can
be  adjusted in the backward pass.

The dataset is static, and all that
the inner loop does 
is deliver the samples to  the
backpropagation algorithm. Note that each
sample is independent
of the other, samples are chosen with 
equal probability. After an image of a white horse is sampled, the
probability that the next image is of a black grouse or a blue moon is
equally (un)likely.

\lstset{label={lst:train_sl}}
\lstset{caption={Network training pseudocode for supervised learning}}
\begin{lstlisting}[language=Python,float]
def train_sl(data, net, alpha=0.001):     # train classifier
    for epoch in range(max_epochs):       # an epoch is one pass 
        sum_sq = 0                 # reset to zero for each pass 
        for (image, label) in data: 
            output = net.forward_pass(image) # predict
            sum_sq += (output - label)**2 # compute error
        grad = net.gradient(sum_sq)       # derivative of error
        net.backward_pass(grad, alpha)    # adjust weights
    return net        
\end{lstlisting}

\lstset{label={lst:qlearn3}}
\lstset{caption={Q-learning pseudocode~\cite{watkins1989learning,sutton2018introduction}}}
\begin{lstlisting}[language=Python,float]
def qlearn(environment, alpha=0.001, gamma=0.9, epsilon=0.05): 
    Q[TERMINAL,_] = 0 # policy
    for episode in range(max_episodes):
        s = s0
        while s not TERMINAL: # perform steps of one full episode
            a = epsilongreedy(Q[s], epsilon) 
            (r, sp) = environment(s, a)
            Q[s,a] = Q[s,a] + alpha*(r+gamma*max(Q[sp])-Q[s,a])
            s = sp
     return Q       

 
\end{lstlisting}

\lstset{label={lst:train_rl}}
\lstset{caption={Network training pseudocode for reinforcement learning}}
\begin{lstlisting}[language=Python,float]
def train_qlearn(environment, Qnet, alpha=0.001, gamma=0.0, epsilon=0.05
    s = s0           # initialize start state
    for epoch in range(max_epochs): # an epoch is one pass 
        sum_sq = 0  # reset to zero for each pass
        while s not TERMINAL: # perform steps of one full episode
            a = epsilongreedy(Qnet(s,a)) # net: Q[s,a]-values
            (r, sp) = environment(a)
            output = Qnet.forward_pass(s, a) 
            target = r + gamma * max(Qnet(sp)) 
            sum_sq += (target - output)**2
            s = sp
        grad = Qnet.gradient(sum_sq) 
        Qnet.backward_pass(grad, alpha) 
    return Qnet    # Q-values   
\end{lstlisting}

\subsubsection{Bootstrapping Q-Values}
Let us now look at Q-learning.
Reinforcement learning chooses the training examples differently. For convergence of
algorithms such as Q-learning, the selection rule must guarantee that
eventually  all states will be sampled by the 
environment~\cite{watkins1989learning}. For large problems, this is
not the case; this condition for convergence to the value function does
not hold.

Listing~\ref{lst:qlearn3}
shows the short version of the bootstrapping tabular Q-learning
pseudocode from the previous chapter.  As in the previous deep learning algorithm, the algorithm consists of a
double loop. The outer loop controls the Q-value convergence episodes, and
each episode consists of a single trace of (time)
steps from the start state to a terminal state. The Q-values are
stored in a Python-array  indexed by $s$ and $a$, since Q is the
state-action value. 
Convergence of the Q-values is assumed to
have occurred when enough episodes have been sampled. The Q-formula
shows how the Q-values are built up by bootstrapping on previous
values, and how Q-learning is learning off-policy, taking the max
value of an action.

A difference with the supervised learning is that in Q-learning
subsequent  samples are not independent. The 
next action is determined by the current policy, and will most likely
be the best action of the state ($\epsilon$-greedy). Furthermore, the next state will
be correlated to the previous state in the trajectory. After a state
of the ball in the upper left corner of the field has been sampled,
the next sample will with very high probability also be of a state
where the ball is close to the upper left corner of the
field. Training can be stuck in local minima, hence the need for exploration.

\subsubsection{Deep Reinforcement Learning Target-Error}
The two algorithms---deep learning and Q-learning---look similar in
structure. Both
consist of a double loop in which a target is optimized, and we can
wonder if bootstrapping can be combined with loss-function
minimization. This is indeed the case, as Mnih et
al.~\cite{mnih2013playing} showed in 2013.  Our  third listing, Listing~\ref{lst:train_rl}, shows a naive deep learning
version of Q-learning~\cite{mnih2013playing,moerland2021lecture},
based on the double loop that now bootstraps Q-values by
minimizing a loss function through adjusting the $\theta$
parameters.

Indeed, a Q-network can be trained with a gradient by minimizing a sequence of
loss functions. The loss
function for
this bootstrap process is quite literally based on the Q-learning
update formula. The loss function is  the squared difference
between the new Q-value $Q_{\theta_t}(s,a)$ from the forward pass and the old update target
$r+\gamma\max_{a'}Q_{\theta_{t-1}}(s',a')$.\footnote{Deep Q-learning
  is a fixed-point iteration~\cite{melo2007convergence}. The gradient of
  this loss function is 
  $\nabla_{\theta_i}\mathcal{L}_i(\theta_i)=\mathbb{E}_{s,a\sim\rho(\cdot);s'\sim\mathcal{E}}
  \left[\left(r +\gamma\max_{a'}Q_{\theta_{i-1}}(s',a') -
      Q_{\theta_i}(s,a)\right)\nabla_{\theta_i}Q_{\theta_i}(s,a)\right]$
  where  $\rho$ is the behavior distribution and $\mathcal{E}$
  the Atari emulator. Further details are in~\cite{mnih2013playing}.}

An important observation is that  the update targets
depend on the previous network weights $\theta_{t-1}$ (the
optimization targets move
during optimization); this is in contrast
with the targets 
used in a supervised learning process, that are fixed before learning
begins~\cite{mnih2013playing}. In other words, the loss function of
deep Q-learning minimizes a moving  target, a target that depends on the network
being optimized.

\subsection{Three Challenges}
Let us have a closer look at the challenges that deep reinforcement
learning faces. There are three problems with our naive deep Q-learner. First,
convergence to the optimal Q-function depends on full coverage of the
state space, yet the state space is too large to sample fully. Second, there is a strong correlation between subsequent
training samples, with a real risk of local optima. Third, the loss function of gradient descent
literally has  a moving target, and bootstrapping may diverge. Let us have a  look at these three
problems in more detail.
\index{coverage, correlation,
  convergence}
\subsubsection{Coverage}\index{coverage}\index{out-of-distribution training}

Proofs that algorithms such as Q-learning converge to the optimal
policy depend on the  assumption of full state space coverage; all
state-action pairs must be  sampled. Otherwise, 
the algorithms will not converge to an optimal action value for each
state. Clearly, in 
large state spaces where  not all states are sampled, this situation does not 
hold, and there is no guarantee of convergence.

\subsubsection{Correlation}\index{correlation}

In  reinforcement learning a sequence of states is generated in an
agent/environment loop.  The states differ only by a single action, one move or one
stone, all other features of the states remain unchanged, and thus,
the values of subsequent
samples are  correlated, which may result in a biased training. The
training may cover only a
part of the state space, especially when greedy action selection increases the
tendency to select a small set of actions and states.  
%
The bias can result in the so-called specialization trap (when there
is too much exploitation, and too
little exploration).

Correlation between subsequent states  contributes to the
low coverage that we discussed before, reducing convergence towards the
optimal Q-function, increasing the probability of local optima and
feedback loops.
%
This happens, for example, when a chess program has been trained
on a particular opening, and the opponent plays a different one. When
test examples are different from  training examples, then 
generalization will be bad. This problem is related to out-of-distribution
training, see for example~\cite{liu2020hybrid}.


\subsubsection{Convergence}\index{convergence}

When we naively apply our deep supervised methods to reinforcement
learning, we encounter the problem that in a bootstrap process, the
optimization target is part of the bootstrap process itself.
Deep supervised learning
uses a static dataset to approximate a 
function, and loss-function targets are therefore stable. However,
deep reinforcement learning uses as bootstrap target the Q-update from
the previous time step, which changes during the optimization.

The loss is the squared difference
between the  Q-value $Q_{\theta_t}(s,a)$  and the old update target
$r+\gamma\max_{a'}Q_{\theta_{t-1}}(s',a')$. Since both depend on parameters
$\theta$ that are optimized, the risk of overshooting the target is
real, and the optimization process can
easily become unstable. 
It has taken  quite some effort to find  algorithms that can tolerate
these moving targets. 

\subsubsection*{Deadly Triad}\index{deadly triad}\label{sec:triad}

Multiple works~\cite{baird1995residual,gordon1999approximate,tsitsiklis1997analysis}
showed  that a combination of off-policy reinforcement 
learning with nonlinear function
approximation (such as deep neural networks)  could cause
Q-values to diverge. Sutton and
Barto~\cite{sutton2018introduction} further analyze  three 
elements  for divergent training: function
approximation, bootstrapping, and  off-policy
learning. Together, they are called  {\em deadly triad}.

Function approximation may  attribute values to states
inaccurately. In contrast to exact tabular methods, that are designed to
identify individual states exactly, neural networks are
designed to  individual \emph{features} of states. These
features can be shared by different states, and values attributed to those features
are shared also by other states. Function approximation may thus cause
 mis-identification
of states, and reward values and Q-values that are not assigned correctly.
If the accuracy of
the approximation of the true function values is good enough, then
states may be identified well enough to reduce or prevent divergent
training processes and loops~\cite{mnih2015human}.  

Bootstrapping of values builds up new values on the basis of
older values. This occurs in
Q-learning and temporal-difference learning where the current value depends on the
previous value. Bootstrapping increases the efficiency of the training because
values do not have to be calculated from the start.  However, errors or
biases in initial values may persist, and  spill over to other states
as values are propagated incorrectly due to function approximation.   Bootstrapping and
function approximation can thus increase divergence.

Off-policy  learning uses a behavior policy that is different from
the target policy that we are optimizing for
(Sect.~\ref{sec:off-policy}). When the behavior  policy is improved, the off-policy 
values may not  improve. Off-policy learning converges generally less well than
on-policy learning as it converges independently from the behavior
policy. With function approximation convergence may be even slower,
due to values being assigned to incorrect states.  

\subsection{Stable Deep Value-Based Learning}\label{sec:stable}
\label{sec:replay}

These considerations 
discouraged  further research  in deep reinforcement
learning for many years. Instead, research focused for some time on linear function approximators,  which have 
better convergence guarantees. Nevertheless, work
on convergent deep reinforcement learning
continued~\cite{sallans2004reinforcement,heess2013actor,bhatnagar2009convergent,maei2010toward},
and algorithms such as neural fitted
Q-learning were developed, which showed some promise~\cite{riedmiller2005neural,lange2010deep,lin1993reinforcement}. After 
the further  results of DQN~\cite{mnih2013playing} showed convincingly that
convergence and stable learning could be achieved in a non-trivial
problem, even more experimental studies were performed 
to find out under which circumstances convergence 
can be achieved and the deadly triad can be overcome. Further convergence and
diversity-enhancing techniques were developed, 
some of which we will cover in Sect.~\ref{sec:rainbow}.\index{neural fitted Q learning}\index{TD-Gammon}\index{Neurogammon}\index{Tesauro, Gerald}\label{sec:tdgam}

Although the theory provides  reasons why function approximation may
preclude stable reinforcement learning, there were, in fact, indications
that stable training 
is possible. Starting at the end of the 1980s,  Tesauro had written a
program that played very strong Backgammon 
based on a neural network. The program was called Neurogammon, and
used   supervised
learning from grand-master games~\cite{tesauro1989neurogammon}. In order to improve the
strength of the program, he switched to temporal
difference reinforcement
learning from self-play
games~\cite{tesauro1995temporal}. TD-Gammon~\cite{tesauro1995td} 
learned by playing against itself, achieving stable learning in a shallow
network. 
TD-Gammon's training used a temporal difference
algorithm similar to Q-learning, approximating the value 
function with a  network with one hidden layer, using raw board input
enhanced with hand-crafted heuristic features~\cite{tesauro1995td}.
Perhaps some form of stable reinforcement
learning was possible, at least in a shallow network?

TD-Gammon's success prompted attempts with TD learning in
checkers~\cite{chellapilla1999evolving} and
Go~\cite{sutskever2008mimicking,clark2014teaching}. Unfortunately the 
success could not be replicated in these games, and it was believed
for some time that Backgammon was a special
case, well suited for reinforcement learning and
self-play~\cite{pollack1997did,schraudolph1994temporal}. 

However, as there came further reports of  successful
applications of deep neural networks  in a reinforcement learning
setting~\cite{heess2013actor,sallans2004reinforcement}, 
more work followed. The  
results in Atari~\cite{mnih2015human} and later in
Go~\cite{silver2017mastering} as well as further
work~\cite{van2018deep} have now provided clear evidence 
that both stable training
and generalizing deep reinforcement learning are indeed possible, and
have improved our understanding of the circumstances that influence
stability and convergence.

Let us have a closer look at
the methods that are used to achieve stable deep reinforcement
learning.

\subsubsection{Decorrelating States}\index{decorrelation}
\label{sec:dqn}

As mentioned in the introduction of this chapter, in 2013 
Mnih et al.~\cite{mnih2013playing,mnih2015human} 
published their work on {\em end-to-end\/} reinforcement
 learning in Atari games. 

The original focus of \gls{DQN} is on
breaking  correlations between subsequent states, and also on
slowing
down  changes to parameters in the training process to improve stability.
%
%
The DQN algorithm has two methods to achieve this: (1) experience replay
and (2) infrequent weight updates.  We will first look at experience replay.

\subsubsection*{Experience Replay}
\index{experience replay}
In reinforcement learning training samples are created in
a sequence of  interactions with the environment,  and subsequent
training states are strongly
correlated to  preceding states. There is a tendency to  train the
network  on too
many samples of a certain kind or in a certain area, and other parts of the state
space remain under-explored. Furthermore, through function
approximation and bootstrapping,
some behavior may be  forgotten.  When an agent  reaches a new level
in a game that is different from previous levels, the agent may forget how to
play the other level. 

We can reduce correlation---and the local minima they cause---by
adding a small amount 
of supervised  learning.  
\index{replay buffer}
To break correlations and to create a more diverse set of training
examples, DQN uses  \emph{experience replay}. Experience replay
introduces a replay buffer~\cite{lin1992self}, a cache of 
previously explored states, from which it samples training states at
random.\footnote{Originally experience replay  is, as so much in
  artificial intelligence,
  a  biologically inspired
  mechanism~\cite{mcclelland1995there,o2010play,lin1993reinforcement}.}
Experience replay stores the last $N$ examples in
the replay memory, and samples uniformly  when
performing updates. A typical number for $N$ is $10^6$
\cite{zhang2017deeper}. By using  a buffer, a dynamic dataset
from which recent training examples are sampled, we train states from
a more diverse set, instead of only from the most recent one.
The goal of experience replay is to increase the independence of subsequent
training examples. The
next state to be trained on is no longer a direct successor of the
current state, but one somewhere in a long history of previous states. In
this way the replay buffer spreads out the learning over more previously seen
states, breaking temporal correlations between samples.
DQN's replay buffer  (1) improves  coverage, and (2) reduces correlation.

 DQN treats all examples equal, old
and recent alike. A form of importance sampling might differentiate between
important transitions, as we will see in the next section.

Note that, curiously, training by experience replay is a form of  off-policy
learning, since the  target parameters are 
different from those used to generate the sample.
Off-policy learning
is one of the three elements of the deadly triad, and we find that
stable learning can actually be improved by a special form of one of
its problems. 

Experience replay  works well  in
Atari~\cite{mnih2015human}. However,   further  analysis of replay
buffers has pointed to possible problems.
Zhang et
al.~\cite{zhang2017deeper} study the deadly triad  with experience
replay, and find that larger networks 
resulted in more instabilities, but also that longer multi-step
returns  yielded fewer unrealistically high reward values. In
Sect.~\ref{sec:rainbow} we will see many further enhancements to
DQN-like algorithms.

\subsubsection{Infrequent Updates of Target Weights}
  The second improvement in DQN is {\em infrequent weight updates,}
introduced in the 2015 paper on DQN~\cite{mnih2015human}. The aim of this improvement is to
reduce divergence that is caused by frequent updates of weights of
the target $Q$-value. Again, the aim is to improve the stability of  the
network optimization  by improving the stability of the $Q$-target in
the loss function.

Every $n$ updates, the network $Q$ is
cloned to obtain  target network $\hat Q$, which is used for
generating the targets for the following $n$ updates to $Q$.
In the original DQN
implementation a single 
set of network weights $\theta$ are used, and the network is
trained on a moving loss-target. Now, with infrequent updates the
weights of the target 
network change much slower than those of the behavior policy,
improving the stability of the Q-targets. 

The second network improves the stability of Q-learning,
where normally an update to $Q_\theta(s_t,a_t)$  also
changes the target at each time step, quite possibly
leading to oscillations and divergence of the policy. Generating the
targets using an older set of parameters adds a delay between the time
an update to $Q_\theta$ is made and the time the update changes the targets,
making oscillations less likely.

\subsubsection{\em Hands On: DQN  and Breakout Gym Example}
To get some hands-on experience with DQN, we will now have a
look at how DQN can be used to play the Atari game Breakout.

The field of deep reinforcement
learning is an open field where most codes of algorithms are freely
shared on GitHub and where test environments are
available. The most widely used environment is Gym, in which benchmarks such as ALE and MuJoCo can be
found, see also Appendix~\ref{ch:env}. 
The open availability of the software
allows
for easy replication, and, importantly, for further
improvement of the methods.
Let us have a closer look at the code of DQN, to experience how it works.

The DQN papers come with source code. The original DQN code
from~\cite{mnih2015human} is available at 
\href{https://github.com/kuz/DeepMind-Atari-Deep-Q-Learner}{Atari
DQN}.\footnote{\url{https://github.com/kuz/DeepMind-Atari-Deep-Q-Learner}}
This code is the original code, in the programming language Lua, which
may be interesting to study, if you are familiar with this language. A
modern reference implementation of DQN, with further improvements,
is in the
(stable) baselines.\footnote{\url{https://stable-baselines.readthedocs.io/en/master/index.html}} The RL Baselines Zoo even provides a collection of
pretrained agents, 
at \href{https://github.com/araffin/rl-baselines-zoo}{Zoo}~\cite{plappert2016kerasrl,geron2017hands}.\footnote{\url{https://github.com/araffin/rl-baselines-zoo}}
The Network Zoo is especially useful if your desired application
happens to be in 
the Zoo, to prevent long  training times. 

%

\subsubsection*{Install Stable Baselines}\index{stable baselines}

The environment is only
half of the reinforcement learning experiment, we also need an
agent algorithm to learn the policy. OpenAI also provides
implementations of  \emph{agent}
algorithms, called the Baselines, at the Gym
GitHub repository \href{https://github.com/openai/baselines}{Baselines}.\footnote{\url{https://github.com/openai/baselines}} Most
algorithms that are covered in this book are present. You can
download them, study the code, and experiment to gain an insight 
into their behavior.

\lstset{label={lst:stable}}
\lstset{caption={Running Stable Baseline PPO on the Gym Cartpole Environment}}
\begin{lstlisting}[language=Python,float]
import gym

from stable_baselines.common.policies import MlpPolicy
from stable_baselines.common.vec_env import DummyVecEnv
from stable_baselines import PPO2

env = gym.make('CartPole-v1')

model = PPO2(MlpPolicy, env, verbose=1)
model.learn(total_timesteps=10000)

obs = env.reset()
for i in range(1000):
    action, _states = model.predict(obs)
    obs, rewards, dones, info = env.step(action)
    env.render()
\end{lstlisting}

In addition to OpenAI's Baselines, there is \emph{Stable} Baselines, a
fork of the OpenAI 
algorithms; it has more documentation and other features. 
It can be
found at \href{https://github.com/hill-a/stable-baselines}{Stable
  Baselines},\footnote{\url{https://github.com/hill-a/stable-baselines}}
and the documentation is at
\href{https://stable-baselines.readthedocs.io/en/master/}{docs}.\footnote{\url{https://stable-baselines.readthedocs.io/en/master/}}

The stable release from the Stable Baselines is installed
by typing
\begin{tcolorbox}
  \verb|pip install stable-baselines|
\end{tcolorbox}
  or
  \begin{tcolorbox}
    \verb|pip install stable-baselines[mpi]|
\end{tcolorbox}
\noindent if support for OpenMPI is desired (a parallel
message passing implementation for cluster computers).
A very quick check to see if everything works is to run the PPO trainer
from Listing~\ref{lst:stable}. PPO is a policy-based algorithm that
will be discussed in the next chapter in Sect.~\ref{sec:ppo}.
The Cartpole should appear again, but should now learn to stabilize
for a brief moment.

\subsubsection*{The DQN Code}

After having studied tabular Q-learning on Taxi in
Sect.~\ref{sec:taxi-q}, let us now see how the network-based DQN works
in practice.  Listing~\ref{lst:dqn} illustrates how easy it is to use the 
Stable Baselines implementation of DQN on the Atari Breakout
environment. (See Sect.~\ref{sec:gym} for installation instructions of
Gym.)

\lstset{label={lst:dqn}}
\lstset{caption={Deep Q-Network Atari Breakout example with Stable Baselines}}
\begin{lstlisting}[language=Python,float]
from stable_baselines.common.atari_wrappers import make_atari
from stable_baselines.deepq.policies import MlpPolicy, CnnPolicy
from stable_baselines import DQN

env = make_atari('BreakoutNoFrameskip-v4')

model = DQN(CnnPolicy, env, verbose=1)
model.learn(total_timesteps=25000)

obs = env.reset()
while True:
    action, _states = model.predict(obs)
    obs, rewards, dones, info = env.step(action)
    env.render()
    
\end{lstlisting}

After you have run the DQN code and seen that it works, it is worthwhile
to study how the code is implemented. Before you dive into the Python
implementation of Stable Baselines, let us look at the pseudocode to
refresh how the elements of DQN work together. See
Listing~\ref{lst:dqn-pseudo}. In this pseudocode we follow the 2015
version of DQN~\cite{mnih2015human}.  (The 2013 version of DQN did not
use the target network~\cite{mnih2013playing}.)

\lstset{label={lst:dqn-pseudo}}
\lstset{caption={Pseudocode for DQN, after \cite{mnih2015human}}}
\begin{lstlisting}[language=Python,float]
def dqn:
    initialize replay_buffer empty
    initialize Q network with random weights
    initialize Qt target network with random weights
    set s = s0
    while not convergence:
        # DQN in Atari uses preprocessing; not shown
        epsilon-greedy select action a in argmax(Q(s,a)) # action selection depends on Q (moving target)
        sx,reward = execute action in environment
        append (s,a,r,sx) to buffer
        sample minibatch from buffer # break temporal correlation
        take target batch R (when terminal) or Qt
        do gradient descent step on Q # loss function uses target Qt network


        
        
\end{lstlisting}

DQN is based on Q-learning, with as extra a replay buffer and a
target network to improve stability and convergence. 
First, at the start of the code, the replay buffer is initialized to empty,
and the weights of the Q network and the separate Q target network are
initialized. The state $s$ is set to the start state.

Next is the optimization loop, that runs until convergence. At the start
of each iteration an action is selected at the state $s$, following an
$\epsilon$-greedy approach. The action is executed in the environment,
and the new state and the reward are stored in a tuple in the replay
buffer. Then, we   train the Q-network. A minibatch is sampled
randomly from the replay buffer, and one gradient descent step is
performed. For this step the loss function is calculated with the
separate Q-target network $\hat Q_\theta$, that is updated  less frequently than
the primary Q-network $Q_\theta$. In this way the loss function
$$\mathcal{L}_t(\theta_t)=\mathbb{E}_{s,a\sim\rho(\cdot)}\Big[\big( \mathbb{E}_{s'\sim\mathcal{E}}(r+\gamma\max_{a'} \hat{Q}_{\theta_{t-1}}(s',a')|s,a) - Q_{\theta_{t}}(s,a)\big)^2\Big]$$
is  more stable, causing better
convergence;   $\rho(s,a)$ is  the behavior distribution over $s$ and $a$,
and $\mathcal{E}$ is the Atari emulator~\cite{mnih2013playing}. 
Sampling the minibatch reduces the correlation that is inherent in
reinforcement learning between subsequent states. 

\subsubsection*{Conclusion}
In summary, DQN was able to successfully learn
end-to-end behavior policies for many different games (although similar and from the
same benchmark set). Minimal prior knowledge was
used to guide the system, and the agent only got to see the pixels and the
game score. The same network architecture and procedure was
used on each game; however,  a network trained for one game could not
be used to play another game. 

The DQN achievement was an important milestone in the history of
deep reinforcement learning.
The main problems that were overcome by Mnih et al.~\cite{mnih2013playing} were training
divergence and learning
instability. 

The nature of most Atari 2600 games is that they require eye-hand
reflexes. The games have some strategic elements,  
credit assignment is mostly over a short term, and can  be learned with a surprisingly simple
neural  network. Most Atari games are more about immediate reflexes
than about longer term reasoning. In this sense, the problem of playing Atari well is not
unlike an image  categorization problem: both problems are to
find the right response  that matches an input
consisting of a set of pixels. Mapping pixels to
categories is not that different from mapping pixels to joystick
actions (see also the observations in~\cite{karpathy2016pong}). 

The Atari results 
have stimulated much
subsequent research. Many blogs 
have been written on reproducing the result, which is  not a straightforward
task, requiring the fine-tuning of many  hyperparameters~\cite{openaibaseline2017}.

\subsection{Improving Exploration}\index{rainbow algorithms}
\index{convergence}\label{sec:rainbow}

The DQN
results have spawned much activity among reinforcement learning
researchers to improve training stability and convergence further, and many
refinements have 
been devised, some of which we will  review in this section.

Many of the topics that are covered by the enhancements are older
ideas that work well in deep reinforcement learning. DQN applies
random sampling of its replay buffer, and one of the first enhancements was prioritized
sampling~\cite{schaul2015prioritized}. It was found that DQN, being an
off-policy algorithm, typically overestimates action values (due to
the max operation, Sect.~\ref{sec:off-policy}). Double DQN addresses
overestimation~\cite{van2016deep}, and dueling DDQN introduces the
advantage function to standardize action
values~\cite{wang2016dueling}. 
Other approaches look at variance in
addition to expected value, the effect of random noise on
exploration was tested~\cite{fortunato2017noisy}, and distributional
DQN showed that networks that use probability distributions work
better than networks that only use single point expected
values~\cite{bellemare2017distributional}.

\begin{table}[t]
  \begin{center}

\begin{tabular}{lccc}
   {\bf Name} & {\bf Principle} & {\bf Applicability} & {\bf Effectiveness}  \\ \hline\hline
DQN \cite{mnih2013playing} & replay buffer & Atari & stable Q learning \\
  Double DQN \cite{van2016deep}& de-overestimate values & DQN  & convergence  \\ 
  Prioritized experience~\cite{schaul2015prioritized}  & decorrelation & replay buffer  & convergence  \\ 
  Distributional \cite{bellemare2017distributional} & probability
                                                      distr & stable gradients  & generalization  \\  
  Random noise \cite{fortunato2017noisy} & parametric noise & stable gradients  & more exploration 
  \\ \hline
\end{tabular}

  \caption{Deep Value-Based Approaches}\label{tab:rainbow}
  \end{center}
\end{table}
In 2017 Hessel et al.~\cite{hessel2017rainbow} performed a large experiment that combined seven
important enhancements. They found that the enhancements worked well
together. The paper has become known as  the Rainbow
paper, since the major
graph showing the cumulative performance over 57 Atari games of the
seven enhancements is 
multi-colored (Fig.~\ref{fig:rainbow}).
Table~\ref{tab:rainbow}
summarizes the   enhancements, 
%
and this section provides an overview of the main ideas.  The
enhancements were tested on  the same  benchmarks (ALE, Gym), and most
algorithm implementations can be found on the OpenAI Gym
\href{https://github.com/openai/baselines}{GitHub} site in the 
baselines.\footnote{\url{https://github.com/openai/baselines}}

\begin{figure}[t]
\begin{center}
\includegraphics[width=8cm]{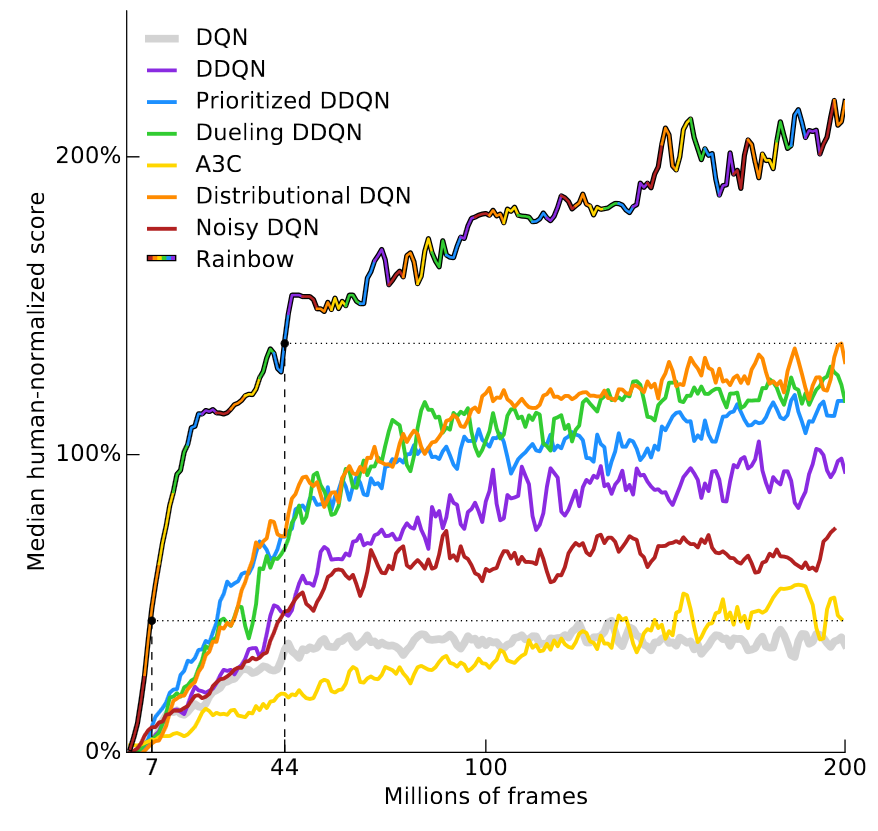}
\caption{Rainbow graph: performance over 57 Atari games~\cite{hessel2017rainbow}}\label{fig:rainbow}
\end{center}
\end{figure}

\subsubsection{Overestimation}
\index{DDQN}
Van Hasselt et al.\ introduce double deep Q learning
(DDQN)~\cite{van2016deep}. \gls{DDQN} is based on the observation that
Q-learning may overestimate action
values.  On the  Atari 2600 games DQN suffers from substantial
over-estimations.
Remember that DQN uses Q-learning. Because of the $\max$ operation in Q-learning this  results in an overestimation of the Q-value. To
resolve this issue, DDQN  uses the Q-Network to choose
the action but uses the separate target Q-Network to evaluate the
action.
Let us compare the training target for DQN
$$y=r_{t+1}+\gamma Q_{\theta_t}(s_{t+1},\argmax_a Q_{\theta_t}(s_{t+1},a)$$
with the training target for DDQN (the difference is a single $\phi$)
$$y=r_{t+1}+\gamma Q_{\phi_t}(s_{t+1},\argmax_a Q_{\theta_t}(s_{t+1},a).$$
The DQN target uses the same set of weights $\theta_t$ twice, for
selection and evaluation; the DDQN target use a separate set of
weights $\phi_t$ for evaluation, preventing overestimation due to
the max operator. Updates are assigned randomly to either set  of weights.

Earlier Van Hasselt et al.~\cite{hasselt2010double} introduced the
double Q learning algorithm in a tabular setting. The later paper
shows that this idea also works with a large deep network. They report
that the DDQN algorithm not only reduces the overestimations but  also leads to 
better performance on several games.
DDQN was tested on 49 Atari games and achieved about twice the
average score of DQN with the same hyperparameters, and four times the
average DQN score with tuned
hyperparameters~\cite{van2016deep}.

\subsubsection*{Prioritized Experience Replay}
DQN samples uniformly  over the entire history in the replay buffer,
where Q-learning uses only the most recent (and important) state. It stands to reason
to see if a solution in between these two extremes performs well.

Prioritized experience replay, or \gls{PEX},\index{prioritized DDQN} is such
an attempt. It was
introduced by Schaul et al.~\cite{schaul2015prioritized}. In the
Rainbow paper PEX is combined with  DDQN, and, as we can see, the blue
line (with PEX) indeed outperforms the purple line.

In DQN experience replay lets agents reuse examples
from the past, although  experience transitions are uniformly
sampled, and actions are simply replayed at the 
same frequency that they were originally experienced, regardless of
their significance. The PEX approach provides a
framework for prioritizing 
experience. Important actions are replayed more frequently, and therefore learning
efficiency is improved.
As measure of importance, standard proportional prioritized replay is used, with the
absolute TD error  to prioritize actions. Prioritized replay is used
widely in value-based deep reinforcement learning.
The measure can be
computed in the distributional setting using the mean action
values. In the Rainbow paper all distributional
variants prioritize actions by the Kullback-Leibler loss~\cite{hessel2017rainbow}.

\subsubsection*{Advantage Function}\index{advantage function}
The original DQN uses a single neural network as function
approximator; DDQN (double deep Q-network) uses a separate target
Q-Network to evaluate an action. Dueling DDQN\index{dueling
  DDQN}~\cite{wang2016dueling}, also known as \gls{DDDQN},
improves on this architecture by using two separate estimators: a value function
and an advantage function $$A(s,a)=Q(s,a)-V(s).$$ Advantage functions
are related to the actor-critic approach (see Chap.~\ref{chap:policy}). An advantage function computes the
difference between the value of an action and the value of the
state. The function  standardizes values on a baseline for the actions of a
state~\cite{grondman2012survey}. 
Advantage functions provide better policy evaluation when    many
actions have similar values.

\subsubsection{Distributional Methods}\label{sec:distrdqn}\label{sec:variance}
 The original DQN learns a single value,
which is the estimated mean of the state
value.\index{distributional DQN} This approach  does not take 
uncertainty  into account. 
To remedy this, distributional Q-learning~\cite{bellemare2017distributional} learns a
categorical probability distribution of discounted returns
instead, increasing exploration.  Bellemare et al.\ design a new
distributional  algorithm which applies
Bellman's equation to the learning of 
distributions, a method called distributional DQN.    Moerland et
al.~\cite{moerland2017efficient,moerland2018potential} propose
uncertain value networks.
Interestingly, a link between the  distributional approach and biology
has been reported. Dabney et
al.~\cite{dabney2020distributional}  showed correspondence between distributional
reinforcement learning algorithms and the dopamine levels in mice,
suggesting that the brain represents possible future rewards  as a probability distribution.

\subsubsection*{Noisy DQN}
Another distributional method is noisy DQN\index{noisy
  DQN}~\cite{fortunato2017noisy}. Noisy DQN  uses stochastic network
layers that add parametric noise to the weights.  The noise induces randomness in the 
agent's policy, which increases exploration. The
parameters that govern the noise are learned by gradient descent together with
the remaining network weights.  In their experiments the standard
exploration heuristics for A3C (Sect.~\ref{sec:a3c}), DQN, and 
dueling agents (entropy reward and $\epsilon$-greedy) were replaced with
NoisyNet. The increased exploration yields substantially higher scores
for  Atari (dark red line).

\section{Atari 2600 Environments}


In their original 2013
workshop paper Mnih et al.~\cite{mnih2013playing} 
achieved human-level play for some of the games. 
Training was performed on  50 million frames in total on seven
Atari games. 
The neural network performed
better than an expert human player on Breakout, Enduro, and Pong. On
Seaqest, Q*Bert, and Space Invaders performance was far below that of
a human. In these games a strategy must be found that extends over
longer time periods. In their follow-up journal  article 
two  years later they were able to achieve human level play for 49 of the 57
games that are in ALE~\cite{mnih2015human}, and performed better than
human-level play in 29 of the 49   games. 

Some of the games still proved  difficult, notably  games
that require  longer-range planning, where long stretches of the
game do not give rewards, such as in Montezuma's Revenge, where the
agent has to walk long distances, and pick up a key to reach new rooms
to enter new
levels. In reinforcement learning terms, delayed credit assignment
over long periods is  
hard. Towards the end of the book we will see Montezuma's Revenge
again, when we discuss hierarchical reinforcement learning methods, in
Chap~\ref{chap:hier}. These methods are specifically developed to take
large steps in the state space. The Go-Explore algorithm was able to
solve Montezuma's Revenge~\cite{ecoffet2019go,ecoffet2021first}.\index{Montezuma's revenge}\label{sec:montezuma}

\subsection{Network Architecture}
End-to-end learning of challenging problems is computationally
intensive. In addition to the two algorithmic innovations, the success
of DQN is also due to the creation of a  specialized efficient training
architecture~\cite{mnih2015human}.

Playing the Atari games  is a computationally intensive task for a
deep neural network: the network trains a
behavior policy directly from 
pixel frame input. 
Therefore, the  training architecture contains  reduction steps. 
To start with, the network  consists of only three hidden
layers (one fully connected, two convolutional), which is simpler than what is used in most supervised learning
tasks.

\begin{figure}[t]
\begin{center}
\includegraphics[width=12cm]{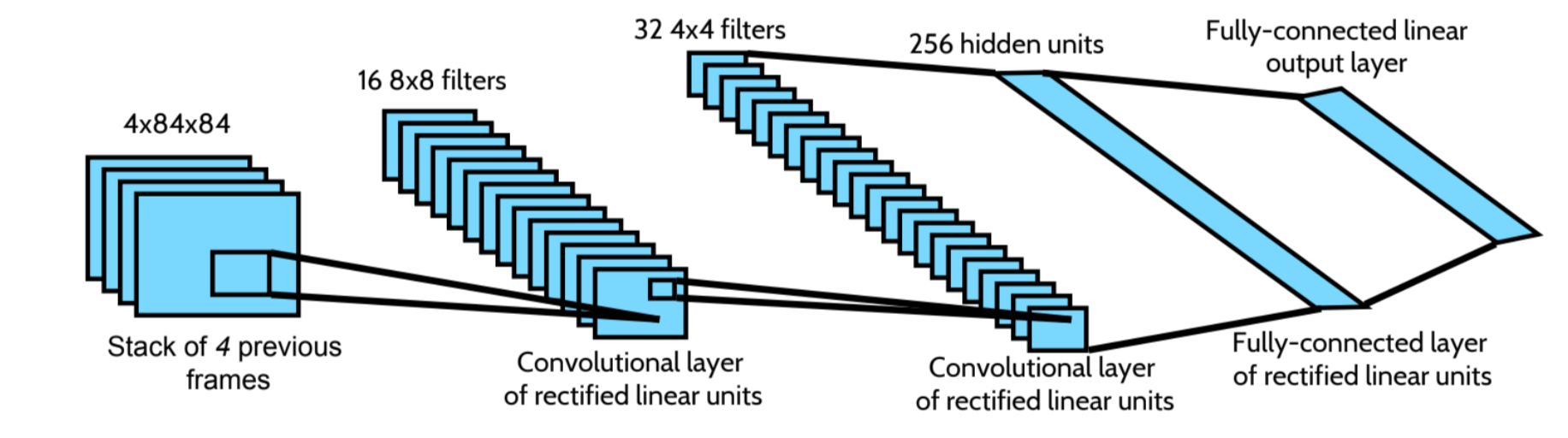}
\caption{DQN architecture \cite{hui2018rl}}\label{fig:dqnarch}
\end{center}
\end{figure}

The pixel-images are high-resolution data. Since working with the full resolution
of $210\times 160$ pixels of $128$ color-values at 60 frames per second would be
 computationally too
intensive,  the images are reduced in resolution. The 
$210\times 160$ with a $128$ color palette is reduced to gray scale and
$110\times 84$ pixels, which is further cropped to $84\times 84$. The first hidden layer convolves
16 $8\times 8$ filters with stride 4 and ReLU neurons. The second hidden layer
convolves 32 $4\times 4$ filters with stride 2 and ReLU neurons. The third
hidden layer is fully connected and consists of 256 ReLU neurons. The
output layer is also fully connected with one output per action (18 joystick
actions). The outputs correspond to the  Q-values of
the individual action. Figure~\ref{fig:dqnarch} shows the architecture
of DQN. The network receives the change in game score
as a number from the emulator, and derivative updates are mapped to
$\{-1,0,+1\}$ to indicate 
decrease, no change, or improvement of the score (the Huber
loss~\cite{openaibaseline2017}). 

To reduce
computational demands further, frame skipping is employed. Only one in
every 3--4 frames was used, depending on the game.
To take game history into account, the net takes as input the last
four resulting frames. This allows movement to be seen by the net.
As optimizer RMSprop is used~\cite{ruder2016overview}.
A variant of $\epsilon$-greedy is used, that starts with an $\epsilon$
of $1.0$ (fully exploring) going down to $0.1$ (90\% exploiting).


\subsection{Benchmarking Atari}
To end the Atari story, we discuss two final algorithms. Of the many  value-based model-free deep reinforcement learning
algorithms that have been developed, one more algorithm
that we discuss is
R2D2\index{R2D2}~\cite{kapturowski2018recurrent}, because of its performance. R2D2 is not part of
the Rainbow experiments, but is a significant further improvement of
the algorithms. R2D2  stands for Recurrent Replay Distributed DQN. It
is built upon 
prioritized distributed replay and $5$-step double
Q-learning. Furthermore, it
uses a dueling network architecture and an LSTM layer after the
convolutional stack. Details about the architecture can be found
in~\cite{wang2016dueling,gruslys2017reactor}. The LSTM uses the 
recurrent state to exploit long-term temporal dependencies, which improve
performance. The authors also report  that the LSTM allows for better
representation learning. R2D2 achieved 
good results on all 57 Atari games~\cite{kapturowski2018recurrent}. 

A more recent benchmark  achievement has been published as
Agent57. Agent57 is the first  program that achieves a score higher than the
human baseline on all 57 Atari 2600 games from ALE. It uses a
controller that adapts the long and short-term behavior of the agent,
training for a range of policies, from very exploitative to very
explorative, depending on the game~\cite{badia2020agent57}.\index{Agent57}  

\subsubsection*{Conclusion}

Progress has come a long way since the replay buffer of
DQN. Performance has been improved greatly in value-based model-free
deep reinforcement learning and now super-human performance in all 57
Atari games of ALE has been achieved.  Many enhancements that improve
coverage, correlation, and convergence have been developed. The
presence of a clear benchmark was instrumental for 
progress so that researchers could clearly see which ideas worked and
why.  The earlier mazes and navigation games, OpenAI's
Gym~\cite{brockman2016openai}, and especially 
the ALE~\cite{bellemare2013arcade}, have enabled this progress.

In the next chapter we will look at the other main branch of
model-free reinforcement learning: policy-based 
algorithms. We will see how they work, and  that they are well suited for a different kind
of application, with continuous action spaces. 

\section*{Summary and Further Reading}
\addcontentsline{toc}{section}{\protect\numberline{}Summary and
  Further Reading}
This has been the first chapter in which we have seen  deep
reinforcement learning algorithms learn complex, high-dimensional, tasks. We
end with a summary and pointers to the literature.

\subsection*{Summary}

The methods that have been discussed in the previous chapter were exact, tabular
methods. 
Most interesting problems
have large state spaces that do not fit into memory. Feature learning
identifies  states by their common features. Function values are not calculated
exactly, but are approximated, with deep learning.

Much of the recent success of reinforcement learning is due to deep learning
methods. 
%
For reinforcement learning a problem arises when states are
approximated. Since in reinforcement learning the next state is
determined by the previous state, algorithms may get stuck in local
minima or run in circles when
values are shared with different states.

Another problem is training convergence.
Supervised learning has a static dataset and training targets are
also static. In reinforcement learning the loss function targets
depend on the parameters that are being optimized. This causes further
instability. DQN 
caused a breakthrough by showing that with a replay buffer and a
separate, more stable, target network, enough stability could be
found for DQN to converge and learn how to play Atari arcade games.

Many further improvements to increase stability 
have been found. The Rainbow paper implements some of these
improvements, and finds that they are complementary, and together
achieve very strong play.

\subsection*{Further Reading}
Deep learning revolutionized reinforcement learning. A comprehensive
overview of the field is provided by Dong et al.~\cite{dong2020deep}. For more on deep
learning, see Goodfellow et al.~\cite{goodfellow2016deep}, a book with much detail
on deep learning; a major journal article
is~\cite{lecun2015deep}. A brief survey
is~\cite{arulkumaran2017deep}. Also see Appendix~\ref{ch:deep}.

In 2013 the Arcade Learning
Environment was
presented~\cite{bellemare2013arcade,machado2018revisiting}. Experimenting
with reinforcement learning was made even more accessible with
OpenAI's Gym~\cite{brockman2016openai}, with  clear and easy to use
Python bindings.

Deep learning versions of value-based tabular algorithms suffer from convergence and
stability problems~\cite{tsitsiklis1997analysis}, yet the idea that stable deep reinforcement learning might be
practical took hold with~\cite{heess2013actor,sallans2004reinforcement}. 
Zhang et
al.~\cite{zhang2017deeper} study the deadly triad  with experience
replay.
Deep gradient TD methods were proven to converge
for evaluating a fixed policy~\cite{bhatnagar2009convergent}.
Riedmiller et al. relaxed the fixed
control policy in  neural fitted Q learning algorithm
(NFQ)~\cite{riedmiller2005neural}. NFQ builds on work on 
stable function approximation~\cite{gordon1995stable,ernst2005tree} and
experience replay~\cite{lin1992self}, and more recently on
least-squares policy iteration~\cite{lagoudakis2003least}.
In 2013 the first DQN paper appeared, showing results on a small
number of Atari games~\cite{mnih2013playing} with the replay buffer to
reduce temporal correlations.
In 2015 the followup 
Nature paper reported results in more games~\cite{mnih2015human}, with
a separate target network to improve training
convergence. A well-known overview paper
is the Rainbow paper~\cite{hessel2017rainbow,justesen2019deep}.

 The use of benchmarks is  of great importance for
reproducible reinforcement learning
experiments~\cite{henderson2018deep,islam2017reproducibility,khetarpal2018re,hutson2018artificial}.
For TensorFlow and Keras,
see~\cite{chollet2015keras,geron2017hands}.

\section*{Exercises}
\addcontentsline{toc}{section}{\protect\numberline{}Exercises}
 We will end
this chapter with some questions to review the concepts that we have
covered. Next are programming exercises to get some more exposure on
how to use the deep reinforcement learning algorithms in practice.

\subsubsection*{Questions}
Below are some questions to check your understanding of this
chapter. Each question is a closed question where a simple, single
sentence answer is expected.

\begin{enumerate}
\item What is Gym?
\item What are the Stable Baselines?

\item The loss function of DQN uses the Q-function as
  target. What is a consequence? 
\item Why is the exploration/exploitation trade-off central in
  reinforcement learning?
\item Name one simple exploration/exploitation method.
\item What is bootstrapping?
\item Describe the architecture of the neural network in DQN.
\item Why is deep reinforcement learning more susceptible to unstable
  learning than deep supervised learning?
\item What is the deadly triad?
\item How does function approximation reduce stability of Q-learning?
\item What is the role of the replay buffer?
\item How can correlation between states lead to local minima?
\item Why should the coverage of the state space be sufficient?
  \item What happens when deep reinforcement learning algorithms do
    not  converge?

        \item How large is the state space of chess estimated to be? $10^{47}$, $10^{170}$
  or $10^{1685}$?
\item How large is the state space of Go estimated to be? $10^{47}$, $10^{170}$
  or $10^{1685}$?
\item How large is the state space of StarCraft estimated to be? $10^{47}$, $10^{170}$
  or $10^{1685}$?

\item What does the rainbow in the Rainbow paper stand for, and what is the main
  message?
  \item Mention three Rainbow improvements that are added to DQN.

\end{enumerate}

\subsubsection*{Exercises}
Let us now start with some exercises. 
If you have not done so already, install Gym, PyTorch\footnote{\url{https://pytorch.org}} or TensorFlow and Keras
(see Sect.~\ref{sec:gym} and \ref{sec:keras} or go to the TensorFlow
page).\footnote{\url{https://www.tensorflow.org}}
Be sure to check  the right versions of Python, Gym, TensorFlow, and the
Stable Baselines to make sure that they work well together.
The exercises
below are designed to be done with Keras.

\begin{enumerate}
\item {\em DQN} Implement DQN from the Stable Baselines on Breakout from
  Gym. Turn off Dueling and Priorities. Find out what the values are for $\alpha$, the training rate, for
  $\epsilon$, the exploration rate, what kind of neural network
  architecture is used, what the replay buffer size is, and how
  frequently the target network is updated.
\item {\em Hyperparameters} Change all those hyperparameters, up, and down, and note the effect on
  training speed, and the training outcome: how good is the result?
  How sensitive is performance to hyperparameter optimization?
  \item {\em Cloud} Use different computers, experiment with GPU versions to speed
    up training, consider Colab, AWS, or another cloud provider with
    fast GPU (or TPU)  machines.
    \item {\em Gym} Go to Gym and try different problems. For what kind of
      problems does DQN work, what are characteristics of problems
      for which it works less well?
\item {\em Stable Baselines} Go to the Stable baselines and implement different agent
  algorithms. Try Dueling algorithms, Prioritized experience replay,
  but also other algorithm, such as Actor critic or
  policy-based. (These algorithms will be explained in the next
  chapter.)
  Note their performance.
\item {\em Tensorboard} With Tensorboard you can follow the training
  process as it progresses. Tensorboard works on log files. Try
  TensorBoard on a Keras exercise and  follow different training
  indicators. Also try TensorBoard on the Stable Baselines and see
  which indicators you can follow.
\item {\em Checkpointing} Long training runs in Keras need checkpointing,
  to save valuable computations in case of a hardware or software
  failure. Create a large training job, and setup checkpointing. Test
  everything by interrupting the training, and try to re-load the
  pre-trained checkpoint to restart the training where it left off. 
\end{enumerate}


\chapter{Policy-Based Reinforcement Learning}\label{ch:pol}\label{chap:policy}
Some of the most successful applications of deep
reinforcement learning have a continuous action space, such as
applications in robotics,  self-driving cars, and real-time strategy
games.

The previous chapters introduced value-based  reinforcement
learning. Value-based methods  find the policy in a two-step
process. First they find the best action-value of a state, for which then the
accompanying actions are  found (by means of $\argmax$).
This works in 
environments with  discrete actions, where the highest-valued action is clearly
separate from the next-best action. 
Examples of  continuous action
spaces  are robot arms that can move over arbitrary 
angles, or poker bets that can be any monetary value. In these
action spaces value-based methods become unstable and 
$\argmax$ is not appropriate.

Another approach works
better: policy-based methods. Policy-based methods do not use a separate value function but find
the policy directly. They start with a policy function, which they
then improve, episode by episode, with  policy gradient methods. 
Policy-based methods are applicable to more domains than
value-based methods. They   work well with deep neural networks
and gradient learning; they are some of the
most popular methods of deep reinforcement learning, and this chapter
introduces you to them.

We start by 
looking at   applications with continuous action spaces.
Next, we look at policy-based agent algorithms. We will introduce basic policy
search algorithms, 
and the policy gradient theorem. We will also discuss  algorithms that combine  value-based and
policy-based  approaches: the so-called Actor critic algorithms. 
At the end of the chapter we discuss larger  environments for
policy-based methods in more depth, where we will discuss progress in
visuo-motor robotics and locomotion environments.

The chapter  concludes with  exercises, a summary, and pointers
to further reading.

\section*{Core Concepts}
\begin{itemize}
\item Policy gradient
\item Bias-variance trade-off; Actor critic
\end{itemize}

\section*{Core Problem}
\begin{itemize}
\item  Find a low variance continuous action policy directly
\end{itemize}

\section*{Core Algorithms}
\begin{itemize}
\item REINFORCE (Alg.~\ref{alg:pg})
\item Asynchronous advantage actor critic  (Alg.~\ref{alg_pg_bootstrapping_baseline})
\item Proximal policy optimization  (Sect.~\ref{sec:ppo})
\end{itemize}

\section*{Jumping Robots}
  


One of the most
intricate problems in robotics is
learning to walk, or more generally, how to perform locomotion. Much work
has been put into making robots walk, run and jump. A
 video of a simulated robot that taught itself to jump over an obstacle course
can be found on YouTube\footnote{\url{https://www.youtube.com/watch?v=hx_bgoTF7bs}}~\cite{heess2017emergence}. 

Learning to walk is a challenge that takes human infants months to
master. (Cats and dogs are quicker.) Teaching robots
to walk is  a challenging problem that is studied
extensively in artificial intelligence and engineering. Movies abound on the internet of robots that try to open
doors, and fall over, or just try to stand upright, and still fall
over.\footnote{See, for example,
  \url{https://www.youtube.com/watch?v=g0TaYhjpOfo}.}

Locomotion of legged robots is a difficult sequential decision
problem. For each leg, many different joints are
involved. They  must be actuated 
in the right order, turned 
with the right force, over the right duration, to the right angle.  Most of these angles,
forces, and durations are continuous. The algorithm
has to decide  how many degrees, Newtons, and seconds, constitute the
optimal policy. All these actions are continuous quantities.
Robot locomotion is a difficult problem, that is studied frequently in policy-based deep reinforcement learning. 

\section{Continuous  Problems}
In this chapter, our actions  are continuous, and stochastic. We
will  discuss  both of these aspects, and
 some of the challenges they pose. We will start with continuous
action policies.

\subsection{Continuous Policies}
In the previous chapter we discussed environments with large state
spaces. We will now move our attention to action spaces.
The action spaces of the problems that we have seen so far---Grid
worlds, mazes, and high-dimensional Atari games---were actually  action
spaces that were small and discrete---we could walk north, east, west,
south, or we could choose from 9 joystick movements.
In board games such as chess the action space is larger, but still
discrete. When you move your 
pawn to e4, you do not move it  to e4\textonehalf.

In this chapter the problems are different. Steering a self driving
car requires turning the steering wheel a certain angle, duration, and
angular velocity, to prevent jerky movements. Throttle movements
should also be smooth and
continuous. Actuation of robot joints is  continuous, as we
mentioned in the introduction of this chapter.
An arm joint can move 1 degree, 2
degrees, or 90 or 180 degrees or anything in between.

An action in a  continuous space is not one of a set of discrete
choices, such as $\{N, E, W, S\}$, but
rather a value over a continuous range, such as $[0,2\pi]$ or
$\mathbb{R}^+$; the number of possible values is 
infinite.
How can we find the optimum value in an infinite space in a finite
amount of time? Trying out all  possible combinations of setting
joint 1 to $x$ degrees and applying force $y$ in motor 2 will take infinitely
long.  
A  solution could be to discretize the actions, although that introduces 
potential  quantization errors. 

When  actions are not discrete,  the $\argmax$
operation can not be used to identify ``the''  best action, and
value-based methods are no longer sufficient. Policy-based
methods find suitable continuous or  stochastic policies directly, without the
intermediate step of a value function and the need for the $\argmax$
operation to construct the final policy.

\subsection{Stochastic Policies}
We will now turn to the modeling of stochastic policies.

When a robot moves
its hand to open a 
door, it must judge the distance correctly. A small
error, and  it may fail (as many
movie clips show).\footnote{Even 
  worse, when a robot thinks it stands still, it may actually be in
  the process of   falling   over (and, of course, robots can
  not think, they only wished they could).}  
Stochastic environments cause stability  problems for value-based
methods~\cite{lillicrap2015continuous}.
Small  perturbations 
in Q-values may lead  to large changes in the
policy of value-based methods. Convergence 
can typically only be 
achieved at slow learning rates, to smooth out the randomness. A
stochastic policy (a target distribution) does not suffer from this problem.
Stochastic policies have another advantage.  By their nature they
perform exploration, without the need to  separately code $\epsilon$-greediness
or other exploration methods,
since a stochastic policy returns a distribution over actions.

Policy-based
methods find suitable stochastic policies directly. A potential
disadvantage of purely episodic
policy-based methods is that they are high-variance; they may find 
local optima instead of global optima, and converge slower than
value-based methods.
Newer (actor critic) methods, such as A3C, TRPO, and PPO, were designed to overcome
these problems. We will discuss these algorithms  later in this chapter.

Before we will explain these  policy-based agent algorithms, we will have a closer
look
at some of the applications for which they are needed.

\subsection{Environments: Gym and MuJoCo}\label{sec:mujoco}\index{MuJoCo}\index{PyBullet}
Robotic experiments play an important role in reinforcement learning. However, because of the cost
associated with real-world robotics 
experiments, in reinforcement learning often simulated robotics
systems are used. This is especially important in model-free methods, that tend to
have a high sample complexity (real robots  wear down
when trials run in the  millions). 
These software simulators
model  behavior of the robot and the effects on the environment, using  physics
models. This prevents the
expense of real experiments with real robots, although some precision
is lost to  modeling error. Two well-known physics models  are \gls{MuJoCo}~\cite{todorov2012mujoco} and
PyBullet~\cite{coumans2019}. They can be used easily via the Gym environment.
\index{MuJoCo}\index{PyBullet}

\subsubsection{Robotics}

\begin{figure}[t]
  \begin{center}
    \begin{tabular}{cc}
      \includegraphics[width=5cm]{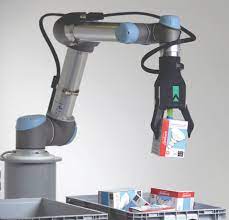}
      &
      \includegraphics[width=5cm]{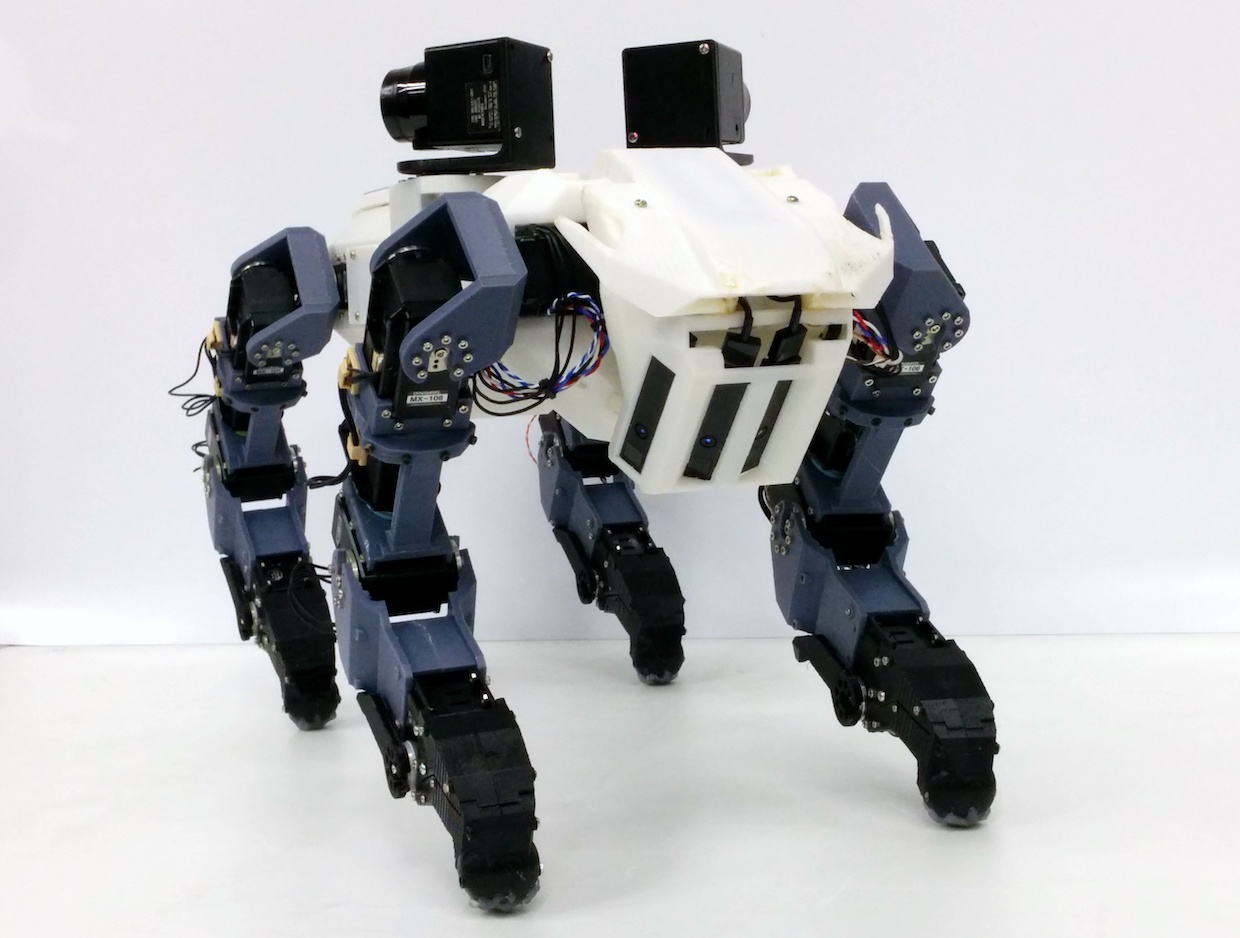}
    \end{tabular}
    \caption{Robot Grasping and Gait \cite{martinelli2019}}\label{fig:robot-arm}
 \end{center}
\end{figure}

Most robotic applications are more complicated than the classics such
as mazes, Mountain car and Cart pole. Robotic control
decisions involve more joints,  directions of
travel, and degrees of freedom,  than a single cart that moves in one
dimension. Typical  problems involve learning of visuo-motor 
skills (eye-hand coordination, grasping), or learning of different locomotion
gaits of multi-legged ``animals.'' Some examples of grasping and
walking are illustrated in Fig.~\ref{fig:robot-arm}.

The environments for these actions are unpredictable to a certain
degree: they require  reactions to disturbances such as bumps
in the road, or the moving of objects in a scene.


%

\subsubsection{Physics Models }
Simulating robot motion  involves modeling  forces, acceleration,
velocity, and movement. It also includes modeling mass and elasticity for bouncing
balls, tactile/grasping mechanics, and the effect of different materials. A
physics mechanics model  needs to simulate the result of
actions in the real world. 
Among the goals of such a simulation is to model grasping,  locomotion, gaits,
and walking and running (see also Sect.~\ref{sec:loco}). 

The simulations should be accurate.
Furthermore, since model-free learning  algorithms often
involve millions of actions,  it is 
important that the physics simulations are fast.
Many different physics environments for model-based robotics have been created, among them
Bullet, Havok, ODE and PhysX, see~\cite{erez2015simulation} for a
comparison. Of the models, 
MuJoCo~\cite{todorov2012mujoco}, and
PyBullet~\cite{coumans2019} are the most popular in
reinforcement learning, especially MuJoCo is used in many
experiments. 

Although MuJoCo calculations are deterministic,  the initial state
of environments is typically randomized, resulting in an overall
non-deterministic environment.
Despite  many code optimizations in MuJoCo, simulating physics is still an
expensive proposition. Most MuJoCo experiments  in the
literature therefore are 
based on stick-like entities, that simulate limited motions, in order
to limit the computational demands.

\begin{figure}[t]
\begin{center}
  \begin{tabular}{cc}
    \includegraphics[height=3cm]{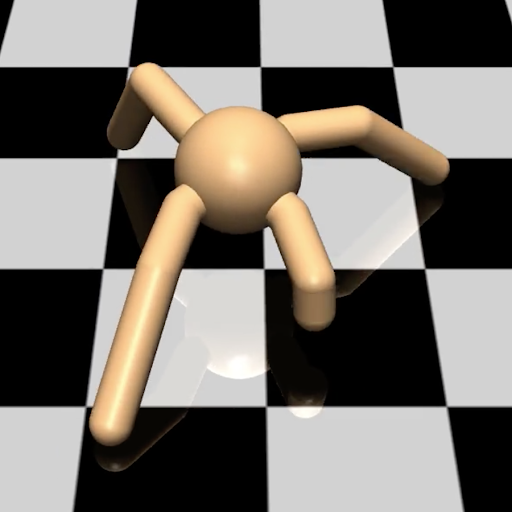}
    &
      \includegraphics[height=3cm]{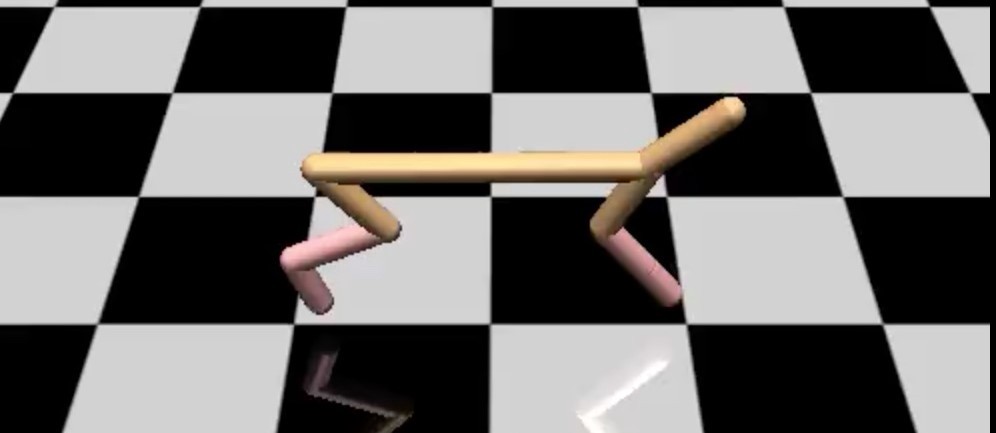}
  \end{tabular}
  \caption{Gym MuJoCo Ant and Half-Cheetah \cite{brockman2016openai}}\label{fig:ant}
\end{center}
\end{figure}


\begin{figure}[t]
\begin{center}
\includegraphics[width=5cm]{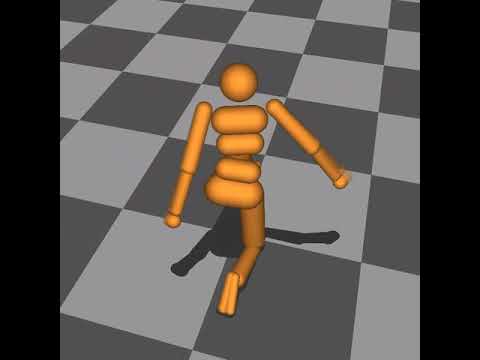}
\caption{Gym MuJoCo Humanoid}\label{fig:humanoid}
\end{center}
\end{figure}

Figures~\ref{fig:ant} and \ref{fig:humanoid} illustrate a few examples of
some of the  common Gym/MuJoCo problems that are 
often used in reinforcement learning: Ant, Half-cheetah, and Humanoid.

\subsubsection{Games}
In  real time video games and certain card games the decisions are
also continuous. For example, in some variants of poker, the size
of monetary bets can be any amount, which makes the action space quite
large (although strictly speaking still discrete). In games such as
StarCraft and Capture the Flag, aspects of  the physical world are
modeled, and movement of agents can vary in duration and speed. The environment
for these games is also stochastic: some information is hidden for the
agent. This increases the size of the state space greatly. We will discuss
these games in Chap.~\ref{chap:multi} when we discuss  multi-agent methods.

\section{Policy-Based Agents}

Now that we have discussed the problems and environments that are used
with policy-based methods, it is time to see how policy-based
algorithms work.
\begin{table}[t]
  \begin{center}
  \begin{tabular}{lll}
    {\bf Name} & {\bf Approach}  &  {\bf Ref}  \\
    \hline\hline
    REINFORCE & Policy-gradient optimization& \cite{williams1992simple}\\
    A3C & Distributed Actor Critic &  \cite{mnih2016asynchronous}\\
    DDPG & Derivative of continuous action function  & \cite{lillicrap2015continuous}\\
    TRPO & Dynamically sized step size & \cite{schulman2015trust} \\
    PPO & Improved TRPO, first order & \cite{schulman2017proximal}\\
    SAC & Variance-based Actor Critic for robustness & \cite{haarnoja2018soft}\\
    \hline
  \end{tabular}
  \caption[Policy-based Algorithms]{Policy-Based Algorithms:
    REINFORCE, Asynchronous Advantage Actor Critic, Deep Deterministic
  Policy Gradient, Trust Region Policy Optimization, Proximal Policy
  Optimization, Soft Actor Critic}\label{tab:pol}
\end{center}
\end{table}
%
Policy-based methods are a popular approach in model-free deep reinforcement
learning. Many algorithms have been developed that perform
well. Table~\ref{tab:pol} lists some of the better known algorithms that will be
covered in this chapter.

We will first provide an intuitive explanation of the
idea behind the basic policy-based approach.  Then we will discuss
some of  the theory behind it, as well as advantages and
disadvantages of the basic policy-based approach. Most of these disadvantages are alleviated by the actor
critic method, that is discussed next.

Let us start with the basic idea behind policy-based methods.

\subsection{Policy-Based Algorithm: REINFORCE}\label{sec:reinforce}


Policy-based approaches learn a parameterized policy, that selects
actions without consulting a value function.\footnote{Policy-based methods may use a value
function to \emph{learn} the policy parameters $\theta$, but do not use it for
\emph{action selection}. 
}
In policy-based methods the policy function is represented directly,
allowing   policies to select a continuous action,
something that is difficult to do  in value-based methods.

\begin{tcolorbox}
{\bf The Supermarket:} To build some intuition on the nature of policy-based methods, let us
think back again at the supermarket navigation task, that we used 
 in Chap.~\ref{chap:feedback}. In this navigation problem we can  try to
assess our current distance  to the supermarket with the
Q-value-function, as we have done before.  The Q-value  assesses the distance of 
each direction to take; it tells us how far each
action is from the goal. We can then use this distance function to
find our path.

In contrast, the policy-based
alternative would be to ask a local the way, who tells us, for example, to go
straight and then left and then right at the Opera House 
and  straight until we  reach the supermarket on our left. The
local just gave us  a full path to follow, without having to
infer which action was the closest and then use that information to
determine the way to go. We can subsequently try to improve this full trajectory.
\end{tcolorbox}


\noindent Let us see how we can optimize  such a direct policy directly, without the
intermediate step of the Q-function.
%
%
%
%
%
%
%
%
We will  develop a  first, generic, policy-based algorithm to see how the pieces 
fit together. The explanation will be intuitive in nature.

The basic framework for policy-based algorithms is 
straightforward. We start with a parameterized policy function
$\pi_\theta$. We first (1) initialize the parameters $\theta$ of the
policy function, (2) sample
a new trajectory $\tau$,  (3) if $\tau$ is a good trajectory, increase  the
parameters $\theta$ towards $\tau$, otherwise decrease them, and (4) keep going until
convergence. Algorithm~\ref{alg_grad_based_opt} provides a framework
in pseudocode. 
Please note the similarity with the codes in the previous chapter
(Listing~\ref{lst:train_sl}--\ref{lst:train_rl}), and especially the
deep learning algorithms, where we also 
optimized function parameters in a  loop. 

The  policy is represented by a set of
parameters $\theta$ (these can be the weights in a neural network). Together, the parameters $\theta$ map the states $S$ to
action probabilities  $A$. When we are given a set of parameters, how should we
adjust them to improve the policy? The basic idea is to randomly sample a new
policy, and if it is better, adjust the parameters a bit in the
direction of this new policy (and away if it is worse). Let us see in
more detail how this idea works. 

To know which policy is best, we need some kind of measure of its quality.
We   denote the quality of the policy that is defined by the
parameters as $J(\theta)$. It is natural to use the value function of
the start state as our measure of  quality 
$$J(\theta)=V^\pi(s_0).$$
We wish to maximize $J(\cdot)$. When the parameters are differentiable, then
all we need to do is to find a way to improve the gradient
$$\nabla_\theta J(\theta)=\nabla_\theta V^\pi(s_0)$$
of this expression to maximize
our objective function $J(\cdot)$.

Policy-based methods  apply gradient-based optimization, using the derivative of the
 objective to find the optimum. Since we are maximizing, we  apply
 gradient \emph{ascent}.  In each time step $t$ of the
 algorithm we perform the following update: 
 $$ \theta_{t+1} = \theta_t + \alpha \cdot \nabla_\theta J(\theta) $$
 for learning rate $\alpha \in \mathbb{R}^+ $ and performance
 objective  $J$, see the 
 gradient ascent algorithm in Alg.~\ref{alg_grad_based_opt}. 
 \begin{algorithm}[t]
   \caption{Gradient ascent optimization} \label{alg_grad_based_opt}
   \begin{algorithmic}
 \State{{\bf Input:} a differentiable objective $J(\theta)$, learning rate $\alpha \in \mathbb{R}^+$, threshold $\epsilon \in \mathbb{R}^+$}
  \State {\bf Initialization}: randomly initialize $\theta$ in $\mathbb{R}^d$ 
  \Repeat
  \State Sample trajectory $\tau$ and compute gradient $\nabla_\theta$
  \State $\theta \gets \theta + \alpha \cdot \nabla_\theta J(\theta)$ 
  \Until {$\nabla_\theta J(\theta)$ converges below $\epsilon$}
 \State \Return  parameters $\theta$
  \end{algorithmic}
 \end{algorithm}

 Remember that $\pi_\theta(a|s)$ is the probability of taking action $a$ in
state $s$. This function $\pi$ is represented by a 
 neural network $\theta$, mapping states $S$ at the input side of
the network to action  probabilities on the output side of the
network. The parameters $\theta$ 
determine the mapping of our function $\pi$.  Our goal is to update the
parameters so that $\pi_\theta$ becomes the optimal policy. The better the
action $a$ is, the more we want to increase the parameters $\theta$.

If we now would know, by some magical way,  the optimal action $a^\star$,
then we could use the gradient to push each parameter $\theta_t, t
\in$ trajectory, of the policy,  in the direction of
the optimal action, as follows
$$\theta_{t+1} = \theta_t+\alpha\nabla\pi_{\theta_t}(a^\star|s).$$
Unfortunately, we do not know which action is  best. We can,
however, take a sample trajectory and use  estimates of the value of
the actions of the sample. This estimate can 
use the regular $\hat{Q}$ function from the previous chapter, or the
discounted return function, or an advantage function (to be introduced shortly). 
Then, by multiplying the push of the parameters (the probability)
with  our estimate, we get
$$\theta_{t+1} = \theta_t+\alpha\hat{Q}(s,a)\nabla\pi_{\theta_t}(a|s) .$$
A problem with this formula is that not only are we going to push
harder on actions with a high value, but also more often, because the
policy $\pi_{\theta_t}(a|s)$ is the probability of action $a$ in state
$s$. Good actions are thus doubly improved, which may cause instability.
We can correct by dividing by the general probability:
$$\theta_{t+1} = \theta_t+\alpha\hat{Q}(s,a)\frac{\nabla\pi_{\theta_t}(a|s)}{\pi_\theta(a|s)}.$$
In fact, we have  now almost arrived at the classic policy-based
algorithm, REINFORCE, introduced by Williams in 1992~\cite{williams1992simple}. In this algorithm our formula is
expressed in a way that  is reminiscent of  a logarithmic cross-entropy loss
function. We can arrive at such a log-formulation by using the basic fact from calculus
that $$\nabla\log f(x)=\frac{\nabla f(x)}{f(x)}.$$  Substituting this
formula into our  equation, we  arrive at
$$\theta_{t+1} =
\theta_t+\alpha\hat{Q}(s,a)\nabla_\theta\log\pi_{\theta}(a|s).$$
This formula is indeed the core of \gls{REINFORCE}, the prototypical
policy-based algorithm, which is shown in full in Alg.~\ref{alg:pg},
 with discounted cumulative reward.

To summarize, the REINFORCE formula pushes the parameters of the policy in the
direction of the better action (multiplied proportionally by the size of
the estimated action-value) to know which action is best.

We have arrived at a method to improve a policy that can be used
directly to indicate the action to take. The method whether the action is
discrete, continuous, or stochastic, without having to go through 
intermediate value or $\argmax$ functions to find it. Algorithm~\ref{alg:pg}
shows the full algorithm, which is called \emph{Monte Carlo policy gradient}. The algorithm is called \emph{Monte Carlo}
because it samples a  trajectory.

\begin{algorithm}[t]
\caption{Monte Carlo policy gradient (REINFORCE)~\cite{williams1992simple}} \label{alg:pg} 
\begin{algorithmic}
\State{{\bf Input:} A differentiable policy $\pi_\theta(a|s)$,
   learning rate $\alpha \in \mathbb{R}^+$, threshold $\epsilon \in
   \mathbb{R}^+$}
  \State {\bf Initialization}:  Initialize parameters $\theta$ in $\mathbb{R}^d$
  \Repeat
 	\State Generate full trace $\tau = \{s_0,a_0,r_0,s_1,..,s_T\}$ following $\pi_\theta(a|s)$ 
	\For{$t \in 0,\ldots,T-1$} \Comment{Do for each step of the episode}
 		\State $R \gets \sum^{T-1}_{k=t} \gamma^{k-t} \cdot r_k$ \Comment{Sum Return from trace}
                \State $\theta \gets \theta + \alpha\gamma^t R \nabla_\theta\log\pi_\theta(a_t|s_t)$ \Comment{Adjust parameters}
	\EndFor
  \Until {$\nabla_\theta J(\theta)$ converges below $\epsilon$}
 \State \Return  Parameters $\theta$
\end{algorithmic}
\end{algorithm}

\subsubsection*{Online and Batch}\index{online updates}\index{batch updates}
The versions of gradient ascent (Alg.~\ref{alg_grad_based_opt}) and
REINFORCE (Alg.~\ref{alg:pg}) that we show, update the parameters inside the
innermost loop. All updates are performed as the time steps of the
trajectory are traversed. This method is called the \emph{online}
approach. When multiple processes work in parallel to update data, the
online approach makes sure that information is used as soon as it is
known.

The policy gradient algorithm can also be formulated  in
\emph{batch}-fashion: all gradients are summed over the states and
actions, and the parameters are updated at the end of the
trajectory.
Since parameter updates can be expensive, the batch
approach can be more efficient. An intermediate form that is
frequently applied in practice is to work with \emph{mini-batches},
trading off computational efficiency for information efficiency.

Let us now take a step back and look at the algorithm and assess how well
it works.

\subsection{Bias-Variance Trade-Off in  Policy-Based Methods}  
Now that we have seen the principles  behind a
policy-based algorithm, let us see how policy-based algorithms
work in practice, and compare  advantages and disadvantages of
the policy-based  approach. 

Let us start with the advantages.
First of all, parameterization is at the core of policy-based methods,
making them a good match for deep learning.
For value-based methods deep
learning had to be retrofitted, giving rise to complications as we saw
in Sect.~\ref{sec:stable}. Second, policy-based methods can easily find 
stochastic policies; value-based methods find  deterministic
policies.
Due to their stochastic nature, policy-based methods
naturally explore, without the need for methods such as
$\epsilon$-greedy, or more involved methods, that may require tuning
to work well.
Third, policy-based methods are effective in large or
continuous action spaces. 
Small changes in $\theta$ lead to small changes in $\pi$, and to small
changes in state distributions (they are smooth).
Policy-based algorithms do not suffer (as much) from convergence and stability
issues that are seen in $\argmax$-based 
algorithms in large or continuous action spaces.

On the other hand, there are disadvantages to the episodic Monte Carlo version
of the REINFORCE algorithm. Remember  that REINFORCE generates a full
random episode
in each iteration, before it assesses the quality. (Value-based
methods use a reward to select the next action in each time
step of the episode.) Because of this, policy-based is low bias, since full random
trajectories are generated. However, they are also high variance, 
since the full trajectory is generated randomly (whereas value-based
uses the value for guidance at each selection step). What are the consequences?
First,  policy evaluation of full trajectories has
low sample efficiency and high variance. As a consequence, 
policy improvement happens infrequently, leading to slow convergence
compared to value-based methods. Second, this approach often finds  a local
optimum, since  convergence to the global optimum takes too long.

Much research has been performed to address the high variance of the episode-based
vanilla policy
gradient~\cite{barto1983neuronlike,konda1999actor,konda2000actor,grondman2012survey}. The
enhancements that have been found have greatly improved 
performance, so much so that 
policy-based approaches---such as A3C, PPO, SAC, DDPG---have become  favorite model-free
reinforcement learning algorithms for many applications.
The enhancements to reduce high
variance  that we discuss are:
\begin{itemize}
\item \emph{Actor critic} introduces within-episode value-based 
  critics based on temporal difference value bootstrapping;
\item \emph{Baseline subtraction} introduces an advantage function to
  lower variance;
\item \emph{Trust regions}  reduce large policy
  parameter changes;
\item \emph{Exploration} is crucial to get out of
  local minima and  for more robust result;  high
  entropy action distributions are often used.
\end{itemize}
Let us have a look at these enhancements.

\subsection{Actor Critic Bootstrapping}\index{actor critic}\index{baseline
  function}\index{advantage function}

The actor critic approach combines value-based elements with the
policy-based method.
 The actor stands for the action, or
policy-based, approach; the critic stands for the value-based
approach~\cite{sutton2018introduction}. 

Action selection in episodic  REINFORCE is random, and hence  low
bias. However,  variance is high, since the full episode is sampled
(the size and direction of the update can strongly vary between different
samples).
The actor critic approach is designed to combine the advantage of the value-based
approach (low variance)
with the advantage of the policy-based approach (low bias). 
Actor critic
methods  are popular because they work well. It is an active field
where many different algorithms
have been developed.

The  variance of policy methods can originate from two sources: (1) high variance in
the cumulative reward estimate, and (2) high variance in the gradient
estimate.
For both problems a solution has been developed:  bootstrapping for
better reward estimates, and 
baseline subtraction to lower the variance of gradient estimates. Both
of these methods use the learned value 
function, which we denote by $V_\phi(s)$. The value function can use a
separate neural network, with separate parameters \gls{phi}, or it can
use a value head on top of the actor parameters $\theta$. In this case
the actor and the critic share the lower layers of the network, and
the network has two separate top heads: a policy and a value head. We
will use $\phi$ for the parameters of the value function, to
discriminate them from the policy parameters \gls{theta}.
\label{sec_actor_critic}

\subsubsection*{Temporal Difference Bootstrapping}
To reduce the variance of the policy gradient, we can increase the
number of traces $M$ that we sample. However, the possible number of
different traces is 
exponential in the length of the trace
for a given
stochastic policy, and we cannot afford to  sample them all for one
update. In practice  the number of sampled
traces  $M$ is small, 
sometimes even  $M=1$, updating the policy parameters from a single
trace. The return of the trace depends on many random action choices;
 the update has  high variance. A solution is to use a principle that
 we known from temporal difference learning,  to bootstrap the value
 function step by step. Bootstrapping uses 
the value function to compute intermediate $n$-step values per episode,
trading-off variance for bias. The $n$-step values are in-between
full-episode Monte Carlo and single step temporal difference targets.

\begin{algorithm}[t]
\caption{Actor critic  with  bootstrapping} \label{alg_pg_bootstrapping}
\begin{algorithmic}
\State{\bf Input:} A policy $\pi_\theta(a|s)$, a value function
$V_\phi(s)$
\State An estimation depth $n$, learning rate $\alpha$, number of episodes $M$
\State {\bf Initialization}: Randomly initialize $\theta$ and $\phi$
 \Repeat
	\For{$i \in 1,\ldots,M$}
 	\State Sample trace $\tau = \{s_0,a_0,r_0,s_1,..,s_{T}\}$ following $\pi_\theta(a|s)$ 
	\For{$t \in 0,\ldots,T-1$}
		\State $\hat{Q}_{n}(s_t,a_t) =  \sum_{k=0}^{n-1}
                \gamma^k \cdot r_{t+k} + \gamma^n \cdot V_\phi(s_{t+n})$ \Comment{$n$-step target}
	\EndFor	
	\EndFor
	\State $\phi \gets \phi - \alpha \cdot \nabla_\phi \sum_t \big(\hat{Q}_{n}(s_t,a_t) - V_\phi(s_t)\big)^2 $ \Comment{Descent value loss}
	\State $\theta \gets \theta + \alpha \cdot \sum_t [ \hat{Q}_{n}(s_t,a_t) \cdot \nabla_\theta \log \pi_\theta(a_t|s_t)]$ \Comment{Ascent policy gradient}
	  \Until {$\nabla_\theta J(\theta)$ converges below $\epsilon$}
\State \Return Parameters $\theta$
\end{algorithmic}
\end{algorithm}

We can use bootstrapping to compute an $n$-step target
\begin{equation*}
\hat{Q}_\text{n}(s_t,a_t) =  \sum_{k=0}^{n-1} r_{t+k} + V_\phi(s_{t+n}),
\end{equation*}
and we can then update the value function, for example on a squared loss
\begin{equation*}
\mathcal{L}(\phi|s_t,a_t) = \big(\hat{Q}_{n}(s_t,a_t) - V_\phi(s_t)\big)^2
\end{equation*}
and update the policy with the standard policy gradient but with that
(improved) value $\hat{Q}_n$
\begin{equation*}
\nabla_\theta \mathcal{L}(\theta|s_t,a_t) =\hat{Q}_{n}(s_t,a_t) \cdot
\nabla_\theta \log \pi_\theta(a_t|s_t) .
\end{equation*}
We are now using the value function prominently in
the algorithm, which is parameterized by a separate set of parameters,
denoted by $\phi$; the policy parameters are still denoted by
$\theta$. The use of both policy and value is what gives the  actor
critic approach its name.

An example algorithm is shown in Alg.~\ref{alg_pg_bootstrapping}. When
we compare this algorithm with Alg.~\ref{alg:pg}, we see how the
 policy gradient ascent update now uses the $n$-step $\hat{Q}_n$ value
estimate instead of the trace return $R$. We also see that this
time the parameter updates are in batch mode, with  separate summations.

\subsection{Baseline Subtraction with Advantage Function}

Another method to reduce the variance of the policy gradient is by
baseline subtraction. Subtracting a baseline from a set of
numbers reduces the variance, but leaves the expectation
unaffected. Assume, in a 
given state with three available actions, that we sample action returns of
65, 70, and 75, respectively.  Policy gradient will then try to push the probability of each action up,
since the return for each action is positive. The above method may
lead to a problem, since we are pushing all actions up (only somewhat harder on
one of them). It might be better if we only push up on actions that are
higher than the average (action 75 is higher than the average of 70 in this example), and push down on actions that are below
average (65 in this example). We can do so through {\it baseline subtraction}.

The most common choice for the baseline is the value function. When we
subtract the value $V$ from a state-action value estimate $Q$, the function is
called the {\it  advantage function}: 
$$ A(s_t,a_t) = Q(s_t,a_t) - V(s_t). $$
The \gls{advantage} function subtracts
the value of the state $s$ from the state-action value. It now estimates how much better a
particular action is compared to the expectation of
a particular state.

We can combine baseline subtraction
with any bootstrapping method to estimate the cumulative reward $\hat{Q}(s_t,a_t)$. We  compute 
$$ \hat{A}_\text{n}(s_t,a_t) = \hat{Q}_n(s_t,a_t)  - V_\phi(s_t) $$
and update the policy with 
$$
\nabla_\theta \mathcal{L}(\theta|s_t,a_t) =\hat{A}_{n}(s_t,a_t) \cdot
\nabla_\theta \log \pi_\theta(a_t|s_t) .
$$
We have now seen the ingredients to construct a full actor critic algorithm. 
An example algorithm is shown in Alg.~\ref{alg_pg_bootstrapping_baseline}. 

\begin{algorithm}[t]
\caption{Actor critic with bootstrapping and baseline  subtraction} \label{alg_pg_bootstrapping_baseline}
\begin{algorithmic}
\State{\bf Input:} A policy $\pi_\theta(a|s)$, a value function
$V_\phi(s)$
\State An estimation depth $n$, learning rate $\alpha$, number of episode $M$
\State {\bf Initialization}: Randomly initialize $\theta$ and $\phi$
 \While{not converged}
	\For{$i = 1, \ldots ,M$}
 	\State Sample trace $\tau = \{s_0,a_0,r_0,s_1,..,s_{T}\}$ following $\pi_\theta(a|s)$ 
	\For{$t = 0, \ldots, T-1$}
		\State $\hat{Q}_{n}(s_t,a_t) =  \sum_{k=0}^{n-1}
                \gamma^k \cdot r_{t+k} + \gamma^n \cdot V_\phi(s_{t+n})$ \Comment{$n$-step target}	
		\State $\hat{A}_\text{n}(s_t,a_t) = \hat{Q}_n(s_t,a_t)  - V_\phi(s_t)$ \Comment{Advantage}
	\EndFor	
	\EndFor
	\State $\phi \gets \phi - \alpha \cdot \nabla_\phi \sum_t \big(\hat{A}_{n}(s_t,a_t)\big)^2 $ \Comment{Descent Advantage loss}
	\State $\theta \gets \theta + \alpha \cdot \sum_t [ \hat{A}_{n}(s_t,a_t) \cdot \nabla_\theta \log \pi_\theta(a_t|s_t)]$ \Comment{Ascent policy gradient}
	\EndWhile	
\State \Return Parameters $\theta$
\end{algorithmic}
\end{algorithm}

\subsubsection*{Generic Policy Gradient Formulation}

With these two ideas we can  formulate an entire spectrum
of policy gradient methods, depending on the type of cumulative reward
estimate that they use. In general, the policy gradient estimator takes the
following form, where we now introduce a new target $\Psi_t$ that we sample
from the trajectories $\tau$: 
$$
\nabla_\theta J(\theta) =  \mathbb{E}_{\tau_0 \sim p_\theta(\tau_0)} \Big[
\sum_{t=0}^n \Psi_t  \nabla_\theta \log \pi_\theta(a_t|s_t)
\Big] 
$$
There is a variety of potential choices for $\Psi_t$, based on the
 use of bootstrapping and baseline substraction:  

\begin{alignat}{3}
\Psi_t &= \hat{Q}_{MC}(s_t,a_t) &&= \sum_{i=t}^\infty \gamma^i \cdot r_i & \text{Monte Carlo target} \nonumber \\
\Psi_t &= \hat{Q}_{n}(s_t,a_t)  &&= \sum_{i=t}^{n-1} \gamma^i \cdot r_i + \gamma^n V_\theta(s_n) & \text{bootstrap ($n$-step target)} \nonumber \\
\Psi_t &= \hat{A}_{MC}(s_t,a_t) &&= \sum_{i=t}^\infty \gamma^i \cdot r_i - V_\theta(s_t) & \text{baseline subtraction} \nonumber \\
\Psi_t &= \hat{A}_{n}(s_t,a_t) &&= \sum_{i=t}^{n-1} \gamma^i \cdot r_i
+ \gamma^n V_\theta(s_n) - V_\theta(s_t) & \text{baseline + bootstrap}
\nonumber \\ 
\Psi_t &= Q_\phi(s_t,a_t) && & \text{Q-value approximation} \nonumber \\ \nonumber
\end{alignat}
Actor critic algorithms are among the most popular model-free reinforcement
learning algorithms in practice, due to their good performance. After
having discussed relevant theoretical background, it is time to look
at how actor critic can be implemented in a practical, high
performance, algorithm. We will start with A3C.

\subsubsection*{Asynchronous Advantage Actor Critic}\label{sec:a3c}\index{A3C}

Many high
performance implementations are based on the actor critic
approach. For large problems the algorithm is typically parallelized
and implemented on a large cluster computer. A well-known parallel
algorithm is Asynchronous advantage actor critic (A3C). \gls{A3C} is  a  framework that uses
asynchronous (parallel and distributed) 
gradient descent for 
optimization of deep neural network
controllers~\cite{mnih2016asynchronous}. 

There is also a non-parallel version of A3C, the synchronous
variant \gls{A2C}~\cite{wu2017scalable}. Together they popularized this
approach to actor critic methods. 
\begin{figure}[t]
\begin{center}
\includegraphics[width=8cm]{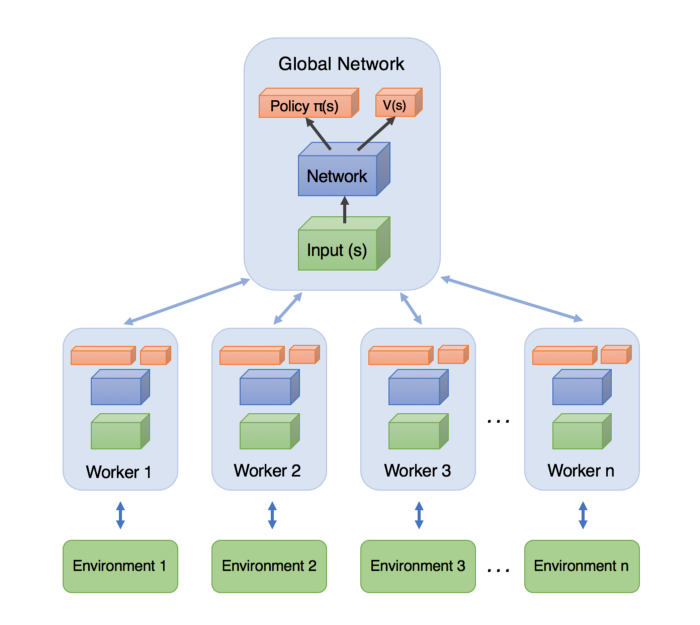}
\caption{A3C network~\cite{juliani2016simple}}\label{fig:a3c}
\end{center}
\end{figure}
Figure~\ref{fig:a3c} shows the distributed architecture of A3C~\cite{juliani2016simple};
Alg.~\ref{alg:a3c} shows the pseudocode, from Mnih et al.~\cite{mnih2016asynchronous}. The A3C network
will estimate both a value function $V_\phi(s)$ and an 
advantage function $A_\phi(s,a)$, as well as a policy function
$\pi_\theta(a|s)$. In the  experiments on
Atari~\cite{mnih2016asynchronous}, the neural networks were 
separate  fully-connected 
policy and value heads at the top (orange in Fig.~\ref{fig:a3c}), followed by joint
convolutional networks (blue). This network architecture
is replicated over the distributed workers.
Each of these workers are run on a separate
processor thread and are synced with global parameters from time to
time.

\begin{algorithm}[t]
\caption{Asynchronous advantage actor-critic pseudocode for each
  actor-learner thread~\cite{mnih2016asynchronous} }
\begin{algorithmic}
\small
\State {\bf Input:} Assume global shared parameter vectors $\theta$ and $\phi$ and global shared counter $T=0$
\State Assume thread-specific parameter vectors $\theta'$ and $\phi'$
\State Initialize thread step counter $t\gets 1$
\Repeat
\State Reset gradients: $d\theta \gets 0$ and $d\phi \gets 0$.
\State Synchronize thread-specific parameters  $\theta'=\theta$ and $\phi'=\phi$ %
\State $t_{start} = t$
\State Get state $s_t$
\Repeat
\State Perform $a_t$ according to policy $\pi (a_t|s_t;\theta')$
\State Receive reward $r_t$ and new state $s_{t+1}$
\State $t \gets t + 1$
\State $T \gets T + 1$
\Until terminal $s_t$ \textbf{or} $t-t_{start}==t_{max}$
\State $R =
    \left\{
    \begin{array}{l l}
      0  \quad & \text{for terminal } s_t\\
        V(s_t,\phi') \quad & \text{for non-terminal } s_t \text{ // Bootstrap from last state}
    \end{array}\right.$
\For {$i \in \{t-1,\ldots,t_{start} \}$}
\State $R \gets r_i + \gamma R$
\State Accumulate gradients wrt $\theta'$: $d\theta \gets d\theta + \nabla_{\theta'} \log\pi(a_i|s_i;\theta') (R - V(s_i;\phi'))$
\State Accumulate gradients wrt $\phi'$: $d\phi \gets d\phi + {\partial\left(R - V(s_i;\phi')\right)^2}/{\partial \phi'}$
\EndFor
\State Perform asynchronous update of $\theta$ using $d\theta$ and of $\phi$ using $d\phi$.
\Until $T > T_{max}$
\end{algorithmic}
\label{alg:a3c}
\end{algorithm}

A3C improves on classic REINFORCE in the following ways: it uses an
advantage actor critic design, it uses deep learning, and it makes
efficient use of parallelism in the training stage.  The gradient
accumulation step at the end of the code can be considered as a parallelized
reformulation of minibatch-based stochastic gradient update: the values
of $\phi$ or $\theta$ are adjusted  in the direction of each
training thread independently. 
A major contribution of A3C comes from its parallelized and
asynchronous architecture: multiple actor-learners are dispatched to
separate instantiations of the environment; they all interact with
the environment and collect experience, and asynchronously push their
gradient updates to a central target network (just as 
DQN).

It was found that the parallel actor-learners have a stabilizing
effect on training. A3C surpassed the previous
state-of-the-art on the Atari domain and succeeded on a wide variety
of continuous motor control problems as well as on a new task of
navigating random 3D mazes using high-resolution visual
input~\cite{mnih2016asynchronous}.

\subsection{Trust Region Optimization}

\index{TRPO}\index{trust region policy optimization}
Another important approach to further reduce the variance of policy methods  is the trust region
approach.
Trust region policy
optimization (\gls{TRPO}) aims to further reduce the high variability in the policy
parameters, by using a special loss function with an additional
constraint on the optimization problem~\cite{schulman2015trust}. 

A naive approach to speed
up an algorithm is to try to increase the step 
size of  hyperparameters, such as the learning rate, and the policy parameters.
This approach will fail to uncover
solutions that are hidden in finer grained trajectories, and the
optimization will converge to  local optima. For this reason the step
size should not be too large. A less naive approach is
to use an adaptive 
step size that depends on the \emph{output} of the optimization
progress. 

Trust regions are used in general  optimization problems to constrain the
update size~\cite{sun2006optimization}. The algorithms work by computing the
quality of the approximation; if it is still good, then the trust
region is expanded. Alternatively, the region can be shrunk  if the
divergence of the new and current policy is getting large. 

Schulman et 
al.~\cite{schulman2015trust} introduced trust region policy
optimization (TRPO) based on this ideas, trying to take the largest
possible parameter improvement
step on a policy, without  accidentally causing performance to collapse.


To this end, as it samples policies, TRPO compares the old and the new
policy:
$$\mathcal{L}(\theta)=\mathbb{E}_t\Big[\frac{\pi_\theta(a_t|s_t)}{\pi_{\theta_{\text{old}}}(a_t|s_t)}
\cdot A_t\Big].$$
In order to increase the learning step size, TRPO tries to maximize this loss function $\mathcal{L}$, subject to the constraint that
the old and the new policy are not too far away. In TRPO the
Kullback-Leibler divergence\footnote{The Kullback-Leibler divergence is a measure of 
distance between probability
distributions~\cite{kullback1951information,bishop2006pattern}.} is used for this purpose:
$$\mathbb{E}_t[{\text{KL}}(\pi_{\theta_{\text{old}}}(\cdot|s_t),\pi_\theta(\cdot|s_t))]
\leq \delta.$$

TRPO   scales to complex high-dimensional
problems. Original experiments demonstrated its 
robust performance on   simulated robotic Swimming, Hopping,
Walking gaits, and  Atari games. TRPO
is  commonly used in experiments and as a  baseline for developing new
algorithms.
A disadvantage of TRPO is  that it is  a complicated algorithm that
uses second order derivatives; 
we will not cover the pseudocode here. 
Implementations can be found at Spinning Up\footnote{\url{https://spinningup.openai.com}} and Stable
Baselines.\footnote{\url{https://stable-baselines.readthedocs.io}}

\label{sec:ppo}\index{PPO}\index{proximal policy optimization}
Proximal policy optimzation (\gls{PPO})~\cite{schulman2017proximal} was
developed as an improvement of TRPO. PPO has some of the benefits of TRPO, but is  simpler to
implement, is more general, has better empirical sample complexity and
has better run time complexity.
It is motivated by the same question as TRPO, to take the
largest possible improvement step on a policy parameter  without  causing
performance collapse. 

There are two  variants of PPO: PPO-Penalty and PPO-Clip.
PPO-Penalty approximately solves a KL-constrained update (like TRPO),
but merely penalizes the KL-divergence in the objective function instead of
making it a hard constraint. 
PPO-Clip does not use a KL-divergence term in the objective and has no
constraints either. Instead it relies on 
clipping in the objective function to remove incentives for the new
policy to get far from the old policy; 
it clips the difference between
the old and the new policy within a fixed range
$[1-\epsilon, 1+\epsilon]\cdot A_t$.

While simpler than TRPO,  PPO is still a  complicated algorithm to
implement, and we omit the code here. The authors of PPO provide an
implementation as a
\href{https://openai.com/blog/openai-baselines-ppo/\#ppo}{baseline}.\footnote{\url{https://openai.com/blog/openai-baselines-ppo/\#ppo}}
Both TRPO and PPO are on-policy algorithms.
Hsu et al.~\cite{hsu2020revisiting} reflect on design choices of PPO.

\subsection{Entropy and Exploration}\index{entropy}
A problem in many deep reinforcement learning experiments where only a
fraction of the state space is sampled, is brittleness: the algorithms
get stuck in local optima, and different
choices for hyperparameters can cause large differences in
performance. Even a different choice for seed for the random number
generator can cause large differences in performance for many algorithms.\index{random number seed}\index{brittleness}

For large problems, exploration is important, in value-based and
policy-based approaches alike.  We must
provide the incentive to sometimes try 
an action which currently seems suboptimal~\cite{moerland2021lecture}.
Too little exploration results in brittle, local, optima. 

When we learn a \emph{deterministic} policy $\pi_\theta(s)\to a$,  we can
manually add exploration noise to the behavior policy. In a continuous action space we can
use Gaussian noise, while in a discrete action space we can use
Dirichlet noise~\cite{kotz2004continuous}. For example, in a 1D continuous action space we
could use: 
$$ \pi_{\theta,\text{behavior}}(a|s) = \pi_\theta(s) + \mathcal{N}(0,\sigma), $$
where $\mathcal{N}(\mu,\sigma)$ is the Gaussian (normal) distribution with
hyperparameters mean $\mu=0$ and standard deviation $\sigma$; $\sigma$
is our exploration hyperparameter.  

\subsubsection*{Soft Actor Critic}\index{SAC}\index{soft actor critic}
When we learn a \emph{stochastic} policy $\pi(a|s)$, then exploration is
already partially ensured due to the stochastic nature of our
policy. For example, when we predict a Gaussian distribution, then
simply sampling from this distribution will already induce variation
in the chosen actions.  
$$ \pi_{\theta,\text{behavior}}(a|s) = \pi_\theta(a|s) $$
However, when there is not sufficient exploration, a potential problem
is the collapse of the policy 
distribution. The distribution then becomes too narrow, and we lose
the exploration pressure that is necessary  for good performance. 

Although we could simply add additional noise,
another common approach is to use {\it entropy regularization}
(see Sect.~\ref{chapter_probability_distributions} for details).
We
then add an additional penalty to the loss function, that 
enforces the entropy $H$ of the distribution to stay larger.
Soft actor critic (\gls{SAC}) is a well-known
algorithm that focuses on
exploration~\cite{haarnoja2018soft,haarnoja2019soft}.\footnote{\url{https://github.com/haarnoja/sac}} SAC  extends the
policy gradient equation 
to 
$$ 
\theta_{t+1} =  \theta_t + 
 R \cdot \nabla_\theta \log \pi_\theta(a_t|s_t) + \eta
\nabla_\theta  H[\pi_\theta(\cdot|s_t)]  
$$
where $\eta \in \mathbb{R}^+$ is a constant that determines the amount of
entropy regularization. SAC ensures that we will move
$\pi_\theta(a|s)$ to the optimal policy, while also ensuring that
the policy stays as wide as possible (trading off
the two against eachother).   Entropy is computed as $H = - \sum_i p_i
\log p_i$ where $p_i$ is the probability of being in state $i$; in SAC entropy is the
negative log of the stochastic policy function
$-\log\pi_\theta(a|s)$.

High-entropy policies favor exploration. 
First, the policy is incentivized to explore more
widely, while giving up on clearly unpromising avenues. Second, with
improved exploration comes improved learning 
speed.

Most
policy-based algorithms (including A3C, TRPO, and PPO) only optimize for
expected value. 
By including entropy  explicitly in the optimization
goal, SAC is able to 
increase the stability of outcome policies, achieving stable results
for different random seeds, and reducing the
sensitivity to hyperparameter settings. Including  entropy into
the optimization goal has been studied widely, see, for example, 
\cite{haarnoja2017reinforcement,kappen2005path,todorov2007linearly,ziebart2008maximum,nachum2017bridging}.

A further element that SAC
uses to improve stability and  sample efficiency is a 
replay buffer.\index{brittleness}
Many policy-based algorithms are  on-policy learners  (including A3C,
TRPO, and PPO). In on-policy algorithms  each policy improvement
 uses feedback on actions  according 
to the most recent version of the behavior policy. On-policy methods
converge well, but 
tend to require many samples to do so.   In contrast, many
value-based algorithms are  off-policy: each
policy improvement can use feedback 
collected at any earlier point during training,  regardless of how the
behavior policy was acting to explore the  environment at the time
when the feedback was obtained. The replay buffer  is such a
mechanism, breaking out of  local maxima. 
Large
replay buffers cause off-policy behavior, improving sample efficiency
by learning from behavior of the past, but also potentially causing
convergence problems.
Like DQN, SAC has overcome these problems, and achieves stable
off-policy performance.

\subsection{Deterministic Policy Gradient}
Actor critic approaches improve the policy-based approach with various
value-based ideas, and with good results. 
Another method to join policy and value approaches is to use a
learned value function as a differentiable target to 
optimize the policy against---we let the policy follow the value
function~\cite{moerland2021lecture}. An example is the {\it deterministic policy gradient}~\cite{silver2014deterministic}.
Imagine we collect data 
$D$ and train a value network $Q_\phi(s,a)$. We can then attempt to optimize the
parameters $\theta$ of a deterministic policy by  optimizing the
prediction of the value network: 
$$J(\theta) =  \mathbb{E}_{s \sim D} \Big[  \sum_{t=0}^n
Q_\phi(s,\pi_\theta(s))  \Big ], $$ 
which by the chain-rule gives the following gradient expression 
\begin{equation*} 
\nabla_\theta J(\theta) =   \sum_{t=0}^n \nabla_a Q_\phi(s,a) \cdot \nabla_\theta \pi_\theta(s) .  
\end{equation*}
In essence, we first train a state-action value network based on
sampled data, and then {\it let the policy follow the value network},
by simply chaining the gradients. Thereby, we push the policy network
in the direction of those actions $a$ that increase the value network
prediction, towards actions that perform better.  


\index{DDPG}\index{deep deterministic policy
  gradient}\label{sec:ddpg} 
Lillicrap
et al.~\cite{lillicrap2015continuous} present Deep deterministic
policy gradient (\gls{DDPG}). It is based on DQN, with the purpose of applying it
to continuous action functions.
 In DQN, if the optimal  action-value function $Q^\star(s,a)$
is known, then the optimal action $a^\star(s)$ can be found via
$a^\star(s) = \argmax_a Q^\star(s,a)$. DDPG uses the derivative of
a continuous function $Q(s,a)$ with respect to the action
argument  to efficiently approximate $\max_a Q(s,a)$.
DDPG is also based on the
algorithms Deterministic policy gradients
(DPG)~\cite{silver2014deterministic} and Neurally fitted Q-learning
with continuous actions
(NFQCA)~\cite{hafner2011reinforcement}, two actor critic algorithms.

\begin{algorithm}[h]
  \caption{DDPG algorithm \cite{lillicrap2015continuous}}
  \label{alg:ddpg}
  \begin{algorithmic}
    \State Randomly initialize critic network $Q_\phi(s, a)$ and actor
    $\pi_\theta(s)$ with weights $\phi$ and $\theta$.
    \State Initialize target network $Q'$ and $\pi'$ with weights $\phi'
    \leftarrow \phi$, $\theta' \leftarrow \theta$
    \State Initialize replay buffer $R$
    \For{episode = 1, M}
      \State Initialize a random process $\mathcal{N}$ for action
      exploration
      \State Receive initial observation state $s_1$
      \For{t = 1, T}
        \State Select action $a_t = \pi_\theta(s_t) + \mathcal{N}_t$
        according to the current policy and exploration noise
        \State Execute action $a_t$ and observe
        reward $r_t$ and observe new state $s_{t+1}$
        \State Store transition $(s_t, a_t,
                r_t, s_{t+1})$ in $R$
        \State Sample a random minibatch of $N$ transitions
               $(s_i, a_i,
        r_i, s_{i + 1})$ from $R$
        \State Set $ y_i = r_i + \gamma Q_{\phi'}(s_{i + 1},
        \pi_{\theta'}(s_{i+1})) $
        \State Update critic by minimizing the loss:
               $L = \frac{1}{N} \sum_i (y_i -
               Q_\phi(s_i, a_i))^2$
        \State Update the actor policy using the sampled policy gradient:
        \begin{equation*}
            \nabla_{\theta} J \approx
            \frac{1}{N} \sum_i
               \nabla_{a} Q_\phi(s, a)|_{s = s_i, a = \mu(s_i)}
               \nabla_{\theta} \pi_\theta(s)|_{s_i}
         \end{equation*}
        \State Update the target networks:
          \begin{equation*}
            \phi' \leftarrow \tau \phi + (1 - \tau) \phi'
          \end{equation*}
          \begin{equation*}
            \theta' \leftarrow \tau \theta +
                (1 - \tau) \theta'
          \end{equation*}
        \EndFor
    \EndFor
  \end{algorithmic}
\end{algorithm}

The pseudocode of DDPG is shown in Alg.~\ref{alg:ddpg}.
DDPG  has been shown to work well on
simulated physics tasks, including 
classic problems such as Cartpole, Gripper, Walker,
and Car driving, being able to learn policies  directly from raw pixel
inputs. DDPG is off-policy  and uses a replay 
buffer and a separate target network to achieve stable deep
reinforcement learning (just as DQN).


DDPG is a popular actor critic algorithm. Annotated pseudocode and efficient
implementations can be found at Spinning Up\footnote{\url{https://spinningup.openai.com}} and Stable
Baselines\footnote{\url{https://stable-baselines.readthedocs.io}} in
addition to the original paper~\cite{lillicrap2015continuous}.

\begin{figure}[t]
  \centering
      \includegraphics[width = 0.8\textwidth]{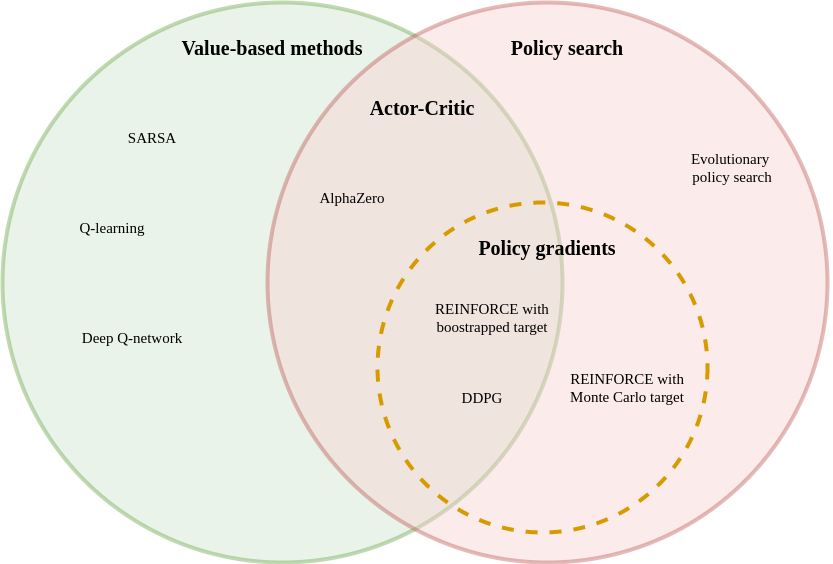}
  \caption[Value-Based, Policy-Based, Actor Critic]{Value-based,
    policy-based and Actor critic methods~\cite{moerland2021lecture}.}
    \label{fig_value_versus_policy}
\end{figure}

\subsubsection*{Conclusion}
We have  seen quite some algorithms that combine the policy and value
approach, and  we have discussed   possible combinations of these
building blocks to construct working
algorithms. Figure~\ref{fig_value_versus_policy} provides a conceptual
map of how the different approaches are related, including two
approaches that will be discussed in later chapters (AlphaZero and
Evolutionary approaches).

Researchers have
 constructed many algorithms and performed experiments to see
when they perform best. 
Quite a number of actor critic algorithms have been developed.   Working
high-performance Python  
implementations can be found on GitHub in the Stable
Baselines.\footnote{\url{https://stable-baselines.readthedocs.io/en/master/guide/quickstart.html}}

\subsection{\em Hands On: PPO and DDPG  MuJoCo Examples}

OpenAI's Spinning Up provides a tutorial on policy
gradient algorithms, complete with TensorFlow and PyTorch versions of
REINFORCE to learn Gym's Cartpole.\footnote{Tutorial:
  \url{https://spinningup.openai.com/en/latest/spinningup/rl_intro3.html\#deriving-the-simplest-policy-gradient}}
with the TensorFlow code\footnote{TensorFlow:
  \url{https://github.com/openai/spinningup/blob/master/spinup/examples/tf1/pg_math/1_simple_pg.py}}
  or PyTorch.\footnote{PyTorch: \url{https://github.com/openai/spinningup/blob/master/spinup/examples/pytorch/pg_math/1_simple_pg.py}}

Now that we have discussed these  algorithms, let us see how
they work in practice, to get a feeling for 
the algorithms and their hyperparameters. 
%
MuJoCo is  the most frequently used physics simulator in policy-based learning
experiments. Gym, the (Stable) Baselines and Spinning up allow us to
run any mix of learning algorithms and experimental environments.
You are  encouraged to try these experiments  yourself.

Please be warned, however, that attempting to install all necessary pieces of
software may invite a minor  version-hell. Different versions of your operating system,
of Python, of GCC, of Gym, of the Baselines, of TensorFlow or
PyTorch, and of MuJoCo all need to line up before you
can see beautiful images of moving arms, legs and jumping
humanoids. Unfortunately not all of these versions are backwards-compatible,
specifically the switch from Python 2 to 3 and from TensorFlow 1 to 2
introduced incompatible language changes. 

Getting everything to work may be  an  effort, and may  require
switching machines, operating systems and languages, but you should really
try. This is the disadvantage of being part of one of the fastest
moving fields in machine learning research. If things do not work with
your current operating system and Python version, in general a combination of Linux Ubuntu
(or macOS), Python 3.7, TensorFlow 1 or 
PyTorch, Gym, and the Baselines may be a good idea to start
with. Search on the GitHub repositories or Stackoverflow when you get
error messages. Sometimes downgrading to the one-but latest version
will be necessary, or fiddling with include or library paths. 

If everything works, then both Spinning up and the Baselines provide convenient scripts that
facilitate mixing and matching algorithms and environments from the
command line.

For example, to run Spinup's PPO on MuJoCo's Walker environment, with a
$32\times 32$ hidden layer, the following command line does the job:
\begin{tcolorbox}
\verb|python -m spinup.run ppo|\\
\verb|--hid "[32,32]" --env Walker2d-v2 --exp_name mujocotest|
\end{tcolorbox}
To train DDPG from the Baselines on the
Half-cheetah, the command is:
\begin{tcolorbox}
\verb|python -m | \verb|base-lines.run|\\
\verb|--alg=ddpg --env=HalfCheetah-v2 --num_timesteps=1e6|
\end{tcolorbox}
All hyperparameters can be controlled via the command line, providing
for a flexible way to run experiments. A final example command line:
\begin{tcolorbox}
\verb|python scripts/all_plots.py|\\
\verb|-a ddpg -e HalfCheetah Ant Hopper Walker2D -f logs/ |\\
\verb|-o logs/ddpg_results|
\end{tcolorbox}
The Stable
Baselines site explains what this command line does.

\section{Locomotion and Visuo-Motor Environments}
We have seen many different policy-based reinforcement learning
algorithms that can be used in agents 
with continuous action spaces. 
%
%
Let 
us have a closer look at the environments that they have been used in,
and how well they perform.

Policy-based methods, and especially the actor critic policy/value
hybrid, work well for many problems, both  with discrete and with
continuous action spaces.
Policy-based methods are often tested on  complex
high-dimensional robotics applications~\cite{kober2013reinforcement}.
Let us have a look at the kind of environments that have been used to
develop PPO, A3C, and the other algorithms. 

Two application categories are robot  
locomotion, and  visuo-motor interaction. 
These two problems have drawn many researchers, and many new algorithms have
been devised, some of which were able to learn impressives
performance.
For each of
the two problems, we will
discuss a few results in more detail. 

\subsection{Locomotion}\label{sec:loco}\index{locomotion}
One of the problems of locomotion of legged entities is the problem of learning
gaits. Humans, with two legs, can walk, run, and jump, amongst
others. Dogs and horses, with four legs, have other gaits, where their
legs may
move in even more interesting patterns, such as the trot, canter, pace and
gallop. The challenges that we pose  robots are often easier. Typical reinforcement
learning tasks are for a one-legged robot to learn to jump, for biped robots
to walk and jump, and for a quadruped to get to learn to use its multitude of
legs in any coordinated fashion that results in forward moving. Learning such policies can be quite computationally
expensive, and a
curious simulated virtual animal has emerged that is cheaper to
simulate:  the  two-legged \emph{half-cheetah},
whose task it is to run forward. We have already seen some of
these robotic creatures in Figs.~\ref{fig:ant}--\ref{fig:humanoid}.


The first approach that we will discuss is by
Schulman et al.~\cite{schulman2015high}. They report 
experiments where human-like bipeds and quadrupeds must learn to stand up and 
learn  running gaits. These are  challenging 3D locomotion tasks
that were formerly attempted with hand-crafted
policies. Figure~\ref{fig:standing} shows a sequence of states.

\begin{figure}[t]
 \begin{center}
 \includegraphics[width=6cm]{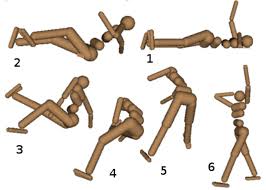}
 \caption{Humanoid Standing Up \cite{schulman2015high}}\label{fig:standing}
 \end{center}
\end{figure}

\begin{figure}[t]
 \begin{center}
 \includegraphics[width=8cm]{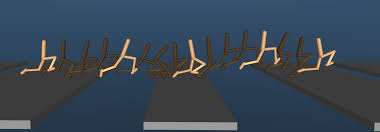}
 \caption{Walker Obstacle Course \cite{heess2017emergence}}\label{fig:obstacle}
 \end{center}
\end{figure}

\begin{figure}[t]
 \begin{center}
 \includegraphics[width=8cm]{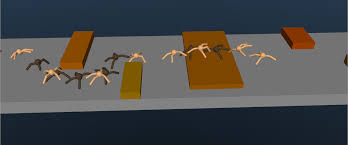}
 \caption{Quadruped Obstacle Course \cite{heess2017emergence}}\label{fig:quadruped}
 \end{center}
\end{figure}

The challenge in
these situations is actually somewhat spectacular: the agent is only
provided with a 
positive reward for moving forward; based on nothing more it has to learn to control
all its limbs  by itself, through trial and error; no hint is given on
how to control a leg or what its purpose is. These results are best
watched in the movies that have been
made\footnote{Such as the movie from the start of this chapter: \url{https://www.youtube.com/watch?v=hx_bgoTF7bs}} about
the learning process.

The authors use an
Advantage actor critic algorithm with trust regions. The algorithm is fully model-free, and learning with
simulated physics was reported to take one to two weeks of real
time.
Learning to walk is quite a complicated challenge, as the movies
illustrate.
They also show the robot  learning
to scale an obstacle run all by itself.


In another study, Heess et al.~\cite{heess2017emergence} report
on end-to-end 
learning of complex robot locomotion from pixel input to (simulated)
motor-actuation.
Figure~\ref{fig:obstacle}  shows how a walker 
scales an obstacle course and Fig.~\ref{fig:quadruped} shows a time lapse of how a quadruped
traverses a course.    Agents
learned to run, jump, crouch and turn as the environment required,
without explicit reward shaping or other hand-crafted features. For
this experiment a distributed version of PPO was 
used. Interestingly, the researchers stress
that the use of a rich---varied, difficult---environment helps to
promote learning of 
complex behavior, that is also robust across a range of tasks.

\subsection{Visuo-Motor Interaction}
Most experiments in ``end-to-end'' learning of robotic locomotion are
set up so that the input is received directly
from features that are derived from the states as calculated by the
simulation software. A step further towards real-world interaction is
to learn directly from camera pixels. We then model eye-hand coordination in
visuo-motor interaction tasks, and the state of the environment has to
be inferred from camera or 
other visual means, and then be translated in joint (muscle)
actuations.

Visuo-motor interaction is a difficult task, requiring many techniques
to work together. 
%
\index{DeepMind control suite}\label{sec:dcs}
Different environments have  been introduced to test algorithms. Tassa
et al.\ 
 report on benchmarking efforts in robot locomotion with
MuJoCo~\cite{tassa2018deepmind}, introducing the DeepMind
control suite, a suite of environments consisting of different MuJoCo
 control 
tasks (see Fig.~\ref{fig:dcs}). 
\begin{figure}[t]
 \begin{center}
 \includegraphics[width=\textwidth]{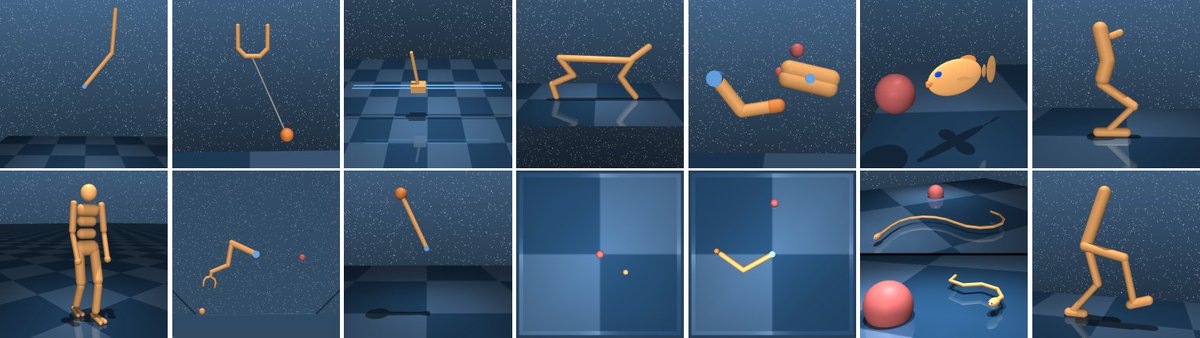}
 \caption[DeepMind Control Suite]{DeepMind Control Suite. Top:
   Acrobot, Ball-in-cup, Cart-pole, Cheetah, Finger, Fish,
   Hopper. Bottom: Humanoid, Manipulator, Pendulum, Point-mass,
   Reacher, Swimmer (6 and 15 links), Walker \cite{tassa2018deepmind}}\label{fig:dcs}
 \end{center}
\end{figure}
The authors also present baseline implementations of learning agents
that use A3C, DDPG and \gls{D4PG} (distributional distributed deep
deterministic policy gradients---an algorithm that extends DDPG).

In addition to learning from state derived features, results are
presented where the agent learns from $84\times 84$ pixel
information, in a
simulated form of visuo-motor interaction. 
%
The DeepMind control suite is especially designed for
further research in the field~\cite{tassa2020dm_control,merel2017learning,merel2018hierarchical,merel2018neural,merel2019deep}. Other
environment suites are Meta-World~\cite{yu2020meta}, Surreal~\cite{fan2018surreal},
RLbench~\cite{james2020rlbench}.  

Visuo-motor interaction is a challenging problem that  remains an
active area of research.

\subsection{Benchmarking}

Benchmarking efforts  are of great importance in the field~\cite{duan2016benchmarking}.
Henderson et al.~\cite{henderson2018deep} published an influential study of the  sensitivity of
outcomes to different hyperparameter settings, and the influence of
non-determinism, by trying to reproduce
many published works in the field. They find large variations in outcomes, and in
general that reproducibilty of results is problematic.  They conclude that \emph{without significance metrics and tighter standardization of
experimental reporting, it is difficult to determine whether
improvements over the prior state-of-the-art are meaningful}~\cite{henderson2018deep}.
Further
studies confirmed these findings~\cite{agarwal2021deep},\footnote{\url{https://github.com/google-research/rliable}} 
and today more works are being 
published with code, hyperparameters, and environments.

Taking inspiration
from the success of the Arcade Learning Environment in game playing,
benchmark suites of continuous control tasks with  high state and
action dimensionality have been introduced~\cite{duan2016benchmarking,garage}.\footnote{The suite in the paper
  is called \href{https://github.com/rll/rllab}{RLlab}. A newer version
  of the suite is named
  \href{https://github.com/rlworkgroup/garage}{Garage}. See also
  Appendix~\ref{ch:env}.}
The tasks include  3D humanoid
locomotion, tasks with 
partial observations, and tasks with hierarchical structure. The
locomotion tasks are: Swimmer, Hopper, Walker, Half-cheetah, Ant,
simple Humanoid and full Humanoid, with the goal being to move forward
as fast as possible. These are difficult tasks because of the high
degree of freedom of movement. Partial observation is achieved by adding noise
or leaving out certain parts of the regular observations. The
hierarchical tasks consist of low level tasks such as learning to
move, and a high level task such as finding the way out of a maze.

\section*{Summary and Further Reading}
\addcontentsline{toc}{section}{\protect\numberline{}Summary and
  Further Reading}
This chapter is concerned with the second kind of model-free
algorithms: policy-based methods. We summarize what we have learned,
and provide pointers to further reading.

\subsection*{Summary}
Policy-based model-free methods are some of the
most popular methods of deep reinforcement learning. For large, continuous 
action spaces, indirect value-based methods are not well suited,
because of the  use of the $\argmax$ function to recover the best action to go with the 
value. Where value-based methods work step-by-step, vanilla policy-based
methods roll out a full future trajectory or episode. Policy-based
methods work with a parameterized current 
policy, which is well suited for a neural network as policy function
approximator.

After the full trajectory has been rolled out, the reward and the
value of the trajectory  is calculated and the policy parameters are updated, using
gradient ascent.
Since the value is only known at the end of an episode, classic policy-based methods have a higher variance
than value based methods, and may converge to a local optimum. The
best known classic policy method is called REINFORCE. 

Actor critic methods add a value network to the policy
network, to achieve the benefits of both approaches. To reduce
variance, $n$-step temporal difference bootstrapping can be added, and a
baseline value can be subtracted, so that we get the so-called \emph{advantage} function
(which subtracts the value of the parent state from the action values
of the future states, bringing their expected value closer to zero).
Well known actor critic methods are A3C, DDPG, TRPO, 
PPO, and SAC.\footnote{Asynchronous advantage actor critic; Deep deterministic
  policy gradients; Trust region policy
  optimization; Proximal policy optimization; Soft actor critic.} A3C features an
asynchronous (parallel, distributed) implementation, DDPG is an
actor critic version of DQN for continous action spaces,  TRPO and PPO
use trust regions to achieve adaptive step sizes in non linear
spaces, SAC optimizes for expected value and entropy of the policy.
Benchmark studies have shown that the performance of these actor
critic algorithm is as good  or better than value-based methods~\cite{duan2016benchmarking,henderson2018deep}. 

Robot learning is among the most popular applications for policy-based
method. Model-free methods have low sample efficiency, and to
prevent the cost of wear after millions of samples, most experiments use a physics simulation as
environment, such as  MuJoCo.
Two main application areas are locomotion (learning to walk, learning
to run) and visuo-motor interaction (learning directly from camera images of
one's own actions). 

\subsection*{Further Reading}
Policy-based methods have been an active research area for some time. Their natural
suitability  for deep function approximation for robotics applications
and other applications with continuous action spaces 
has spurred a large interest in the research community. The
classic policy-based algorithm is Williams'
REINFORCE~\cite{williams1992simple}, which is based on  the policy gradient
theorem, see~\cite{sutton2000policy}. Our explanation is based on
~\cite{ecofet2018,deshpande2019,franccois2018introduction}.
Joining policy and value-based
methods as we do in actor critic is discussed in Barto et
al.~\cite{barto1983neuronlike}. Mnih et
al.~\cite{mnih2016asynchronous} introduce a modern efficient parallel
implementation named A3C. After the success of DQN a version for the
continuous action space of policy-based methods was introduced as
DDPG by Lillicrap et al.~\cite{lillicrap2015continuous}. Schulman et
al.\ have worked on trust regions, yielding efficient popular
algorithms TRPO~\cite{schulman2015trust} and
PPO~\cite{schulman2017proximal}.

Important benchmark studies of policy-based methods are Duan et
al.~\cite{duan2016benchmarking} and Henderson et
al.~\cite{henderson2018deep}. These papers have stimulated
reproducibility in reinforcement learning research. 

Software environments that are used in testing policy-based methods are MuJoCo~\cite{todorov2012mujoco} and
PyBullet~\cite{coumans2019}. Gym~\cite{brockman2016openai} and the
DeepMind control suite~\cite{tassa2018deepmind} incorporate MuJoCo and
provide an easy to use Python interface. An active research community
has emerged around the DeepMind control suite.

\section*{Exercises}
\addcontentsline{toc}{section}{\protect\numberline{}Exercises}
We have come to the end of this chapter, and it is time to test our
understanding with questions, exercises, and a summary.

\subsubsection*{Questions}
Below are some quick questions to check your understanding of this
chapter. For each question a simple, single sentence answer is
sufficient.
\begin{enumerate}
\item Why are value-based methods difficult to use in continuous action spaces?

\item What is MuJoCo? Can you name a few example tasks?
  
\item What is an advantage of  policy-based methods?
\item What is a disadvantage of full-trajectory policy-based methods?
\item What is the difference between actor critic and vanilla
  policy-based methods?
\item How many parameter sets are used by actor critic? How can they
  be represented in a neural network?
  \item Describe the relation between Monte Carlo REINFORCE, $n$-step
    methods, and temporal difference bootstrapping.
\item What is the advantage function?
\item Describe a MuJoCo task that methods such as PPO can learn to perform
  well.
  \item Give two actor critic approaches to further improve upon
    bootstrapping and advantage functions, that are used in
    high-performing algorithms such as PPO and SAC.
\item Why is learning robot actions from image input hard?
\end{enumerate}

\subsubsection*{Exercises}
Let us now look at programming exercises. If you have not already done
so, install MuJoCo  or
PyBullet, and install the DeepMind control
suite.\footnote{\url{https://github.com/deepmind/dm_control}}
We will use agent algorithms from the Stable
baselines. Furthermore, browse the examples directory of the DeepMind control suite
on GitHub, and study the Colab notebook.




\begin{enumerate}
\item \emph{REINFORCE} Go to the Medium
  blog\footnote{\url{https://medium.com/@ts1829/policy-gradient-reinforcement-learning-in-pytorch-df1383ea0baf}}
  and reimplement REINFORCE. You can choose PyTorch, or TensorFlow/Keras,
  in which case you will have to improvise. Run the algorithm on an
  environment with a 
  discrete action space, and compare with DQN. Which works better? Run
  in  an environment with a continuous action space. Note that Gym
  offers a discrete and a continuous version of Mountain Car. 
\item \emph{Algorithms}
  Run REINFORCE on a Walker environment from the Baselines. Run
  DDPG, A3C, and PPO. Run them for different time steps. Make
  plots. Compare training speed, and outcome quality.  Vary
  hyperparameters to develop an intuition for their effect.

\item \emph{Suite} Explore the DeepMind control suite. Look around and see
  what environments have been provided, and  how you can use
  them. Consider extending an environment. What learning challenges
  would you like to introduce? First do a survey of the literature
  that has been published about the DeepMind control suite.


\end{enumerate}


\chapter{Model-Based Reinforcement Learning}\label{chap:learned}\label{chap:model}

The previous  chapters discussed model-\emph{free} methods, and we saw
their success in video games and simulated robotics. In model-free
methods the agent updates a policy  directly from the
feedback that the environment provides on its actions. The
environment performs the state transitions and calculates the
reward. A disadvantage of deep model-free  
methods is that they can be slow to train; for stable convergence or low
variance often  millions of environment
samples are needed before the policy function converges to a high quality optimum.

In contrast, with model-\emph{based} methods the agent first builds its own  internal
transition model  from the environment feedback. The agent can then use this local
transition model  to find out about the
effect of actions on states
and rewards. The agent can use a planning
algorithm to play \emph{what-if} games, and  generate
policy updates, all without causing any state changes in the environment.
This approach 
promises higher quality at lower sample complexity. Generating 
policy updates from the internal model is called planning or
\emph{imagination}. 

Model-based methods update the policy
indirectly: the agent first 
learns a local transition model from the environment, which the
agent  then uses to update the policy. 
Indirectly learning  the  policy function has two
consequences. On the positive side, as soon as the agent  has its own
model of the state transitions of the world, it can 
learn the best policy for free,
without further incurring the cost of acting in
the environment.  Model-based methods thus may have a lower
sample complexity.
The downside is that the learned transition model
may be inaccurate, and the resulting policy may be of low quality.
No matter how many samples can be taken for free from the model,
if the agent's local transition model does not reflect the
environment's real transition model, then the locally learned policy
function  will not work in  the environment. Thus, dealing with uncertainty
and model bias are important elements in model-based reinforcement learning. 

The idea to first learn an internal representation of the
environment's transition function 
has been conceived many years ago, and   transition  models have been
implemented in many different ways. Models can be 
tabular, or they can be based on various kinds of deep learning, as we
will see.

This chapter will start
%
with an example showing how model-based methods work.
%
Next, we describe in more detail different kinds of model-based
approaches; approaches that focus on \emph{learning} an accurate  model,
and approaches for \emph{planning} with an imperfect model.
Finally, we describe application 
environments for which model-based methods have been used in practice,
to see how well the approaches perform. 

The chapter is concluded with  exercises, a summary, and pointers
to further reading.

\section*{Core Concepts}
\begin{itemize}
\item Imagination 
\item Uncertainty models
\item World models, Latent models 
\item Model-predictive control 
\item Deep end-to-end  planning and learning
\end{itemize}

\section*{Core Problem}
\begin{itemize}
\item  Learn and use  accurate transition models for high-dimensional problems
\end{itemize}

\section*{Core Algorithms}
\begin{itemize}
\item Dyna-Q (Alg.~\ref{alg:dyna-q})
  \item Ensembles and model-predictive control (Alg.~\ref{alg:metrpo},
    \ref{alg:nagabandi})
\item Value prediction networks (Alg.~\ref{alg:dplan})
\item Value iteration networks (Sect.~\ref{sec:vin})
\end{itemize}
\section*{Building a Navigation Map}
To illustrate basic concepts of model-based reinforcement learning, we
return to the supermarket example.

Let us compare how model-free and model-based methods  
 find their way to the supermarket in a new city.\footnote{We use 
distance to the supermarket as negative reward, in order to formulate this as a
distance minimization problem, while still being able to reason in our
familiar reward maximization setting.} In this example we will use value-based
Q-learning; our policy $\pi(s, a)$ will be derived directly from
the $Q(s,a)$ values with $\argmax$, and  writing $Q$ is in this sense equivalent to writing $\pi$.

\emph{Model-free} Q-learning: the agent picks the start state $s_0$, and uses (for
example) an
$\epsilon$-greedy  behavior policy on the action-value function $Q(s,a)$ to select the next
action. The environment  then executes the action,
computes  the next state $s'$ and  reward $r$, and returns these to the
agent. The agent  updates its
action-value function $Q(s,a)$ with the familiar update rule 
$$Q(s,a)\leftarrow Q(s,a) +\alpha[r+\gamma\max_aQ(s',a) - Q(s,a)].$$
The agent  repeats this procedure until the values in the $Q$-function no
longer change greatly.

Thus we  pick our start
location in the city,  perform one walk along a block in an
$\epsilon$-greedy direction, and record
the reward  and the new state at which we arrive. We use the
information to update the 
policy, and from our new location, we walk again in an
$\epsilon$-greedy direction using the policy. If we
find the supermarket, we start over  again, trying to find a shorter
path, until our policy values no 
longer change (this 
may take many environment interactions). Then
the best policy is the path with the shortest distances. 

\emph{Model-based} planning and learning: the agent uses the $Q(s,a)$ function 
as behavior policy as before to sample the new state and reward from
the environment, and to update the policy ($Q$-function). In addition,
however, the agent will record the new state and reward in a local
transition  $T_a(s,s')$ and  reward
function $R_a(s,s')$. Because the agent now has these local entries
we can  also sample from our local
 functions to update the policy. We  can choose: sample from the (expensive) environment
transition function, or from the (cheap) 
local transition function. There is a caveat with sampling locally,
however.  The  local functions may contain fewer entries---or only
high variance entries---especially in the early stages, when few environment
samples have been performed. The usefulness of the local functions
increases as more environment samples are performed.

Thus, we now have a local map on
which to record the  new states and rewards. We will use this
map to peek, as often as we like and at no cost, at a location on that
map, to update the
$Q$-function. As more environment samples come in, the map will have
more and more  locations for which a distance to the
supermarket is recorded.
When glances at the map 
do not improve the  policy anymore, we have to walk in the environment again,
and, as before, update the map and the policy.


In conclusion, model-free finds all  policy updates  outside the agent,
from the environment 
feedback; model-based also\footnote{One option is to
  only update the policy from the agent's
  internal transition model, and not by the environment samples
  anymore. However, another option is to keep using the environment
  samples to also update the policy in the model-free way. Sutton's
  Dyna~\cite{sutton1990integrated} approach is a well-known example of
  this last, hybrid,  approach. Compare
  also Fig.~\ref{fig:imagination} and Fig.~\ref{fig:learn}.} uses policy updates from within the agent, using
information from its local map (see Fig.~\ref{fig:modelfreemodelbased}). In both methods all
updates to  the  policy are ultimately derived from the
environment feedback; model-based  offers a different way to use the
information to update the policy, a way that may be more
information-efficient, by  keeping  information from each 
sample within the agent transition model and re-using that information.

  

\section{Dynamics Models of  High-Dimensional Problems}

 \begin{figure}[t]
 \begin{center}
 \begin{tikzpicture}[->,scale=1]
   \node (i) at (2,2)  {\bf Value/Policy};
   \node (j) at (3,0) {\bf Environment};
   \node (k) at (0,0) {\bf Model};


   \draw[bend left,->,very thick]  (i) to node [auto] {\em acting} (j);
   \draw[bend left,->,very thick]  (j) to node [auto] {\em model learning} (k);
   \draw[bend left=60,->,very thick,in=135]  (k) to node [auto] {\em planning} (i);
   \draw[bend left,->,very thick]  (j) to node [auto] {\em direct RL} (i);
\end{tikzpicture}
 \caption{Direct and Indirect Reinforcement Learning~\cite{sutton2018introduction}}\label{fig:modelfreemodelbased}
 \end{center}
\end{figure}
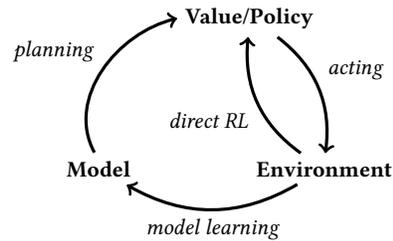

The application environments for model-based reinforcement learning are the
same as for model-free;
our goal, however, is   to solve larger and
more complex problems in the same amount of time, by virtue of the lower
sample complexity and, as it were, a deeper understanding of the environment.

\subsubsection*{Transition Model and Knowledge Transfer}
\index{learning}
The principle of model-based learning is as follows.
Where model-free methods sample  the environment to  learn the state-to-\emph{action}
policy function $\pi(s,a)$ based on action rewards,  model-based
methods sample the environment to learn the state-to-\emph{state}
transition function $T_a(s,s')$ based on action rewards. Once
the accuracy of  this local transition function is good enough, the
agent can sample from this local function to improve the policy
$\pi(s,a)$ as often as it likes, without incurring the cost of actual
environment samples. 
In the model-based approach, the agent builds its
own local state-to-state transition (and reward) model of the
environment, so that, in theory at least, it does not need the environment
anymore. 

This brings us to another
reason for the interest in  model-based methods. For
sequential decision problems,  knowing the transition function is
a natural way 
of capturing \emph{the essence} of how the environment works---$\pi$
gives the next action, $T$ gives the next state.

This is
useful, for example,  when we switch to a related  environment. When
the transition function of the environment is 
known by the agent, then the agent can be  adapted quickly, without having
to  learn a whole new policy by sampling the environment. 
When a  good local transition
function of the domain is known by the agent, then new,  but related, problems might be
solved efficiently. Hence,
model-based reinforcement learning may contribute to efficient transfer
learning (see  Chap.~\ref{chap:meta}). 

\subsubsection*{Sample Efficiency}
\index{sample efficiency}\label{sec:sample}
The sample efficiency of an agent algorithm tells us how many environment
samples it needs for the policy to reach a certain
accuracy.

To achieve  high sample efficiency, model-based methods learn a
dynamics model. Learning high-accuracy high-capacity  
models of high-dimensional problems requires a high number of training
examples, to prevent overfitting (see
Sect.~\ref{sec:overfitting}). Thus, reducing overfitting in 
learning the transition model
would  negate (some of) the advantage of the low sample complexity that model-based
learning of the policy function achieves.
Constructing accurate deep transition models can be difficult in practice,
and for many complex sequential decision problems the best  results are often achieved
with model-free methods, although deep model-based methods are becoming
stronger (see, for example, Wang et al.~\cite{wang2019benchmarking}).  

\section{Learning and Planning Agents}
The promise of model-based reinforcement learning is to find a high-accuracy
behavior policy   at a 
low cost, by building a local model of the world. This will only work if
the learned transition model provides accurate predictions, and if the
extra cost of planning with the model is reasonable. 

Let us  see which solutions have been
developed for deep model-based reinforcement learning. 
In
Sect.~\ref{sec:hdenv} we will have a closer look at the performance in
different environments. First, in this section, we will look at four
different algorithmic approaches, and at a classic  approach:
Dyna's tabular imagination.

\subsubsection*{Tabular Imagination}
A classic   approach is Dyna~\cite{sutton1990integrated},
which popularized the idea of  model-based reinforcement learning. In Dyna,  environment
samples are  used in a hybrid model-free/model-based manner,   to train 
the transition model, use planning to improve the policy, while also training the policy function
directly. 

Why is Dyna a hybrid approach?
Strict model-based methods  update 
the policy only by planning using the agent's transition model, see
Alg.~\ref{lst:back}. In Dyna,
however, environment samples are used to \emph{also} update the policy
directly (see Fig.~\ref{fig:imagination} and Alg.~\ref{lst:dyna}). Thus we get a hybrid approach combining
model-based and model-free learning. This hybrid model-based planning is  called 
\emph{imagination} because  looking ahead  with the agent's own
dynamics model resembles imagining environment samples outside the
real environment inside the ``mind'' of the agent. In this approach
the imagined  samples augment the real (environment) samples
at no sample  cost.\footnote{The term \emph{imagination} is used
  somewhat loosely in the field. In  a strict sense imagination refers
  only to
  updating the policy from the internal model by planning. In a wider
  sense imagination  refers to  hybrid schemes where the policy is
  updated from both the internal model and the environment. Sometimes
  the term \emph{dreaming} is  used  for agents imagining environments.}
  


Imagination is a mix of
model-based and model-free reinforcement learning. Imagination
performs regular direct reinforcement learning, where the environment
is sampled with actions according to the behavior policy, and the
feedback is used to update the same behavior policy.
Imagination also uses the environment sample to update the dynamics
model $\{T_a, R_a\}$. This extra model is also sampled, and provides extra
updates to the behavior policy, in between the model-free
updates. 

%
%
\begin{algorithm}[t]
  \begin{algorithmic}
    \Repeat
    \State Sample environment $E$ to generate data $D=(s, a, r', s')$ 
    \State Use $D$ to learn $M=T_a(s,s'),R_a(s,s')$ \Comment{learning}
    \For{$n=1,\ldots, N$}
    \State Use $M$ to update policy $\pi(s, a)$  \Comment{planning}
    \EndFor
    \Until $\pi$ converges
  \end{algorithmic}
  \caption{Strict Learned Dynamics Model}\label{lst:back}
\end{algorithm}

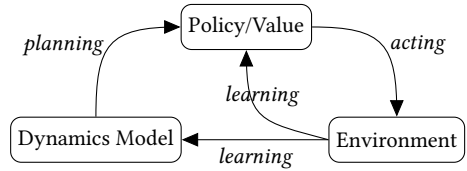
\begin{figure}[t]
  \begin{center}
    \begin{tikzpicture}[>=triangle 45,
  desc/.style={
		scale=1.0,
		rectangle,
		rounded corners,
		draw=black, 
		}]

  \node [desc,minimum height=0.6cm] (env) at   (4,0.5) {Environment};
  \node [desc,minimum height=0.6cm] (tm) at   (0,0.5) {Dynamics Model};
  \node [desc,minimum height=0.6cm] (pol) at   (2,2) {Policy/Value};
  \draw (env.west) edge[->,in=0,out=180,looseness=2.5] node[below]
  {\em learning} (tm.east);
  \draw (env.west) edge[->,in=270,out=165,looseness=1.5] node[above]
  {\em learning} (pol.south);
  \draw (pol.east) edge[->,out=0,in=90,looseness=1.5] node[right] {\em acting} (env.north);
  \draw (tm.north) edge[->,out=90,in=180,looseness=1.5] node[left]
  {\em planning} (pol.west);

\end{tikzpicture}
    \caption{Hybrid Model-Based Imagination}\label{fig:imagination}
  \end{center}
\end{figure}

\begin{algorithm}[t]

  \begin{algorithmic}[t]
    \Repeat
    \State Sample env $E$ to generate data $D=(s, a, r', s')$ 
    \State {\em Use $D$ to update policy $\pi(s, a)$}   \Comment{learning}
    \State Use $D$ to learn $M = T_a(s,s'), R_a(s,s')$  \Comment{learning}
    \For{$n=1,\ldots, N$}
    \State Use $M$ to update policy $\pi(s, a)$ \Comment{planning}
    \EndFor
    \Until $\pi$ converges
  \end{algorithmic}
  \caption{Hybrid Model-Based Imagination}\label{lst:dyna}

\end{algorithm}

The diagram in Fig.~\ref{fig:imagination} shows how sample feedback is
used both for updating the policy directly and for updating the model,
which then updates the policy, by planning ``imagined'' feedback. In
Alg.~\ref{lst:dyna} the general imagination approach is
shown as pseudocode.

 Sutton's
Dyna-Q~\cite{sutton1990integrated,sutton2018introduction}, which is
shown in more detail in
Alg.~\ref{alg:dyna-q}, is a  concrete implementation of the
imagination approach. Dyna-Q uses the Q-function   as behavior
policy $\pi(s)$ to perform $\epsilon$-greedy sampling of the environment. It then
updates this policy with the reward, and an explicit model $M$. When
the model $M$ has been updated, it is used $N$ 
times by planning  with random actions to update the
Q-function. The pseudocode shows the learning steps (from
environment $E$) and $N$ planning steps (from model $M$). In both cases the
Q-function state-action values are updated. The best action is then derived
from  the Q-values as usual.

Thus, we see that  the number of updates to the policy can
be  increased without more  environment samples. By choosing the value for
$N$, we can
tune how many of the policy updates will be environment samples, and
how many will be model samples. In the larger problems
that we will see later in this chapter, the ratio of environment-to-model samples is
often set at, for example, $1 : 1000$, greatly reducing sample complexity. The
questions then become, of course: \emph{how good is the model}, and: \emph{how far is
 the resulting policy from a model-free baseline?}

\begin{algorithm}[t]
  
        \begin{algorithmic}
            \State Initialize $Q(s,a) \rightarrow \mathbb{R}$ randomly
            \State Initialize $M(s,a) \rightarrow \mathbb{R}\times S$ randomly \Comment{Model}
            \Repeat
                \State Select $s \in S$ randomly
                \State $a \gets \pi(s)$ \Comment{$\pi(s)$ can be
                  $\epsilon$-greedy$(s)$ based on $Q$}
                \State $(s',r) \gets E(s, a)$ \Comment{Learn new
                  state and reward from environment}
                \State $Q(s, a) \gets  Q(s, a) + \alpha \cdot [r +
                \gamma \cdot \max_{a'} Q(s', a') - Q(s,a)]$
                \State $M(s, a) \gets (s', r)$
                \For{$n = 1, \dots, N$}
                    \State Select $\hat{s}$ and $\hat{a}$ randomly
                    \State $(s', r) \gets M(\hat{s}, \hat{a})$
                    \Comment{Plan imagined state and reward from  model}
                    \State $Q(\hat{s}, \hat{a}) \gets  Q(\hat{s}, \hat{a}) + \alpha \cdot [r + \gamma \cdot \max_{a'} Q(s', a')-Q(\hat{s},\hat{a})]$
                \EndFor
            \Until $Q$ converges
        \end{algorithmic}

  \caption{Dyna-Q~\cite{sutton1990integrated}} 
  \label{alg:dyna-q}
\end{algorithm}

\subsubsection*{\em Hands On: Imagining Taxi Example}\label{sec:taxi-example}
It is time to illustrate how Dyna-Q works with an example. For that, we
turn to one of our favorites, the Taxi world.

Let us see what the effect of imagining with a model can be. Please
refer to Fig.~\ref{fig:taxi6}. We  use our 
simple maze example, the Taxi maze, with zero imagination ($N=0$), and with large
imagination ($N=50$).
 Let us assume that the reward at
all states 
returned by the environment is $0$, except for  the goal, where the
reward  is $+1$. In  states the usual  actions are present (north,
east, west, south),  except at borders or walls.

\begin{figure}[t]
\begin{center}
\includegraphics[width=5cm]{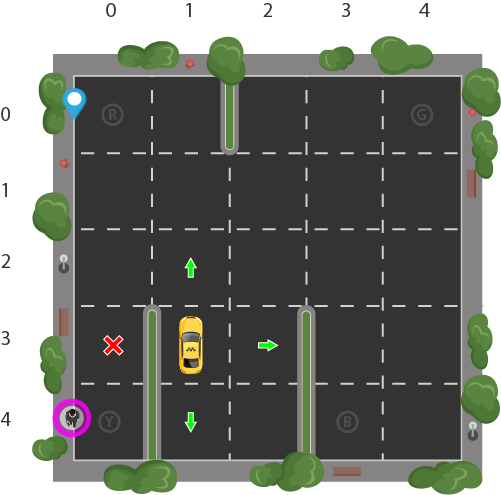}
\caption{Taxi world~\cite{learn}}\label{fig:taxi6}
\end{center}
\end{figure}

When $N=0$ Dyna-Q performs exactly Q-learning,
randomly sampling action rewards, building up the Q-function, and
 using the Q-values following the $\epsilon$-greedy policy for
action selection. The purpose of the Q-function is to act as a vessel
of information to find the goal. How does our vessel get filled with
information? Sampling starts off randomly, and the Q-values fill
slowly, since the reward landscape is flat, or sparse: only the goal state
returns $+1$, all other states return $0$. In order to fill the
Q-values with actionable information on where to find the goal, first
the algorithm must be lucky enough to choose a state
next to the goal, including the appropriate action to reach
the goal. Only then the first useful reward information is found and the first
non-zero step towards finding the goal can be entered into the Q-function.
We conclude that, with $N=0$, the Q-function is filled up slowly,  due to sparse rewards.

What happens when we turn on planning? When we set $N$ to a high
value, such as $50$, we perform $50$ planning steps for each learning step.
As we can see in the algorithm, the model is built alongside the
$Q$-function, from environment returns. As long as the $Q$-function is
still fully
zero, then planning with the model will also be useless. But as soon as one
goal entry is entered into $Q$ and $M$, then planning will start to
shine: it will perform $50$ planning samples on the M-model, probably
finding the goal information, and possibly building up an entire
trajectory filling states in the $Q$-function with actions towards the
goal.

In a way, the model-based planning amplifies any useful reward  information that the agent
has learned from the environment, and plows it back quickly into the policy
function.
The policy is learned much quicker, with fewer environment samples.

\subsubsection*{Reversible Planning and Irreversible
  Learning}\index{reversible actions}\index{planning and learning}

\index{planning}\index{learning}
Model-free methods sample the environment 
and learn the policy function $\pi(s,a)$ directly, in one step. Model-based methods sample the
environment to learn the policy indirectly, using a dynamics model
$\{T_a, R_a\}$ 
(as we see in Fig.~\ref{fig:modelfreemodelbased} and~\ref{fig:learn},
and in Alg.~\ref{lst:back}).

\begin{figure}[t]
  \centering
  \begin{tabular}{cc}
    \begin{tikzpicture}[>=triangle 45,
  desc/.style={
		scale=1.0,
		rectangle,
		rounded corners,
		draw=black, 
		}]

  \node [desc,minimum height=0.6cm] (env) at   (4.5,0.5) {Environment};
  \node [desc,minimum height=0.6cm] (tm) at   (0.5,0.5) { Dynamics Model};
  \node [desc,minimum height=0.6cm] (pol) at   (2.5,2) {Policy/Value};
  \draw (env.west) edge[->,in=0,out=180,looseness=2.5] node[below]
  {\em learning} (tm.east);
  \draw (pol.east) edge[->,out=0,in=90,looseness=1.5] node[right] {\em acting} (env.north);
  \draw (tm.north) edge[<->,out=90,in=180,looseness=1.5] node[left]
  {\em planning} (pol.west);

\end{tikzpicture}
    &
      \begin{tikzpicture}[>=triangle 45,
  desc/.style={
		scale=1.0,
		rectangle,
		rounded corners,
		draw=black, 
		}]

  \node [desc,minimum height=0.6cm] (tm) at   (0,0.5) {Environment};
  \node [desc,minimum height=0.6cm] (pol) at   (0,2) {Policy/Value};
  \draw (tm.west) edge[->,in=180,out=180,looseness=2.5] node[left]
  {\em learning} (pol.west);
  \draw (pol.east) edge[->,out=0,in=0,looseness=2.5] node[right] {\em acting} (tm.east);

\end{tikzpicture}
      \end{tabular}
    \caption{Model-based (left) and Model-free (right). Learning
      changes the environment state irreversibly (single arrow); planning changes the
      agent state reversibly (undo, double arrow)}\label{fig:learn}
  
\end{figure}
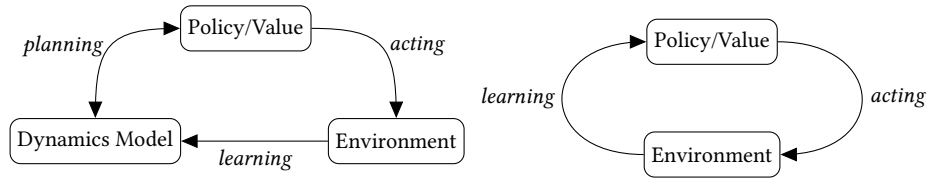



It is useful to step back for a moment to consider the place of
learning and 
planning algorithms in the reinforcement learning paradigm.  Please
refer to Table~\ref{tab:planlearn} for a summary of
differences between planning and learning. 

Planning  with an internal transition model is \emph{reversible}. When
the agent uses its own 
transition model to perform local actions on a local state, then the
actions can be \emph{undone}, since the  agent applied them to a copy in its
own memory~\cite{moerland2020framework}.\footnote{In our dreams we can undo our
actions, play \emph{what-if}, and imagine alternative
realities.}\index{reversible actions}
 Because of  this local state memory, the agent can return to the old state, 
reversing the local state change caused by the local action that it has just
performed. The agent can then try an 
alternative action (which it can also reverse). The agent can  use
tree-traversal methods to 
traverse the state space, backtracking to try other states.

\begin{table}[t]
  \begin{center}
  \begin{tabular}{lll}
    &{\bf Planning} & {\bf Learning}  \\
    \hline\hline
    Transition model in: & Agent&  Environment\\
    Agent can Undo:& Yes & No\\
    State is:  & Reversible by agent & Irreversible by agent\\
    Dynamics: & Backtrack & Forward only\\
    Data structure: & Tree & Path\\
    New state: & In agent & Sample from environment\\
    Reward:    & By agent & Sample from environment\\
    Synonyms: & Imagination, simulation & Sampling, rollout\\
    \hline
  \end{tabular}
  \caption{Difference between Planning and Learning}\label{tab:planlearn}
\end{center}
\end{table}

\index{planning}\index{learning}\index{irreversible actions}
In contrast to planning, \emph{learning} is done when the agent does
not  have access to its own
transition function $T_a(s,s')$. The agent can get reward information by sampling
real actions in the environment. These actions are not played out inside the
agent but executed in the actual environment; they are irreversible
and can not be undone by the agent. Learning
uses actions that irreversibly change the state of the
environment. Learning does not permit backtracking; learning
algorithms learn a policy by repeatedly sampling the environment.

Note the similarity between learning and planning: learning
samples rewards from the external environment, planning from the
internal model; both
use the samples to update the policy function
$\pi(s,a)$.

\subsubsection*{Four Types of Model-Based Methods}
In model-based reinforcement learning the challenge is to learn deep,
high-dimen\-sional transition models 
from limited 
data. Our methods should be able to account for model uncertainty, and
plan over these 
models to achieve  policy and value functions that perform as well or
better than model-free methods.
Let us look in more detail at specific model-based reinforcement
learning methods to see how this can be achieved.

Over the years, many different approaches for high-accuracy
high-dimensional model-based
reinforcement learning have been devised.
Following~\cite{plaat2021high}, we group the methods  into four main
approaches. 
We start with two approaches for learning the model, and
then two approaches for planning, using the model.
For each  we
will take  a few representative papers from the literature 
that we describe in more depth. After we have done so, we will look at
their performance in different environments. But let us start with the
methods for learning a deep model first.

\subsection{Learning the Model}

In model-based approaches, the transition model is learned
from sampling the environment.
%
%
%
%
%
If this model is not accurate, then planning will
not  improve the value or policy function, and the method will
perform worse than model-free methods. When  the
learning/planning ratio is set to $1/1000$, as it is in some
experiments, inaccuracy in the models
will reveal itself quickly in a low accuracy policy function.

Much research has focused on  achieving  high
accuracy dynamics models for high-dimensional problems.
Two methods to achieve better
accuracy are \emph{uncertainty modeling} and \emph{latent models}. We
will start with uncertainty modeling.

\subsubsection{Modeling Uncertainty}
\index{Gaussian processes}
The variance of the transition model can be reduced by  increasing the number of
environment samples, but there are also other approaches that we will
discuss. 
A popular approach for smaller problems is to use 
Gaussian processes, where the dynamics model is learned
by giving an estimate of the function and of the uncertainty around
the function with a covariance matrix on the entire
dataset~\cite{bishop2006pattern}. A Gaussian model can be learned 
from few data points,  
and the transition model can be used to plan the policy
function successfully. 
%
%
An example of this approach is the \gls{PILCO} system, which stands
for Probabilistic Inference for Learning
Control~\cite{deisenroth2011pilco,deisenroth2013gaussian}. This system
was effective on  Cartpole and Mountain car, but does not scale to larger problems.
%
%

We can also sample from a trajectory distribution optimized
for cost, and 
use that to train the policy, with a policy-based
method~\cite{levine2013guided}. Then we can  optimize
policies with the aid of  locally-linear models and a 
stochastic trajectory optimizer.
This is the approach that is used in  Guided policy search (GPS), which been shown to train
complex policies with thousands of 
parameters, learning tasks in  MuJoCo such as  Swimming, Hopping
and Walking.

\begin{algorithm}[t]
  
        \begin{algorithmic}
            \State Initialize policy $\pi_\theta$ and the
            models $\hat{m}_{\phi_1}, \hat{m}_{\phi_1}, \ldots,
            \hat{m}_{\phi_K}$ \Comment{ensemble}
            \State Initialize an empty dataset $D$
            \Repeat
                \State $D\gets$ sample with $\pi_\theta$ from
                environment $E$
                \State Learn models $\hat{m}_{\phi_1}, \hat{m}_{\phi_1}, \ldots,
            \hat{m}_{\phi_K}$ using $D$ \Comment ensemble
                \Repeat
                    \State $D'\gets$ sample with $\pi_\theta$ from $\{\hat{m}_{\phi_i}\}^K_{i=1}$
                 
                    \State Update $\pi_\theta$ with TRPO using $D'$ \Comment{planning}
                    \State Estimate performance of trajectories
                    $\hat{\eta}_\tau(\theta,\phi_i)$ for $i=1, \ldots, K$  
                \Until{performance converges}
            \Until $\pi_\theta$ performs well in environment $E$

        \end{algorithmic}

  \caption{Planning with an Ensemble of Models \cite{kurutach2018model}} 
  \label{alg:metrpo}
\end{algorithm}
Another popular method to reduce variance in machine learning is the ensemble
method. Ensemble methods combine multiple learning 
algorithms to achieve  better predictive performance; for example, a
random forest of decision trees often has better predictive
performance than a single decision
tree~\cite{bishop2006pattern,opitz1999popular}. In deep model-based
methods the ensemble methods
are used to estimate the variance and account for it during planning.  A number of researchers
have reported good results with ensemble methods on larger
problems~\cite{clavera2018model,janner2019trust}. For example, Chua et al.\   use  an ensemble of probabilistic
neural network models~\cite{chua2018deep} in their approach named
Probabilistic ensembles with trajectory sampling (\gls{PETS}).\index{PETS}  They report good results on high-dimensional
simulated robotic  tasks (such as Half-cheetah and Reacher).
Kurutach et
al.~\cite{kurutach2018model} combine an ensemble of models with
TRPO, in ME-TRPO.\footnote{A video is available at
  \url{https://sites.google.com/view/me-trpo}. The code is  
  at \url{https://github.com/thanard/me-trpo}. A blog post is at \cite{hui2018model}.}
In ME-TRPO an ensemble of deep neural networks is used to  maintain model
uncertainty, while
TRPO is used to control the model parameters. 
In the planner, each imagined step is sampled from the
ensemble predictions (see 
%
Alg.~\ref{alg:metrpo}).

Uncertainty modeling tries to improve the accuracy of high-dimensional
models by probabilistic methods.
A different approach, specifically designed for high-dimensional deep models, is the latent model approach, 
which we will discuss next.

\subsubsection{Latent Models}\label{sec:latent}
Latent models focus on dimensionality reduction of
high-dimensional problems. 
The idea
behind latent models is that in most high-dimensional environments
some elements  are less important, such as buildings in the background
that never move and that have no relation with the reward. We can 
abstract  these  unimportant elements away from the model, 
reducing the effective dimensionality of the space. Latent models do so by
learning to represent  the elements of the input and the
reward.
Since planning and learning are now possible in a
lower-dimensional latent space, the sampling complexity of
learning from the latent models improves.
\index{value prediction network}

Even though latent model approaches are often complicated designs, many
works have been published that show good 
results~\cite{kaiser2019model,hafner2018learning,hafner2019dream,sekar2020planning,silver2017predictron,hafner2020mastering,ha2018world}.
  Latent models use multiple neural 
  networks, as well as different learning and planning algorithms.

  To
 understand this approach, we 
will briefly  discuss one such latent-model approach: the Value prediction network
(VPN) by Oh et al.~\cite{oh2017value}.\footnote{See
  \url{https://github.com/junhyukoh/value-prediction-network} for the
  code.}
\gls{VPN} uses four differentiable functions, that are trained to predict the 
value~\cite{grimm2020value}, Fig.~\ref{fig:oh} shows how the core functions.
The core idea in VPN is not to
learn directly in the actual observation space, but first to
transform the  state respresentations to a smaller latent representation
model, also known as abstract model. The other functions, such as
value, reward, and next-state, then work on these smaller latent states,
instead of on the more complex high-dimensional states.  In this way, planning and learning occur
in a space 
where states are encouraged only to contain the elements that influence
value changes.
Latent space
is lower-dimensional, and training and planning become more efficient.

The four functions in VPN are: (1) an encoding
function, (2) a reward function, (3) a value function, and (4) a transition
function. All functions are parameterized with their own set of
parameters. To distinghuish these latent-based functions from the
conventional observation-based functions $R, V, T$ they are denoted
 as $f^{enc}_{\theta_e}, f^{reward}_{\theta_r},
f^{value}_{\theta_v}, f^{trans}_{\theta_t}$.

\begin{itemize}
\item The encoding function $f^{enc}_{\theta_e}:s_{actual}\rightarrow
s_{latent}$ maps the observation $s_{actual}$ to the abstract state
using  neural network $\theta_e$, such as a CNN for visual
observations. This is the function that performs the dimensionality reduction.
\item The latent-reward function $f^{reward}_{\theta_r}:(s_{latent},o)\rightarrow r,
\gamma$ maps the latent state $s$ and option $o$ (a kind of action) to the
reward and discount factor. If the option takes $k$ primitive actions,
the network should predict the discounted sum of the $k$ immediate
rewards as a scalar. (The role of options is  explained in the paper~\cite{oh2017value}.) The network also predicts option-discount factor
$\gamma$ for the number of steps taken by the option.
\item The latent-value function
  $f^{value}_{\theta_v}:s_{latent}\rightarrow
  V_{\theta_v}(s_{latent})$ maps the abstract state to its value using
  a separate neural network $\theta_v$. This value is the value of the
  latent state, not of the actual observation state $V(s_{actual})$.
\item The latent-transition function $f^{trans}_{\theta_t}:(s_{latent},o)\rightarrow
  s'_{latent}$ maps the latent state  to the next latent state,
  depending also on the option.
\end{itemize}
Figure~\ref{fig:oh} shows how the core functions work together in the
smaller, latent, space; with $x$ the
observed actual state, and $s$ the encoded latent
state~\cite{oh2017value}.

\begin{figure}[t]
\begin{center}
\includegraphics[width=7cm]{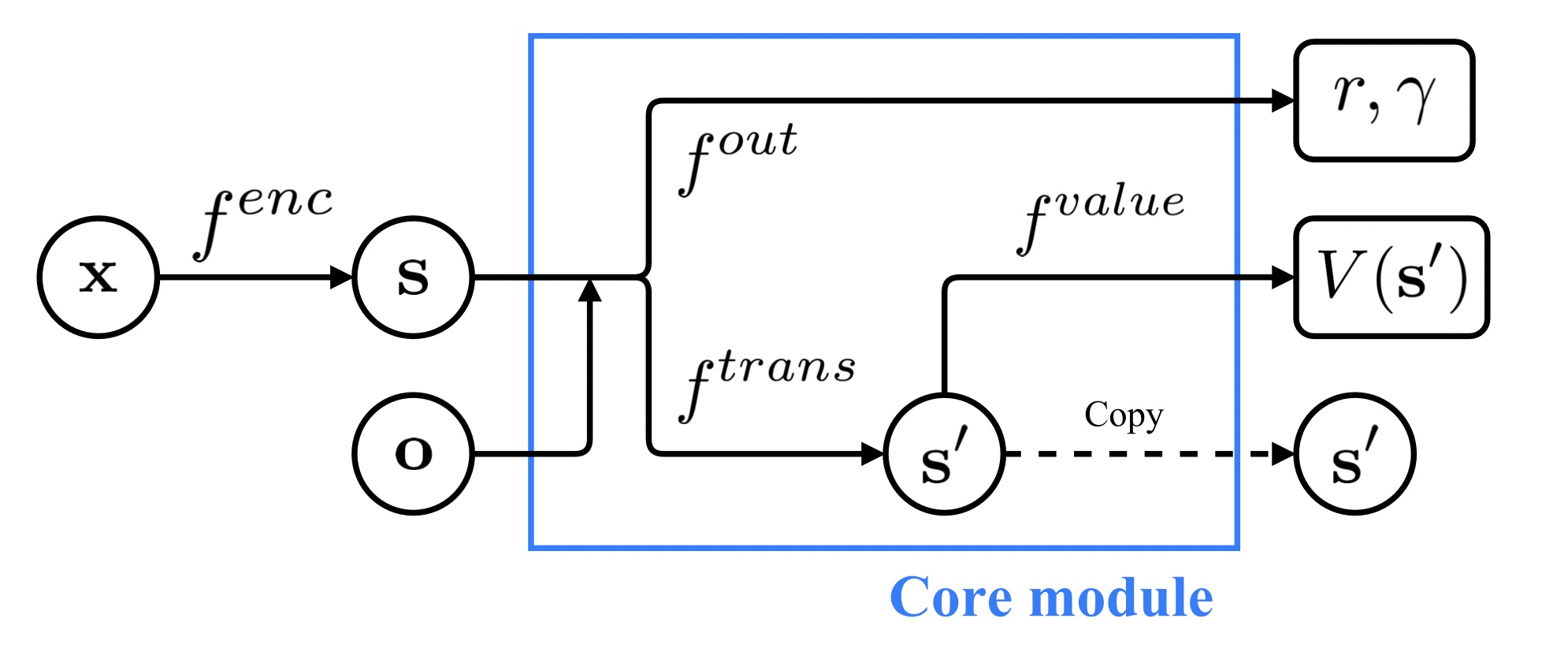}
\caption{Architecture of latent model~\cite{oh2017value}}\label{fig:oh}
\end{center}
\end{figure}

The figure shows a single rollout step, planning one step ahead.
However,  a model also allows looking further into the future, by
performing multi-step rollouts. Of course, this requires a highly
accurate model, otherwise the accumulated inaccuracies diminish the
accuracy of the far-into-the-future lookahead. Algorithm~\ref{alg:dplan}
shows the pseudocode for a $d$-step planner for the value prediction
network.

\begin{algorithm}[t]

  \begin{algorithmic}
            \Function{Q-Plan}{s, o, $d$}
            \State  $r, \gamma, V(s'), s' \rightarrow
            f^{core}_\theta(s, o)$ \Comment{Perform the four latent functions}
            \If{$d=1$}
            \State \Return{$r+\gamma V(s')$}
            \EndIf
            \State $A\leftarrow b$-best options based on $r+\gamma
            V_\theta(s')$ \Comment{See paper for other expansion strategies}
            \For{$o'\in A$}
            \State $q_{o'} \leftarrow $ Q-Plan$(s',o',d-1)$
            \EndFor
            \State \Return{$r+\gamma[\frac{1}{d}V(s')+\frac{d-1}{d}\max_{o'\in
                A} q_{o'}]$}
        \EndFunction
  \end{algorithmic}
  \caption{Multi-step planning~\cite{oh2017value}}\label{alg:dplan}

\end{algorithm}

The networks are trained 
with $n$-step Q-learning and TD search~\cite{silver2012temporal}. Trajectories are generated with an $\epsilon$-greedy policy
using the planning algorithm from Alg.~\ref{alg:dplan}. 
VPN achieved good results on Atari games such as Pacman
and Seaquest, outperforming model-free DQN, and outperforming
observation-based planning in stochastic domains.

Another relevant approach is
  presented in a sequence of papers by Hafner et
  al.~\cite{hafner2019dream,hafner2018learning,hafner2020mastering}. Their
  PlaNet and Dreamer approaches use  latent models based on a Recurrent
  State Space Model (RSSM), that consists of a 
transition model, an observation model, a variational encoder and a reward model,
to improve consistency between one-step and multi-step predictions in
latent space~\cite{karl2016deep,buesing2018learning,doerr2018probabilistic}.

The latent-model approach reduces the dimensionality of the
observation space. Dimensionality reduction is related to unsupervised 
learning (Sect.~\ref{sec:unsup}), and autoencoders 
(Sect.~\ref{sec:vae}). 
The latent-model approach is also related to world models, a term used
by Ha and Schmidhuber~\cite{ha2018recurrent,ha2018world}. 
World models are inspired
by the manner in which humans are thought to construct a mental model of
the world in which we live. Ha et al.\ implement world models using
generative recurrent neural networks that 
generate states for simulation using a variational autoencoder~\cite{kingma2013auto,kingma2019introduction} and a
recurrent network. Their approach learns a compressed spatial and temporal
representation of the environment. By using features extracted from the world model as inputs to
the agent, a compact and simple policy can be trained to solve a
task, and planning occurs in the compressed  world.
The term
\emph{world model}  goes back to 1990, see
Schmidhuber~\cite{schmidhuber1990making}. 

Latent models and world models achieve promising results and are,
despite their complexity, an active area of  research, see, for
example~\cite{zhang2021world}.
In the next section we will further discuss the performance of latent
models, but we will first look at two methods for planning with deep
transition models.

\subsection{Planning with the Model}
We have discussed in some depth methods to improve the
accuracy of models. 
We will now switch from how to \emph{create} deep models, to how to \emph{use} them. We will
describe two planning approaches that are designed to be forgiving for
models that contain inaccuracies.
%
The planners try to reduce the
impact of the 
inaccuracy of the model, for example, by planning
ahead with a limited horizon, 
and by re-learning and re-planning at each step of the
trajectory. We will start with planning with a limited horizon.

\subsubsection{Trajectory Rollouts and Model-Predictive Control}

At each planning step,  the local transition
model $T_a(s) \rightarrow s'$ computes the new state, using the local
reward to update the policy.
Due
to the inaccuracies of the internal model,  planning algorithms  that
perform many steps will quickly accumulate model
errors~\cite{gu2016continuous}.
Full  rollouts  of long, inaccurate, trajectories are therefore problematic.
We can  reduce the  impact of accumulated model errors by not
planning too far ahead. For example, Gu et al.~\cite{gu2016continuous} perform experiments with locally linear
models that roll out planning trajectories of length 5 to 10. This reportedly works well for
MuJoCo tasks Gripper and Reacher. 

In another experiment,  Feinberg et al.~\cite{feinberg2018model}  allow
imagination to a fixed look-ahead  
depth, after which value estimates 
are split into a near-future model-based
component and a distant future model-free component (Model-based value
expansion, MVE).\label{sec:mve} They experiment
with  horizons of 1, 2, and 10, and find that 10 generally performs best
on typical MuJoCo tasks such as Swimmer, Walker, and Half-cheetah. The
sample complexity in their experiments is better than model-free methods such as
DDPG~\cite{silver2014deterministic}. 
Similarly good results are reported
by others~\cite{janner2019trust,kalweit2017uncertainty}, with a model
horizon that is much shorter than the task horizon.

\subsubsection*{Model-Predictive Control}\label{sec:mpc}
Taking the idea  of shorter trajectories for planning than for learning 
further, we arrive at decision-time
planning~\cite{leonetti2016synthesis}, also known as Model-predictive control
(MPC)~\cite{kwon1983stabilizing,garcia1989model}. Model-predictive
control is a well-known approach  in 
process engineering, to control complex processes with
frequent re-planning over a limited time horizon. Model-predictive
control  uses the fact that  many real-world processes are
approximately linear over a small operating
range (even though they can be highly non-linear over a longer
range). In MPC the model is optimized for a limited time into the
future, and then 
it is re-learned after each environment step. In this way small
errors do not get a chance to accumulate and influence the outcome greatly.
Related to MPC are other local
re-planning methods. All try to reduce the impact of the use of an
inaccurate model by not planning too far into the future and by
updating the model frequently.
Applications are found in the automotive industry and in
aerospace, for example for terrain-following and obstacle-avoidance
algorithms~\cite{kamyar2014aircraft}.

MPC has been
used in various deep  model learning
approaches. Both Finn et al.\ and Ebert et al.~\cite{finn2017deep,ebert2018visual} use a form of
MPC in the planning for their Visual foresight robotic manipulation
system. The MPC part uses a model that
generates the corresponding sequence of future frames based on an
image to select the least-cost sequence of actions.
This approach is able to  perform
multi-object manipulation, pushing, picking and placing, and
cloth-folding tasks (which adds the difficulty of material that
changes shape as it is being manipulated).

Another approach is to use ensemble models for learning the transition
model, with MPC for planning. PETS~\cite{chua2018deep} uses
probabilistic ensembles~\cite{lakshminarayanan2017simple} for
learning, based on cross-entropy-methods
(CEM)~\cite{de2005tutorial,botev2013cross}.  In MPC-fashion  only the 
first action from the CEM-optimized  sequence is used, re-planning at every
environment-step. 
Many model-based approaches combine MPC and the ensemble method, as
we will also see in the overview in Table~\ref{tab:overview} at the end of the next
section. 
\begin{algorithm}[t]
  
        \begin{algorithmic}
            \State Initialize  the model $\hat{m}_\phi$ 
            \State Initialize an empty dataset $D$
            \For{$i=1, \ldots, I$}
                \State $D \gets E_a$ \Comment{sample action from environment}
                \State Train $\hat{m}_\phi(s,a)$ on $D$ minimizing the error by gradient descent 
                \For{$t=1, \ldots, T$ Horizon}
                    \Comment{planning}
                    \State{ $A_t \gets \hat{m}_\phi$}
                    \Comment{estimate optimal action sequence with  finite MPC horizon}
                    \State Execute first action $a_t$ from sequence $A_t$
                    \State $D\gets (s_t,a_t)$ 
                \EndFor
            \EndFor
        \end{algorithmic}

  \caption{Neural Network Dynamics
for Model-Based Deep Reinforcement Learning (based on \cite{nagabandi2018neural})} 
  \label{alg:nagabandi}
\end{algorithm}
Algorithm~\ref{alg:nagabandi} shows in pseudocode an example of  Model-predictive
control 
(based on~\cite{nagabandi2018neural}, only the model-based part is shown).\footnote{The code is at \url{https://github.com/anagabandi/nn_dynamics}.}

MPC is a simple and effective planning method that is well-suited for
use with inaccurate models, by restricting the planning horizon and by
re-planning. 
It has also been used with success in combination with
latent models~\cite{hafner2018learning,kaiser2019model}.
\index{model-predictive control}

It is now time to look at the final method, which is a very different
approach to planning.

%
\subsubsection{End-to-end Learning and Planning-by-Network}
Up until now, the \emph{learning} of the dynamics model and its \emph{use} are
performed by separate algorithms.
\label{sec:vin}
In the previous subsection differentiable  transition models were learned through
backpropagation and then the models were used by a conventional
hand-crafted procedural
planning algorithm, such
as  depth-limited search,  with  hand-coded
selection and backup rules.

A  trend in
machine learning is to replace \emph{all} hand-crafted algorithms by 
differentiable  approaches, that are trained by example,
end-to-end. These differentiable approaches often  are more general
and perform better than their hand-crafted versions.\footnote{Note
  that here we use the term \emph{end-to-end} 
  to indicate the use of differentiable methods for the learning and
  use of a deep dynamics model---to replace  hand-crafted planning
  algorithms to use the learned model. Elsewhere, in supervised
  learning, the term \emph{end-to-end} is used differently, to describe  learning 
  both features and their use from raw pixels for classification---to replace 
  hand-crafted feature recognizers to pre-process the raw pixels and
  use in a hand-crafted machine learning algorithm.} We could ask
the question if it would  be   
possible to make the planning phase differentiable as well? Or,  to
see if the planning rollouts can be implemented in a 
single computational model, the
neural network?

At first sight, it may seem strange to think of a neural
network as something that can  perform planning and backtracking,
since we often think of a neural network as a state-less mathematical
function.
%
%
Neural networks normally perform  transformation and filter
activities to achieve selection or classification.  Planning consists of action selection and
state unrolling. 
Note, however, that  recurrent neural networks and LSTMs contain implicit
state, making them a candidate to be used for planning (see
Sect.~\ref{sec:lstm}). Perhaps it is not so strange to try to
implement planning in a neural network.
Let us have a look at attempts to perform planning with  a neural network.


Tamar et al.~\cite{tamar2016value} introduced Value Iteration Networks
(VIN), convolutional networks for planning in Grid worlds. A \gls{VIN} is a 
differentiable multi-layer convolutional network that can execute the steps
of a simple planning algorithm~\cite{niu2018generalized}.
The core  idea it that in a Grid world, value
iteration
can be implemented by a multi-layer convolutional
network: each layer does a step of lookahead~(refer back to
Listing~\ref{lst:val-it} for value iteration). The value iterations are rolled-out in
the network layers $S$ with $A$ channels, and the CNN architecture
is shaped specifically for each problem task.
Through backpropagation the model learns the value
iteration parameters including the transition function. The aim is to
learn a general model, that can 
navigate in unseen environments.

Let us  look in more detail at the value iteration algorithm. 
It is a simple algorithm that consists of a doubly nested loop over
states and actions, 
calculating the sum of rewards $\sum_{s'\in S}
T_a(s,s')(R_a(s,s') + \gamma V[s'])$ and a subsequent maximization
operation $V[s] = \max_a(Q[s,a])$. This double loop is iterated to
convergence.
The insight is that each iteration  can  be implemented by passing the previous value
function $V_n$ and reward function $R$ through a convolution layer and
max-pooling layer. In this way, each channel in the convolution
layer corresponds to the Q-function for a specific action---the innermost loop---and
the convolution kernel weights correspond to the 
transitions. Thus by recurrently applying a convolution layer $K$ 
times, $K$ iterations of value iteration are  performed. 

The value iteration module is simply a neural network  that has the
capability of  approximating a value iteration computation. 
Representing value iteration in this form makes learning the MDP parameters and
 functions natural---by backpropagating through the network,
as in a standard CNN. 
In this way, the classic value iteration algorithm can be approximated
by a neural network.

Why  would we want to have
a fully differentiable algorithm that can only give an approximation,
if we have a perfectly 
good classic procedural implementation that can calculate the value function $V$
exactly?

The reason is  generalization.
The exact algorithm only works for known transition
probabilities. The neural network can learn $T(\cdot)$ when it is
not given, from the environment, and  it
learns the reward and value functions at the same time. By learning
all functions all at once in an end-to-end fashion, the dynamics and
value functions might be better integrated than when a separately
hand-crafted planning algorithm  uses the results of a learned
dynamics model. Indeed, reported results do indicate good
generalization  to unseen problem instances~\cite{tamar2016value}.

The idea of planning by gradient descent has existed for some
time---actually, the idea of learning \emph{all} functions by example has
existed for some time---several authors explored learning approximations of dynamics in
neural networks~\cite{kelley1960gradient,schmidhuber1990line,ilin2007efficient}. 
%
The VINs  can be used for discrete and continuous path planning, and have
been tried in  Grid world problems and natural language
tasks.

Later work has extended 
the approach to other applications of more irregular shape, by adding
abstraction
networks~\cite{schleich2019value,srinivas2018universal,silver2017predictron}.
The addition of latent models increases the power and versatility of  end-to-end learning of
planning and transitions even further. 
Let us look briefly in more detail at  one such extension of VIN,
to illustrate how  latent models and planning go together. TreeQN by Farquhar et
al.~\cite{farquhar2018treeqn} is a fully
differentiable model learner and planner, using observation
abstraction so that the approach works on  applications that are less
regular than mazes.

TreeQN consists of five differentiable functions, four of which we have
seen in the previous section in Value Prediction
Networks~\cite{oh2017value}, Fig.~\ref{fig:oh} on page~\pageref{fig:oh}.
\begin{itemize}
\item The encoding function consists of a series of convolutional
  layers that  embed the actual state in a lower dimensional state
  $s_{latent} \gets f^{enc}_{\theta_e}(s_{actual})$
\item The transition function uses  a
  fully connected layer per action to calculate the next-state
  representation $s'_{latent} \gets f^{trans}_{\theta_t}(s_{latent},
  a_i)^I_{i=0}$. 
\item The reward function predicts the immediate reward for every
  action $a_i \in A$ in state $s_{latent}$ using a ReLU layer 
  $r \gets f^{reward}_{\theta_r}(s'_{latent})$.  

\item The value function of a state is estimated with a vector of weights
  $V(s_{latent}) \gets w^\top s_{latent} + b$.
\item The backup function applies a softmax function\footnote{
The
  softmax function normalizes an input vector of real numbers to a
  probability distribution $[0,1]$; $ p_\theta(y|x) = \text{softmax}(f_\theta(x)) = \frac{e^{f_{\theta}(x)}}{\sum_k e^{f_{\theta,k}(x)}} $}
  recursively to
  calculate the tree backup value $b(x) \gets \sum_{i=0}^I x_i\,
  \mbox{softmax}(x)_i$.
\end{itemize}
These functions together can learn a model, and can also execute
$n$-step Q-learning, to use the model to update a policy. Further
details can be found in~\cite{farquhar2018treeqn} and the
GitHub code.\footnote{See
  \url{https://github.com/oxwhirl/treeqn} for the code of TreeQN.}
TreeQN has been applied on games such as box-pushing and some Atari games,
and outperformed model-free DQN.

A limitation of VIN is
that the tight connection between problem domain, iteration algorithm,
and network architecture limited the applicability to other problems.
Another system that addresses this limitation os Predictron. Like TreeQN, the   Predictron~\cite{silver2017predictron} introduces an abstract model to reduce
this limitation. 
As in VPN, the latent model  consists
of four differentiable components: a representation model, a next-state model, a reward model, and a
discount model.
The goal of the abstract model in Predictron is to
facilitate value prediction (not state prediction) or prediction of
pseudo-reward functions that can encode special events, 
such as \emph{staying alive} or \emph{reaching the next room}.
The planning part rolls forward its internal model $k$ steps. 
Unlike VPN, Predictron uses joint parameters. 
The Predictron has been applied to procedurally generated mazes and a
simulated pool domain. In both cases it out-performed model-free
algorithms.

End-to-end model-based learning-and-planning is an active area of
research. Challenges include understanding the relation between
planning and learning~\cite{anand2021procedural,grill2020monte}, achieving performance that is competitive
with classical planning algorithms and with model-free methods, and
generalizing the class of applications.  In Sect.~\ref{sec:hdenv} more methods
will be shown.


\subsubsection*{Conclusion}
In the previous sections we  have discussed two  methods to reduce the inaccuracy of
the model, and two methods to reduce the impact of the use of an
inaccurate model.
%
%
%
We have seen a  range of different approaches to
model-based algorithms. Many of the algorithms were developed
recently. Deep model-based reinforcement 
learning is an  active area of research. 

Ensembles and MPC have improved the performance of model-based
reinforcement learning.
The goal of latent or world models is to learn the essence of the domain,
reducing the dimensionality, and for
end-to-end, to also include the planning part in the learning. Their
goal is generalization in a 
fundamental sense. Model-free learns a policy of which action to take
in each state. Model-based methods learn the transition
model, from state (via action) to state. Model-free teaches you how to
best respond to actions in your world, 
model-based helps  you to understand your world. By learning the
transition model (and possibly even how to best plan with it) it is
hoped that new generalization methods can be learned.

The goal of model-based methods is to get to know the environment so
intimitely that the sample complexity can be reduced while staying
close to the solution
quality of model-free methods.  A second goal is
that the generalization power of the methods improves so much, that new classes of
problems can be solved. 
The literature is rich and contains many
experiments of these approaches on different environments.
Let us now look at the environments to see  if we have succeeded.

\section{High-Dimensional Environments}\label{sec:hdenv}

We have now looked in some detail at  approaches for 
deep model-based reinforcement learning. Let us now
change our perspective from the agent to the environment, and look  at 
the kinds of environments that can be solved with these
approaches.

\subsection{Overview of Model-Based Experiments}
The main goal of model-based reinforcement learning is to learn the transition
model accurately---not just the policy function that finds the best action, but
the function that finds the next state.  
By learning the full essence of the environment a substantial reduction of sample
complexity can be achieved. Also, the hope is that the model allows us
to solve new classes of problems. 
%
In this section we will try to answer the question if these approaches
have  succeeded. 

The answer to this question can be measured in training time and in
run-time performance. For \emph{performance}, most benchmark domains provide easily
measurable quantities, such as the score in an Atari game. For
model-based approaches, the scores achieved by state-of-the-art
model-free algorithms such as DQN, DDPG, PPO, SAC and A3C are a useful
baseline. For \emph{training time}, the reduction in sample complexity is an
obvious choice. However, many model-based approaches use a fixed
hyperparameter to determine the relation between external environment samples
and internal model samples (such as $1 : 1000$). Then the number of time
steps needed for high performance to be reached 
becomes an important measure, and this is indeed published by most
authors. For model-free methods, we often see time steps in the millions per
training run, and sometimes even billions.  With so many
time steps it becomes quite important how much 
processing each time step takes. For
model-based methods, individual time steps may take longer than for
model-free, since  more processing for learning and planning has
to be performed. In the end, wall-clock  time is 
important, and this is also often published. 

There are two additional questions.  First  we are interested in knowing whether a model-based approach allows new types
of problems to be solved, that could not be solved by model-free
methods. 
Second is the question of
brittleness. In many experiments the numerical results are quite
sensitive to 
different settings of hyperparameters (including the random seeds). 
This is the case in many model-free and 
model-based results~\cite{henderson2018deep}. However, when the transition model is
accurate, the variance may diminish, and some model-based approaches
might be more robust.\index{brittleness}\index{robustness}

\begin{table}[t]
  \begin{center}\footnotesize
    \begin{tabular}{lllllc} 
      {\bf Name}&{\bf Learning}&{\bf Planning}\qquad&{\bf Environment} &
                                                                   {\bf Ref}\\
      \hline\hline
      PILCO&Uncertainty& Trajectory
                                              &Pendulum &\cite{deisenroth2011pilco}\\ 
      iLQG& Uncertainty &MPC& Small&\cite{tassa2012synthesis}\\ 
      GPS& Uncertainty& Trajectory&Small&\cite{levine2014learning}\\ 
      SVG&Uncertainty& Trajectory& Small&\cite{heess2015learning}\\
VIN &CNN&e2e&Mazes&\cite{tamar2016value}\\
      VProp & CNN&e2e&Mazes&\cite{nardelli2018value}\\
    Planning &CNN/LSTM&e2e&Mazes&\cite{guez2019investigation}\\
     TreeQN &  Latent& e2e&Mazes&\cite{farquhar2018treeqn}\\
      I2A &Latent&e2e&Mazes&\cite{racaniere2017imagination}\\
              Predictron & Latent&e2e&Mazes&\cite{silver2017predictron}\\
              World Model &Latent & e2e&Car Racing&\cite{ha2018world}\\ \hline
       Local Model& Uncertainty& Trajectory\qquad& MuJoCo&\cite{gu2016continuous}\\
      Visual Foresight & Video Prediction&MPC&Manipulation&\cite{finn2017deep}\\
      PETS & Ensemble&MPC& MuJoCo&\cite{chua2018deep}\\
      MVE    & Ensemble& Trajectory& MuJoCo&\cite{feinberg2018model}\\
      Meta Policy &Ensemble& Trajectory & MuJoCo&\cite{clavera2018model}\\
      Policy Optim  &Ensemble& Trajectory &MuJoCo&\cite{janner2019trust}\\
              PlaNet &Latent& MPC&MuJoCo&\cite{hafner2018learning}\\
              Dreamer &Latent& Trajectory&MuJoCo&\cite{hafner2019dream}\\
              Plan2Explore
                &Latent&Trajectory&MuJoCo&\cite{sekar2020planning}\\
      L$^3$P&Latent&Trajectory&MuJoCo&\cite{zhang2021world}\\
      \hline
              Video-prediction &Latent&Trajectory&Atari&\cite{oh2015action}\\
              VPN &Latent&Trajectory &Atari&\cite{oh2017value}\\
              SimPLe &Latent&Trajectory&Atari&\cite{kaiser2019model}\\
              Dreamer-v2 &Latent& Trajectory&Atari&\cite{hafner2020mastering}\\
      MuZero &Latent&e2e/MCTS&Atari/Go&\cite{schrittwieser2020mastering}\\
      \hline\\
    \end{tabular}
    \caption{Model-Based Reinforcement Learning Approaches~\cite{plaat2021high}}\label{tab:overview}
  \end{center}
\end{table}

Table~\ref{tab:overview} lists 26 experiments with  model-based
methods~\cite{plaat2021high}. In addition to the name, the table provides an
indication of the type of model learning that the agent  uses, of the
type of planning, and of the application environment
in which it was  used. The categories in the table are described in the
previous section, where e2e means end-to-end.

In the table, the approaches are grouped by environment. At the
top are smaller applications such as mazes and navigation tasks. In
the middle are larger MuJoCo tasks. At the bottom are high-dimensional
Atari tasks. 
Let us look in more depth at the three groups of environments: 
small navigation, robotics, and Atari games.

\subsection{Small Navigation Tasks} 
We see that a few approaches use smaller 2D Grid world navigation tasks such 
as mazes, or block
puzzles, such as Sokoban, and Pacman. Grid world tasks are some of the oldest
problems in reinforcement learning, and they are used frequently to
test out new ideas. Tabular imagination approaches such as Dyna, and some latent model
and end-to-end learning and planning, have been evaluated with these environments. They
typically achieve good results, since the problems are of moderate
complexity.

Grid world navigation problems are quintessential sequential decision problems. 
Navigation problems are typically low-dimensional, and no
visual recognition is involved; transition functions are 
easy to learn.

Navigation tasks are also used
for latent model and end-to-end
learning. 
Three latent model approaches in Table~\ref{tab:overview} use navigation
problems.
I2A deals with model imperfections by introducing a latent model, based
on Chiappa et al.\ and Buesing et al.~\cite{chiappa2017recurrent,buesing2018learning}. I2A is applied to Sokoban and
Mini-Pacman by~\cite{racaniere2017imagination,buesing2018learning}.
Performance  compares favorably to  model-free learning and to planning
algorithms (MCTS).

Value iteration networks introduced the concept of end-to-end differentiable
learning and planning~\cite{tamar2016value,niu2018generalized}, 
after~\cite{kelley1960gradient,schmidhuber1990line,ilin2007efficient}. 
Through backpropagation the model learns to perform value
iteration. The aim  to learn a general model that can
navigate in unseen environments was achieved, although different
extensions were needed for more complex environments.

\subsection{Robotic  Applications}

Next, we look at  papers that use MuJoCo to model continuous
robotic problems.
Robotic problems  are high-dimensional  problems with continuous
action spaces. MuJoCo is used by most experiments in this category to 
simulate the physical behavior of robotic movement and the environment. 

Uncertainty modeling with ensembles and MPC re-planning
try to reduce or contain inaccuracies. 
The combination of ensemble
methods with MPC is well
suited for robotic problems, as we have seen in individual
approaches such as PILCO and PETS.

Robotic applications are more complex than Grid worlds; model-free
methods can take many time steps to find good policies. It is
important to know if model-based methods succeed in reducing sample
complexity in these problems. 
When we have a closer look at how well uncertainty modeling and MPC
succeed at achieving our first goal, we find a mixed picture.

A benchmark study by Wang et al.~\cite{wang2019benchmarking} looked
into the performance of ensemble methods and Model-predictive control
on MuJoCo tasks. It finds that these methods mostly find good
policies, and  do so in significantly fewer time 
steps than model-free methods, typcially in 200k time steps versus 1
million for model-free. So, it would appear that the lower sample
complexity is achieved.
However, they also note that per time step, the more complex model-based methods
perform more processing than the simpler model-free methods. Although
the sample complexity may be lower, the wall-clock time  is not,
and  model-free methods such as
PPO ans SAC are still much faster for many problems. Furthermore, the score that the
policy achieves varies greatly for different problems, and is
sensitive to different hyperparameter values.

Another finding is that in some experiments with a large number of
time steps, the performance of model-based methods plateaus well below
model-free performance, and the
performance of the model-based methods themselves differs
substantially. There is  a need for further research in deep
model-based methods, especially into robustness of results. More
benchmarking studies are needed that compare different methods.

\subsection{Atari Games Applications}\label{sec:dreamer}
Some experiments in Table~\ref{tab:overview} use the Arcade learning environment (ALE). ALE features
high-dimensional inputs, and provides one of the
most challenging  environments of 
the table.
Especially latent models  choose Atari games to showcase their
performance, and some do indeed achieve impressive results, in that
they  are  able to solve new problems, such as playing all 57
Atari games well (Dreamer-v2)~\cite{hafner2020mastering} and learning
the rules of Atari and chess (MuZero)~\cite{schrittwieser2020mastering}.

Hafner et al.\ published the papers
\emph{Dream to control: learning behaviors by latent imagination}, and
\emph{Dreamer v2}~\cite{hafner2019dream,hafner2020mastering}. Their
work extends the work on VPN and PlaNet with more advanced latent models and 
reinforcement learning methods~\cite{oh2017value,hafner2018learning}.
Dreamer uses an
actor-critic approach to learn behaviors that 
consider rewards beyond the horizon.  Values are
backpropagated through the value model, similar to
DDPG~\cite{lillicrap2015continuous} and
Soft actor critic~\cite{haarnoja2018soft}.

An important advantage of model-based reinforcement learning is that
it can generalize to unseen environments with similar
dynamics~\cite{sekar2020planning}. The Dreamer experiments showed that
 latent models are indeed more robust to 
unseen environments than model-free methods.
Dreamer is tested with applications from the DeepMind control suite
(Sect.~\ref{sec:dcs}).
\label{sec:muzero}

Value prediction networks are another latent approach. 
They outperform model-free DQN on mazes and Atari games such as Seaquest, QBert,
Krull, and Crazy Climber. 
Taking the development of end-to-end learner/planners such as VPN and
Predictron further is the work on 
MuZero~\cite{schrittwieser2020mastering,grimm2020value,hubert2021learning}. 
In
MuZero a new architecture is used to learn the transition functions for a
range of different games, from  Atari to the board games chess, shogi
and Go. MuZero learns
the transition model for all games from interaction with the
environment.\footnote{This is a somewhat unusual usage of board
  games. Most researchers use
  board games because the transition function is given (see next
  chapter). MuZero instead does \emph{not} know the rules of chess, but
  starts from scratch  learning the rules from interaction with the environment.}  
The
MuZero model includes different
modules: a representation, dynamics, and prediction function. 
Like AlphaZero, MuZero uses a refined version of  MCTS for
planning (see Sect.~\ref{sec:mcts} in the next chapter). 
This MCTS
planner is used in
a self-play training loop for policy improvement.
MuZero's achievements are impressive: it
is able to learn the rules of Atari games as well as   board games,
learning to play the games from scratch, in conjunction with  learning the
rules of the games. The MuZero
achievements have created follow up work to provide more insight into
the relationship between actual  and latent
representations, and to reduce the  computational demands~\cite{grimm2020value,vries2021visualizing,babaeizadeh2020models,hessel2021muesli,anand2021procedural,schrittwieser2021online,grill2020monte,ye2021mastering}.

Latent models reduce observational dimensionality to
a smaller model to perform planning in latent space. 
End-to-end learning and planning is able to learn new problems---the
second of our two goals: it is able to learn
to generalize navigation tasks, and to learn the rules of chess and
Atari.  These
are new problems, that are out of reach for model-free methods (although
the sample complexity of MuZero is quite large).

\subsubsection*{Conclusion}

In deep model-based reinforcement learning benchmarks drive progress. We
have seen  good results with ensembles and local re-planning in
continuous problems, and with latent models in discrete problems. In some
applications, both the first goal, of better sample complexity, and
the other goals, of learning new applications and reducing brittleness, are achieved.

The experiments used many different environments within the ALE and
MuJoCo suites, from hard to harder. 
In the next two chapters we will study multi-agent
problems, where we encounter a new set of benchmarks, with a state
space of many combinations, including hidden information and
simultaneous actions. These provide even more complex challenges for
deep reinforcement learning methods.

\subsection{\em Hands On: PlaNet Example}\label{sec:planet}

Before we go to the next chaper, let us take a closer look at how one of these methods achieves
efficient learning of a complex high-dimensional task. We will look at
PlaNet, a well-documented project by Hafner et al.~\cite{hafner2018learning}. Code is
available,\footnote{\url{https://github.com/google-research/planet}}
scripts are available, videos are available, and a blog is
available\footnote{\url{https://planetrl.github.io}} inviting us to
take the experiments further.
The name of the work is \emph{Learning latent dynamics from pixels}, which
describes what the algorithm does: use high
dimensional visual input, convert it to latent space, and plan in
latent space to learn robot locomotion dynamics.

\begin{figure}[t]
    \centering{\includegraphics[width=\textwidth]{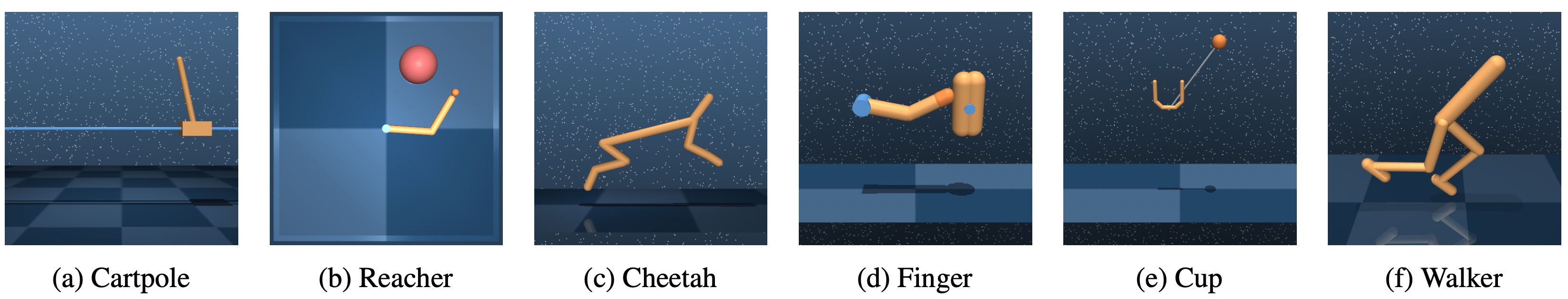}
     \caption{Locomotion tasks of PlaNet~\cite{hafner2018learning}}\label{fig:planet-exp}}
\end{figure}

PlaNet solves continuous control tasks that include  contact dynamics, partial
observability, and sparse rewards. 
The applications used in the PlaNet experiments are: (a) Cartpole (b)
Reacher (c) Cheetah (d) Finger (e) Cup and (f) Walker (see
Fig.~\ref{fig:planet-exp}).
The Cartpole task is a  swing-up task, with a fixed viewpoint. The
cart  can be out of sight, requiring the agent to remember
information from previous frames. The Finger spin task  requires
predicting the location of two separate objects and their interactions. 
The Cheetah task involves learning to run. It  includes contacts of
the feet with the ground that requires  a model to predict multiple futures.
The Cup task must catch a ball in a cup. It provides a sparse reward
signal once the ball is caught, requiring  accurate predictions far
into the future.
The Walker task involves a simulated robot that begins lying on the
ground, and must learn to stand up and then walk.
PlaNet
performs well on these tasks. On DeepMind control tasks it achieves
higher accuracy than an A3C or an D4PG agent. It reportedly does so
using 5000\% fewer interactions with the environment on average.  

It is instructive to experiment with PlaNet. The code can be found on
GitHub.\footnote{\url{https://github.com/google-research/planet}}
Scripts are available to run the experiments with simple one line
commands:
\begin{tcolorbox}
  \verb|python3 -m planet.scripts.train --logdir /path/to/logdir|
  \verb|--params '{tasks:  [cheetah_run]}'|
\end{tcolorbox}
As usual, this does require having the right versions of the right libraries 
installed, which may be a challenge and may require some creativity on
your part. The required versions are listed on the GitHub page.
The blog also
contains videos and pictures of what to expect, including comparisons to
model-free baselines from the DeepMind Control Suite (A3C, D4PG).

The experiments show the viability of the idea to use rewards and values
to compress actual states into lower dimensional latent states, and
then plan with these latent states. Value-based compression reduces
details in the high-dimensional actual states as noise that is not
relevant to improve the value function~\cite{grimm2020value}. To help
understand how the actual state map to the latent states, see, for
example~\cite{vries2021visualizing,li2017visualizing,karpathy2015visualizing}.

\section*{Summary and Further Reading}
\addcontentsline{toc}{section}{\protect\numberline{}Summary and Further Reading}
This has been a diverse chapter. We will summarize the chapter, and
provide references 
for further reading.  
\subsection*{Summary}

\label{sec:dis}

Model-free methods sample the environment using the rewards to learn
the  policy function, providing actions for all states for an environment. 
Model-based methods use the rewards to learn the transition function,
and then use planning methods to sample the policy from this internal
model. Metaphorically speaking: model-free learns how to \emph{act} in the
environment, model-based  learns how to \emph{be} the environment. The
learned transition model acts as a multiplier on the
amount of information that is used from each environment sample. A
consequence is that model-based methods have a lower sample
complexity, although, when the agent's transition model does not
perfectly reflect the environment's transition function, the
performance of the policy may be worse than a model-free policy (since
that always uses the environment to sample from).

Another, and perhaps more important aspect of the model-based
approach, is generalization. Model-based reinforcement learning builds
a dynamics model of the 
domain. This model can be used multiple times, for new problem
instances, but also for related problem classes. By learning the
transition and reward model, model-based reinforcement 
learning may be better at capturing the essence of a domain  than 
model-free methods, and thus be able to generalize to variations of
the problem.

Imagination showed how to learn
a model and use it to fill in extra samples based on the model (not
the environment). For problems where tabular methods work, imagination  
can be many times more efficient than model-free methods.

When the agent has access to the transition model, it can
apply reversible planning algorithms, in additon to one-way learning with
samples. There is a large literature on backtracking and tree-traversal
algorithms. Using a look-ahead of more than one step
can increase the quality of the reward even more.
When the problem size increases, or when we perform deep multi-step
look-ahead, the accuracy of the model  becomes critical. For
high-dimensional problems high capacity networks are used that require
many samples to prevent overfitting. 
Thus a trade-off exists, to keep
sample complexity low.

Methods such as PETS aim to take the uncertainty of the model into
account in order to increase modeling accuracy. Model-predictive
control methods re-plan at each environment step to prevent over-reliance on the
accuracy of the model.
Classical tabular approaches and Gaussian Process approaches have
been quite succesful in achieving low sample complexity for small
problems~\cite{sutton2018introduction,deisenroth2013survey,kober2013reinforcement}.

Latent models  observe that in many
high-dimensional problems the factors that influence changes in the
value function are often lower-dimensional. For example, the
background scenery
in an image may be irrelevant for the quality of play in a game, and
has no effect on the value.  Latent models
use an encoder to translate the high-dimensional actual state space
into a lower-dimensional latent state space. Subsequent planning and
value functions work on the (much smaller) latent space.

Finally, we considered end-to-end model-based algorithms. These fully
differentiable algorithms not only learn the dynamics model, but also
learn the planning algorithm that uses the model. 
The work on Value iteration networks~\cite{tamar2016value} inspired
recent work on end-to-end learning,
where both the  transition model and the planning algorithm are
learned, end-to-end. Combined with latent models (or World
models~\cite{ha2018world}) impressive results 
were achieved~\cite{silver2017predictron}, and the model and planning
accuracy was improved to the extent that tabula rasa self-learning of
game-rules was achieved, in Muzero~\cite{schrittwieser2020mastering} for 
both chess, shogi, Go, and Atari games.

\subsection*{Further Reading}
Model-based reinforcement learning promises more sample efficient
learning. The field  has a long history. For  exact tabular
methods Sutton's Dyna-Q is a classical approach that illustrates the
basic concept of model-based learning~\cite{sutton1990integrated,sutton1991dyna}.

There is an extensive literature on the  approaches that were
discussed in this chapter.  For
uncertainty modeling
see~\cite{deisenroth2011pilco,deisenroth2013gaussian,tassa2012synthesis}, and for
ensembles~\cite{opitz1999popular,chua2018deep,clavera2018model,janner2019trust,kurutach2018model,kaiser2019model,levine2013guided,levine2014learning,heess2015learning,finn2017deep}. For
Model-predictive
control see~\cite{nagabandi2018neural,garcia1989model,mayne2000constrained,kouvaritakis2016model,gu2016continuous,feinberg2018model,azizzadenesheli2018surprising}.

Latent models is an active field of research. Two of the earlier works
are~\cite{oh2015action,oh2017value}, although the ideas go back to
World models~\cite{ha2018world,kelley1960gradient,schmidhuber1990line,ilin2007efficient}. Later, a sequence of PlaNet and
Dreamer papers was
influential~\cite{hafner2018learning,hafner2019dream,sekar2020planning,hafner2020mastering,zhang2018solar}.

The literature on end-to-end learning and planning is also extensive,
starting with VIN~\cite{tamar2016value}, see~\cite{anthony2017thinking,silver2018general,silver2017mastering,silver2017predictron,nardelli2018value,guez2019investigation,farquhar2018treeqn,schrittwieser2020mastering,feng2020solving}.

As applications became more challenging, notably in robotics, other methods were developed,
mostly based on uncertainty, see for
surveys~\cite{deisenroth2013survey,kober2013reinforcement}. Later, as
high-dimensional problems became prevalent, latent 
and end-to-end methods were developed. 
The basis for the section on environments is an overview of recent
model-based approaches~\cite{plaat2021high}. Another survey is~\cite{moerland2020model}, a comprehensive
benchmark study is~\cite{wang2019benchmarking}.

\section*{Exercises}
\addcontentsline{toc}{section}{\protect\numberline{}Exercises}
Let us go to the Exercises.
\subsubsection*{Questions}
Below are
first some quick questions to check your understanding of this
chapter. For each question a simple, single sentence answer is sufficient.


\begin{enumerate}
\item What is the advantage of model-based over model-free methods?
\item Why may the sample complexity of model-based methods suffer in high-dimensional problems?
\item Which functions are part of the dynamics model?
\item Mention four deep model-based approaches.
\item Do model-based methods achieve better sample complexity than model-free?
\item Do model-based methods achieve better performance than model-free?
\item In Dyna-Q the policy is updated by two mechanisms: learning
  by sampling the environment and what other mechanism?
\item Why is the variance of ensemble methods lower than of the individual
  machine learning approaches that are used in the ensemble?
\item What does model-predictive control do and why is this approach
  suited for models with lower accuracy?
\item What is the advantage of planning with latent models over
  planning with actual models?
\item How are latent models trained?
\item Mention four typical modules that constitute the latent model.
\item What is the advantage of end-to-end planning and learning?
\item Mention two end-to-end planning and learning methods.
\end{enumerate}

\subsubsection*{Exercises}
It is now time to introduce a few programming exercises. The main
purpose of the exercises is to become more familiar with the methods that
we have covered in this chapter. By playing around with the algorithms
and trying out different hyperparameter settings you will develop some
intuition for the effect on performance and run time of the different
methods.

The experiments may become computationally expensive. You may want to
consider running them in the cloud, with Google Colab, Amazon AWS, or
Microsoft Azure. They may have student discounts, and they will have
the latest GPUs or TPUs for use with TensorFlow or PyTorch.

\begin{enumerate}
\item \emph{Dyna} Implement  tabular Dyna-Q for the Gym Taxi environment. Vary
  the amount of planning $N$ and see how performance is influenced.
\item \emph{Keras} Make a function approximation version of Dyna-Q and Taxi, with
  Keras. Vary the capacity of the network and the amount of
  planning. Compare against a pure model-free version, and note the
  difference in performance for different tasks and in computational demands. 
\item \emph{Planning} In Dyna-Q,  planning has so far been with single step model
  samples. Implement a simple depth-limited multi-step look-ahead
  planner, and see how 
  performance is influenced for the different look-ahead depths.
\item \emph{MPC} Read the paper by  Nagabandi et al.~\cite{nagabandi2018neural}
  and download the
  code.\footnote{\url{https://github.com/anagabandi/nn_dynamics}}
  Acquire the right versions of the libraries, and run the code with
  the supplied scripts, just for the MB (model-based) versions. Note
  that plotting is also supported by the scripts. Run with different
  MPC horizons. Run with different ensemble sizes. What are the
  effects on performance and run time for the different applications?
    
\item \emph{PlaNet} Go to the PlaNet blog and read it
(see previous section).\footnote{\url{https://planetrl.github.io}} Go to the PlaNet GitHub site and download and
install the code.\footnote{\url{https://github.com/google-research/planet}} Install the DeepMind control suite,\footnote{\url{https://github.com/deepmind/dm_control}} and all
necessary versions of the support libraries.

Run Reacher and Walker in PlaNet, and compare against the model-free methods
D4PG and A3C. Vary the size of the encoding network and note the
effect on performance and run time. Now turn off the encoder, and run
with planning on actual states (you may have to change network sizes
to achieve this). Vary the capacity of the latent model,
    and of the value and reward functions. Also vary the amount of
    planning, and note its effect.
  \item \emph{End-to-end} As you have seen, these experiments are computationally
    expensive. We will now turn to end-to-end planning and
    learning (VIN and MuZero). This exercise is also
    computationally expensive. Use small applications, such as small
    mazes, and  Cartpole. Find and download a MuZero implementation from GitHub
    and explore using the experience that you have gained from the
    previous exercises. Focus on gaining insight into the shape of the
    latent space. Try
    MuZero-General~\cite{muzero-general},\footnote{\url{https://github.com/werner-duvaud/muzero-general}}
      or a MuZero visualization~\cite{vries2021visualizing} to get insight
    into latent
    space.\footnote{\url{https://github.com/kaesve/muzero}} (This is a
    challenging exercise, suitable for a term project or thesis.)
\end{enumerate}


\chapter{Two-Agent Self-Play}\label{chap:given}\label{chap:self}
Previous chapters were concerned with how a single  agent can learn
optimal behavior for its environment. This chapter is
different. We  turn to problems where two agents operate whose
behavior  will both be modeled (and,
in the next chapter, more than two).

Two-agent problems are interesting for two reasons. First, the world
around us is full of active entities that interact, and modeling two
agents and their interaction is a
step closer to understanding the real world than modeling a single
agent. Second, in two-agent problems exceptional results were 
achieved---reinforcement learning agents teaching themselves to become
stronger than human world champions---and by studying these
methods we may find a way to achieve similar results  in other problems.

The kind of interaction that we model in this chapter is zero-sum: 
my win is your loss and vice versa.
These two-agent zero-sum dynamics  are fundamentally different from
single-agent dynamics. In single agent problems the environment lets you probe
it, lets you learn how it works, and lets you find good actions.
Although the environment may not be your friend, it is also
not working against you. In two-agent zero-sum problems 
the environment does try  to win from
you, it actively changes its replies to minimize your reward, based on
what it learns from your behavior.
When learning our optimal policy we should take all possible counter-actions into
account.
 
%
A popular way to do so is to implement the environment's actions  with self-play:
we replace   the
environment  by a copy of ourselves. In this way we let ourselves play
against an opponent  that has all the
knowledge that we currently have, and agents learn from eachother.

We start with a short review of two-agent problems, after which we
dive into  self-learning. We look at the situation when both
agents know the transition function perfectly, so that model accuracy
is no longer a problem. 
This is the case, for
example, in games such 
as chess and Go, where the rules of the game determine how we can go
from one state to another.  

In self-learning the environment is used 
to generate training examples for the agent to train a better
policy, after which the better agent policy is used in this
environment to train the agent, and
again, and again,
creating a virtuous cycle of self-learning and mutual improvement. It
is possible for an agent to teach itself to play a game without any 
prior knowledge at all, so-called  \emph{tabula rasa} learning, learning from a blank
slate. 

The self-play systems that we describe in this chapter use
model-based methods, and  combine planning and
learning approaches. There is
a planning algorithm that we have mentioned a few times, but have
not yet explained in detail.
In this chapter we will discuss Monte Carlo Tree Search, or MCTS, a
highly popular planning algorithm. MCTS can be used
in single agent and in two-agent situations, 
and  is the core of many successful applications,
including MuZero and the  self-learning AlphaZero series of programs. We will
explain how self-learning and self-play work in AlphaGo Zero, and why
they work so well. We 
will then discuss the concept of curriculum learning, which is behind
the success of self-learning.

The chapter is concluded with  exercises, a summary, and pointers
to further reading.

\section*{Core Concepts}
\begin{itemize}
\item Self-play
\item Curriculum learning
\end{itemize}

\section*{Core Problem}
\begin{itemize}
\item Use a given transition model for self-play, in order to become
  stronger  than the current best  players
\end{itemize}

\section*{Core Algorithms}
\begin{itemize}
\item Minimax (Listing~\ref{lst:minimax})
\item Monte Carlo Tree Search (Listing~\ref{lst:mcts})
\item AlphaZero tabula rasa learning (Listing~\ref{lst:selfplay})
\end{itemize}

\section*{Self-Play in Games}
 
\index{TD-Gammon}\index{backgammon}
We have seen in Chap.~\ref{chap:model} that when the agent has a
transition model of the environment, it can achieve greater
performance, especially when the model has high accuracy.  What if the
accuracy of our model were perfect, if the agent's transition function
is the same as the environment's, how far would that bring us? And what
if we   could improve our environment as part of our learning process, can
we then transcend our teacher, can  the sorcerer's apprentice outsmart the wizard?

To set the scene for this chapter, let us describe the first game where this has
happened: backgammon.\index{Tesauro, Gerald}
  
\begin{figure}[t]
  \begin{center}
        \includegraphics[width=5.5cm]{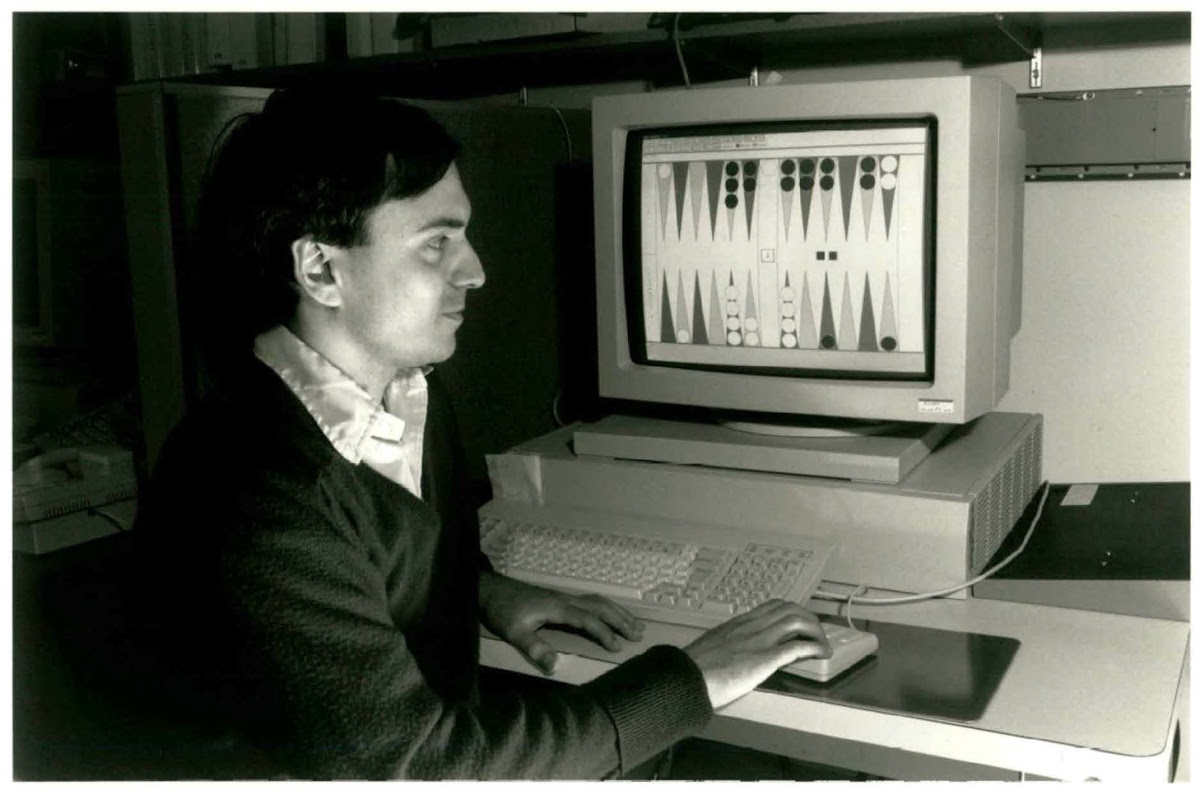}
\caption{Backgammon and Tesauro}\label{fig:bg}
\end{center}
\end{figure}

\subsubsection*{Learning to Play Backgammon}
In Sect.~\ref{sec:tdgam} we briefly discussed research into
backgammon. Already in the early 1990s, the program TD-Gammon achieved stable
reinforcement learning with a shallow 
network. This work was started at the end of the 1980s by Gerald Tesauro,
a researcher at IBM laboratories. Tesauro was faced with the
problem of getting a program to learn beyond the capabilities of any
existing entity. 
(In Fig.~\ref{fig:bg} we see Tesauro in front of his program; image by
IBM Watson Media.)

In the 1980s computing was  different. Computers were
slow,   datasets were small, and neural networks were  shallow. Against this background,
the success of Tesauro is quite  remarkable.

His programs were based on neural networks that learned good patterns of
play. His first program, Neurogammon, was trained using supervised
learning, based on games of human experts. In supervised
learning the model cannot become stronger than the human games
it is trained on. Neurogammon achieved an intermediate level of
play~\cite{tesauro1989neurogammon}. His second program,
TD-Gammon, was based on reinforcement learning, using temporal
difference learning and self-play. Combined with
hand-crafted heuristics and some planning, in 1992 it played at  human
championship 
level, becoming the first computer program to do so in a game of
skill~\cite{tesauro2002programming}. 

TD-Gammon is named after temporal difference learning because it updates  its
neural net after each move, reducing the difference between the
evaluation of previous and current positions. 
The neural
network used a single hidden layer with up to 80 units. 
TD-Gammon initially learned from a state of zero knowledge, \emph{tabula rasa}.
Tesauro describes TD-Gammon's  self-play as follows: \emph{The move that is selected is  the move with maximum expected
outcome for the side making the move. In other words, the neural
network is learning from the results of playing against itself. This
self-play training paradigm is used even at the start of learning,
when the network's weights are random, and hence its initial strategy
is a random strategy}~\cite{tesauro1995temporal}.


TD-Gammon performed tabula rasa learning, its
neural network weights initiliazed to small random numbers. It reached
world-champion level purely  by playing against
itself, learning the game as it played along. 

Such autonomous self-learning is one of the main goals of artificial
intelligence. TD-Gammon's
success  inspired many  researchers to try 
neural networks and self-play approaches, culminating eventually, many
years later, in high-profile results in  Atari~\cite{mnih2013playing} and 
AlphaGo~\cite{silver2016mastering,silver2017mastering}, which we will
describe  in this chapter.\footnote{
A modern
reimplementation of TD-Gammon in TensorFlow is available on GitHub at
\href{https://github.com/fomorians/td-gammon}{TD-Gammon} \url{https://github.com/fomorians/td-gammon}}

In Sect.~\ref{sec:self:env} two-agent zero-sum environments will be
described. Next, in Sect.~\ref{sec:self:agents} the tabula rasa
self-play method is described in detail. In
Sect.~\ref{sec:self:bench} we focus on the achievements of the
self-play methods. Let us now
start with two-agent zero-sum problems.

\section{Two-Agent Zero-Sum Problems}\label{sec:self:env}
Before we look  into self-play algorithms, let us look for a moment  at the
two-agent games that have  fascinated artificial intelligence
researchers  for such a long time.

Games come in many shapes and sizes. Some are easy, some are
hard.  
The characteristics of games are described
in a fairly standard taxonomy.
Important characteristics of games are: the number of players,
whether the game is zero-sum or non-zero-sum, whether it is perfect or imperfect information,
what the complexity of taking decisions is, and what the state space complexity
is. We will  look at these characteristics  in more detail.

\begin{itemize}
\item \emph{Number of Players}
One of the most important elements of a game is the number of players.
One-player games are normally called puzzles, and are modeled as a
standard MDP. The goal of a puzzle is
to find a solution. 
%
Two-player games are ``real'' games.    Quite a number of two-player 
games exist that provide a nice balance between being too easy and being
too hard for players (and for computer programmers)~\cite{czarnecki2020real}. 
Examples of two-player games that are popular in AI are chess,
checkers, Go, Othello, and shogi. 

Multi-player games are played by three or more players.   Well-known examples of
multiplayer games are the card games bridge and poker, and
strategy games such as Risk, Diplomacy, and StarCraft.

\item \emph{Zero Sum versus Non Zero Sum}
An important aspect of a game is whether it is competitive or
cooperative. Most two-player games are competitive: the win ($+1$) of
player A is the loss ($-1$) of player B. These games are called {\em
  zero sum\/} because the sum of the wins for the players remains a
constant zero. Competition is an important element in the real world,
and these games provide a useful model for the 
study of conflict and strategic behavior. 

In contrast, in cooperative games the players win if they can find win/win
situations. 
Examples of cooperative games are Hanabi, bridge, 
Diplomacy~\cite{kraus1994negotiation,de2018challenge}, 
poker and Risk. The next chapter will discuss multi-agent and
cooperative games. 

\item \emph{Perfect versus Imperfect Information}
In perfect information games all relevant information is known to all
players. This is the case in typical board games such as chess and
checkers. In imperfect information games some information may be hidden from
some players. This is the case in  card games such as bridge
and poker, where not all cards are known to all players. Imperfect
information games can be modeled as partially observable Markov
processes, POMDP~\cite{oliehoek2016concise,shani2013survey}.\index{POMDP}
A special form of (im)perfect information games are games of chance, such as
backgammon and Monopoly, in which dice play an important role.
There is no hidden information in these games, and these games are
sometimes considered to be perfect information games,
despite the uncertainty present at move time. Stochasticity is not the
same as imperfect information.



\item \emph{Decision Complexity}
The difficulty of playing a game depends on the complexity of the
game. The decision complexity is the number of end positions that define the
value (win, draw, or loss) of the initial game position (also known as
the  critical tree or proof
tree~\cite{knuth1975analysis}). The larger the
number of actions in a position, the larger the decision
complexity. Games with small board sizes such as tic tac toe ($3\times
3$) have a smaller complexity than games with larger boards, such as
gomoku ($19\times 19$). When the action space is very large, it can
often be
treated as a continuous action space. In poker, for example, the
monetary bets can be of any size, defining an action size that is
practically continuous.

\item \emph{State Space Complexity}
The state space complexity of a game is the number of legal positions
reachable from the initial position of a game. State space and
decision complexity are normally positively correlated, since games
with high decision complexity typically have high state space complexity. Determining the exact state space
complexity of a game is a nontrivial task, since  positions may be
illegal or unreachable.\footnote{For example, the maximal state space
  of tic tac toe is $3^9=19683$ positions (9 squares of 'X', 'O', or blank),
  where only 765 positions remain if we remove symmetrical and illegal
  positions~\cite{schaefer2002}.} For many games
approximations of the state space have been calculated.
In general, games with a larger state space complexity are harder to
play (``require more intelligence'') for humans and computers. Note
that the dimensionality of the states may not correlate with the size
of the state space, for example, the rules of some of the simpler Atari games
limit the number of reachable states, although the states themselves
are high-dimensional (they consist of many video pixels).
\end{itemize}

\subsubsection*{Zero-Sum Perfect-Information Games}\index{zero-sum game}
Two-person zero-sum games of perfect information, such as chess,
checkers, and Go,  are among the oldest
applications of artificial intelligence.
Turing and Shannon published the first ideas on how to write a
program to play chess more than 70 years
ago~\cite{turing1953digital,shannon1988programming}.
To study strategic reasoning in artificial intelligence,
these games are frequently used. Strategies,
or policies, determine the outcome. 
Table~\ref{tab:games} summarizes some of the games that have played an important
role in artificial intelligence research.

\begin{table}[t]
  \begin{center}
\begin{tabular}{lcccc}
  {\bf Name}&{\bf board}&{\bf state space}&{\bf zero-sum}&{\bf
                                                           information}\\
  \hline\hline
  Chess&$8\times 8$&$10^{47}$&zero-sum&perfect\\ 
  Checkers&$8\times 8$&$10^{18}$&zero-sum&perfect\\ 
  Othello&$8\times 8$&$10^{28}$&zero-sum&perfect\\ 
  Backgammon&24 &$10^{20}$&zero-sum&chance\\ 
  Go&$19\times 19$&$10^{170}$&zero-sum&perfect\\ 
  Shogi&$9\times 9$&$10^{71}$&zero-sum&perfect\\ 
  Poker&card&$10^{161}$&non-zero&imperfect\\ 
  \hline
\end{tabular}
  
  \caption{Characteristics of games}\label{tab:games}
  \end{center}
\end{table}

\begin{figure}[t]
  \begin{center}
\includegraphics[width=10cm]{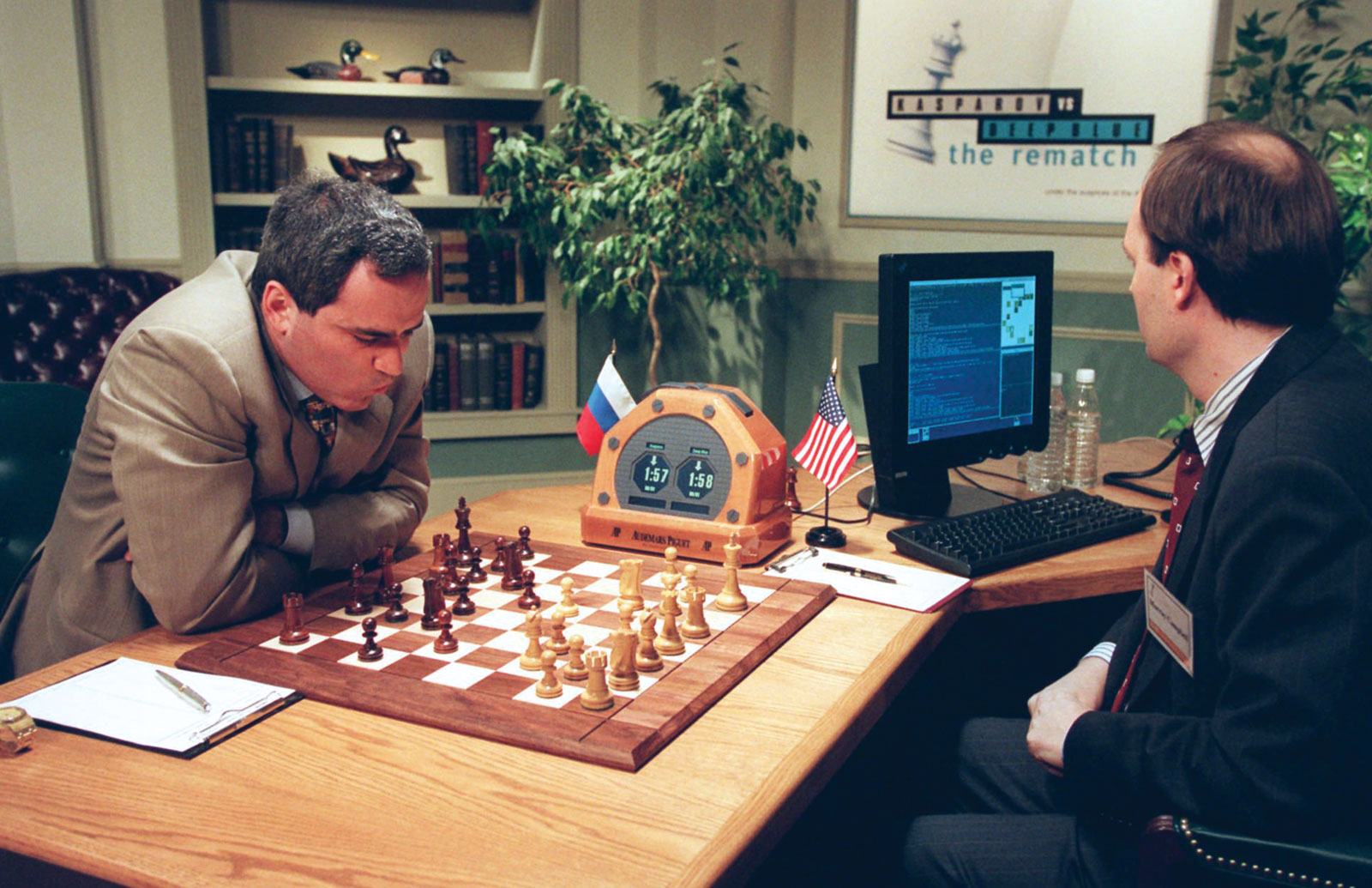}
  \caption{Deep Blue and Garry Kasparov in May 1997 in New York}\label{fig:kasparov}
  \end{center}
\end{figure}

\subsection{The Difficulty of Playing Go}

After the 1997 defeat of   chess world champion Garry Kasparov by IBM's Deep Blue
computer (Fig.~\ref{fig:kasparov}; image by Chessbase), the game of Go
(Fig.~\ref{fig:go}) became the next   benchmark 
game, the \emph{Drosophila}\footnote{\emph{Drosophila Melanogaster} is also
  known as the fruitfly, a favorite
  species of genetics researchers to test their theories, because
  experiments produce quick and clear answers.} of AI,  and research activity in Go
intensified significantly.

The game of Go is more difficult than chess. It is played on a larger
board ($19\times 19$ vs.\ $8\times 8$), the action space is larger
(around 250 moves available in a position versus some 25 in chess), the
game takes longer (typically 300 moves versus 70) and the state space complexity
is much larger: $10^{170}$ for Go, versus $10^{47}$ for chess.
Furthermore, rewards in Go are sparse. Only at the end of a long game,
after many moves have been played, is the outcome (win/loss) known. Captures are
not so frequent in Go, and no good efficiently
computable heuristic has been found. In chess, in contrast, the material
balance in chess can be calculated  efficiently, and gives a good
indication of how far ahead we are. For the computer, much of the playing in Go
happens in the dark. In contrast, for humans, it can be argued
that the visual patterns of Go may be somewhat easier to interpret
than the deep combinatorial lines of chess.

For reinforcement learning,  credit assignment in
Go is challenging. Rewards only occur after  a long
sequence of moves, and it is unclear which moves contributed the most to
such an outcome, or whether all moves contributed equally. Many games
will have to be played to acquire enough outcomes.
In conclusion, Go is  more difficult  to master with a
computer than chess.


Traditionally, computer Go programs followed the conventional chess
design of a minimax search  with a heuristic evaluation function, that,
in the case of Go,
was based on the influence of stones (see
Sect.~\ref{sec:minimax} and
Fig.~\ref{fig:influence})~\cite{millen1981programming}.
This chess approach, however, did not work for Go, or at least not well enough. The level of play
was stuck at mid-amateur level for many years.

The main problems were the  large branching factor, and the absence of
an efficient and good evaluation function. 

Subsequently, Monte Carlo Tree Search was developed, in 2006. MCTS is a variable depth
adaptive search algorithm, that did not need a heuristic function, but
instead used random playouts to estimate board strength.
MCTS programs caused the level of play to  improve from 10 kyu to 2-3
dan, and even stronger on the 
small $9\times 9$ board.\footnote{Absolute beginners
  in Go start at 30 kyu, progressing to 10 kyu, and advancing to 1
  kyu (30k--1k). Stronger amateur players then
  achieve 1 dan, progressing to 7 dan, the highest amateur rating for
  Go (1d--7d). Professional Go players have a rating from 1 dan to 9 dan,
  written as 1p--9p.}\index{kyu (Go rank)}\index{dan (Go rank)}
However, again, at that point, performance stagnated, and researchers
expected that world champion level play was still many years into the future.
Neural networks had been tried, but were slow, and did not improve
performance much.

\begin{figure}[t]
    \centering{\includegraphics[width=\textwidth]{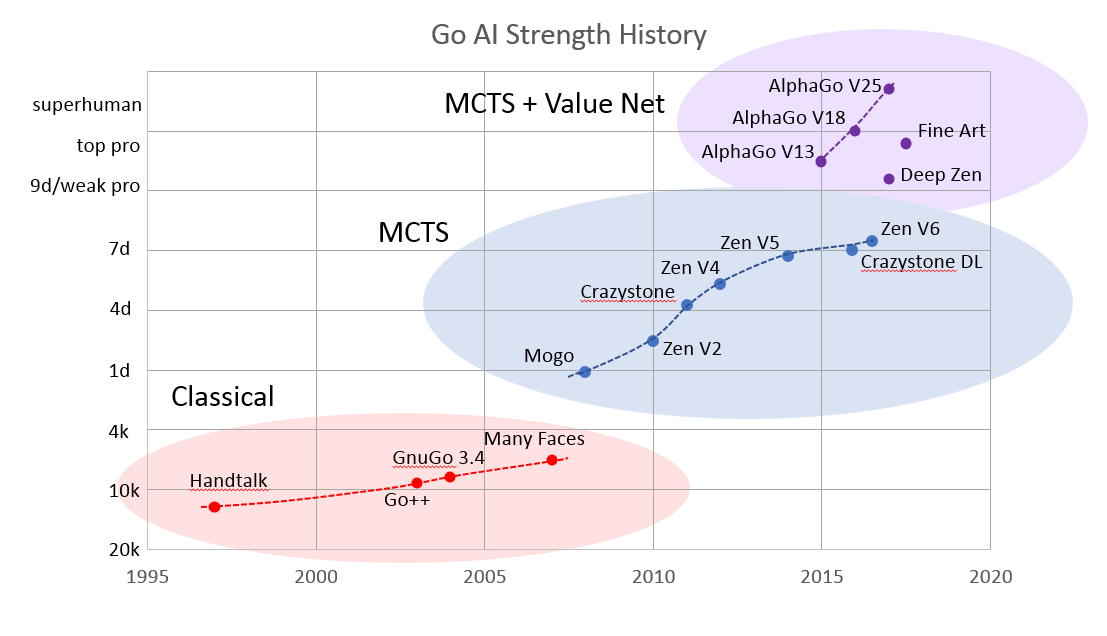}
     \caption{Go Playing Strength of Top Programs over the Years~\cite{reddit2017}}\label{fig:gostrength}}
\end{figure}

\subsubsection*{Playing Strength in Go}
Let us  compare the three programming paradigms  of a few
different Go
programs that have been written over the years (Fig.~\ref{fig:gostrength}).  The programs fall into three
categories. First are the
programs that use heuristic planning, the minimax-style
programs. GNU Go is a well-known example of this group of
programs. The heuristics in these programs are hand-coded. The level
of play of these programs was at
medium amateur level. Next  come the 
MCTS-based programs. They reached strong amateur level.  Finally come the AlphaGo
programs, in which MCTS is  combined with deep self-play. These reached super-human
performance. The figure also shows other  programs that follow a
related approach.

Thus, Go provided  a large and sparse state space, providing a highly
challenging  test, to see how far self-play with a perfect transition
function can come. Let us have a closer look at the  achievements of AlphaGo.

\subsection{AlphaGo Achievements}

\begin{figure}[t]
  \begin{center}
\includegraphics[width=5cm]{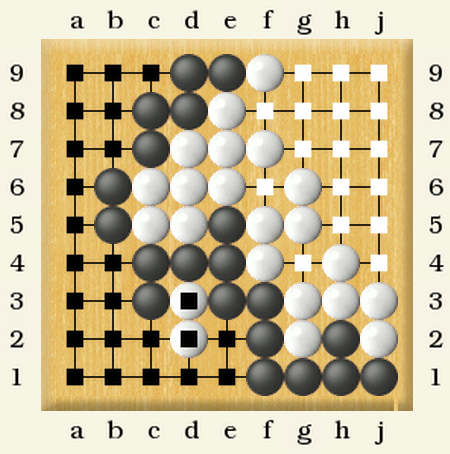}
  \caption{Influence in the game of Go. Empty intersections are marked
  as being part of Black's or White's Territory}\label{fig:influence}
  \end{center}
\end{figure}

In 2016, after decades of research, the  effort in Go paid off.
In the years 2015--2017 the DeepMind AlphaGo team 
played three matches
in which it beat all human champions that it played, Fan Hui, Lee
Sedol, and Ke Jie.
The breakthrough performance of AlphaGo came as a surprise.  Experts in computer games had expected grandmaster level play to be at least ten years away.

%
The techniques used in AlphaGo are the result of many
years of research, and cover a  wide range of topics.  The game
of Go worked very well as \emph{Drosophila}. 
Important new algorithms were developed, most notably Monte Carlo Tree
Search (MCTS), as well as major progress was made in deep reinforcement
learning. We will 
 provide a high-level  overview of the research  that
culminated in  AlphaGo (that beat the champions), and its successor,
AlphaGo Zero (that learns Go tabula rasa). First we will describe the
Go matches.

\begin{figure}[t]
\begin{center}
\includegraphics[width=\textwidth]{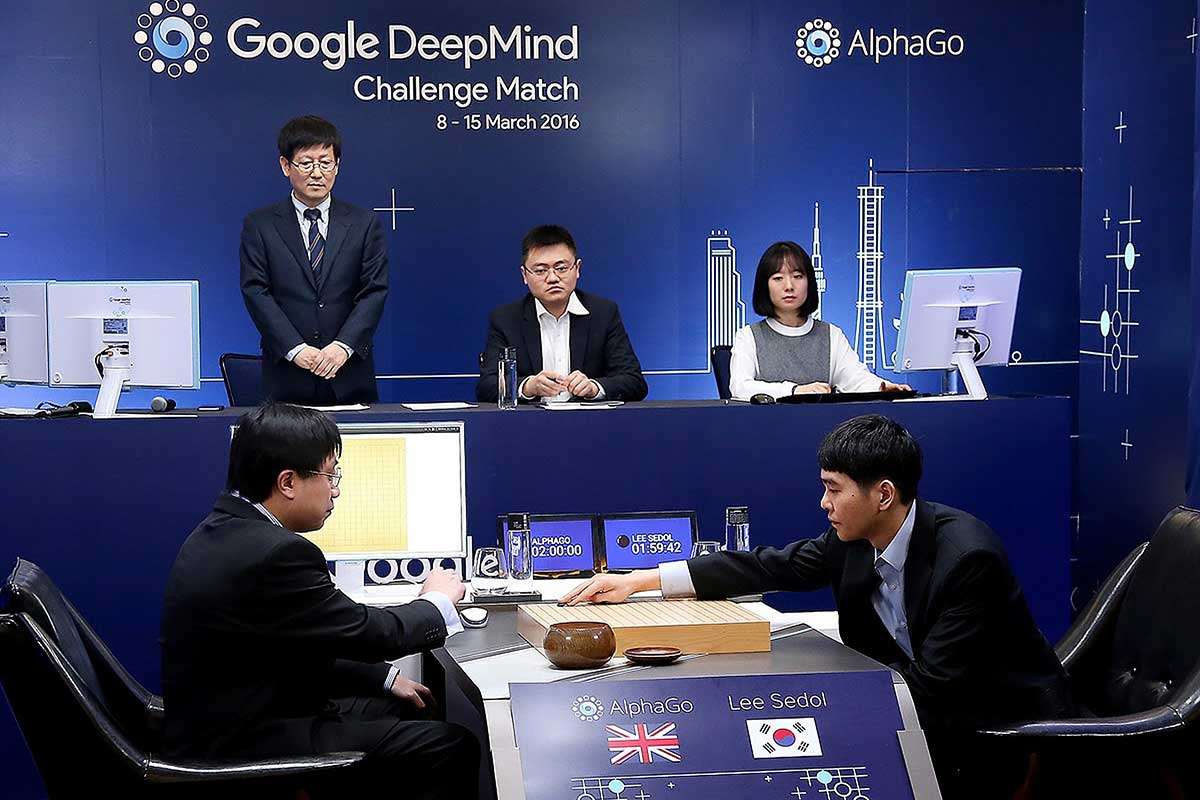}
\caption{AlphaGo versus Lee Sedol in 2016 in Seoul}\label{fig:fourseasons}
\end{center}
\end{figure}

The games against Fan Hui were   played in October 2015 in London
as part of the development effort of AlphaGo.  Fan Hui is the 2013, 2014, and
2015 European Go Champion, then rated at 2p dan.
The games against Lee Sedol were played in May 2016 in Seoul, and
were widely covered by the media (see Fig.~\ref{fig:fourseasons};
image by DeepMind). Although there is no official
worldwide ranking in international Go, 
in 2016 Lee Sedol was widely considered  one of the four best
players in the world.
A year later another match was played, this time in China, against the Chinese
champion Ke Jie, who was ranked  number one in
the Korean, Japanese, and Chinese ranking systems at the time of the
match.
All three matches were won convincingly by AlphaGo. Beating the best Go players appeared on the
cover of the journal Nature, see 
Fig.~\ref{fig:nature}.

\begin{figure}[t]
\begin{center}
\includegraphics[width=6cm]{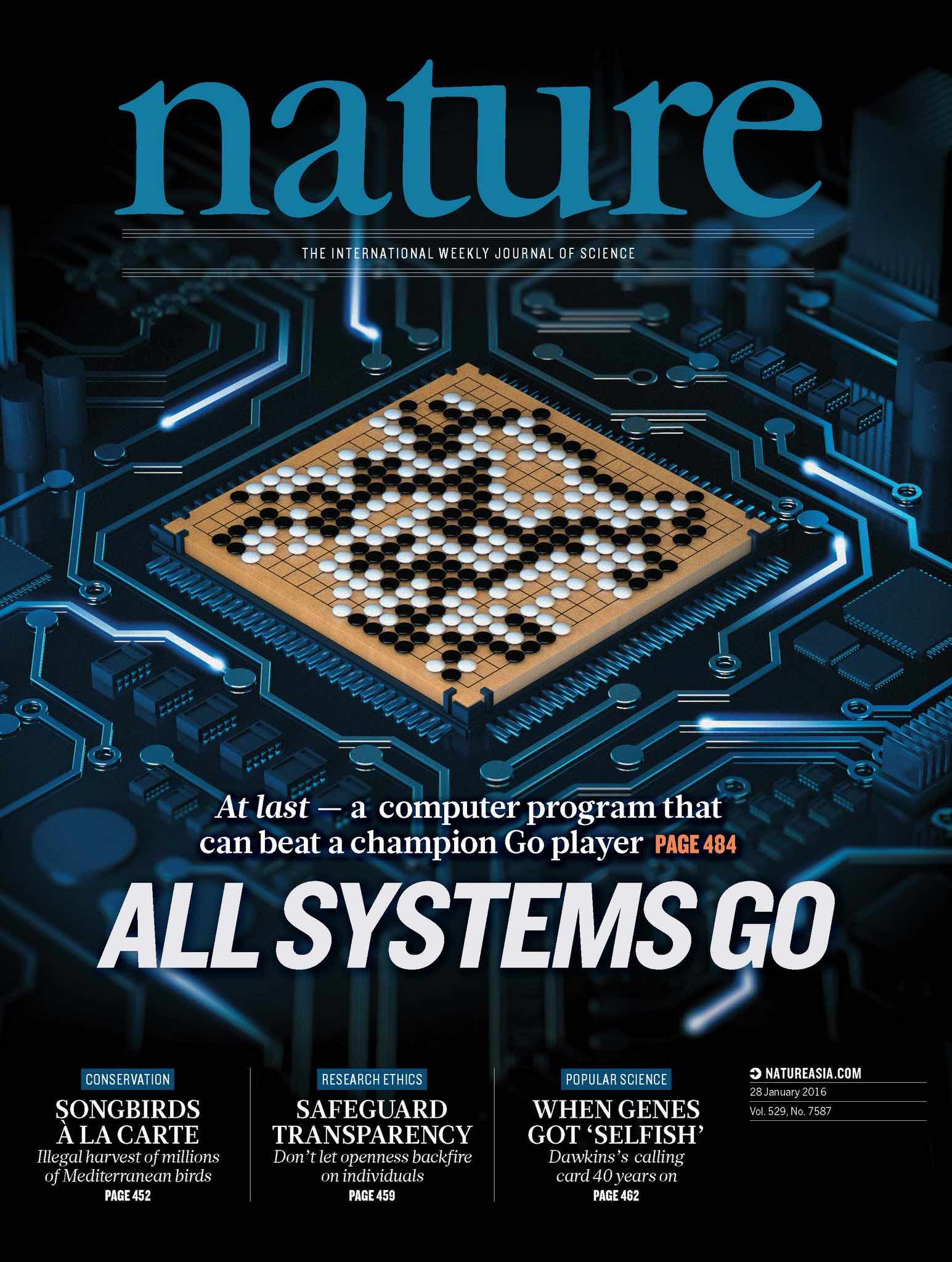}
\caption{AlphaGo  on the Cover of Nature}\label{fig:nature}
\end{center}
\end{figure}

The AlphaGo series of programs actually consists of three programs:
AlphaGo, AlphaGo Zero, and AlphaZero. AlphaGo is the program that beat
the human Go champions. It consists of a combination of supervised
learning from grandmaster games and from self-play games. The second
program, AlphaGo Zero, is a full re-design, based solely on
self-play. It performs tabula rasa learning of Go, and plays stronger
than AlphaGo. AlphaZero is a
generalization of this program that also plays chess and
shogi. Section~\ref{sec:self:bench} will describe the programs in more detail.

Let us now have an in-depth look at the self-play algorithms as
featured in AlphaGo Zero and AlphaZero.

\section{Tabula Rasa Self-Play Agents}\label{sec:self:agents}


Model-based reinforcement learning showed us that by learning a local
transition model, good  sample
efficiency can be achieved when the accuracy of the model is
sufficient. When we have perfect knowledge of the transitions, 
as we have in this chapter, 
then we can plan far into the future, without error. 

In regular agent-environment
reinforcement learning  the complexity of the environment does not
change as the agent learns, and as a consequence, the intelligence of the  agent's policy
is limited by the complexity of the environment.
However, in  self-play  a  cycle of mutual improvement
occurs;  the intelligence of the
environment  improves \emph{because} the agent is learning. With self-play, we can create a system that can
transcend the original environment, and keep growing, and growing, in a virtuous
cycle of mutual learning. Intelligence emerging out of nothing.
This is the kind of system that  is needed when we wish beat the best known entity in  a
certain domain,  since copying from a teacher will not help us to
transcend it.

Studying how such a high level of play is  achieved is interesting,
for three reasons: (1) it
is exciting to follow an AI success story, (2) it is interesting to see
which techniques were used and how it is possible to achieve
beyond-human intelligence, and (3) it is interesting to see if we can
learn a few techniques that can be used in other domains, beyond two-agent
zero-sum games, to see if we can achieve super-intelligence there as
well.


Let us have a
closer look at  the self-learning agent
architecture that is used by AlphaGo Zero. We will see that two-agent self-play
actually consists of three levels of self-play: move-level self-play,
example-level self-play, and tournament-level self-play.

First, we will discuss the general architecture, and how it creates
a cycle of virtuous improvement. Next, we will describe the levels in detail.

\subsubsection*{Cycle of Virtuous Improvement}

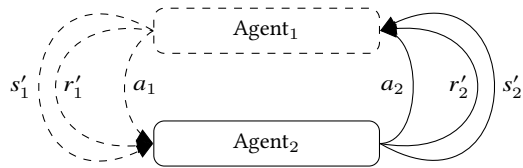
\begin{figure}[t]
\begin{center}
\begin{tikzpicture}[>=triangle 45,
  desc/.style={
		scale=1.0,
		rectangle,
		rounded corners,
		draw=black, 
		}]

  \node [desc,minimum width=3cm,minimum height=0.6cm] (tm) at   (0,0.5) {Agent$_2$};
  \node [desc,minimum width=3cm,dashed,minimum height=0.6cm] (pol) at   (0,2) {Agent$_1$};
  \draw (tm.west) edge[<-,in=170,out=190,dashed,looseness=3] node[right] {$r_1'$} (pol.west);
  \draw (tm.west) edge[<-,in=150,out=210,dashed,looseness=4] node[left] {$s_1'$} (pol.west);
  \draw (pol.east) edge[<-,out=0,in=0,looseness=1] node[left] {$a_2$} (tm.east);
  \draw (tm.east) edge[->,in=10,out=350,looseness=3] node[left] {$r_2'$} (pol.east);
  \draw (tm.east) edge[->,in=30,out=330,looseness=4] node[right] {$s_2'$} (pol.east);
  \draw (pol.west) edge[->,out=210,in=150,dashed,looseness=1] node[right] {$a_1$} (tm.west);

\end{tikzpicture}
\caption{Agent-Agent World}\label{fig:agent-agent}
\end{center}
\end{figure}

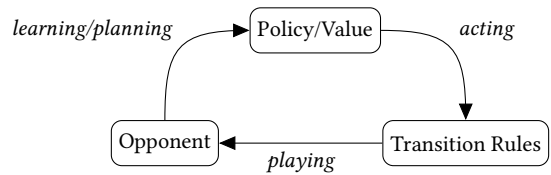
\begin{figure}[t]
  \begin{center}
    \begin{tikzpicture}[>=triangle 45,
  desc/.style={
		scale=1.0,
		rectangle,
		rounded corners,
		draw=black, 
		}]

  \node [desc,minimum height=0.6cm] (tm) at   (4,1) {Transition Rules};
    \node [desc,minimum height=0.6cm] (env) at   (0,1) {Opponent};
  \draw (env.east) edge[<-,in=180,out=0,looseness=2.5] node[below]
  {\em playing} (tm.west);

  \node [desc,minimum height=0.6cm] (pol) at   (2,2.5) {Policy/Value};
  \draw (env.north) edge[->,in=180,out=90,looseness=1.5] node[auto] (tour)
  {\em learning/planning} (pol.west);
  
  
  
  \draw (pol.east) edge[->,in=90,out=0,looseness=1.5] node[auto] {\em acting} (tm.north);
  
\end{tikzpicture}
    \caption{Playing with Known Transition Rules}\label{fig:given}
  \end{center}
\end{figure}

In contrast to the agent/environment model, 
we now have two agents
(Fig.~\ref{fig:agent-agent}). 
In comparison with the model-based
world of Chap.~\ref{chap:model} (Fig.~\ref{fig:given}) our learned model has been replaced by
perfect knowledge of the transition rules, and the environment is now
called opponent: the negative version of the same
agent playing the role of agent$_2$.

The goal in this chapter is to reach the highest possible performance
in terms of level of play, without using any hand-coded  domain
knowledge.  
In applications such as chess and
Go a perfect transition model is present. Together with a learned
reward function and a learned policy function, we can  create
a self-learning system in which a virtuous cycle of ever improving
performance occurs. Figure~\ref{fig:selfplay} illustrates  such a
system: (1) the searcher  
uses the evaluation network to estimate reward values and policy
actions, and  the search
results are used in games against the opponent in self-play, (2) the game results are
then collected in a buffer, which is used to train the evaluation
network in self-learning, and (3) by playing a tournament against a
copy of ourselves a virtuous cycle of ever-increasing
function improvement is created.

\subsubsection*{AlphaGo Zero Self-Play in Detail}
Let us look in more
detail at how self-learning works in AlphaGo Zero.
\begin{figure}[t]
  \centering{
    \vspace{2cm}
    \begin{tabular}{lr}
      \begin{minipage}{5cm}
        \vspace{-2cm}    
        \begin{tikzpicture}[>=triangle 45,
  desc/.style={
                scale=1.0,
		rectangle,
		rounded corners,
		draw=black,
		}]
  \node [desc] (search) at    (0,2) {{search}};
  \node [desc] (eval) at    (3,2) { {eval}};
  \node [desc] (train) at       (1.5,0.5) {{train}};

  \draw (search.south) edge[semithick,->,out=270,in=180,looseness=1] node[left] {game examples}  (train.west);
  
  \path (eval.south) edge[loop,semithick,<-,in=0,out=270,looseness=1]
  node[right] {new  net}  (train.east);
  \path[desc] (search.east) edge[semithick,loop above,<-,min distance=7mm,in=160,out=20,looseness=1] node[above] {reward}  (eval.west);
    \path[desc] (search.north) edge[semithick,loop above,->,min
    distance=13mm,in=45,out=135] node {opponent games}  (search.north);

\end{tikzpicture}
        \hspace{2cm} \ \ \ \ 
        \end{minipage}
      & \begin{minipage}{5cm}
        \vspace{-2cm}
        \hspace{2cm}
        \begin{tikzpicture}[>=triangle 45,
  desc/.style={
		scale=1.0,
		rectangle,
		rounded corners,
		draw=black, 
		}]

  \node [desc,minimum width=2cm,minimum height=0.6cm] (n0) at
  (0,1.5) {$\mbox{net}_0$};
  \node [desc,minimum width=2cm,minimum height=0.6cm] (n1) at
  (0,3) {$\mbox{net}_1$};
  \node [desc,minimum width=2cm,minimum height=0.6cm] (n2) at
  (0,4.5) {$\mbox{net}_2$};
  \node [desc,minimum width=2cm,minimum height=0.6cm] (n3) at
  (0,6) {$\mbox{net}_3$};
  \node [desc,minimum width=2cm,minimum height=0.6cm] (n4) at
  (0,7.5) {$\ldots$};

  \draw [semithick,->] (n0) -- node [left] {MCTS} (n1);
  \draw [semithick,->] (n1) -- node [left] {MCTS} (n2);
  \draw [semithick,->] (n2) -- node [left] {MCTS} (n3);
  \draw [semithick,->] (n3) -- node [left] {MCTS} (n4);

\end{tikzpicture}
        \end{minipage}
    \end{tabular}
     \caption{Self-play loop improving quality of net}\label{fig:selfplay}}
\end{figure}
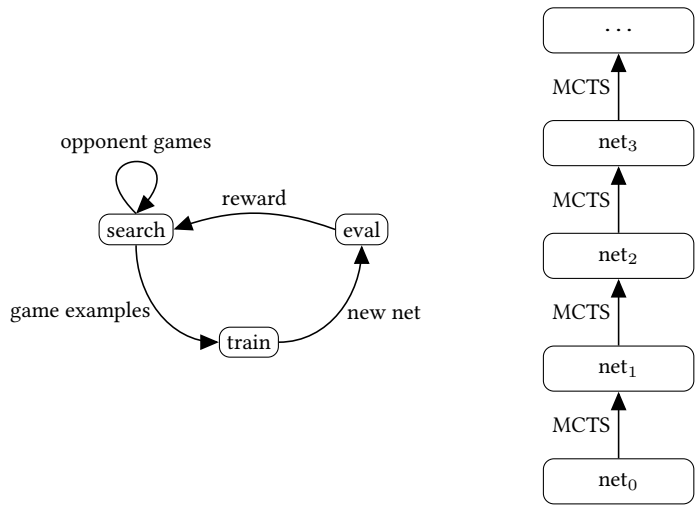
%
%
%
AlphaGo Zero uses a model-based actor critic approach with a planner
that improves  a single value/policy network. For 
policy improvement it uses MCTS, for learning a single deep
residual network with  a policy head and a value head 
(Sect.~\ref{sec:resnet}), see Fig.~\ref{fig:selfplay}. 
MCTS improves the quality of the training
examples in each iteration (left panel), and the net is trained with these better
examples, improving its quality (right panel). 

%
%
The output of
MCTS is used to train the evaluation network, whose output is then used as
 evalution function in that same MCTS. A   loop is
wrapped around the search-eval functions to keep training the network
with the game results, creating a learning curriculum. Let us put
these ideas
into pseudocode.

\subsubsection*{The Cycle in  Pseudocode}\index{self-play}\index{self-learning}

Conceptually   self-play is as ingenious as it is
elegant:  a double training loop around an MCTS player with a neural network
as evaluation and policy function that help MCTS.
%
Figure~\ref{fig:selfplay2} and Listing~\ref{lst:selfplay} show the
self-play loop in  detail. The 
numbers in the figure correspond to the line numbers in the pseudocode.

\lstset{label={lst:selfplay}}
\lstset{caption={Self-play pseudocode}}
\begin{lstlisting}[language=Python,float]
for tourn in range (1, max_tourns):      # curric. of tournaments
    for game in range(1, max_games):     # play a tourn. of games 
        trim(triples)       # if buffer full: replace old entries
        while not game_over():  # generate the states of one game
            move = mcts(board, eval(net))  # move is (s,a) pair
            game_pairs += move
            make_move_and_switch_side(board, move)
        triples += add(game_pairs, game_outcome(game_pairs))     # add to buf
    net = train(net, triples)   # retrain with (s,a,outc) triples
\end{lstlisting}

\begin{figure}[t]
    \centering{\begin{tikzpicture}[>=triangle 45,
  desc/.style={
                scale=0.9,
		rectangle,
		rounded corners,
		draw=black,
		}]
  \node [desc] (search) at    (0,2.15) {5 game\_pairs $\leftarrow$ \bf mcts};
  \node [desc] (eval) at      (6,2.15) {5 pol/val $\leftarrow$ {\bf eval}(net(state))};
  \node [desc] (train) at     (3,0.5) {9 net $\leftarrow$ {\bf train}(net, triples)};

  \draw (search.south) edge[semithick,->,out=270,in=180,looseness=1]
  node[left] {\small 8 triples}  (train.west);
  \path (eval.south) edge[loop,semithick,<-,in=0,out=270,looseness=1]
  node[right] {\small \ \ 1 tourn: {\em iterate with new\/} net}  (train.east);
  \path[desc] (search.east) edge[semithick,loop above,<-,min distance=7mm,in=160,out=20,looseness=1] node[above] {\small   policy/value}  (eval.west);
  \path[desc] (search.east) edge[semithick,loop below,->,min distance=7mm,in=200,out=340,looseness=1] node[below] {\small  state}  (eval.west);
    \path[desc] (search.north) edge[semithick,loop above,->,min
    distance=13mm,in=45,out=135] node {\small 2/4 \em game\_pairs}  (search.north);
\end{tikzpicture}
     \caption{A diagram of self-play with line-numbers}\label{fig:selfplay2}}
\end{figure}
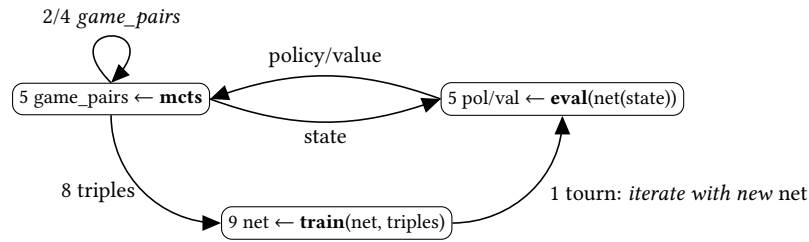

Let us perform an outside-in walk-through of this system. 
Line 1 is the main self-play loop. It controls how long the
execution of the curriculum of self-play tournaments will continue. 
Line 2 executes
the training episodes, the tournaments of self-play games after which
the network is retrained.
Line 4 plays such a game to create  ({\em state, action\/}) 
pairs for each move, and the outcome of the game. Line 5 calls MCTS to
generate an action 
in each  state. MCTS performs the simulations where it uses
the policy head of the net in P-UCT selection, and the value head of the
net at the MCTS
leaves. Line 6 appends the state/action pair to the list of game
 moves. Line 7 performs the move on the board, and switches color
to the other player, for the next move in the while loop. At line 8 a
full game has ended, and the outcome is known. Line 8 adds the outcome of each game to the
({\em state, action})-pairs, to make the ({\em state, action, outcome})-triples for the network to train
on. Note that since the network is a two-headed policy/value net,
both an action and an outcome are needed for network  training.
On the last
line this triples-buffer
is then used to 
train the network. The newly trained network is  used in the next
self-play iteration as the evaluation function by the searcher. 
With this net another 
tournament is played, using the searcher's look-ahead to generate a
next batch of higher-quality examples, resulting in a sequence of stronger and stronger
networks (Fig.~\ref{fig:selfplay} right panel).

In the pseudocode we see the three self-play loops where the principle
of playing against a copy of yourself is used:
\begin{enumerate}
\item \emph{Move-level}:  in the MCTS playouts, our opponent actually is a copy
  of ourselves (line 5)---hence, self-play at the level of \emph{game moves}
\item \emph{Example-level}:  the input for self-training the approximator for the policy
  and the reward functions is generated by our own games (line 2)---hence,
  self-play at the level of the  \emph{value/policy network}.
 \item \emph{Tournament-level}:  the self-play loop creates a training curriculum
   that starts tabula rasa and ends at world champion level. The
   system trains at the level of the \emph{player} against itself (line 1)---hence, self-play, of the third kind.
\end{enumerate}
All three of these levels  use their own kind of self-play, of which we
will  describe the details in the following sections.
We start with move-level self-play.

\subsection{Move-Level Self-Play}\index{minimax}\index{heuristic}\index{search-eval}\index{evaluation function}\label{sec:minimax}

At the innermost level, we 
use the agent to play against itself, as its own opponent. Whenever it
is my opponent's turn to move, I play its 
move, trying to find the best move for my opponent (which will be the
worst possible move for me). This scheme uses the same knowledge for
player and opponent. This is different from the real world, where the
agents are different, with different brains, different reasoning
skills, and different experience. Our scheme is symmetrical: when we assume that our agent plays
a strong game, then the opponent is also assumed to play strongly, and
we can hope to learn from the strong counter play. (We thus assume that
our agent plays with the same knowledge as we have; we are not trying
to consciously exploit opponent weaknesses.)\footnote{There is
  also research into opponent modeling, where we try to exploit our
  opponent's
  weaknesses~\cite{he2016opponent,billings1998opponent,ganzfried2011game}. Here,
  we assume an identical opponent, which often works best in chess and
  Go.}

\subsubsection{Minimax}
This principle of generating the counter play by playing yourself
while switching perspectives has been used since the start of
artificial intelligence. It is known as minimax.

The games of chess, checkers and Go are challenging games. 
The architecture that has been used to program  chess and checkers
players has been the same since the earliest paper designs of
Turing~\cite{turing1953digital}: a search routine based on minimax
which searches to a certain depth, and an evaluation
function to estimate the score of  board positions using
heuristic rules of thumb when this depth is reached. In chess and checkers, for
example, the number of pieces on the board of a player is a crude but 
effective approximation of the strength of a state for that
player.  
Figure~\ref{fig:se-arch} shows a diagram of this classic
search-eval architecture.\footnote{Because the agent knows the transition
  function $T$, it can calculate the new state $s'$ for each action
  $a$. The reward $r$ is calculated at terminal states, where it is
  equal to the value $v$. Hence, in
  this diagram the search function provides the state to the eval
  function. See~\cite{turing1953digital,plaat2020learning} for an explanation of the
  search-eval architecture.}

\begin{figure}[t]
\begin{center}
\begin{tikzpicture}[>=triangle 45,
  desc/.style={
		scale=1.0,
		rectangle,
		rounded corners,
		draw=black, 
		}]

  \node [desc,minimum width=3cm,minimum height=0.6cm] (tm) at   (0,0.5) {eval};
  \node [desc,minimum width=3cm,minimum height=0.6cm] (pol) at   (0,2) {search};
  \draw (pol.east) edge[<-,out=330,in=30,looseness=1] node[right] {\em value} (tm.east);
  \draw (pol.west) edge[->,out=210,in=150,looseness=1] node[left]
  {\em action/state} (tm.west);

\end{tikzpicture}
\caption{Search-Eval Architecture of Games}\label{fig:se-arch}
\end{center}
\end{figure}
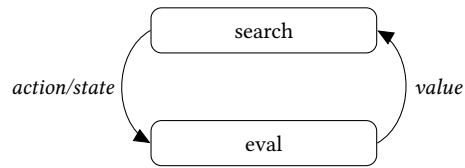

Based on this 
principle many successful search algorithms have been developed, of
which alpha-beta is the best known~\cite{knuth1975analysis,pearl1984heuristics}.
Since the size of the state space is exponential in the depth of
lookahead, however,  many enhancements had to be developed to manage
the size of the state space and to allow  deep lookahead to
occur~\cite{plaat2020learning}.

\begin{figure}[t]
  \begin{center}
    \begin{tikzpicture}
\tikzstyle{max node}=[rectangle,draw,inner sep=6]
\tikzstyle{min node}=[circle,draw,inner sep=4]

\node[max node](0){2}
          child[sibling distance=30mm]{node[min node]{1}
                  child[sibling distance=10mm]{node[max node]{6}}
                  child[sibling distance=10mm]{node[max node]{1}}
                  child[sibling distance=10mm]{node[max node]{3}}}
          child[sibling distance=30mm]{node[min node]{2}
                  child[sibling distance=10mm]{node[max node]{3}}
                  child[sibling distance=10mm]{node[max node]{4}}
                  child[sibling distance=10mm]{node[max node]{2}}}
          child[sibling distance=30mm]{node[min node]{1}
                  child[sibling distance=10mm]{node[max node]{1}}
                  child[sibling distance=10mm]{node[max node]{6}}
                  child[sibling distance=10mm]{node[max node]{5}}};
\end{tikzpicture}
\caption{Minimax tree}\label{fig:minimax}
\end{center}
\end{figure}

The word \emph{minimax} is a  contraction of maximizing/minimizing
(and then reversed for easy  pronunciation). It means that in zero-sum
games the two players alternate making moves, and that on even moves,
when player A is to choose a move, the best move is the one that
maximizes the score for player A, while on odd moves the best move
for player B is the move that minimizes the score for player
A.

Figure~\ref{fig:minimax} depicts this situation in a tree. The
score values in the nodes are chosen to show how minimax works. At the
top is the root of the tree, level 0, a square node where player A is to
move.

Since we assume that all players rationally choose the best
move, the value of the root node is determined by the value of the
best move, the maximum of its children. Each child, at level 1, is a
circle node where player B chooses its best move, in order to minimize
the score for player A. The leaves of this tree, at level 2, are again
max squares (even though there is no child to choose from
anymore). Note how for each circle node the value is the minimum of
its children, and for the square  node, the value is the maximum
of the tree circle nodes.

\index{NNUE}
Python pseudocode for a recursive minimax procedure
is shown in Listing~\ref{lst:minimax}. Note the extra hyperparameter
\verb|d|. This
is the search depth counting upwards from the leaves. At depth $0$ are the leaves, where
the heuristic evaluation function is called to score the
board.\footnote{The heuristic evaluation function is originally a
  linear combination of hand-crafted heuristic rules, such as material
  balance (which side has more pieces) or center control. At first, the linear combinations
  (coefficients) were not only hand-coded, but also
  hand-tuned. Later they were trained by supervised
  learning~\cite{baxter2000learning,quinlan1983learning,thrun1995learning,fogel2005further}.
  More recently,
  NNUE was introduced  as a non-linear neural network to
  use as evaluation function in an alpha-beta framework~\cite{nasu2018efficiently}.}
Also note that the code for making
moves on the board---transitioning actions into the new states---is not
shown in the listing. It is assumed to happen inside the children
dictionary. We frivolously mix
actions and states in these sections, since  an action fully
determines which  state will follow.
(At the end of this chapter, the exercises  provide more detail about
move making and unmaking.) 

\lstset{label={lst:minimax}}
\lstset{caption={Minimax code~\cite{plaat2020learning}}}
\begin{lstlisting}[language=Python,float]
INF = 99999 # a value assumed to be larger than eval ever returns

def minimax(n, d):
    if d <= 0:
        return heuristic_eval(n)
    elif n['type'] == 'MAX':
        g = -INF
        for c in n['children']:  
            g = max(g, minimax(c, d-1))
    elif n['type'] == 'MIN':
        g = INF
        for c in n['children']:  
            g = min(g, minimax(c, d-1))
    return g

print("Minimax value: ", minimax(root, 2))

            

        
\end{lstlisting}

AlphaGo  Zero uses MCTS, a more advanced search algorithm
than minimax,
that we will discuss shortly.


\subsection*{Beyond Heuristics}\index{alpha-beta}\index{transposition table}


Minimax-based procedures traverse the state space by recursively
following all actions in each state that they
visit~\cite{turing1953digital}. Minimax works just
like a standard depth-first search procedure, such as we have been
taught in our undergraduate algorithms and data structures
courses. 
It is a straightforward, rigid, approach, that searches all branches of the node to
the same search depth. 

To focus the search effort on promising parts of the tree,
researchers have subsequently introduced many algorithmic
enhancements, such as alpha-beta cutoffs, iterative deepening, 
transposition tables, null windows, and null
moves~\cite{knuth1975analysis,plaat1996best,korf1985depth,slate1983chess,donninger1993null}.

\begin{figure}[t]
  \begin{center}
    \includegraphics[width=5cm]{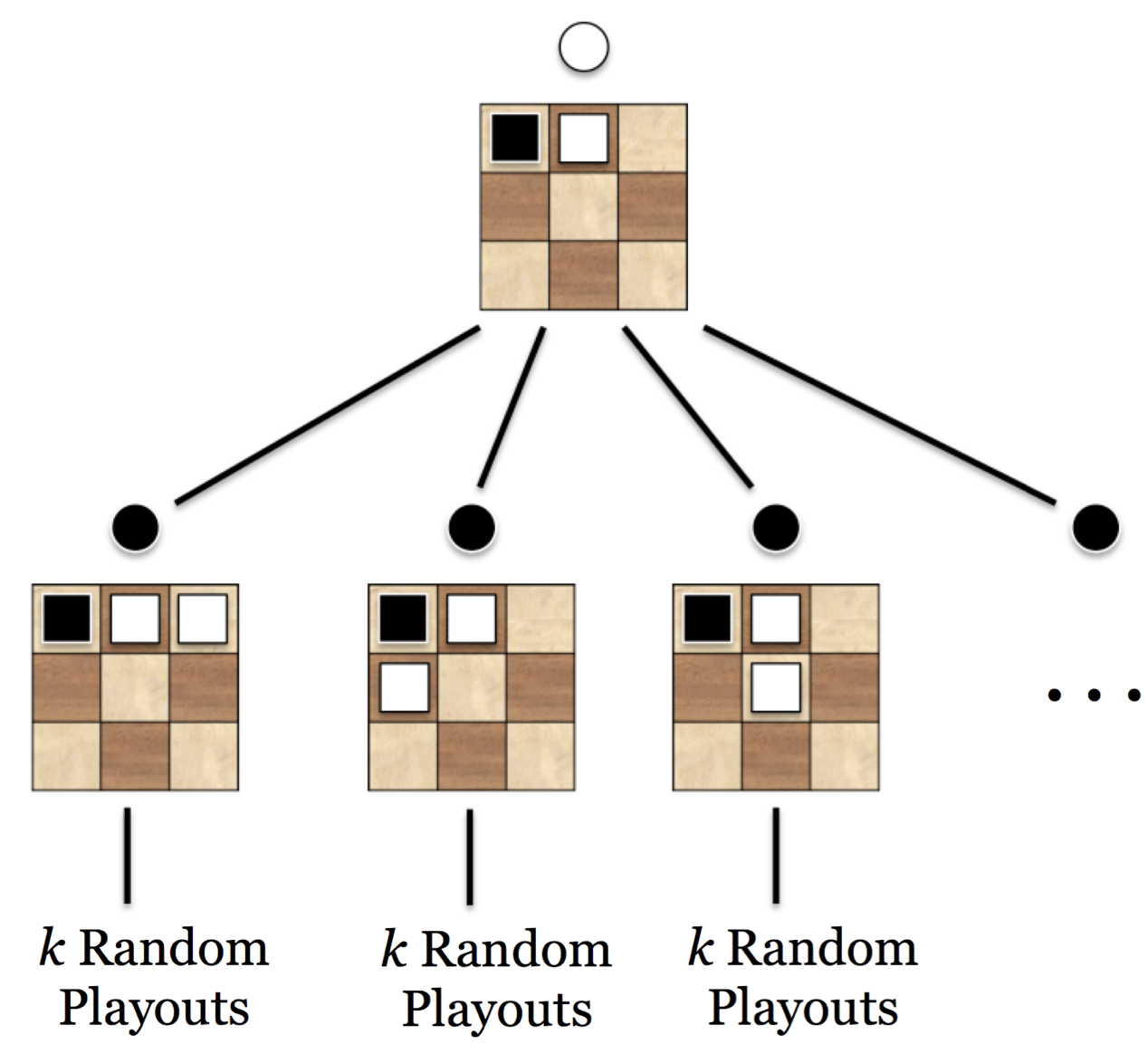}
  \caption{Three Lines of Play \cite{beljaards2017}}\label{fig:linesofplay}
  \end{center}
\end{figure}

In the early 1990s experiments with a different approach started,
based on random
playouts of a single line of play~\cite{abramson1990expected,bouzy2004monte,brugmann1993monte}
(Fig.~\ref{fig:linesofplay} and \ref{fig:treepath}). In
Fig.~\ref{fig:treepath} this different approach is illustrated. We see
a  search of a single line of
play versus a search of a full subtree. It turned out that
averaging many such playouts could also be used to approximate the
value of the root, in addition to the classic recursive tree search approach. In 2006, a tree version of this
approach was introduced that proved successful in Go. This algorithms
was called Monte Carlo Tree
Search~\cite{coulom2006efficient,browne2012survey}. Also in that year
Kocsis and Szepesv\'ari created a selection rule for the
exploration/exploitation trade-off  that performed well and 
converged to the minimax value~\cite{kocsis2006bandit}. Their  rule is called
UCT, for upper confidence bounds applied to
trees.

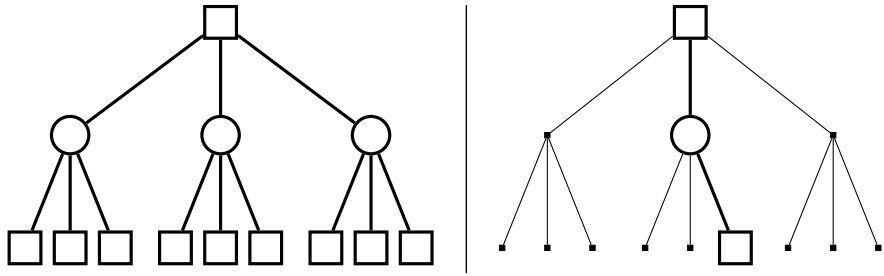
\begin{figure}[t]
  \begin{center}
\begin{tabular}{c|c}
  \begin{tikzpicture}
\tikzstyle{max node}=[rectangle,draw,inner sep=6,very thick]
\tikzstyle{min node}=[circle,draw,inner sep=5,very thick]
\tikzstyle{cut node}=[rectangle,draw,fill,inner sep=2,very thick]
\node[max node](0){}
          child[sibling distance=20mm,very thick]{node[min node]{}
                         child[sibling distance=6mm,very thick]{node[max node]{}}
                         child[sibling distance=6mm,very thick]{node[max node]{}}
                         child[sibling distance=6mm,very thick]{node[max node]{}}}
          child[sibling distance=20mm,very thick]{node[min node]{}
                         child[sibling distance=6mm,very thick]{node[max node]{}}
                         child[sibling distance=6mm,very thick]{node[max node]{}}
                         child[sibling distance=6mm,very thick]{node[max node]{}}}
          child[sibling distance=20mm,very thick]{node[min node]{}
                         child[sibling distance=6mm,very thick]{node[max node]{}}
                         child[sibling distance=6mm,very thick]{node[max node]{}}
                         child[sibling distance=6mm,very thick]{node[max node]{}}};
                     \end{tikzpicture} 
\ \ \ \ \ & \ \ \ \ \
  \begin{tikzpicture}
\tikzstyle{max node}=[rectangle,draw,inner sep=6]
\tikzstyle{min node}=[circle,draw,inner sep=5]
\tikzstyle{cut node}=[rectangle,draw,fill,inner sep=1]
\node[max node,very thick](0){}
          child[sibling distance=19mm,thin]{node[cut node]{}
                         child[sibling distance=6mm,thin]{node[cut node]{}}
                         child[sibling distance=6mm,thin]{node[cut node]{}}
                         child[sibling distance=6mm,thin]{node[cut node]{}}}
          child[sibling distance=19mm,very thick]{node[min node]{}
                         child[sibling distance=6mm,thin]{node[cut node]{}}
                         child[sibling distance=6mm,thin]{node[cut node]{}}
                         child[sibling distance=6mm,very thick]{node[max node]{}}}
          child[sibling distance=19mm,thin]{node[cut node]{}
                         child[sibling distance=6mm,thin]{node[cut node]{}}
                         child[sibling distance=6mm,thin]{node[cut node]{}}
                         child[sibling distance=6mm,thin]{node[cut node]{}}};
                     \end{tikzpicture}
\end{tabular}  
  
  \caption{Searching a Tree versus Searching a Path}\label{fig:treepath}
  \end{center}
\end{figure}
\subsubsection{Monte Carlo Tree Search}
Monte Carlo Tree Search  has two main advantages over minimax
and alpha-beta. First, MCTS is based on averaging single lines of play,
instead of recursively traversing subtrees. The computational
complexity of a path from the root to a leaf is polynomial in the
search depth. The computational complexity of a tree is exponential in
the search depth. Especially in
applications with many actions per state it is much easier to manage
the search time with an algorithm that expands one path at a
time.\footnote{Compare chess and Go: in chess the typical number of
  moves in a position is 25, for Go this number is 250. A chess-tree
  of depth 5 has $25^5=9765625$ leaves. A Go-tree of depth 5 has
  $250^5=976562500000$ leaves. A depth-5 minimax search in Go would
  take prohibitively long; an MCTS search of 1000 expansions
  expands the same number of paths from root to leaf in both games. } 

Second, MCTS does not need a heuristic evaluation function. It plays
out a line of play in the game from the root to an end position. In
end-positions the score of the game, a win or a loss, is known. By
averaging many of these playouts   the value of
the root is approximated. Minimax has to cope with an exponential search tree, which
it cuts off after a certain search depth, at which point it uses the
heuristic to estimate the scores at the leaves. There are, however,
games where no efficient heuristic evaluation function can be found. In this
case MCTS has a clear advantage, since it works without a heuristic
score function.

MCTS has proven to be successful in many different applications. 
Since its introduction in 2006  MCTS has transformed the field of
heuristic search. Let us see in more detail how it works.

\index{Monte Carlo Tree Search}\index{MCTS}\label{sec:mcts}
Monte Carlo Tree Search  consists of four
operations: select, expand, playout, and backpropagate (Fig.~\ref{fig:mcts}). The third operation (playout)
 is also called {\em rollout}, {\em simulation}, and {\em sampling}. Backpropagation
is sometimes called {\em backup}. Select is the downward policy action
trial part, backup is the upward
error/learning part of the algorithm. We will discuss the operations
in more detail in a short while.

\begin{figure}[t]
    \centering{
    \includegraphics[width=\textwidth]{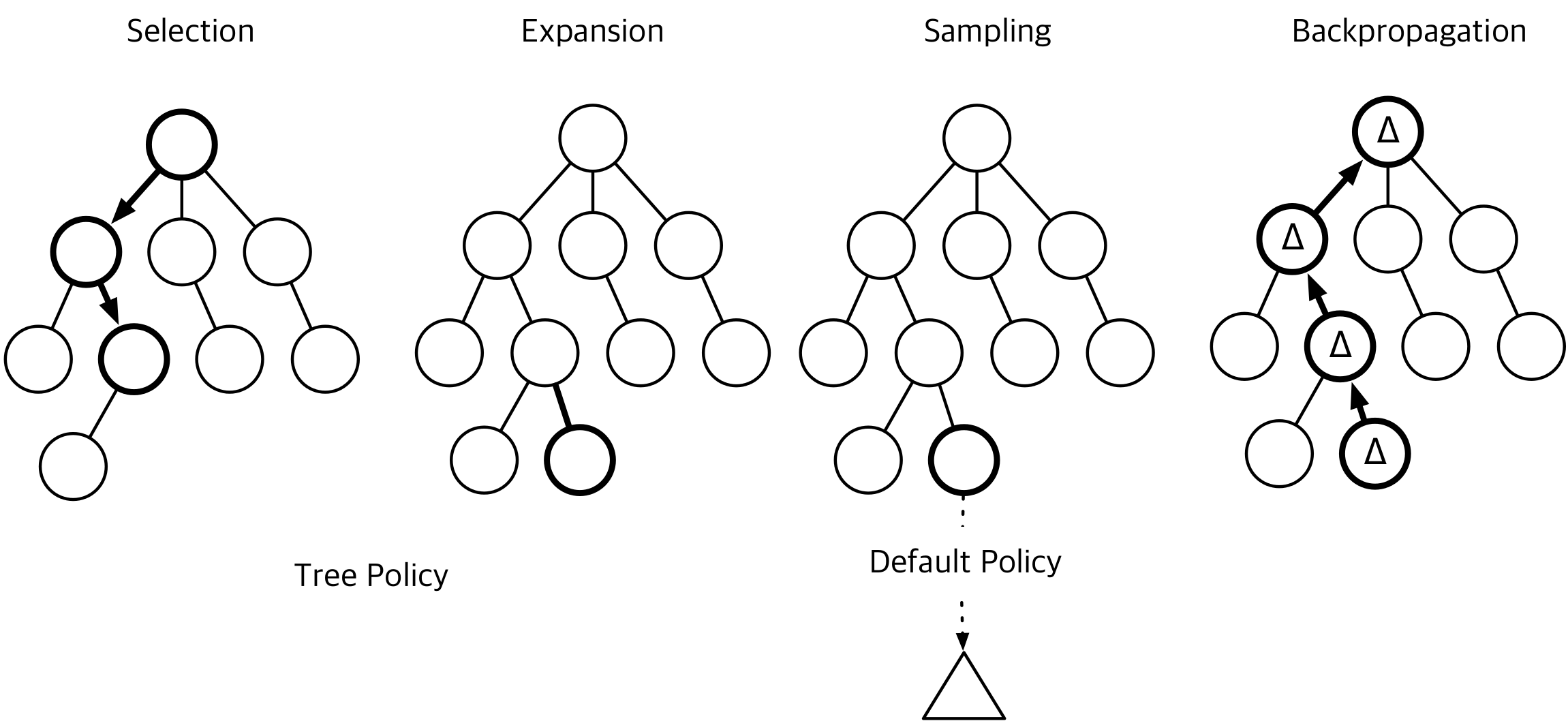}
     \caption{Monte Carlo Tree Search \cite{browne2012survey}}\label{fig:mcts}}
\end{figure}

MCTS is a succesful planning-based reinforcement learning algorithm, with
an advanced exploration/exploitation selection rule.
MCTS starts from the initial state $s_0$, using the transition function to
generate successor states. In MCTS the state
space is traversed iteratively, and the tree data structure is built in a
step by step fashion, node by node, playout by playout.  A typical size of an MCTS search
is to do 1000--10,000 iterations. In MCTS each iteration starts at
the root $s_0$, traversing a path in the  tree down to the leaves using a
selection rule, expanding a new node, and performing a random playout. The
result of the playout is then propagated back to the root. During the
backpropagation, statistics at all internal nodes are updated. These
statistics are then used in future iterations by the selection rule
to go to the currently most interesting part of the tree.

The statistics consist of two counters: the win count $w$ and the visit
count $v$. During backpropagation, the visit count $v$ at all nodes
that are on the path back from the leaf to the root are
incremented. When the result of the playout was a win, then the win
count $w$ of those nodes is also incremented. If the result was a loss,
then the win count is left unchanged.

The selection rule uses the win rate $w/v$ and the visit count $v$ to decide
whether to exploit high-win-rate parts of the tree or to explore
low-visit-count parts. An often-used selection rule is UCT
(Sect.~\ref{uct}). It is this selection rule that governs  
the exploration/exploitation trade-off in MCTS.

\subsubsection*{The Four MCTS Operations}
Let us  look in more detail at the four operations. Please refer to
Listing~\ref{lst:mcts} and Fig.~\ref{fig:mcts}~\cite{browne2012survey}.
%
%
As we see in the figure and the listing, the main 
steps are repeated as long as there is time left. Per step, the activities 
are as follows.

\lstset{label={lst:mcts}}
\lstset{caption={MCTS pseudo-Python \cite{browne2012survey,czarnogorski2018monte}}}
\begin{lstlisting}[language=Python,float]
def monte_carlo_tree_search(root):
    while resources_left(time, computational power):
        leaf = select(root) # leaf = unvisited node 
        simulation_result = rollout(leaf)
        backpropagate(leaf, simulation_result)
    return best_child(root) # or: child with highest visit count

def select(node):
    while fully_expanded(node):
        node = best_child(node)    # traverse down path of best UCT nodes
    return expand(node.children) or node # no children/node is terminal 

def rollout(node):
    while non_terminal(node):
        node = rollout_policy(node)
    return result(node) 

def rollout_policy(node):
    return pick_random(node.children)

def backpropagate(node, result):
    if is_root(node) return 
    node.stats = update_stats(node, result) 
    backpropagate(node.parent)
   
def best_child(node, c_param=1.0):
    choices_weights = [
        (c.q / c.n) + c_param * np.sqrt((np.log(node.n) / c.n))    # UCT
        for c in node.children
    ]
    return node.children[np.argmax(choices_weights)]
\end{lstlisting}

\begin{enumerate}
\item \emph{Select}\index{MCTS!select}
In the selection step the tree is traversed from the root node down
until a leaf of the MCTS search tree is reached where a new child is selected that is not part of
the tree yet. At each internal state the selection rule is
followed to determine which action to take and thus which state to
go to next. The UCT rule works well in many applications~\cite{kocsis2006bandit}.

The selections at these states are part of the policy $\pi(s)$ of
actions of  the state. 

\item \emph{Expand}\index{MCTS!expand}
Then, in the expansion step, a child is added to the tree. In most
cases only one child is added. In some MCTS versions all successors of a leaf
are added to the tree~\cite{browne2012survey}.

\item \emph{Playout}\index{MCTS!playout}\label{sec:playout}\index{pattern database}
Subsequently, during the playout  step random moves are played
in a form of self-play 
until the end of the game is reached. 
(These
nodes are not added to the MCTS tree, but their search result is, in
the backpropagation step.) The reward $r$ of 
this simulated game is $+1$ in case of a win for the first 
player, $0$ in case of a draw, and $-1$ in case of a win for the
opponent.\footnote{Originally, playouts were random (the Monte Carlo part in
the name of MCTS) following
Br\"ugmann's~\cite{brugmann1993monte} and Bouzy and
Helmstetter's~\cite{bouzy2004monte} original approach. In practice,
most Go playing programs improve on the random playouts by using
databases of small $3\times 3$ patterns with best
replies and other fast heuristics~\cite{gelly2006modification,coulom2007monte,chaslot2010monte,silver2009reinforcement,culberson1998pattern}. 
Small amounts of domain knowledge are used after all, albeit not in
the form of a heuristic evaluation function.} 

\item \emph{Backpropagation}\index{MCTS!backup}
In the backpropagation step, reward $r$ is propagated back upwards in
the tree, through the 
nodes that were traversed down previously. Two counts are
updated: the visit count, for all nodes, and the win count, depending
on the reward value. Note that in a two-agent game, nodes in the MCTS
tree alternate color. If white has won, then only white-to-play
nodes are incremented; if black has won, then only the black-to-play nodes.


MCTS is on-policy: the values that are backed up are those of the
nodes that were selected. 
\end{enumerate}


\subsubsection*{Pseudocode}

Many websites
 contain
useful 
resources on MCTS, including  example
\href{https://int8.io/monte-carlo-tree-search-beginners-guide/}{code}
(see Listing~\ref{lst:mcts}).\footnote{\url{https://int8.io/monte-carlo-tree-search-beginners-guide/}} The
 pseudocode in the listing is  
from an example  program for game play. The MCTS algorithm can be coded in many different
ways.
For 
implementation details, see~\cite{czarnogorski2018monte} and the
comprehensive survey~\cite{browne2012survey}.

MCTS is a popular algorithm. An easy  way to use it in Python is by
installing it from a pip package (\verb|pip install mcts|).

\subsubsection*{Policies}
At the end of the search, after the predetermined  iterations have been
performed, or when time is up, MCTS returns the value and the action
with the highest visit count. An alternative would be to return
  the action with the highest win rate. However, the visit count takes
  into account the win rate (through UCT) and the number of
  simulations on which it is based. A high 
  win rate may be based on a low number of simulations, and can thus
  be high variance. High visit counts will be low variance. Due to
  selection rule, high
  visit count implies high win-rate with high confidence, while high
  win rate may be low confidence~\cite{browne2012survey}.  The action
  of this initial state $s_0$ constitutes the deterministic policy $\pi(s_0)$.

\subsubsection*{UCT Selection}\label{sec:uct}
The adaptive exploration/exploitation behavior of MCTS is governed by the
selection rule, for which often UCT is chosen. UCT is an adaptive
exploration/exploitation rule that 
achieves high performance in many different domains.

UCT was introduced in 2006 by  Kocsis and
Szepesv\'ari~\cite{kocsis2006bandit}.\index{Kocsis, Levente}\index{Szepesv\'ari, Csaba}
The paper provides a theoretical guarantee of
eventual convergence to the minimax
value. The selection rule was named UCT,  for upper confidence
bounds for multi-armed bandits applied to trees. (Bandit theory was
also mentioned in  Sect.~\ref{sec:multiarmed}).
\index{UCT}\label{uct}\index{multi-armed bandit}

The selection rule determines the way in which the
current values of the children influence which part of the tree will
be explored more.   The UCT formula
is 
\begin{equation}
 \mbox{UCT}(a)=\frac{w_{a}}{n_{a}}+C_{p}\sqrt{\frac{\ln n}{n_{a}}}  \label{eq:uct}
 \end{equation}
where $w_{a}$ is the number
of wins in child $a$, $n_{a}$ is the number of times child $a$ has
been visited, $n$ is the number of times the parent node has been
visited, and $C_{p}\geq0$ is a constant, the tunable
exploration/exploitation hyperparameter. The first term in the UCT
equation, the win rate $\frac{w_{a}}{n_{a}}$, is the exploitation term. A child with a high
win rate receives  through this term an exploitation bonus. The second
term $\sqrt{\frac{\ln n}{n_{a}}}$ is for
exploration. A child that has been visited infrequently has a higher
exploration term. The level of exploration can be 
adjusted by the $C_{p}$ constant.  A low $C_p$ does little exploration;
a high $C_p$ has more exploration.
The selection rule then is to
select the child with the highest UCT sum (the familiar $\argmax$
function of  value-based methods).

The UCT formula balances {\em win rate\/} $\frac{w_{a}}{n_{a}}$ and
{\em ``newness''\/} $\sqrt{\frac{\ln n}{n_{a}}}$ in the selection of
nodes to expand.\footnote{The square-root term is a measure of the
  variance (uncertainty) of the action value. The use of the natural
  logarithm ensures that, since increases get smaller over
  time, old actions are selected less frequently. However, since logarithm values
  are unbounded, eventually all actions will be 
  selected~\cite{sutton2018introduction}.}  Alternative selection rules have been
proposed,  such as 
Auer's UCB1~\cite{auer2002using,auer2002finite,auer2010ucb} and
P-UCT~\cite{rosin2011multi,matsuzaki2018empirical}.

\subsubsection*{P-UCT}\index{P-UCT}
We should note that the
MCTS that is used in the  AlphaGo Zero program is a little different.
MCTS is used
inside the training loop, as an integral 
part of the self-generation of training examples, to enhance the
quality of the examples for every self-play iteration, using both
value and policy inputs to guide the search.

Also, in the AlphaGo Zero program MCTS backups rely fully on the value function
approximator;
no playout is performed anymore.  The MC part in the
name of MCTS, which stands for the Monte Carlo playouts, really has
become a misnomer for this neural network-guided tree searcher.

Furthermore, selection in self-play MCTS is  different.  UCT-based
node selection now also uses the input from the policy head of the trained function
approximators, in addition to the win rate and newness. 
What remains is that through the UCT mechanism MCTS can focus its search effort greedily
on the part with the highest win rate, while at the same time balancing
exploration of parts of the tree that are underexplored.

The  formula that is  used to incorporate input from the policy head of the
deep network is a variant of
P-UCT~\cite{silver2017mastering,moerland2018a0c,rosin2011multi,matsuzaki2018empirical}\index{P-UCT}
(for \emph{predictor}-UCT). Let
us compare
P-UCT with UCT. 
 The P-UCT formula adds the policy head $\pi(a|s)$ to Eq.~\ref{eq:uct}
$$
\mbox{P-UCT}(a)=\frac{w_{a}}{n_{a}}+C_{p}\pi(a|s)\frac{\sqrt{n}}{1+n_{a}} .
 $$
P-UCT adds the  $\pi(a|s)$ term specifying the probability of the
action $a$ to the exploration part of the UCT formula.\footnote{Note
  further the small differences under the square root (no logarithm, and the 1
 in the denominator) also change the UCT function profile somewhat,
 ensuring correct behavior at unvisited
 actions~\cite{moerland2018a0c}.}

\subsubsection*{Exploration/Exploitation}\label{sec:explexpl}

The search process of MCTS is guided by the statistics values in the tree. MCTS
discovers during the search where the promising parts of the tree
are. The tree expansion of MCTS is  inherently variable-depth and
variable-width (in contrast to minimax-based algorithms such as
alpha-beta, which are inherently fixed-depth and fixed-width).
In
Fig.~\ref{fig:mctstree} 
we see a snapshot of the search tree of an actual MCTS optimization.  Some
parts of the tree are searched more deeply than
others~\cite{vermaseren2000new}. 

\begin{figure}[t]
    \centering{\includegraphics[width=\textwidth]{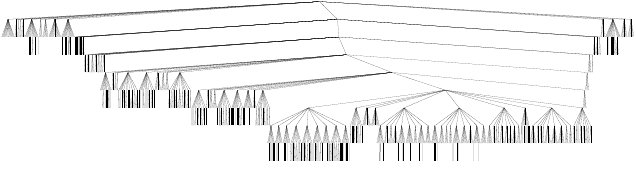}
     \caption{Adaptive MCTS tree \cite{kuipers2013improving}}\label{fig:mctstree}}
\end{figure}

An important element of MCTS is the exploration/exploitation trade-off, that
can be tuned with the \gls{cp} hyperparameter. The effectiveness of MCTS in different
applications 
depends  on the value of this hyperparameter. Typical initial choices for
Go programs are $C_p = 1$ or $C_p=0.1$~\cite{browne2012survey}, although in AlphaGo we
 see highly explorative choices such as $C_p=5$. In general,
 experience has learned that when
compute power is low,  $C_p$ should be low, and when more compute
power is available, more exploration (higher $C_p$) is
advisable~\cite{browne2012survey,kuipers2013improving}.

\subsubsection*{Applications}
MCTS was introduced in
2006~\cite{coulom2007monte,coulom2009monte,coulom2006efficient} in the
context of computer Go programs, following work by Chang et
al.~\cite{chang2005adaptive}, Auer et al.~\cite{auer2002finite}, and
Cazenave and Helmstetter~\cite{cazenave2005combining}. The introduction of MCTS improved performance of Go programs
considerably, from medium amateur to strong amateur. Where the
heuristics-based GNU Go program played around 10 kyu, Monte Carlo programs
progressed to 2-3 dan  in a few years' time.

Eventually, on the small  $9\times 9$ board, Go
programs achieved very strong play. 
On the large $19\times 19$ board, performance did not improve much beyond 
the 4-5 dan level, despite
much effort by researchers. It was thought that perhaps the large action space
of the $19\times 19$ board was too hard for MCTS. Many enhancements
were considered, for the playout phase, and for the selection. As the
AlphaGo results show, a crucial enhancement was the introduction of deep
function approximation. 


After its introduction, MCTS quickly proved successful in other applications, both two agent and
single agent: for video games~\cite{chaslot2008monte}, for single
player applications~\cite{browne2012survey}, and for many other games.
%
%
Beyond games, MCTS revolutionized the world of heuristic
search~\cite{browne2012survey}. Previously, in order to achieve
best-first search, one had 
to find a domain specific heuristic to guide the search in a smart
way. With MCTS this is no longer necessary. Now a general method
exists that  finds the promising parts of the search without a
domain-specific heuristic, just by using statistics of the search
itself.

There is a deeper relation between   UCT and reinforcement
learning. Grill et al.~\cite{grill2020monte} showed how the second
term of P-UCT acts as a regularizer on model-free policy
optimization~\cite{abdolmaleki2018maximum}. In particular, Jacob et
al.~\cite{jacob2021modeling} showed how MCTS can be used to achieve
human-like play in chess, Go, and Diplomacy, by regularizing
reinforcement learning with supervised learning on human games.

\subsubsection*{MCTS in AlphaGo Zero}

For policy improvement, AlphaGo Zero
uses a version of on-policy MCTS that does not use random playouts
anymore.
To increase exploration, Dirichlet noise is added to the P-UCT value at the root
node, to ensure that all moves may be tried. The $C_p$ value of MCTS in AlphaGo is 5, heavily favoring
exploration. In AlphaGo Zero
the value depends on the stage in the learning; it grows during
self-play. In each
self-play iteration 25,000 games are played. For each move, MCTS
performs 1600 simulations. In total over a three-day course of training 4.9
million games were played, after which AlphaGo Zero outperformed the
previous version, AlphaGo~\cite{silver2017mastering}.

\subsubsection*{Conclusion}
We have taken a look into the planning part of AlphaGo Zero's self-play
architecture.
MCTS consists of a move selection and  a
statistics backup phase, that corresponds to the behavior (trial) and
learning (error) from reinforcement learning.
MCTS is an important algorithm in reinforcement learning, and we have
taken a detailed look at the algorithm.

Move-level self-play is our first self-play procedure; it is a
  procedure that
plays itself to generate its counter moves.
The move-level planning  is only one part of the self-play
picture. Just as important is the learning part. Let us have a look at
how AlphaGo Zero achieves its function approximation. For this, we
move to the second level of self-play: the example level.

\subsection{Example-Level Self-Play} 

Move-level self-play creates an environment for us that can play our
counter-moves. Now we need a mechanism 
to learn from these actions. AlphaGo Zero follows the actor
critic principle to approximate both value and
policy functions. It approximates these functions  using a single deep
residual neural network with a value-head and a policy-head (Sect.~\ref{sec:residual}). The policy
and the value approximations are incorporated in MCTS in the
selection and backup step.

In order to learn, reinforcement learning needs training examples. The training examples
are generated at the self-play move level. Whenever a move is played, the
$\langle s,a\rangle$ state-action pair is recorded, and whenever a full game has been
played to the end, the outcome $z$ is known, and the outcome is added to
all pairs of game moves, to create $\langle s,a,z\rangle$ triples. The triples are
stored in the replay buffer, and sampled randomly  to train the value/policy net.
The
actual implementation in AlphaGo Zero contains many more elements to
improve the learning stability.

The player is designed to become
stronger than the opponent, and this occurs at the example-level. Here
it uses MCTS to improve the current policy, improving it with  moves
that are winning against the opponent's moves.

Example-level self-play is our second self-play procedure, 
the examples  are generated in the self-play games and are used to
train the network that is used to play the moves by the two players.

\subsubsection{Policy and Value Network}\index{residual network}
The first AlphaGo program uses three separate neural networks: for
the rollout policy, for the value, and for the selection
policy~\cite{silver2016mastering}. AlphaGo Zero uses a single network,
that is tightly integrated in MCTS. Let us have a closer look at this single
network.

The network is trained on the example triples  $\langle s, a,
z\rangle$ from the replay buffer. These triples contain search results
of the board states of the game, and the two loss function targets:
$a$ for the  actions
that MCTS predicts for each board states, and $z$ for the outcome of the
game (win or loss) when it came to an end. The action $a$ is the
policy loss, and the outcome $z$ is the value loss. All triples for a
game consist of the same outcome $z$, and the different  actions that were
played at each state.

AlphaGo Zero uses a dual-headed residual network (a convolutional network with extra
skip-links between layers, to improve
regularization, see
Sect.~\ref{sec:residual}~\cite{he2016deep,cazenave2018residual}).
Policy and value loss contribute 
equally to the loss function~\cite{wang2019alternative}. The network is trained by stochastic
gradient descent. L2
regularization is used to reduce overfitting. The network has 19
hidden layers, and an input layer 
and two output layers,  for policy and value.
The size of the mini-batch for updates is 2048. This batch is
distributed over 64 GPU workers, each with 32 data entries. The
mini-batch is sampled uniformly over the last 500,000 self-play
games (replay buffer). The learning rate started at 0.01 and went down to 0.0001
during self-play. More details of the AlphaGo Zero network are
described in~\cite{silver2017mastering}.

Please note the size of the replay buffer, and the long training time. Go is a complex game, with sparse rewards. Only at the end of a long game the win or loss is known, and attributing this sparse reward to the many individual moves of a game is difficult, requiring many games to even out errors.  

MCTS is an on-policy algorithm that makes use of guidance in two
places: in the downward action selection 
and in the upward value backup. 
In AlphaGo Zero the function approximator  returns both elements:
a policy for the action selection and a value for the
backup~\cite{silver2017mastering}.

For tournament-level self-play to succeed,
the training process must (1) cover enough of the
state space,  must (2) be stable, and must (3) converge. Training
targets must be sufficiently 
challenging to learn, and sufficiently diverse. The purpose of MCTS is
to act as a policy improver in the actor critic setting, to generate
learning targets of sufficient quality and diversity for the agent to
learn.

Let us have a closer look at these aspects, to get a broader
perspective on why it was so difficult get self-play to work in Go.

\subsubsection{Stability  and Exploration}

Self-play has a long history in artificial intelligence, going back to
TD-Gammon, 30 years ago.  
Let  us look at the challenges in achieving a
strong level of play.

Since in AlphaGo Zero all learning is  by reinforcement, the training
process must now be even more stable than in AlphaGo, which also used
supervised learning from grandmaster games. The slightest problem
in overfitting 
or correlation between states can throw off the  coverage,
correlation, and convergence. AlphaGo Zero
uses  various forms of exploration to achieve stable reinforcement learning. 
%
Let us summarize how stability is achieved.
\begin{itemize}
\item \emph{Coverage} of the sparse state space is improved by playing a  large 
  number of diverse games. The quality of the states is further improved by
  MCTS look-ahead. MCTS
searches  for good training
samples, improving the quality and diversity of the  covered states.  
The exploration
part of MCTS should make sure that enough new and unexplored parts of
the state space are covered. Dirichlet noise is added  at the root
node and the $C_p$ parameter in the P-UCT formula,
that controls the level of exploration,
has been set quite high, around 5 (see also Sect.~\ref{sec:explexpl}). 

\item \emph{Correlation} between subsequent states is reduced
through the use of an experience replay buffer, as
in DQN and Rainbow algorithms. The replay buffer breaks correlation
between subsequent training examples. Furthermore, the MCTS search
also breaks correlation, by searching deep in the tree to 
find better states.

\item \emph{Convergence} of the
training is improved by using on-policy MCTS, and by taking small training
steps. Since the  learning rate is small, training target stability is
higher, and the  risk of
divergence is reduced. A disadvantage is that  convergence is quite
slow and requires many training 
games.
\end{itemize}
By using  these measures together, stable generalization and convergence
are achieved.
Although self-play is conceptually simple, achieving stable and
high-quality self-play in a game as complex and sparse as Go, required slow training
with a  large number of games,  and quite some hyperparameter tuning.
There are many hyperparameters whose values must be set
correctly, for the full
list, see~\cite{silver2017mastering}. Although the values are
published~\cite{silver2017mastering}, the reasoning behind the values
is not always clear.
Reproducing the AlphaGo
Zero results  is not easy, and  much time is  spent in tuning and
experimenting  to reproduce the AphaGo Zero
results~\cite{prasaf2018lessons,straus2018alphazero,nair2017learning,tian2017elf,PhoenixGo2018}.

\subsubsection*{Two Views}
At this point it is useful to step back and reflect on the
self-play architecture.

There are two different views. The one view,
planning-centric, which we have followed so far, is of a searcher
that is helped by a learned evaluation function (which trains on
 examples from games played against itself). In addition, there is
move-level self-play (opponent's moves are generated with an inverted replica of itself) and there is tournament-level self-play (by the value learner).

The alternative view, learning-centric, is that a policy is learned by
generating game examples from self-play. In order for these examples
to be of high quality, the policy-learning is helped by a policy
improver, a planning function that performs lookahead to create better
learning targets (and the planning is performed by making moves
by a copy of the player). In addition,  there is tournament-level self-play (by the
policy learner) and there is move-level self-play (by the
policy improver).

The difference in viewpoint is who comes first: the planning viewpoint favors
the searcher, and the learner is there to help the planner; the
reinforcement learning viewpoint favors the policy learner, and the
planner is there to help improve the policy.
Both viewpoints are equally valid, and
both viewpoints are equally valuable.  Knowing of
the other viewpoint deepens our understanding of how these 
complex self-play algorithms work.

This concludes our discussion of the second type of self-play, the
example-level, and we move on to the third type: tournament-level self-play.

\subsection{Tournament-Level Self-Play} 


At the top level, a tournament of self-play games is played between
the two (identical) players. The player is designed to increase in
strength, learning from the examples at the second level, so that the player can achieve a
higher level of play. In tabula rasa self-play, the players  start from scratch. By
becoming progressively stronger, they also become stronger opponents
for each other, and their mutual level of play can increase. A virtuous cycle
of ever increasing intelligence will emerge.

For this ideal of artificial intelligence to become reality, many
stars have to line up. After TD-Gammon, many researchers have tried to
achieve this goal in other games, but were unsuccessful.

Tournament-level self-play is only possible when move-level self-play and
example-level self-play work. For move-level self-play to work, both
players need to have access to the transition function, which must be
completely accurate. For example-level self-play to work, the player
architecture must be such that it is able to learn a stable policy of a high
quality (MCTS and the network have to mutually improve eachother).

Tournament-level self-play is the third self-play procedure, where a
tournament is created with  games starting from easy learning tasks,
changing to harder tasks, increasing all the way to world champion
level training. This third  procedure  allows
reinforcement learning to transcend the level of play
(``intelligence'') of previous teachers.

\subsubsection*{Curriculum Learning}\label{sec:days}
\index{curriculum learning}\index{intrinsic motivation}\index{developmental robotics}
As mentioned before, the AlphaGo effort consists of three programs: AlphaGo, AlphaGo Zero, and AlphaZero.
The first AlphaGo program used supervised learning based on grandmaster games, followed
by reinforcement learning on self-play games. The second program,
AlphaGo Zero, used reinforcement learning only, in a self-play
architecture that starts from zero knowledge.  The first program
trained many weeks, yet the second program needed only a few days to
become stronger than the 
first~\cite{silver2017mastering,silver2016mastering}.

Why did the self-play approach of AlphaGo Zero learn faster than the
original AlphaGo that could benefit from all the knowledge of Grandmaster games?
%
Why is
self-play faster than a combination of supervised and reinforcement
learning? The reason is a phenomenon called
curriculum learning: self-play is faster because it creates a sequence
of  learning tasks that are ordered from easy to hard. Training such
an ordered sequence of small tasks is quicker than one large unordered
task. 

Curriculum learning starts the training process  with  easy concepts
before the hard
concepts are learned; this is, of course, the way in which  humans learn.  Before
we learn to run, we learn to walk; before we 
learn about multiplication, we learn about 
addition. In curriculum
learning the examples are ordered in batches from easy to
hard. Learning such an ordered sequence of batches goes better since
understanding the easy concepts helps understanding of the harder concepts;
learning everything all at once typically takes longer and may result
in lower accuracy.\footnote{Such a sequence of related learning tasks
  corresponds to a meta-learning problem. In meta-learning the aim is
  to learn a new task fast,  by using the knowledge learned from
  previous, related, tasks; see Chap.~\ref{chap:meta}.}
\label{sec:selfplay}

\subsubsection{Self-Play Curriculum Learning}
  \label{sec:easy}
\label{sec:curriculum}
\label{sec:curr}\index{curriculum learning}\label{sec:curr:gen}

In ordinary deep reinforcement learning  the network tries
to solve a fixed problem in one large step, using environment samples
that are not
sorted from easy to 
hard.
With  examples that are not sorted, the program  has to
achieve the optimization step from  
no knowledge to human-level play in one big, unsorted, 
leap, by optimizing many times over challenging  samples  where
the error function is large.
Overcoming such a
large training step (from beginner to advanced) 
costs much training time.

In contrast, in AlphaGo Zero,  the
network is trained in many small steps, starting against a very weak
opponent, just as a human child learns to play the game by playing against a
teacher that plays simple moves. 
As our level of play increases, so does the difficulty of the moves that our teacher proposes to us. 
Subsequently,
harder problems are generated and trained for, refining  the network that
has already been pretrained with the easier examples.

Self-play naturally generates a curriculum with examples from easy to
hard. The learning network is always in lock step with the training
target---errors are low throughout the training. 
As a consequence,  training times go down and the  playing
strength goes up.

\begin{figure}[t]
    \centering{\includegraphics[width=\textwidth]{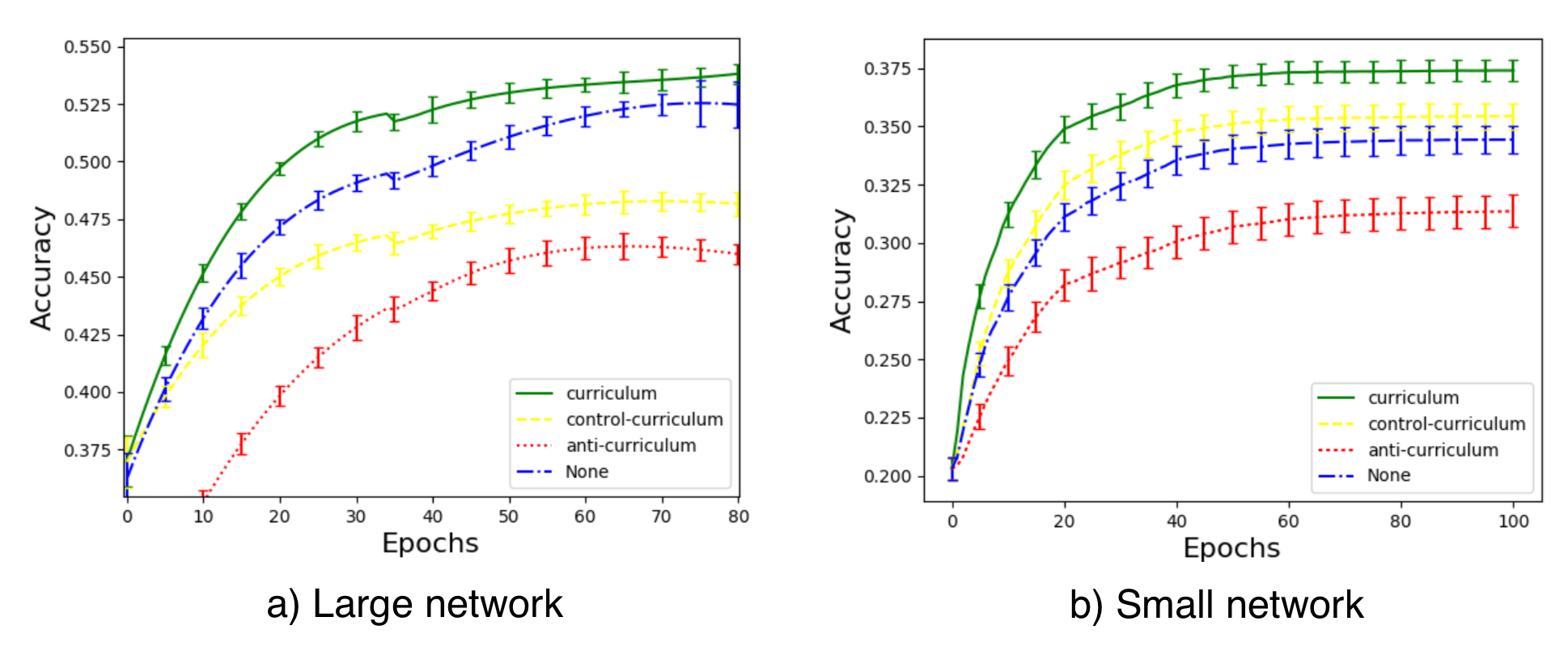}
     \caption{Effectiveness of a Sorted Curriculum~\cite{weinshall2018curriculum}}\label{fig:cur}}
\end{figure}

\subsubsection{Supervised and Reinforcement Curriculum Learning}
Curriculum learning has been studied before in psychology and education science.
Selfridge et al.~\cite{selfridge1985training} 
first connected curriculum learning to machine learning, where they
trained the proverbial Cartpole controller. First they trained the
controller on long and light poles, while  gradually
moving towards shorter and heavier poles.
Schmidhuber~\cite{schmidhuber1991curious} proposed a related concept, to improve
exploration for  world models by  artificial curiosity. Curriculum
learning was subsequently applied to match the order of training examples to
the growth in model capacity in 
different supervised learning
settings~\cite{elman1993learning,krueger2009flexible,bengio2009curriculum}. 
Another related development is in developmental robotics, where
curriculum learning can help to self-organize open-ended developmental
trajectories~\cite{oudeyer2007intrinsic}, related to
intrinsic motivation (see Sect.~\ref{sec:intrinsic}). The  AMIGo
approach uses
curriculum learning  to generate subgoals in hierarchical
reinforcement learning~\cite{campero2020learning}.

To  order the training examples from easy to hard, we need a measure
to quantify the difficulty of the task. One idea is  to use the
minimal loss with respect to some of the upper layers of a high-quality
pretrained model~\cite{weinshall2018curriculum}.  In a supervised learning experiment,
Weinshall et al.\ compared the effectiveness of curriculum learning on a set of
test images (5 images of mammals from CIFAR100). Figure~\ref{fig:cur}
shows the accuracy of a curriculum ordering (green), no curriculum
(blue), randomly ordered groups (yellow) and the labels sorted in
reverse order (red). Both networks are regular networks with multiple
convolutional layers followed by a fully connected layer. The large
network has 1,208,101 parameters, the small network has 4,557
parameters.  We can clearly see the effectiveness of  ordered learning~\cite{weinshall2018curriculum}.

\subsubsection*{Procedural Content Generation}\index{procedural content generation}

Finding a good way to
order the sequence 
of examples is often difficult. A possible method  to generate a
sequence of tasks that are related is by using procedural content generation
(PCG)~\cite{shaker2016procedural,brockhausen2021procedural}.
Procedural content generation uses randomized algorithms to
generate images and other content for computer games; the difficulty
of the examples can often be 
controlled.
It is frequently used to automatically generate different levels in
games, so that they do not have to be all created manually by the game
designers and
programmers~\cite{smith2015analog,togelius2013procedural}.\footnote{See also
generative adversarial networks and deep dreaming, for a
connectionist approach to content generation, Sect.~\ref{sec:onepixel}.}

The Procgen benchmark suite has been built upon procedurally generated
games~\cite{cobbe2020leveraging}. Another popular benchmark is the General video game AI
competition (GVGAI)~\cite{liebana2019general}.
Curriculum learning reduces overfitting to single tasks. Justesen et
al.~\cite{justesen2018illuminating} have used  GVGAI to show
that a policy easily overfits to specific games, and that training
over a  curriculum  improves its generalization to 
levels that were designed by humans. MiniGrid is a procedurally
generated world that can be used for hierarchical reinforcement
learning~\cite{chevalierminimalistic,raileanu2020ride}.

\subsubsection*{Active Learning}\index{active learning}
Curriculum learning is related also related to active
learning.

Active learning is a type of machine learning that is in-between
supervised and reinforcement learning. Active learning is relevant
when labels are in principle available (as in supervised learning)
but at a cost.

Active learning performs a kind of iterative
supervised learning, in which the agent
can choose to query which labels to reveal during the learning process. Active learning is related
to reinforcement learning and to curriculum learning, and is for
example of
interest for studies into recommender systems, where acquiring more
information may come at a cost~\cite{settles2009active,rubens2015active,das2016incorporating}.

\subsubsection*{Single-Agent Curriculum Learning}
Curriculum learning has been studied for many years. A problem
 is that it is difficult to find an ordering of tasks from easy to
hard in most learning situations. In two-player self-play the
ordering comes natural, and the successes
have inspired recent work on  single-agent curriculum learning. 
For example, Laterre et al.\ introduce the Ranked Reward method for solving bin packing
problems~\cite{laterre2018ranked} and Wang et al.\ presented a method for
Morpion Solitaire~\cite{wang2020tackling}. Feng et al.\ 
use an AlphaZero based approach to solve hard Sokoban
instances~\cite{feng2020solving}. Their model is an 8 block standard
residual network, with MCTS as planner. 
They create a curriculum by constructing simpler subproblems from hard
instances, using the fact that Sokoban problems have a natural
hierarchical structure. 
This approach was able to solve
harder Sokoban instances than had been solved before. Florensa et
al.~\cite{florensa2018automatic} study the generation of goals for
curriculum learning using a generator network (GAN).\index{GAN}

\subsubsection*{Conclusion}

Although curriculum learning has been studied in artificial
intelligence and in psychology for some time, 
it has not been a popular method, since  it is difficult to find well-sorted training
curricula~\cite{mitchell1980need,mitchell2006discipline,wang2015basic}. 
Due to the self-play results, curriculum learning is
now attracting more interest, see~\cite{narvekar2020curriculum,weng2020curriculum}.
%
Work is reported in single-agent
problems~\cite{narvekar2020curriculum,feng2020solving,doan2019line,laterre2018ranked},
and in multi-agent  games, as we will see in Chap.~\ref{chap:multi}. 

After this detailed look at self-play algorithms, it is time to look
in more detail at the environments and benchmarks for self-play.

\section{Self-Play Environments}\index{tabula rasa}\label{sec:self:bench}
Progress in reinforcement learning is determined to a large extent by
the application domains that provide the learning challenge. The
domains of checkers, backgammon, and especially chess and Go, have provided
highly challenging domains, for which  substantial 
progress was achieved, inspiring many researchers.

The previous section provided an overview of the planning and 
learning algorithms. In this section we will have a closer look at the
environments and systems that are used to benchmark these algorithms.

\begin{table}[t]
  \begin{center}
  \begin{tabular}{lll}
    {\bf Name} & {\bf Approach}  &  {\bf Ref}  \\
    \hline\hline
    TD-Gammon & Tabula rasa self-play, shallow network, small alpha-beta search& \cite{tesauro1995td}\\
    AlphaGo & Supervised, self-play, $3\times$CNN, MCTS &  \cite{silver2016mastering}\\
    AlphaGo Zero & Tabula rasa self-play, dual-head-ResNet, MCTS  & \cite{silver2017mastering}\\
    AlphaZero & Tabula rasa self-play, dual-head-ResNet, MCTS on Go, chess, shogi  & \cite{silver2018general}\\
    \hline
  \end{tabular}
  \caption{Self-Play Approaches}\label{tab:az}
\end{center}
\end{table}
Table~\ref{tab:az} lists the AlphaGo and self-play approaches that we
discuss in this chapter.
First we will discuss the three AlphaGo programs, and then we will
list open self-play frameworks. We will start with the first program,
AlphaGo.

\subsection{How to Design a World Class Go Program?}
Figure~\ref{fig:ag0perf} shows the playing strength of
traditional programs (right panel, in red) and different versions of the
AlphaGo programs, in blue. We see how much stronger the 2015, 2016, and 2017
versions of AlphaGo are than the earlier heuristic minimax program GnuGo,
and two MCTS-only programs Pachi and Crazy Stone.

\begin{figure}[t]
    \centering{\includegraphics[width=\textwidth]{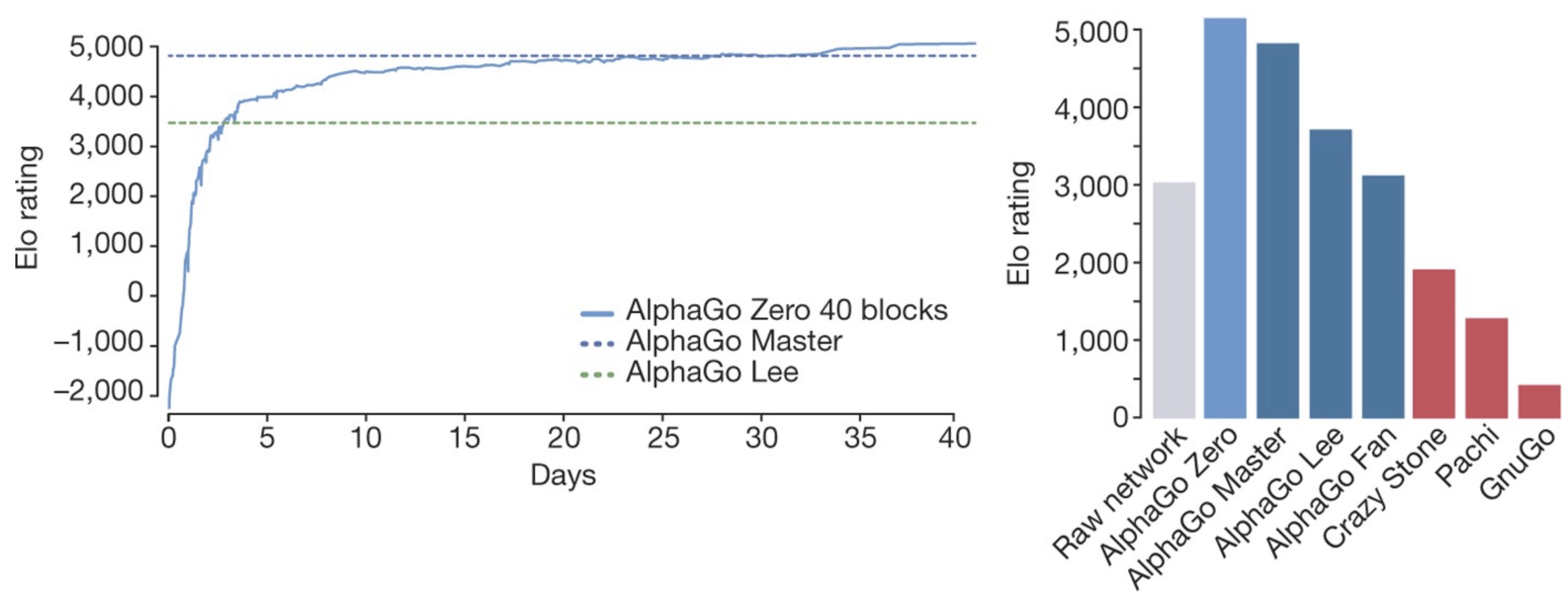}
     \caption{Performance of AlphaGo Zero~\cite{silver2017mastering}}\label{fig:ag0perf}}
\end{figure}

How did the AlphaGo
authors  design such a strong Go program? Before AlphaGo, the
strongest 
programs used the Monte Carlo Tree Search planning
algorithm, without neural networks. For some time, neural networks
were considered to be too
slow for use as value function in
MCTS, and random playouts were used, often improved with small
pattern-based heuristics~\cite{gelly2006modification,gelly2008achieving,chaslot2008monte,coulom2009monte,enzenberger2010fuego}. Around
2015 a few researchers tried to
improve performance of MCTS by using deep learning evaluation
functions~\cite{clark2015training,anthony2017thinking,gelly2012grand}.
These efforts were strenghthened by the strong
results in Atari~\cite{mnih2015human}.

The AlphaGo team also tried to use neural networks. Except for backgammon,  pure self-play
approaches had not been shown to work well, and the AlphaGo team did the sensible thing
to pretrain the network with the games of human grandmasters, using
supervised learning. Next, a large number of self-play games were used to further train
the networks. In total, no less than three neural networks were used: one for the MCTS
playouts, one for the policy function, and one for the value
function~\cite{silver2016mastering}.

\begin{figure}[t]
  \centering{\includegraphics[width=\textwidth]{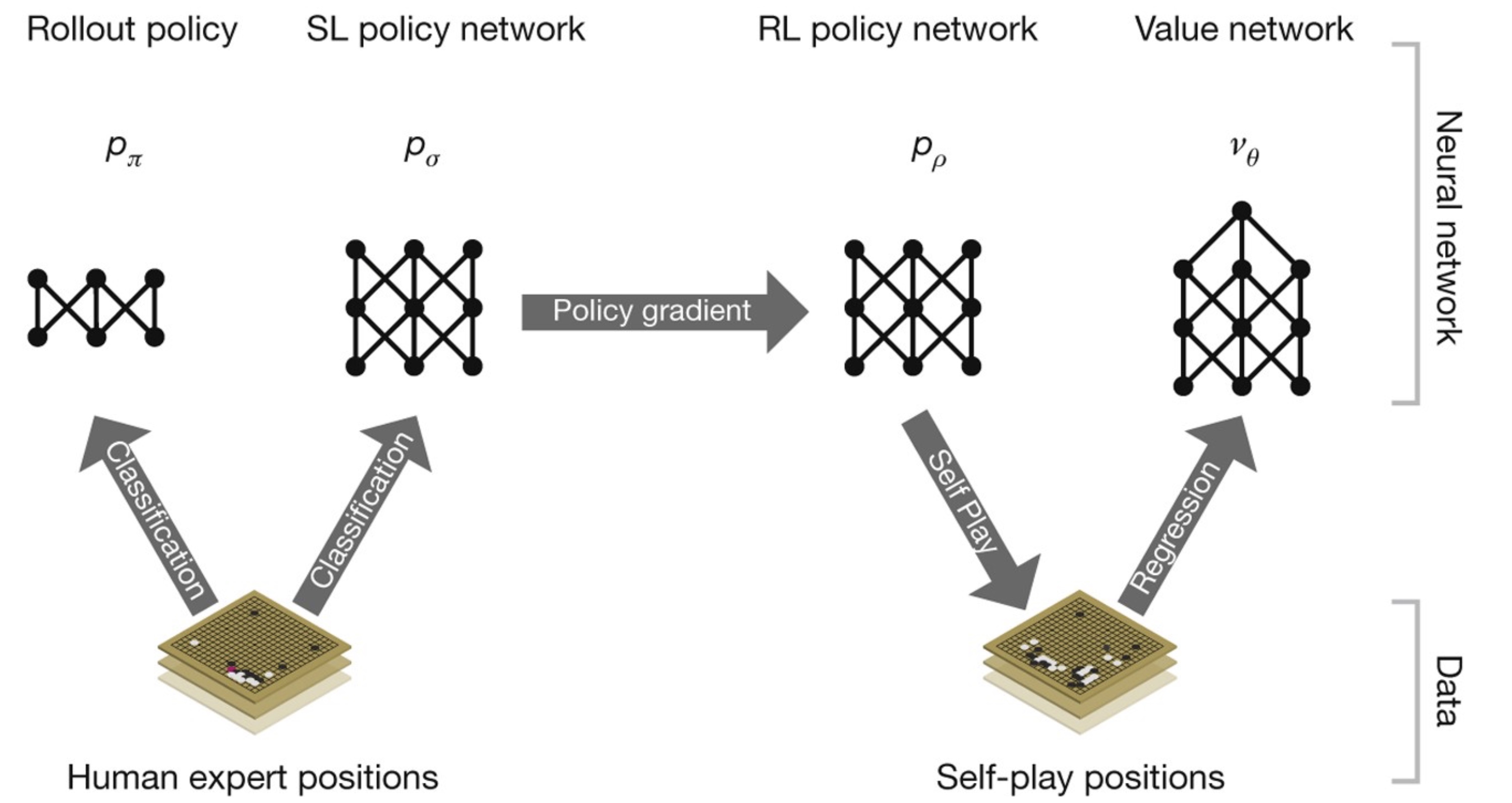}
\caption{All AlphaGo Networks~\cite{silver2016mastering}}\label{fig:alphago-training}}
\end{figure}

Thus, the original AlphaGo program consisted of three neural networks
and used both supervised learning and reinforcement learning. The
diagram in Fig.~\ref{fig:alphago-training} illustrates the AlphaGo
architecture.  Although this design 
made sense at the time given the state of the art of the field, and although it
did convincingly beat the three strongest human Go players, managing
and tuning such a complicated piece of software is quite difficult. The
authors of AlphaGo tried to improve performance further by
simplifying their design. Could TD-gammon's elegant  self-play-only design
be replicated in Go after all?

Indeed, a year later a reinforcement
learning-only version was ready that learned to play Go from zero
knowledge, tabula rasa: no grandmaster games, only
self-play, and just a single neural
network~\cite{silver2017mastering}. Surprisingly, this simpler version 
played  stronger and learned faster. The program was called AlphaGo 
Zero, since it learned from zero-knowledge;
not a single grandmaster game was learned from, nor was there any heuristic
domain knowledge hand coded into the program. 

This new result, tabula rasa learning in Go with a
pure self-play design, inspired much further research in self-play
reinforcement learning.


\subsection{AlphaGo Zero  Performance}

In their paper Silver et al.~\cite{silver2017mastering} describe that learning
progressed smoothly throughout the training. AlphaGo Zero
outperformed the original AlphaGo after just 36 hours. The training time for the
version of  AlphaGo that played Lee Sedol was  several
months. Furthermore, AlphaGo Zero used a single machine with 4 tensor
processing units, whereas AlphaGo Lee was distributed over many
machines and used 48 TPUs.\footnote{\gls{TPU} stands for tensor processing unit, a
  low-precision design specifically developed for fast neural network
  processing.} Figure~\ref{fig:ag0perf} shows the
performance of AlphaGo Zero. Also shown is the performance of the raw
network, without MCTS search. The importance of MCTS is large, around
2000 Elo points.\footnote{The
basis of the Elo rating is pairwise
comparison~\cite{elo1978rating}. Elo is
often used to compare playing strength in board games.}\index{Elo, Arpad}

AlphaGo Zero's reinforcement learning is truly learning Go knowledge from
scratch, and, as the development team discovered, it did so in a way
similar to how humans are discovering the intricacies of the game. In
their paper~\cite{silver2017mastering} they published a picture of how this
knowledge acquisition progressed (Fig.~\ref{fig:joseki}).

Joseki are standard corner
openings that all Go players become familiar with as they  learn
to play the game. There are
beginner's and advanced joseki. Over the course of its learning,
AlphaGo Zero did  learn joseki, and it learned them from
beginner to advanced. It is interesting to see how it did so, as it
reveals the 
progression of AlphaGo Zero's Go intelligence. Figure~\ref{fig:joseki}
shows sequences from games played by the program.  Not to anthropomorphize too
much,\footnote{Treat as if human} but you can see the little program getting smarter.

\begin{figure}[t]
    \centering{\includegraphics[width=\textwidth]{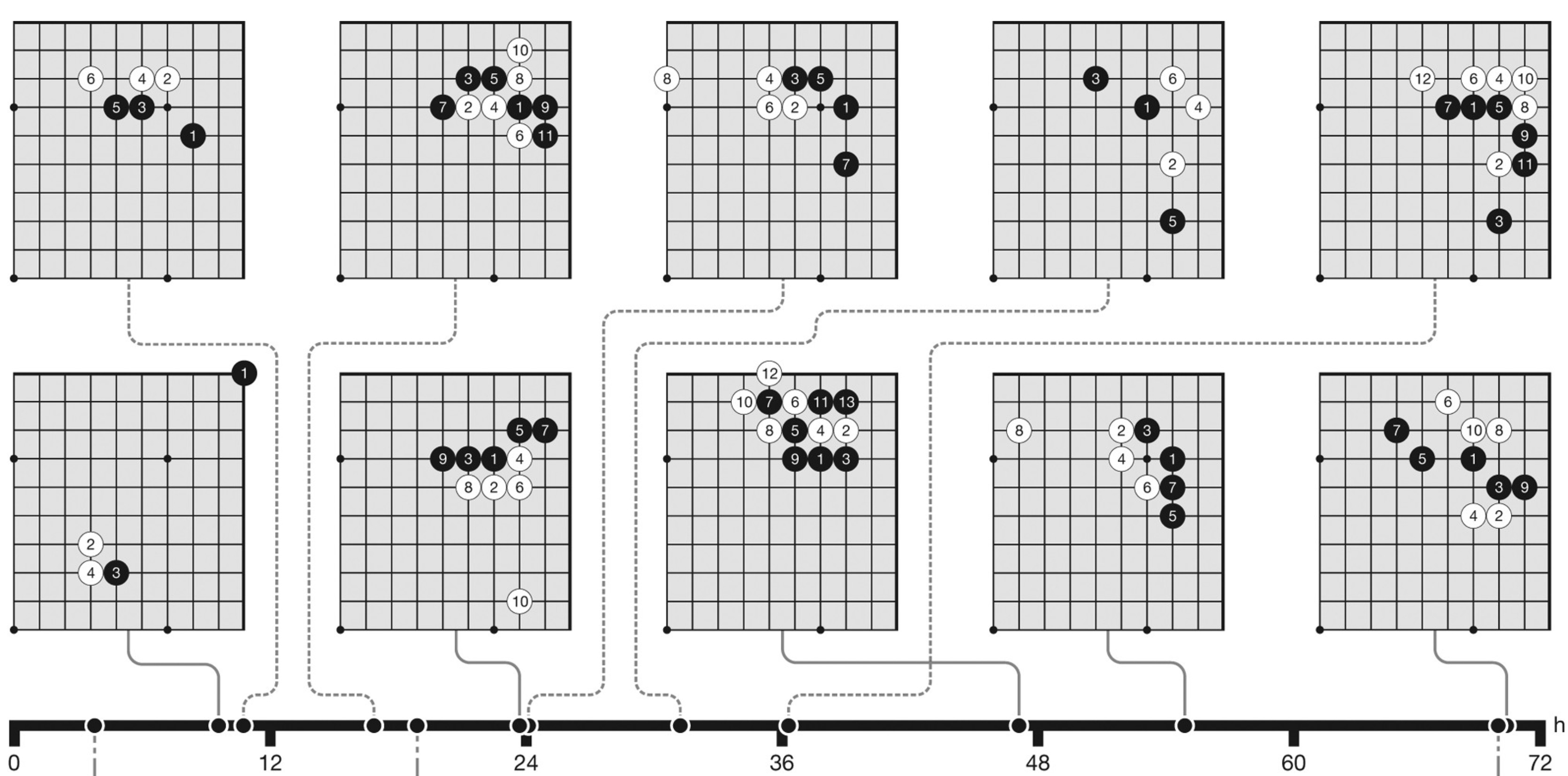}
     \caption[AlphaGo Zero Learning Joseki]{AlphaGo Zero is Learning Joseki in a Curriculum from
       Easy to Hard~\cite{silver2017mastering}}\label{fig:joseki}}
\end{figure}

The top row shows five joseki that AlphaGo Zero discovered. The first
joseki is one of the standard beginner's openings in  Go theory. As
we move to the right, more difficult joseki are learned, with 
stones being played in looser configurations.
The bottom row shows five joseki favored at different stages of the
self-play training. It starts with a preference for a weak corner
move. After 10 more hours of training, a better 3-3 corner sequence
is favored. More training reveals more, and better, variations.

AlphaGo Zero discovered a remarkable level of Go knowledge during
its self-play training process. This knowledge included not only fundamental
elements of human Go knowledge, but also nonstandard strategies
beyond the scope of traditional Go knowledge.

For a human Go player, it is remarkable to see this kind of progression in
computer play, reminding them of the time when they
discovered these joseki themselves. With such evidence of the computer's
learning, it is hard not to anthropomorphize AlphaGo
Zero. 

\subsection{AlphaZero}
The AlphaGo story does not end with AlphaGo Zero.
A year after AlphaGo Zero, a version was created with different input
and output layers 
that  learned to play chess and shogi (also known as Japanese
chess, Fig.~\ref{fig:shogi}). AlphaZero  uses the same MCTS and deep reinforcement learning architecture
as for learning to play Go (the only differences are the input and 
output layers)~\cite{silver2018general}. This new 
program, AlphaZero, beat the
strongest chess and shogi programs, Stockfish and Elmo. Both these
programs followed  a conventional heuristic
minimax design, optimized by hand and machine learning, and improved
with many heuristics for decades. AlphaZero used zero 
knowledge, zero grandmaster games, and zero hand-crafted heuristics,
yet it played stronger. The AlphaZero architecture allows not only
very strong play, but is also a general architecture, suitable for
three different games.\footnote{Although an AlphaZero version that has learned
  to play Go, cannot play chess. It has to re-learn chess from scratch, with
  different input and output layers.}

 \begin{figure}[t]
 \begin{center}
 \includegraphics[width=5cm]{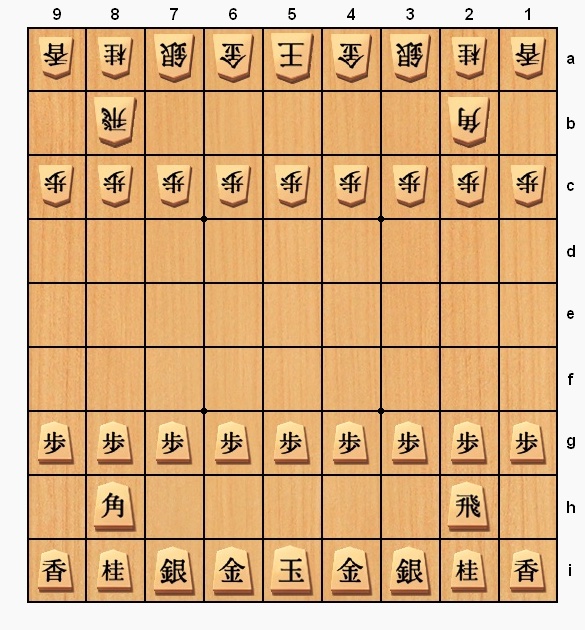}
 \caption{A Shogi Board}\label{fig:shogi}
 \end{center}
 \end{figure}

The Elo rating of AlphaZero in chess, shogi, and Go is shown in
Fig.~\ref{fig:azelo},~\cite{silver2017mastering,silver2016mastering}.
AlphaZero is
stronger than the other programs. In chess the difference is the
smallest. In this field the program  has benefited from  a large
community of researchers that have worked intensely on improving
performance of the heuristic alpha-beta approach. For shogi the difference  is larger.

\begin{figure}[t]
    \centering{\includegraphics[width=\textwidth]{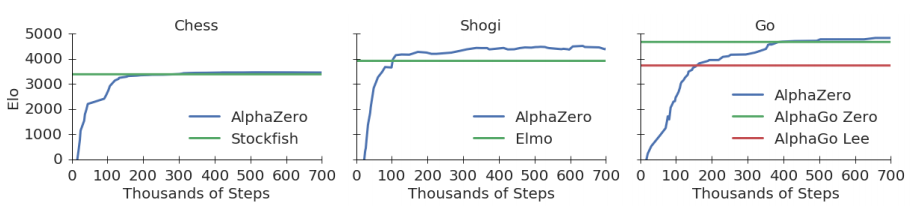}
     \caption{Elo rating of AlphaZero in Chess, Shogi, and Go~\cite{silver2018general}}\label{fig:azelo}}
\end{figure}

\subsubsection*{General Game Architecture}
\index{Shogi}\index{Elmo}\index{Stockfish}
AlphaZero can play three different games with the same architecture. 
The three games are quite different. Go is a
static game of strategy. Stones do not move and are
rarely captured. Stones, once played, are of strategic
importance. In chess the pieces move. Chess is a  dynamic game
where tactics are important. Chess also features sudden death, a
check-mate can occur in the middle of the game  by capturing the
king.
Shogi is even more dynamic since captured pieces can be returned to the
game, creating even more complex game dynamics.

It is testament to the generality of AlphaZero's architecture, that
games that differ so much in tactics and 
strategy can be learned so successfully.
Conventional programs must be purposely developed for each game, with
different search hyperparameters and different heuristics. Yet the 
MCTS/ResNet self-play architecture is able to learn all three from
scratch.

\subsection{Open Self-Play  Frameworks}
Tabula rasa learning for the game of Go is a remarkable achievement
that inspired many researchers. 
The code of AlphaGo Zero  and AlphaZero, however, is not public.  Fortunately, the
scientific
publications~\cite{silver2017mastering,silver2018general} provide many
details, allowing other researchers to reproduce similar results.

Table~\ref{tab:self-env} summarizes some of the self-learning
environments, which we will briefly discuss.

\begin{table}[t]
  \begin{center}\small
  \begin{tabular}{p{1.4cm}lp{6cm}c}
    {\bf Name} & {\bf Type} & {\bf URL} & {\bf Ref.}\\
    \hline\hline
    AlphaZero General& AlphaZero in Python &
                                              \url{https://github.com/suragnair/alpha-zero-general}
                                        &  \cite{nair2017learning} \\ 
    ELF & Game framework & \url{https://github.com/pytorch/ELF} & \cite{tian2017elf}  \\ 
    Leela & AlphaZero for Chess, Go &
                                     \url{https://github.com/LeelaChessZero/lczero} 
                                        &  \cite{pascutto2017leela} \\ 
    PhoenixGo &  AlphaZero-based Go prog. & \url{https://github.com/Tencent/PhoenixGo}
                                        & \cite{PhoenixGo2018} \\ 
    PolyGames &  Env. for Zero learning & \url{https://github.com/facebookincubator/Polygames} & \cite{cazenave2020polygames} \\ \hline
  \end{tabular}
  \caption{Self-learning environments}\label{tab:self-env}
\end{center}
\end{table}

\begin{itemize}
  \item \emph{A0G: AlphaZero General}\label{sec:a0g}
Thakoor et al.~\cite{nair2017learning} created\index{A0G}\index{AlphaZero General}
a self-play system  called AlphaZero
General (A0G).\footnote{\url{https://github.com/suragnair/alpha-zero-general}} It
is implemented in Python for TensorFlow, Keras, and PyTorch, and
suitably scaled down for smaller computational resources. It has
implementations for $6\times 6$ Othello, tic tac toe, gobang, and
connect4, all small games of significantly less complexity than
Go. Its main network architecture is a four layer CNN followed by 
two fully connected layers. The code is  easy to understand
in an afternoon of study, and is well suited for educational purposes.
The  project write-up provides some \href{https://github.com/suragnair/alpha-zero-general/raw/master/pretrained\_models/writeup.pdf}{documentation}~\cite{nair2017learning}.

\item \emph{Facebook ELF}
ELF stands for Extensible Lightweight Framework. It is a framework for game
research  in C++ and Python~\cite{tian2017elf}. Originally developed
for real-time strategy games by Facebook, it includes the Arcade Learning Environment and
the Darkforest\footnote{\url{https://github.com/facebookresearch/darkforestGo}}
Go program~\cite{tian2015better}. ELF can be found
on GitHub.\footnote{\url{https://github.com/pytorch/ELF}}
ELF  also contains the self-play program OpenGo~\cite{ELFOpenGo2018}, a reimplementation of  
AlphaGo Zero (in C++).\index{ELF}\index{OpenGo}

\item \emph{Leela}
Another reimplementation of AlphaZero is
 Leela. Both a chess and a Go version of Leela
 exist. The \href{https://github.com/LeelaChessZero/lczero}{chess}
 version is based on chess engine
 Sjeng. The Go\footnote{\url{https://github.com/gcp/leela-zero}}
 version is based on Go engine Leela.  Leela does not come with
 trained weights of the network. Part of Leela  is a community effort
 to compute these weights.\index{Leela}

\item \emph{PhoenixGo}\index{PhoenixGo}
PhoenixGo is a strong self-play Go program by Tencent~\cite{PhoenixGo2018}. It
is based on the AlphaGo Zero architecture.\footnote{\url{https://github.com/Tencent/PhoenixGo}} A
trained network is available as well.

\item \emph{Polygames}\index{Polygames}
PolyGames~\cite{cazenave2020polygames} is an environment for Zero-based learning (MCTS with deep
reinforcement learning) inspired by AlphaGo Zero. Relevant learning
methods are implemented, and bots for hex, Othello, and Havannah have
been implemented. PolyGames can be found on
GitHub.\footnote{\url{https://github.com/facebookincubator/Polygames}} A library of games is provided, as well as
a checkpoint zoo of neural network models. 
\end{itemize}

\subsection{\em Hands On: Hex in Polygames Example}\index{Hex}\index{Polygames}
Let us get some hands-on experience with MCTS-based self-play. We will 
implement the game of Hex with the PolyGames
suite. Hex is a  simple and fun board game invented independently
by Piet Hein and John Nash in the 1940s. Its simplicity makes it easy to learn and
play, and also a popular choice  for mathematical analysis. The game
is played on a hexagonal board, player 
A wins if its moves connect the right to the left side, and player B
wins if top connects to bottom (see Fig.~\ref{fig:hex}; image by Wikimedia). A simple page
with resources is
here;\footnote{\url{https://www.maths.ed.ac.uk/~csangwin/hex/index.html}}
extensive strategy and background books have been written about hex~\cite{hayward2019hex,browne2000hex}.
We use hex because it is simpler than Go, to get you up to speed
quickly with
self-play learning; we will also use PolyGames. 

\begin{figure}[t]
    \centering{\includegraphics[width=5cm]{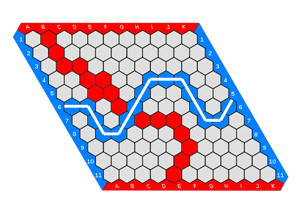}
     \caption{A Hex Win for Blue}\label{fig:hex}}
\end{figure}

Click on the
link\footnote{\url{https://github.com/facebookincubator/Polygames}}
and start by reading the  introduction to PolyGames on
GitHub. Download the paper~\cite{cazenave2020polygames} and
familiarize yourself with the concepts behind Polygames. Clone the repository and build it by following the
instructions. Polygames uses PyTorch,\footnote{\url{https://pytorch.org}} so install that too (follow the
instructions on the Polygames page).

Polygames is interfaced
via the pypolygames  Python package. The games, such as Hex, can be
found in \verb|src/games| and are  coded in
C++ for speed. The command
\begin{tcolorbox}
  \verb|pypolygames train|
\end{tcolorbox}
is used to train a
game and a model.

The command
\begin{tcolorbox}
  \verb|pypolygames eval|
\end{tcolorbox}  
is used to use
a previously trained model.

The command
\begin{tcolorbox}
  \verb|pypolygames human|
\end{tcolorbox}
allows a human to play against a trained model.

Type
\begin{tcolorbox}
  \verb|python -m pypolygames {train,eval,traineval,human} --help|
\end{tcolorbox}  
for help with each of the commands \verb|train|, \verb|eval|,
\verb|traineval|, or \verb|human|.

A command to start training a Hex model with the default options is:
\begin{tcolorbox}
  \verb|python -m pypolygames train --game_name="Hex"|
\end{tcolorbox}
Try loading a
pre-trained model from the zoo.
Experiment with different training options, and try playing against
the model that you just trained.
When everything works, you can also try training different games. Note
that more complex games may take (very) long to train.

\section*{Summary and Further Reading}
\addcontentsline{toc}{section}{\protect\numberline{}Summary and Further Reading}
We will now summarize the chapter, and provide pointers for further reading.
\subsection*{Summary}
For two-agent zero-sum games, when the transition function is given by the
rules of the game, a special kind of 
reinforcement learning becomes possible. Since the agent
can perfectly simulate the moves of the opponent, accurate   planning
far into the future
becomes possible, allowing strong policies to be learned. Typically, the second agent becomes the
environment. Previously environments were
static, but they  will now evolve as the agent is learning, creating a
virtuous cycle of increasing (artificial) intelligence in agent (and
environment). The promise of this self-play setup is to achieve high
levels of intelligence in a specific field.
The challenges to overcome instability, however, are  large, since
this kind of self-play combines different kinds of unstable learning
methods. Both TD-Gammon and AlphaGo Zero have overcome these
challenges, and we have described their approach in quite some detail.


Self-play is  a combination
of  planning, learning, and a self-play loop.
%
The self-play loop in AlphaGo Zero uses MCTS to generate high-quality examples,
which are used to train the neural net. This new neural net is then used in a
further self-play iteration to generate more difficult games, and
refine the network further (and again, and again, and again).
Alpha(Go) Zero thus learns starting at zero knowledge, \emph{tabula rasa}. 

Self-play makes use of many reinforcement learning techniques. In
order to ensure stable learning, exploration is
important. MCTS is used for deep planning. The exploration
parameter in MCTS is set high, and convergent training is achieved by
a low learning rate $\alpha$. Because of these hyperparameter
settings, and because of the sparse rewards in Go, many games
have to be played. The computational demands of stable self-play are large.

AlphaGo Zero uses function approximation of two functions: value and
policy. Policy is used to help guide action selection in the P-UCT
selection operation in MCTS, and value is used instead of random
playouts to provide the value function at the leaves of the MCTS tree.
MCTS has been changed significantly to work in the self-play setting. Gone are the
random playouts that gave MCTS the name Monte Carlo, and much of the
performance is due to a high-quality policy and value approximation
residual network. 

Originally, the AlphaGo program (not AlphaGo Zero) used grandmaster
games in supervised learning in addition to using  reinforcement
learning; it started from the knowledge of grandmaster games. Next
came AlphaGo Zero, which does not use grandmaster games or any other
domain specific knowledge. All learning is based on reinforcement
learning,  playing itself to build up the knowledge of the game
from zero knowledge. A third experiment has been published, called
AlphaZero (without the ``Go''). In this paper the same network
architecure and MCTS design (and the same learning hyperparameters)
were used to learn three  games: chess, shogi, and Go. This
presented the AlphaZero architecture as a general learning
architecture, stronger than the best alpha-beta-based chess and shogi programs.

Interestingly, the all-reinforcement learning AlphaGo Zero architecture
was not only stronger than the supervised/reinforcement hybrid
AlphaGo, but also faster: it learned world champion level play in
days, not weeks. Self-play learns quickly because of curriculum
learning. It is more efficient to learn a large problem in many small
steps, starting with easy problems, ending with hard ones, then in one
large step. Curriculum learning works  both for humans and for artificial
neural networks.

\subsection*{Further Reading}

One of the main interests of artificial intelligence is the study of
how intelligence emerges out of simple, basic, interactions. 
In self-learning systems this is happening~\cite{leibo2019autocurricula}.  

The work on AlphaGo is a landmark achievement in artificial intelligence. 
The primary sources of information for AlphaGo are the three AlphaGo/AlphaZero
papers by Silver et
al.~\cite{silver2016mastering,silver2017mastering,silver2018general}. The
systems are complex, and so are the papers and their supplemental
methods sections. Many blogs have been written about AlphaGo that are  more
accessible.
A movie has been made about AlphaGo.\footnote{\url{https://www.alphagomovie.com}}
There are also  explanations on YouTube.\footnote{\url{https://www.youtube.com/watch?v=MgowR4pq3e8}}

A large literature on  minimax and minimax enhancements for
games exists, an overview is in~\cite{plaat2020learning}.
A book devoted to building your own state-of-the-art self-learning Go bot is \emph{Deep Learning and the Game of Go} by Pumperla
and Ferguson~\cite{pumperla2019}, which came out before PolyGames~\cite{cazenave2020polygames}.

MCTS has been a landmark algorithm by itself in artificial
intelligence~\cite{coulom2006efficient,browne2012survey,kocsis2006bandit}.
In the contexts of MCTS, many researchers worked on combining MCTS with
learned patterns, especially to improve the random rollouts of
MCTS. Other developments, such as asymmetrical  and continuous MCTS, are~\cite{moerland2018monte,moerland2018a0c},
or parallelizations such as~\cite{mirsoleimani2015scaling}.

Supervised learning on grandmaster games 
was used to improve playouts and also to improve UCT selection. Gelly and Silver published notable works in this
area~\cite{gelly2006modification,silver2007reinforcement,gelly2008achieving}. 
Graf et al.~\cite{graf2015adaptive} describe experiments with adaptive
playouts in MCTS with deep learning.
Convolutional neural nets were also used in Go by Clark and
Storkey~\cite{clark2014teaching,clark2015training}, who had used a CNN for supervised learning
from a database of human professional games, showing that it
outperformed GNU Go and scored wins against Fuego, a strong open
source Go program~\cite{enzenberger2010fuego} based on MCTS without
deep learning.

Tesauro's success inspired many others to try temporal difference learning.
Wiering et al. and Van der
Ree~\cite{wiering2010self,van2013reinforcement} report on self-play
and TD learning in Othello and Backgammon.
The program Knightcap~\cite{baxter1999knightcap,baxter2000learning}\index{Knightcap} and
Beal et al.~\cite{beal2000temporal} also use temporal difference learning on evaluation
function features. Arenz~\cite{arenz2012monte} applied MCTS to chess.
Heinz  reported on self-play experiments in
chess~\cite{heinz2000new}.

Since the AlphaGo results many  other applications of machine learning
have been
shown to be successful.
There is interest in theoretical physics~\cite{ruijl2014hepgame,paparo2014quantum}, chemistry~\cite{jumper2021highly}, and pharmacology, specifically for
retrosynthetic molecular design~\cite{segler2018planning} and drug
design~\cite{van2011compound}.  High-profile results have been
achieved by AlphaFold, a program that can predict protein structures~\cite{jumper2021highly,senior2020improved}.

To learn more about 
curriculum learning, see~\cite{weng2020curriculum,bengio2009curriculum,matiisen2017teacher,florensa2018automatic}.
Wang et al.~\cite{wang2019alternative} study the optimization
target of a dual-headed self-play network in AlphaZeroGeneral.
The success of self-play has led to interest in curriculum learning
in single-agent
problems~\cite{feng2020solving,doan2019line,duan2016rl,laterre2018ranked}. The
relation between classical single-agent and two-agent search is
studied by \cite{schaeffer2001unifying}.

\section*{Exercises}
\addcontentsline{toc}{section}{\protect\numberline{}Exercises}
To review your knowledge of self-play, here are some questions and exercises.
We start with  questions to check your understanding of this
chapter. Each question is a closed question where a simple, one
sentence answer is possible.

\subsubsection*{Questions}
\begin{enumerate}
\item What are the differences between AlphaGo, AlphaGo Zero, and
  AlphaZero?
\item What is MCTS?
\item What are the four steps of MCTS?
\item What does UCT do?
\item Give the UCT formula.   How is P-UCT different?   
\item Describe the function of each of the four operations of MCTS.
\item How does UCT achieve trading off exploration and exploitation?
  Which inputs does it use?
\item When $C_p$ is small, does MCTS explore more or exploit more?
\item For small numbers of node expansions, would you prefer more
  exploration or more exploitation?    
\item What is a double-headed network? How is it different from
  regular actor critic?
\item Which three elements make up the self-play loop? (You may draw a
  picture.)
\item What is tabula rasa learning?
\item How can tabula rasa learning be faster than reinforcement
  learning on top of supervised learning of grandmaster games?
\item What is curriculum learning?
\end{enumerate}

\subsubsection*{Implementation: New or
  Make/Undo}\index{make-move}\index{undo-move}
You may have noticed that the minimax and MCTS pseudocode in the figures lacks
implementation details for performing actions, to arrive at successor
states. Such board manipulation and move
making details are  important for creating a
working program.

Game playing programs typically call the search routine with the
current board state, often indicated with parameter \verb|n| for the
new node. This board can be created and allocated anew in each search
node, in a  value-passing style (local variable). Another option is to pass a
reference to the board, and to apply a \verb|makemove| operation on
the board, placing the stone on the board
before the recursive call, and an \verb|undomove| operation removing
the stone from the board when it returns back out of the recursion (global variable).
This reference-passing style may be quicker
if allocating the memory for a new board is an expensive operation. It may also be
more difficult to implement correctly, since the makemove/undomove
protocol must be followed strictly on all relevant places in the code. If capture moves cause many
changes to the board, then these must be remembered for the subsequent
undo. 

For parallel implementations in a shared memory at least all parallel threads must have
their own copy of a value-passing style board. (On a distributed memory
cluster the separate machines will have their own copy of the board by
virtue of the distributed memory.)

\subsubsection*{Exercises}
For the programming exercises we use PolyGames. See the previous
section on how to install PolyGames. If training takes a long time,
consider using the GPU support of Polygames and Pytorch.
\begin{enumerate}
  \item \emph{Hex} Install PolyGames and train a Hex player with self-play.  Experiment with different
    board sizes. Keep the training time constant, and draw a graph
    where you contrast playing strength against board size. Do you
    see a clear correlation that the program is stronger on smaller
    boards?
    \item \emph{Visualize} Install the visualization support \verb|torchviz|. Visualize
      the training process using the \verb|draw_model| script.
    \item \emph{Hyperparameters} Use different models, and try different hyperparameters, as
      specified in the PolyGames documentation.
      \item \emph{MCTS} Run an evaluation tournament from the trained Hex model
        against a pure MCTS player. 
 See below for tips on make and undo of moves.
      How many nodes can MCTS  search in a reasonable search
      time? Compare the MCTS Hex player against the
      self-play player. How many games do you need to play to have
      statistically significant results? Is the random seed randomized
      or fixed? Which is stronger: MCTS or trained Hex?
      
    \end{enumerate}
    

\chapter{Multi-Agent Reinforcement Learning}\label{chap:team}\label{chap:multi}

On this planet, in our societies, millions of people live and work
together. Each  individual has their own individual set of
goals and performs their actions accordingly. Some of these goals are
shared.  When we want to  achieve shared goals  we
organize ourselves in teams, groups, companies, organizations and
societies. In many intelligent species---humans, mammals,
birds, insects---impressive displays of  collective
intelligence emerge from such organization~\cite{woolley2010evidence,holldobler2009superorganism,seeley1989honey}. We are learning as individuals, and we are
learning as groups. This  setting is studied in multi-agent learning:
through their  independent actions, agents learn to interact  and
to compete and cooperate with other agents, and form groups.

Most research in reinforcement learning  has focused on single
agent problems.  Progress has been
made in many topics, such as path-finding, robot locomotion, and video
games. In addition,  research 
has been performed in  two-agent problems, such as
competitive two-person board games.
Both single-agent and two-agent  problems are questions  of
optimization. The goal is to find the policy with the highest reward,
 the shortest
path, and the best moves and counter moves. The basic setting is one of reward
optimization in the face of
natural adversity or competitors, modeled by the environment.

As we move closer toward modeling real-world problems, we encounter
another category of sequential decision making  
problems, and that is the category of multi-agent decision
problems. Multi-agent decision making is a difficult problem,

Agents that share the same goal  might collaborate,  and
finding a policy with the highest reward for oneself or for the group may 
include achieving win/win solutions with other agents.  
Coalition forming and collusion are an integral part of the field of multi-agent
reinforcement learning.

In real-world decision making both 
competition and cooperation are important. If we want our agents to behave
realistically in settings with multiple agents,  then they should
understand cooperation in order to perform well.

From a computational perspective, studying the emergence of group behavior and
collective intelligence
is challenging: the environment for the agents consists of many
other agents; goals may move, many interactions have to be modeled in order to be
understood, and the world to be optimized against is constantly changing. 
%
A lack of compute power has held
back experimental research in 
multi-agent reinforcement learning for some time.  The recent growth in
compute power and  advances in deep reinforcement learning methods
are making progress increasingly possible.

Multi-agent decision making problems that have been studied
experimentally include the games of 
bridge, poker, Diplomacy,
StarCraft,  Hide and Seek, and Capture the Flag. 
%
To a large extent, the algorithms for
multi-agent reinforcement learning are still being developed. 
This 
chapter is 
full of challenges and development.

We will start by reviewing the theoretical framework and defining
multi-agent problems. We will 
look deeper at cooperation and 
competition, and introduce stochastic games,  extensive-form games,
and the Nash equilibrium and Pareto optimality. 
We will discuss population-based methods
and curriculum learning in multi-player teams.  
Next, we will discuss environments on which multi-agent reinforcement
learning methods can be tested,  such as poker and StarCraft. In these
games, often team structure plays an important role. The next chapter
discusses hierarhical methods, which can be used to model team
structures.

This chapter is concluded, as usual, with  exercises, a summary, and pointers
to further reading.

\section*{Core Concepts}
\begin{itemize}
\item Competition
\item Cooperation
\item Team learning
\end{itemize}

\section*{Core Problem}
\begin{itemize}
\item Find efficient methods for large competitive and cooperative
  multi-agent problems
\end{itemize}

\section*{Core Algorithms}
\begin{itemize}
\item Counterfactual regret minimization (Sect.~\ref{sec:cfr})
\item Population-based algorithms (Sect.~\ref{sec:evo-alg})
\end{itemize}

\section*{Self-Driving Car}

To illustrate the key components of multi-agent decision making, let
us assume a self driving car is on its way to the supermarket and is approaching an intersection.
What action should it take? What is the best outcome for the
car? For other agents? For society?

The goal of the car is to exit the intersection safely
and reach the destination.  Possible decisions are going straight or
turning left or right into another lane. All these actions  consist of
sub-decisions. At each time step, the  car can move by steering,
accelerating and braking.   The car must be
able to detect objects, such as traffic lights, lane markings, and
other cars. Furthermore, we aim to find a
 policy that can control the car to make a sequence of
manoeuvres to achieve the goal. In a decision-making setting such as
this, two additional challenges arise.

The first challenge is that we should be able to anticipate the actions
of other agents. During the decision-making process, at each time
step, the 
robot car should consider not only the immediate value of its own current
action but also adapt to consequences of  actions by  other
agents. (For example,
it would not be good to choose a certain direction that appears
safe early on, stick to it, and not adapt the policy as new information comes
in, such as a car heading our way.)

The second challenge is that these  other
agents, in turn, may  anticipate the actions of our agent in choosing
their actions. Our agent needs to take their anticipatory behavior into account
in its own policy---recursion at the policy level.
Human
drivers, for example, often predict likely movements of other cars
and then take action in response (such as giving way,  or accelerating
while merging into another lane).\footnote{Human drivers have a \emph{theory of mind}
of other drivers. Theory of mind, and the related concept of mirror
neurons~\cite{gallese1998mirror}, are a psychological theory of
empathy and understanding, that allows a limited amount of prediction
of  future behavior. Theory of mind studies how individuals
simulate in their minds the actions of others, including their simulation
of our actions (and of our simulations, etc.)~\cite{baron1985does,hernandez2017survey}.}

Multi-agent reinforcement learning 
addresses the sequential decision making problem of  multiple
autonomous agents that operate in a common stochastic environment,
each of which aims to maximise their own long-term reward by interacting
with the environment and with other agents. Multi-agent reinforcement
learning combines the fields of multi-agent
systems~\cite{shoham2008multiagent,wooldridge2009introduction} and  reinforcement
learning.

\section{Multi-Agent Problems}

We have seen  impressive results of single-agent reinforcement
learning in recent years. In addition, 
important results have been achieved in games such as Go and poker, and in
autonomous driving. These application domains involve the
participation of more than one agent.
The study of interactions between more than one
agent has a long history
in fields such as economics and social choice.


Multi-agent problems introduce many new kinds of possibilities, such as
cooperation and simultaneous actions. These phenomena invalidate some of the
assumptions of the theory behind single-agent reinforcement
learning, and  new theoretical concepts have been developed.

To begin our study of multi-agent problems, we will start with game theory.

\subsubsection*{Game Theory}\index{game theory}
Game theory is the  study of strategic
interaction among rational decision-making agents. Game theory originally
addressed only the theory of two-person zero-sum games, and was later extended to  cooperative
games of multiple rational players, such as  
simultaneous action, non-zero sum, and imperfect-information games.
A classic work in game theory is John von Neumann  and Oskar
Morgenstern's \emph{ Theory of Games and Economic
  Behavior}~\cite{von1944theory,von2007theory}, published first in 1944. 
This book has laid the foundation for the mathematical study of 
economic behavior and social choice. Game theory
has been instrumental in developing the theory of common goods and
the role of government in a society of independent self-interested rational
agents. In mathematics and artificial intelligence, game theory has given a
formal basis for the computation of strategies in multi-agent systems.



The Markov decision process that we have used to formalize
single-agent reinforcement learning assumes perfect information.
Most multi-agent reinforcement learning problems, however, are imperfect
information problems, where some of the information is private, or
moves are simultaneous.
We will have to extend our MDP model appropriately to be able to
model imperfect information.

\subsubsection*{Stochastic Games and Extensive-Form Games}


One direct generalization of MDP that captures the interaction of
multiple agents is the \emph{Markov game}, also known as the stochastic
game~\cite{shapley1953stochastic}. Described  by
Littman~\cite{littman1994markov}, the 
framework of Markov games has long been used to express 
multi-agent reinforcement learning algorithms~\cite{lanctot2017unified,tuyls2018generalised}.

This multi-agent version of an MDP is defined as follows~\cite{zhang2019multi}. At time $t$,
each agent $i \in N$ executes an action 
$a^i_t$, for the state $s_t$ of the system. The system then transitions to
state $s_{t+1}$, and rewards each agent $i$ with reward $R_{a_t}^i(s_t,s_{t+1})$. The goal of
agent $i$ is to optimize its own long-term reward, by finding the policy
$\pi^i : S \rightarrow \mathbb{E}(A^i)$ as a mapping from the state
space to a distribution over the action space, such that $a^i_t \sim \pi^i(\cdot|s_t)$. Then, the
value-function $V^i : S \rightarrow R$ of agent $i$ becomes a function of the joint
policy $\pi : S \rightarrow \mathbb{E}(A)$ defined as $\pi(a|s) = \Pi_{i\in N} \pi^i(a^i |s)$.

\begin{figure}
\begin{center}
\includegraphics[width=\textwidth]{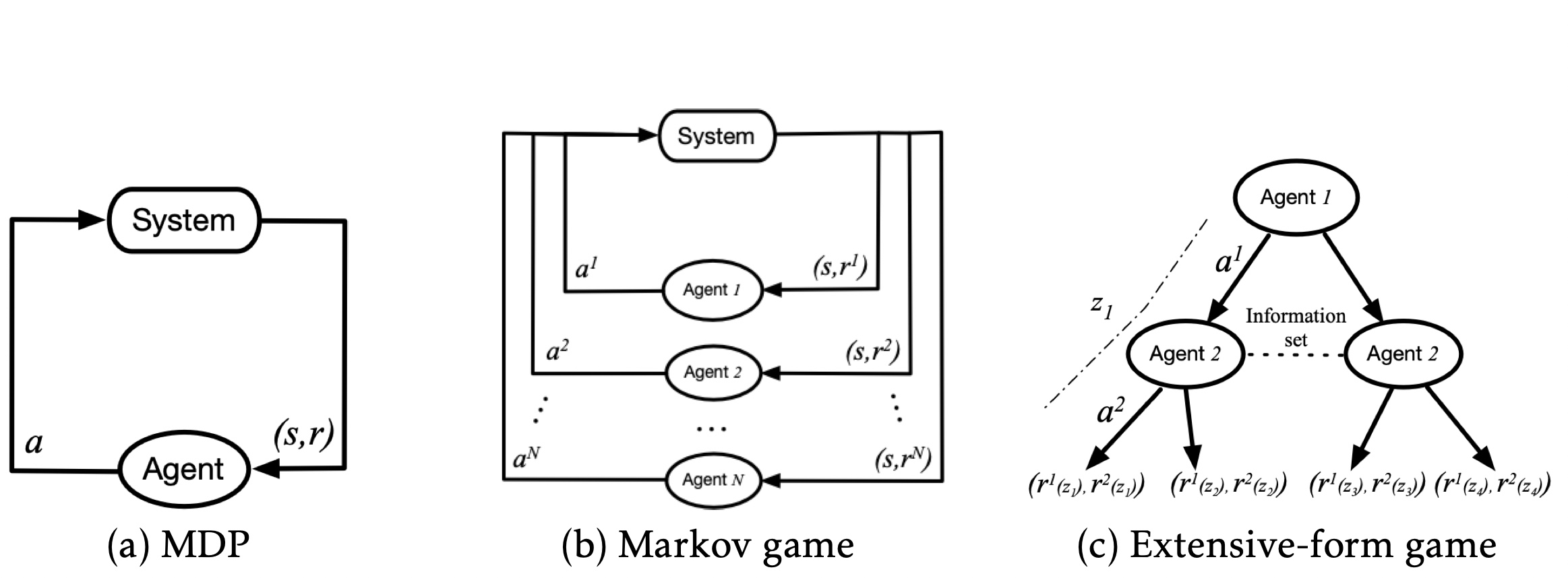}
\caption{Multi-Agent Models~\cite{zhang2019multi}}\label{fig:marl1}
\end{center}
\end{figure}

To visualize imperfect information in multiple agents, new diagrams are
needed. Figure~\ref{fig:marl1} shows three schematic diagrams. First,
in (a),  the familiar 
 agent/environment diagram is shown that we used in earlier
chapters. Next, in (b),  the multi-agent version of this diagram for Markov
games is shown.  Finally, in (c)   an
extensive-form game tree is shown~\cite{zhang2019multi}. Extensive form
game trees are introduced to model imperfect information. Choices by
the agent are shown as solid links, and the hidden private
information of the other agent is  shown dashed, as the information set of possible situations.

In (a), the agent observes the state $s$, performs action $a$, and receives
reward $r$ from the environment. In
(b), in the
Markov game, all agents choose their actions $a^i$ simultaneously and
receiving their individual reward $r^i$. In (c), in  a
two-player extensive-form game, the agents make decisions on
choosing actions $a^i$. They receive their individual reward
$r^i(z)$ at the end of the game, where $z$ is the score of a terminal
node for the branch that resulted from the information set.  The
information set is indicated with a dotted line, to signal 
stochastic behavior, the environment (or other agent) chooses an unknown
outcome amongst the dotted actions. The
extensive-form notation is designed for expressing 
imperfect information games, where all possible (unknown) outcomes are
represented in the information set. In order to calculate the value function,
the agent has to regard all possible different choices in the
information sets; the unknown choices of the opponents create a  large state space.

\subsubsection*{Competitive, Cooperative, and Mixed Strategies}
\index{cooperative}\index{competitive}\index{mixed}\index{general-sum}
Multi-agent reinforcement learning problems fall into three groups:
problems with competitive behavior, with cooperative behavior,  and
with mixed behavior.
%
In the competitive setting, the reward of the agents 
sums up to zero.  A win for one agent is a loss for another. In the
cooperative setting, agents collaborate to optimize a common long-term
reward.  A win for one agent is a win for all agents (example: an embankment being built around an
area prone to flooding; the embankment benefits all agents in the
area).   The mixed setting involves both cooperative and 
competitive agents, with so-called \emph{general-sum} rewards, some
action may lead to win/win, some other to win/loss.

These three behaviors are a useful guide to navigate the landscape of
multi-agent algorithms, and we will do so in this chapter. Let us have
a closer look at each type of 
behavior.

\subsection{Competitive Behavior}\index{Nash
  equilibrium}
One of the hallmarks of the  field of game theory is a result by John
Nash, who 
defined the conditions for a stable (and in a certain sense optimal)
solution among multiple rational non-cooperative
agents. The 
Nash equilibrium is defined as a situation in which no agent has
anything to gain by changing  its own strategy. For two competitive agents
the Nash equilibrium is the minimax strategy.

In single-agent reinforcement learning the goal is to find the policy
that maximizes the cumulative future reward of the agent. In multi-agent
reinforcement learning the goal is to find a combined policy of all
agents that simultaneously achieves that goal:
 the multi-policy that for each agent  maximizes their cumulative
future reward.
If you have a set of competitive (near-)zero exploitability strategies this is
called a Nash equilibrium.

The Nash equilibrium characterizes an equilibrium point $\pi^\star$, from
which none of the agents has an incentive to deviate. In other words,
for any agent $i \in N$, the policy $\pi^{i,\star}$ is the best-response
to $\pi^{-i,\star}$, where $-i$ are all agents except $i$~\cite{hernandez2017survey}.

An agent that follows a Nash strategy is guaranteed
to do no worse than tie,  against any other opponent strategy. For
games of imperfect information or chance, such as many card games, this is an \emph{expected}
outcome. Since the cards are randomly dealt, there is no theoretical
guarantee that a Nash strategy  will win or draw every single hand, although
on average, it cannot do worse than tie against the other agents.

If the opponents also play a Nash  strategy then all will tie. If the
opponents make mistakes, however, then they can lose some hands,
allowing the Nash equilibrium strategy to win. Such a mistake by the
opponents would be a deviation from the Nash strategy, following a
hunch or other non-rational reason, despite
having no theoretical incentive to do so. A Nash equilibrium
plays perfect defence. It does not try to  exploit the opponent
strategy's flaws, and instead just wins when the opponent makes
mistakes~\cite{bowling2015heads,lanctot2009monte}.

The Nash strategy gives the best possible  outcome we can achieve
against our adversaries when they work against us. 
In the sense that a Nash strategy is on average unbeatable, it is considered to
be an optimal strategy, and solving a game is equivalent to computing a
Nash equilibrium.

In a few moments we will introduce a method to
calculate Nash strategies, called counterfactual regret
minimization, but we will first  look at cooperation.

\subsection{Cooperative Behavior}\index{Pareto optimum}
In single-agent reinforcement learning, reward functions return scalar
values: an action results in a win or a loss, or a single numeric
value. In multi-agent reinforcement learning
the reward functions may still return a scalar value, but the
functions may be different for each agent; the overall reward function
is a vector of the individual rewards.
In fully cooperative stochastic games, the individual rewards are the
same, and all agents have the same goal: to maximize the common
return.

When choices in our problem are made in a decentralized manner by a set of individual
decision makers, the problem can be modeled naturally as a decentralized partially
observable Markov decision process (dec-POMDP)~\cite{oliehoek2016concise}.
Large dec-POMDPs are  hard to solve; in general dec-POMDPs  are known to be
NEXP-complete. These problems are not solvable with polynomial-time
algorithms and searching directly for an optimal solution in the
policy space is intractable~\cite{bernstein2002complexity}.

\begin{figure}[t]
    \centering{\includegraphics[width=6cm]{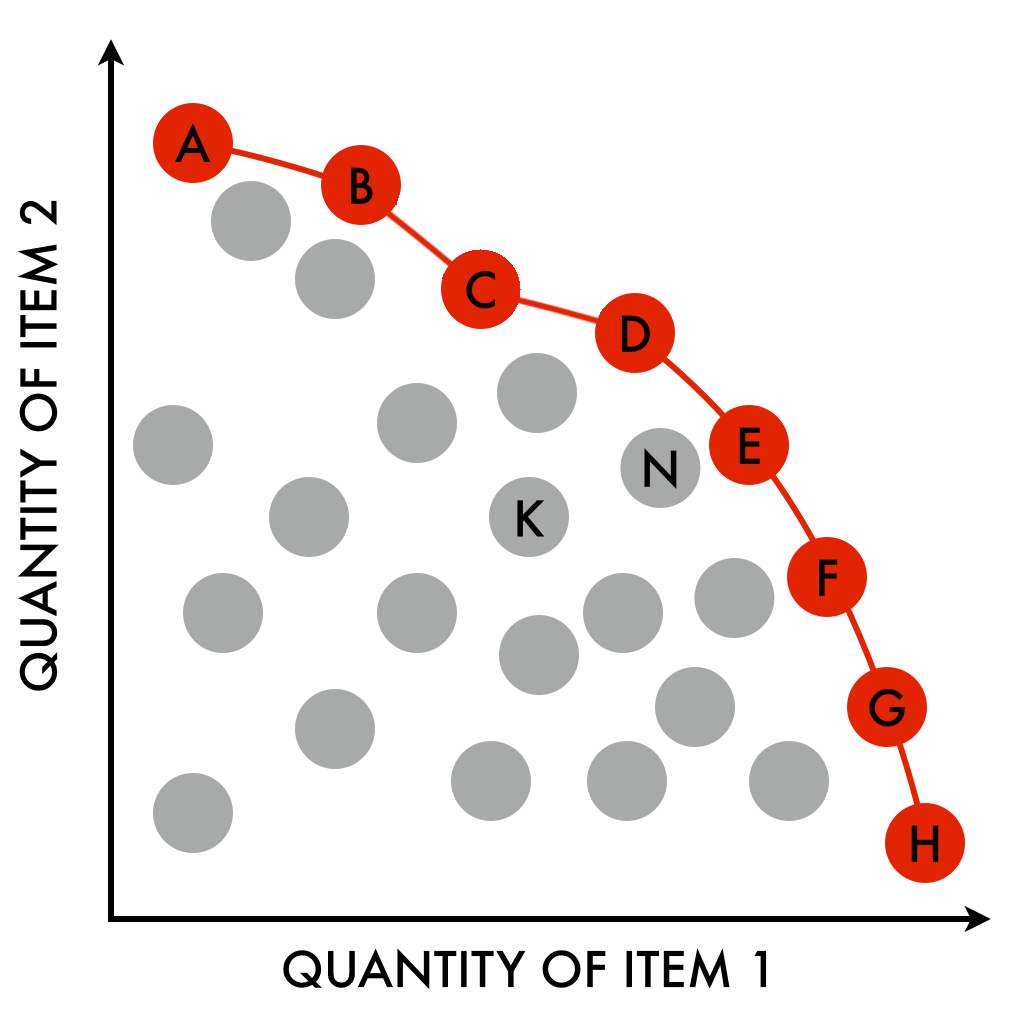}
     \caption{Pareto Frontier of Production-Possibilties}\label{fig:pareto}}
\end{figure}

A central concept in cooperative problems is Pareto efficiency, after
the work of Vilfredo Pareto, who studied economic efficiency and income
distributions.
The Pareto efficient solution is the situation
where no cooperative agent  can be better off without making at
least one other agent  worse off. All non-Pareto efficient
combination are dominated by the Pareto efficient solution. The Pareto
frontier is the set of all Pareto efficient 
combinations, usually drawn as a
curve. Figure~\ref{fig:pareto} shows
a situation where we can choose to produce different quantities of two separate
goods, item 1 and item 2 (image by Wikimedia). The grey items close to the origin are
choices for which better choices exist. The red symbols all represent 
production combinations that are deemed more favorable by
consumers~\cite{sickles2019measurement}. 

In multi-objective reinforcement learning, different agents can have
different preferences. A policy is  Pareto-optimal when
deviating from the policy will make at least one agent worse
off. We will compare Nash and Pareto soon by looking at how they are related
in the prisoner's dilemma (Table~\ref{tab:pd}).

The Pareto optimum is the best possible outcome for us where  we do
not hurt others, and others do not hurt us. It is a cooperative
strategy.  Pareto equilibria assume
communication and trust. In some sense, 
it is the opposite of the non-cooperative Nash strategy. Pareto
calculates the situation for  an  cooperative world, Nash for
an  competitive world.

\subsubsection*{Multi-objective reinforcement learning}
Many real-world problems involve the optimization of multiple,
possibly conflicting objectives.
In a fully cooperative setting, all agents 
share such a common reward function.
When the reward function consists of different values for each agent, the problem is
said to be multi-objective: different agents may have different
preferences or objectives. Note that in principle single-agent problems can also have
multiple ojectives, although in this book all single agent problems
have had single, scalar, rewards. Multi-objective problems become
especially relevant in multi-agent problems, where different
agents may have different reward preferences.

In decentralized decision processes, the agents must communicate their
preferences to eachother. Such heterogeneity  
necessitates the incorporation of communication protocols into
multi-agent reinforcement learning, and the analysis of
communication-efficient algorithms~\cite{hayes2021practical,ropke2021communication}. 

Multi-objective reinforcement
learning~\cite{roijers2013survey,gronauer2021multi,hernandez2019survey,ruadulescu2020multi} is a generalization of standard
reinforcement learning 
where the scalar reward signal is extended to multiple feedback
signals, such as one for each agent. 
A multi-objective reinforcement learning algorithm optimizes multiple
objectives at the same time, and algorithms such as Pareto Q-learning
have been introduced for 
this purpose~\cite{van2014multi,roijers2021following}.


\subsection{Mixed Behavior}\index{prisoner's dilemma}
Starting as a theory for non-cooperative games, game theory has
progressed to include cooperative games, where it analyzes optimal
strategies for groups of individuals, assuming that they can
communciate or  enforce
agreements between them about proper strategies. 
The prisoner's
dilemma is a well-known example of this type of
problem~\cite{kuhn1997prisoner}.\footnote{\url{https://plato.stanford.edu/entries/prisoner-dilemma/}}
The prisoner's dilemma is  thought to be invented by John
Nash, although in a more neutral (non cops and robbers) setting.

The prisoner's dilemma is as follows (see Table~\ref{tab:pd}). Two
burglars, called Row and Column, have been arrested for breaking into
a bank. Row and Column  understand that they
cannot trust the other person, hence they operate in a non-cooperative
setting. The police officer is offering the following options: \emph{If
you both confess, then you will both get a light sentence of 5
years. If you both keep quiet, then I  will only have evidence  to get you a short sentence
of 2 years. However, if one of you confesses, and
the other stays silent, then the confessor will walk free, and the one
keeping quiet goes to prison for 10 years. Please tell me your choice
tomorrow morning.}

This leaves our two criminals with a tough, but clear, choice. If the
other stays silent, and I confess, then I walk free; if I also stay
silent, then we get 2 years in prison, so confessing is better for me
in this case. If the other confesses, and I also
confess, then we both get 5 years; if I  then stay silent, then I
get 10 years in prison, so confessing is again better for me. 
Whatever the other chooses, confessing gives the 
lighter sentence, and since they can not coordinate their action, they will
each independently confess. Both will get 5 years in prison, even
though both would have only gotten 2 years if both would have stayed
silent. The police is happy, since both confess and the case is solved.

If they would
have been able to communicate, or if they would have trusted
eachother, then both would have stayed silent, and both would have gotten off with a 2
year sentence.

\begin{table}[t]
  \begin{center}
  \begin{tabular}{l||c|c|}
    & {\bf Confess}  & {\bf Silent}  \\
    & { Defect}  & {Cooperate}  \\
    \hline\hline
    {\bf Confess}  &  $(-5, -5)$ & $(0,-10)$  \\ 
    {Defect}  &   {\em Nash} &  \\ \hline
    {\bf Silent} &  $(-10, 0)$ & $(-2,-2)$ \\
    {Cooperate} &   &  {\em Pareto}  \\
    \hline
  \end{tabular}
  \caption{Prisoner's Dilemma}\label{tab:pd}
\end{center}
\end{table}

The dilemma  faced by the criminals is that, whatever the other
does, each is better off confessing than remaining silent. However,
both also know that if the other could have been trusted, or if they
could have coordinated their answer, then a
better outcome would have been within reach. 

The dilemma illustrates individual self-interest against the interest
of the group. In the literature on game theory, confessing is also
known as defecting, and staying silent is the cooperative choice.
The confess/confess situation (both defect) is the optimal
non-cooperative strategy, the Nash equilibrium,
because for each agent that stays silent a strategy exists where the
sentence will be made worse (10 years)  when the other agent
confesses. Hence the agent will not stay silent but confess, so as to
limit the loss.

Silent/silent (both cooperate) is Pareto optimal at 2 years for both,
since going to all other cases 
will  make at least one agent worse off.

The Nash strategy is non-cooperative, non-communication,
non-trust, while Pareto is the  cooperative,
communication,  trust,  outcome.

\subsubsection*{Iterated Prisoner's Dilemma}\index{Tit for
  Tat}\index{iterated prisoner's dilemma}\label{sec:tit-for-tat}
The prisoner's dilemma is a one-time game. What would happen if we
play this game repeatedly, being able to identify and remember the choices of our
opponent? Could some kind of communication or trust arise, even if the
setting  is initially non-cooperative?

This question is answered by the iterated version of the prisoner's
dilemma.  Interest in the iterated prisoner's dilemma grew after a
series of publications following a computer
tournament~\cite{axelrod1988further,axelrod1981evolution}. The tournament 
was organized by political scientist Robert Axelrod, around 1980. Game
theorists were invited to send in computer programs to play iterated
prisoner's dilemmas. The programs were organized in a tournament and
played eachother hundreds of times. The goal was to see which strategy
would win, and if cooperation would emerge in this simplest of
settings. A range of programs were entered, some using elaborate
response strategies based on psychology, while others used advanced machine learning
to try to predict the opponent's actions.

Surprisingly, one of the simplest strategies won. It was
submitted by Anatol Rapoport, a mathematical psychologist who
specialized in the  modeling of social interaction.
Rapoport's program played a strategy known as \emph{tit for tat}. It
would start by cooperating (staying silent) in the first round, and in
the next rounds if would play whatever the opponent did in the
previous round. Tit for tat thus rewards cooperation with cooperation,
and punishes defecting with defection---hence the name.

Tit for tat wins if it is paired with a cooperative opponent, and does
not loose too much by stubbornly cooperating with a non-cooperative
opponent. In the long run, it either ends up in the Pareto optimum or
in the Nash equilibrium.
Axelrod attributes the success of tit for tat to a number of
properties. First, it is nice, that is, it is never the first to
defect. In his tournament, the top eight programs played nice
strategies. Second, tit for tat is also retaliatory, it is difficult to
exploit the strategy by the non-cooperative strategies. Third, tit for
tat is forgiving, when the opponent plays nice, it rewards the play
by reverting to being nice, being willing to forget past
non-cooperative behavior. Finally, the rule has the advantage of being
clear and predictable. Others easily learn its behavior, and adapt to
it, by being cooperative, leading to the mutual win/win Pareto optimum.

\subsection{Challenges}

After this brief overview of theory, let us see how well practical
algorithms are able to solve multi-agent problems.  With the
impressive results in single-agent and two-agent reinforcement
learning,  the interest  (and the expectations) in multi-agent problems have 
increased. In the next section, we will have a closer look at
algorithms for multi-agent problems, but let us first look at the
challenges that these algorithms face.

Multi-agent problems are studied in competitive, cooperative, and
mixed settings. 
The main challenges faced by multi-agent reinforcement learning are
threefold: (1) partial observability, (2)  nonstationary
environments, and (3) large state space. All three of these aspects
increase the size of the state space.
We will discuss these challenges in order.

\subsubsection{Partial Observability}
Most multi-agent settings are imperfect information settings, where
agents have some private information that is not revealed to other
agents. The private information can be in the form of hidden cards
whose value is unknowns, such as in poker, blackjack or bridge. In
real time strategy games players often do not see the entire map of
the game, but parts of the world are obscured.
Another reason for imperfect
information can be that the rules of the game allow simultaneous
moves, such as in Diplomacy, where all agents are determining their next
action at the same time, and agents have to act without full knowledge
of the other actions.

All these situations require that all
possible states of the world have to considered, increasing the number of
 possible states greatly in comparison to perfect information
games. Imperfect information is best expressed as an
extensive-form game.  
In fact, given the size of most
multi-agent systems, it quickly becomes unfeasible to communicate and
keep track of all the information of all agents, even if all agents
would make their state and intentions public (which they rarely do).

Imperfect information increases the size of the state space, and computing the
unknown outcomes quickly becomes unfeasible.

\subsubsection{Nonstationary Environments}
Moreover, as all agents are improving their
policies according to their own interests concurrently, the agents are
faced with a nonstationary environment. 
The environment's dynamics are determined by the joint action space of
all agents, in which the best policies of agents depend on the best policies
of the other agents.  This mutual dependence creates an unstable
situation. 

In a single agent setting a single state needs to be
tracked to calculate the next action. In a multi-agent setting, all
states and all agents' policies need to be taken into account, and
mutually so.   Each agent faces a
moving target problem.

\index{non-stationarity}
In multi-agent reinforcement learning the agents
learn concurrently, updating their behavior policies
concurrently and often simultaneously~\cite{hernandez2017survey}. 
Actions taken by one agent affect the reward of other 
agents, and therefore of the next state. This invalidates the Markov
property, which states that
all information that is necessary to determine the next state is
present in the current state and the agent's action. The powerful
arsenal of single-agent
reinforcement theory must be adapted before it can be used.

To handle nonstationarity,   agents must account
for the joint action space of the other agents' actions. The size of
the space increases exponentially with the number of agents. In two-agent settings the agent has to consider
all possible replies of a single opponent to each of its own moves, increasing the state
space greatly.   In multi-agent settings the number of replies increases
even more, and computing
solutions to these  problems quickly becomes quite expensive. A
large number of agents complicates  convergence analysis and increases
the computational demands substantially.

On the other hand, agents may
learn from each other, line up their goals, and
collaborate. Collaboration and group forming reduce the number of independent agents
that must be tracked.

\subsubsection{Large State Space}
In addition to the problems caused by partial observability and
nonstationarity, the size of the state space is significantly
increased in a multi-agent setting, simply because
every additional agent exponentially increases the state
space. 
%
Furthermore, the action space of multi-agent reinforcement learning is
a joint action space whose size 
increases exponentially with the number of agents. The size of the
joint action  space often  causes
scalability issues.

Solving such a large state space to optimality is infeasible for all
but the smallest multi-agent problems, and much work has been done to
create models (abstractions) that make  simplifying, but sometimes realistic,  assumptions to
reduce the size of the state space. 
In the next section we will look at some of these assumptions.

\section{Multi-Agent Reinforcement Learning Agents}
In the preceding section the problem of multi-agent reinforcement
learning has been introduced, as well as game theory and the links
with social choice theory.  We have also seen that the introduction of
multiple agents has, unsurprisingly, increased the size of the state
space even more than for single-agent state spaces. 
Recent work has introduced some approaches for solving multi-agent
problems, which we  will introduce here.

First, we will discuss an approach based on planning, with the name
Counterfactual regret minimization (\gls{CFR}). This algorithm is  successful
for computing complex social choice problems, such as occur in the
game of poker. CFR is suitable for competitive multi-agent problems.

Second, we will discuss cooperative reinforcement learning
methods that are  also suitable for mixed multi-agent
problems. 

Third, we discuss population-based approaches such as evolutionary
strategies and swarm computing~\cite{back1991survey,fogel1994introduction,holland1975adaptation}. These approaches are inspired by
computional  behavior that occurs in nature, such as in flocks of birds and
societies of insects~\cite{blum2008swarm,dorigo2007swarm,bonabeau1999swarm}. Such methods are well-known from single-agent
optimization problems; indeed, population-based methods are  successful in solving
many complex and large single-agent optimization problems (including stochastic gradient descent
optimization)~\cite{salimans2017evolution}. In this section, we will
see that evolutionary methods 
are also a natural match for mixed multi-agent problems,
although they are typically used for cooperative problems with many
homogeneous agents.

Finally, we will discuss an approach based on multi-play
self-learning, in which evolutionary and hierarchical aspects are
used. Here different  groups of
reinforcement 
learning agents are trained against eachother. This approach has been
highly successful in the games of Capture the Flag and StarCraft, and
is suitable for mixed multi-agent problems. 

Let us start with counterfactual regret minimization.

\subsection{Competitive Behavior}\label{sec:cfr}\index{competitive behavior}
The first setting that we will discuss is the competitive
setting. This setting is still close to single and
two-agent reinforcement learning, that are also based on competition.

\subsubsection{Counterfactual Regret Minimization}\index{CFR}\index{counterfactual regret
  minimization}

Counterfactual regret minimization (CFR) is an iterative method for
approximating the Nash equilibrium of an 
extensive-form game~\cite{zinkevich2008regret}. CFR is suitable for
imperfect information games (such as poker) and computes a strategy
that is (on average)
non-exploitable, and that is therefore robust in a competitive
setting. Central in counterfactual regret minimization is the notion
of regret. 
%
%
Regret is the loss in expected 
reward that an agent suffers for not having selected the  best
strategy, with respect to fixed choices by the other players. Regret can only be known in
hindsight. We can, however, statistically
sample  expected regret, by averaging the regret that did not happen.
CFR finds the  Nash
equilibrium  by comparing two hypothetical players against eachother, where
the opponent  chooses the action that minimizes our value.

%
CFR is a statistical algorithm that
converges to a Nash equilibrium. Just like minimax, it is a self-play
algorithm that finds the optimal strategy under the assumption of
optimal play by both sides. Unlike minimax, it is suitable for
imperfect information 
games, where information sets describe a set of possible worlds
that the opponent may hold. Like MCTS it samples,  repeating this
process for billions of games, improving its 
strategy each time. As it plays, it gets closer and closer towards an
optimal strategy for the game: a strategy that can do no worse than
tie against any opponent~\cite{johanson2012efficient,tammelin2014solving,lanctot2009monte}.
The
quality of the strategy that it computes is measured by its
\emph{exploitability}. Exploitability  is the maximum amount that a
perfect counter-strategy could win (on expectation) against it.

Although the Nash equilibrium is a strategy that is theoretically
proven to be not exploitable,
in practice, typical human play is far from the theoretical optimum,
even for top players~\cite{brown2018superhuman}.
Poker programs based on counterfactual regret minimization started
beating the world's best human players in heads-up limit hold'em in
2008, even though these programs programs were still very much
exploitable by this worst-case measure~\cite{johanson2012efficient}.

Many  papers on counterfactual regret minimization are quite
technical, and the codes for algorithms are too long to explain here in
detail. CFR is in important algorithm that is essential for
understanding progress in poker.
To make the work on poker and CFR more accessible, 
introductory papers and blogs have been
written. Trenner~\cite{trenner2020cfr} has written a blog\footnote{\url{https://ai.plainenglish.io/building-a-poker-ai-part-6-beating-kuhn-poker-with-cfr-using-python-1b4172a6ab2d}} in which CFR
is used to play Kuhn poker, one of the simplest Poker variants.
\lstset{label={lst:cfr},numbers=left}
\lstset{caption={Counter-factual regret minimization~\cite{trenner2020cfr}}}
\begin{lstlisting}[language=Python,float]
def cfr(
        self, 
        cards: List[str], 
        history: str, 
        reach_probabilities: np.array, 
        active_player: int) -> int:
    if KuhnPoker.is_terminal(history):
        return KuhnPoker.get_payoff(history, cards)

    my_card = cards[active_player]
    info_set = self.get_information_set(my_card + history)

    strategy = info_set.get_strategy(reach_probabilities[active_player])
    opponent = (active_player + 1) % 2
    counterfactual_values = np.zeros(len(Actions))

    for ix, action in enumerate(Actions):
        action_probability = strategy[ix]

        new_reach_probabilities = reach_probabilities.copy()
        new_reach_probabilities[active_player] *= action_probability

        counterfactual_values[ix] = -self.cfr(
            cards, history + action, new_reach_probabilities, opponent)

    node_value = counterfactual_values.dot(strategy)
    for ix, action in enumerate(Actions):
        counterfactual_regret[ix] = \
            reach_probabilities[opponent] * (counterfactual_values[ix] - node_value)
        info_set.cumulative_regrets[ix] += counterfactual_regret[ix]

    return node_value

def train(self, num_iterations: int) -> int:
    util = 0
    kuhn_cards = ['J', 'Q', 'K']
    for _ in range(num_iterations):
        cards = random.sample(kuhn_cards, 2)
        history = ''
        reach_probabilities = np.ones(2)
        util += self.cfr(cards, history, reach_probabilities, 0)
    return util
\end{lstlisting}
The CFR pseudocode and the function that calls it iteratively are shown in Fig.~\ref{lst:cfr}; for the
other routines, see the blog. The CFR code works as follows. 
First it checks for being in  a terminal state and returns the
payoff, just as a regular tree traversal code does.
Otherwise, it retrieves the information set and the current
regret-matching strategy. It uses the reach probability, which is the probability
that we reach the current node according to our strategy in the
current iteration.  Then CFR loops over the possible actions 
(lines 17–24), computes the new reach probabilities for the next game
state and calls itself recursively.  As there are $2$ players
taking turns in Kuhn poker, the utility for one player is exactly $-1$ times the
utility for the other, hence the minus sign in front of the \verb|cfr()|
call. What is computed here for each action is  the
counterfactual value. When the loop over all possible actions
finishes, the value of the node-value of the current state is computed
(line
26), with our current strategy. This value is the sum of the counterfactual
values per action, weighted by the likelihood of  taking this
action. Then the cumulative counterfactual regrets are updated by adding
the node-value times the reach probability of the
opponent. Finally, the node-value is returned. 

 Another accessible blog post where the algorithm is explained 
step by step has been written by 
Kamil Czarnog\`orski,\footnote{
  \url{https://int8.io/counterfactual-regret-minimization-for-poker-ai/}}
with code on GitHub.\footnote{
\url{https://github.com/int8/counterfactual-regret-minimization/blob/master/games/algorithms.py}}

\subsubsection{Deep Counterfactual Regret Minimization}
Counterfactual regret minimization is a  tabular
algorithm that traverses the extensive-form game tree from root to
terminal nodes, coming closer to the Nash equilibrium  with each
iteration. Tabular algorithms do not scale well to large problems,
and  researchers often have to use domain-specific heuristic abstraction
schemes~\cite{sandholm2015abstraction,ganzfried2015endgame,brown2015hierarchical}, 
alternate methods for regret updates~\cite{tammelin2014solving}, or
sampling variants~\cite{lanctot2009monte} to achieve acceptable
performance. 

For  large problems a deep learning version of the algorithm has
been developed~\cite{brown2019deep}.
The goal of deep counterfactual regret minimization is to approximate
the behavior of the tabular algorithm without
calculating  regrets at each individual information set. It  generalizes across
similar infosets using  approximation of the value function via a deep 
neural network with alternating player updates.

\subsection{Cooperative Behavior}\index{MADDPG}\index{multi-agent DDPG}\index{cooperation}
CFR is an algorithm for the competitive setting.
The Nash-equilibrium defines the competitive win/lose multi-agent case---the
$(-5,-5)$ situation of the prisoner's dilemma of Table~\ref{tab:pd}.

We will now move to the cooperative setting. As we have seen, in a
cooperative setting,  win/win situations are possible, with higher rewards, both for society as a
whole and for the individuals. The Pareto optimum
for the prisoner's dilemma example is 
$(-2,-2)$, only achievable through norms, trust or cooperation by the agents
(as close-knit criminal groups  aim to achieve, for example,  through a code of
silence), see also  Leibo et al.~\cite{leibo2017multi}.

The achievements in
single-agent reinforcement learning inspire researchers
 to achieve similar results in multi-agent.
However, partial observability and nonstationarity
create a  computational challenge.  Researchers have tried many
different approaches, some of which we will cover, although  the size of
problems for which the current agorithms work is
still limited. 
Wong et al.~\cite{wong2021survey} provide a review of these
approaches, open problems in cooperative reinforcement learning
are listed by Dafoe et al.~\cite{dafoe2020open}.
First we will discuss approaches based on
single-agent reinforcement learning methods, next we
will discuss approaches based on opponent modeling,
communication, and psychology~\cite{wong2021survey}.

\subsubsection{Centralized Training/Decentralized Execution}

While the dec-POMDP model offers an appropriate framework for cooperative
sequential decision making under uncertainty, 
solving a large dec-POMDP is intractable~\cite{bernstein2002complexity}, and
therefore many relaxations of the problems have been developed,
where some elements, such as communication, or training, are
centralized, to increase tractability~\cite{wong2021survey,tan1993multi}.

One of the easiest approaches to train a policy for a multi-agent
problem is to train the
 collaborating agents with a centralised controller, effectively
reducing a decentralized multi-agent computation to a centralized
single-agent computation. In this approach all agents send their 
observations and local policies to a central controller, that now has
perfect information, and that decides
which action to take for each agent. 
However, as  large collaborative problems are computationally
 expensive, the single controller becomes overworked, and this approach
does not scale.

On the other extreme, we can ignore communication and
nonstationarity and let agents train separately. In this approach the agents  learn an
individual action-value function and view other agents as part of the
environment. This approach simplifies the computational demands at the
cost of gross oversimplification, ignoring multi-agent interaction.

An in-between approach is centralized training and decentralized
execution~\cite{kraemer2016multi}. Here agents can access
extra information during training, such as other agents’ observations,
rewards, gradients and parameters. However, they execute their policy
decentrally based on their local observations. The local computation
and inter-agent
communication mitigate nonstationarity while still  modeling partial
observability and (some) interaction. This approach stabilises the
local policy learning  of agents, 
even when other agents’ policies are changing~\cite{kraemer2016multi}.

When value functions are learned centrally,  how should this
function then be used for decentral execution by the agents? A popular method is value-function factorization. The Value
decomposition networks method (VDN) decomposes the central value function as
a sum of individual value functions~\cite{sunehag2017value}, who are
executed greedily by the agents. QMIX and QTRAN are two methods that
improve on VDN by allowing nonlinear
combinations~\cite{rashid2018qmix,son2019qtran}. Another approach is
Multi-agent variational exploration (MAVEN) which improves the
inefficient exploration problem of QMIX using a latent space model~\cite{mahajan2019maven}.

\index{MADDPG}
Policy-based methods focus on actor critic approaches, with a
centralized critic training decentralized actors. Counterfactual
multi-agent (COMA) uses such a centralized critic to approximate the
Q-function that has access to the  actors that train the
behavior policies~\cite{foerster2018counterfactual}.

Lowe et al.~\cite{lowe2017multi} introduce a multi-agent version of a
popular off-policy single-agent deep policy-gradient algorithm DDPG
(Sect.~\ref{sec:ddpg}), called \gls{MADDPG}. It considers action
policies of other agents and their coordination. MADDPG uses an ensemble of
policies for each agent. It uses a decentralized actor, centralized
critic approach, with deterministic policies. MADDPG works for both
competitive and collaborative multi-agent problems. An extension for
collision avoidance is presented by Cai et
al.~\cite{cai2021safe}. A popular on-policy single-agent method is
PPO. Han et al.~\cite{han2021multiagent} achieve sample efficient results modeling the continuous
Half-cheetah task as a model-based multi-agent problem. Their
model-based, multi-agent, work is inspired by
MVE~\cite{feinberg2018model} (Sect.~\ref{sec:mve}). Yu et
al.~\cite{chao2021surprising} achieve good results in cooperative
multi-agent games (StarCraft, Hanabi, and Particle world) with MAPPO.\index{MAPPO}

Modeling cooperative behavior in reinforcement learning in a way that is computationally
feasible  is  an active area of research. 
Li et al.~\cite{li2020deep} use implicit coordination graphs to model
the structure of interactions. They use graph neural networks to model
the coordination graphs~\cite{guestrin2001multiagent} for StarCraft
and traffic environments~\cite{samvelyan2019starcraft}, allowing
scaling of interaction patterns that are learned by the graph
convolutional network.

\subsubsection{Opponent Modeling}\index{opponent modeling}
The state space of multi-agent problems is large, yet the previous
approaches  tried to learn this large space with adaptations of
single-agent algorithms.  Another
approach is to reduce the size of the state space, for
example by explicitly modeling opponent behavior in the agents. These
models can then be used to guide the agent's decision making, reducing
the state space that it has to traverse.  Albrecht and Stone have
written a survey of approaches~\cite{albrecht2018autonomous}.

One approach to reduce the state space is to assume a set of
stationary policies between which agents
switch~\cite{everett2018learning}. The Switching
agent model (SAM)~\cite{zheng2018deep} learns an opponent model from observed trajectories
with a Bayesian network. The Deep reinforcement open network
(DRON)~\cite{he2016opponent} uses two networks, one to learn the
Q-values, and the other to learn the opponent policy representation.

\index{theory of mind}
Opponent modeling is related to  the psychological Theory of
Mind~\cite{premack1978does}. According to this Theory, people
attribute mental states to others, such as beliefs, intents, and
emotions. Our theory of the minds of others helps us to analyze and predict
their behavior.  Theory of mind also holds that we assume that the
other has theory of mind; it allows for a nesting of beliefs of the
form: ``I believe that you believe that I believe''
\cite{van2015narrative,van2016lazy,van2019recursive}. Building on
these concepts, Learning with 
opponent-learning awareness (LOLA) anticipates opponent's
behavior~\cite{foerster2017learning}. Probabilistic recursive
reasoning (PR2) models our own and our opponent's behavior as a
hierarchy of perspectives~\cite{wen2019probabilistic}. Recursive
reasoning has been shown to lead to faster convergence and better
performance~\cite{moreno2021neural,dai2020r2}. Opponent modeling is also an active
area of research.

\subsubsection{Communication}
Another  step  towards modeling the 
real world is taken when we explicitly 
model communication between agents. A fundamental question is how
language between agents emerges when no predefined communication
protocol exists, and how syntax and meaning evolve out of
interaction~\cite{wong2021survey}. A basic approach to communication is
with referential games: a sender sends two images and a message; the
receiver then has to identify which of the images was the
target~\cite{lazaridou2016multi}. Language also emerges in more
complicated versions, or in negotiation between
agents~\cite{cao2018emergent,kottur2017natural}.

Another area where multi-agent systems are frequently used is the
study of coordination, social
dilemmas, emergent phenomena and evolutionary
processes, see, for example~\cite{tampuu2017multiagent,eccles2019learning,leibo2017multi}. 
%
%
In the card game bridge~\cite{smith1998computer} bidding strategies have been
developed to signal  to the other player in
the team which cards a player has~\cite{smith1998computer}. In the game of Diplomacy, an
explicit negotion-phase is part 
of each game round~\cite{de2018challenge,kraus1988diplomat,paquette2019no,anthony2020learning}. Work is ongoing to design communication-aware
variants of  reinforcement learning
algorithms~\cite{simoes2020multi,hausknecht2016cooperation}. 

\subsubsection{Psychology}
Many of the key ideas in reinforcement learning, such as operant
conditioning and trial-and-error, originated in cognitive
science. Faced with the large state space, multi-agent reinforcement
learning  methods are moving towards human-like agents. In addition to
opponent modeling, studies focus
on coordination, pro-social behavior, and intrinsic motivation. A
large literature exists on emergence of social norms and cultural evolution in multi-agent
systems~\cite{leibo2017multi,boyd1988culture,axelrod1997complexity,dawkins2017selfish}. To 
deal with nonstationarity and large states spaces,  humans use heuristics and
approximation~\cite{gigerenzer1996reasoning,marewski2010good}. However,
heuristics can lead to biases and suboptimal
decision-making~\cite{gilovich2002heuristics}. It is interesting to
see how  multi-agent modeling is discovering concepts from
psychology. More research in this area is likely to improve the
human-like behavior of artificial agents.

\subsection{Mixed Behavior}\index{evolutionary
  strategies}\index{population-based strategies}\label{sec:evo}\label{sec:swarm}


To discuss solution methods for agents in the  mixed setting,  we
will look at one important  approach that is again inspired by
biology: population-based algorithms.

Population-based methods such as evolutionary algorithms and swarm
computing work by evolving (or optimizing) a large number of agents at
the same time. 
We will look
closer at evolutionary algorithms and at swarm computing, and then we
will look at the role they play in multi-agent reinforcement learning.

\subsubsection{Evolutionary Algorithms}\index{evolutionary
  algorithms}\label{sec:evo-alg}
Evolutionary algorithms are inspired by bio-genetic 
processes of reproduction: mutation, recombination, and
selection~\cite{back1996evolutionary}. 
Evolutionary  algorithms work with large populations of simulated
individuals, which  typically makes it easy  to parallelize and run them
on large computation clusters.

Evolutionary algorithms often achieve good results in optimizing
diverse problems. For example, in optimizing single agent problems, an
evolutionary approach would model the problem as a population of
individuals, in which each individual represents a candidate solution
to the problem. The candidate's quality is determined by a fitness
function, and the best candidates are selected for reproduction. New
candidates are created through crossover and mutation of genes, and the cycle
starts again, until the quality of candidates stabilizes.  In this way an evolutionary algorithm iteratively
approaches the optimum. Evolutionary algorithms are randomized
algorithms, that can circumvent local optima.

Although they are best known for solving 
single agent optimization problems, we will use them here  to model
multi-agent problems.

\begin{algorithm}[t]
    \caption{Evolutionary Framework \cite{back1996evolutionary}}
    \begin{algorithmic}[1] 
            \State Generate the initial population randomly
            \Repeat{}
                \State Evaluate the fitness of each individual of the population
                \State  Select the fittest individuals for reproduction
                \State  Through crossover and mutation generate  new individuals
                \State  Replace the least fit individuals by the new individuals
        \Until {terminated}
    \end{algorithmic}
\label{alg:evo}
\end{algorithm}

Let us look in more detail at how an evolutionary algorithm works (see Alg.~\ref{alg:evo})~\cite{back1996evolutionary}. First an
initial population is generated. The fitness of each individual is
computed, and the fittest individuals are selected for reproduction,
using crossover and mutation to create a new generation of
individuals. The least fit individuals of the old populations are
replaced by the new individuals.

%
Compared to reinforcement learning, in evolutionary algorithms the
agents are typically homogeneous, in the 
sense that the reward (fitness) function for the individuals is the
same.  Individuals do have  different genes, and
thus differ in their behavior (policy). In reinforcement learning
there is a single current behavior policy, where an evolutionary approach has many
candidate policies (individuals).
The fitness function can engender in principle both competitive and
cooperative 
behavior between individuals, although a typical optimization scenario
is to select 
a single individual with the genes  for the highest
fitness (survival of the fittest competitor).\footnote{Survival of the fittest
  \emph{cooperative group} of individuals can also be 
achieved with an appropriate fitness function~\cite{ma2018survey}.}

Changes to genes  of individuals (policies) occur explicitly via crossover and (random)
mutation, and implicitly via selection for fitness. In reinforcement
learning the reward is used more directly as a policy goal; in
evolutionary algorithms the fitness does not directly influence the
policy of an individual, only its survival. 

Individuals in  evolutionary algorithms are passive entities that do
not communicate or act, although they do combine to create new
individuals.

There are  similarities and differences  between evolutionary and
multi-agent reinforcement learning
algorithms. 
First
of all, in both approaches the goal is to find the optimal solution,
the policy that maximizes (social) reward.  In reinforcement learning
this occurs by learning a policy through
interaction with an environment, in  evolutionary algorithms by
evolving a population through survival of the fittest. Reinforcement
learning deals with a limited number of agents whose policy determines their actions,
evolutionary algorithms deals with many individuals whose genes
determine 
their survival. Policies are improved using a reward function that
assesses how good actions are, genes mutate and combine, and individuals are selected
using a fitness function. Policies are improved ``in place'' and
agents do not die, in evolutionary computation the traits
(genes) of the best individuals
are selected and copied to new individuals in the next generation
after which 
the old generation does die.

Although different at first sight, the two approaches share many
traits, including the main goal: optimizing behavior. Evolutionary
algorithms are inherently multi-agent, and may work well in finding
good solutions in large and nonstationary sequential decision problems.

\subsubsection{Swarm Computing}\index{swarm
  computing}\index{decentralized computing}\index{Ant colony
  optimization}\index{collective intelligence}
Swarm computing is related to evolutionary algorithms~\cite{beni1993swarm,blum2008swarm}. Swarm computing focuses
on emerging behavior in 
decentralized, collective, self-organized systems. Agents are
typically simple, numerous, and homogeneous, and interact locally with each
other and the environment. Biological examples of swarm intelligence
are   behavior in ant colonies, bee-hives, flocks of birds, and
schools of fish, Fig.~\ref{fig:flock}~\cite{sunehag2019reinforcement};
image by Wikimedia. Behavior is typically cooperative through
decentralized communication mechanisms. In artificial swarm
intelligence, individuals are sometimes able to
imagine the behavior of other individuals, when they have a Theory of
  mind~\cite{baron1985does,gallese1998mirror}.\index{theory of mind} Swarm
intelligence, or collective intelligence in general, is a form of
decentralized computing, as opposed to 
reinforcement learning, where external algorithms calculate optimal behavior
in a single classical centralized algorithm~\cite{woolley2010evidence}.

Although both approaches work for the mixed setting,  evolutionary
algorithms tend to stress competition and survival of the
fittest (Nash), where swarm computing stresses cooperation and survival of
the group (Pareto).

\begin{figure}[t]
    \centering{\includegraphics[width=5cm]{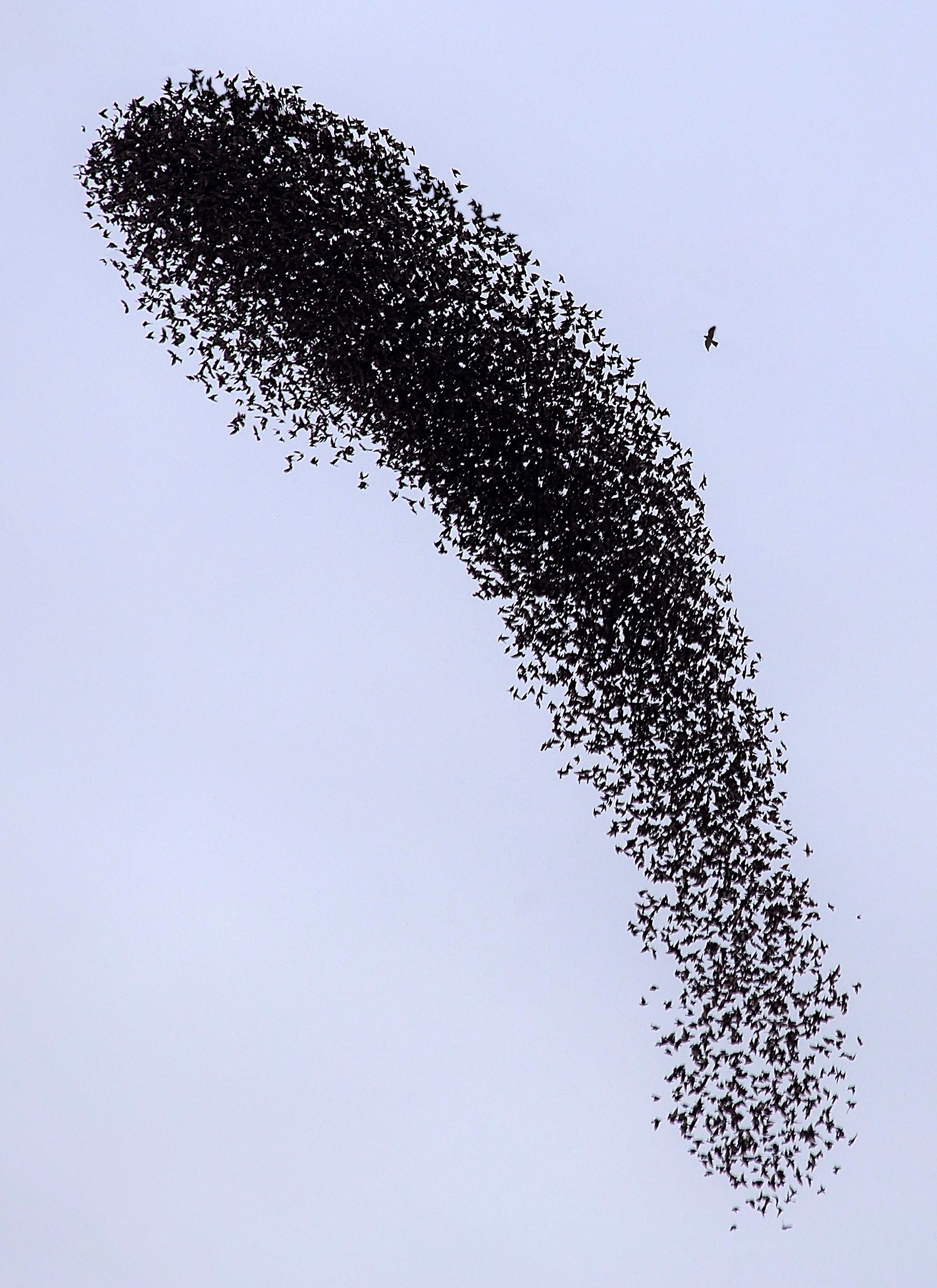}
     \caption{A Flock of Starlings, and One Predator}\label{fig:flock}}
\end{figure}

A well-known example of artificial swarm
intelligence is Dorigo's Ant colony optimization
algorithm (\gls{ACO}) which is a probabilistic optimization algorithm modeled
after the pheromone-based communication of biological
ants~\cite{dorigo1992optimization,dorigo2006ant,dorigo1997ant}. 

Emergent behavior in multi-agent reinforcement learning is
specifically studied
in~\cite{liu2019emergent,jaderberg2017population,mordatch2018emergence,heess2017emergence,bansal2017emergent,leibo2019autocurricula}. For
decentralized
algorithms  related to solving multi-agent
problems see, for example~\cite{oliehoek2016concise,zhang2018fully,omidshafiei2017deep,oliehoek2013incremental}.


\subsubsection{Population-Based Training}

Translating traditional value or policy-based reinforcement learning algorithms
to the multi-agent setting is non-trivial. It is interesting to see
that, in contrast, evolutionary algorithms, first designed
for single-agent optimization using a population of candidate
solutions, translate so naturally to the multi-agent setting, where a
population of agents is used to find a shared solution that is optimal for
society.

In evolutionary algorithms agents are typically homogeneous, although they
can  be heterogeneous,   with different
fitness functions. The fitness functions may be competitive, or
cooperative. In 
the latter case the increase of reward for one agent can also imply an
increase for other agents, possibly for the entire group, or
population.
%
%
Evolutionary algorithms are quite effective optimization
algorithms. Salimans et al.~\cite{salimans2017evolution} report that
evolution strategies  rival the performance of 
standard reinforcement learning  techniques on modern 
benchmarks, while being easy to parallelize.
In particular, evolutionary strategies are  simpler to implement
(there is no need for 
backpropagation),  are easier to scale in a distributed setting, 
do not suffer in settings with sparse rewards, and have fewer
hyperparameters. 

\begin{algorithm}[t]
    \caption{Population Based Training \cite{jaderberg2017population}}
    \label{pbtalg}
    \begin{algorithmic} 
        \Procedure{Train}{$\mathcal{P}$} \Comment{initial population $\mathcal{P}$}
        \State Population $\mathcal{P}$, weights $\theta$,
        hyperparameters $h$, model evaluation $p$, time step $t$
            \For{($\theta$, $h$, $p$, $t$) $\in \mathcal{P}$ (asynchronously in parallel)}
            \While{not end of training} 
                \State $\theta \gets \mathtt{step}(\theta|h)$ \Comment{one step of optimisation using hyperparameters $h$}
                \State $p \gets \mathtt{eval}(\theta)$ \Comment{current model evaluation}
                \If{$\mathtt{ready}(p,t,\mathcal{P})$}
                \State $h', \theta' \gets \mathtt{exploit}(h,\theta,p,\mathcal{P})$\Comment{use the rest of population for improvement}
                \If{$\theta \neq \theta'$}
                \State $h, \theta \gets \mathtt{explore}(h', \theta', \mathcal{P})$ \Comment{produce new hyperparameters $h$}
                \State $p \gets \mathtt{eval}(\theta)$  \Comment{new model evaluation}
                \EndIf
                \EndIf
            \State update $\mathcal{P}$ with new $(\theta, h, p, t + 1)$ \Comment{update population}
            \EndWhile\label{euclidendwhile}
            \EndFor
            \State \textbf{return} $\theta$ with the highest $p$ in $\mathcal{P}$
        \EndProcedure
    \end{algorithmic}
\label{alg:pbt}
\end{algorithm}

Evolutionary algorithms are a form of population-based
computation that can be used to compute strategies that are optimal in
a game-theoretic Nash sense, or that try to find strategies that
out-perform other agents.
The evolutionary approach to optimization is efficient, robust, and
easy to parallelize, and thus has some advantages over the stochastic
gradient approach. 
How does this approach relate to multi-agent reinforcement learning,
and  can it  be used to find efficient
solutions for multi-agent reinforcement learning problems?

In recent years a number of research teams have reported successful
efforts in creating reinforcement learning players for multi-agent
strategy games (see Sect.~\ref{sec:maenv}). These were all large research efforts, where a range
of different approaches was used, from self-play reinforcement learning,
cooperative learning, hierarchical modeling, and evolutionary  computing.

Jaderberg et al.~\cite{jaderberg2017population} report success in Capture the Flag games with a
combination of ideas from evolutionary algorithms and self-play
reinforcement learning. Here a population-based approach of self play
is used where teams of diverse agents are trained in tournaments
against each other.
Algorithm~\ref{alg:pbt} describes this Population based training
(\gls{PBT}) approach in more detail. Diversity is enhanced through
mutation of 
policies, and performance is improve through culling of
under-performing agents.

Population based training uses two methods. The first is
\verb|exploit|, which decides whether a worker should abandon the
current solution and  focus on a more promising one. The second is \verb|explore|, which, given the current solution and hyperparameters,
proposes new solutions to  explore the solution space. 
Members of the population are trained in parallel. Their weights
$\theta$ are updated and \verb|eval| measures their current
performance. When a member of 
the population is deemed \emph{ready} because it has reached a certain
performance threshold, its weights and hyperparameters are updated by
\verb|exploit| and \verb|explore|, to  replace the current
weights with the weights that have the highest recorded performance in
the rest of the population, and to randomly perturb the
hyperparameters with noise. After \verb|exploit| and \verb|explore|,
iterative training continues as before until convergence.

Let us have a look at this fusion approach for training leagues of
players.

\subsubsection{Self-Play Leagues}\index{league learning}
Self-play league learning 
refers to the training of a multi-agent league of individual game
playing characters. As we will see in the next section, variants have
been used to play games such as StarCraft, 
Capture the Flag,  and Hide and Seek.

League learning
 combines population-based training with self-play training as in
AlphaZero. In league learning, the agent plays against a league of different
opponents, while being part of a larger team of agents that is being
trained. The team of agents is  managed to have enough
diversity in order to provide a stable training goal to reduce
divergence or local minima. League learning employs evolutionary 
concepts such as mutation of behavior policies and culling of
under-performing agents from the population. The team employs
cooperative strategies, in addition to competing against the other
teams. Agents are trained in an explicit hierarchy.

The goal of self-play league training is to find stable policies for all agents, that maximize their team reward. In mixed and cooperative
settings the teams of agents may benefit from each others' strength
increase~\cite{lowe2017multi}.
Self-play league learning  also uses aspects of hierarchical
reinforcement learning, a topic that will be covered in the next chapter.

In the next section we will look  deeper into how self-play league learning
is implemented in specific multi-player games.

\section{Multi-Agent  Environments}\label{sec:maenv}

Reinforcement learning has achieved quite a few imaginative results in
which it  has succeeded in emulating behavior that
approaches human behavior in the real world. 
In this chapter we have made a step towards modeling  more
behavior that is even closer to the real world. 
Let us summarize in this section the results in four different multi-agent games, some of 
which have been published in  prestigious scientific journals.

We will use the familiar sequence of (1) 
competitive, (2) cooperative, and (3)  mixed environments. For each we will sketch the problem,
outline the algorithmic approach, and summarize the achievements.

\begin{table}[t]
  \begin{center}
  \begin{tabular}{lcll}
    {\bf Environment} & {\bf Behavior}&{\bf Approach}  & {\bf Ref}  \\
    \hline\hline
    Poker &  Competitive& (Deep) Counterfactual regret minimization & \cite{bowling2015heads,brown2019superhuman,moravvcik2017deepstack}  \\
    Hide and Seek & Cooperative& Self-play, Team hierarchy & \cite{baker2019emergent} \\ 
    Capture the Flag & Mixed&Self-play, Hierarchical, Population-based  & \cite{jaderberg2019human} \\
    StarCraft II &  Mixed&Self-play, Population-based   & \cite{vinyals2019grandmaster} \\ 
    \hline
  \end{tabular}
  \caption{Multi-Agent Game Approaches}\label{tab:multigame}
\end{center}
\end{table}
Table~\ref{tab:multigame} lists the multi-agent games and their
dominant approach.

\subsection{Competitive Behavior: Poker}\label{sec:poker}\index{poker}\index{Libratus}


Poker is a popular imperfect-information game. It is played competitively
and  human championships are organized regularly.
Poker has been studied for some time in artificial
intelligence, and  computer poker championships have been conducted
regularly~\cite{bard2013annual,billings2002challenge}.
Poker is a 
competitive game. Cooperation (collusion, collaboration between two players
to the detriment of a third player) is possible, but often does not
occur in practice; finding the Nash equilibrium therefore is a
successful approach to playing the game in practice. The method of counterfactual regret minimization has
been developed specifically to make progress in poker.

There are many variants of poker that are regularly played. No-limit
Texas hold’em is a  popular variant; the  two-player 
version is called Heads Up and is easier to analyse because no
opponent collusion can occur.  Heads-up 
no-limit Texas hold’em (HUNL) has been the primary AI benchmark for 
imperfect-information game play for several years.

Poker has hidden information (the face-down cards).
Because of this, agents are faced with
a large number of possible states;  poker is a game that is far
more complex than chess or checkers. 
The state space of the two-person HUNL version is reported to be around 
$10^{161}$~\cite{brown2018superhuman}. A further complication
is that during the course of the game information is revealed by
players through their bets; high bets indicate good cards, or the wish
of the player to make the opponent believe that this is the case
(bluffing).  Therefore, a player  must choose  between
betting high on good cards, and on doing the opposite,  so that the
opponent does not find out too much, and can counter-act.

In 2018, one of the top two-player poker programs, Libratus, 
defeated top human professionals in HUNL in a 20-day,
120,000-hand competition featuring a \$200,000 prize
pool. Brown et al.~\cite{brown2018superhuman} describe the
architecture of Libratus. The program consists of three main modules,
one for computing a quick CFR Nash-policy using a smaller version of the
game, a second module for constructing a finer-grained strategy once a
later stage of the game is reached, and a third module to enhance the
first policy by filling in missing branches.

In the experiment against  top
human players, Libratus analyzed the bet
sizes that were played most often by its
opponents at the end of each day of the competition. 
The programs  would then calculate a response overnight, in order  
to improve
as the competition proceeded.

\begin{figure}[t]
\begin{center}
\includegraphics[width=5cm]{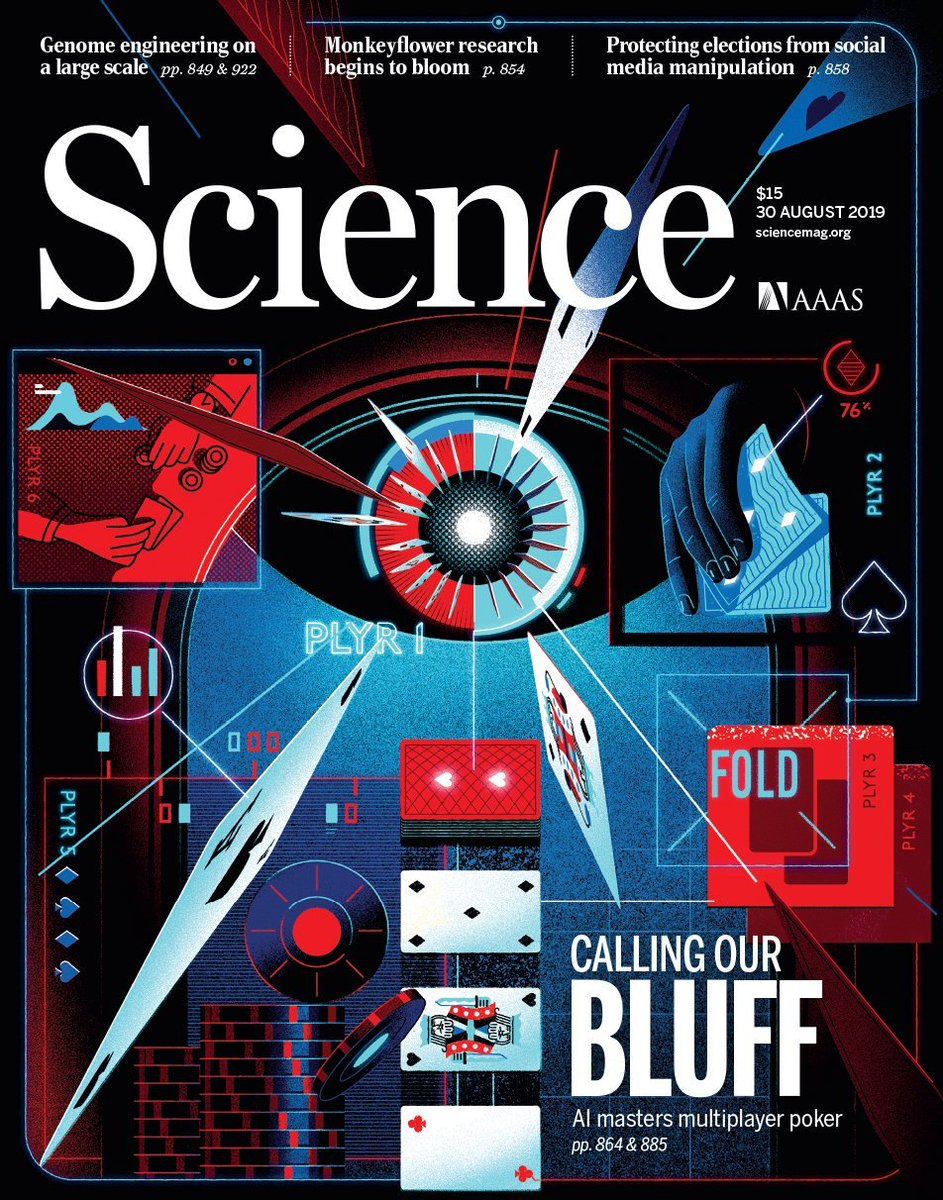}
\caption{Pluribus  on the Cover of Science}\label{fig:pluribus}
\end{center}
\end{figure}

Two-agent Libratus was originally based on heuristics,
abstraction, and tabular  counterfactual regret
minimization---not on deep
reinforcement learning. For multi-agent poker the program Pluribus was
developed. For 
Pluribus,  the authors used \emph{deep} 
counterfactual regret minimization~\cite{brown2019deep}, with a 
7-layer neural network that followed the AlphaZero self-play approach.  Pluribus defeated top players in six-player
poker~\cite{brown2019superhuman} (Fig.~\ref{fig:pluribus}).\index{Pluribus}\index{DeepStack}  
Pluribus  uses an AlphaZero approach of  self-play in combination
with search. 
Another top program, DeepStack, also uses randomly generated games to train 
a deep value function network~\cite{moravvcik2017deepstack}. 



\subsection{Cooperative Behavior: Hide and Seek}
\index{Hide and Seek}\label{sec:hide}

In addition to competition, cooperative behavior is part of 
real-world behavior. Indeed, cooperation is what defines our social
fabric, and much of our society consists of different ways in which we
organize ourselves in families, groups, companies, parties, filter
bubbles, and
nations. The question of how voluntary cooperation can emerge between
individuals has been studied extensively, and the work in tit-for-tat
(Sect.~\ref{sec:tit-for-tat}) 
is just one of many studies in this fascinating
field~\cite{axelrod1997complexity,boyd1988culture,woolley2010evidence,leibo2017multi,sabater2002reputation}.
  





One study into emergent cooperation has been performed with a version
of the game of Hide and Seek.
Baker et al.\cite{baker2019emergent} report on an experiment in which
they used  MuJoCo to build
a new multi-agent game environment. The environment  was created with
the specific purpose of studying how cooperation
emerged out of the combination of a few simple rules and reward
maximation (see Fig.~\ref{fig:six}).  


In the Hide and Seek experiment, 
the game is played on a randomly generated grid-world where props are
available, such as boxes. The
reward function stimulates hiders to  avoid the line of sight 
of seekers, and vice versa. There are objects scattered throughout the
environment that the agents can grab and  lock in place.
The environment contains one to three hiders and one to three seekers,
there are three 
to nine movable boxes, some elongated. There are also two movable ramps. Walls
and rooms are static and are randomly generated. Agents can see, move, and
 grab objects.
A good way to
understand the challenge is to view the videos\footnote{\url{https://www.youtube.com/watch?v=kopoLzvh5jY&t=10s}} on the Hide and Seek blog.\footnote{\url{https://openai.com/blog/emergent-tool-use/}}

With only  a visibility-based reward function, the agents are able to learn many
different skills, including collaborative tool use. For example, hiders learn
 to create shelters by barricading doors 
or constructing multi-object forts, so that the seekers can never see
them anymore, until, as a counter strategy, seekers
learned to use ramps to jump into the shelter. 
In effect, 
out of the agents' interaction  a training curriculum emerges in which
the agents learn tasks, 
many of which
require sophisticated tool use and coordination.

Hide and Seek features cooperation (inside the team) and
competition (between the hiders and the seekers). It uses a self-play version of
PPO for policy learning. It is interesting to
see how easy cooperation emerges. The game does not
have explicit communication for the team to coordinate cooperation,
all cooperative behavior emerges out of basic interaction between
agents that are guided by their reward functions. 
Cooperative strategies thus emerge out of the game design:  the homogeneous
reward functions for each team and the environment in which blocks are
present and follow  laws of physics.

\begin{figure}[t]
\begin{center}
\includegraphics[width=\textwidth]{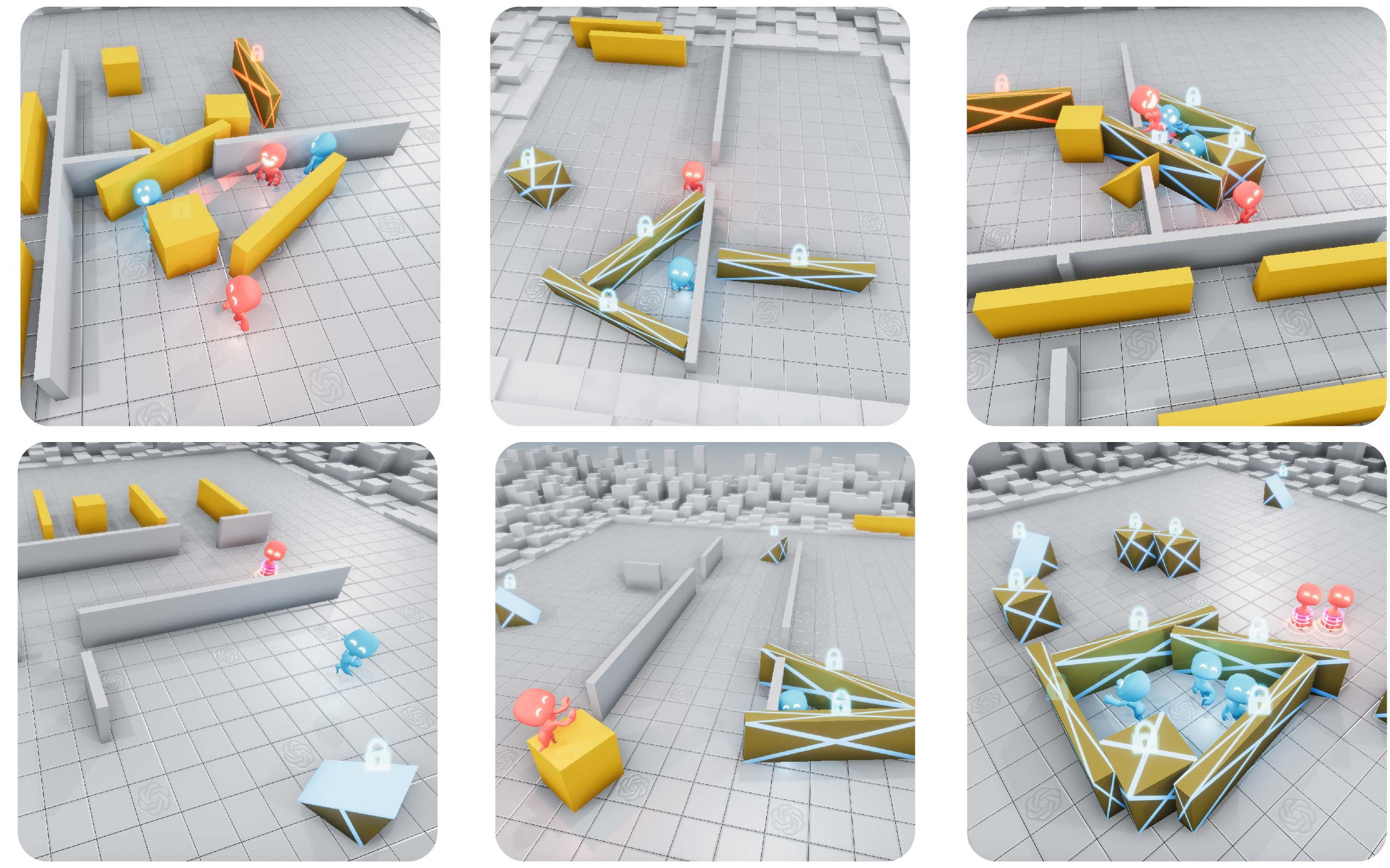}
\caption[Six Strategies in Hide and Seek]{Six Strategies in Hide and
  Seek: Running and Chasing; Fort Building; Ramp Use; Ramp Defense;
  Box Surfing; Surf Defense (left-to-right, top-to-bottom) \cite{baker2019emergent}}\label{fig:six}
\end{center}
\end{figure}

\index{reward is enough}
During play the agents essentially construct an autocurriculum for
themselves~\cite{leibo2019autocurricula,baker2019emergent}. Six different behavior strategies are reported,
each more advanced than the other, increasing the competitive pressure to find counter
strategies for the opponent (see Fig.~\ref{fig:six}). 

Baker et al.~\cite{baker2019emergent} report that initially, the hiders and seekers learn the basic strategy of running
away and chasing. However, after much training (25 million episodes),
hiders start  to use boxes to  construct  shelters behind which they
hide. Then, after another 75 million episodes, the seekers learn  to
move and use ramps to jump over obstacles into  the  shelter. A mere
10 million episodes later, the hiders learn to defend by moving the
ramps to the edge and lock them in place out of range of the
shelters. Then, after a long  270 million  episodes of training,
the seekers learned \emph{box-surfing}. They moved a box to the edge
of the play area next to the locked ramps. One seeker then used the
ramp to climb on top of the box and the other seekers push it to the
shelter, where the seeker could peek over te edge and see the hiders.
Finally,  in response, the hiders locked all of the boxes in place
before building their shelter, and they were safe from the seekers.

The Hide and Seek experiment is interesting because of the emergence
of diverse behavior strategies. The strategies   emerged out of a basic reward function and
random exploration (see also the \emph{reward is enough} argument~\cite{silver2021reward}).

The emergence of strategies out of basic reward and exploration
suggests an evolutionary process. However, Hide and
Seek does not employ population-based training or evolutionary
algorithms,  in contrast to the work in Capture the
Flag, in the next
section.

\subsection{Mixed Behavior: Capture the Flag and StarCraft}
The world around us exhibits a mix of competitive and cooperative
behavior. 
Team collaboration is an important aspect of human life, and it has
been studied extensively in biology, sociology and artificial
intelligence. It emerges 
(evolves) out of the 
most basic settings---the need to achieve an ambitious goal---as we just saw.
In recent years many research groups have studied the mixed
multi-agent model in real time strategy games, such as Capture the
Flag, and StarCraft. We will discuss both games.

\subsection*{Capture the Flag}\index{Capture the Flag}


First we will discuss the game Capture the Flag, which is played in a
Quake III Arena (see Fig.~\ref{fig:ctf}).
 \begin{figure}[t]
 \begin{center}
 \includegraphics[width=\textwidth]{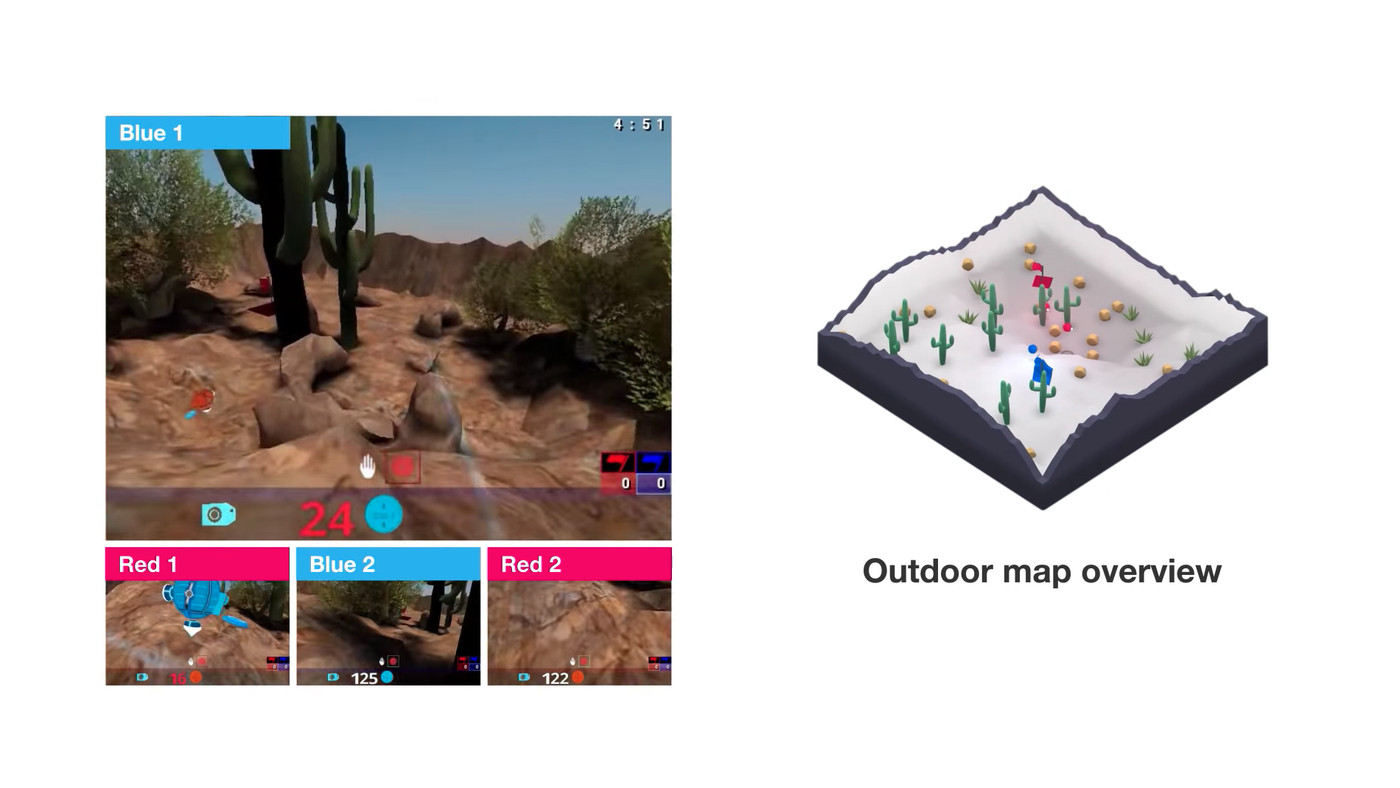}
 \caption{Capture the Flag \cite{jaderberg2019human}}\label{fig:ctf}
 \end{center}
\end{figure}
Jaderberg et al.\ have reported on an extensive experiment with this
game ~\cite{jaderberg2019human}, in which  the agents  learn from scratch  to see, act,
cooperate, and compete. In this experiment the agents are trained with
population-based self-play~\cite{jaderberg2017population}, Alg.~\ref{alg:pbt}.  The agents in the population are all
different (they have different genes). The population learns  by playing against each
other, providing increased diversity of teammates and
opponents, and a more stable and faster learning process than
traditional single-agent deep reinforcement learning methods. In total  30 different
bots were created and pitted  against each other. Agents are
part of a team, and the reward functions form a 
hierarchy. A two-layer optimization process optimizes
the internal rewards  for winning, and uses reinforcement learning on
the internal rewards to learn the  policies. 

In Capture the Flag,  the bots start by acting randomly. After 450,000
games,  a bot strategy was found that performed well,  
 and
they  developed  cooperative strategies, such as  following team mates
in order to outnumber opponents, and loitering near the enemy base when
their team mate has the flag. Again, as in Hide and Seek, cooperative
strategies emerged out of the basic rules, by combining 
environment feedback and survival of the fittest.

The work on Capture the Flag is notable since it demonstrated that
with only pixels  as input an agent can learn
to play  competitively in a rich multi-agent environment. To do
so it used a combination of population based training, internal reward
optimization, and temporally hierarchical reinforcement learning (see
next chapter). 

\subsection*{StarCraft}\label{sec:starcraft}\index{StarCraft}\index{AlphaStar}
The final game that we will discuss in this chapter is StarCraft. 
StarCraft is a multi-player real-time strategy game of even larger
complexity. The state space has been estimated to be on the order of
$10^{1685}$~\cite{ontanon2013survey}, a very large number.  StarCraft
features multi-agent decision making under uncertainty, spatial and temporal
reasoning,  competition, team-collaboration, opponent modeling, and
real-time planning. 
%
Fig.~\ref{fig:starcraft} shows a picture of a StarCraft II scene.

Research on StarCraft has been ongoing for some
time~\cite{ontanon2013survey}, a special StarCraft multi-agent
challenge has been introduced~\cite{samvelyan2019starcraft}. A team from DeepMind has
created a program 
called  AlphaStar.
In a series of test matches held in December 2018, DeepMind's AlphaStar
beat two top players in  two-player single-map matches, using a different
user interface. 
 
AlphaStar plays the full game of StarCraft II. 
The neural
network was initially trained by supervised learning from anonymized
human games, that were then further trained by playing against other
AlphaStar agents, using a  population-based version of self-play
reinforcement
learning~\cite{vinyals2019grandmaster,jaderberg2017population}.\footnote{Similar
  to the first approach in AlphaGo, where self-play reinforcement
  learning was also bootstrapped by supervised learning from human games.}  
These agents are  used to seed a multi-agent reinforcement learning
process. A continuous competitive league was created, with the agents of the
league playing games in competition against each other. By
branching from existing competitors, new competitors were 
added. Agents learn from games against other
competitors. Population-based learning was taken further, creating a
process that explores the very large  space of StarCraft game play
by pitting agents against strong opponent strategies, and retaining
strong early strategies.
   
Diversity in the league is increased by giving each agent  its own
learning objective, such as which competitors it should focus on and
which game unit it should build. A  form of prioritized league-self-play actor-critic
 training is used, called prioritized fictitious self-play---details
 are in~\cite{vinyals2019grandmaster}. 
AlphaStar was trained on a custom-built  scalable distributed
training system  using Google's  tensor processing units (TPU).
The AlphaStar league was run for 14 days. In this training, each agent
experienced the equivalent of  200 years of real-time StarCraft play.

In StarCraft  players can choose to play one  of three alien
 races: Terran, Zerg or Protoss.  AlphaStar was trained to  play
 Protoss only, to reduce training time, although the same training
 pipeline could be applied to any race.
AlphaStar was first  tested
against a human grandmaster named TLO, a top
professional Zerg player and a grandmaster level Protoss
player. 
The human player remarked: \emph{I was surprised by how
strong the agent was. AlphaStar takes well-known strategies and turns
them on their 
head. The agent demonstrated strategies I had not thought of before,
which means there may still be new ways of playing the game that we
haven’t fully explored yet.}

\subsection{\em Hands On: Hide and Seek in the Gym Example}

Many of the research efforts reported in these final chapters 
describe significant efforts by  research teams working on 
complicated and large games. These games represent
the frontier of artificial intelligence, and research teams use all
available computational and software engineering power that they
can acquire to achieve the best results.  Also, typically a
large amount of time is spent in training, and in finding the right
hyperparameters for the learning to work.

Replicating results of this scale  is highly challenging, and most research
efforts are focused on replicating the results on a smaller, more
manageable  scale, with   more manageable computational resources.

In this section we will try to replicate some aspects with  modest
computational requirements. We will focus on Hide and Seek
(Sect.~\ref{sec:hide}).
The original code for the Hide and Seek experiments is on GitHub.\footnote{
\url{https://github.com/openai/multi-agent-emergence-environments}}
Please visit and install the code. Hide and Seek uses MuJoCo and the
mujoco-worldgen package. Install them and the dependencies with
\begin{tcolorbox}
\verb|pip install -r mujoco-worldgen/requirements.txt|
\verb|pip install -e mujoco-worldgen/|\\ 
\verb|pip install -e multi-agent-emergence-environments/|
\end{tcolorbox}

Examples of environments can be found in the \verb|mae_envs/envs| folder.
You can also build your own environments, starting from the Base
environment, in \verb|mae_envs/envs/base.py|, and then adding boxes,
ramps, as well as the appropriate wrappers. Look in the other
environments for how to do this.

Try out  environments by using the \verb|bin/examine| script; example
usage:
\begin{tcolorbox}
  \verb|bin/examine.py base|
\end{tcolorbox}
See further the instructions in the
GitHub repository.  



\subsubsection*{Multiplayer Environments}
\index{Ape-X DQN}\index{Petting Zoo}\index{Google Research Football}\index{IMPALA}

To conclude this hands-on section, we mention  a multiplayer implementation of the
Arcade Learning Environment. It is presented by Terry et al.~\cite{terry2020multiplayer},
who also present baseline performance results for a multiplayer
version of DQN, Ape-X DQN, which performed well
elsewhere~\cite{horgan2018distributed,bard2020hanabi}.
The environment is also presented as part of PettingZoo, a multi-agent
version of Gym~\cite{terry2020pettingzoo,tylkin2021learning,klijn2021coevolutionairy}.

Another multi-agent research environment is the Google Football
Research Environment~\cite{kurach2020google}. A physics-based
simulator is provided, as well as three baseline implementations (DQN,
IMPALA, and PPO).

\section*{Summary and Further Reading}
\addcontentsline{toc}{section}{\protect\numberline{}Summary and Further Reading}
We will now summarize the chapter and provide pointers to further reading.
\subsection*{Summary}
Multi-agent reinforcement learning learns optimal policies for
environments that consist of multiple agents.  The optimal policy of
the agents is
influenced by the policies of the other agents, whose policy is also being
optimized. This gives rise to the problem of nonstationarity, and
agent behavior violates the  Markov property. 

Multi-agent reinforcement learning adds the element of cooperative behavior to
the repertoire of reinforcement learning, which now consists of
competition, cooperation, and mixed behavior.  The field is
closely related to game theory---the basis of the study of rational
behavior in economics. A famous problem of game theory is the
prisoner's dilemma. A famous result of non-cooperative game theory is
the Nash equilibrium, which is defined as the joint strategy where no
player has anything to gain by changing their own strategy. A famous
result from cooperative game-theory is the Pareto optimum, the
situation where no individual  can be better off without making
someone else worse off.

When agents have private information, a multi-agent problem is
partially observable. Multi-agent problems  can be modeled by
stochastic games or as extensive form games. 
%
The behavior of agents is ultimately
determined by the reward functions, that can be homogeneous, or
heterogeneous. When agents have different reward functions, multi-agent
becomes multi-objective reinforcement learning.

The regret of an action is the amount of reward that is missed by the
agent for not choosing the actions with the highest payoff. A regret
minimization algorithm is the stochastic and multi-agent equivalent of 
minimax. Counterfactual-regret minimization is an approach for finding
Nash strategies in competitive multi-agent games, such as poker.

Variants of single-agent algorithms are often used for cooperative
multi-agent situations. 
The large state space due to nonstationarity and
partial observability precludes solving large problems. Other promising
approaches are opponent modeling and explicit communication modeling.

Population-based methods such as evolutionary algorithms and swarm
intelligence are used frequently in multi-agent systems. These
approaches are suitable for homogeneous reward 
functions and competitive, cooperative, and mixed
problems. Evolutionary methods evolve a population of agents,
combining behaviors, and selecting the best according to some fitness
function. Evolutionary methods are a natural fit for parallel
computers and are among the most popular and successful optimization
algorithms. Swarm intelligence often introduces (rudimentary) forms of
communication between agents, such as in Ant colony optimization 
where agents communicate through artificial pheromones to indicate
which part of the solution space they have traveled.

For some of the most complicated problems that have recently been
tackled, such as StarCraft, Capture the Flag, and Hide and Seek,
hierarchical and evolutionary principles are  often combined in league
training, where leagues of teams of agents are trained in a self-play
fashion, and where the fittest agents survive.
Current achievements require large amounts of computational power,
future work is trying to reduce these requirements.

\subsection*{Further Reading}
Multi-agent learning is a widely studied field. Surveys---both on early
multi-agent reinforcement learning and on deep multi-agent
reinforcement learning---can be found
in~\cite{gronauer2021multi,busoniu2008comprehensive,yang2020overview,albrecht2017multiagent,albrecht2018autonomous,tampuu2017multiagent,tuyls2012multiagent,hernandez2019survey,hernandez2017survey,wong2021survey}.
After Littman~\cite{littman1994markov},
Shoham et
al.~\cite{shoham2003multi} look deeper into MDP modeling.

The classic work on game theory is Von Neumann and
Morgenstern~\cite{von1944theory}. Modern introductions
are~\cite{myerson2013game,davis2012game,ganzfried2011game}. 
Game theory underlies much of the theory of rational behavior in
classical economics. Seminal works of John Nash
are~\cite{nash1950equilibrium,nash1950bargaining,nash1951non}. In 1950
he introduced the Nash equilibrium in his dissertation of 28 pages,
which won him the Nobel prize in Economics in 1994. A 
biography and film have been made about the  life of John
Nash~\cite{nasar2011beautiful}.

The game of rock-paper-scissors plays an important role in game
theory, and the  study of computer poker~\cite{walker2004official,billings2002challenge,rubin2011computer,brown2019superhuman}.
Prospect
theory~\cite{kahneman2013prospect}, introduced in 1979, studies 
human behavior in the face of uncertainty, a topic that evolved into
the field of behavioral
economics~\cite{wilkinson2017introduction,mullainathan2000behavioral,cartwright2018behavioral}. Gigerenzer  
introduced fast and frugal heuristics to explain human decision
making~\cite{gigerenzer1996reasoning}.

For more intriguing works on the field of  evolution of
cooperation and the emergence of social norms, see, for
example~\cite{boyd1988culture,axelrod1981evolution,axelrod1997complexity,axelrod1986evolutionary,axelrod1997dissemination,henrich2008five,heylighen1998makes}.

More recently multi-\emph{objective} reinforcement learning has been
studied, a survey is~\cite{liu2014multiobjective}. In this field the
more realistic assumption is adopted that agents have different
rewards functions, leading to different Pareto
optima~\cite{van2014multi,mossalam2016multi,wiering2014model}.
Oliehoek et al.\ have written a concise introduction to decentralized
multi-agent modelling~\cite{oliehoek2016concise,oliehoek2012decentralized}.

Counterfactual regret minimization has been fundamental for the
success in computer
poker~\cite{brown2018superhuman,brown2019superhuman,johanson2012efficient,zinkevich2008regret}. An
often-used Monte Carlo version is published
in~\cite{lanctot2009monte}. A combination
with function approximation is studied in~\cite{brown2019deep}.

Evolutionary algorithms have delivered highly successful optimization
algorithms. Some entries to this vast field are~\cite{eiben2015evolutionary,back1996evolutionary,back1991survey,back1997handbook,back1993overview}.
A related field is swarm intelligence, where 
communication between homogeneous agents is taking
place~\cite{kennedy2006swarm,eberhart2001swarm,dorigo2007swarm,beni2020swarm}. 
For further research in multi-agent systems refer
to~\cite{wooldridge2009introduction,van2008multi}.  For collective
intelligence, see, for example~\cite{kennedy2006swarm,dorigo2006ant,gallese1998mirror,woolley2010evidence,plaat2010vlinder}.

Many other works report on evolutionary algorithms in
a reinforcement learning
setting~\cite{such2017deep,khadka2018evolutionary,moriarty1999evolutionary,whiteson2012evolutionary,chrabaszcz2018back,conti2017improving,salimans2017evolution,wierstra2008natural}. Most
of these approaches concern single agent approaches, although some are
specifically applied to multi agent approaches~\cite{khadka2019evolutionary,khadka2019collaborative,singh2010intrinsically,liu2019emergent,jaderberg2017population,klijn2021coevolutionairy}.

Research into  benchmarks is active. Among  interesting approaches
are Procedural content generation~\cite{togelius2013procedural}, 
MuJoCo Soccer~\cite{liu2019emergent}, and the Obstacle Tower
Challenge~\cite{juliani2019obstacle}. 
There is an extensive literature on computer poker. See, for example~\cite{billings2002challenge,billings2004game,gilpin2006competitive,rubin2011computer,bard2013annual,bowling2015heads,billings2004game,gilpin2006competitive,bowling2009demonstration,sandholm2010state,moravvcik2017deepstack}.
StarCraft research can be found
in~\cite{vinyals2017starcraft,samvelyan2019starcraft,vinyals2019grandmaster,samvelyan2019starcraft,ontanon2013survey}. Other games
studies  are~\cite{tekofsky2015past,toubman2014dynamic}. Approaches
inspired by results in poker and Go are now also being applied with
success in
no-press Diplomacy~\cite{paquette2019no,anthony2020learning}.

\section*{Exercises}
\addcontentsline{toc}{section}{\protect\numberline{}Exercises}
Below are a few quick questions to check your understanding of this
chapter. For each question a simple, single sentence answer is sufficient.

\subsubsection*{Questions}
\begin{enumerate}
\item Why is there so much interest in multi-agent reinforcement learning?
\item What is one of the main challenges of multi-agent reinforcement
  learning?
\item What is a Nash strategy?
\item What is a Pareto Optimum?  
\item In a competitive multi-agent system, what algorithm can be used
  to calculate a Nash   strategy?
\item What makes it difficult to calculate the solution for a game
  of imperfect information?
  \item Describe the Prisoner's dilemma.
  \item Describe the iterated Prisoner's dilemma.
\item Name two multi-agent card games of imperfect information.
\item What is the setting with a heterogeneous reward function
  usually called?
\item Name three kinds of strategies that can occur a multi-agent
  reinforcement learning.
\item Name two solution methods that are appropriate for solving mixed
  strategy games.
\item What AI method is named after ant colonies, bee swarms, bird
  flocks, or fish schools? How does it work in general terms?
\item Describe the main steps of an evolutionary algorithm.
\item Describe the general form of Hide and Seek and three strategies
  that emerged from the interactions of the hiders or seekers.
\end{enumerate}

\subsubsection*{Exercises}
Here are  some programming exercises  to become more familiar with the methods that
we have covered in this chapter. 

\begin{enumerate}
\item \emph{CFR}
  Implement counterfactual regret minimization for a Kuhn poker
  player. Play against the program, and see if you can win. Do you see
  possibilities to extend it to a more challenging version of poker?

\item \emph{Hide and Seek}
  Implement  Hide and Seek with cooperation and   competition. Add
  more types of objects. See if other cooperation behavior emerges.

\item \emph{Ant Colony} Use the DeepMind Control Suite and setup a collaborative level and
  a competitive level, and implement Ant Colony Optimization. Find
  problem instances on the web, or in the original
  paper~\cite{dorigo2006ant}. Can you implement more swarm algorithms?

  \item \emph{Football} Go to the Google Football
    blog\footnote{\url{https://ai.googleblog.com/2019/06/introducing-google-research-football.html}}
    and implement algorithms for football agents.\index{Google
      Football} Consider using a population-based approach.

\item \emph{StarCraft} Go to the StarCraft Python
  interface,\footnote{\url{https://github.com/deepmind/pysc2}} and
  implement a StarCraft player (highly challenging)~\cite{samvelyan2019starcraft}.
    
\end{enumerate}


\chapter{Hierarchical Reinforcement
  Learning}\label{chap:hier}\index{hierarchical reinforcement learning}
The goal of artificial intelligence is to understand and create intelligent 
behavior; the goal of deep reinforcement learning is to find a behavior
policy for ever larger  sequential decision 
problems.

But how does real 
intelligence  find these policies? One of the things that
humans are good at, is dividing a complex task into simpler subproblems,
and then solving those tasks, one by one, and combining them as the
solution to the larger problem. These subtasks are of 
different scales, or granularity, than the original problem. For example, when planning a trip
from your house to a hotel room in a far-away city, you typically only plan the start
and the end in terms of footsteps taken in a certain direction. The
in-between part may contain different modes of transportation that get
you to your destination quicker, with macro steps, such as taking a trainride or
a flight. During this macro, you  do not try out
footsteps in different directions. Our trip---our policy---is a combination of fine-grain
primitive actions and  coarse-grain macro actions. 

Hierarchical reinforcement learning studies this  
real-world-inspired approach to problem solving. It provides formalisms and
algorithms to  divide    problems 
into larger subproblems, and then  plans with these  
subpolicies, as if they were subroutines.

In principle the hierarchical approach can exploit structure in all sequential decision
problems, although  some problems are easier than others.  Some  environments can 
be subdivided into smaller problems in a natural way, such as
navigational tasks on maps, or path-finding tasks in
mazes. Multi-agent problems also naturally divide into hierarchical
teams, and have large state spaces where hierarchical methods may
help. For other
problems, however, it can be hard to find efficient macros, or it can 
be computationally intensive to find good combinations of macro steps
and primitive steps.

Another aspect of hierarchical methods is that since macro-actions take large 
steps, they may miss the global minimum.  The  best policies found
by hierarchical methods may be less optimal than those found by
``flat'' approaches (although they may get there much quicker).

In this chapter we will start with an example to capture the flavor of
hierarchical problem solving. Next, we will look at a theoretical
framework that is used to model hierarchical algorithms, and at a few
examples of algorithms. Finally, we will
look deeper at  hierarchical environments.

The chapter ends with  exercises, a summary, and pointers
to further reading.

\section*{Core Concepts}
\begin{itemize}
\item Solve large, structured, problems by divide and conquer
\item Temporal abstraction of actions with options
\end{itemize}

\section*{Core Problem}
\begin{itemize}
\item Find  subgoals and subpolicies efficiently, to perform
  hierarchical abstraction
\end{itemize}

\section*{Core Algorithms}
\begin{itemize}
\item Options framework (Sect.~\ref{sec:options})
\item Option critic (Sect.~\ref{sec:optioncritic})
\item Hierarchical actor critic (Sect.~\ref{sec:hac})
\end{itemize}

\section*{Planning a Trip}
Let us see how we  plan a major trip to visit
a friend that lives in another city, with a hierarchical method. The
method would break up the trip in 
different parts. The first part would be to walk to your closet and
get your things, and then to get out and  get your bike.  You would go to the
train station, and park your bike. You would then take the train to
the other city, possibly changing trains 
enroute if that would be necessary to get you there faster. Arriving
in the city your friend would meet you at the station and would drive
you to their house.

A ``flat'' reinforcement learning method would have at its disposal
actions consisting of footsteps in certain directions. This would make
for a large  space of possible policies, although the fine
grain at which the policy would be planned---individual
footsteps---would sure be able to find the optimal shortest route. 

The hierarchical method has at its disposal a wider variety of
actions---macro actions: it can plan a bike ride, a  train trip, and getting a ride by
your friend. The route may not be the shortest possible (who knows if
the train follows the shortest route between the two cities) but
planning will be much faster than painstakingly optimizing footstep by
footstep.

\section{Granularity of the Structure of Problems}\index{temporal abstraction}
In hierarchical reinforcement learning the granularity of 
abstractions is larger than the fine grain of the primitive
actions of the environment. When we are preparing a meal, we reason in
large action chunks: chop onion, cook spaghetti, instead of reasoning
about the actuation of individual muscles in our hands and arms. Infants
learn to use their muscles to performs certain tasks as they grow up until it becomes second nature.

We generate subgoals that act as temporal  abstractions, and
subpolicies that are  macro's of multiple ordinary
actions~\cite{schmidhuber1991learning}. Temporal abstraction allows us
to reason about actions of different time scales, sometimes with course grain
actions---taking a train---sometimes with fine grain actions---opening
a door---mixing macro actions with primitive actions. 
%
%
%

Let us look at  advantages and disadvantages of the hierarchical  approach.

\subsection{Advantages}\label{sec:hrl-adv}
We will start with the advantages of
hierarchical methods~\cite{flet2019promise}. 
First of all, hierarchical reinforcement learning simplifies problems
through \emph{abstraction}. Problems are abstracted into a higher level of
aggregation.
Agents create subgoals and solve fine grain subtasks first. Actions are abstracted into
larger macro actions to solve these subgoals; agents use temporal
abstraction.

Second, the temporal abstractions increase  \emph{sample
efficiency}. The number of interactions with the environment is
reduced because subpolicies are learned to solve subtasks, reducing
the environment interactions. Since subtasks are learned, they can
be transfered to other problems, supporting transfer learning.

Third,  subtasks reduce  brittleness due to overspecialization
of policies. Policies become more \emph{general}, and are able to adapt to
changes in the environment more easily.

Fourth, and    most importantly, the higher level of
abstraction allows  agents  to solve
\emph{larger}, more complex problems. This is a reason why
for  complex  multi-agent games  such as StarCraft, where teams
of agents must be managed, hierarchical approaches are used.

Multi-agent reinforcement learning often exhibits a hierarchical
structure;  problems  can
be organized such that each agent is assigned  its own
subproblem,  or  the agents
themselves may be structured or organized 
in teams or groups.  There can be
cooperation within the teams or competition between the teams, or the
behavior can be  fully cooperative or fully competitive. 

Flet-Berliac~\cite{flet2019promise}, in a recent
overview,  
summarizes the promise of hierarchical reinforcement learning as
follows: (1) achieve long-term 
credit assignment through faster learning and better generalization,
(2)  allow structured exploration by exploring with sub-policies
rather than with primitive actions, and (3) perform transfer learning
because different levels of hierarchy can encompass different
knowledge.

\subsection{Disadvantages}\label{sec:hrl-dis}
There are also disadvantages and challenges associated with hierarchical
reinforcement learning. First of all, it works better when there is  \emph{domain knowledge} available about
structure in the domain.  Many hierarchical methods assume that domain
knowledge is available  to subdivide
 the environment so that hierarchical reinforcement learning can be
applied. 

Second, there is  \emph{algorithmic complexity} to be solved. Subgoals
must be identified in the problem environment, subpolicies must be
learned, and termination conditions are needed. These algorithms must
be designed, which costs programmer effort.

Third, hierarchical approaches
introduce a new type of actions, macro-actions. Macros are
combinations of primitive actions, and their use can greatly improve
the performance of the policy. On the other hand,  the number of
possible combinations of actions 
is exponentially large in their
length~\cite{backstrom1995planning}. For larger problems enumeration
of all 
possible macros is out of the question, and the overall-policy function
has to be approximated. 
Furthermore, 
at each decision point in a hierarchical planning or learning
algorithm we now have the option to consider 
if any
of the macro actions improves the current policy. The
\emph{computational complexity} of the
planning and learning choices   increases by the introduction of
the macro actions~\cite{backstrom1995planning}, and approximation
methods must be used. The efficiency  gains of the
hierarchical behavioral policy must outweigh the higher cost of finding this
policy. 

Fourth, the \emph{quality} of a behavioral policy that
includes macro-actions may be less than that of a policy consisting
only of primitive actions. The macro-actions may skip over possible
shorter routes, that the primitive actions  would have found.

\subsubsection*{Conclusion}
There are advantages and disadvantages to hierarchical reinforcement learning.
Whether an efficient policy can be constructed
and whether its accuracy is good enough depends on the problem at
hand, and also on the quality of the algorithms that are used to find this policy.

For a long time,  finding good
subgoals has been a major challenge. With recent algorithmic advances,
especially in function approximation,
important progress has been made. We will discuss these advances in
the next section.

\section{Divide and Conquer for Agents}\label{sec:hier}

To discuss hierarchical reinforcement learning, first we will discuss
a  model, the options framework, that formalizes the concepts of
subgoals and subpolicies. Next, we will describe the main 
challenge of hierarchical reinforcement learning, which is sample
efficiency. Then we will discuss the main part of this chapter: algorithms for finding subgoals and
subpolicies, and finally we will provide an overview of algorithms
that have been developed in the field.

\subsection{The Options Framework}
\label{sec:options}

A hierarchical reinforcement learning algorithm tries to solve
sequential decision problems more efficiently by identifying common
substructures and re-using subpolicies to solve them. The hierarchical
approach has three challenges~\cite{rafati2019learning,kulkarni2016hierarchical}:
find  subgoals,
find a meta-policy over these subgoals, and
find  subpolicies for these subgoals.
%

Normally, in reinforcement learning, the agent follows in each state the
action that is indicated by the policy. In 1999, Sutton, Precup and
Singh~\cite{sutton1999between} introduced the options framework. This
framework introduces  formal constructs with which subgoals and
subpolicies can be incorporated elegantly into the reinforcement learning
setting. 
The idea of options is simple. Whenever a
 state is reached that is a subgoal, then, in addition to following a
primitive action suggested by the main policy, the option can be taken. This means
that not the main action policy is followed, but the option policy, a macro
action consisting of a different subpolicy specially aimed at satisfying
the subgoal in one large step. In this way macros are incorporated into the
reinforcement learning framework.

We have been using the terms \emph{macro} and \emph{option}
somewhat loosely until now; there is, however,
a difference between macros and options. 
A macro is any group of actions, possibly open-ended. An option is a group of actions with
a termination condition. Options take in environment observations and
output actions until a termination condition is met.

\begin{figure}[t]
\begin{center}
\includegraphics[width=4cm]{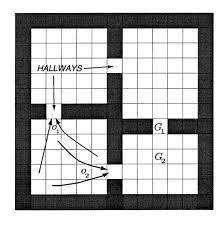}
\caption{Multi-Room Grid~\cite{sutton1999between}}\label{fig:rooms}
\end{center}
\end{figure}

Formally, an option \gls{omega-hier}  has three
elements~\cite{alver2018ml}. Each option $\omega=\langle I,\pi,
\beta\rangle$ has the following triple:
\begin{itemize}
\item[\gls{initiation}] \qquad\quad The initiation set
  $I\subseteq S$ are the states that the
  option can start from
\item[$\pi_\omega(a|s)$]  \qquad\quad The
  subpolicy $\pi : S\times A\rightarrow [0,1] $ internal to this particular
  option
\item[\gls{termination}]  \qquad\quad The
  termination condition $\beta : S\rightarrow [0,1] $  tells us if
   $\omega$ terminates in  $s$
\end{itemize}
The set of all options is denoted as \gls{options}. In the options framework,
there are thus two types of policies: the
(meta-)policy \emph{over} options $\pi_\Omega(\omega|s)$ and the subpolicies
$\pi_\omega(a|s)$. The subpolicies $\pi_\omega$  are short 
macros to get from $I_\omega$ to $\beta_\omega$ quickly, using the
previously learned macro (subpolicy). Temporal abstractions mix
actions of different granularity, short and long, primitive action and
subpolicy. They allow traveling from $I$ to $\beta$ without additional
learning,  using a previously provided or learned subpolicy.  

One of the problems for which the options framework works well, is
room navigation in a Grid world (Fig.~\ref{fig:rooms}). In a regular reinforcement
learning problem the agent would learn
to move step by step. In hierarchical reinforcement learning the doors
between rooms are bottleneck states, and are natural  subgoals.  Macro
actions (subpolicies) are to move to a door in one multi-step
action (without considering alternative actions along the way). Then we can  go to a
different room, if we choose the appropriate option, using another
macro, closer to where the main goal is located. The four-room 
problem from the figure is  used in many research works in hierarchical
reinforcement learning.

\subsubsection*{Universal Value Function}\index{universal value function}

In the original options framework the process of identifying the
subgoals (the 
hallways, doors) is external. The subgoals have to be provided
manually, or by other
methods~\cite{precup1997planning,hauskrecht2013hierarchical,laird1986chunking,stolle2002learning}.  
Subsequently, methods have been published to learn these subgoals. 

Options are goal-conditioned subpolicies.
More recently a generalization to parameterized options has been
presented in the universal value function,  by
Schaul et al.~\cite{schaul2015universal}. Universal value functions
provide a unified theory for goal-conditioned parameterized value
approximators $V(s,g,\theta)$.

\subsection{Finding Subgoals}
\label{sec:challenge}

Whether the hierarchical method improves over a traditional flat
method depends on a number of factors. First, there should be enough
repeating structure in the domain to be exploited (are there  many rooms?),
second,  the algorithm must find  appropriate subgoals (can it find
the doors?), third,  the options that are found must repeat many times
(is the puzzle played frequently enough for the option-finding cost
to be offset?), and, finally,  subpolicies must be found that give  enough
improvement (are the rooms large enough that options outweigh
actions?).

The original options framework assumes that the structure of the domain is
obvious, and that the subgoals are  given. When this is
not the case, then the subgoals must be found by the algorithm.
Let us look at an overview of  approaches, both tabular and with deep
function approximation.

\subsection{Overview of Hierarchical Algorithms}
\label{sec:overview}

The options framework provides a convenient formalism for temporal
abstraction.
In addition to
the algorithms that can  construct 
policies consisting of individual actions, we need algorithms that find the subgoals, and learn the
subpolicies.
%
Finding efficient algorithms for the three tasks is important in order to
be able to achieve an efficiency advantage over ordinary ``flat''
reinforcement learning.

\begin{table}[t]
  \begin{center}
  \begin{tabular}{lllccl}
     &   &  & {\bf Find} & {\bf Find} &  \\
        {\bf Name} & {\bf Agent}  & {\bf Environment} & {\bf
                                                        Subg}&{\bf Subpol} & {\bf Ref}  \\
    \hline\hline
    STRIPS & Macro-actions & STRIPS planner & - &-&\cite{fikes1972learning}\\
    Abstraction Hier. & State abstraction & Scheduling/plan.&
                         +&+& \cite{knoblock1990learning}\\
    HAM & Abstract machines & MDP/maze  &-&-& \cite{parr1998reinforcement}\\
    MAXQ & Value function decomposition& Taxi &-&-& \cite{dietterich2000hierarchical} \\
    HTN & Task networks & Block world &-&-&
                                            \cite{currie1991plan,ghallab2004automated}\\
    Bottleneck & Randomized search & Four room &+&+&
                                            \cite{stolle2002learning}\\
    \hline
    Feudal & manager/worker, RNN & Atari &+&+& \cite{vezhnevets2017feudal,dayan1993feudal}\\
    Self p. goal emb. & self play subgoal & Mazebase,
                                                    AntG&+&+ &
                                                                \cite{sukhbaatar2018learning}\\
    Deep Skill Netw. & deep skill array, policy distillation &
                                                                  Minecraft
                                                     &+&+&
                                                       \cite{tessler2017deep}\\
    STRAW & end-to-end implicit plans & Atari &+&+&\cite{vezhnevets2016strategic}\\
    HIRO & off-policy & Ant maze &+&+& \cite{nachum2018data}\\
    Option-critic & policy-gradient & Four room &+&+& \cite{bacon2017option}\\
    HAC & actor critic, hindsight exper. repl.& Four room ant 
                                                     &+&+&
                                                       \cite{levy2019learning,andrychowicz2017hindsight}\\
    Modul. pol. hier. & bit-vector, intrinsic motivation & FetchPush
                                                     &+&+&
                                                       \cite{pashevich2018modulated}\\                                                    
    h-DQN & intrinsic motivation & Montezuma's R. &-&+& \cite{kulkarni2016hierarchical}\\
    Meta l. sh. hier. & shared primitives, strength
                                       metric & Walk, crawl &+&+& \cite{frans2017meta} \\
    CSRL & model-based transition dynamics & Robot tasks &-&+& \cite{li2017efficient}\\
    Learning Repr. & unsup. subg. disc.,
                               intrinsic motiv. & Montezuma's R. &+&+&
                                                                       \cite{rafati2019learning}\\
    AMIGo&Adversarially intrinsic goals& MiniGrid PCG&+&+&\cite{campero2020learning}\\
    \hline
  \end{tabular}
  \caption[Hierarchical Reinforcement Learning Approaches]{Hierarchical Reinforcement Learning Approaches (Tabular and Deep)}\label{tab:hrl}
\end{center}
\end{table}

Hierarchical reinforcement learning is based on subgoals. It implements a
top-level policy over these subgoals, and subpolicies to solve the subgoals. The
landscape of subgoals determines to a great extent the efficiency of
the algorithm~\cite{dwiel2019hierarchical}.  
%
%
%
In recent years  new
algorithms have been developed to find 
sub-policies for options, and the field has received a renewed
interest~\cite{pateria2021hierarchical}.  Table~\ref{tab:hrl} shows a list of approaches.   The table
starts with  classic tabular
approaches (above the line). It
continues with  
more recent deep learning approaches. 
%
%
We will now look at some of the algorithms. 

\subsubsection{Tabular Methods}

Divide and conquer is a natural method to exploit hierarchical problem
structures. 
A famous early planning system is STRIPS, the Stanford Research
Insititute Problem Solver, designed by Richard Fikes and Nils Nilsson
in the 1970s~\cite{fikes1972learning}. STRIPS created an extensive
language for expressing planning problems, and was quite
influential. Concepts from STRIPS are at the basis of most modern 
planning systems, action languages, and  knowledge
representation systems~\cite{gelfond1998action,baral2003knowledge,van2008handbook}.
The concept of macros as open-ended groups of actions was used in
STRIPS to create higher-level primitives, or subroutines. 

Later planning-based approaches are Parr and Russell's hierarchical abstract
machines~\cite{parr1998reinforcement} and Dietterich's
MAXQ~\cite{dietterich2000hierarchical}. Typical applications of these
systems are the blocks world, in which a robot arm has to manipulate
blocks, stacking them on top of each other, and the taxi world, which
we have seen in earlier chapters. An overview of these and other
early  approaches can be found in  Barto et
al.~\cite{barto2003recent}.

Many of these early approaches focused on macros (the subpolicies), and
require  that the experimenters 
identify the subgoals in a planning
language. For problems where no such obvious subgoals are available, Knoblock~\cite{knoblock1990learning} showed how abstraction
hierarchies can be generated, although Backstrom et
al.~\cite{backstrom1995planning} found that doing so can be
exponentially less efficient.   For small room problems, however,
 Stolle and
Precup~\cite{stolle2002learning} showed that subgoals can be found in
a more efficient way,  using a short randomized search to
find bottleneck states that can be used as subgoals.
This approach finds subgoals 
automatically, and efficiently, in a rooms-grid world.

Tabular hierarchical methods were mostly applied to small and
low-dimensional problems, and have difficulty finding subgoals,
especially for large problems. The advent of deep function
approximation methods attracted renewed  interest in hierarchical methods.


\subsubsection{Deep Learning}

Function approximation can potentially reduce the problem of
exponentially exploding search spaces that plague tabular methods,
especially for subgoal discovery. 
Deep learning  exploits similarities between states using
commonalities between features, and allows larger problems to be
solved. Many new methods were developed.
The deep learning approaches in hierarchical reinforcement learning
typically are end-to-end: they generate both appropriate subgoals and
their policies. 

Feudal networks is an older idea from Dayan and Hinton in which an
explicit control hierarchy is built of managers and workers that work
on tasks and subtasks, organized as in a feudal
fiefdom~\cite{dayan1993feudal}. This idea was used 15 years later
as a model for hierarchical deep reinforcement
learning by Vezhnevets et al.~\cite{vezhnevets2017feudal}, out-performing
non-hierarchical A3C on Montezuma's Revenge, and performing well on
other Atari games, achieving a similar score as
Option-critic~\cite{bacon2017option}. The approach uses a manager that
sets abstract goals (in latent space) for workers.
The feudal idea was
also used as inspiration for a 
multi-agent cooperative 
reinforcement learning design~\cite{ahilan2019feudal},   on proof of
concept cooperative multi-agent problems on 
pre-specified hierarchies.

Other deep learning approaches include deep skill
networks~\cite{tessler2017deep}, off-policy 
approaches~\cite{nachum2018data}, and self-play~\cite{sukhbaatar2018learning}.  The latter uses an intrinsic
motivation approach to learn both a low level actor and the
representation of the state
space~\cite{pere2018unsupervised}. Subgoals are learned at the higher
level, after which policies are trained at the lower level. Application environments for deep
learning have become
more 
challenging, and now include Minecraft, and robotic tasks such as ant navigation
in multiple rooms, and maze navigation. The approaches outperform
basic non-hierarchical approaches such as DQN.

In STRAW Vezhnevets et
al.~\cite{vezhnevets2016strategic} learns a model of  
macros-actions, and is evaluated on text recognition tasks and on 
Atari games such as PacMan and Frostbite, showing promising results.
Zhang et al.~\cite{zhang2021world} use world models to learn latent
landmarks (subgoals) for graph-based planning (see also Sect.~\ref{sec:latent}).



\label{sec:optioncritic}
Almost two decades after the options framework was introduced,  Bacon et
al.~\cite{bacon2017option} introduced  the
option-critic approch. Option-critic extends the options
framework with  methods to learn the option subgoal and subpolicy, so that it does
not have to be provided externally anymore. The options are learned
similar to 
actor critic using a gradient-based approach.  The
intra-option policies and termination functions, as 
well as the policy over options are learned simultaneously.   The user
of the Option-critic  
approach has to specify how many options have to be learned. The
Option-critic paper reports good results for experiments in 
a four-room environment  with 4 and with
8 options (Fig.~\ref{fig:4rooms}).
\begin{figure}[t]
\begin{center}
\includegraphics[width=7cm]{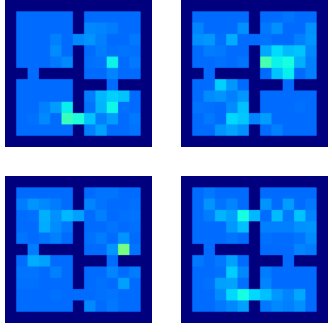}
\caption[Option-Critic]{Termination Probabilities Learned with 4
  Options by Option-Critic~\cite{bacon2017option};  Options Tend to
Favor Squares Close to Doors}\label{fig:4rooms}
\end{center}
\end{figure}
Option critic  learns options in an end-to-end fashion
that scales to  larger domains, outperforming DQN in four ALE games
(Asterix, Seaquest, Ms.\ Pacman, Zaxxon)~\cite{bacon2017option}. 

\label{sec:hac}
Levy et al.~\cite{levy2019learning}
presented an  approach based on Option critic,  called Hierarchical
actor critic. This approach  can learn the goal-conditioned policies at different levels
concurrently, where previous approaches  had to
learn them in a bottom up fashion. 
In addition, Hierarchical actor critic uses a method for learning the multiple levels of
policies for sparse rewards, using Hindsight
experience replay~\cite{andrychowicz2017hindsight}. In typical robotics
tasks, the reinforcement learning algorithm learns more from a
successful outcome (bat hits the ball) than from an unsuccessful
outcome (bat misses the ball, low). In this failure case a human
learner would  draw the conclusion  that we can now reach another
goal, being bat misses the ball if we aim low. Hindsight
experience replay allows learning to take place by incorporating such
adjustments of the goal using the benefit of hindsight, so that the
algorithm can 
now also learn from failures, by pretending that they were the goal that
you wanted to reach, and learn from them as if they were.

Hierarchical actor critic has been  evaluated  on  grid world tasks
and more complex 
simulated robotics environments, using a 3-level hierarchy.

\index{AMIGo}
A final approach that we mention is AMIGo~\cite{campero2020learning},
which is related to intrinsic motivation. It uses a teacher to
adversarially generate goals for a student. The student is trained
with increasingly challenging goals to learn general skills. The
system effectively builds up an automatic curriculum of goals. It is
evaluated on MiniGrid, a parameterized world that is generated by procedural content
generation~\cite{raileanu2020ride,chevalierminimalistic}. 

\subsubsection*{Conclusion}
Looking back at the list of advantages and disadvantages at the start
of this chapter, we  see a range of interesting and creative ideas that
achieve the advantages (Sect.~\ref{sec:hrl-adv}) by providing
methods to address the
disadvantages (Sect.~\ref{sec:hrl-dis}). In general, the tabular
methods are restricted to smaller problems, and often need to be
provided with subgoals. Most of the newer deep learning methods find
subgoals by themselves, for which then subpolicies are found. Many
promising methods have been discussed, and most report to outperform
one or more 
flat baseline algorithms. 

The promising results stimulate further research in deep hierarchical
methods, and more
benchmark studies of large problems are needed. Let us have a closer look at the
environments that have been used so far.

\section{Hierarchical Environments}
Many  environments for hierarchical reinforcement
learning exist, starting with mazes and the four-room environment
from the options paper. Environments have evolved with the rest of the
field of reinforcement learning;  for hierarchical reinforcement
learning no clear favorite benchmark has emerged,
although Atari en MuJoCo tasks are often used. In the following we  will review some of
the environments that are used in   algorithmic 
studies. Most hierarchical environments are smaller than typically
used for model-free flat reinforcement learning, although some
studies do use  complex environments, such as StarCraft.

\subsection{Four Rooms and Robot Tasks}
Sutton et al.~\cite{sutton1999between}  presented the four rooms problems to
illustrate how the options model worked (Fig.~\ref{fig:4rooms8};  left panel). This
environment has been used frequently in subsequent papers on
reinforcement learning.
The rooms are connected by  hallways. Options point the way to these
hallways, which lead to the goal $G_2$ of the environment. A hierarchical algorithm should
identify the hallways as the subgoals, and create subpolicies for each
room to go to the hallway subgoal (Fig.~\ref{fig:4rooms8}; right panel).

\begin{figure}[t]
  \begin{center}
    \begin{tabular}{cc}
      \includegraphics[width=4cm]{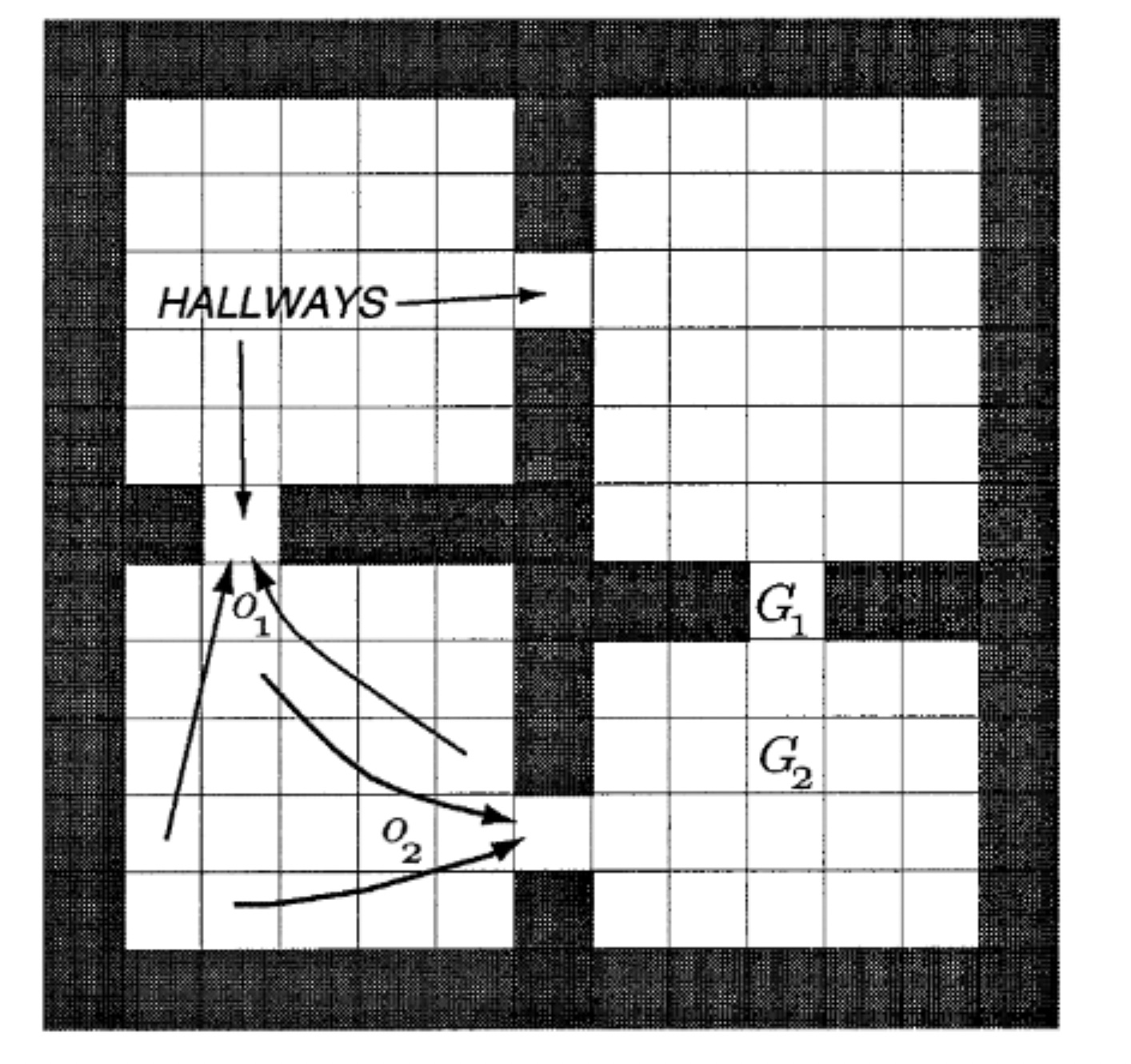}
      &
        \includegraphics[width=5cm]{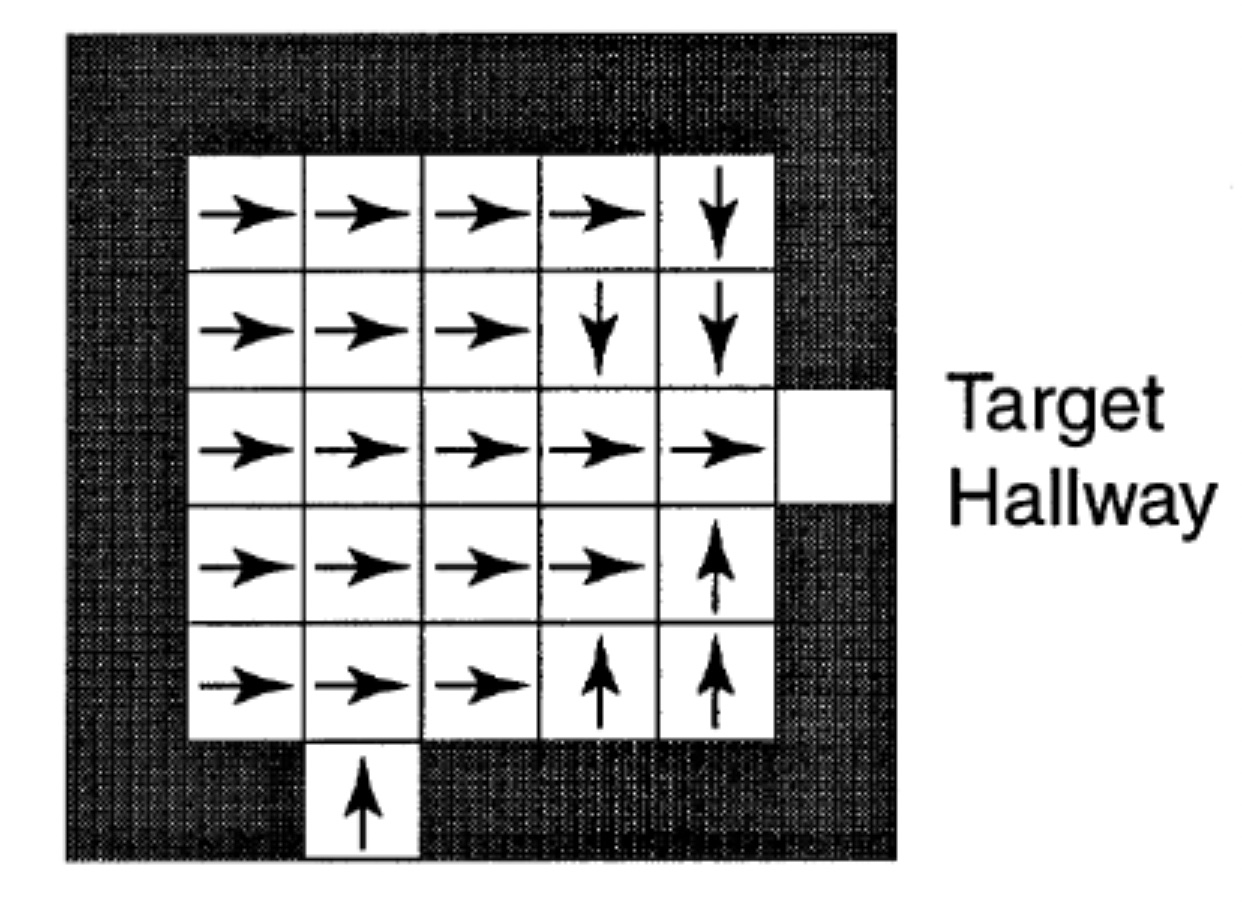}
    \end{tabular}
  \end{center}
  \caption{Four Rooms, and One Room with Subpolicy and Subgoal~\cite{sutton1999between}}\label{fig:4rooms8}
\end{figure}



The four-room environment is a toy environment with which 
algorithms can be explained. More complex versions can be created by
increasing the dimensions of the grids and by increasing the number of
rooms.

\begin{figure}[t]
  \begin{center}
    \includegraphics[width=8cm]{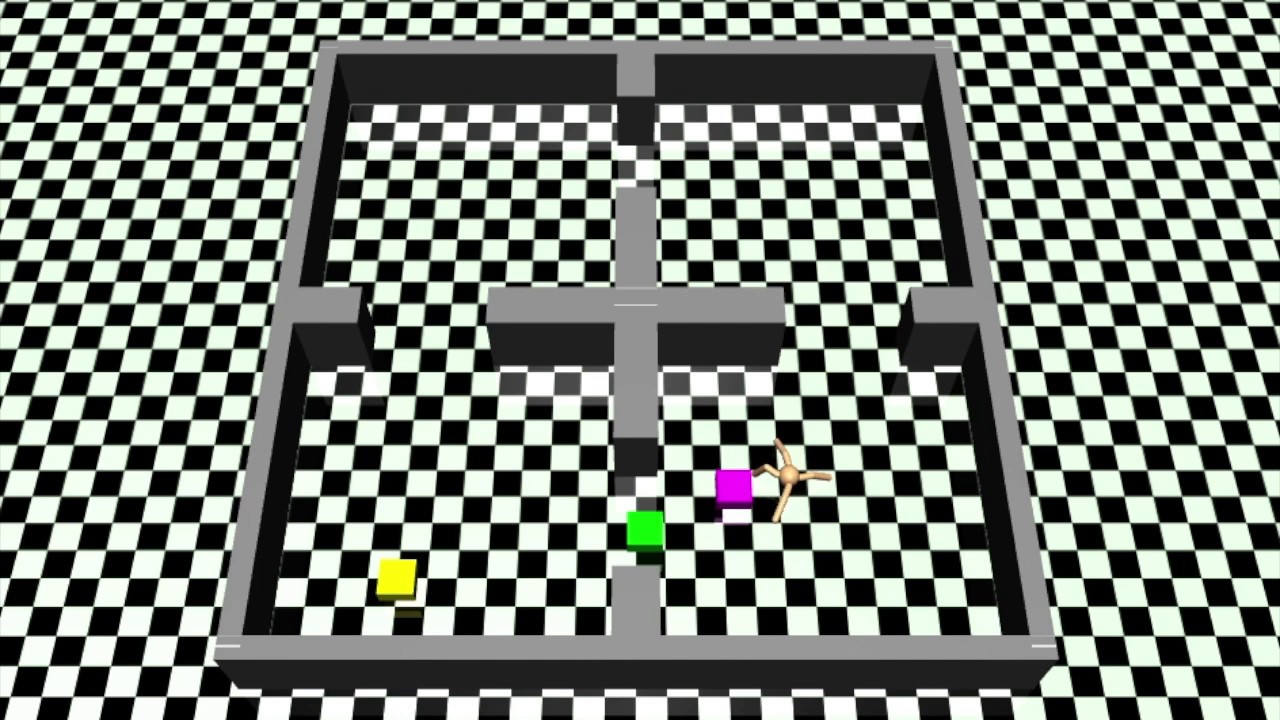}
  \end{center}
  \caption{Ant in Four Rooms~\cite{levy2019learning}}\label{fig:hacimage}
\end{figure}

The Hierarchical actor critic paper uses the four-room environment as
a basis for a robot to crawl through. The agent has to learn both the
locomotion task and solving the four-room problem
(Fig.~\ref{fig:hacimage}).
\begin{figure}[t]
  \begin{center}
    \includegraphics[width=\textwidth]{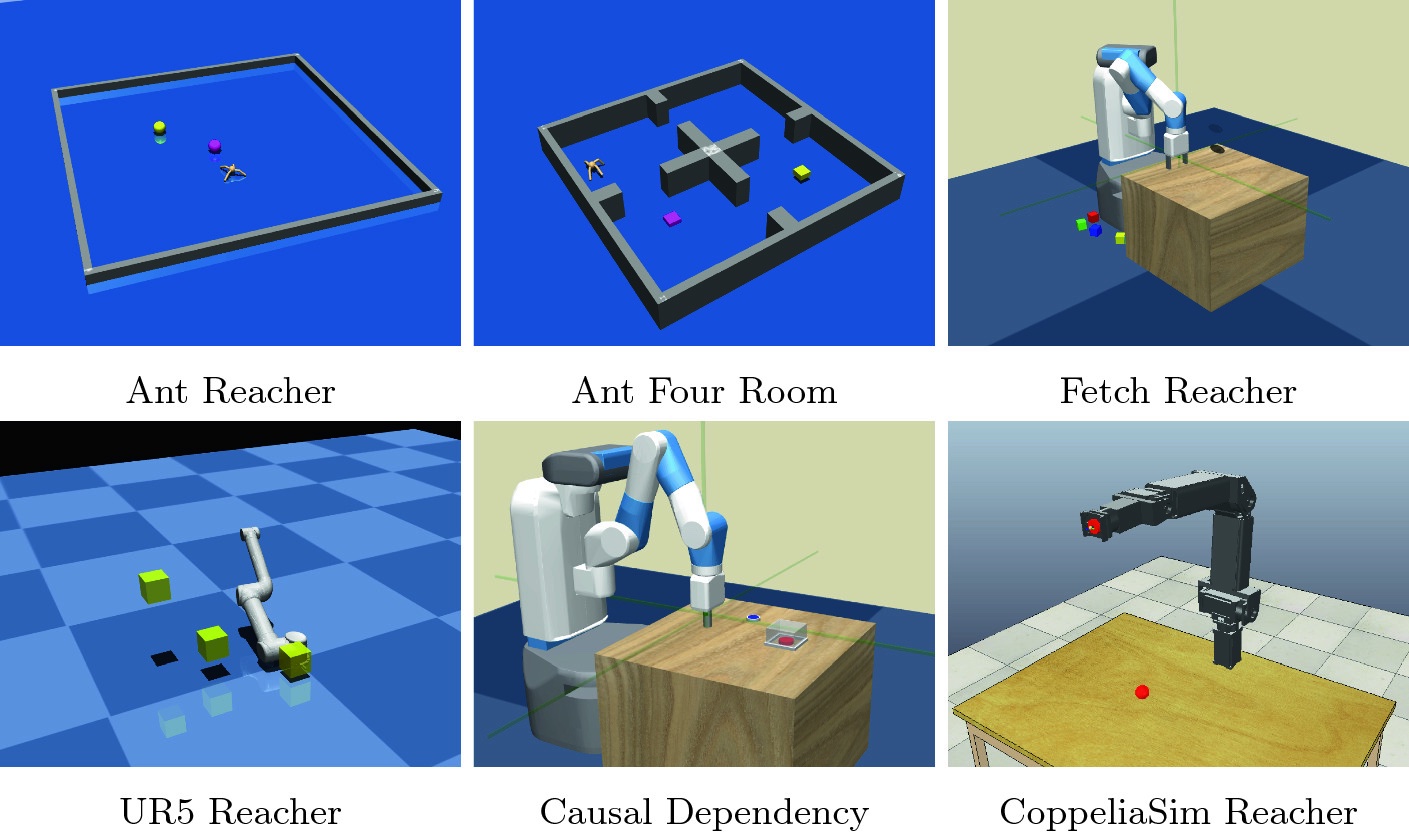}
  \end{center}
  \caption{Six Robot Tasks~\cite{roder2020curious}}\label{fig:robothier}
\end{figure}
Other environments that are used for hierarchical reinforcement
learning are robot tasks, such as shown in
Fig.~\ref{fig:robothier}~\cite{roder2020curious}.

\subsection{Montezuma's Revenge}\index{Montezuma's revenge}\index{intrinsic
  motivation}\label{sec:intrinsic}\index{goal-conditioned}
One of the most difficult situations for reinforcement learning is
when there is  little reward signal, and when it is delayed. The game of Montezuma's
Revenge consists of long stretches in which the agent has to walk
without the reward changing. Without smart exploration methods this
game cannot be solved. Indeed, the game has long been a test bed for
research into 
goal-conditioned and exploration methods.

\begin{figure}[t]
  \begin{center}
    \includegraphics[width=7cm]{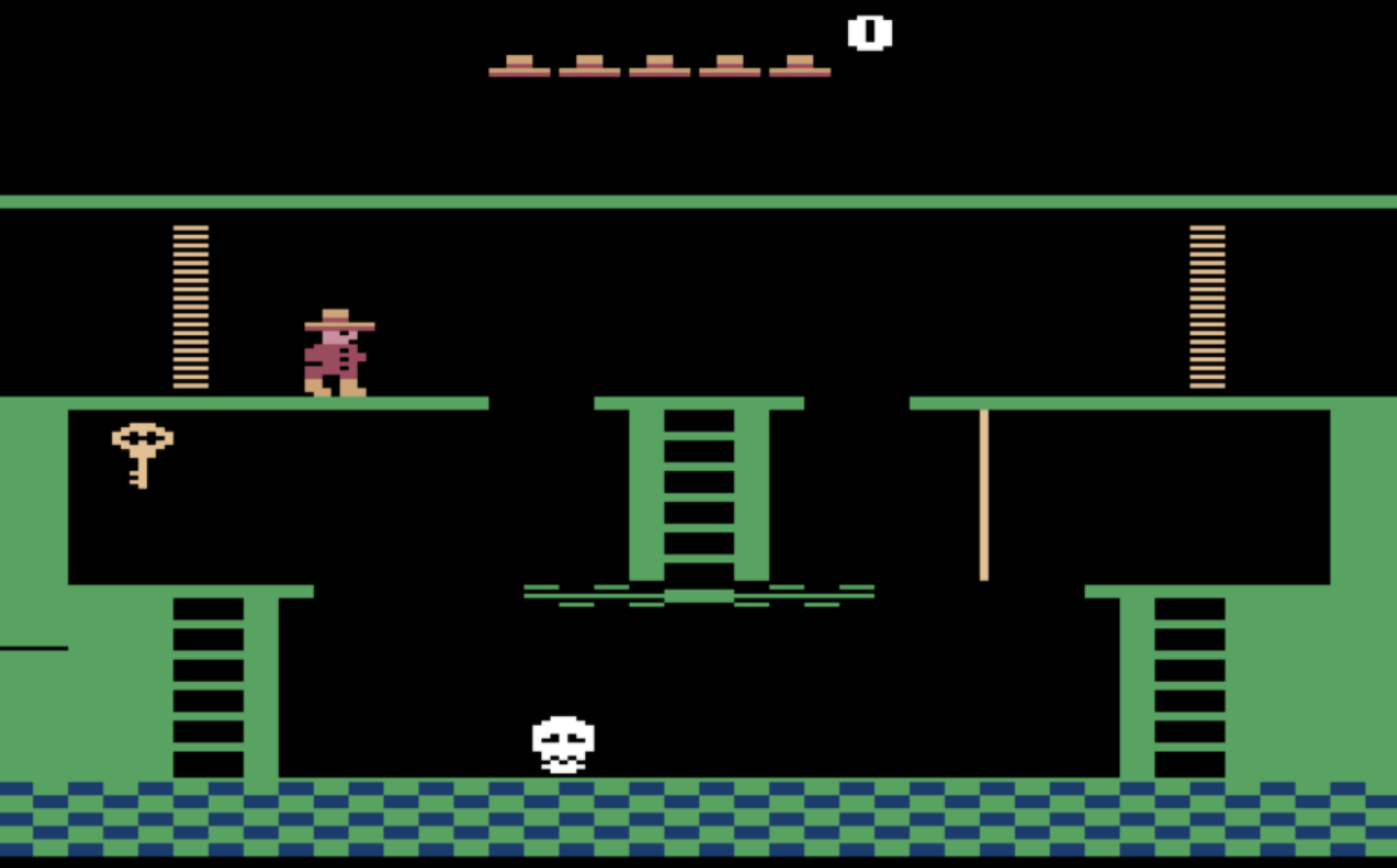}
  \end{center}
  \caption{Montezuma's Revenge \cite{bellemare2013arcade}}\label{fig:montezuma}
\end{figure}

For the state in Fig.~\ref{fig:montezuma}, the player has to go
through several rooms while collecting items. However, to
pass through doors (top right and top left corners), the player needs
the  key. To pick up the key, the player has to  climb down the
ladders  and move  towards the key. This is  a long and
complex sequence before receiving the  reward increments for
collecting the key. Next, the player has to go to the door to collect
another increase in reward.
Flat reinforcement learning algorithms struggle with this environment.
For hierarchical reinforcement the long stretches without a reward
 can be an opportunity to show the usefulness of the option,
jumping through the space from states where the reward changes to
another reward change. To do so, the algorithm has to be able to identify the key
as a subgoal.

Rafati and
Noelle~\cite{rafati2019learning} learn subgoals in Montezuma's
Revenge, and so do Kulkarni et
al.~\cite{kulkarni2016hierarchical}.
Learning to choose promising subgoals is a challenging problem by itself.  Once subgoals
are found, the subpolicies can be learned by introducing a
reward signal for achieving the subgoals. Such intrinsic rewards are
related to \emph{intrinsic motivation} and the psychological concept
of curiosity~\cite{aubret2019survey,oudeyer2009intrinsic}.

\begin{figure}[t]
 \begin{center}
 \includegraphics[width=10cm]{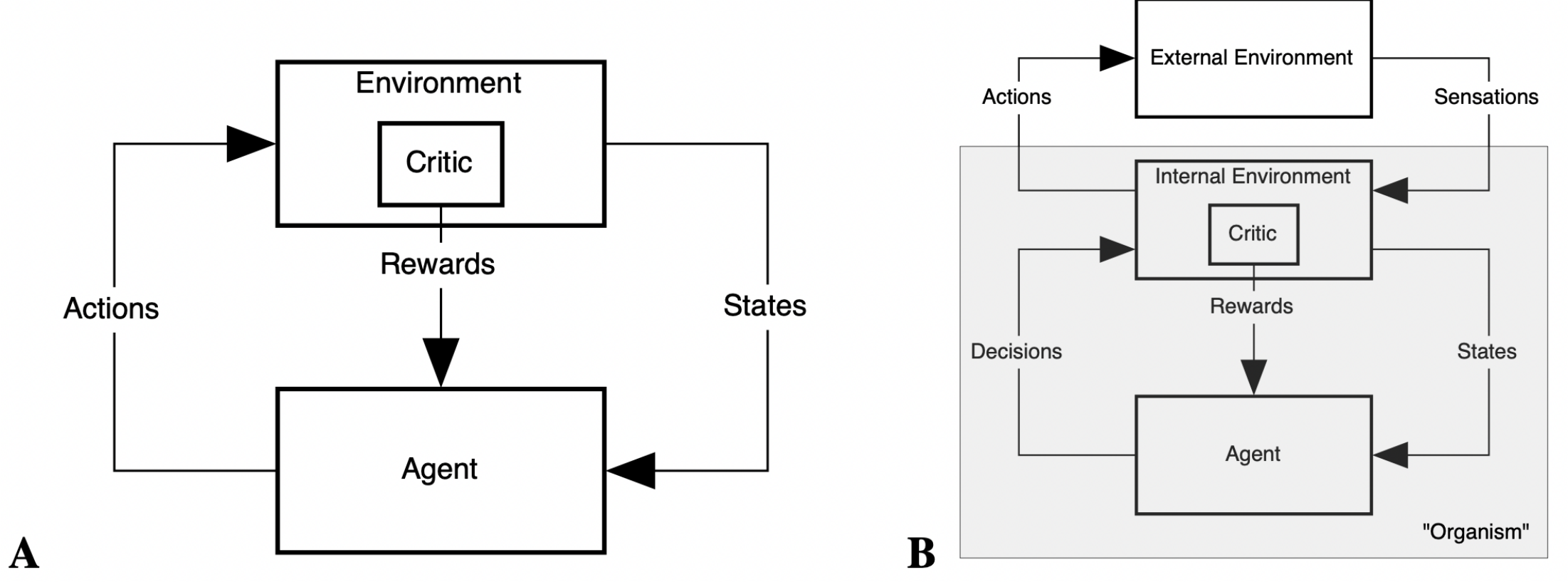}
 \caption{Intrinsic Motivation in Reinforcement Learning~\cite{singh2005intrinsically}}\label{fig:intrinsic}
 \end{center}
 \end{figure}

Figure~\ref{fig:intrinsic} illustrates the idea behind intrinsic
motivation. In ordinary reinforcement learning, a critic in the environment
provides rewards to the agent. When the agent has an internal
environment where an internal critic provides rewards, these internal
rewards provide an intrinsic motivation to the agent. This mechanism aims to more
closely model exploration 
behavior in animals and humans~\cite{singh2005intrinsically}.
For example, during curiosity-driven activities, children use  knowledge to
generate intrinsic goals while playing,  building  block
structures, etc. While doing this, they  
construct subgoals such as putting a lighter entity on top of a
heavier entity in order to build a
tower~\cite{kulkarni2016hierarchical,singh2005intrinsically}. Intrinsic
motivation is an active field of research. A recent survey is~\cite{aubret2019survey}.

Montezuma's Revenge has also been
used as benchmark for the 
Go-Explore algorithm, that has achieved  good results in sparse reward
problems, using a goal-conditioned policy with cell
aggregation~\cite{ecoffet2021first}. Go-Explore performs a
planning-like form of backtracking, combining elements of planning and
learning in a  different way than AlphaZero.

\subsection{Multi-Agent Environments}
Many multi-agent problems are a natural match for hierarchical
reinforcement learning, since agents often work together in teams or
other hierarchical structure. For multi-agent hierarchical problems a
multitude of different environments are used.

Makar et al.~\cite{makar2001hierarchical,ghavamzadeh2006hierarchical} study cooperative
multi-agent learning, and use small tasks such as a  two-agent cooperative trash
collection task, the dynamic rescheduling of automated guided vehicles in
a factory, as well as an environment in which agents communicate
amongst eachother.  Han et al.~\cite{han2019multi} use a multi-agent
Taxi environment. Tang et al.~\cite{tang2018hierarchical} also use
robotic trash collection.

Due to the computational complexity of multi-agent and hierarchical
environments, many of the environments are of lower dimensionality than what
we see in model-free and  model-based single-agent
reinforcement learning.
There are a few exceptions, as we saw in the previous chapter (Capture
the Flag and StarCraft). However, the algorithms that were
used for these environments were  based on population-based self-play algorithms that are
well-suited for parallelization;   hierarchical
reinforcement algorithms of the type that we have discussed in this
chapter were of less importance~\cite{vinyals2019grandmaster}.

\subsection{\em Hands On:  Hierarchical
  Actor Citic Example}\index{hierarchical actor critic}
%
%
%

The research reported in this chapter is of a more manageable scale
than in some other chapters
. Environments are smaller, computational demands are more
reasonable.  Four-room experiments and experiments with movement of
single robot arms invites experimentation and tweaking. Again, as in
the other chapters, the code of most papers can be found online on
GitHub.

Hierarchical reinforcement learning is  well suited for
experimentation because the environments are small, and the concepts
of hierarchy, team, and subgoal, are intuitively appealing. Debgugging
one's implementation should be just that bit easier when the desired
behavior of the different pieces of code is clear.

To get you started with hierarchical reinforcement learning we will go
to HAC: Hierarchical actor critic~\cite{levy2019learning}. 
Algorithm~\ref{alg:HAC} shows the pseudocode, where TBD is the subgoal
in hindsight~\cite{levy2019learning}. 
 \begin{algorithm}[t]
 \caption{Hierarchical Actor Critic \cite{levy2019learning}}\label{alg:HAC}
 \begin{algorithmic}

 \State \textbf{Input:}
    
      Key agent parameters: number of levels in hierarchy $k$, maximum subgoal horizon $H$, and subgoal testing frequency $\lambda$ 
    
 \State \textbf{Output:}
       $k$ trained actor and critic functions $\pi_0, ..., \pi_{k-1}, Q_0, ... , Q_{k-1}$

 \State
 \For{$M$ episodes} \Comment{Train for M episodes}
     \State $s \leftarrow$ $S_{\mbox{init}}$, $g$ $\leftarrow$ $G_{k-1}$ \Comment{Sample initial state and task goal}
     \State train-level($k-1$, $s$, $g$) \Comment{Begin training}
     \State Update all actor and critic networks
 \EndFor
 \State
 \Function{train-level}{$i::$ level, $s::$ state, $g::$ goal}
 \State $s_i \gets s$, $g_i \gets g$ \Comment{Set current state and goal for level $i$}
 \For{$H$ attempts or until $g_n$, $i \leq n < k$ achieved}
     \State $a_i$ $\gets$ $\pi_i(s_i, g_i)$ + noise (if not subgoal testing) \Comment{Sample (noisy) action from policy}
     \If{$i > 0$} 
         \State Determine whether to test subgoal $a_i$
         \State $s_{i}' \gets$ train-level$(i-1, s_i, a_i)$ \Comment{Train level $i-1$ using subgoal $a_i$}
     \Else
         \State Execute primitive action $a_0$ and observe next state $s_{0}'$
     \EndIf
     \State  \Comment{Create replay transitions}
     \If{$i > 0$ and $a_i$ missed} 
         \If{$a_i$ was tested} \Comment{Penalize subgoal $a_i$}
             \State Replay\_Buffer$_i \gets [s = s_i, a = a_i, r = $ Penalty$, s' = s_{i}', g = g_i, \gamma = 0]$
         \EndIf
         \State $a_i \gets s_i'$ \Comment{Replace original action with action executed in hindsight}
     \EndIf

     \State \Comment{Evaluate executed action on current goal and hindsight goals}
     \State Replay\_Buffer$_i \gets [s = s_i, a = a_i, r \in \{-1,0\}, s' = s_{i}', g = g_{i}, \gamma \in \{\gamma, 0\}]$
     \State HER\_Storage$_i \gets [s = s_i, a = a_i, r = $TBD$, s' =
     s_{i}', g = $TBD$, \gamma = $TBD$]$ 
     \State $s_i \gets s_i'$
 \EndFor
 \State Replay\_Buffer$_i \gets$ Perform HER using HER\_Storage$_i$ transitions

 \State \textbf{return} $s_i'$\Comment{Output current state}
 \EndFunction

 \end{algorithmic}
 \end{algorithm}
A blog\footnote{\url{http://bigai.cs.brown.edu/2019/09/03/hac.html}}
with animations has been written, a
video\footnote{\url{https://www.youtube.com/watch?v=DYcVTveeNK0}} has
been made of the results, and the code can be found in
GitHub.\footnote{\url{https://github.com/andrew-j-levy/Hierarchical-Actor-Critc-HAC-}}

To run the hierarchical actor critic experiments, you need MuJoCo and
the required Python wrappers. The code is TensorFlow 2 compatible. When you
have cloned the repository, run the experiment with
\begin{tcolorbox}
  \verb|python3 initialize_HAC.py --retrain|
\end{tcolorbox}

  which will train a UR5 reacher
agent with a 3-level hierarchy. Here is a video\footnote{\url{https://www.youtube.com/watch?v=R86Vs9Vb6Bc}} that shows how it
should look like after 450 training episodes. You can watch your
trained agent with the command
\begin{tcolorbox}
\verb|python3 initialize_HAC.py --test --show| 
\end{tcolorbox}
The README at the GitHub repository contains more suggestions on what
to try.
You can try different hyperparameters, and you can modify the designs,
if you feel like it. Happy experimenting!

\section*{Summary and Further Reading}
\addcontentsline{toc}{section}{\protect\numberline{}Summary and Further Reading}
We will now summarize the chapter and provide pointers to further reading.
\subsection*{Summary}
A typical reinforcement learning algorithm moves in small steps. For a
state, it picks an action, gives it to the environment for a new
state and a reward, and processes the reward to pick a new
action. Reinforcement learning works step by small step.
In contrast, consider the following problem:
in the real world, when we plan a trip from A to B, we use abstraction
to reduce the 
state space, to be able to reason at a higher level. We
do not reason at the level of footsteps  to take, but we first
decide on the mode of transportation to get close to our goal, and
then we fill in the different parts of the journey with small steps.

Hierarchical reinforcement learning tries to mimic this idea:
conventional reinforcement learning  works at the level of a single
state; {\em hierarchical\/} reinforcement 
learning performs  abstraction, solving subproblems in
sequence. Temporal abstraction is described in a  paper by
Sutton et al.~\cite{sutton1999between}.
Hierarchical reinforcement learning uses the principles of divide and
conquer to make solving large problems feasible. It finds 
subgoals in the space that it  solves with subpolicies (macros or
options).

Despite the appealing
intuition, progress in hierarchical reinforcement learning was
initially slow. Finding these new subgoals and
subpolicies is a computationally intensive problems that is
exponential in the number of actions, and in some situations
it is quicker to use conventional ``flat'' reinforcement learning
methods, unless domain knowledge can be exploited. The advent of deep
learning provided a boost to 
hierarchical reinforcement learning, and much progress is being
reported in important tasks such as learning subgoals
automatically, and finding subpolicies.

Although popular  for single-agent reinforcement learning,
hierarchical methods are also used in  multi-agent
problems. Multi-agent problems often feature agents that work in
teams, that cooperate within, and compete between the teams. Such an
agent-hierarchy is a natural fit for hierachical solution
methods. 
Hierarchical
reinforcment learning remains a promising technique.

\subsection*{Further Reading}

Hierarchical reinforcement learning, and subgoal finding, have a rich and
long history~\cite{florensa2018automatic,parr1998reinforcement,sutton1999between,precup1997planning,dietterich2000hierarchical,hauskrecht2013hierarchical,laird1986chunking,barto2003recent,pateria2021hierarchical};
see also Table~\ref{tab:hrl}.
Macro-actions
are a basic
approach~\cite{hauskrecht2013hierarchical,randlov1999learning}. Others, using macros,
are~\cite{xu2019macro,xiao2020macro,durugkar2016deep}. The options
framework has provided a boost to the development of the
field~\cite{sutton1999between}. Other approaches are
MAXQ~\cite{dietterich1998maxq} and Feudal
networks~\cite{vezhnevets2017feudal}.

Earlier tabular approaches are~\cite{florensa2018automatic,parr1998reinforcement,sutton1999between,precup1997planning,dietterich2000hierarchical,hauskrecht2013hierarchical,laird1986chunking}.

Recent method are
Option-critic~\cite{bacon2017option} and Hierarchical
actor-critic~\cite{levy2019learning}.
%
%
There are many  deep learning methods for finding subgoals and subpolicies~\cite{levy2019learning,panov2018automatic,florensa2018automatic,pertsch2020long,nachum2018data,frans2017meta,veeriah2021discovery,schaul2015universal,sunehag2017value,daniel2016probabilistic}. Andrychowicz
et al.~\cite{andrychowicz2017hindsight}
introduce Hindsight experience replay, which can improve performance
for hierarchical methods.

Instrinsic motivation 
is a concept from developmental neuroscience that has come to
reinforcement learning with the purpose of providing learning signals in
large spaces. It is
related to curiosity. Botvinick et
al.~\cite{botvinick2009hierarchically} have written an overview of
hierarchical reinforcement learning and neuroscience. Aubret et al.~\cite{aubret2019survey}
provide a survey of intrinsic motivation for reinforcement
learning. Instrinsic motivation is used
by~\cite{kulkarni2016hierarchical,rafati2019learning}.
Intrinsic motivation is closely related to goal-driven
reinforcement learning~\cite{schaul2015universal,ryan2000intrinsic,oudeyer2008can,oudeyer2009intrinsic,oudeyer2007intrinsic}.

\section*{Exercises}
\addcontentsline{toc}{section}{\protect\numberline{}Exercises}
It is time for the Exercises to test your knowledge.
\subsubsection*{Questions}
Below are some quick questions to check your understanding of this
chapter. For each question a simple, single sentence answer should be sufficient.

\begin{enumerate}
\item Why can hierarchical reinforcement learning be faster?
\item Why can hierarchical reinforcement learning be slower?
\item Why may hierarchical reinforcement learning give an answer of
  lesser quality?
  \item Is hierachical reinforcement more general or less general?
\item What is an option?
\item What are the three elements that an option consists of?
\item What is a macro?
\item What is intrinsic motivation?
\item How do multi agent and hierarchical reinforcement learning fit together?
\item What is so special about Montezuma's Revenge?
\end{enumerate}

\subsubsection*{Exercises}
Let us go to the programming exercises  to become more familiar with the methods that
we have covered in this chapter.

\begin{enumerate}
\item \emph{Four Rooms}
  Implement a hierarchical solver for the four-rooms
  environment. You can code the hallway subgoals using domain
  knowledge. Use a simple tabular, planning, approach. How will you
  implement the subpolicies?

\item \emph{Flat}
  Implement a flat planning or Q-learning-based
    solver for 4-rooms. Compare this program to the tabular
    hierarchical solver. Which is quicker? Which of the two does fewer environment
    actions? 
  \item \emph{Sokoban} Implement a Sokoban solver using a hierarchical
    approach  (challenging). The challenge in Sokoban is that there can be dead-ends
    in the game that you create rendering the game
    unsolvable (also see the
    literature~\cite{shoham2021solving,grinsztajn2021there}). Recognizing these dead-end
    moves is important. What are the subgoals? Rooms, or each box-task
    is one 
    subgoal, or can you find a way to code dead-ends as subgoal? How
    far can you get? Find Sokoban levels.\footnote{\url{http://sneezingtiger.com/sokoban/levels.html}}\footnote{\url{http://www.sokobano.de/wiki/index.php?title=Level_format}}\footnote{\url{https://www.sourcecode.se/sokoban/levels}} 
    \item \emph{Petting Zoo} Choose one of the easier multi-agent problems from the
      Petting Zoo~\cite{terry2020pettingzoo}, introduce teams, and write a hierarchical
      solver. First try a tabular planning approach, then look at
      hierarchical actor critic (challenging).
    \item \emph{StarCraft} The same as the previous exercise, only now with
      StarCraft (very challenging).
\end{enumerate}


\chapter{Meta-Learning}\label{chap:txl}\label{chap:meta}


Although current deep reinforcement learning methods have obtained great
successes, training times for most interesting problems are high;
they are often measured in weeks or
months, consuming  time and resources---as you may have noticed while
doing some of the exercises at the end of the chapters. 

Model-based methods aim to  reduce the sample complexity in
order to speed up learning---but still, for each new task a new network has to
be trained from scratch. In this chapter we turn to another
approach, that aims to \emph{re-use} information learned in earlier training
tasks from a closely related problem. 
When humans learn a new task, they do not learn from a blank slate.
Children learn to walk and then they learn to run; they follow a training curriculum, and
they remember. Human learning
 builds on existing knowledge, using
knowledge from previously learned tasks to facilitate the
learning of  new tasks. 
In machine learning such transfer of previously learned knowledge from one  task to another
is called \emph{transfer learning}. We will study it in this chapter. 

Humans learn continuously.  When learning a new task, we do not start  from
scratch, zapping our minds first to emptiness. Previously learned
task-representa\-tions allow us to learn 
new representations for new tasks quickly; in effect, we have learned to learn.
Understanding  how we (learn to) learn has intrigued artificial intelligence
researchers since  the
early days,  
 and it is the topic of
this chapter.

The fields of transfer learning and meta-learning
are tightly related. For both, the goal is to speed up learning a new
task, using previous knowledge. In transfer learning, we pretrain our
parameter network with knowledge from a single  task. In meta-learning, we
use multiple  related tasks.


In this chapter, we first discuss 
the concept of
lifelong learning, something that is quite familiar to human
beings. Then we discuss transfer learning, followed by meta-learning. Next, we discuss some of the benchmarks that are used to
test transfer learning and meta-learning.

The chapter is concluded with  exercises, a summary, and pointers
to further reading.

\section*{Core Concepts}
\begin{itemize}
\item Knowledge transfer
\item Learning to learn
\end{itemize}

\section*{Core Problem}
\begin{itemize}
\item Speed-up learning with knowledge from related  tasks
\end{itemize}

\section*{Core Algorithms}
\begin{itemize}
\item Pretraining (Listing~\ref{lst:pretrain1})
\item Model-Agnostic Meta-Learning (Alg.~\ref{alg:mamlrl})
\end{itemize}

\section*{Foundation Models}

Humans are good at meta-learning. We  learn new tasks more
easily after we have learned other tasks. Teach us to walk, and we learn how
to run. Teach us to play the violin, the viola, and the cello, and we
more easily learn to play the double bass (Fig.~\ref{fig:violin}).

\begin{figure}[t]
\begin{center}
\includegraphics[width=7cm]{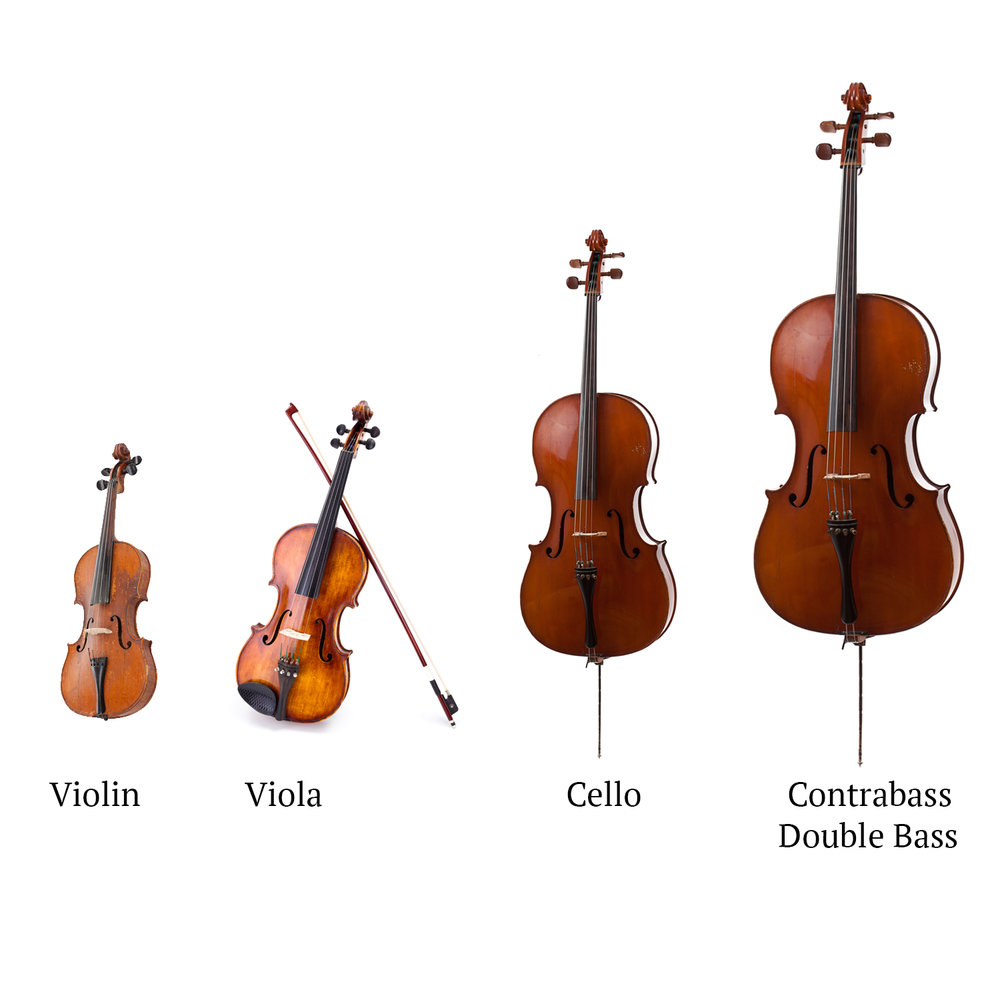}
\caption{Violin, viola, cello and double bass}\label{fig:violin}
\end{center}
\end{figure}

Current deep learning networks are large, with many layers of neurons and millions
of parameters. For new problems,  training large networks  on large
datasets or environments  takes time, up to weeks,
or months---both for supervised and for reinforcement learning. In order to shorten training
times for subsequent networks, these are often pretrained, using
foundation models~\cite{bommasani2021opportunities}.  With
pretraining, some of the exisiting
weights of another network are used as starting point for finetuning a network
on a new dataset, instead of using a randomly initialized
network.\index{foundation models}

Pretraining works especially well on deeply layered architectures. The
reason is that the ``knowledge'' in the layers goes from generic to
specific: lower layers contain 
generic filters such as lines and curves, and upper layers contain
more specific filters such as ears, noses, and mouths (for a face
recognition
application)~\cite{lecun2015deep,mahajan2018exploring}. These lower
layers contain more generic information that is well suited for
transfer to other tasks.
\begin{figure}[t]
\begin{center}
\includegraphics[width=\textwidth]{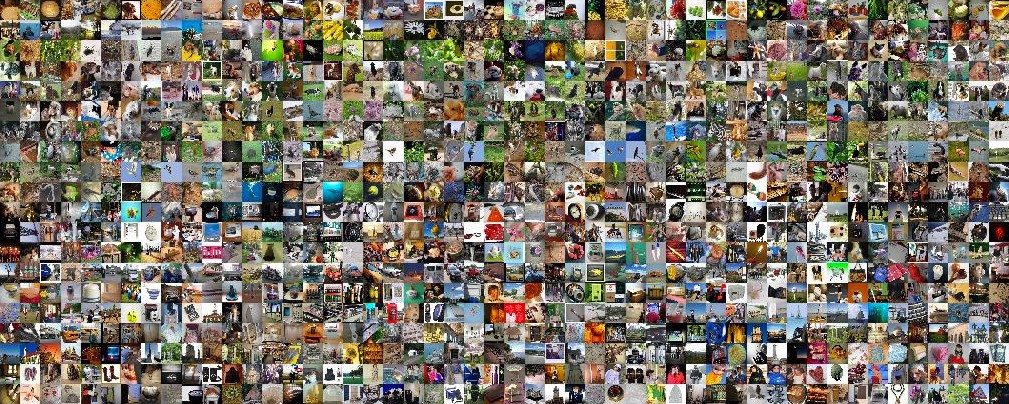}
\caption{Imagenet Thumbnails \cite{russakovsky2015imagenet}}\label{fig:imagenet7}
\end{center}
\end{figure}

Foundation models are large models in a certain
field, such as image recognition, or natural language processing, that are trained
extensively on large datasets. Foundation models contain general
knowledge, that can be specialized for a certain purpose.
The world of applied deep learning has moved from training a net from scratch for a
certain problem, to taking a part of an existing net that is trained
for a related problem and then finetuning it on the new task.
%
%
%
Nearly all state-of-the-art visual perception approaches rely on the
same approach: (1) pretrain a convolutional network on a large,
manually annotated image classification dataset and (2) finetune the
network on a smaller, task-specific
dataset~\cite{girshick2014rich,donahue2014decaf,zeiler2014visualizing,agrawal2014analyzing,huh2016makes},
see Fig.~\ref{fig:imagenet7} for some thumbnails of  Imagenet~\cite{russakovsky2015imagenet}. For natural
language recognition,  pretraining is also the norm---for example, for large-scale
pretrained language models such as Word2vec~\cite{mikolov2013efficient,mikolov2013distributed}, BERT~\cite{devlin2018bert} and
GPT-3~\cite{radford2018improving}.\index{foundation model}

In this chapter, we will study pretraining, and more.

\section{Learning to Learn Related Problems}

Training times for modern deep networks are large. 
Training AlexNet for ImageNet took 5-6 days on 2 GPUs
in 2012~\cite{krizhevsky2012imagenet}, see Sect.~\ref{sec:alexnet}. 
In reinforcement learning,
training AlphaGo took 
weeks~\cite{silver2016mastering,silver2017mastering}, in natural
language processing, training also takes a long
time~\cite{devlin2018bert}, 
even excessively long as in the case of GPT-3~\cite{brown2020language}. 
%
%
%
Clearly, some solution is needed.
Before we look closer at transfer learning, let us  have a 
look at the bigger picture: lifelong learning. 
  
\index{lifelong learning}

When humans learn a new task,  learning  is based on previous
experience. Initial learning by infants
of elementary skills in vision, speech, and locomotion 
takes years. Subsequent learning of new skills  builds on the
previously acquired skills. Existing knowledge is
adapted, and new skills are learned based on previous skills.

Lifelong learning remains a
long-standing challenge for machine learning; in current methods  the continuous acquisition of
information  often leads
to interference of
concepts or catastrophic forgetting~\cite{silver2013lifelong}. This limitation represents 
a major drawback for deep networks that typically learn
representations from stationary batches of training data.
Although  some advances have
been made in narrow domains, significant
advances are necessary to approach generally-applicable lifelong
learning.

\begin{table}[t]
  \begin{center}
  \begin{tabular}{l|l|l}
    {\bf Name} & {\bf Dataset} & {\bf Task}  \\
    \hline\hline
    Single-task Learning &  $D_{train} \subseteq D,  D_{test}
                           \subseteq D$ & $T = T_{train}= T_{test}$  \\
    Transfer Learning &  $D_1\gg D_2$ & $T_1\neq T_2$  \\
    Multi-task Learning &  $D_{train} \subseteq D,  D_{test}
                           \subseteq D$  & $T_1\neq T_2$  \\
    Domain Adaptation  &  $D_1\neq D_2$  & $T_1=T_2$  \\
    Meta-Learning  &  $\{D_1, \ldots,  D_{N-1}\} \gg D_N$  &
                                                             $T_1,\ldots,
                                                             T_{n-1}
                                                             \neq T_N$  \\
    \hline

  \end{tabular}
  \caption[Different kinds of supervised learning]{Different Kinds of Supervised Learning. A typical \emph{single learning
    task} is to classify pictures of animals into different
    classes. The dataset is split into a train and a testset, and the
    task (loss function) is the same at train and test time. A
    \emph{transfer learning} task uses part of the knowledge (network
    parameters) that are learned on a large dataset to initialize a second
  network, that is trained subsequently (fine tuned) on a different
  dataset. This second dataset/learning task is related to the first
  task, so the learning the second task goes faster. For example,
  having learned to recognize cars, may be useful to speedup recognizing
trucks. In \emph{multi-task learning} several related tasks are trained at the
same time, possibly benefitting from better regularization. An example
could be training spam filters for different users at the same
time. \emph{Domain adaptation} tries to adapt the network to a new dataset of
related examples, such as images of pedestrians in different light
conditions. \emph{Meta-learning} tries to learn meta knowledge, such
as hyperparameters over a sequence of related (larger) learning tasks, so that
a new learning task goes faster. In deep meta-learning these
hyperparameters include the initial
network parameters, in this sense meta-learning can be considered to
be multi-job transfer learning.} \label{tab:transfer}
\end{center}
\end{table}

Different approaches have been developed.
Among the methods are meta-learning, domain
adaptation, multi-task learning, and pretraining.
Table~\ref{tab:transfer} lists these
approaches, together with regular single task learning. The learning
tasks are formulated  using datasets, as in a regular supervised
setting. The table shows how 
the lifelong learning methods differ in their training and test
dataset, and the different learning tasks.

The first line shows
regular single-task learning. For single-task learning, the
training and test dataset are both drawn from the same distribution
(the datasets do not contain the same examples, but they are drawn from the same
original dataset and the data distribution is expected to be the same),
and the task to perform is the same for training and test.

In the next
line, for
transfer learning, networks trained on 
one dataset are used to speedup training for a different task,
possibly using a much smaller dataset~\cite{pan2010survey}. Since the
datasets are not drawn from the same master dataset, their
distribution will differ, and there typically is only an informal
notion of how ``related'' the datasets are. However, in practice transfer learning
often provides significant speedups, and transfer
learning, pretraining and finetuning are currently used in many
real-world training tasks, sometimes using large foundation models as a basis.

In multi-task learning, more than one task is learned from one
dataset~\cite{caruana1997multitask}. The tasks are often related, such
as classification tasks of different, but related, classes of images,
or learning spam filters for different email-users. Regularization may
be improved when  a neural network is trained on related tasks at the
same time~\cite{baxter2000model,ciliberto2015convex}. 

So far, our learning tasks were trying to speedup learning different
tasks with related data. Domain adaptation switches this around: the task remains the
same, but the data changes. In domain adaptation, a different dataset is used to perform the same
task, such as recognizing pedestrians in different light
conditions~\cite{tommasi2016learning}. 

In  meta-learning,  both datasets and tasks are different, although
not too different. In meta-learning, a sequence of datasets and learning tasks is
generalized to learn a new (related) task
quickly~\cite{brazdil2008metalearning,hospedales2020meta,huisman2020deep,schaul2010metalearning}. The
goal of meta-learning is to learn hyperparameters over a sequence of
learning tasks.

\section{Transfer Learning and Meta-Learning Agents}

We will now introduce transfer learning and meta-learning algorithms.
%
Where in normal learning we
would intialize our parameters randomly, in transfer learning we
initialize them with (part of) the training results of another
training task. This other task is related in some way to
the new task, 
for example, a task to recognize images of dogs playing in a
forest is initalized on a dataset of  dogs playing in a park.
The parameter transfer of the old task is called \gls{pretraining}, the second phase, where
the network learns the new task on the new dataset, is called \gls{finetuning}. The
pretraining will hopefully allow the new task to train faster.

Where \gls{transfer learning} \emph{transfers} knowledge from a \emph{single} previous 
task, \gls{meta-learning} aims to \emph{generalize} knowledge from \emph{multiple} previous
learning tasks.  
Meta-learning tries to learn hyperparameters over these related learning
tasks, that tell the algorithm how to learn the new task. Meta-learning thus aims to learn to learn. In deep meta-learning approaches, the initial network parameters are
typically part of the hyperparameters. Note that in transfer learning
we also use (part of) the parameters to speed up learning (finetuning) a new,
related task. We can say that deep meta-learning generalizes transfer
learning by learning 
the initial parameters over not one but a sequence of related
tasks~\cite{hospedales2020meta,huisman2021survey}. (Definitions are
still in flux, however, and different authors and different fields have
different definitions.)

Transfer learning has become part of the standard
approach in machine learning, meta-learning is still an area of active
research. We will
look into meta-learning shortly, after we have
 looked into transfer learning, multi-task learning, and domain adaptation.

\subsection{Transfer Learning}\label{sec:transfer}\index{transfer learning}

\index{multi-task learning}
Transfer learning aims to improve the process of learning new tasks
using the experience gained by solving similar  problems~\cite{pratt1993discriminability,thrun2012learning,thrun2012explanation,parisi2019continual}. 
Transfer learning aims
to transfer past experience of  source tasks and
use  it to boost learning in a related target
task~\cite{pan2010survey,zhuang2020comprehensive}.

In transfer learning, we first train a base network on a base dataset
and task, and then we repurpose some of the learned features to a second target network to be trained on a target dataset and
task. This process works better if the features are general,
meaning suitable to both base and target tasks, instead of specific to
the base task. 
This form of transfer learning is called
inductive transfer. The scope of possible models (model
bias) is narrowed in a beneficial way by using a model fit on a
different but related task.

First we will look at task similarity, then at transfer learning,
multi-task learning, and domain adaptation.

\subsubsection{Task Similarity}

Clearly, pretraining
works  better when the tasks are
similar~\cite{caruana1997multitask}. Learning to play the viola 
based on the violin is more similar than learning the tables of multiplication based
on  tennis. 
%
Different  measures can be used to measure the similarity
of examples and features in datasets, from linear one-dimensional
measures to non-linear multi-dimensional measures. 
Common measures are the cosine similarity
for real-valued vectors and the radial basis function
kernel~\cite{tan2016introduction,vert2004primer}, but many more
elaborate measures have been devised.

Similarity measures are also used to devise meta-learning algorithms,
as we will see later.

\subsubsection{Pretraining and Finetuning}

  

  
When we want to transfer knowledge, we can  transfer the weights of
the network,  and then start
re-training with the new dataset.
Please refer back to Table~\ref{tab:transfer}.
In pretraining   the new dataset is smaller
than  the old dataset   $D_1\gg D_2$, and we train for a new task
$T_1\neq T_2$, which we want to train faster. This works when the new
task is different, but similar, 
so that the old dataset $D_1$ contains useful information for the new
task $T_2$.

To learn new image recognition problems, it is common to use a deep learning
model pre-trained for a large and challenging image classification
task such as the ImageNet 1000-class photograph classification
competition.
%
%
%
%
Three examples of pretrained models  include:
the Oxford VGG Model,
Google's Inception Model,
Microsoft's ResNet Model.
For more examples, see the Caffe Model
Zoo,\footnote{\url{https://caffe.berkeleyvision.org/model_zoo.html}}
or other zoos\footnote{\url{https://modelzoo.co}} where more pre-trained
models are shared.

Transfer learning is effective because the images were trained on a 
corpus that requires the model to make predictions on a
large number of classes, requiring the model to be general, and since it
efficiently learns to extract features in order to
perform well. 
 %


Convolutional neural network features are more generic in lower
layers, such as color blobs or Gabor filters, and more specific to the
original dataset in higher 
layers.
Features must
eventually transition from general to specific in the last layers of
the network~\cite{yosinski2014transferable}. 
Pretraining
copies some of the layers to the new task. Care should be taken how
much of the old task network to copy. It is relatively safe to copy
the more general lower layers. Copying the more specific higher layers
may  be detrimental to performance.

In natural language processing a similar situation occurs. 
In natural language processing, a word embedding is used that is a
mapping of words to a high-dimensional continuous vector  where
different words with a similar meaning have a similar vector
representation. 
Efficient algorithms exist to learn these  word representations. Two
examples of common  pre-trained word models
trained on very large datasets of text documents include Google’s
Word2vec model~\cite{mikolov2013efficient} and Stanford’s GloVe
model~\cite{pennington2014glove}. 


\subsubsection{\em Hands-on: Pretraining Example}\label{sec:pretraining}
Transfer learning and pretraining have become a standard approach to
learning new tasks, especially when only a small dataset is present,
or when we wish to limit  training time. Let us have a look at a hands-on
example that is part of the Keras distribution (Sect.~\ref{sec:keras}). The Keras transfer
learning example provides a basic Imagenet-based approach, getting
data from TensorFlow DataSets (TFDS). The example follows a supervised
learning approach, although the learning and fine-tuning phase can
easily be substituted by a reinforcement learning setup.
The Keras transfer learning example is at the Keras
site,\footnote{\url{https://keras.io/guides/transfer_learning/}} and
can also be run in a Google
Colab.

The most common incarnation of transfer learning in the context of
deep learning is the following worfklow: 
\begin{enumerate}
\item Take layers from a previously trained model.
\item Freeze them, so as to avoid destroying any of the information they contain during future training rounds.
\item Add some new, trainable layers on top of the frozen layers. They
  will train on the new dataset using  the old features as predictions.
\item Train the new layers on your new (small) dataset.
\item A last, optional, step, is fine-tuning of the frozen layers, which consists of
  unfreezing the entire model you obtained above, and
  re-training it on the new data with a very low learning rate. This
  can potentially achieve meaningful improvements, by incrementally
  adapting the pretrained features to the new data.
\end{enumerate}
%
%
%
%
%
Let us look at
how this workflow works in practice in Keras. At the Keras site we find an accessible example code for pretraining (see
Listing~\ref{lst:pretrain1}).  

First, we instantiate a base model with pretrained weights. (We do not
include the classifier on top.)
Then, we freeze the base model (see Listing~\ref{lst:pretrain2}).
\lstset{label={lst:pretrain1}}
\lstset{caption={Pretraining in Keras (1): instantiate model}}
\begin{lstlisting}[language=Python,float]
base_model = keras.applications.Xception(
    weights='imagenet',  # Load weights pre-trained on ImageNet.
    input_shape=(150, 150, 3),
    include_top=False)  # Do not include the ImageNet classifier at the top.

base_model.trainable = False
\end{lstlisting}
Next, we create a new model on top, and train it.

\lstset{label={lst:pretrain2}}
\lstset{caption={Pretraining in Keras (2): create new model and train}}
\begin{lstlisting}[language=Python,float]
inputs = keras.Input(shape=(150, 150, 3))
# The base_model is running in inference mode here,
# by passing `training=False`. This is important for fine-tuning
x = base_model(inputs, training=False)
# Convert features of shape `base_model.output_shape[1:]` to vectors
x = keras.layers.GlobalAveragePooling2D()(x)
# A Dense classifier with a single unit (binary classification)
outputs = keras.layers.Dense(1)(x)
model = keras.Model(inputs, outputs)


model.compile(optimizer=keras.optimizers.Adam(),
              loss=keras.losses.BinaryCrossentropy(from_logits=True),
              metrics=[keras.metrics.BinaryAccuracy()])
model.fit(new_dataset, epochs=20, callbacks=..., validation_data=...)

\end{lstlisting}

The Keras
example
contains this example 
and others, including fine-tuning. Please go to the Keras site and
improve your experience with 
pretraining in practice.

\subsubsection{Multi-task Learning}\index{multi-task learning}
Multi-task learning is related to transfer learning. In multi-task
learning a  single network is trained \emph{at the same time} on multiple
related tasks~\cite{thrun1996learning,caruana1997multitask}.


In multi-task learning the learning process of one task benefits from
the simultaneous learning of the related task.
This approach is effective when the tasks have some
commonality, such as learning to recognize breeds of dogs and breeds
of cats. In multi-task learning, related learning
tasks are learned at the same time, whereas in transfer learning they
are learned in sequence by different networks. 
Multi-task learning improves
regularization by  requiring the algorithm to perform well on a
related learning task instead of penalizing all overfitting
uniformly~\cite{evgeniou2004regularized,argyriou2007multi}.
%
The two-headed AlphaGo Zero
network optimizes for value and
for policy at the same time in the same
network~\cite{pan2010survey,caruana1997multitask}. A multi-headed
architecture is often used in multi-task learning, although in AlphaGo
Zero the two
heads are trained for two related aspects (policy and value) of the
same task (playing Go games).

Multi-task
learning has  been applied with success to Atari
games~\cite{kelly2018emergent,kelly2017multi}.

\subsubsection{Domain Adaptation}\index{domain adaptation}
Domain adaptation is necessary  when there
is a change in the data
distribution between the training dataset  and the test dataset
(domain shift). This problem is related to 
out-of-distribution learning~\cite{levine2020offline}. Domain shifts
are common in practical 
applications of artificial intelligence, such as when items must be
recognized in different light conditions, or when the background
changes. Conventional machine-learning 
algorithms often have difficulty adapting to such changes.

\begin{figure}[t]
\begin{center}
\includegraphics[width=9cm]{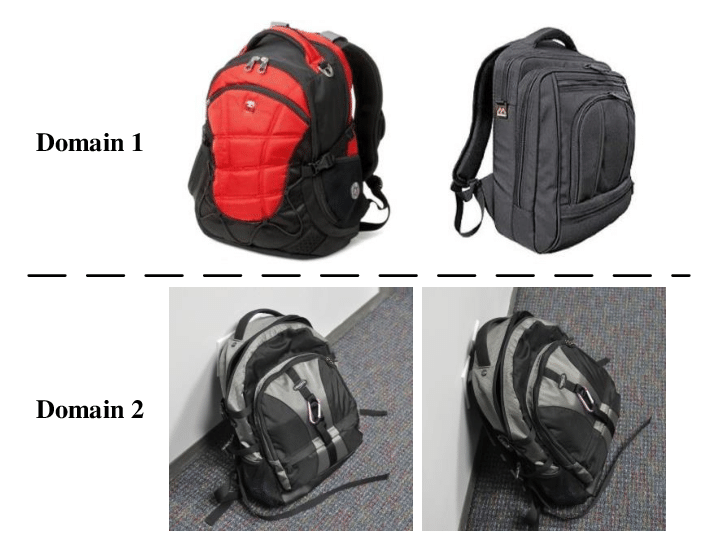}
\caption[Domain Adaptation]{Domain Adaptation: Recognizing Items in Different
  Circumstances is Difficult~\cite{guo2016simple}}\label{fig:adapt}
\end{center}
\end{figure}

The goal of domain adaptation is to compensate for the
variation among two data distributions, to be able to reuse
information  from a source domain on a  different target
domain~\cite{tommasi2016learning}
, see
Fig.~\ref{fig:adapt} for a backpack in different circumstances.
As was indicated in Table~\ref{tab:transfer},
domain adaptation applies to the situation where the tasks
are the same  $T_1= T_2$ but the datasets are different $D_1\neq D_2$,
although still somewhat similar. For
example, the task may be to recognize a 
backpack, but in a different orientation,  or to recognize a
pedestrian, but
in different lighting. 

Domain adaptation can be seen as the opposite of
pretraining. Pretraining 
uses the same dataset for a different task, while
domain adaptation adapts to a new dataset for the same task~\cite{carr2018domain}.

In natural language processing, examples of domain shift are an
algorithm that has been trained on news items that is then applied to
a dataset of biomedical
documents~\cite{daume2009frustratingly,sun2016return}, or a spam
filter that is trained on a certain group of email users, which is
deployed to a new target user~\cite{ben2007analysis}. Sudden changes in the environment (pandemics, severe weather) can also upset
machine learning algorithms. 

There are different techniques to overcome domain shift~\cite{csurka2017domain,zhang2019transfer,xu2020transfer,weiss2016survey}. In visual
applications, adaptation can be achieved by 
re-weighting  the samples of the first dataset,
or clustering them  for visually coherent sub-domains. Other
approaches try to find transformations
that map the source distribution to the target, or  learn  a classification 
model and a feature transformation jointly~\cite{tommasi2016learning}. 
Adversarial techniques where feature representations are encouraged to be
difficult to distinguish can be used to achieve
adaptation~\cite{tzeng2017adversarial,wulfmeier2017addressing,du2020dual},
see also Sect.~\ref{sec:gan}.

\subsection{Meta-Learning}\index{meta learning}\index{in-distribution
  learning}

Related to transfer learning is meta-learning.
Where the focus in 
transfer learning  is on transferring  parameters from a single donor
task to 
a receiver  task for further finetuning, in meta-learning the focus is on
using the knowledge of a number of tasks to learn how to learn a new
task faster and better.
Regular machine learning learns examples \emph{for one}  task, meta-learning
aims to learn \emph{across} tasks. Machine learning learns parameters
that approximate the function, meta-learning learns \emph{hyper}parameters \emph{about} the
learning-function.\footnote{Associating  base learning with
   parameter learning, and meta-learning wth hyperparameter learning
  appears to give us a clear distinction; however, in practice the distinction is not so
  clear cut: in deep meta-learning the initialization of the
  regular parameters is considered to be an important ``hyper''parameter.} This is
often expressed as \emph{learning 
  to learn}, a phrase that has defined the field ever since its
introduction~\cite{schmidhuber1987evolutionary,thrun1996learning,thrun2012learning}.
The
term \emph{meta-learning} has been used for many different contexts,
not only for deep learning, but also for tasks ranging from
hyperparameter optimization, algorithm selection, to automated machine
learning. (We will briefly look at these  in
Sect.~\ref{sec:automl}.) 

Deep meta reinforcement learning is an active area of research. Many algorithms are
being developed, and much progress is being made. Table~\ref{tab:meta}
lists nine algorithms that have been proposed for deep meta
reinforcement learning (see  the surveys~\cite{hospedales2020meta,huisman2021survey}).

\subsubsection{Evaluating Few-Shot Learning  Problems}\index{support
  set}\index{query set}\index{N-way-k-shot learning}\index{few-shot learning}

One of the challenges of  lifelong machine  learning is  to judge
the performance of an algorithm. Regular train-test generalization does
not capture the speed of adaptation of a meta-learning algorithm.

For this reason, meta-learning tasks are typically evaluated on their
\gls{few-shot learning} ability. In few-shot learning, we test if a learning
algorithm can be made to recognize examples from classes from which
it has seen only few examples  in training. In few-shot learning 
prior knowledge is available in the network.

To translate few-shot
learning to a human setting, we can think of a situation where a human
plays the double  bass after only a few minutes of training on the
double bass, but 
after years on the violin, viola, or cello.

Meta-learning algorithms are often evaluated with few
shot learning tasks, in which the algorithm must recognize items of which it
has only seen a few examples. This is formalized in the
$N$-way-$k$-shot
approach~\cite{chen2019closer,lake2011one,wang2020generalizing}.
Figure~\ref{fig:nway} illustrates this process. Given a large dataset $\mathcal{D}$, a smaller
training dataset $D$ is sampled from this dataset. The $N$-way-$k$-shot classification problem
constructs  training dataset $D$ such that it consists of $N$ classes, of
which, for each class, $k$ examples are present in the dataset. Thus,
the cardinality of $|D|=N \cdot k$. 

A full $N$-way-$k$-shot few-shot learning meta task $\mathcal{T}$
consists of many episodes in which base tasks $\mathcal{T}_i$ are
performed. A base task consists of a
training set and a test set, to test generalization. In few-shot
terminology, the training set is called the support set, and the test
set is called the query set. The support set has size $N\cdot k$, the
query set consists of a small number of examples. The meta-learning algorithm can learn
from the episodes of $N$-way-$k$-shot query/support base tasks, until
at meta-test time the generalization of the meta-learning algorithm is
tested with another query, as is illustrated in the figure.

\begin{figure}[t]
\begin{center}
\includegraphics[width=\textwidth]{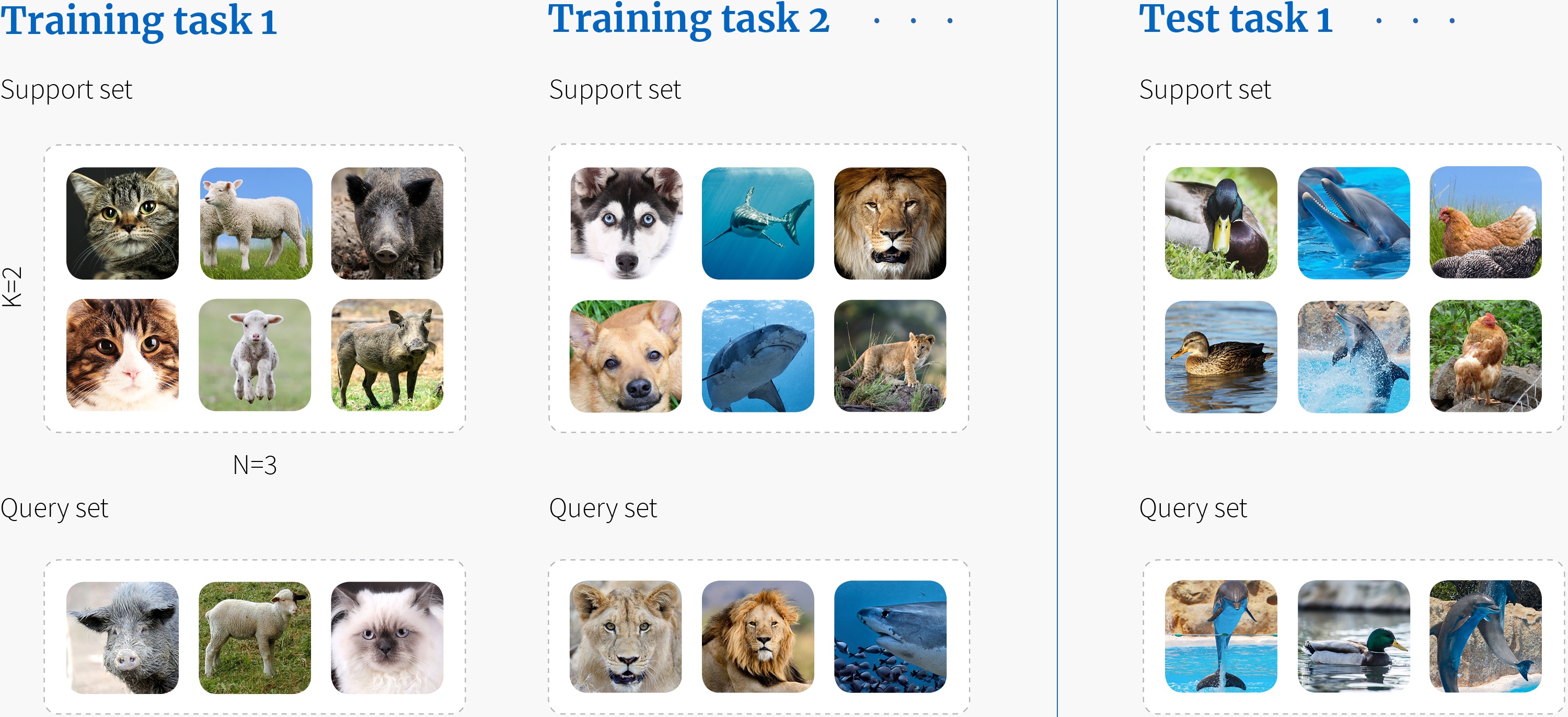}
\caption{$N$-way-$k$-shot learning~\cite{borealis2019}}\label{fig:nway}
\end{center}
\end{figure}

\begin{table}[t]
  \begin{center}
  \begin{tabular}{llll}
    {\bf Name} & {\bf Approach}  & {\bf Environment} & {\bf Ref}  \\
    \hline\hline
    Recurr. ML & Deploy recurrent networks on RL problems & - & \cite{duan2016rl,wang2016learning}\\
    Meta Netw. & Fast reparam. of base-learner by distinct meta-learner &  O-glot, miniIm.& \cite{munkhdalai2017meta}\\
    SNAIL & Attention mechanism coupled with temporal conv. &  O-glot, miniIm.  & \cite{mishra2017simple}\\
    LSTM ML & Embed base-learner parameters in cell state of LSTM&  miniImageNet & \cite{ravi2016optimization} \\
    MAML & Learn initialization weights $\theta$ for fast adaptation &  O-glot, miniIm. & \cite{finn2017model}\\
    iMAML & Approx. higher-order gradients, indep. of optim. path &  O-glot, miniIm. & \cite{rajeswaran2019meta}\\
    Meta SGD & Learn both the initialization and updates &  O-glot, miniIm. & \cite{li2017meta}\\
    Reptile & Move init. towards task-specific updated
              weights & O-glot, miniIm.& \cite{nichol2018first}\\
    BayesMAML & Learn multiple initializations $\Theta$, jointly optim.  SVGD & miniImagaNet &\cite{yoon2018bayesian}\\
    \hline
  \end{tabular}
  \caption{Meta Reinforcement Learning Approaches \cite{huisman2021survey}}\label{tab:meta}
\end{center}
\end{table}

\subsubsection{Deep Meta-Learning Algorithms}






We will now turn our attention to deep meta-learning algorithms. 
The meta-learning field is still young and active, 
nevertheless,
the field is converging on a set of definitions that we will present
here. We start our explanation in a supervised setting.

Meta-learning is concerned with learning a task $\mathcal{T}$ from a set of
base-learning tasks $\{\mathcal{T}_1,\mathcal{T}_2,\mathcal{T}_3,\dots\}$
so that a new (related) meta-test-task will reach a high accuracy
quicker. Each  
base-learning task \gls{task} consists of a dataset
$D_i$  and a learning 
objective, the loss function $\mathcal{L}_i$. Thus we get $\mathcal{T}_i=(D_i,
\mathcal{L}_i)$. Each dataset consists of pairs of inputs and labels
$D_i=\{(x_j,y_j)\}$, and is split into a training and a testset $D_i=\{D_{{\mathcal{T}_i},train},D_{{\mathcal{T}_i},test}\}$. On each training dataset a parameterized model
$\hat{f}_{\theta_i}(D_{i,{train}})$ is approximated, with a loss function
$\mathcal{L}_i(\theta_i,D_{i,{train}})$.  The model $\hat{f}$ is approximated
with a deep  
learning algorithm, that is governed by a set of hyperparameters
\gls{omega-hyp}. The particular hyperparameters vary from algorithm to
algorithm, but frequently encountered hyperparameters are the learning
rate $\alpha$,  the initial  parameters $\theta_0$, and  algorithm
constants. 
 
This conventional machine learning algorithm is called the \emph{base
  learner}. Each base learner task approximates a model $\hat{f}_i$ by
finding the optimal parameters $\theta_i^\star$ to
minimize the loss function on its data set  
$$\mathcal{T}_i = \hat{f}_{\theta_i^\star}=\argmin_{\theta_i}\mathcal{L}_{i,\omega}(\theta_i,D_{i,train})$$
while the learning algorithm is governed by hyperparameters
$\omega$. 

\subsubsection*{Inner and Outer Loop Optimization}
One of the most popular deep meta-learning approaches of the last few 
years is optimization-based meta-learning~\cite{huisman2021survey}.
This approach
optimizes the initial  parameters 
$\theta$ of the network for fast learning of new tasks. Most 
optimization-based techniques do so 
by approaching meta-learning as a two-level optimization problem. At
the \emph{inner} level, a base learner makes task-specific updates to
$\theta$ for the different observations in the training set. At the
\emph{outer} level, the meta-learner optimizes hyperparameters 
$\omega$ across 
a sequence of base tasks where the loss of each
task is evaluated using the \emph{test} data from the base tasks
$D_{{\mathcal{T}_i},test}$~\cite{ravi2016optimization,hospedales2020meta,lee2018gradient}. 


The inner loop optimizes the parameters $\theta$, and the outer loop
optimizes the hyperparameters $\omega$ to find the best performance on
the set of base tasks $i=0, \ldots, M$ with the appropriate test data:
$$ \omega^\star=\underbrace{\argmin_\omega
  \mathcal{L}^{\mbox{meta}}}_{\textrm{outer loop}} (\underbrace{\argmin_{\theta_i}
\mathcal{L}^{\mbox{base}}_\omega (\theta_i,D_{i,train})}_{\textrm{inner loop}},D_{i,test}) .$$
The inner loop optimizes $\theta_i$ within the datasets $D_i$ of the
tasks $\mathcal{T}_i$ , and the
outer loop optimizes $\omega$ across the tasks and datasets.

The meta loss function
optimizes for the meta objective, which can be accuracy, speed, or
another goal over the set of base tasks (and datasets). The outcome of the meta
optimization is a set of optimal hyperparameters $\omega^\star$.

In optimization-based meta-learning  the most important hyperparameters
$\omega$
are the  optimal initial
parameters $\theta_0^\star$. When the meta-learner   only optimizes  the
initial parameters as hyperparameters ($\omega=\theta_0$)  then the
inner/outer formula simplifies as follows:
$$ \theta_0^\star=\underbrace{\argmin_{\theta_0}
  \mathcal{L}}_{\textrm{outer loop}} (\underbrace{\argmin_{\theta_i}
\mathcal{L}
(\theta_i,D_{i,train})}_{\textrm{inner loop}},D_{i,test}) .$$ In this
approach we meta-optimize the initial parameters $\theta
_0$, such that the loss function  performs well on the \emph{test} data of the base
tasks. Section~\ref{sec:maml} describes MAML, a well-known example of
this approach.
 

Deep meta-learning approaches are sometimes categorized as  (1)
similarity-metric-based, (2) model-based, and (3) optimization-based~\cite{huisman2021survey}. 
We will now have a closer look at two of the nine meta reinforcement
learning algorithms from Table~\ref{tab:meta}. We will look 
at Recurrent meta-learning and MAML; the former is a model-based
approach, the latter optimization-based.

\subsubsection{Recurrent Meta-Learning}
For meta reinforcement learning approaches to be able to learn to learn, they must
be able to remember what they have learned  across subtasks. Let us
see how  Recurrent meta-learning learns across tasks.

\begin{figure}[t]
    \centering
    \includegraphics[width=\linewidth]{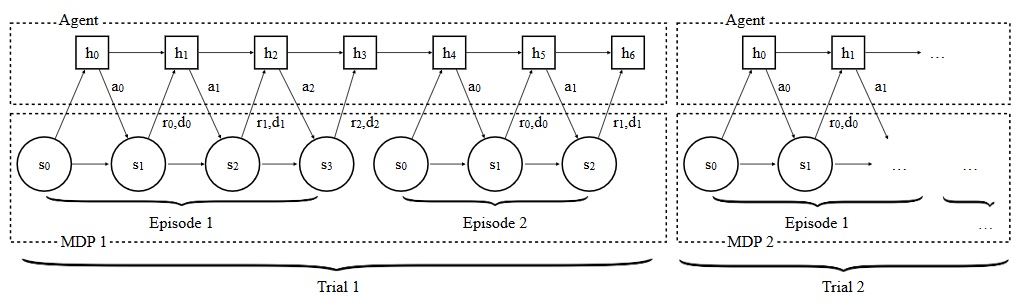}
    \caption{Workflow of recurrent meta-learners in reinforcement
      learning contexts. State, action, reward, and termination flag at time
      step $t$ are denotednby $s_{t}, a_t, r_{t},$ and $d_{t}$, $h_{t}$
      refers to the hidden state \cite{duan2016rl}.} 
    \label{fig:rlRNN}
\end{figure}

Recurrent meta-learning uses recurrent neural networks to
remember this knowledge~\cite{duan2016rl, wang2016learning}. 
The recurrent network serves as dynamic storage for the learned task
embedding (weight vector). 
The recurrence can be implemented by an LSTM \cite{wang2016learning}
or by  gated recurrent units \cite{duan2016rl}. The choice of recurrent
neural meta-network (meta-RNN) determines how well it adapts
to the subtasks, as it
gradually accumulates knowledge about the base-task structure.

Recurrent meta-learning 
 tracks variables $s,a,r,d$ which denote state, action,
reward, and termination of the episode.
For each 
task $\mathcal{T}_{i}$, Recurrent meta-learning inputs the set of 
environment variables $\{s_{t+1},a_{t},r_{t},d_{t}\}$ into a meta-RNN at
each time step $t$.  The
meta-RNN outputs an action and a hidden state $h_t$. Conditioned on the hidden
state $h_{t}$, the meta network outputs  action $a_{t}$. The goal is to
maximize the expected reward in each trial (Fig.~\ref{fig:rlRNN}).
Since Recurrent meta-learning  embeds information from
previously seen inputs in  hidden state, it is regarded as a model-based
meta-learner~\cite{huisman2021survey}.  

Recurrent meta-learners performed almost as well as model-free baselines
on simple $N$-way-$k$-shot reinforcement learning tasks
\cite{wang2016learning, duan2016rl}. However, the performance 
degrades in more complex problems, when dependencies 
span a longer horizon.

\subsubsection{Model-Agnostic Meta-Learning}\index{MAML}\label{sec:maml}
The Model-agnostic meta-learning approach (\gls{MAML})~\cite{finn2017model}  is an optimization approach that is model-agnostic:
it can be used for different learning problems, such as
classification, regression, and reinforcement learning.

\begin{figure}[t]
\begin{center}
\includegraphics[width=5.5cm]{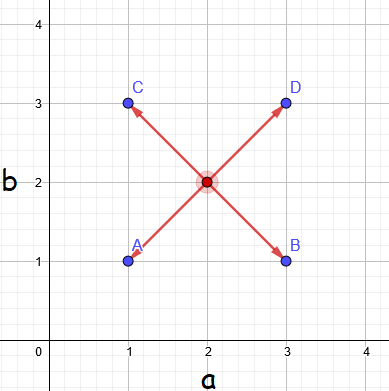}
\caption{The Optimization approach aims to learn parameters from which
  other tasks can be learned quickly. The intuition behind
  Optimization approaches such as MAML is that when our meta-training set
  consists of tasks A, B, C, and D, then if our meta-learning
  algorithm adjusts parameters $a$ and $b$ to $(2, 2)$, then they 
  are close to either of the four tasks, and can be adjustly quickly
  to them, with few examples~(after \cite{huisman2021survey,finn2017model}).}\label{fig:four}
\end{center}
\end{figure}

As mentioned, the optimization view of meta-learning is especially
focused on optimizing the initial parameters $\theta$. The intuition behind the
optimization view can be illustrated with a simple regression example (Fig.~\ref{fig:four}). Let us 
assume that  we are faced with multiple linear regression problems
$f_i(x)$. The  model has two parameters: $a$ and $b$,
$\hat{f}(x) = a\cdot x + b$. When the meta-training set consists of
four tasks, A, B, C, and D, then we wish to optimize to a single set
of parameters $\{a, b\}$ from which we can quickly learn the optimal
parameters for each of the four tasks. In Fig.~\ref{fig:four} the
point in the middle represents this combination of
parameters. The point is the closest to the four different
tasks. This is how  
Model-agnostic meta-learning works~\cite{finn2017model}: by exposing
our model to various base tasks, we  update the parameters 
$\theta=\{a,b\}$ to  good initial parameters $\theta_0$ that facilitate
quick meta-adaptation.


Let us look at the process of training a deep learning model’s parameters from a
feature learning
standpoint~\cite{finn2017model,huisman2021survey}, where the goal is  that a
few gradient steps can produce good results on a
new task.  We build a feature representation that is broadly suitable for
many tasks, and by then
fine-tuning the parameters slightly (primarily
updating the top layer weights) we achieve 
good results---not unlike transfer learning. MAML finds parameters $\theta$ that are
easy and fast to finetune, allowing the adaptation to happen in an
embedding space that is well suited for fast learning. To put it
another way, MAML's goal is to  find the
point in the middle of Fig.~\ref{fig:four}, from where the other tasks are
easily reachable.

\newcommand{\task}{\mathcal{T}}
\newcommand{\loss}{\mathcal{L}}
\newcommand{\inp}{\mathbf{x}}
\newcommand{\learner}{\pi}
\newcommand{\lossi}{\loss_{\task_i}}
\newcommand{\action}{a}
\newcommand{\reward}{R}

How does MAML
work~\cite{huisman2020deep,raghu2019rapid,finn2017model}? Please 
refer to the pseudocode for  MAML in Alg.~\ref{alg:mamlrl}. The
learning task is an episodic  Markov decision process
 with horizon $T$, where the learner is allowed to query a
limited number of sample trajectories for few-shot learning.
Each reinforcement learning task $\task_i$ contains an initial state distribution
$p_i(s_1)$ and a transition distribution $p_i(s_{t+1}|s_t,\action_t)$.
 The loss $\lossi$ corresponds to the (negative) reward function
$\reward$.  
The model
being learned, $\learner_\theta$, is a policy  from states
$s_t$ to a distribution over actions $\action_t$ at each timestep $t
\in \{1,...,T\}$. The loss for task $\task_i$ and policy
$\learner_\theta$ takes the familiar form of the objective (Eq.~\ref{eq_rl_objective}):
\begin{align}
\label{eq:rl}
\lossi( \learner_\theta) = - \mathbb{E}_{s_t, \action_t \sim
  \learner_\theta, p_{\task_i}}\left[ \sum_{t=1}^T \reward_i(s_t,
  \action_t, s_{t+1})  \right].
\end{align}
In $k$-shot reinforcement learning, $k$ rollouts from $\learner_\theta$ and
task $\task_i$, $(s_1,\action_1,...s_T)$, and the 
rewards $\reward(s_t, \action_t)$, may be used for adaptation on a
new task $\task_i$.  MAML uses  TRPO
to estimate the gradient both for the policy gradient update(s) and the
meta optimization~\cite{schulman2015trust}.

\begin{algorithm}[t]
\caption{MAML for Reinforcement Learning~\cite{finn2017model}}
\label{alg:mamlrl}
\begin{algorithmic}
{\footnotesize
\Require $p(\task)$: distribution over tasks
\Require $\alpha$, $\beta$: step size hyperparameters
\State randomly initialize $\theta$
\While{not done}
\State Sample batch of tasks $\task_i \sim p(\task)$
  \ForAll{$\task_i$}
      \State Sample $k$ trajectories $\mathcal{D}=\{(s_1,\action_1,...s_T)\}$ using $f_\theta$ in $\task_i$
      \State Evaluate $\nabla_\theta \lossi(\learner_\theta)$ using $\mathcal{D}$ and $\loss_{\task_i}$ in Equation~\ref{eq:rl}
      \State Compute adapted parameters with gradient descent: $\theta_i'=\theta-\alpha \nabla_\theta  \loss_{\task_i}(  \learner_\theta )$
      \State Sample trajectories $\mathcal{D}_i'=\{(s_1,\action_1,...s_T)\}$ using $f_{\theta_i'}$ in $\task_i$ 
 \EndFor
 \State Update $\theta \leftarrow \theta - \beta \nabla_\theta \sum_{\task_i \sim p(\task)}  \loss_{\task_i} ( \learner_{\theta_i'})$ using each $\mathcal{D}_i'$ and $\loss_{\task_i}$ in Eq.~\ref{eq:rl}
\EndWhile
}
\end{algorithmic}
\end{algorithm}

The goal is to quickly learn new 
concepts, which is equivalent to achieving a minimal loss in few
gradient update steps. The number of gradient steps  has to be
specified in advance.
For a single
gradient update step 
gradient
descent  produces updated parameters 
\begin{align*}
\theta_i'=\theta-\alpha \nabla_\theta  \loss_{\task_i}(  \learner_\theta )
\end{align*}
specific to task $i$. The meta loss of one gradient step across tasks is
\begin{align}
  \label{eq:metaobj}
  \theta \leftarrow \theta - \beta \nabla_\theta \sum_{\task_i \sim
  p(\task)}  \loss_{\task_i} ( \learner_{\theta_i'})
\end{align} where $p(\Tau)$ is a probability distribution over
tasks. This expression contains an inner gradient
$\nabla_{\boldsymbol{\theta}}
\mathcal{L}_{\Tau_{i}}(\pi_{\theta_i'})$. 
Optimizing this meta loss requires 
computing second-order gradients when backpropagating the
meta gradient through the gradient operator in the meta objective (Eq.~\ref{eq:metaobj}), which is computationally
expensive~\cite{finn2017model}. Various  algorithms have been 
inspired by MAML and aim to improve optimization-based meta-learning
further~\cite{nichol2018first,huisman2020deep}. 

In meta reinforcement learning the goal  is
to  quickly find a policy for a new environment
using only a small amount of experience. 
MAML has gained attention within the field of deep
meta-learning,  due to its simplicity (it requires two
hyperparameters), its general applicability, and its strong
performance. 

\subsubsection{Hyperparameter Optimization}\index{hyperparameter optimization}\index{inductive bias}\label{sec:automl}

Meta-learning has been around for a long time, long before deep
learning became popular. It has been applied to
classic machine learning tasks, such as regression, decision trees, support
vector machines, clustering algorithms, Bayesian networks, evolutionary
algorithms, and local
search~\cite{bishop2006pattern,brazdil2008metalearning,vilalta2002perspective}.
The hyperparameter view on meta-learning originated here.

Although this is a book about deep learning, it is interesting to
briefly discuss this non-deep background, also because
hyperparameter optimization is an important technology to find a
good set of hyperparameters in reinforcement learning experiments. 

Machine
learning algorithms have hyperparameters that govern their
behavior, and 
finding the optimal setting for these
hyperparameters has long been called meta-learning. A naive approach
is to enumerate all  combinations and run the machine learning problem
for them.  For all but the smallest hyperparameter spaces such a grid
search will be  prohibitively slow.  Among the smarter
meta-optimization approaches are random search, Bayesian optimization,
gradient-based optimization, and evolutionary optimization.

This  meta-algorithm  approach has given
rise to algorithm configuration research, such as
SMAC~\cite{hutter2011sequential},\footnote{\url{https://github.com/automl/SMAC3}}
ParamILS~\cite{hutter2009paramils},\footnote{\url{http://www.cs.ubc.ca/labs/beta/Projects/ParamILS/}}
irace~\cite{lopez2016irace},\footnote{\url{http://iridia.ulb.ac.be/irace/}}\index{SMAC}\index{SATzilla}\index{ParamILS}\index{irace} and algorithm
selection research~\cite{rice1976algorithm,kerschke2019automated},
such as
SATzilla~\cite{xu2008satzilla}.\footnote{\url{http://www.cs.ubc.ca/labs/beta/Projects/SATzilla/}}
Well-known hyperparameter optimization packages include
scikit-learn~\cite{pedregosa2011scikit},\footnote{\url{https://scikit-learn.org/stable/}}
scikit-optimize,\footnote{\url{https://scikit-optimize.github.io/stable/index.html}}
 nevergrad~\cite{nevergrad},\footnote{\url{https://code.fb.com/ai-research/nevergrad/}}
 and
 optuna~\cite{akiba2019optuna}.\footnote{\url{https://optuna.org}}\index{scikit-learn}\index{scikit-optimize}\index{nevergrad}\index{optuna}
Hyperparameter optimization and algorithm configuration have grown
into the field of automated machine learning, or AutoML.\footnote{\url{https://www.automl.org}} The AutoML field is a
large and active field of meta-learning, an overview book is~\cite{hutter2019automated}.\index{AutoML}

All machine learning algorithms have a bias, different algorithms
perform better on different types of problems.
Hyperparameters constrain this algorithm bias. This so-called
inductive bias reflects the set of assumptions about the data on
which the algorithms  are based. Learning algorithms  perform better when this
bias matches the learning problem (for example: CNNs work
well on image problems, RNNs on language problems).
 Meta-learning changes
this inductive bias,  either by choosing a different  learning algorithm, by
changing the network initialization, or by other
means, allowing the algorithm to be adjusted to work well on different
problems.

\subsubsection{Meta-Learning and Curriculum Learning}
There is in interesting connection between meta-learning and
curriculum learning. Both approaches aim to improve the speed and
accuracy of
learning, by learning from a set of subtasks.

In meta-learning, knowledge is gained from subtasks, so that  the learning of a
new, related, task, can  be quick.
In curriculum learning (Sect.~\ref{sec:curriculum}) we aim to learn
quicker by dividing a large and difficult learning task into a set of
subtasks, ordered from easy to hard.

Thus we can conclude that curriculum
learning is a form of meta-learning where the subtasks are ordered from easy to
hard, or, equivalently, that meta-learning is unordered curriculum
learning.

\subsubsection{From Few-Shot  to Zero-Shot
  Learning}\label{sec:zsl}\index{zero-shot learning}\index{few-shot learning}
Meta-learning  uses information from previous learning
tasks to learn new tasks quicker~\cite{thrun2012learning}. Meta-learning algorithms are often evaluated in a few-shot setting, to see
how well they do in image classification problems when they are shown
only few training examples. This
\emph{few-shot learning} problem aims to correctly classify queries with  little
previous support for the new class. In the previous sections we have
discussed how meta-learning algorithms aim to achieve few-shot
learning. A discussion of meta-learning would not be complete without
mentioning \gls{zero-shot learning}.

Zero-shot learning (\gls{ZSL}) goes a step further than  few-shot learning. In
zero-shot learning an example has to be  recognized  as belonging to 
a class without ever having been trained on  an example of this
class~\cite{larochelle2008zero,palatucci2009zero}.
In zero-shot learning  
the classes covered by training instances and the classes we aim to
classify are disjoint. This  may sound like an impossibility---how can you recognize something you have never seen before?---yet it is
something that we, humans, do all the time: having learned
to pour coffee in a cup, we can also pour tea, even if we never have
seen tea  before. (Or,  having learned
to play the violin, the viola, and the cello, we can play
the double bass, to some degree, even if we have never played the
double bass before.)

Whenever we recognize something that we have  not  seen before,
we are actually using extra information (or features). If we recognize a red beak in a
picture of a bird-species that we have never seen before, then the
concepts ``red'' and ``beak'' are known to us, because we have
learned them in other contexts. 


Zero-shot
learning recognizes new categories of instances without training
examples. Attribute-based zero-shot learning uses  separate high-level attribute
descriptions of the new categories, based on  categories previously
learned in the dataset. 
Attributes are an intermediate representation that enables
parameter sharing between
classes~\cite{akata2013label}.
The  extra information can be in the form of
textual description of the class---red, or beak---in addition to the visual
information~\cite{lampert2009learning}. The  learner must  be able to match text
with image information.

Zero-shot learning
approaches are designed to learn this intermediate semantic layer,
the attributes, and apply them at inference time to predict new
classes, when they are provided with  descriptions in terms of these attributes.
Attributes correspond to
high-level properties of the objects which are shared across multiple
classes (which can be detected by machines and which can be understood
by humans). Attribute-based image classification is
a label-embedding problem where each class is embedded in the space of
attribute vectors.
As an example, if the classes correspond to animals,
possible attributes include \emph{has paws}, \emph{has stripes} or \emph{is
black.}

\section{Meta-Learning Environments}
Now that we have seen how transfer learning and meta-learning can be implemented, it is time to
look at  some of the
environments that are used to evaluate the  algorithms. We will list
important datasets, environments, and foundation models for images,
behavior, and text, expanding our scope beyond pure reinforcement learning.
We will look at how well the  approaches succeed in
generalizing quickly to new  machine learning tasks.

Many benchmarks have been introduced to test transfer and
meta-learning algorithms. Benchmarks for conventional machine 
learning algorithms aim to offer a variety of \emph{challenging} learning
tasks. Benchmarks for meta-learning, in contrast, aim to offer \emph{related} learning
tasks. Some benchmarks are parameterized, where the difference
between tasks can be controlled.

Meta-learning aims to learn new and related tasks quicker, trading off 
speed versus accuracy.
This raises the question  how fast and how accurate  meta-learning algorithms are
under different circumstances. In answering these questions we must
keep in mind that the closer the learning tasks are, the easier the
task is, and the quicker and the more accurate results will
be. Hence, we should carefully look at which dataset a benchmark uses
when we compare results.

\begin{table}[t]
  \begin{center}
  \begin{tabular}{lccc}
    {\bf Name} & {\bf Type} & {\bf Domain} & {\bf Ref}  \\
    \hline\hline
    ALE &  single  & games & \cite{bellemare2013arcade}  \\
    MuJoCo &  single  & robot & \cite{todorov2012mujoco}  \\
    DeepMind Control & single & robot & \cite{tassa2018deepmind} \\ \hline
    BERT &  transfer & text & \cite{devlin2018bert} \\
    GPT-3 &  transfer & text & \cite{radford2018improving} \\ \hline
    ImageNet &transfer& image & \cite{fei2009imagenet}\\
    Omniglot & meta & image & \cite{lake2011one} \\
    Mini-ImageNet & meta & image & \cite{vinyals2016matching} \\
    Meta-Dataset & meta & image & \cite{triantafillou2019meta} \\
    Meta-World & meta  & robot & \cite{yu2020meta} \\
    Alchemy & meta & unity & \cite{wang2021alchemy} \\
    \hline
  \end{tabular}
  \caption[Meta-Learning Datasets and Environments]{Datasets, Environments, and Models for Meta-Learning in
    Images,  Behavior, and Text}\label{tab:meta-env}
\end{center}
\end{table}

Table~\ref{tab:meta-env} lists some of the environments that are often
used for meta-learning experiments. Some are regular deep learning
environments designed for single-task learning (``single''), some are transfer learning and pretraining
datasets (``transfer''), and some datasets and environments are
specifically designed for meta-learning experiments (``meta''). 

We will now describe them in more detail.
%
ALE (Sect.~\ref{sec:ale}), MuJoCo (Sect.~\ref{sec:mujoco}) and the 
DeepMind control suite (Sect.~\ref{sec:dcs}) are originally single
task deep learning environments. They are also being used in
meta-learning  experiments and few-shot learning, often with moderate
results, since the tasks are typically not very similar (Pong is not
like Pac-Man).

\subsection{Image Processing}\index{Omniglot}
Traditionally, two datasets have emerged as de facto benchmarks for
few-shot image
learning: Omniglot~\cite{lake2015human}, and
mini-ImageNet~\cite{russakovsky2015imagenet,vinyals2016matching}.

Omniglot is a dataset for one-shot learning. This dataset contains
1623 different handwritten characters from 50 different
alphabets
and contains 20 examples per class
(character)~\cite{lake2015human,lake2019omniglot}. Most recent 
methods obtain very high accuracy on Omniglot, rendering 
comparisons between them mostly uninformative.

Mini-ImageNet uses the same setup as Omniglot for testing, consisting of
60,000 colour images of size $84\times 84$ with 100 classes
(64/16/20 for
train/validation/test) and contains 600 examples per class~\cite{vinyals2016matching}. Albeit
harder than Omniglot,  most recent
methods achieve similar accuracy when controlling for
model capacity. Meta-learning algorithms such as Bayesian Program
Learning and MAML achieved accuracies
comparable to human performance on Omniglot and ImageNet, with
accuracies in the high nineties and error rates as low
as a few percent~\cite{lake2019omniglot}. Models trained on the
largest datasets, such as ImageNet, are used as foundation
models~\cite{bommasani2021opportunities}. Pre-trained models can be
downloaded from the Model zoo.\footnote{\url{https://modelzoo.co}}

These benchmarks  may be too homogeneous for testing meta-learning. In contrast,
real-life learning experiences are heterogeneous: they vary in terms
of the number of classes and examples per class, and are
unbalanced. Furthermore, the Omniglot and Mini-ImageNet benchmarks
 measure  within-dataset 
generalization. For meta-learning, we are eventually after models that
can generalize to entirely new distributions. For this reason, new
datasets are being developed specifically for meta-learning. 

\subsection{Natural Language Processing}\index{BERT}\index{GPT-3}\index{transformers}
In natural language processing BERT is a well-known pretrained model.
\gls{BERT} stands for Bidirectional encoder representations from 
transformers~\cite{devlin2018bert}. It is designed to pretrain deep bidirectional
representations from unlabeled text by jointly conditioning on two
contexts. 
BERT has shown that transfer learning  can 
work well in natural language tasks.
BERT can be used for classification tasks such as sentiment analysis,
question answering tasks, and named entity recognition. BERT is a
large model, with 345 million
parameters~\cite{rothman2021transformers}.

An even larger pretrained transformer model is the Generative
pretrained transformer 3, or 
\gls{GPT-3}~\cite{radford2018improving}, with 175 billion parameters.  The quality of the text
generated by GPT-3 is exceptionally good, it is difficult to distinguish from that written by
humans.  In
this case, it appears that  size matters.  OpenAI provides a public interface where you can see for
yourself how well it
performs.\footnote{\url{https://openai.com/blog/openai-api/}}

BERT and GPT-3 are large models, that are used more and more as a
foundation model as a basis for other experiments to pretrain on.

\subsection{Meta-Dataset}\index{meta dataset benchmark}

A recent dataset specifically designed 
for meta-learning is Meta-dataset~\cite{triantafillou2019meta}.\footnote{\url{https://github.com/google-research/meta-dataset}} 
Meta-dataset is a 
set of datasets, consisting of: Imagenet, Omniglot,
Aircraft, Birds, Textures, Quick Draw, Fungi, Flower, Traffic Signs,
and MSCOCO. 
Thus, the datasets provide a  more heterogeneous
challenge than earlier single-dataset experiments. 
 
Triantafillou et al.\ report results with Matching networks~\cite{vinyals2016matching},
Prototypical networks~\cite{snell2017prototypical}, first-order MAML~\cite{finn2017model},
and Relation networks~\cite{sung2018learning}.
As can be expected, the accuracy  on the larger (meta) dataset is much
lower than on previous homogeneous datasets.
They
find that a variant of MAML performs best, although for classifiers that are trained on other datasets most
accuracies are between 40\% and 60\%, except for Birds and Flowers,
which scores in the 70s and 80s, closer to the single dataset results
for Omniglot and Imagenet.
Meta-learning for heterogeneous datasets remains a challenging task 
(see also~\cite{chen2019closer,tian2020rethinking}).

\begin{figure}[t]
  \begin{center}
    \includegraphics[width=\textwidth]{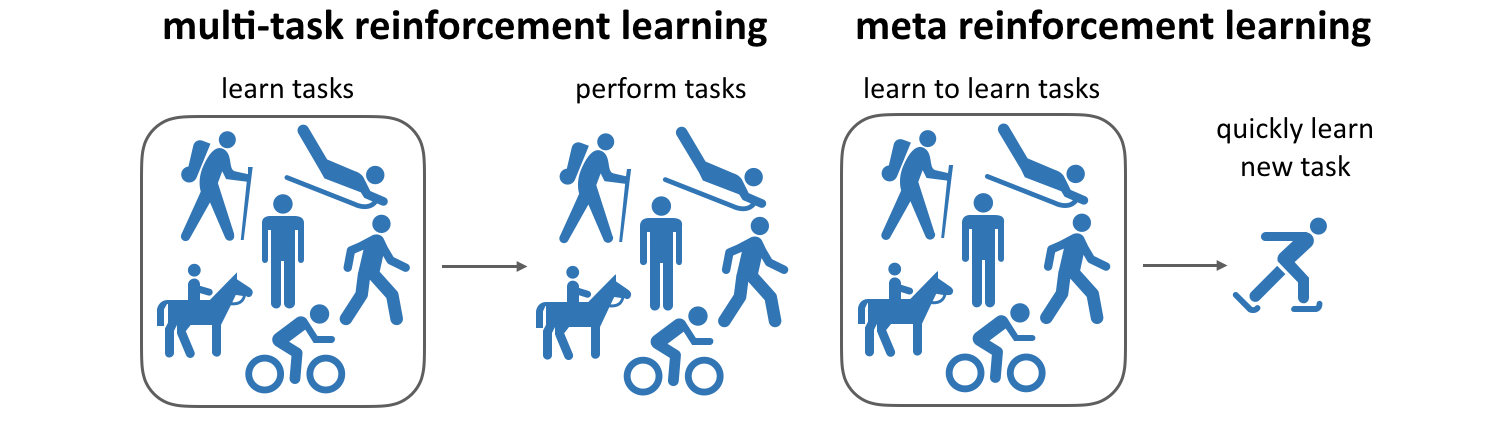} 
 \caption{Multi Task and Meta Reinforcement Learning~\cite{yu2020meta}}\label{fig:multimeta}
 \end{center}
\end{figure}

\begin{figure}[t]
  \begin{center}
    \includegraphics[width=\textwidth]{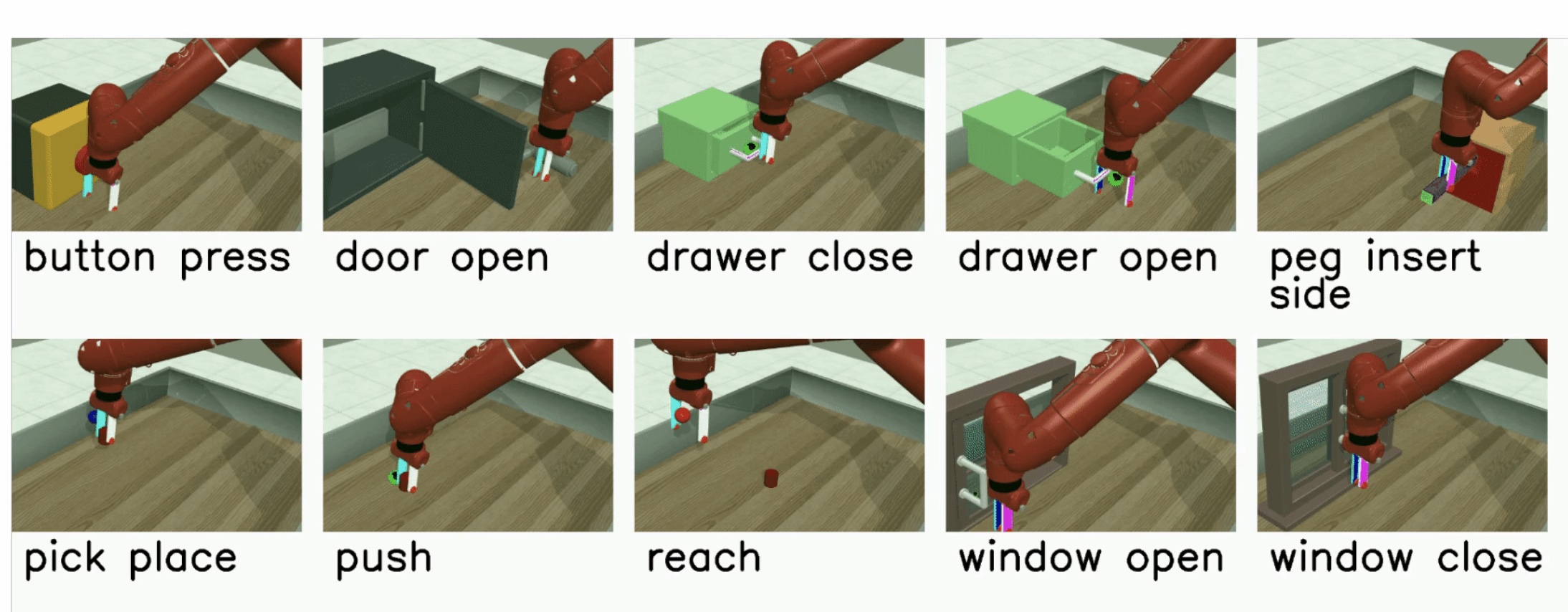}
 \caption{Meta-World Tasks~\cite{yu2020meta}}\label{fig:metaworld}
 \end{center}
\end{figure}

\subsection{Meta-World}\index{meta world benchmark}\label{sec:metaworld} 
For deep reinforcement learning  two traditionally popular environments are
ALE  and MuJoCo. The games in the ALE
benchmarks typically
differ considerably, which makes the ALE test set  challenging for
meta-learning, and  little success has been reported. (There are a
few exceptions that apply transfer learning (pretraining) to
DQN~\cite{parisotto2015actor,mittel2019visual,sobol2018visual}
succeeding in multitask
learning 
in a set of  Atari games that all move a ball.)

Robotics tasks, on the other hand, are more easily parameterizable. Test
tasks can be generated with the desired level of similarity, making
robotic tasks amenable to meta-learning testing.  Typical tasks such as reaching and learning
different walking gaits,
are more related than two Atari games such as, for example, Breakout
and Space Invaders.  

To provide a better benchmark that is more challenging  for meta reinforcement learning,  Yu et al.\ introduced Meta-World~\cite{yu2020meta},\footnote{\url{https://meta-world.github.io}} a benchmark  for multi-task and meta
reinforcement learning~(see Fig.~\ref{fig:multimeta} for their
pictorial explanation of the difference between multi-task and meta-learning).
Meta-World   consists of
50 distinct  manipulation tasks with a robotic arm~(Fig.~\ref{fig:metaworld}). The tasks are
designed to be different, and contain structure, which can be
leveraged for transfer to  new tasks.

When the authors of Meta-World
evaluated six state-of-the-art meta-reinforce\-ment  and
multi-task learning algorithms on these tasks, they found that
generalization of existing algorithms to heterogeneous tasks is
limited. They tried PPO, TRPO, SAC, RL$^2$\cite{duan2016rl}, MAML, and
PEARL~\cite{rakelly2019efficient}. Small variations of tasks such as
different object positions can be learned with reasonable success, but
the algorithms struggled to
learn multiple tasks at the same time, even with as few as ten
distinct training tasks. In contrast to more limited meta-learning benchmarks, Meta-World  emphasizes generalization to
 new tasks and interaction scenarios, not 
 just a parametric variation in goals.

\subsection{Alchemy}\index{alchemy benchmark}
A final meta reinforcement learning benchmark that we will discuss is
Alchemy~\cite{wang2021alchemy}. The Alchemy benchmark is  a procedurally generated 3D video game~\cite{shaker2016procedural}, implemented
in Unity~\cite{juliani2018unity}. Task generation is parameterized, and varying degrees of
similarity and hidden structure can be chosen. The process by which Alchemy levels are  created is
accessible to researchers, and  a perfect Bayesian ideal observer can
be implemented.

Experiments with two  agents are reported, VMPO and
IMPALA. The VMPO~\cite{song2019v,parisotto2020stabilizing} agent is based on a
gated transformer
network. The IMPALA~\cite{espeholt2018impala,jaderberg2017population} agent is
based on population-based  training with an LSTM core network. Both
agents are strong deep learning methods, although not necessarily for
meta-learning. Again,
in both agents
meta-learning became more difficult as learning tasks became more
diverse. The reported performance on meta-learning was weak.

The Alchemy platform can be found on
GitHub.\footnote{\url{https://github.com/deepmind/dm_alchemy}} A
human-playable interface is part of the environment.

\subsection{\em Hands-on: Meta-World Example}\index{meta world benchmark}
Let us get experience with running  Meta-World
benchmark environments. We will run a few popular agent algorithms on
them, such as PPO and TRPO. 
Meta-World is as easy to use as Gym. The code can be found on
GitHub,\footnote{\url{https://github.com/rlworkgroup/metaworld}} and
the accompanying implementations of a suite of agent algorithms is
called Garage~\cite{garage}.\footnote{\url{https://github.com/rlworkgroup/garage}}
Garage runs on PyTorch and on TensorFlow.

Standard Meta-World benchmarks include multi-task and meta-learning
setups, named MT1, MT10, MT50, and ML1, ML10, ML50. 
The Meta-World benchmark can be installed  with
pip
\begin{tcolorbox}
  \verb|pip install git+https://github.com/rlworkgroup/|

  \verb|metaworld.git@master\#egg=metaworld|
\end{tcolorbox}
Note that Meta-World is a robotics benchmark and needs MuJoCo, so you
have to install that too.\footnote{\url{https://github.com/openai/mujoco-py\#install-mujoco}}
\lstset{label={lst:metaworld}}
\lstset{caption={Using Meta-World}}
\begin{lstlisting}[language=Python,float]
import metaworld
import random

print(metaworld.ML1.ENV_NAMES)  # Try available environments

ml1 = metaworld.ML1('pick-place-v1') # Construct the benchmark

env = ml1.train_classes['pick-place-v1']()  
task = random.choice(ml1.train_tasks)
env.set_task(task)  # Set task

obs = env.reset()  # Reset environment
a = env.action_space.sample()  # Sample an action
obs, reward, done, info = env.step(a)  # Step the environoment 
\end{lstlisting}
The GitHub site contains brief example instructions on the usage  of
the benchmark, please refer to
Listing~\ref{lst:metaworld}. The benchmark can be used to test the
meta-learning performance of your favorite algorithm, or you can use one
of the baselines provided in Garage.

\subsubsection*{Conclusion}

In this chapter we have seen different approaches to learning new and
different tasks with few or even zero examples. Impressive results
have been achieved, although  major challenges remain to learn
general adaptation when tasks are more diverse. As is often the case
in new fields, many different approaches have been tried. 
The field of meta-learning is an active
field of research, aiming to reduce one of the main problems of
machine learning, and  many new
methods will continue to be developed.

\section*{Summary and Further Reading}
\addcontentsline{toc}{section}{\protect\numberline{}Summary and Further Reading}
We will now summarize the chapter and provide pointers to further reading.
\subsection*{Summary}
This chapter is concerned
with learning new tasks faster and with smaller
datasets or lower sample complexity.    Transfer learning is
concerned with transferring knowledge that 
has been learned to solve a taks, to another task, to allow quicker
learning. A popular transfer learning approach is pretraining, where
 some network layers are copied to intialize a network for a new
task, followed by fine tuning, to improve performance on the new task,
but with a
smaller dataset.

Another approach is meta-learning, or learning to
learn. Here  knowledge of how a sequence of previous tasks is learned
is used to learn a new task quicker. Meta-learning learns
\gls{hyperparameters} of the different tasks. In deep meta-learning, the set of
initial network parameters is usually considered to be such a 
hyperparameter. Meta-learning aims to learn hyperparameters that can
learn a new task with only a few
training examples, often using $N$-way-$k$-shot learning.
For deep few-shot learning
the Model-Agnostic Meta-Learning (MAML) approach is well-known, and has inspired  follow-up work.

Meta-learning is of great importance in machine learning.
For tasks that are
related, good results are reported. For more challenging benchmarks, where
tasks are less related (such as pictures of animals from very different
species),  results are reported that are weaker. 

\subsection*{Further Reading}
Meta-learning is a highly active field of research. Good entry points
are~\cite{zintgraf2019fast,hospedales2020meta,huisman2020deep,huisman2021survey,botvinick2019reinforcement}.  
Meta-learning has attracted much attention in artificial
intelligence, both in supervised  learning and in reinforcement  learning. 
Many books and surveys have been written about the field of meta-learning, see, for example~\cite{brazdil2008metalearning,schweighofer2003meta,vanschoren2018meta,schaul2010metalearning,weng2020meta}. 

There has been
active research interest in meta-learning algorithms for some time,
see, for example 
\cite{schmidhuber1987evolutionary,bengio1990learning,schmidhuber1996simple,vilalta2002perspective}. 
Research into transfer learning and meta-learning has a long history,
starting with Pratt and Thrun~\cite{pratt1993discriminability,thrun2012learning}.
Early surveys into the field
are~\cite{pan2010survey,taylor2009transfer,weiss2016survey}, more
recent surveys are~\cite{zhang2019transfer,zhuang2020comprehensive}.
Huh et al.\ focus on
ImageNet~\cite{huh2016makes,raina2006constructing}.
Yang et al.\ study the relation between transfer learning and curriculum
learning with Sokoban~\cite{yang2021transfer}.

Early principles of meta-learning are described by Schmidhuber~\cite{schmidhuber1987evolutionary,schmidhuber1996simple}.
Meta-learning surveys
are~\cite{schaul2010metalearning,vilalta2002perspective,brazdil2008metalearning,schweighofer2003meta,vanschoren2018meta,weng2020meta,gupta2018meta,hospedales2020meta,huisman2020deep,huisman2021survey}.  
Papers on similarity-metric meta-learning are~\cite{koch2015siamese,vinyals2016matching,snell2017prototypical,sung2018learning,garcia2017few,shyam2017attentive}.
Papers on model-based meta-learning are~\cite{duan2016rl,wang2016learning,santoro2016meta,munkhdalai2017meta,mishra2017simple,edwards2016towards,garnelo2018conditional}.
Papers on optimization-based meta-learning are many~\cite{ravi2016optimization,li2017learning,finn2017model,antoniou2018train,rajeswaran2019meta,li2017meta,nichol2018first,rusu2018meta,finn2019online,grant2018recasting,finn2018probabilistic,yoon2018bayesian,bertinetto2018meta}.

Domain adaptation is studied
in~\cite{amodei2016concrete,donahue2014decaf,csurka2017domain,zhang2019transfer,brockhausen2021procedural}.
Zero-shot learning is an active and promising field. Interesting
papers are~\cite{chichilicious,huzero,larochelle2008zero,palatucci2009zero,stork2021large,akata2013label,xian2018zero,lampert2009learning,romera2015embarrassingly,do2005transfer,raina2006constructing}.
Like zero-shot learning, few-shot learning is also a
popular area of meta-learning
research~\cite{higgins2017darla,kuo2020encoding,sohn2018hierarchical,oh2017zero,stork2021large}.
Benchmark papers
are~\cite{triantafillou2019meta,yu2020meta,chen2019closer,tian2020rethinking,wang2021alchemy}.

\section*{Exercises}
\addcontentsline{toc}{section}{\protect\numberline{}Exercises}
It is time to test our understanding of transfer learning and meta-learning with some exercises and questions.
\subsubsection*{Questions}
Below are some quick questions to check your understanding of this
chapter. For each question a simple, single sentence answer is sufficient.

\begin{enumerate}
\item What is the reason for the interest in meta-learning and transfer learning?
\item What is transfer learning?
  \item What is meta-learning?
\item How is meta-learning different from multi task learning?
\item Zero-shot learning aims to identify classes that it has not
  seen before. How is that possible?
\item Is pretraining a form of transfer learning?
\item Can you explain learning to learn?
\item Are the initial network parameters also hyperparameters? Explain.
\item What is an approach for zero-shot learning?
\item As the diversity of tasks increases, does meta-learning
  achieve good results?
\end{enumerate}

\subsubsection*{Exercises}
Let us go to the programming exercises  to become more familiar with the methods that
we have covered in this chapter.
Meta-learning and transfer learning experiments are often very
computationally expensive.  You may need to scale down dataset
sizes, or skip some exercises as a last resort.

\begin{enumerate}
\item  \emph{Pretraining} Implement 
  pretraining and fine tuning in  the Keras pretraining example from
  Sect.~\ref{sec:pretraining}.\footnote{\url{https://keras.io/guides/transfer_learning/}}
    Do the exercises as suggested, including finetuning on the cats
    and dogs training set. Note the uses of preprocessing, data
    augmentation and regularization (dropout and batch
    normalization). See the effects of increasing the number of layers
    that you transfer on training performance and speed.
  \item \emph{MAML} Reptile~\cite{nichol2018first} is a meta-learning approach
    inpspired by MAML, but first order, and faster, specifically
    designed for few-shot learning. The Keras website contains a
    segment on
    Reptile.\footnote{\url{https://keras.io/examples/vision/reptile/}}
     At the start a number of
    hyperparameters are defined: learning rate, step size, batch size,
    number of meta-learning iterations, number of evaluation
    iterations, how many shots, classes, etcetera.   Study the effect of
    tuning 
    different hyperparameters, especially the ones related to few-shot
    learning: the classes, the shots, and the number of iterations. 

    To delve deeper into few-shot learning, also have a look at the
    MAML code, which has a section on reinforcement
    learning.\footnote{\url{https://github.com/cbfinn/maml_rl/tree/master/rllab}}
    Try different environments.

\item \emph{Meta World} As we have seen in Sect.~\ref{sec:metaworld}, Meta World~\cite{yu2020meta} is an elaborate benchmark suite for meta
  reinforcement learning. Re-read the section, go to GitHub, and
  install the
  benchmark.\footnote{\url{https://github.com/rlworkgroup/metaworld}}
  Also go to Garage to install the agent algorithms so that you are
  able to test their
  performance.\footnote{\url{https://github.com/rlworkgroup/garage}}
  See that they work with your PyTorch or TensorFlow setup.
  First try running the ML1 meta benchmark for PPO. Then try MAML, and
  RL2. Next, try the more elaborate meta-learning benchmarks. Read the
  Meta World paper, and see if you can
  reproduce their results.

\item \emph{ZSL} We go from few-shot learning to zero-shot
  learning. One of the ways in which zero-shot learning works is by
  learning attributes that are shared by classes. Read the papers \emph{Label-Embedding for Image
  Classification}~\cite{akata2013label}, and \emph{An embarrassingly simple approach to zero-shot
  learning}~\cite{romera2015embarrassingly}, and go to the
code~\cite{chichilicious}.\footnote{\url{https://github.com/sbharadwajj/embarrassingly-simple-zero-shot-learning}}
Implement it, and try to understand   how attribute learning
works. Print the attributes for the classes, and use the different
datasets. Does MAML work for few-shot learning?
 (challenging) 
\end{enumerate}


\chapter{Further Developments}\label{chap:conc}

We have come to the end of this book. We will reflect on what we have
learned. In this chapter we will review the main themes and essential lessons, and we will
look to the future. 

Why do we study deep reinforcement learning? Our inspiration is the dream of
artificial intelligence; to understand human intelligence and to create
intelligent behavior that can supplement our own, so that together we
can grow. For reinforcement learning our goal is to learn from the
world, to learn increasingly complex behaviors for increasingly
complex sequential decision problems. The preceding chapters have
shown us that many successful algorithms were inspired by  how
humans learn.

Currently many environments consist of games and simulated robots, in the
future  this may include human-computer interactions and
collaborations in teams with real humans.

\section{Development of Deep Reinforcement Learning}
Reinforcement learning has made a remarkable transition, from a method
that was used to learn small tabular toy problems, to learning
simulated robots how to walk, to playing the largest
multi-agent real time strategy games, and beating the best humans in
Go and  poker.
The reinforcement learning paradigm is a  framework in which
many  learning algorithms have been developed. The framework
is able to incorporate powerful ideas from other fields, such as deep learning, and
autoencoders.

To appreciate the versatility of reinforcement learning, and now that
we have studied the field in great detail, let us have a
closer look at how the developments in the field have proceeded over time.

\subsection{Tabular Methods}
Reinforcement learning starts with a simple agent/environment loop, where an environment
performs the agent's actions, and returns a reward (the model-free approach). We use a  Markov decision process to
formalize reinforcement learning. The value and policy 
functions are initially implemented in a tabular fashion, limiting 
the method to small environments, since the agent has to fit the
function representations in memory. Typical
environments are Grid world, Cartpole and Mountain car; a typical algorithm to find the optimal policy function
is tabular Q-learning. Basic principles in the design of these algorithms are
 exploration, exploitation and
on-policy/off-policy learning. Furthermore,  imagination, as a form of
model-based reinforcement learning, was developed.

This part of the field forms a well-established and stable basis, that
is, however, only suitable for learning small single-agent
problems. The advent of deep learning caused the field to shift into a
higher gear.

\subsection{Model-free Deep Learning}
Inspired by breakthroughs  in supervised image recognition, deep
learning was also applied to Q-learning, causing the Atari breakthrough for
deep reinforcement learning. The basis of the success of deep learning
in reinforcement learning are methods to break correlations and improve
convergence (replay buffer and a separate target
network). The DQN algorithm~\cite{mnih2015human} has become quite well known. Policy-based and actor critic approaches 
work well with deep learning, and are also applicable to continuous
action spaces. Many model-free actor critic variants have been
developed~\cite{schulman2017proximal,lillicrap2015continuous,haarnoja2018soft,mnih2016asynchronous},
they are often tested on  simulated robot applications. Algorithms often
reach good quality optima, but model-free algorithms have a high
sample complexity. Tables~\ref{tab:rainbow} and \ref{tab:pol} list these algorithms.

This part of the field---deep model-free value and policy-based algorithms---can
now be considered as well-established, with mature
algorithms whose behavior is well understood, and with good
results for high-dimensional single-agent environments. Typical
high-dimensional environments are the Arcade Learning Environment, 
and  MuJoCo simulated physics locomotion tasks.

\subsection{Multi-Agent Methods}
Next, more advanced methods are covered. In Chap.~\ref{chap:model} model-based algorithms 
combine planning and learning to improve sample efficiency. For
high-dimensional visual environments they  use uncertainty modeling and
latent models or world models, to reduce the dimensionality for planning. The algorithms are listed in Table~\ref{tab:overview}.

Furthermore, the step from
single-agent to multi-agent is made, enlarging the type of problems
that can be modeled, getting closer to real-world problems. 
The strongest human Go players are beaten with a model-based self-play
combination of MCTS and a deep actor critic algorithm. The self-play setup performs curriculum
learning, learning from previous learning tasks ordered from easy to hard, and is in that sense a
form of meta-learning. Variants are shown in Table~\ref{tab:az}.

For multi-agent and imperfect information problems, deep reinforcement learning is
used to study competition, emergent collaboration, and hierarchical
team learning. In these areas reinforcement learning comes close to
multi-agent systems and population based methods, such as swarm
computing.
Parallel population-based methods may be able to learn the policy
quicker than gradient-based methods, and they may be a good
fit for multi-agent problems.

Also,  imperfect information multi-agent problems are studied, such as
poker; for competitive games counterfactual regret minimization was developed.
 Research into cooperation is  continuing. Early
strong results are reported in StarCraft using team collaboration and
team competition. 
In addition,  research is being performed into emergent
social behavior, connecting the fields of reinforcement learning to
swarm computing and multi-agent systems. Algorithms and experiments are
listed in Table~\ref{tab:multigame}. A  list of hierarchical approaches is shown in Table~\ref{tab:hrl}.

In human learning, new concepts are learned based on old concepts.
Transfer learning from foundation models, and meta-learning, aim to
re-use existing knowledge, or even learn to learn.  
Meta-learning, curriculum learning, and hierarchical learning are
emerging as techniques to conquer ever larger state
spaces. Table~\ref{tab:meta} shows meta-learning approaches.  

All these areas should be considered as  advanced reinforcement
learning, where  active research is  
still very much occurring. New algorithms  are  being developed, and
experiments typically require large amounts of compute
power. Furthermore, results are less robust, and require much
hyperparameter tuning. More advances are needed, and expected.

\subsection{Evolution of Reinforcement Learning}
In contrast to supervised learning, which learns from  a fixed dataset, reinforcement learning is a
mechanism for learning by doing, just as children learn. The agent/environment framework has
turned out to be a versatile approach, that can be augmented and
enhanced when we try new problem domains, such as high-dimensions,
multi-agent or imperfect information. Reinforcement learning has encompassed methods from 
supervised learning (deep learning) and
unsupervised learning (autoencoders), as well as from population-based
optimization.

In this sense reinforcement learning has evolved from being a single-agent
Markov decision process to a framework
for learning, based on agent and environment. Other approaches can be
hooked into this framework, to learn new fields, and to 
improve performance. These additions can  interpret high dimensional states
(as in DQN), or  shrink a state space  (as with latent
models). 
When  accomodating self-play, the framework provided us with a curriculum learning
sequence, yielding world class levels of play in two-agent games. 


\section{Main Challenges}
Deep reinforcement learning is being used to understand more real world sequential
decision making situations. Among the applications that motivate these developments
are self driving cars and other autonomous operations, image and speech recognition, 
decision making, and, in general, acting naturally.

What will the future bring for deep reinforcement learning?
The main challenge for deep reinforcement learning is to
manage the combinatorial explosion that occurs 
when a sequence of decisions  is chained together.
Finding  the right kind of inductive bias can  exploit structure in
this  state space.

We list three major challenges for  current and future  research in deep
reinforcement learning:
 \begin{enumerate}
 \item Solving larger problems faster
 \item Solving problems with more agents
 \item Interacting with people
 \end{enumerate}
%
The following techniques 
address these  challenges:
\begin{enumerate}
\item Solving larger problems faster
  \begin{itemize}
  \item Reducing sample complexity with latent-models
  \item Curriculum learning in self-play methods
  \item Hierarchical reinforcement learning
  \item Learn from previous tasks with transfer learning and meta-learning 
  \item Better exploration through intrinsic motivation
  \end{itemize}
\item Solving problems with more agents
  \begin{itemize}
  \item Hierarchical reinforcement learning
  \item Population-based self-play league methods
\end{itemize}
\item Interacting with people
\begin{itemize}
\item Explainable AI
\item Generalization 
\end{itemize}
\end{enumerate}
Let us have a closer look at these techniques, to see what future
developments can be expected for them.

\subsection{Latent Models}\index{latent models}
Chapter~\ref{chap:model} discussed model-based deep reinforcement
learning methods. In model-based methods a transition model is learned
that is then used with planning to
augment the policy function, reducing sample complexity. A problem for model-based methods in
high-dimensional problems is that high-capacity networks need many
observations in order to prevent overfitting, negating the potential reduction in sample complexity.

One of the most promising model-based methods is
the use of autoencoders to create latent models, that compress or
abstract from irrelevant 
observations, yielding a lower-dimensional latent state model that can be used for
planning in a reduced state space. The reduction in sample complexity
of model-based approaches can thus be maintained. Latent models create
compact world representations, that are also used in hierarchical and
multi-agent problems, and further work is ongoing. 
%

A second development in model-based deep reinforcement learning is
the use of end-to-end planning and learning of the
transition model. Especially for elaborate self-play designs such as
AlphaZero, where an MCTS planner is integrated in  self-learning, the
use of end-to-end learning is advantageous, 
as the work on MuZero has shown. Research is ongoing in this field
where planning and learning are combined~\cite{schrittwieser2020mastering,muzero-general,vries2021visualizing,schrittwieser2021online,hubert2021learning,hessel2021muesli,moerland2021intersection}.

\subsection{Self-Play}\index{self-play}
Chapter~\ref{chap:given} discussed learning by self-play in two-agent
games. In many two-agent games the transition function is given.
When the environment of an agent is played by the opponent with
the exact same transition function, a self-learning self-play system
can be constructed in which both agent and environment improve
eachother. We have discussed examples of cycles of continuous
improvement, from tabula-rasa to world-champion level.

After earlier results in backgammon~\cite{tesauro1995td}, AlphaZero
achieved landmark results in Go, chess, and
shogi~\cite{silver2017mastering,silver2018general}. The AlphaZero
design includes an MCTS planner in a self-play loop that improves a
dual-headed deep residual network~\cite{silver2018general}.  The design has spawned much
further research, as well as inspired general interest in artificial
intelligence and reinforcement learning~\cite{duan2016rl,leibo2019autocurricula,segler2018planning,narvekar2020curriculum,laterre2018ranked,feng2020solving,schrittwieser2020mastering,jumper2021highly}.

\subsection{Hierarchical Reinforcement Learning}\index{hierarchical methods}

Team play is important in multi-agent problems, and hierarchical approaches
can  structure the
environment in a hierarchy of agents. Hierarchical reinforcement
learning methods are also applied to 
single agent problems, using principles of divide and conquer. Many
large single-agent problems are hierarchically structured. Hierarchical
methods aim to make use of this structure by dividing large problems
into smaller subproblems; they group primitive actions into macro actions. 
When a policy has been found
with a solution for a certain subproblem, then this can be re-used  when
the subproblem surfaces again.
Note
that for some problems it is difficult to find  a hierarchical structure that can be exploited
efficiently. 

Hierarchical reinforcement learning has been studied for some
time~\cite{chung2016hierarchical,flet2019promise,sutton1999between}. 
Recent work has been reported on successful methods for deep
hierarchical
learning and population-based training~\cite{li2019hierarchical,levy2019learning,nachum2018data,roder2020curious},
and more is to be expected.

\subsection{Transfer Learning and Meta-Learning}\index{meta learning}
Among the major challenges of  deep reinforcement learning is
the long training time.
Transfer learning and meta-learning aim to reduce
the long training times, by transferring learned knowledge from
existing to new (but related) tasks, and by learning to learn from the
training of previous tasks, to speedup learning new (but related) tasks.

In the fields of image recognition and natural language processing it
has become common practice to use networks that are pretrained on
ImageNet~\cite{deng2009imagenet,fei2009imagenet} or
BERT~\cite{devlin2018bert} or other large pretrained networks~\cite{nadkarni2011natural,bird2009natural}.
Optimization-based methods such as MAML  learn better initial network
parameters for new tasks, and have spawned much further research. 

Zero-shot learning is a meta-learning approach where outside information is
learned, such as attributes or a textual description for image content, that is then
used to recognize individuals from a new
class~\cite{romera2015embarrassingly,sohn2018hierarchical,xian2018zero,akata2013label}. Meta-learning is a highly active field where more results can be expected.

 \begin{figure}[t]
    \centering{\includegraphics[width=11cm]{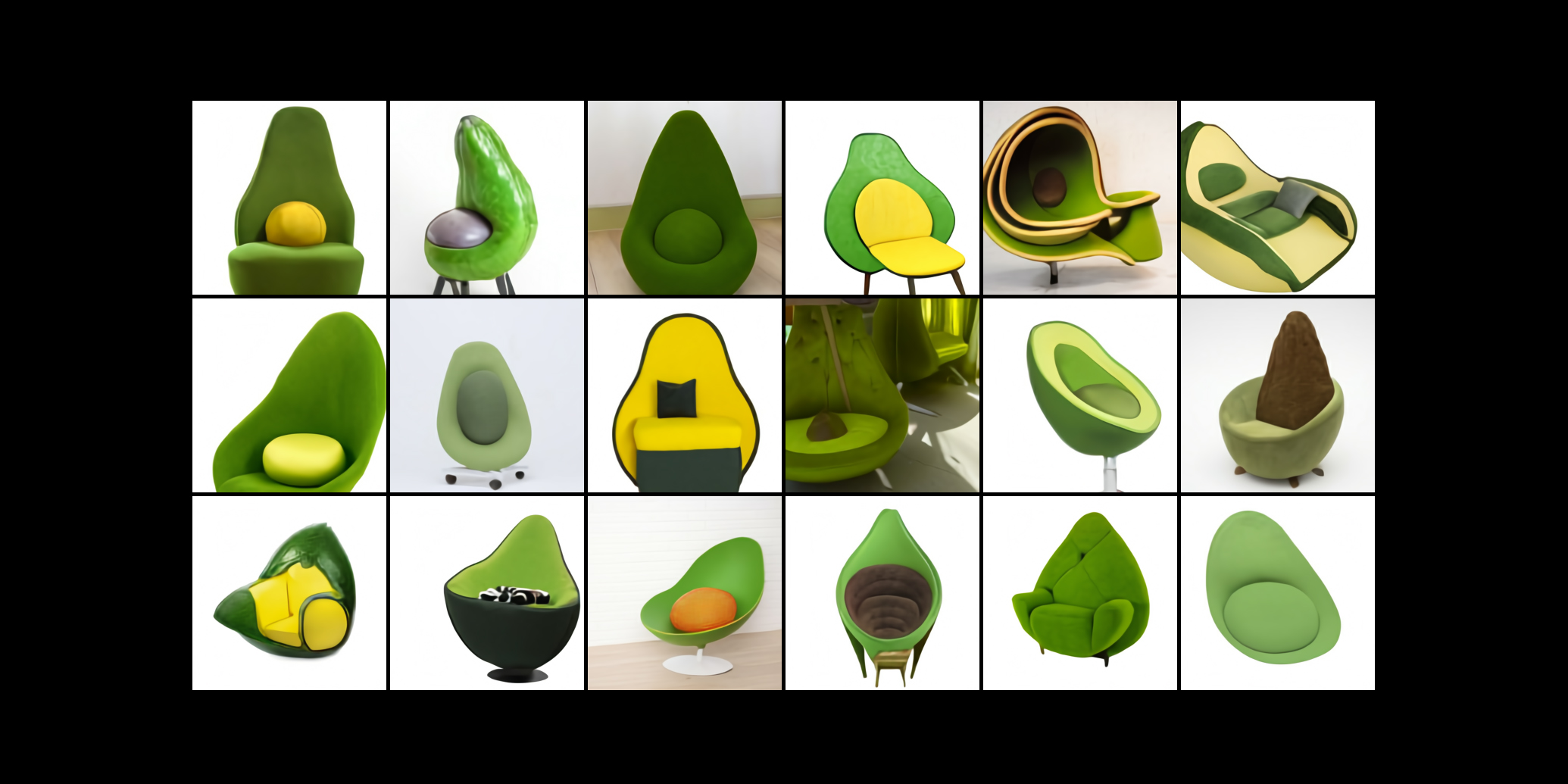}
     \caption{DALL-E, an Algorithm that draws Pictures based on
       Textual Commands \cite{radford2021learning}}\label{fig:dalle}}
 \end{figure}

Foundation models are large models, such as ImageNet for image
recognition, and GPT-3 in natural language processing, that are
trained extensively on large datasets. They contain general 
knowledge, that can be specialized for a certain more specialized
task. They can also be used for multi-modal tasks, where text and
image information is combined. The DALL-E project is able to create
images that go with textual descriptions. See Fig.~\ref{fig:dalle} for
amusing or beautiful examples (``an armchair in the shape of an
avocado'')~\cite{radford2021learning}. GPT-3 has also been used to
study zero-shot learning, with success, in the CLIP
project~\cite{ramesh2021zero}.  

\subsection{Population-Based Methods}\index{population based self-play}
Most reinforcement learning has focused on one or two-agent
problems. However, the world around us is full of many-agent problems.
The major  problem of  multi-agent reinforcement learning is to model
the large, nonstationary, problem space. 

Recent work applies self-play methods in a multi-agent
setting, where entire populations of agents are trained against
eachother. This approach combines aspects of evolutionary algorithms (combining and
 mutating policies, as well as culling of underperforming
agent policies) and hierarchical methods (modeling of team
collaboration).

Population-based training of leagues of agents has achieved 
success in  highly complex multi-agent games: StarCraft~\cite{vinyals2019grandmaster} and Capture the
Flag~\cite{jaderberg2019human}.
Population-based methods, combining evolutionary principles with self-play, curriculum learning, and hierarchical methods are active areas
of
research~\cite{salimans2017evolution,jaderberg2017population,miikkulainen2019evolving,volz2018evolving,khadka2019collaborative}. 

\subsection{Exploration and Intrinsic Motivation}\index{intrinsic motivation}
The prime motivator for learning in reinforcement learning is
reward. However, reward is sparse in  many sequential decision
problems. Reward shaping tries to augment the reward function with
heuristc knowledge. 
%
%
%
%
Developemental psychology argues that   learning is (also) based on
curiosity, or
intrinsically  motivated. Intrinsic motivation is a basic curiosity drive to explore for
exploration's sake, deriving satisfaction from the exploration process
itself. 

The field of intrinsic motivation is relatively new in reinforcement
learning. Links 
with hierarchical reinforcement learning and the options framework are
being explored, as well as with models of
curiosity~\cite{schmidhuber1991possibility,santucci2020intrinsically,colas2019curious}. 

Intrinsic motivation in reinforcement
learning can be used for exploring open-ended
environments~\cite{aubret2019survey,singh2005intrinsically} (see also
Sect.~\ref{sec:intrinsic}). 
Intrinsically motivated goal-conditioned  algorithms can train agents
to learn to represent, generate and pursue their own
goals~\cite{colas2020intrinsically}. 
%
The success of the Go-Explore algorithm in domains with sparse rewards
also stresses the importance of exploration in reinforcement
learning~\cite{ecoffet2021first}.


\subsection{Explainable AI}\index{explainable AI}\index{XAI}\label{sec:xai}


Explainable AI (\gls{XAI}) is closely related to the topics of planning and
learning that we discuss in this book, and to natural language
processing.

When a human expert suggests an answer, this expert can be questioned to explain the
reasoning behind the answer. This is a  desirable property,
and enhances how much we trust the answer. Most
clients receiving advice, be it financial or  medical,
put greater trust in  a well-reasoned explanation than in a \emph{yes} or \emph{no}
answer without any explanation. 

Decision support systems that are based on classic symbolic AI can
often be made to provide such reasoning easily. For example,
interpretable models~\cite{proencca2020interpretable}, decision
trees~\cite{quinlan1986induction}, graphical
models~\cite{jordan1998learning,lauritzen1996graphical}, and 
search trees~\cite{coelho1986automated} can be traversed and the choices at decision points
can be recorded and used to translate in a human-understandable argument.

Connectionist approaches  such as deep learning, in contrast, are
less interpretable. Their accuracy, however, is typically much higher
than the classical approaches. Explainable AI aims to combine the ease of
interpreting symbolic AI with the accuracy of connectionist
approaches~\cite{gunning2017explainable,doran2017does,browne2019strategic}.

The work on soft decision trees~\cite{frosst2017distilling,hinton2015distilling} and adaptive neural
trees~\cite{tanno2018adaptive} has shown how hybrid approaches of planning and learning can try to
build an explanatory decision tree based on a neural network.
 These
works build in part  on  model 
compression~\cite{cheng2017survey,bucilua2006model}\index{model
  compression}\index{adaptive neural trees}\index{soft decision trees}
and belief
networks~\cite{heckerman1995learning,neapolitan2004learning,scutari2009learning,de2002ant,bellemare2013bayesian,teyssier2012ordering,brochu2010tutorial,asmuth2009bayesian}.\index{knowledge representation}\index{interpretable models} 
Unsupervised methods can  be used to find interpretable
models~\cite{proencca2020interpretable,rudin2019stop,vellido2012making}. Model-based
reinforcement learning methods aim to  perform deep
sequential planning in learned world models~\cite{ha2018world}.

\subsection{Generalization}\index{generalization}
Benchmarks drive algorithmic progress in artificial intelligence. Chess and poker
have given us deep heuristic search; ImageNet has driven deep supervised
learning; ALE has given us deep Q-learning; MuJoCo and 
the DeepMind control suite have driven actor critic methods; Omniglot,
MiniImagenet and Meta-World drive meta-learning; and StarCraft and
other multi-agent games drive hierarchical and population-based
methods.

As the work on explainable AI indicates, there is a  trend in
reinforcement learning to study  problems that are closer
to the real world, through model-based methods, multi-agent methods, meta-learning
and hierarchical methods. 

As deep reinforcement learning will be more widely applied to real-world
problems, generalization becomes important, as
is argued by Zhang et al.~\cite{zhang2021understanding}. Where supervised learning
experiments have a clear training set/test set separation to measure
generalization, in reinforcement learning agents often fail to
generalize beyond the environment they were trained
in~\cite{packer2018assessing,cobbe2018quantifying}. Reinforcement learning agents often overfit
on their training
environment~\cite{zhang2018dissection,zhang2018study,whiteson2011protecting,farebrother2018generalization,ghosh2021generalization}. This
becomes especially diffcult in sim-to-real
transfer~\cite{zhao2020sim}. One benchmark specifically designed to increase
generalization is  Procgen. It aims to increase
environment diversity through procedural content
generation, providing 16 parameterizable
environments~\cite{cobbe2020leveraging}.

Benchmarks will continue to drive progress in artificial intelligence,
especially for generalization~\cite{kirk2021survey}.

\section{The Future of Artificial Intelligence}
This book has covered the stable basis of
deep reinforcement learning, as well as active areas of research.
Deep reinforcement learning is a highly active field, and many more
developments will follow. 

We have  seen complex methods for solving sequential
decision  problems, some of which are easily solved on a daily
basis by humans in the world around us. In certain problems, such
as backgammon, chess, checkers, poker, and Go, computational methods have now
surpassed human ability. In most other endeavours, such as pouring water from a
bottle in a cup,  writing poetry, or falling in love,
humans still reign supreme.

Reinforcement learning is inspired by
biological learning, yet computational and biological methods for
learning are still far apart. Human 
intelligence is general and broad---we know much about many different
topics, and we use our general knowledge of previous tasks when learning new
things. Artificial intelligence is specialized and deep---computers can
be  good at  certain tasks, but their intelligence is narrow, and
learning from other tasks 
is  a challenge.

Two conclusion are clear. First, for humans, hybrid intelligence, where
human general intelligence is augmented by specialized artificial
intelligence,  can be highly beneficial. Second, for AI, the field of
deep reinforcement learning is
taking cues from human learning in hierarchical methods, curriculum learning, learning to learn, and multi-agent
cooperation.

The future of artificial intelligence is human.

%



\part*{Appendices}
\appendix

\chapter{Mathematical Background}\label{app:math}
This appendix  provides essential  mathematical background and establishes the notation that we use in this book.
We start with notation of
sets and functions, then we discuss probability distributions,
expectations, and information theory. You will most
likely have seen some of these in previous courses.  We will also discuss how to differentiate through an
expectation, which frequently appears in machine learning.

This appendix is based on Moerland~\cite{moerland2021lecture}.



\section{Sets and Functions} \label{sec_preliminaries}
We start at the beginning, with sets and functions.

\subsection{Sets}
\subsubsection*{Discrete set}
A discrete set is a set of countable elements.
\begin{tcolorbox}
{\bf Examples}:
\begin{itemize} 
\item \makebox[5cm]{${X}=\{1,2,..,n\}$ \hfill}  (integers)
\item \makebox[5cm]{${X}=\{$up, down, left, right$\}$ \hfill} (arbitrary elements)
\item \makebox[5cm]{${X}=\{0,1\}^d$ \hfill} (d-dimensional binary space)
\end{itemize}
\end{tcolorbox}

\subsubsection*{Continuous set}
A continuous set is a set of connected elements.
\begin{tcolorbox}
{\bf Examples}:
\begin{itemize} 
\item \makebox[5cm]{${X}=[2,11]$ \hfill} (bounded interval)
\item \makebox[5cm]{${X} = \mathbb{R}$ \hfill} (real line)
\item \makebox[5cm]{${X}=[0,1]^d$ \hfill}  ($d$-dimensional hypercube)

\end{itemize} 
\end{tcolorbox}


\subsubsection*{Conditioning a set}
We can also condition within a set, by using $:$ or $|$. For example,
the {\it discrete probability $k$-simplex}, which is what we actually
use to define a discrete probability distribution over $k$ categories,
is given by: 

$${X}=\{ x \in [0,1]^k : \sum_k x_k=1 \}.$$
This means that $x$ is a vector of length $k$, consisting of entries
between 0 and 1, with the restriction that the vector sums to 1. 

\ \\
\subsubsection*{Cardinality and dimensionality}

It is important to distinguish the cardinality and dimensionality of a set:

\begin{itemize}
\item The {\it cardinality} (size) counts the number of elements in a
  vector space, for which we write $|{X}|$.  
\item The {\it dimensionality} counts the number of dimensions in the
  vector space ${X}$, for which we write
  $\text{Dim}({X})$.  
\end{itemize}

\begin{tcolorbox}
{\bf Examples}:
\begin{itemize}

\item The discrete space ${X}=\{0,1,2\}$ has cardinality
  $|{X}|=3$ and dimensionality $\text{Dim}({X})=1$.  

\item The discrete vector space ${X}=\{0,1\}^4$ has
  cardinality $|{X}|=2^4=16$ and dimensionality
  $\text{Dim}({X})=4$.  
\end{itemize}  

\end{tcolorbox}

\subsubsection*{Cartesian product}

We can combine two spaces by taking the Cartesian product, denoted by
$\times$, which consists of all the possible combinations of elements
in the first and second set: 

$$ {X} \times {Z} = \{ (x,z) : x \in {X}, z \in {Z} \} $$

\noindent We can also combine discrete and continuous spaces through Cartesian products. 

\begin{tcolorbox}
{\bf Example}:
Assume ${X} = \{20,30\}$ and ${Z} = \{0,1\}$. Then

$$ {X} \times {Z} = \{(20,0),(20,1),(30,0),(30,1) \}$$

Assume ${X} = \mathbb{R}$ and ${Z} = \mathbb{R}$. Then
${X} \times {Z} = \mathbb{R}^2$.   

\end{tcolorbox}

\subsection{Functions}
\begin{itemize}
\item A function $f$ maps a value in the function's {\it domain}
  ${X}$ to a (unique) value in the function's {\it
    co-domain}/{\it range} ${Y}$, where ${X}$ and
  ${Y}$ can be discrete or continuous sets.  
\item We write the statement that $f$ is a function from ${X}$
  to ${Y}$ as $$f: {X} \to {Y}$$ 
\end{itemize}

\begin{tcolorbox}
{\bf Examples}:
\begin{itemize}
\item $y = x^2$ maps every value in domain ${X} \in
  \mathbb{R}$ to range ${Y} \in \mathbb{R}^+$ (see Fig.~\ref{fig:yx2})

\end{itemize}
\end{tcolorbox}
\vspace{0.5cm}
\begin{figure}[t]
  \centering
  \includegraphics[width = 0.4\textwidth]{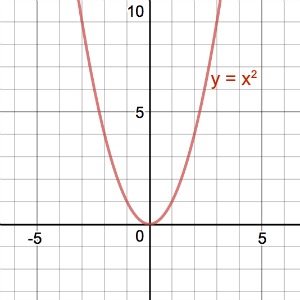}
  \caption{$y=x^2$}\label{fig:yx2}
\end{figure}

\section{Probability Distributions} \label{chapter_probability_distributions}
A probability distribution is a mathematical function that gives the
probability of the occurrence of a set of possible outcomes. The set
of possible outcomes is called the {\it sample space}, which can be
discrete or continuous, and is denoted by ${X}$. For example,
for flipping a coin ${X} =
\{\text{heads},\text{tails}\}$. When we actually sample the variable,
we get a particular value $x \in {X}$. For example, for the
first coin flip $x_1 = \text{heads}$. Before we actually sample the
outcome, the particular outcome value is still unknown. We say that it
is a {\it random variable}, denoted by $X$, which always has an
associated probability distribution $p(X)$.  

\begin{center}
\begin{tabular}{p{3.5cm} p{4.5cm}}
Sample space (a set) & ${X}$ \newline \\

Random variable &  $X$ \newline \\

Particular value &  $x$ \newline \\

\end{tabular}
\end{center}

Depending on whether the sample space is a discrete or continuous set,
the distribution $p(X)$ and the way to represent it differ. We detail
both below, see also Fig.~\ref{fig_discrete_continuous}.  

\subsection{Discrete Probability Distributions}
\begin{itemize}
\item A {\it discrete variable} $X$ can take values in a discrete set
  ${X} = \{1,2,..,n\}$. A particular value that $X$ takes is
  denoted by $x$.  
\item Discrete variable $X$ has an associated {\it probability mass
    function}: $p(X)$, where $p:{X} \to [0,1]$. Each possible
  value $x$ that the variable can take is associated with a
  probability $p(X=x) \in[0,1]$. (For example, $p(X=1)=0.2$, the
  probability that $X$ is equal to 1 is $20\%$.)  
\item Probability distributions always sum to 1: $\sum_{x \in {X}} p(x) = 1$. 
\end{itemize}

\subsubsection*{Parameters}
We represent a probability distribution with {\it parameters}. For a
discrete distribution of size $n$, we need $n{-}1$ parameters, 
$\{p_{x=1},..,p_{x=n{-}1}\}$, where $p_{x=1} = p(x{=}1)$. The
probability of the last category follows from the sum to one
constraint,  $p_{x=n} = 1 - \sum_{i=1}^{n-1} p_{x=i}$.  

\begin{tcolorbox}
{\bf Example}:
A discrete variable $X$ that can take three values (${X} 
=\{1,2,3\}$), with associated probability distribution $p(X=x)$: 

\begin{center}
\begin{tabular}{c c c}
$p(X=1)$ & $p(X=2)$ & $p(X=3)$ \\
\hline
$0.2$ & $0.4$ & $0.4$ \\
\end{tabular}
\end{center}
\end{tcolorbox}

\subsubsection*{Representing discrete random variables}
It is important to realize that we always represent a discrete
variable {\it as a vector} of probabilities. Therefore, the above
variable $X$ does not really take values ${X} =\{1,2,3\}$,
because 1, 2 and 3 are arbitrary categories (category two is not
twice as much as the first category). We could just as well have
written ${X} =\{a,b,c\}$. Always think of the possible values
of a discrete variable as separate entries. Therefore, we should
represent the value of a discrete variable as a vector of
probabilities. In the data, when we observe the ground truth, this
becomes a {\it one-hot encoding}, where we put all mass on the
observed class.  

\vspace{0.3cm}
\begin{tcolorbox}
{\bf Example}:
In the above example, we had  (${X} =\{1,2,3\}$). Imagine we sample $X$ three times and observe 1, 2 and 3, respectively. We would actually represent these observations as

\begin{center}
\begin{tabular}{c c}
Observed category\qquad &\qquad Representation  \\
\hline
1 & $(1,0,0)$  \\
2 & $(0,1,0)$  \\
3 & $(0,0,1)$  \\
\end{tabular}
\end{center}

\end{tcolorbox}

\subsection{Continuous Probability Distributions}
\begin{itemize}
\item A {\it continuous variable} $X$ can take values in a continuous
  set,   ${X} = \mathbb{R}$ (the real line), or ${X} = [0,1]$ (a bounded interval). 
\item Continuous variable $X$ has an associated {\it probability 
    density function}: $p(X)$, where $p:{X} \to \mathbb{R}^+$
  (a positive real number).  
\item In a continuous set, there are infinitely many values that the
  random value can take. Therefore, the {\it absolute} probability of
  any particular value is 0.  
\item We can only define absolute probability on an interval, 
  $p(a < X \leq b) = \int_a^b p(x)$. (For example, $p(2 < X \leq
  3)=0.2$,  the probability that $X$ will fall between 2 and 3 is
  equal to $20\%$.)  
\item The interpretation of an individual value of the density, like
  $p(X=3) = 4$, is only a {\it relative} probability. The higher the
  probability $p(X=x)$, the higher the relative chance that we would
  observe $x$.  
\item Probability distributions always sum to 1: $\int{x \in
    {X}} p(x) = 1$ (note that this time we integrate instead of
  sum).   
\end{itemize}

\subsubsection*{Parameters}
We need to represent a continuous distribution with a parameterized
function, that for every possible value in the sample space predicts a
relative probability. Moreover, we need to obey the sum to one
constraint. Therefore, there are many {\it parameterized continuous
  probability densities}. An example is the Normal distribution. A
continuous density is a function $p:{X} \to \mathbb{R}^+$ that
depends on some parameters. Scaling the parameters allows variation in
the location where we put probability mass.  

\begin{figure}[t]
  \centering
      \includegraphics[width = 1.0\textwidth]{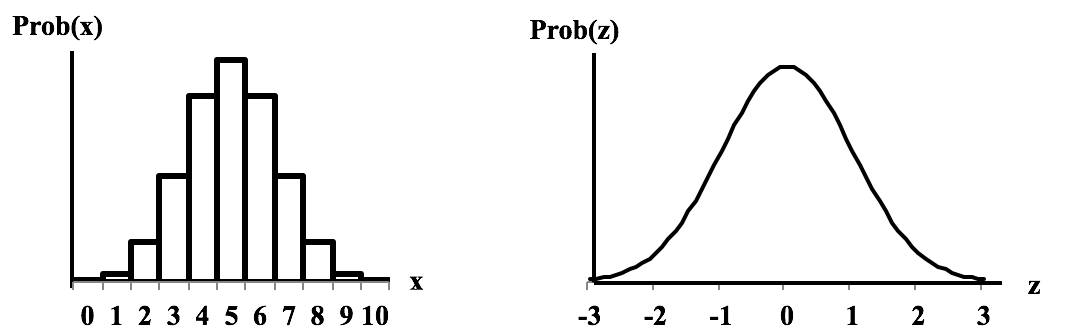}
  \caption[Discrete and Continuous]{Examples of discrete (left) versus
    continuous (right) probability distibution \cite{jawlik2016statistics}} 
    \label{fig_discrete_continuous}
\end{figure}

\begin{center}
\begin{table}[t]
\footnotesize
\begin{minipage}{\textwidth} 
\caption{Comparison of discrete and continuous probability distributions}
\vspace{0.3cm} \label{table_discrete_continuous}
\begin{tabular}{p{2.7cm}|P{4.0cm} P{4.0cm} }
 & \bf Discrete distribution & \bf Continuous distribution \\
\hline

Input/sample space & Discrete set,\newline  ${X}=\{0,1\}$, \newline with size $n = |{X}|$ \newline & Continuous set,\newline  ${X} = \mathbb{R}$ \newline \\

Probability function  & Probability mass function (pmf) \newline $p:
                        {X} \to [0,1]$ \newline \newline such
                        that $\sum_{x\in{X}} p(x) = 1$ &
                                                                 Probability
                                                                 density
                                                                 function
                                                                 (pdf)
                                                                 \newline
                                                                 $p:
                                                                 {X}
                                                                 \to
                                                                 \mathbb{R}$
                                                                 \newline
                                                                 \newline
                                                                 such
                                                                 that
                                                                 $\int_{x\in{X}}
                                                                 p(x)
                                                                 = 1$
                                                                 \newline
  \\  

Possible parametrized distributions \newline & Various, but only need simple Discrete \newline & Various,  \newline Normal, Logistic, Beta, etc.  \\

Parameters & $\{p_{x=1},..,p_{x=n{-}1}\}$ \newline & Depends on distribution, \newline  for normal: $\{\mu,\sigma\}$ \newline \\

Number of parameters & $n{-}1 = |{X}|{-}1 $ \newline (Due to
                       sum to 1 constraint)\footnote{Due to the sum to
                       1 constraint, we need one parameter less than
                       the size of the sample space, since the last
                       probability is 1 minus all the others: $p_n =
                       1-\sum_{i=1}^{n-1} p_i$.} & Depends on
                                                   distribution,\newline
                                                    for normal: 2
                                                   \newline \\

Example distribution function & $p(x=1)=0.2$ \newline $p(x=2)=0.4$
                                \newline $p(x=3)=0.4$ \newline  &
                                                                  e.g. for normal $p(x|\mu,\sigma) = \frac{1}{\sigma\sqrt{2\pi}} \exp \Big( - \frac{(x-\mu)^2}{2\sigma^2}  \Big) $  \\ 

Absolute probability & $p(x=1)=0.2$ \newline  & $p(3 \leq x < 4)=
                                                \int_3^4 p(x) = 0.3$
                                                \newline (on
                                                interval)\footnote{Note
                                                that for continuous
                                                distributions,
                                                probabilities are only
                                                defined on {\it
                                                intervals}. The
                                                density function
                                                $p(x)$ only gives
                                                relative
                                                probabilities, and
                                                therefore we may have
                                                $p(x) >1$, like
                                                $p(x=3)=5.6$, which is
                                                of course not possible
                                                (one should not
                                                interpret it as an
                                                absolute
                                                probability). However,
                                                $p(a \leq x < b) =
                                                \int_a^b p(x) < 1 $ by
                                                definition.}\newline
  \\ 

Relative probability & -  & $p(x=3)=7.4$ \\

\end{tabular}
\end{minipage}
\end{table}
\end{center}

\begin{tcolorbox}
{\bf Example}:
A variable $X$ that can take values on the real line with
distribution $$p(x;\mu,\sigma) = \frac{1}{\sigma\sqrt{2\pi}} \exp
\Big( - \frac{(x-\mu)^2}{2\sigma^2}  \Big).$$  

Here, the mean parameter $\mu$ and standard deviation $\sigma$ are the
parameters. We can change them to change the shape of the
distribution, while always ensuring that it still sums to one. We draw
an example normal distribution in Fig.~\ref{fig_discrete_continuous},
right.   

The differences between discrete and continuous probability distributions are summarized in Table \ref{table_discrete_continuous}.
\end{tcolorbox}

\subsection{Conditional Distributions}
\begin{itemize}
\item A conditional distribution means that the distribution of one
  variable depends on the value that another variable takes.  

\item We write $p(X|Y)$ to indicate that the value of $X$ depends on the value of $Y$. 
\end{itemize}

\begin{tcolorbox}
{\bf Example}:
For discrete random variables, we may store a conditional distribution
as a table of size $|{X}| \times |{Y}|$. A variable
$X$ that can take three values (${X} =\{1,2,3\}$) and variable
$Y$ that can take two values (${Y} =\{1,2\}$). The conditional
distribution may for example be: 

\begin{center}
\begin{tabular}{c|c c c}
 & $p(X=1|Y)$ & $p(X=2|Y)$ & $p(X=3|Y)$ \\
\hline
$Y=1$ & 0.2 & 0.4 & 0.4 \\
$Y=2$ & 0.1 & 0.9 & 0.0 \\
\end{tabular}
\end{center}

Note that for each value $Y$, $p(X|Y)$ should still sum to 1, it
is a valid probability distribution. In the table above, each row
therefore sums to 1.  
\end{tcolorbox}

\begin{tcolorbox}
{\bf Example}:
We can similarly store conditional distributions for continuous random
variables, this time mapping the input space to the parameters of a
{\it continuous} probability distribution. For example, for $p(Y|X)$
we can assume a Gaussian distribution ${N}(\mu(x),\sigma(s))$,
where the mean and standard deviation depend on $x\in
\mathbb{R}$. Then we can for example specify: 

$$\mu(x) = 2 x,\quad \sigma(x) = x^2 $$

and therefore have

$$ p(y|x) = {N}(2x,x^2) $$

Note that for each value of $X$, $p(Y|X)$ still integrates to 1,  it is a valid probability distribution. 
\end{tcolorbox}

\subsection{Expectation}
We also need the notion of an {\it expectation}. 

\subsubsection{Expectation of a Random Variable}
The expectation of a random variable is essentially an average. For a
discrete variable, it is defined as: 

\begin{equation}
\mathbb{E}_{X \sim p(X)}[f(X)] = \sum_{x \in {X}} [ x \cdot p(x) ]
\end{equation}
For a continuous variable, the summation becomes integration. 

\begin{tcolorbox}
{\bf Example}:

Assume a given $p(X)$ for a binary variable:
\begin{center}
\begin{tabular}{c|c}
$x$ & $p(X=x)$ \\
\hline
0 & 0.8 \\
1 & 0.2 \\
\end{tabular}
\end{center}

The expectation is
\begin{align}
\mathbb{E}_{X \sim p(X)} [f(X)] &= 0.8 \cdot 0 + 0.2 \cdot 1 = 0.2
\end{align}

\end{tcolorbox}

\subsubsection{Expectation of a Function of a Random Variable}
More often, and also in the context of reinforcement learning, we will
need the expectation {\it of a function of the random variable},
denoted by $f(X)$. Often, this function maps to a continuous output.  

\begin{itemize}
\item Assume a function $f: {X} \to \mathbb{R}$, which, for
  every value $x \in {X}$ maps to a continuous value $f(x) \in
  \mathbb{R}$. 

\item The expectation is then defined as follows:
\begin{equation}
\mathbb{E}_{X \sim p(X)}[f(X)] = \sum_{x \in X} [ f(x) \cdot p(x) ]
\end{equation}
\end{itemize}

\noindent For a continuous variable the summation again becomes
integration. The formula may look complicated, but it essentially
reweights each function outcome by the probability that this output
occurs, see the example below.  

\begin{tcolorbox}
{\bf Example}:

Assume a given density $p(X)$ and function $f(x)$:
\begin{center}
\begin{tabular}{c|cc}
$x$ & $p(X=x)$ & $f(x)$ \\
\hline
1 & 0.2 & 22.0 \\
2 & 0.3 & 13.0 \\
3 & 0.5 & 7.4
\end{tabular}
\end{center}

The expectation of the function can be computed as 
\begin{align}
\mathbb{E}_{X \sim p(X)} [f(X)] &= 22.0 \cdot 0.2 + 13.0 \cdot 0.3 + 7.4 \cdot 0.5 \nonumber \\
&= 12.0 \nonumber
\end{align}

\end{tcolorbox}

The same principle applies when $p(x)$ is a continuous density, only
with the summation replaced by integration.

\subsection{Information Theory}

Information theory studies the amount of information that is present in
distributions, and the way that we can compare distributions.  

\subsubsection{Information}\label{sec:ce}
The {\it information} $I$ of an event $x$ observed from distribution $p(X)$ is defined as:

$$ I(x) = -\log p(x). $$

In words, the more likely an observation is (the higher $p(x)$), the
less information we get when we actually observe the event. In other
words, the information of an event is the (potential) reduction of
uncertainty. On the two extremes we have: 

\begin{itemize}
\item $p(x) = 0 \quad : \quad  I(x) = -\log 0 = \infty$
\item $p(x) = 1 \quad : \quad I(x) = -\log 1 = 0 $
\end{itemize}

\begin{figure}[t]
  \centering
      \includegraphics[width = 0.5\textwidth]{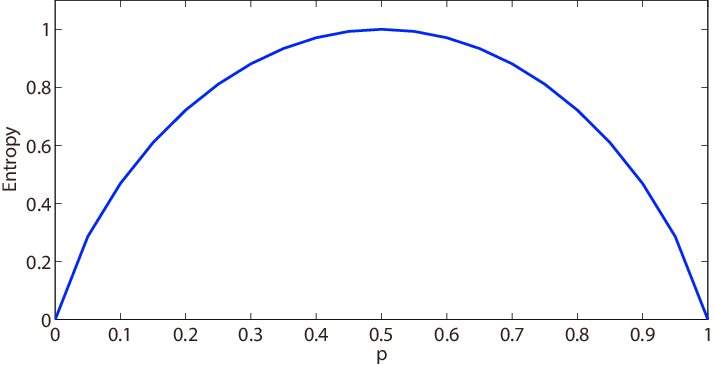}
  \caption[Entropy]{Entropy of a binary discrete variable. Horizontal
    axis shows the probability that the variable takes value 1, the
    vertical axis shows the associated entropy of the
    distribution. High entropy implies high spread in the
    distribution, while low entropy implies little spread.} 
    \label{fig_entropy}
\end{figure}

\subsubsection{Entropy}
We define the \gls{entropy} $H$ of a discrete distribution $p(X)$ as

\begin{align} 
H[p] &= \mathbb{E}_{X\sim p(X)} [ I(X) ] \nonumber \\ 
&= \mathbb{E}_{X\sim p(X)} [ - \log p(X) ] \nonumber \\
&= - \sum_{x} p(x) \log p(x) \nonumber \\
\end{align}

If the base of the logarithm is 2, then we measure it in {\it
  bits}. When the base of the logarithm is $e$, then we measure the
entropy in {\it nats}. The continuous version of the above equation is
called the {\it continuous entropy} or {\it differential entropy}.  

Informally, the entropy of a distribution is a measure of the amount of
``uncertainty'' in a distribution, i.e., a measure of its ``spread.''
We can nicely illustrate this with a binary variable (0/1), where we
plot the probability of a 1 against the entropy of the distribution
(Fig.~\ref{fig_entropy}). We see that on the two extremes, the entropy
of the distribution is 0 (no spread at all), while the entropy is
maximal for $p(x=1) = 0.5$ (and therefore $p(x=0)=0.5$), which gives
maximal spread to the distribution.  

\begin{tcolorbox}
{\bf Example}: The entropy of the distribution in the previous example is:

\begin{align}
H[p] &= - \sum_{x} p(x) \log p(x) \nonumber \\
&= - 0.2 \cdot \ln 0.2 - 0.3 \cdot \ln0.3 - 0.5 \cdot \ln 0.5 = 1.03 \text{ nats}
\end{align}

\end{tcolorbox}

\subsubsection{Cross-entropy}\index{cross-entropy}\label{sec:cem}
The cross-entropy is defined between two distributions $p(X)$ and
$q(X)$ defined over the same support (sample space). The cross-entropy
is given by: 

\begin{align} 
H[p,q] &= \mathbb{E}_{X\sim p(X)} [-  \log q(X) ] \nonumber \\ 
&= - \sum_{x} p(x) \log q(x) \nonumber \\
\end{align}

When we do maximum likelihood estimation in supervised learning, then
we actually minimize the cross-entropy between the data distribution
and the model distribution. 

\subsubsection{Kullback-Leibler Divergence}\index{KL divergence}
For two distributions $p(X)$ and $q(X)$ we can also define the {\it
  relative entropy}, better known as the Kullback-Leibler (KL)
divergence $D_\text{KL}$: 

\begin{align} 
D_\text{KL}[p||q] &= \mathbb{E}_{X\sim p(X)} \Big[-  \log \frac{q(X)}{p(X)} \Big] \nonumber \\ 
&= - \sum_{x} p(x) \log \frac{q(x)}{p(x)} \nonumber \\
\end{align}

The Kullback-Leibler divergence is a measure of the distance between
two distributions. The more two distributions depart from eachother,
the higher the KL-divergence will be. Note that the KL-divergence is
not symmetrical,  $D_\text{KL}[p||q] \neq D_\text{KL}[q||p] $ in
general.  

Finally, we can also rewrite the KL-divergence as an entropy and
cross-entropy, relating the previously introduced quantities: 

\begin{align} 
D_\text{KL}[p||q] &= \mathbb{E}_{X\sim p(X)} \Big[-  \log \frac{q(X)}{p(X)} \Big] \nonumber \\ 
&= \sum_{x} p(x) \log p(x) - \sum_{x} p(x) \log q(x) \nonumber \\
&= H[p] + H[p,q] \nonumber \\
\end{align}
Entropy, cross-entropy and KL-divergence are common in many machine
learning domains, especially to construct loss functions. 


\section{Derivative of an Expectation} \label{sec_derivative_of_expectation}\label{sec:deriv}

A key problem in gradient-based optimization, which appears in parts
of machine learning, is getting the gradient of an expectation. We
will here discuss one well-known method:\footnote{Other methods to
  differentiate through an expectation is through the {\it
    reparametrization trick}, as for example used in variational
  auto-encoders, but we will not further treat this topic here.} the
{\it REINFORCE} estimator (in reinforcement learning), which is in
other fields also know as the {\it score function estimator}, {\it
  likelihood ratio method}, and {\it automated variational inference}.  

Assume that we are interested in the gradient of an expectation, where
the parameters appear in the distribution of the
expectation:\footnote{If the parameters only appear in the function
  $f(x)$ and not in $p(x)$, then we can simply push the gradient
  through the expectation} 
\begin{equation} 
\nabla_\theta \mathbb{E}_{x \sim p_\theta(x)} [ f(x) ] \label{eq_grad_of_exp}
\end{equation}
We cannot sample the above quantity, because we have to somehow move
the gradient inside the expectation (and then we can sample the
expectation to evaluate it). To achieve this, we will use a simple
rule regarding the gradient of the log of some function $g(x)$:  
\begin{equation}
\nabla_x \log g(x) = \frac{\nabla_x g(x)}{g(x)} \label{eq_logder}
\end{equation}
This results from simple application of the chain-rule. 

We will now expand Eq. \ref{eq_grad_of_exp}, where we midway apply the above log-derivative trick. 
\begin{align}
\nabla_\theta \mathbb{E}_{x \sim p_\theta(x)} [ f(x) ] &= \nabla_\theta \sum_x f(x) \cdot p_\theta(x) && \text{definition of expectation} \nonumber \\
&=  \sum_x f(x) \cdot  \nabla_\theta p_\theta(x) && \text{push gradient through sum} \nonumber \\
&=  \sum_x f(x) \cdot p_\theta(x) \cdot \frac{\nabla_\theta p_\theta(x)}{p_\theta(x)} && \text{multiply/divide by $p_\theta(x)$} \nonumber \\
&=  \sum_x f(x) \cdot p_\theta(x) \cdot \nabla_\theta \log p_\theta(x) && \text{log-der.\ rule (Eq. \ref{eq_logder})} \nonumber \\
&=  \mathbb{E}_{x \sim p_\theta(x)} [f(x) \cdot \nabla_\theta \log  
                                                                                                                                          p_\theta(x)]
                                                                                                      &&
                                                                                                         \text{rewrite into expectation} \nonumber  
\end{align}
What the above derivation essential does is {\it pushing the
  derivative inside of the sum}. This equally applies when
we change the sum into an integral. Therefore, for any $p_\theta(x)$,
we have: 
\begin{align}
\nabla_\theta \mathbb{E}_{x \sim p_\theta(x)} [ f(x) ] =
  \mathbb{E}_{x \sim p_\theta(x)} [f(x) \cdot \nabla_\theta \log
  p_\theta(x)] \label{eq_log_der_trick} 
\end{align}
This is known as the log-derivative trick, score function estimator,
or REINFORCE trick. Although the formula may look complicated, the
interpretation is actually  simple. We explain this idea in
Fig.~\ref{fig_reinforce}. In Sect.~\ref{sec:reinforce} we  apply this
idea to reinforcement learning.  

\begin{figure}[t]
  \centering
      \includegraphics[width = 1.0\textwidth]{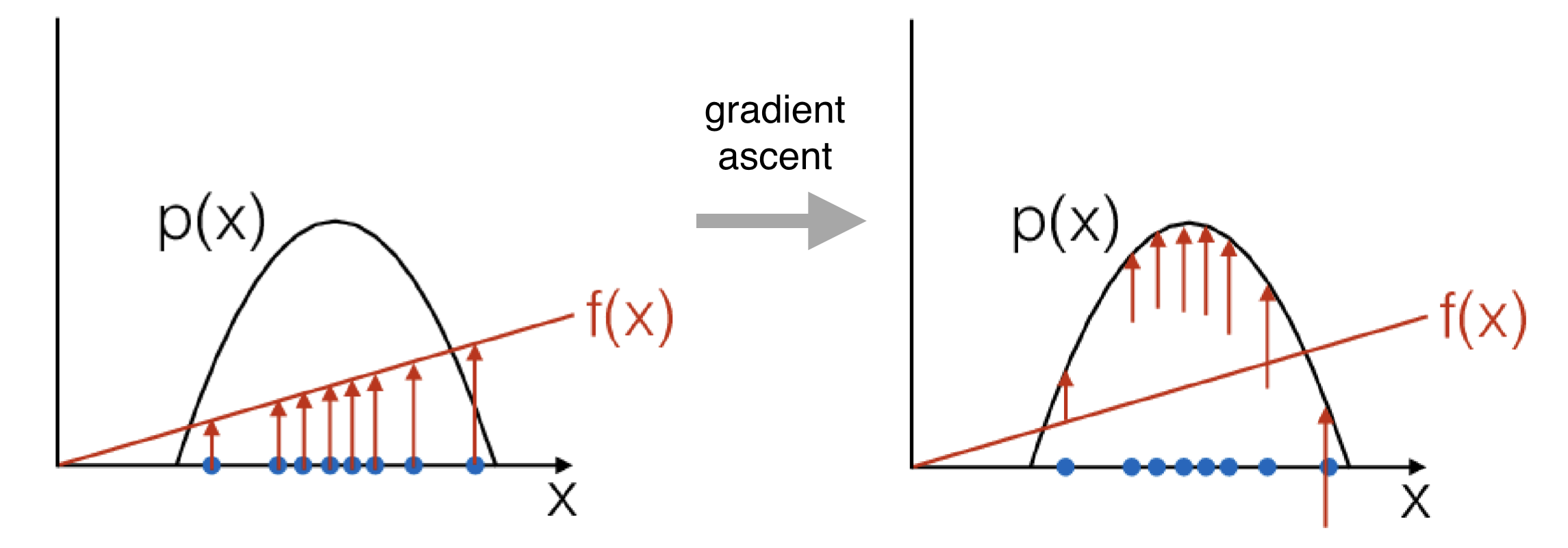}
  \caption[REINFORCE intuition]{Graphical illustration of REINFORCE
    estimator. Left: Example distribution $p_\theta(x)$ and function
    $f(x)$. When we evaluate the expectation of
    Eq. \ref{eq_log_der_trick}, we take $m$ samples, indicated by the
    blue dots (in this case $m=8$). The magnitude of $f(x)$ is shown
    with the red vertical arrows. Right: When we apply the gradient
    update, each sample pushes up the density at that location, but
    the magnitude of the push is multiplied by $f(x)$. Therefore, the
    higher $f(x)$, the harder we push. Since a density needs to
    integrate to 1, we will increase the density where we push hardest
    (in the example on the rightmost sample). The distribution will
    therefore shift to the right on this update.} 
    \label{fig_reinforce}
\end{figure}

\section{Bellman Equations}\label{sec:bellman}
Bellman noted that the value function can be written  in 
  recursive form, because the value is also defined at the next
states. In his work on dynamic programming~\cite{bellman1957dynamic},
he  derived recursive 
 equations for $V$ and $Q$.   

The Bellman equations for state-values and state-action values are: 
\begin{align}
  V^\pi(s) &= \mathbb{E}_{a \sim \pi(\cdot|s)} \mathbb{E}_{s'\sim T_a(s)} \big[r_a(s,s') + \gamma \cdot V^\pi(s') \big]  \nonumber
\end{align}
\begin{align}
Q^\pi(s,a) &= \mathbb{E}_{s' \sim T_a(s)} \big[ r_a(s,s') + \gamma \cdot \mathbb{E}_{a' \sim \pi(\cdot|s')} [Q^\pi(s',a')] \big] \nonumber 
\end{align}
Depending on whether the state and action space are discrete or continuous respectively, we write out these equations differently. 
For a discrete state space and discrete action space, we write the expectations as summations: 
\begin{align}
V(s) &= \sum_{a \in {A}} \pi(a|s) \Big[ \sum_{s'\in{S}} T_a(s,s') \big[ r_a(s,s') + \gamma \cdot V(s') \big] \Big] \nonumber
\end{align}
\begin{align}
Q(s,a) &= \sum_{s'\in{S}} T_a(s,s') \big[ r_a(s,s') + \gamma \cdot \sum_{a \in {A}} \pi(a|s) [Q(s',a')] \big] \nonumber 
\end{align}
For continuous state and action spaces, the summations over policy and transition  are replaced by integration: 

\begin{align}
V(s) &= \int_{a} \pi(a|s) \Big[ \int_{s'} T_a(s,s') \big[ r_a(s,s') + \gamma \cdot V(s') \big] \diff s' \Big]  \diff a \nonumber
\end{align}
The same principle applies to the Bellman equation for state-action values:

\begin{align}
Q(s,a) &=  \int_{s'} T_a(s,s') \big[ r_a(s,s') + \gamma \cdot \int_{a'} [ \pi(a'|s') \cdot Q(s',a')\big] \diff a' \Big]  \diff s' \nonumber
\end{align}
We may also have  a continuous state space (such as visual
input) with a discrete action space (such as pressing buttons in a
game):

\begin{align}
Q(s,a) &=  \int_{s'} T_a(s,s') \big[ r_a(s,s') + \gamma \cdot \sum_{a'} [ \pi(a'|s') \cdot Q(s',a')\big] \Big]  \diff s' \nonumber
\end{align}


\chapter{Deep  Supervised Learning}\label{ch:deep}\label{app:deep}

This appendix provides a chapter-length overview of essentials of
machine learning and deep learning.
Deep reinforcement learning uses much of the machinery of deep
supervised learning, and a good understanding of \gls{deep supervised learning}
is essential.
This appendix should provide you with the basics, in case your
knowledge of basic machine learning and deep learning is rusty. In
doubt? Try to answer the questions on page~\pageref{sec:dlq}. 

We will start with machine
learning basics. We will discuss 
training and testing, accuracy, the confusion matrix,
generalization, and overfitting.

Next, we will provide an overview of deep learning. We will  look into
neural networks, error functions, training by gradient descent,
end-to-end feature learning, and the curse of dimensionality. For neural networks
we will discuss convolutional networks, recurrent networks, LSTM, and
measures against overfitting.

Finally, on the practical
side, we will discuss TensorFlow, Keras, and PyTorch.
%
%
%
We will start with methods for solving large and complex problems.



\section{Machine Learning}
\label{sec:ml}
The goal of
machine learning is \emph{generalization}: to create an accurate, predictive, model of the
world. Such an accurate model is said to generalize well to the
world.\footnote{Generalization is closely related to the concepts of
  overfitting, regularization, and smoothness, as we will see in
  Sect.~\ref{sec:smooth}.}
In machine learning we operationalize 
this goal  with two datasets, a training set and a test set.

The field of machine learning aims to fit a function to approximate
an input/output relation. This can be, for example, a regression
function or a classification function. The most basic form of machine learning is
when a dataset \gls{data} of input/output pairs is given. We call this form \emph{supervised
learning}, since the learning process is \emph{supervised} by the output
values.

In machine learning we often deal with large problem domains. We are interested in
a  function that works not just on the particular values on which it was
trained, but also in the rest of the problem domain from which the
data items were taken.

In this section 
we will first see how such a learning process for
generalization works; for notation and  examples we follow
from~\cite{moerland2021lecture}. Next, we will discuss the specific problems of
large domains. Finally, we will discuss the phenomenon of overfitting
and how it relates to the bias-variance trade-off.

\subsection{Training Set and Test Set}
\label{sec:train}\label{sec:traintest}

A machine learning algorithm must learn a function $\hat{f}(x) \rightarrow
y$ on a training set from  input values $x$ (such as images) to approximate 
corresponding output values $y$ (such as image labels).   The goal
of the machine
learning algorithm is to learn this function $\hat{f}$, such that it performs
well on the training data and  generalizes well  to the test data.
Let us see how we can measure how well the machine learning methods generalize.


%




\begin{table}[t]
  \begin{center}
    \begin{tabular}{cc}
      \begin{tabular}{ll||cc|}
        & & \multicolumn{2}{c|}{Predicted class} \\
        & & P & N \\ \hline\hline
        Actual & P & TP & FN\\
        class  & N & FP & TN\\ \hline
      \end{tabular} \hspace{1cm} & \hspace{1cm}
      \begin{tabular}{ll||cc|}
        & & \multicolumn{2}{c|}{Predicted class} \\
        & & Cat & Dog \\ \hline\hline
        Actual & Cat & 122 & 8\\
        class  & Dog & 3 & 9\\ \hline
      \end{tabular}       
    \end{tabular}
  \end{center}
  \caption{Confusion Matrix}\label{tab:confusion}
\end{table}

\index{machine
  learning}\index{precision}\index{recall}\index{accuracy}\index{error
  value}\index{function approximation}\index{training set}\index{test set}\index{generalization}
The notions that are used to assess  the quality of the approximation are
as follows. Let us assume that our problem is a classification
task. The elements that our function correctly predicts are called the
true positives (TP). Elements that are correctly 
identified as not belonging to a class are the true
negatives (TN). Elements that are mis-identifed as belonging to a class are
the false positives (FP), and elements that belong to the class but that the
predictor misclassifies are the false
negatives (FN).

\begin{tcolorbox}
  {\bf Confusion Matrix}:
The number of true positives divided by the total number of
positives (true and false, TP/(TP+FP)) is called the \emph{precision} of the
classifier. The number of true positives divided by the size of the
class (TP/(TP+FN)) is the \emph{recall}, or the number of relevant elements that
the classifier could find. The term \emph{accuracy} is  defined
as the total number of true positives and negatives  divided by the
total number of predictions
(TP+TN)/(TP+TN+FP+FN): how well correct elements are
predicted~\cite{bishop2006pattern}. These numbers are often shown in
the form of a \emph{confusion matrix} (Table~\ref{tab:confusion}). The
numbers in the cells represent the number of true positives, etc., of
the experiment.
\end{tcolorbox}

In machine learning, we are often interested in the
\gls{accuracy} of a method on the training set, and whether the accuracy on
the test set is the same 
as on the training set (generalization).

In regression and classification  problems
the term \emph{error value} is used to indicate the difference between the
true output value $y$ and the predicted output value $\hat{f}(x)$.
\index{generalization}\index{$k$-fold cross validation}\index{cross validation}
Measuring how well a function  generalizes is typically performed using 
a method called $k$-fold cross-validation. This  works as
follows. When a  dataset of input-output examples $(x,y)$ is present,
this set is split into a large training set and a smaller hold-out test set,
typically 80/20 or 90/10. The approximator is trained on the
training set until a certain level of accuracy is achieved.  For
example, the approximator may be a neural network whose \gls{parameters}
\gls{theta}
are iteratively adjusted by gradient descent so that the error value  on the training set is reduced to a
suitably low value.
The approximator function is said to \emph{generalize} well, when
the  accuracy of the approximator  is about the same on
the test set as it is on the training set. (Sometimes a third dataset
is used, the validation set, to allow stopping before overfitting occurs, see Sect.~\ref{sec:earlystop}.)

Since
the test set and the training set contain examples that are drawn from
the same original dataset, 
approximators can be expected to be able to generalize
well. The testing is said to be \emph{in-distribution} when training
and test set are from the same distribution. 
Out-of-distribution generalization to different problems (or transfer learning)
is more difficult, see Sect.~\ref{sec:transfer}.

\subsection{Curse of Dimensionality}
\label{sec:curse}
\index{state space}\index{curse of dimensionality}

The \emph{state space} of a problem is the space of all possible different
states (combinations of  values of variables).
State spaces grow exponentially with the number of dimensions (variables); high-dimensional problems have large state spaces, and 
modern machine learning algorithms have to be 
able to learn functions in such large state spaces. 

\begin{tcolorbox}
{\bf Example}: A classic problem in AI is image
classification: predicting what type of object is pictured in an
image. Imagine that we have low-resolution greyscale
images of $100\times 100$ pixels, where each pixel takes a discrete value
between 0 and 255 (a byte). Then, the input space $X \in
\{0,1,\ldots,255\}^{100\cdot100}$, the machine learning  problem has dimensionality
$100\cdot100=10000$, and the state space has size $256^{10000}$.  
\end{tcolorbox}
When the input space $X$ is high dimensional, we can never
store the entire state space (all possible pixels with all
possible values) as a table. 
The effect of an exponential need for observation data as the
dimensionality grows, has been called the \emph{curse of dimensionality}
by Richard Bellman~\cite{bellman1957dynamic}.
The curse of dimensionality states that the cardinality (number of
unique points) of a space scales exponentially in the dimensionality
of the problem.  
In a formula, we have that $$|X| \sim \exp (
\text{Dim}(X)).$$
Due to the curse of dimensionality, the size of a table to 
represent a  function increases quickly when the size of the input
space increases. 

\begin{tcolorbox}
{\bf Example}: Imagine we have a discrete input space $X =
\{0,1\}^D$ that maps to a real number, $Y = \mathbb{R}$, and
we want to store this function fully as a table. We will show the required
size of the table and the required memory when we use 32-bit (4 byte)
floating point numbers:

\begin{center}
\begin{tabular}{P{0.3\textwidth}|P{0.3\textwidth}|P{0.3\textwidth}}
$\text{Dim}(X)$ & $|X|$ & Memory \\
\hline
1 & $2$ & 8 Byte \\
5 & $2^5 = 32$ &  128 Byte \\
10 & $2^{10} = 1024 $ & 4KB \\
20 & $2^{20} \approx 10^6 $ & 4MB\\
50 & $2^{50} \approx 10^{15} $ & 4.5 million TB\\
100 &$2^{100} \approx 10^{30} $ & $5\cdot 10^{21}$ TB \\
265 &$2^{265} \approx 2\cdot 10^{80} $ & - \\

\end{tabular}
\end{center}

The table shows how quickly exponential growth
develops. At a discrete space of size 20, it appears that we are
doing alright, storing 4 Megabyte of information. However, at a size
of 50, we suddenly need to store  4.5 million Terabyte. We can
hardly imagine the numbers that follow. At an input dimensionality of
size 265, our table would have grown to size $2\cdot 10^{80}$, which
is close to
the estimated number of atoms in the universe.  

\end{tcolorbox}
Since the size of the state space grows exponentially, but the number
of observations typically does not, most  state spaces
in large machine learning problems are sparsely populated with observations. An important challenge
for machine learning algorithms is to fit good predictive models on
sparse data. 
To reliably estimate a function, each variable needs a certain number
of observations. 
The number 
of samples that are needed to maintain statistical significance increases exponentially as the number of
dimensions grows.
A large amount
of training data is required to ensure that there are several samples
for each combination of values;\footnote{Unless we introduce some bias into the problem,
  by assuming smoothness, implying that there are dependencies between variables, and that
the ``true'' number
of independent variables is  smaller than the number of pixels.}
however, datasets rarely grow exponentially.

\subsection{Overfitting and the Bias-Variance Trade-Off}
\index{bias-variance trade-off}
\label{sec:hi-dim}\label{sec:high}

Basic statistics  
tells us that, according to the law of large numbers, the more
observations we have of an experiment, the more reliable the 
estimate of their value is (that is, the average will be close to the
expected value)~\cite{bishop2006pattern}. 
This   has important implications
for the study of large problems where we would like to have confidence in the estimated
values of our parameters.

In small toy problems
the number of variables of the model is also
small. Single-variable linear regression
problems model the function as a straight line $y=a\cdot x+b$, with
only one independent variable $x$ and two parameters $a$ and $b$. Typically, when regression is
performed with a small
number of independent variables, then a
relatively large number of observations is available per variable,
giving  confidence in the estimated parameter values.  A  rule of
thumb is that there should be  5 or more training examples for each
variable~\cite{theodoridis2009pattern}.

In machine learning dimensionality typically means the number of
variables  of a model. In statistics, dimensionality can be a relative concept:
the ratio of the number of variables  compared to the number of
observations. 
In practice, the size of
the observation dataset is limited, and then  absolute and relative
dimensionality do not differ much. In this book we follow the
machine learning definition of dimensions meaning variables. 

Modeling high-dimensional problems well typically requires a model with many
parameters, the so-called \emph{high-capacity} models.
Let us look deeper into the consequences of working with
high-dimensional problems.
%
%
%
\begin{figure}[t]
  \centering{\includegraphics[width=8cm]{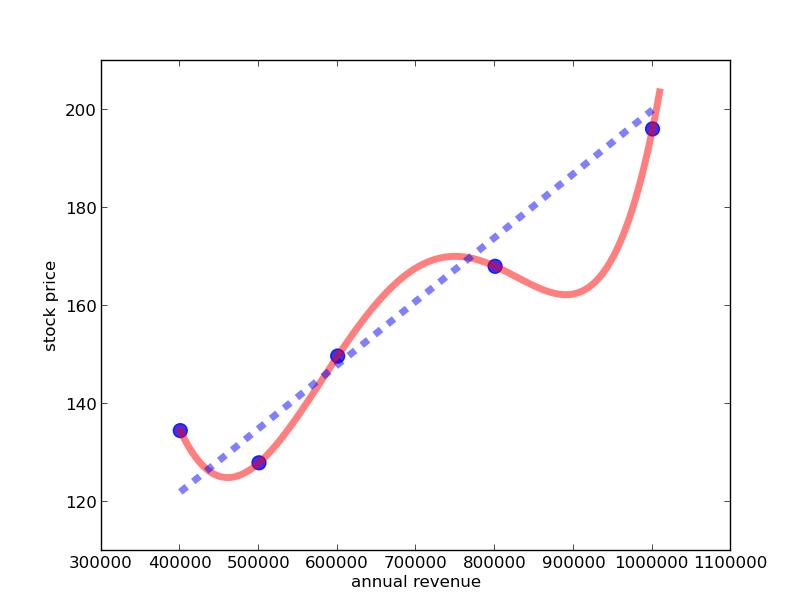}\caption[Overfitting]{Curve
      fitting:
    does the curvy red line or the straight dashed blue line best generalize the information in the data points?}\label{fig:overfitting}}
\end{figure}
%
\index{underfitting}\index{overfitting}
%
To see how to best fit our model, let us consider machine learning as a
curve fitting 
problem,  see Fig.~\ref{fig:overfitting}. In many problems, the observations are 
measurements of an underlying natural process. The observations
therefore contain some measurement noise. The goal is (1) that the
approximating curve fits the (noisy) observations as accurately as
possible, but, (2) in order to generalize well, to aim to fit the
signal, not the noise.

 \begin{table}[t]
 \begin{center}
   \begin{tabular}{ccccc}
     {\bf Environment  } &
     & {\bf Model}
     & {\bf Relative}
     & {\bf Chance of} \\ 
     {\bf  Observations } &
     & {\bf Variables}
     & {\bf Dimensionality }
     & {\bf } \\ \hline\hline
 $n$ & $>$ & $d$ & low & underfitting, high bias\\ 
 $n$ & $<$ & $d$ & high & overfitting, high variance\\ \hline \\
 \end{tabular}
 \caption{Observations, Variables, Relative Dimensionality, Overfitting}\label{tab:capacity}
 \end{center} 
 \end{table}

How complex should the approximator be to faithfully capture the
essence of a natural, noisy, process?  You can think of this question as: how many
parameters $\theta$ the network should have, or, if the approximator
is a polynomial, what the degree $d$ of the polynomial should be? The
complexity of the approximator is also called the capacity of 
the model, for the amount of information that it can contain (see
Table~\ref{tab:capacity}). 
When the capacity  of the approximator $d$ is lower than the
number of observations $n$, then the model is likely to \emph{underfit}: the
curve is too simple and cannot reach all observations well, the error
is large and the accuracy is low.\footnote{Overfitting can be reduced in the
loss function and training procedure, see Sect.~\ref{sec:overfitting}.}

Conversely, when the number of coefficients $d$ is as high or
higher than the number of observations $n$, then many machine learning
procedures wil be able to find a good fit. 
The error on the
training set will be zero, and training accuracy reaches $100\%$. Will this
high-capacity approximator generalize well to unseen states? Most likely it will not,
since the training and test observations are from a real-world process and
observations contain  noise from the training set. The training
noise will be modeled perfectly, and the trained function
will have low accuracy on the test set and other datasets. A high-capacity
model ($d>n$)  that trains well but tests poorly 
is said to \emph{overfit} on the training data.

\begin{figure}[t]
  \centering{\includegraphics[width=\textwidth]{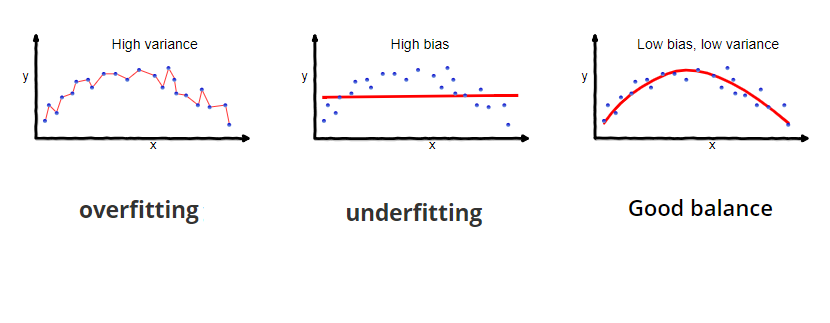}\caption[Bias-variance
    Trade-off]{Bias-variance
     trade-off; few-parameter model: high 
     bias;  many-parameter model: high variance}\label{fig:biasvariance}}
\end{figure}

Underfitting and \gls{overfitting} are related to the so-called bias-variance
trade-off, see Fig.~\ref{fig:biasvariance}.  High-capacity models
fit ``peaky'' high-variance curves that can fit signal and noise of the
observations, and tend to overfit. Low-capacity models fit
``straighter'' low-variance curves that have
higher bias and tend to underfit the observations. Models with a capacity of $d
\approx n$ can have both good bias and variance~\cite{bishop2006pattern}.

Preventing underfitting and overfitting is a matter of matching the capacity of our
model so that $d$ matches $n$. 
This is a delicate trade-off, since reducing the
capacity also reduces expressive power of the models.
To reduce overfitting, we can use regularization, and
many regularization methods  have 
been devised. Regularization has the effect of dynamically adjusting
capacity to the number 
of observations.

\subsubsection*{Regularization---the World is Smooth}\index{smoothness assumption}\label{sec:smooth}
Generalization is closely related to regularization. Most functions in the real world---the functions that we
wish to learn in machine learning---are smooth: near similar input
leads to near similar output.  Few real-world functions are jagged, or 
locally random (when they are, they are often contrived examples).

This has lead to the introduction of regularization methods, to
restrict or smooth the behavior of models. Regularization methods
allow us to
use high capacity models, learn
the complex function, and then later introduce a smoothing procedure, to reduce too
much of the randomness and jaggedness. 
Regularization may for example restrict the
weights of variables by moving the values closer to zero, but many
other methods exist, as we will see.
Different techniques for regularization have been developed for
high-capacity deep neural models, and we discuss them in
Sect.~\ref{sec:regularization}.  The goal is to filter out random
noise, but at the same time to allow meaningful trends to be recognized, to smooth the
function without restricting complex shapes~\cite{goodfellow2016deep}.

Before we delve into regularization methods, 
let us have a look in more detail at how to implement parameterized
functions, for which we will use   neural networks and deep learning.

\section{Deep Learning}
\label{sec:dl}
\index{neural networks}\label{sec:ann}

\begin{figure}[t]
\begin{center}
\includegraphics[width=7cm]{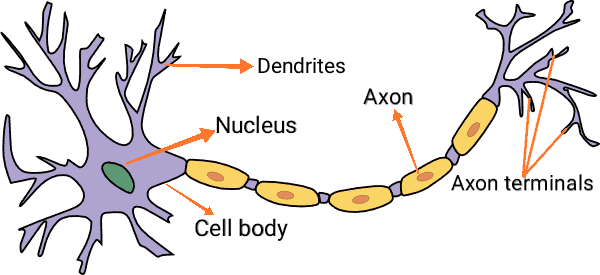}
\caption{A Single Biological neuron \cite{quoc2018}}\label{fig:bnn}
\end{center}
\end{figure}

The architecture of artificial neural networks is inspired by the
architecture of biological neural networks, such as the human
brain. Neural networks consist of neural core cells that are connected by
 nerve cells  \cite{bear2007neuroscience}. Figure~\ref{fig:bnn} shows a
drawing of a biological neuron, with a nucleus, an axon, and dendrites~\cite{quoc2018}.

Figure~\ref{fig:nn} shows a simple fully connected artificial neural
network, with an input layer of two neurons, 
an output layer of two neurons, and a single hidden layer of five neurons. A neural network with a single hidden layer
is called shallow. When the network has more  hidden layers
it is called deep (Fig.~\ref{fig:layers}).

\begin{tcolorbox}
In this section, we  provide an overview of artificial neural
networks and their training algorithms. We provide enough detail to
understand the deep reinforcement learning concepts in this
book. Space does provide a limit to how deep we can go.
Please refer to specialized deep learning literature for more
information, such
as~\cite{goodfellow2016deep}.

We provide a conceptual overview, which should be enough to
successfully use 
existing high quality deep learning packages, such as
TensorFlow\footnote{\url{https://www.tensorflow.org}}~\cite{abadi2016tensorflow}
or
PyTorch~\cite{paszke2019pytorch}.\footnote{\url{https://pytorch.org}}
\end{tcolorbox}

\begin{figure}[t]
\begin{center}
\newcommand{\inputnum}{2} 
\newcommand{\hiddennum}{5}  
\newcommand{\outputnum}{2} 

\begin{tikzpicture}

\foreach \i in {1,...,\inputnum}
{
	\node[circle, 
		minimum size = 6mm,
		fill=orange!30] (Input-\i) at (0,-\i) {};
}

\foreach \i in {1,...,\hiddennum}
{
	\node[circle, 
		minimum size = 6mm,
		fill=teal!50,
		yshift=(\hiddennum-\inputnum)*5 mm
	] (Hidden-\i) at (2.5,-\i) {};
}

\foreach \i in {1,...,\outputnum}
{
	\node[circle, 
		minimum size = 6mm,
		fill=purple!50,
		yshift=(\outputnum-\inputnum)*5 mm
	] (Output-\i) at (5,-\i) {};
}

\foreach \i in {1,...,\inputnum}
{
	\foreach \j in {1,...,\hiddennum}
	{
		\draw[->, shorten >=1pt] (Input-\i) -- (Hidden-\j);	
	}
}

\foreach \i in {1,...,\hiddennum}
{
	\foreach \j in {1,...,\outputnum}
	{
		\draw[->, shorten >=1pt] (Hidden-\i) -- (Output-\j);
	}
}

\foreach \i in {1,...,\inputnum}
{            
	\draw[<-, shorten >=1pt] (Input-\i) -- ++(-1,0)
		node[left]{$x_{\i}$};
}

\foreach \i in {1,...,\outputnum}
{            
	\draw[->, shorten >=1pt] (Output-\i) -- ++(1,0)
		node[right]{$y_{\i}$};
}

\end{tikzpicture}
\caption{A Fully Connected Shallow Neural Network with 9 Neurons}\label{fig:nn}
\end{center}
\end{figure}
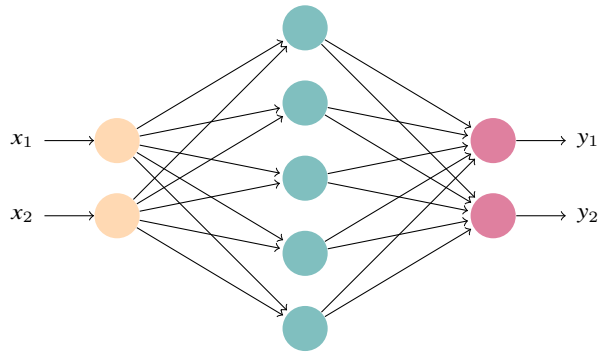

\begin{figure}[t]
  \centering{\includegraphics[width=10cm]{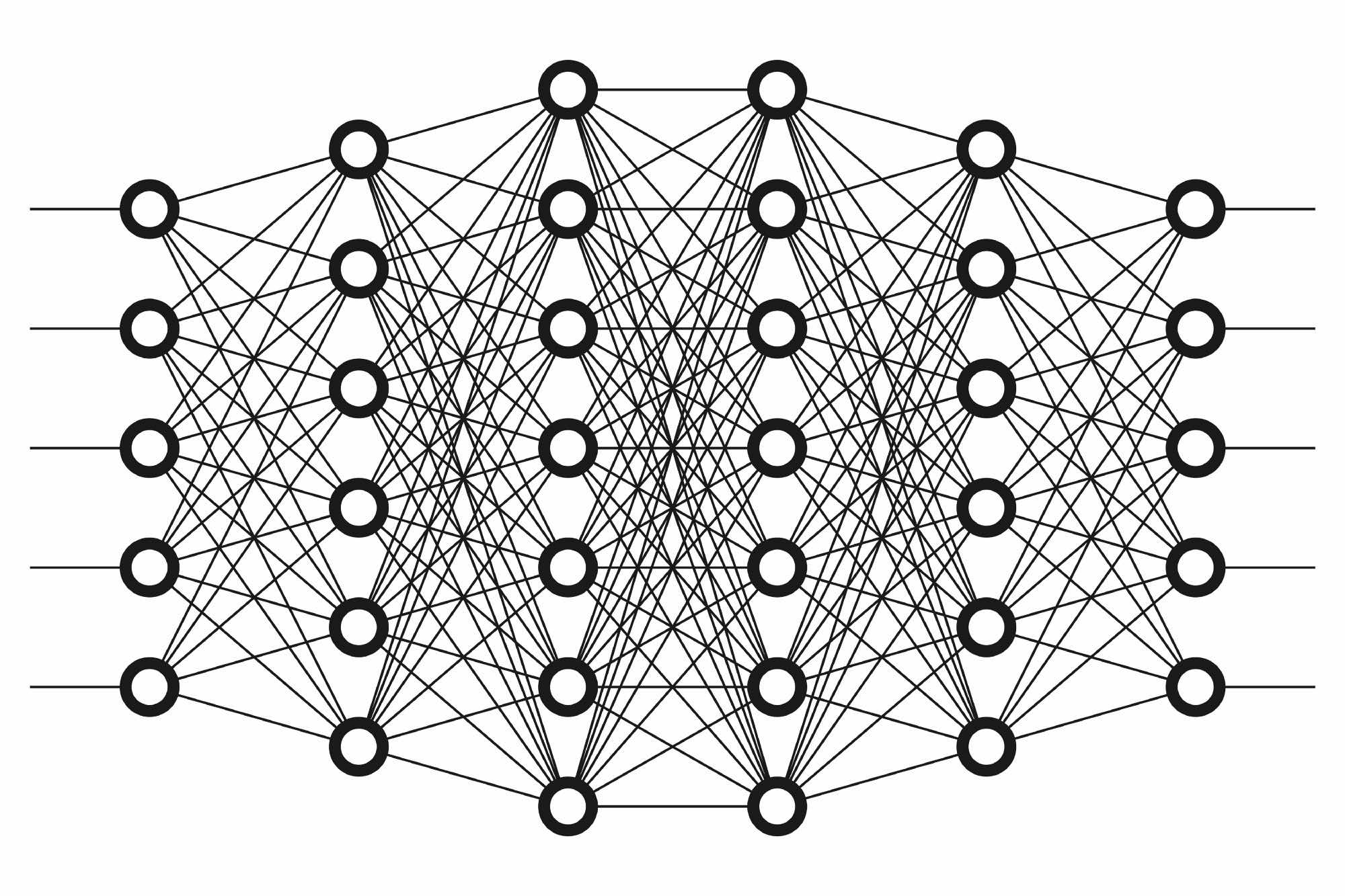}\caption{Four
      fully connected hidden layers~\cite{saha2018}}\label{fig:layers}}
\end{figure}


\subsection{Weights, Neurons}
\label{sec:algo}\index{backpropagation}\index{SGD}\index{epoch}

Neural networks consist of neurons and connections, typically
organized in layers. The neurons
process their input signals as a weighted
combination, producing an output
signal.  This calculation is called  the activation function, or
squashing function, since it is non-linear. Popular
activation functions are the rectified linear unit (ReLU: partly
linear, partly zero), the hyperbolic tangent, and the
sigmoid or logistic function $\frac{1}{1+e^{-a}}$ for neuron
activation $a$.
The neurons are connected
by  weights. At each neuron $j$ the incoming weights $ij$ are summed $\sum$
and then processed  by
the activation function $\sigma$. The output $o$ of neuron $j$ is therefore:  $$o_j = \sigma(\sum_i o_iw_{ij})$$  for weight $ij$ of
predecessor neuron $o_i$. The outputs of this layer of neurons are
fed to the
inputs for the weights of the next layer.

\subsection{Backpropagation}
The neural network as a whole is a parameterized  function $f_\theta(x)\rightarrow \hat{y}$ that
converts input to an output approximation. The behavior depends on the parameters
$\theta$, also known as the network weights. The parameters are adjusted such that the
required input-output relation is achieved (\emph{training} the
network). This is done by 
minimizing the error (or loss) function that calculates the
difference between the network output $\hat{y}$ and the training target $y$.

\lstset{label={lst:train}}
\lstset{caption={Network training pseudocode for supervised learning}}
\begin{lstlisting}[language=Python,float]
def train_sl(data, net, alpha=0.001):     # train classifier
    for epoch in range(max_epochs):       # an epoch is one pass 
        sum_sq = 0                 # reset to zero for each pass 
        for (image, label) in data: 
            output = net.forward_pass(image) # predict
            sum_sq += (output - label)**2 # compute error
        grad = net.gradient(sum_sq)       # derivative of error
        net.backward_pass(grad, alpha)    # adjust weights
    return net        
\end{lstlisting}

\index{gradient descent}
The training process consists of training {\em epochs,} individual passes in
which the network weights are optimized towards the target, using a
method called gradient descent (since the goal is to minimize the
error function). An epoch is
one complete  pass over the training data. Epochs are usually done in batches. Since training is an
iterative optimization process, it is typical to train for multiple
epochs. 
Listing~\ref{lst:train} shows  simplified pseudocode for
the gradient descent training algorithm (based
on~\cite{goodfellow2016deep}). When training starts, the weights of
the 
network are initialized to small random numbers. 
Each epoch consists of a forward pass
(recognition, usage)  and a backward pass (training, adjustment). The forward pass
is just the regular recognition operation for which the network is designed.
The input layer is exposed to the input
(the image), which is then propagated through the network to the output
layers, using the weights and activation functions. The output layer provides the answer, by
having a high value at the neuron corresponding to the right label
(such as Cat or Dog, or the correct number), so that an error  can be
calculated to be used to adjust the weights in the backward pass.

The listing
shows a basic version of gradient descent, that calculates the
gradient over all examples in the dataset, and then updates the weights. Batch versions of gradient descent update
the weights after smaller
subsets, and are typically quicker. 

\subsubsection*{Loss Function}
At the output
layer the propagated value $\hat{y}$ is compared with the other part of the 
example pair, the label $y$. The difference with the label is calculated, yielding
the {\em error}. The error function is also known as
the loss function \gls{loss}.
Two common error functions are the mean squared error $\frac{1}{n}\sum_{i}^n(y_i-\hat{y}_i)^2$ (for
regression) and the cross-entropy error $-\sum_i^M y_i\log \hat{y}_i$ (for classification
of $M$ classes).   
The backward pass uses the difference between the
forward recognition outcome and the true label to adjust the weights,  so that the
error becomes smaller. This method uses the gradient of the error
function over the weights,
and is called gradient descent.
The parameters are adjusted as follows:
$$\theta_{t+1}=\theta_t-\alpha\nabla_{\theta_t}{\cal L}_D(f_{\theta_t})$$
where $\theta$ are the network parameters, $t$ is the optimization time
step, $\alpha$ is the learning rate, $\nabla_{\theta_t}$ are the current
gradient of the loss function of data $\mathcal{L}_D$, and $f_\theta$ is the
parameterized objective function. 

The training process can be stopped when the error has been reduced below a
certain threshold for a single example, or when the loss on an entire validation set has dropped
sufficiently. More elaborate stopping criteria can be used 
in relation to overfitting (see Sect.~\ref{sec:overfitting}).

Most neural nets are trained using a stochastic
version of   gradient descent, or
SGD~\cite{song2018mean}. SGD samples a minibatch of size smaller than
the total dataset, and thereby computes a noisy estimate of the true
gradient. This is  faster per update step, and  does not
affect the direction of the gradient too much.  See Goodfellow et
al.~\cite{goodfellow2016deep} for details.
\index{stochastic gradient descent}\index{SGD}

\subsection{End-to-end Feature Learning}

\index{feature discovery}
Let us now look in more detail at how neural networks can be used to
implement end-to-end feature learning.

We can approximate a function  through the discovery of common
features in states. Let us, again, concentrate on image recognition.
Traditionally, feature discovery  was a manual
process.
Image-specialists
would painstakingly pour over images to identify 
common features in a dataset, such as lines, squares, circles, and
angles, by hand. They would 
write small pre-processing algorithms
to recognize the features that  were then used with classical machine learning
methods such as decision trees, support vector machines, or principal component analysis to construct
recognizers to classify an image.
This hand-crafted method is a labor-intensive and error prone
process, and researchers have worked to find algorithms for the full image
recognition process, end-to-end. For this to work,  also the features
must be learned.

For example, if the function approximator consists of the sum of $n$ features, then with hand-crafted features, only the coefficients
 $c_i$ of the features
in the  function $$h(s) = c_1\times f_1(s) + c_2\times f_2(s) + c_3\times f_3(s) +
\ldots + c_n\times f_n(s)$$ are learned. In end-to-end learning,  the
coefficients $c_i$
and the features $f_i(s)$ are learned.

\index{end-to-end learning}
Deep end-to-end learning has achieved great success in image recognition, speech
recognition, and  natural language
processing~\cite{krizhevsky2012imagenet,graves2013speech,xiong2018microsoft,devlin2018bert}.
End-to-end learning is the learning of a classifier directly from
high-dimensional, raw, un-pre-processed,
pixel data, all the way to the classification layer,  as opposed  to learning
pre-processed data from  intermediate (lower dimensional) hand-crafted
features.

\begin{figure}[t]
    \centering{\includegraphics[width=6cm]{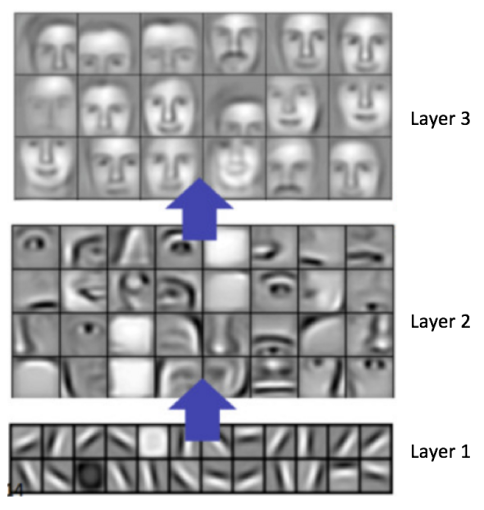}
     \caption{Layers of features of increasing complexity~\cite{lee2009convolutional}}\label{fig:leelayers}}
\end{figure}

We will now see in more detail how neural networks can perform automated feature
discovery.
The  hierarchy of network layers together can 
recognize a hierarchy of low-to-high level
concepts~\cite{lecun2015deep,lee2009convolutional}. For example, in
face recognition (Fig.~\ref{fig:leelayers}) the first hidden layer may 
encode edges; the second layer then composes and 
encodes simple structures of edges; the third layer may encode higher-level concepts such as noses or
eyes; and the fourth layer may work at the abstraction level of a
face. Deep feature learning finds what to abstract at
which level on its own~\cite{bengio2013representation}, and can come
up with classes of intermediate concepts, that
work, but look   counterintuitive upon 
inspection by humans.

\lstset{label={lst:lenet}}
\lstset{caption={LeNet-5 code in Keras~\cite{lecun1998gradient,weill}}}
\begin{lstlisting}[language=Python,float]

def lenet_model(img_shape=(28, 28, 1), n_classes=10, l2_reg=0.,
	weights=None):

	# Initialize model
	lenet = Sequential()

	# 2 sets of CRP (Convolution, RELU, Pooling)
	lenet.add(Conv2D(20, (5, 5), padding="same",
		input_shape=img_shape, kernel_regularizer=l2(l2_reg)))
	lenet.add(Activation("relu"))
	lenet.add(MaxPooling2D(pool_size=(2, 2), strides=(2, 2)))

	lenet.add(Conv2D(50, (5, 5), padding="same",
		kernel_regularizer=l2(l2_reg)))
	lenet.add(Activation("relu"))
	lenet.add(MaxPooling2D(pool_size=(2, 2), strides=(2, 2)))

	# Fully connected layers (w/ RELU)
	lenet.add(Flatten())
	lenet.add(Dense(500, kernel_regularizer=l2(l2_reg)))
	lenet.add(Activation("relu"))

	# Softmax (for classification)
	lenet.add(Dense(n_classes, kernel_regularizer=l2(l2_reg)))
	lenet.add(Activation("softmax"))

	if weights is not None:
		lenet.load_weights(weights)

	# Return the constructed network
	return lenet


\end{lstlisting}

Towards the end of the 1990s the work on neural networks moved into
deep learning, a term coined by 
Dechter in~\cite{dechter1986learning}.  LeCun et
al.~\cite{lecun1998gradient} published an influential paper on deep convolutional nets.  The paper introduced the architecture
LeNet-5,\index{LeNet-5}  a 
seven-layer convolutional neural net trained to classify handwritten MNIST
digits from $32\times 32$ pixel images. 
\index{CNN}\index{convolutional network}\index{MNIST}\index{Keras}
Listing~\ref{lst:lenet} shows a modern rendering of  LeNet in Keras. The code straightforwardly lists the layer definitions.

End-to-end learning is computationally quite demanding.
After the turn of the century,  methods, datasets,  and compute power had
improved to such an extent that full raw, un-pre-processed 
pictures could be learned, without the intermediate step of
hand-crafting features.
End-to-end learning proved very powerful,
achieving higher  accuracy in image recognition than previous methods,
and even higher than human test subjects~\cite{krizhevsky2012imagenet}.
In natural language
processing, deep transformer
models such as BERT and GPT-2 and 3 have  reached equally impressive results~\cite{devlin2018bert,brown2020language}.\index{BERT}\index{GPT-3}\index{AlexNet}

\begin{table}[t]
  \begin{center}
  \begin{tabular}{llll}
    & {\bf Function} & {\bf Input}  & {\bf Output}   \\
    \hline\hline
    Dataset & Regression & continuous (number) & continuous (number) \\
     & Classification & discrete (image) & discrete (class label)\\ \hline
    Environment& Value $V$ & state & continuous (number)\\
    & Action-value $Q$& state $\times$ action& continuous (number)\\
    & Policy $\pi$ & state & action(-distribution)\\
    \hline
  \end{tabular}
  \caption{Functions that are Frequently Approximated}\label{tab:func-approx}
\end{center}
\end{table}

\subsubsection*{Function Approximation}
Let us have a look at the different kinds of functions that we wish to
approximate in machine learning.
%
\label{sec:func-approx}
The most basic function
establishing an input/output relation is regression, which  outputs a
continuous number. Another important function is classification, which
outputs a discrete number.
Regression and classification are often
learned through supervision, with a dataset of examples (observations)
and labels.

In reinforcement learning, three functions are typically
approximated: the value function $V(s)$, that relates states to their
expected cumulative future rewards, the action-value function $Q(s,a)$
that relate actions to their values, and the policy function $\pi(s)$
that relates states to an action (or $\pi(a|s)$ to an action
distribution).\footnote{In
  Chap.~\ref{chap:model}, on model-based learning, we  also
  approximate the transition function $T_a(\cdot)$ and the reward
  function $R_a(\cdot)$.} In reinforcement learning, the functions $V, Q, \pi$
are learned through reinforcement by the environment.
Table~\ref{tab:func-approx} summarizes these functions.

\subsection{Convolutional Networks}\index{convolutional networks}

The first neural networks consisted of fully connected layers (Fig.~\ref{fig:layers}).
In image recognition, the input layer of a neural network is typically
connected directly to the input image. Higher resolution images
therefore need a higher number of input neurons. If all layers would
have more neurons, then the width of the network grows quickly. Unfortunately, growing a fully connected network (see Fig.~\ref{fig:nn}) by
increasing its width (number of neurons
per layer)  will increase the number of parameters quadratically.

The naive solution of high-resolution problem learning is to increase
the capacity of the model $m$. 
However, because of the problem of overfitting, as $m$ grows, so must
the number of examples, $n$. 

The solution lies in using a sparse interconnection structure instead
of a fully connected network.
Convolutional neural nets (CNNs) take their inspiration from biology. 
The
visual cortex in animals and humans is not fully connected, but
locally connected~\cite{hubel1963shape,hubel1968receptive,matsugu2003subject}. Convolutions
efficiently exploit  prior knowledge about the structure of the
data: patterns reoccur at different locations in the data (translation
invariance), and therefore we can share parameters by moving a
convolutional window over the image.  

A CNN consists
of convolutional operators or filters. A typical convolution operator
has a small {\em receptive field\/} (it only connects to a limited number of
neurons, say $5\times 5$), whereas a fully connected neuron connects to
all neurons in the layer below. Convolutional filters detect the
presence of local patterns. The next layer thus acts as a feature
{\em map}.  A CNN layer can be seen as a set of learnable 
filters, invariant for local transformations~\cite{goodfellow2016deep}. 

Filters can be used to identify  features. Features are basic
elements such as edges,  straight lines, round lines, curves, and 
colors. To work as a curve detector, for example, the  filter should
have a pixel 
structure with high values  indicating a shape of a curve. By
then multiplying and adding these filter values with the pixel values,
we can detect whether the shape is present.  The sum of the 
multiplications in the input
image  will be large if there is a shape that resembles the curve in the filter.

This filter can only detect a certain shape of curve. Other
filters can detect other shapes. 
Larger activation maps can 
recognize more elements in the input image. Adding more filters 
increases the size of the network, which effectively enlarges the
activation map. 
The filters in the first network layer process (``convolve'') the
input image and fire (have high values) when a specific feature that it is
built to detect is in the input image. Training a convolutional net is
training a filter that consists of layers of subfilters.

By going through the convolutional layers of the network, increasingly
complex features can be represented in the activation maps.
%
%
 Once they are trained, they can be used for as many recognition tasks
 as needed. A recognition task consists of a single quick forward pass
 through the network.

Let us spend some more time on understanding these filters.

\subsubsection*{Shared Weights}
\index{weight sharing}\index{shared weights}
In CNNs the filter parameters are shared in a layer. Each layer thus
defines a filter operation. A filter is
defined by few parameters but is applied to many pixels of the image; 
each filter is replicated across the entire
visual field. These replicated units share the same parameterization
(weight vector and bias) and form a feature map. This means that all
the neurons in a given convolutional layer respond to the same feature
within their specific response field. Replicating units in this way
allows for features to be detected regardless of their position in the
visual field, thus constituting the property of translation invariance.

This weight sharing is also
important to prevent an increase in the number of weights in deep and
wide nets, and to 
prevent overfitting,  as we shall see later.

Real-world images  consist of repetitions of many smaller elements. Due to this
so-called {\em translation invariance}, the same patterns reappear
throughout an image. 
CNNs can take advantage of this. The weights of 
the links are 
shared, resulting in a large reduction in the number of weights that
have to be trained. Mathematically CNNs put
constraints on what the weight values can be. This is a significant advantage of CNNs, since the
computational requirements of training the weights of  fully
connected layers are prohibitive. In addition, statistical
strength is gained, since the effective data per weight increases.

Deep CNNs work well in image recognition tasks, for visual filtering
operations in spatial dependencies, and for feature recognition
(edges, shapes)~\cite{lecun1989backpropagation}.\footnote{Interestingly, this paper
was already published in 1989. The deep learning revolution happened
twenty years later, when publicly available datasets, more efficient
algorithms, and more compute power in the form of GPUs were available.}

\subsubsection*{CNN Architecture}
Convolutions recognize features---the deeper the network, the more
complex the features. A typical
CNN architecture consists of a number of stacked convolutional
layers. In the
final layers, fully connected layers are used to then classify the inputs.

In the convolutional layers, by connecting only locally, the number of
weights is dramatically 
reduced in comparison with a fully connected net. The ability  of a
single  neuron  to
recognize different features, however, is less than that of a fully connected
neuron. 

By stacking many such locally connected layers on top of each other we
can achieve the desired nonlinear 
filters whose joint effect becomes increasingly global.\footnote{Nonlinearity is essential. If all neurons performed
  linearly, then there would be no need for layers.
  Linear  recognition functions cannot discriminate between  cats and
  dogs.}
The neurons become responsive to a larger
region of pixel space, so that the network first creates
representations of small parts of the input, and  from these
representations create
larger areas. The network  can 
recognize and represent increasingly complex concepts without an
explosion of weights.

\begin{figure}[t]
  \centering{\includegraphics[width=12cm]{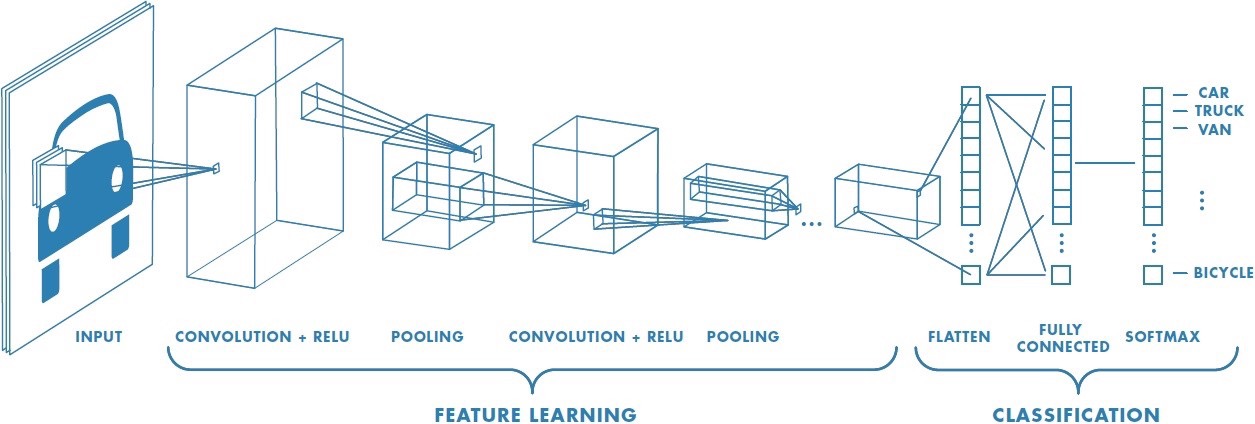}\caption{Convolutional
      network example architecture~\cite{saha2018}}\label{fig:cnn}}
\end{figure}

A typical CNN architecture consists of an architecture of multiple
layers of convolution, max pooling, and ReLU layers, topped off by a
fully connected layer (Fig.~\ref{fig:cnn}).\footnote{Often with a softmax function. The
  softmax function normalizes an input vector of real numbers to a
  probability distribution $[0,1]$; $ p_\theta(y|x) = \text{softmax}(f_\theta(x)) = \frac{e^{f_{\theta}(x)}}{\sum_k e^{f_{\theta,k}(x)}} $}\index{softmax}\label{sec:softmax}

\begin{figure}[t]
  \centering{\includegraphics[width=6cm]{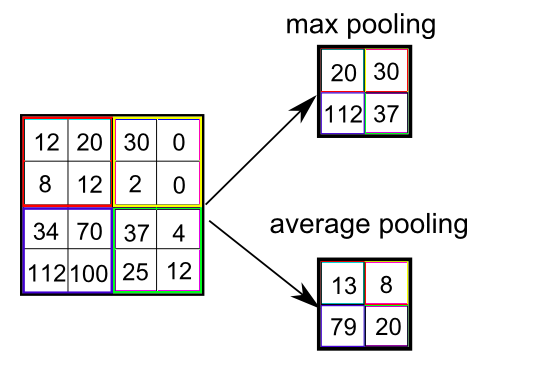}\caption{Max
      and average $2\times 2$ pooling~\cite{goodfellow2016deep}}\label{fig:maxpooling}}
\end{figure}

\subsubsection*{Max Pooling}\index{max pooling}
A further method for reducing the number of weights is weight pooling.
Pooling is a kind of
nonlinear downsampling (expressing the information in lower
resolution with fewer bits).  Typically, a $2\times 2$ block is
sampled down to a scalar
value (Fig.~\ref{fig:maxpooling}). Pooling reduces the dimension of the
network. The most frequently used form is max pooling. It
is an important component for object
detection~\cite{cirecsan2012multi} and is an integral part of most CNN
architectures.
Max pooling also allows small translations, such as shifting the object
by a few pixels, or scaling, such as putting the object closer to the camera.

\subsection{Recurrent Networks}\index{RNN}\index{recurrent
  neural network}\index{captioning challenge}\index{LSTM}\index{text to image}\label{sec:lstm}
Image recognition  has had a large
impact on network architectures, leading to  innovations  such as convolutional nets for spatial data.

Speech recognition and   time series analysis   have also caused new architectures to
be created, for  sequential data. Such sequences can be
modeled by recurrent neural
nets (RNN)~\cite{bertolami2009novel,fernandez2007application}. Some of the better known RNNs are Hopfield
networks~\cite{hopfield1982neural},   and long
short-term memory (\gls{LSTM})~\cite{hochreiter1997long}.

\begin{figure}[t]
\begin{center}
\includegraphics[width=2cm]{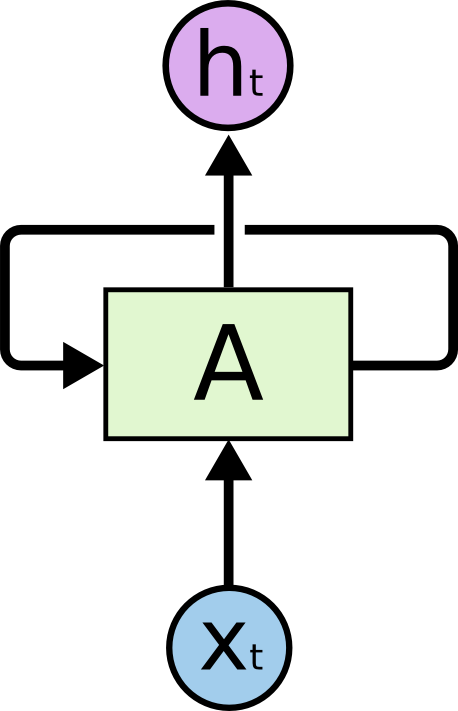}
\caption[RNN]{RNN $x_t$ is an input vector, $h_t$ is the
  output/prediction, and $A$ is the RNN \cite{colah2015}}\label{fig:RNN-rolled}
\end{center}
\end{figure}


Figure~\ref{fig:RNN-rolled} shows a basic recurrent neural network. An
RNN neuron is the same as a normal neuron, with input, output, and
activation function. However, RNN neurons have an extra pair of looping input/output
connections. 
Through this structure, the values of the parameters
in an RNN can evolve. In effect, RNNs have a variable-like state.

\begin{figure}[t]
\begin{center}
\includegraphics[width=10cm]{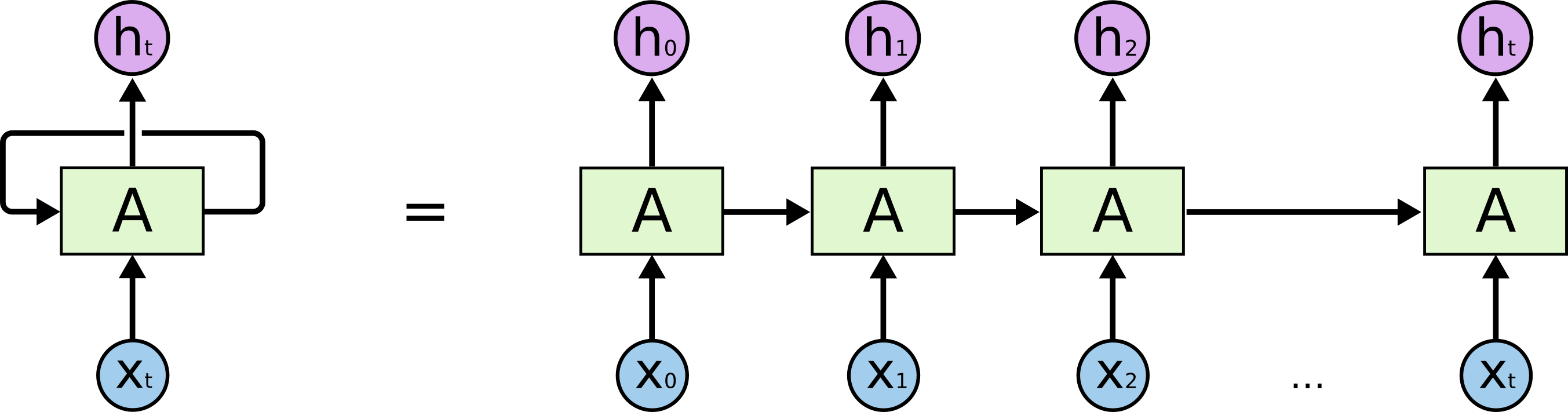}
\caption{RNN unrolled in time \cite{colah2015}}\label{fig:RNN-unrolled}
\end{center}
\end{figure}


To understand how RNNs work, it helps to unroll
the network, as has been done in Fig.~\ref{fig:RNN-unrolled}. The
recurrent neuron loops have 
been drawn as a straight line to show the network in a deeply layered style, with 
connections between the layers. In reality the layers are  time steps in
the processing of the recurrent connections.
In a sense, an RNN is a deeply layered neural net folded into a single layer
of recurrent neurons. 

Where deep convolutional networks are successful
in image classification, RNNs are used for tasks with a sequential nature, such
as captioning challenges. In a captioning task the network is shown a
picture, and then has to come up with a textual description that makes
sense~\cite{vinyals2015show}. 

\begin{figure}[t]
\begin{center}
\includegraphics[width=12cm]{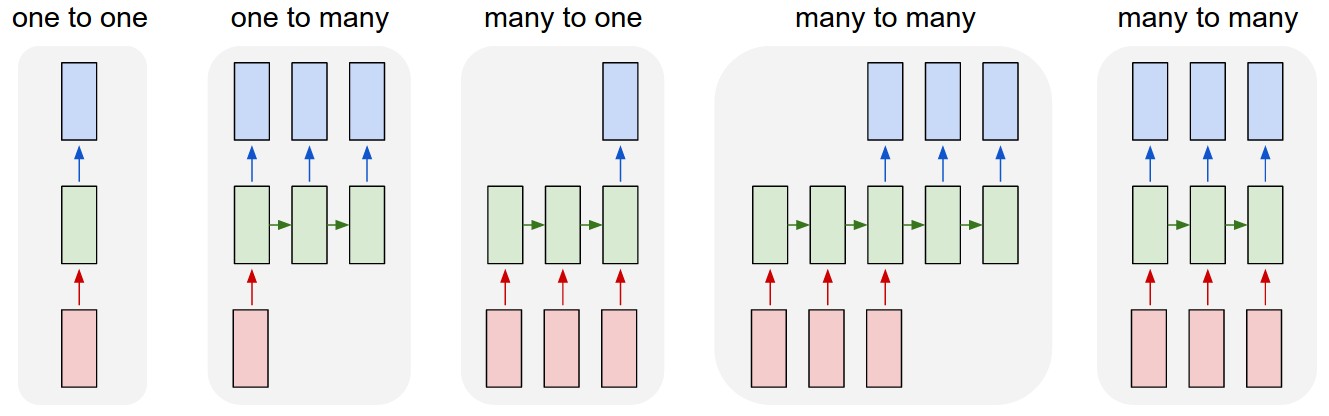}
\caption{RNN configurations \cite{karpathy2015}}\label{fig:RNN-many}
\end{center}
\end{figure}


The main innovation of  recurrent nets is that they allow us to work
with sequences of vectors. Figure~\ref{fig:RNN-many} shows
different 
combinations of sequences that we will discuss now, following an accessible and
well-illustrated blog on the different RNN
configurations written by Karpathy~\cite{karpathy2015}. There can be sequences in the input, in
the output, or in both. 
The figure shows different rectangles. Each rectangle is a vector. 
Arrows represent computations, such as matrix
multiply. Input vectors are in red, output vectors are in blue, and
green vectors hold the state. From left to
right we see:
\begin{enumerate}
  \item  {\em One to one\/}, the standard network  without RNN. This network
maps a fixed-sized input to fixed-sized output, such as an image
classification task (picture in/class out).
\item {\em One to many\/} adds a sequence in the output. This can be an
image captioning task that takes an image and outputs a sentence of 
words.
\item {\em Many to one\/} is the opposite, with a sequence in the input. Think for example of
sentiment analysis (a sentence is classified for words with negative
or positive emotional meaning).
\item {\em Many to many\/} has both a sequence for input and
a sequence for output. This can be the case in machine
translation, where a sentence in English is read and then a sentence
in Fran\c{c}ais is produced.
\item {\em Many to many\/} is a related but different situation, with synchronized
input and output sequences. This can be the case in video 
classification where  each frame of the video should be labeled.
\end{enumerate}


\begin{figure}[t]
\begin{center}
\includegraphics[width=7cm]{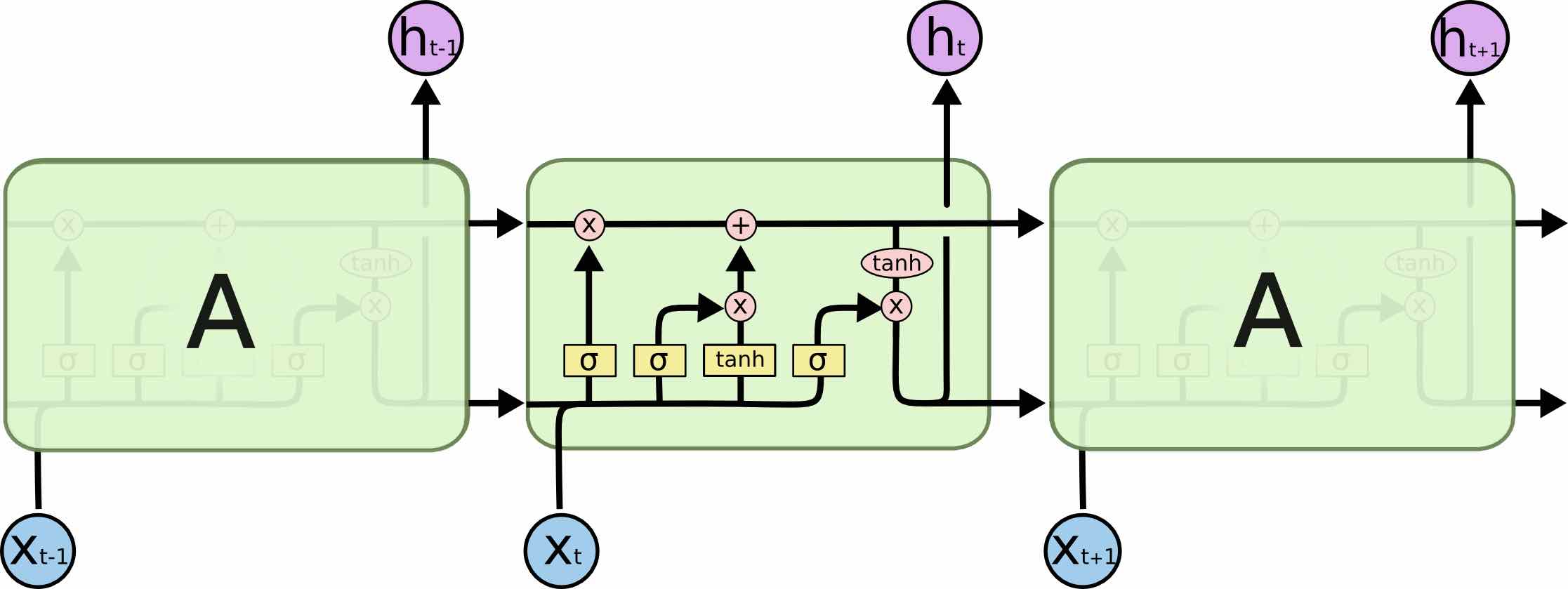}
\caption{LSTM \cite{karpathy2015}}\label{fig:LSTM3}
\end{center}
\end{figure}

\subsubsection*{Long Short-Term Memory}\index{LSTM}
Time series prediction is more complex than  conventional regression
or classification. It adds  the complexity of a sequence dependence among the input
variables.
 
LSTM  (long short-term memory) is a more  powerful type of neuron designed to handle
sequences. Figure~\ref{fig:LSTM3} shows the LSTM module,  allowing
comparison with a simple RNN. 
LSTMs are designed for sequential problems, such as time series, and
planning.
LSTMs were introduces by Hochreiter and
Schmidhuber~\cite{hochreiter1997long}.

 RNN training suffers from the vanishing gradient problem. For
 short-term  sequences this problem may be controllable by the same
 methods 
 as for deep CNN~\cite{goodfellow2016deep}. For  
 long-term remembering LSTM are better suited.
%
%
LSTMs are frequently used to solve diverse
problems~\cite{schmidhuber2002learning,graves2005bidirectional,greff2017lstm,gers1999learning},
and we  encounter them at many places throughout this book.

\subsection{More Network Architectures}\label{sec:residual}\label{sec:resnet}
Deep learning is a highly active field of research, in which many
advanced network architectures have been developed. We will describe
some of the better known architectures.


\subsubsection*{Residual Networks}
\begin{figure}[t]
    \centering{\includegraphics[width=11cm]{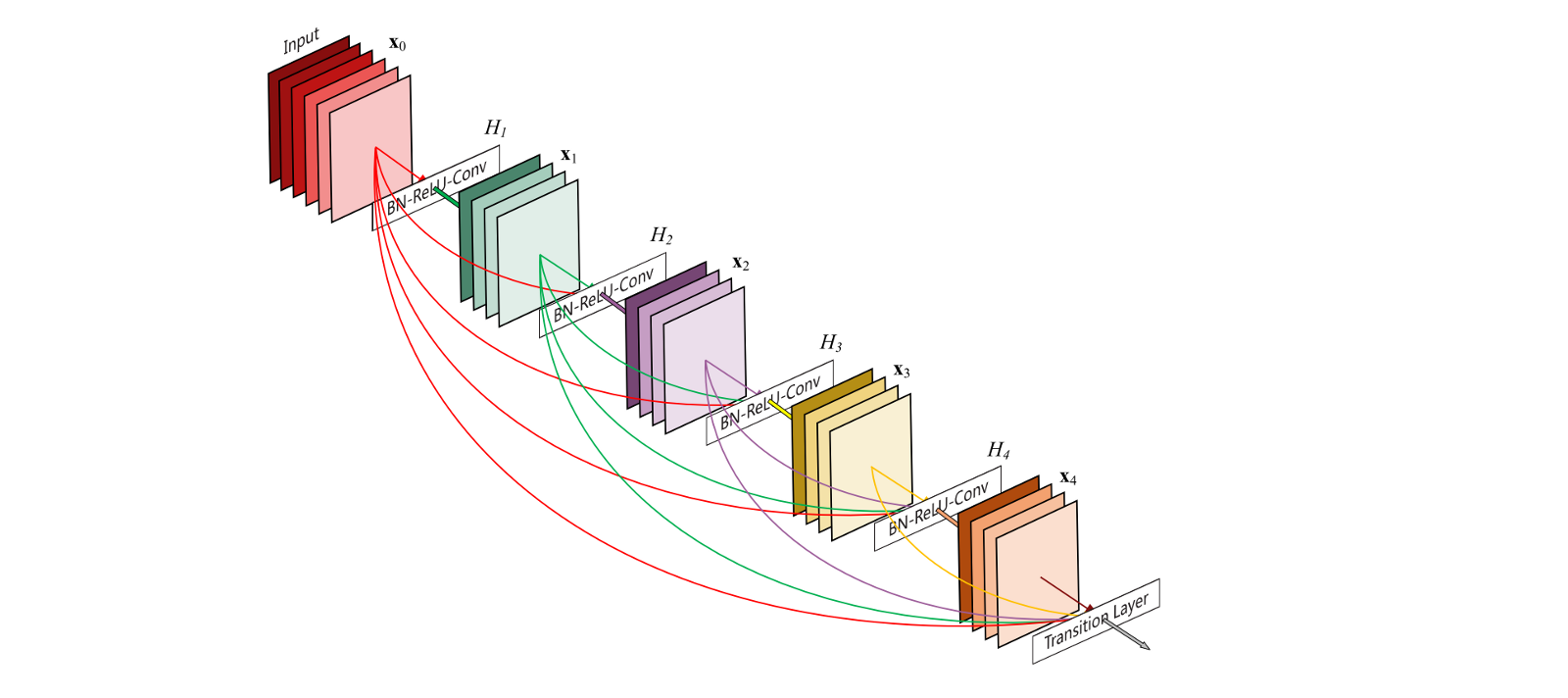}
     \caption{Residual Net with Skip Links~\cite{he2016deep}}\label{fig:resnet}}
 \end{figure}


An important innovation on CNNs is the residual network architecture,
 or {\em ResNets}. This idea
 was introduced in the 2015 ImageNet challenge, which He et al.~\cite{he2016deep} won with a
 very low error rate of 3.57\%. This error rate is actually lower than
 what most humans achieve:
 5--10\%.\index{ResNet}\index{residual network} ResNet has
 no fewer than 152 layers.

Residual nets  introduce {\em skip links\/}. Skip links are connections skipping one or
 more layers, allowing the training to go directly to other layers,
 reducing the effective depth of the network
 (Fig.~\ref{fig:resnet}). Skip links create a mixture of a shallow and a deep
network, preventing the accuracy degradation and vanishing gradients
of deep networks~\cite{goodfellow2016deep}.



\subsubsection*{Generative Adversarial Networks}
\index{deep fake}\label{sec:onepixel}\label{sec:gan}\index{generative adversarial networks}
Normally neural networks are used in  forward mode, to discriminate input
images into classes, going from high dimensional to low
dimensional. Networks can also be run backwards, to \emph{generate} an image
that 
goes with a certain class, going from low dimensional to high dimensional.\footnote{Just like the decoding phase of
 autoencoders.} Going from small to large implies many
possibilities for the image to be instantiated.  Extra
input is needed to fill in the degrees of freedom.



\begin{figure}[t]
  \centering{\includegraphics[width=\textwidth]{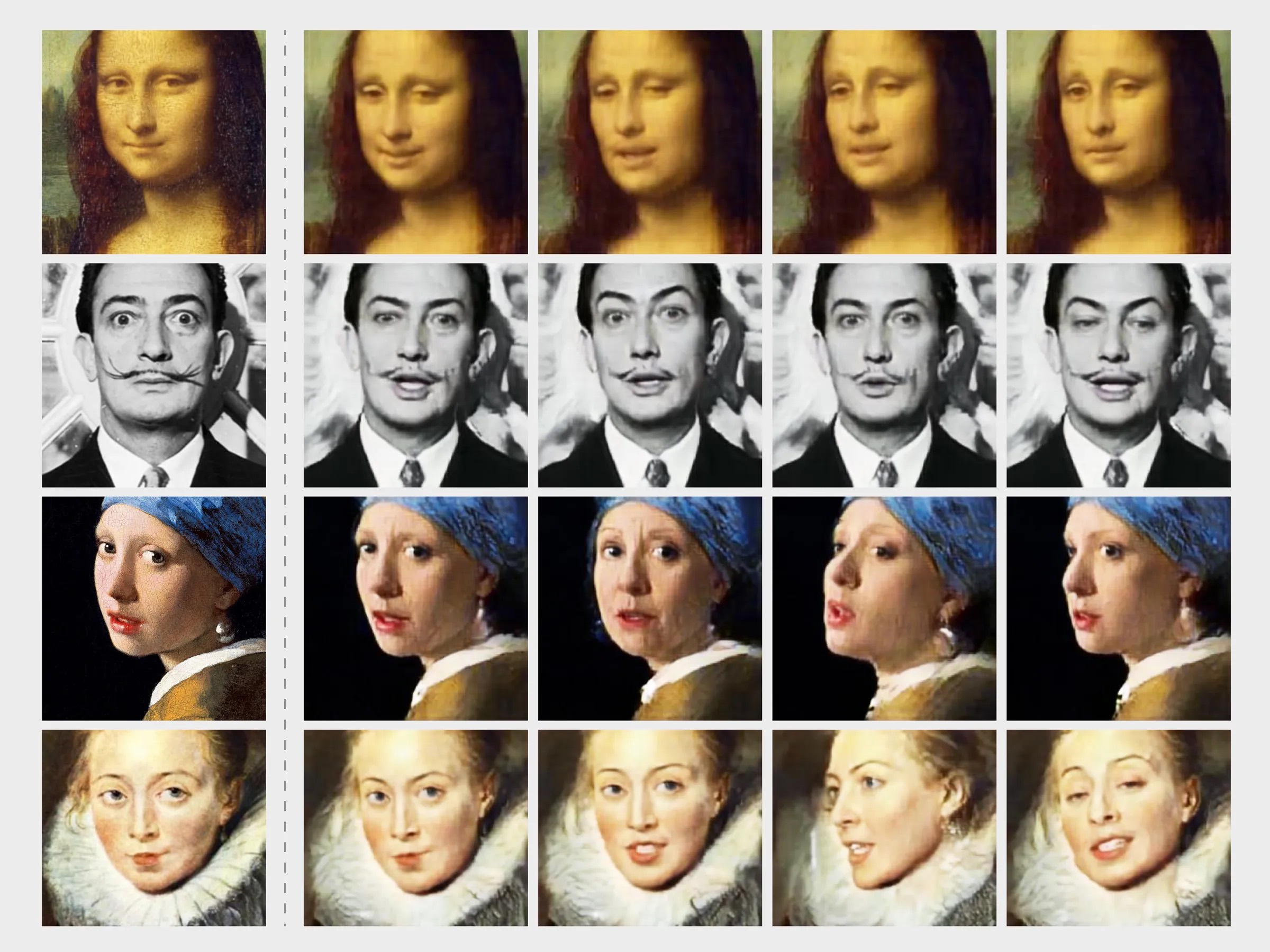}
     \caption{Deep fakes~\cite{yang2019exposing}}\label{fig:deepfake}}
\end{figure}

\begin{figure}[t]
    \centering{\includegraphics[width=\textwidth]{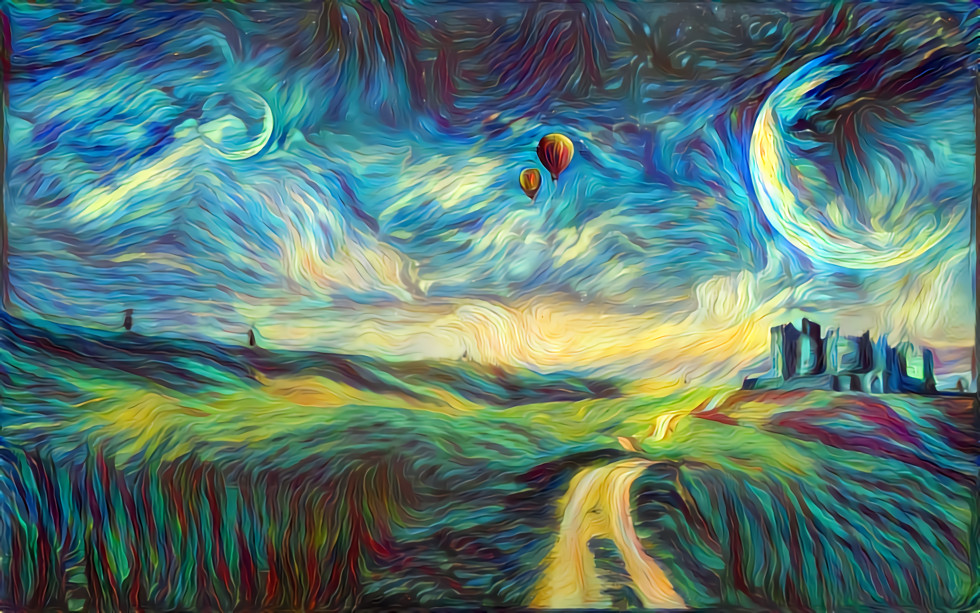}
     \caption{Deep dream~\cite{karras2017progressive}}\label{fig:deepdream}}
\end{figure}

Running the recognizers
backwards, in generative mode, has created an active research area called deep generative
modeling. An important  type of generative model  that has made quite an impact
is the Generative adversarial network, or
\gls{GAN}~\cite{goodfellow2014generative}.\index{GAN}

Deep networks are susceptible 
to adversarial attacks.
A well-known problem of the image recognition process is that it is brittle.
It was found that if an image is slightly perturbed,
and imperceptibly to the human 
eye, deep networks can easily be fooled to characterize an image as the wrong
category~\cite{szegedy2013intriguing}.\index{adversarial attack}\index{one-pixel problem} This brittleness is known as the
one-pixel problem: changing a single unimportant pixel in an image
could cause classifiers to switch from  classifying an image from cat to
dog~\cite{su2019one,szegedy2013intriguing}.

GANs are used to generate input images that are slightly different
from the original input. 
GANs generate adversarial examples whose  purpose it is 
to fool the discriminator (recognizer). 
The first network, the generator, generates an image. The second network, the
discriminator, tries to recognize the image. The goal for the generator is
to mislead the discriminator, in order to improve the robustness of
the  discriminator.
In this way, GANs can be used to make
image recognition more robust.   The
one-pixel problem   has spawned an active area of research to
understand this problem, and to make deep networks more robust.\index{brittleness}

Another use of 
generative networks is to generate
artificial photo-realistic images known as {\em deep fake\/}
images~\cite{yang2019exposing} and {\em deep
  dreaming}~\cite{karras2017progressive}, see Fig.~\ref{fig:deepfake}
and
Fig.~\ref{fig:deepdream}.\footnote{\href{https://deepdreamgenerator.com}{Deep
    Dream Generator} at \url{https://deepdreamgenerator.com}}
GANs have significantly increased our theoretical
understanding of deep learning.\index{deep fake}\index{deep dreaming}

\subsubsection*{Autoencoders}\label{sec:vae}
\begin{figure}[t]
    \centering{\includegraphics[width=9cm]{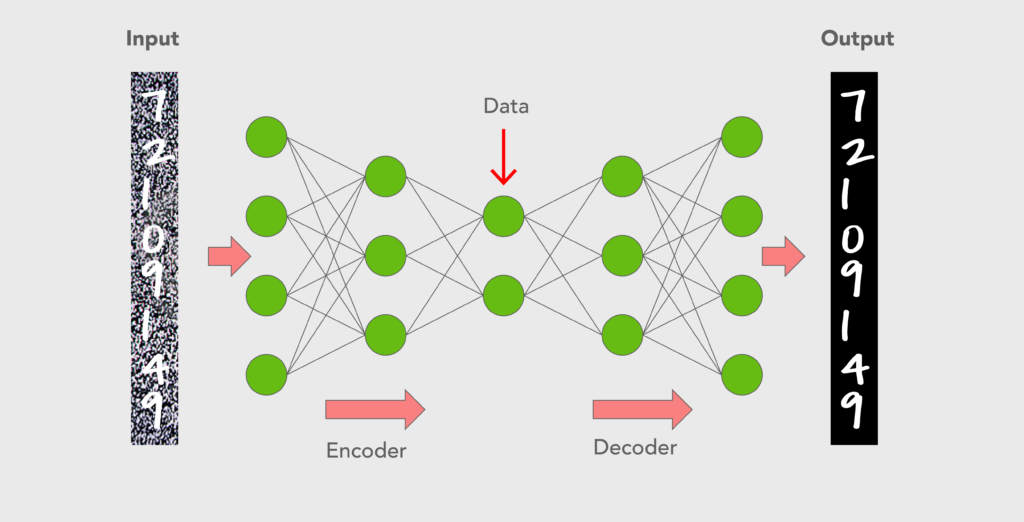}
     \caption{Autoencoder, finding the ``essence'' of the data in the
       middle \cite{mujtaba2020}}\label{fig:vae}}
 \end{figure}

\index{VAE}\index{variational
  autoencoder}\index{autoencoder}\index{unsupervised learning}
Autoencoders and variational autoencoders (VAE) are used in deep learning for
dimensionality reduction, in unsupervised learning~\cite{kramer1991nonlinear,kingma2013auto,kingma2019introduction}. An autoencoder network has  a
butterfly-like architecture, with the same number of neurons in the
input and the output layers, but a decreasing layer-size as we go to
the center (Fig.~\ref{fig:vae}). The input (contracting) side is said to perform a
discriminative action, such as image classification, and the other
(expanding) side is generative 
~\cite{goodfellow2014generative}. When an image is fed to  both the
input and the output of the autoencoder,  results in the
center layers are exposed to  a compression/decompression process,
resulting 
in the same image, only smoothed. The discriminative/generative
process performed by  autoencoders reduces the
dimensionality, and is generally thought of as going to the  ``essence'' of a problem,
de-noising it~\cite{hinton2006reducing}.

Autoencoding illustrates a deep relation between supervised and
unsupervised learning. The architecture of an autoencoders consists of
an encoder and a decoder. The encoder is a regular discriminative
network, as is common in supervised learning. The decoder is a
generative network, creating high dimensional output from low
dimensional input. Together, the encoder/decoder butterfly perform
dimensionality reduction, or compression, a form of
unsupervised learning\cite{grunwald2007minimum}.

Autoencoders and generative networks are an active area of research.

\subsubsection*{Attention Mechanism}\index{attention networks}

\begin{figure}[t]
\begin{center}
\includegraphics[width=\textwidth]{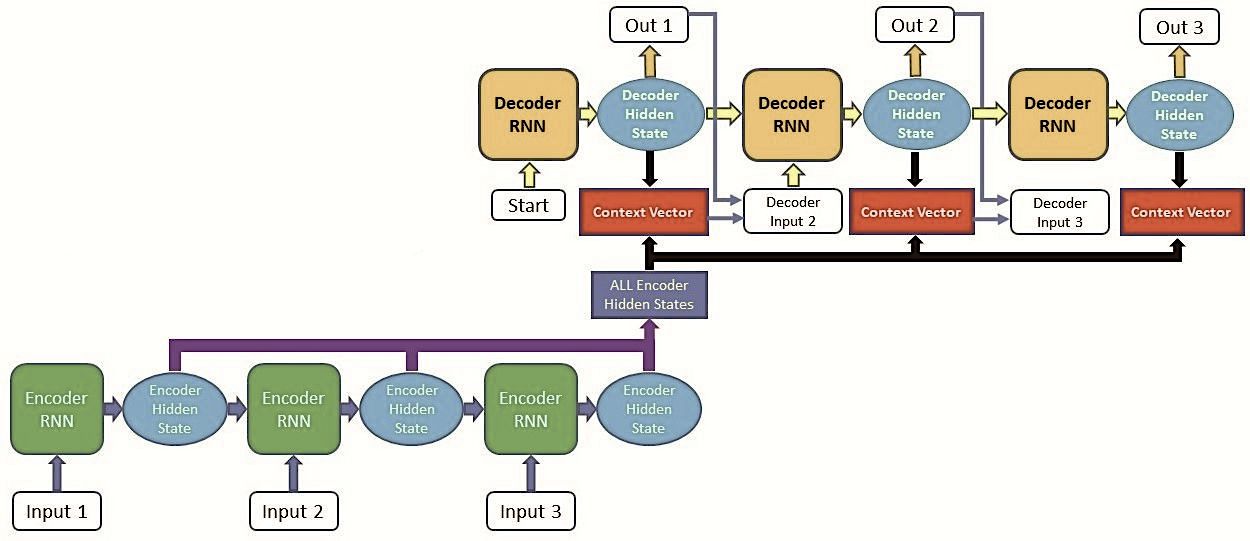}
\caption{Attention Architecture \cite{bahdanau2014neural,loye2019}}\label{fig:att-arch}
\end{center}
\end{figure}

\begin{figure}[t]
\begin{center}
\includegraphics[width=10cm]{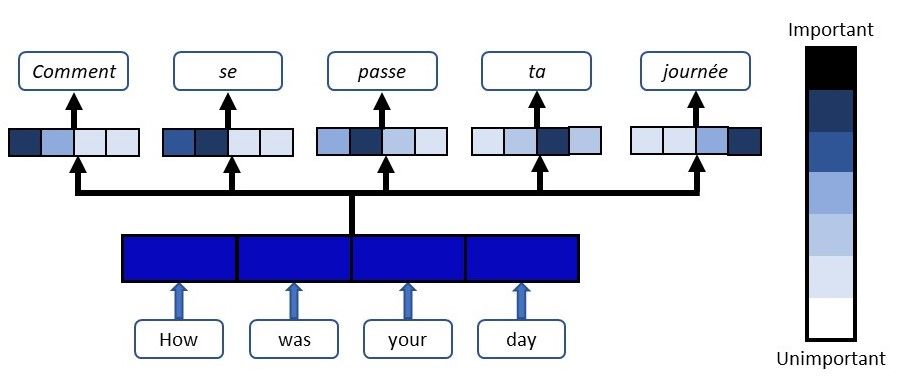}
\caption{Example for Attention Mechanism \cite{loye2019}}\label{fig:attention}
\end{center}
\end{figure}

Another important  architecture is the attention mechanism.
Sequence to sequence learning occurs in many applications of  machine
 learning, for example, in machine translation. The dimensionality of
 the output of basic RNNs is the same 
 as the input.

 The {\em attention\/} architecture allows a flexible
 mapping between input and output
 dimensionality~\cite{sutskever2014sequence}. It
 does so by augmenting the RNNs with an extra
 network that focuses  attention over the sequence of encoder
 RNN states~\cite{bahdanau2014neural}, see Fig.~\ref{fig:att-arch}. The extra network focuses
 perception and memory access. It has been shown to achieve state of
 the art results in machine translation and natural language
 tasks~\cite{bahdanau2014neural}, see Fig.~\ref{fig:attention}. The
 attention mechanism is especially useful for time 
 series forecasting and translation.

\subsubsection*{Transformers}\index{transformers}
Finally, we discuss the transformer architecture. The transformer
architecture is introduced by Vaswani et
al.~\cite{vaswani2017attention}. 
Transformers are based on the concept of self-attention:
they use attention encoder-decoder models, but weigh the
influence of different parts of the data. They are central to the
highly successful BERT~\cite{devlin2018bert} and
GPT-3~\cite{radford2018improving,brown2020language} natural language
models, but have also been successful in imaginative text-to-image
creation. 

Transformer-based foundation models  play an important role in
multi-modal (text/image)
learning~\cite{ramesh2021zero,radford2021learning}, such as the one
that we saw in Fig.~\ref{fig:dalle}. Treating deep reinforcement learning
as a sequence problem, transformers are also being applied to
reinforcement learning,  with some early
success~\cite{chen2021decision,janner2021reinforcement}.  
Transformers are an active area of research. For a detailed
explanation of how they work, see, for
example~\cite{alammer2018,bloem2019,vaswani2017attention}.

\subsection{Overfitting}\index{dropouts}\index{regularization}\index{overfitting}\label{sec:dropouts}\label{sec:overfitting}\label{sec:regularization}
Now that we have discussed  network architectures, it is time to
discuss an important problem of deep neural networks: overfitting, and
how we can reduce it.
Overfitting in neural networks can be reduced  in a number
of ways. Some of the methods are 
aimed at restoring the balance between the number of network
parameters and the number of training examples, others on data
augmentation and capacity reduction.
Another approach is to look at
the  training process. Let us list the most popular approaches~\cite{goodfellow2016deep}.

  \index{data augmentation}
\begin{itemize}
\item {\em Data Augmentation}
Overfitting occurs when there are more parameters in the network than
examples to train on. 
The training dataset
is increased through manipulations such as rotations, reflections, 
noise, rescaling, etc. A disadvantage of this method is that the
computational cost of training increases. 
 


\item
  {\em Capacity Reduction}
Another  solution to overfitting lies in the realization that
overfitting is a result of the network having too large a capacity; the
network has too many parameters. A cheap way of preventing this
situation is to reduce the capacity of the network, by reducing the
width and depth of the network.

\item 
  {\em Dropout}
A popular method to reduce overfitting is to introduce dropout layers
into the networks. Dropout reduces the effective capacity of the
network by stochastically dropping a certain percentage of neurons
from the backpropagation
process~\cite{hinton2012improving,srivastava2014dropout}. Dropout is
an effective and computationally efficient  method to reduce
overfitting~\cite{goodfellow2016deep}.

\item
  {\em L1 and L2 Regularization}
  Regularization involves adding an extra term to the loss
function that forces the network to not be too
complex. The term penalizes the model for  using too high weight
values.  This limits  flexibility, but also encourages
 building solutions based on multiple features. Two popular versions
 of this method are L1 and L2
 regularization~\cite{ng2004feature,goodfellow2016deep}.

\index{early stopping}\index{holdout validation
  set}\label{sec:earlystop}
\item
  {\em Early Stopping}
Early stopping is
based on the observation that overfitting can be regarded as a
consequence of so-called overtraining (training that progresses beyond
the signal, into the noise). By terminating the training
process earlier, for example 
by using a higher stopping threshold for the error function, we can
prevent overfitting from occurring~\cite{caruana2001overfitting,prechelt1998automatic,prechelt1998early}.
A convenient and popular way is to add a third set to the training set/test set
duo which then becomes a training set, a test set, and a holdout
validation set.
After each training epoch, the network is evaluated 
against the holdout validation set, to see  if under- or
overfitting occurs, and if we should stop training.
In this way, overfitting can be
prevented dynamically during
training~\cite{prechelt1998early,bishop2006pattern,goodfellow2016deep}. 


\item
{\em Batch Normalization}
Another method is batch normalization. Batch normalization periodically normalizes the
input to the layers~\cite{ioffe2017batch}. This has many benefits, including a
reduction of overfitting. 
\end{itemize}
Overfitting and regularization are an important topic of research. In
fact, a basic
question is why large neural networks  
perform so well at all. The network capacity is in the  millions to billions
of parameters, much larger than the number of observations, yet networks
perform well. There appears to be a regime, beyond where performance
suffers due to overfitting, where performance increases as we continue
to increase the capacity of our
networks~\cite{belkin2018overfitting,belkin2018reconciling}. 
Belkin et al.\ have performed   studies on interpolation in
SGD. Their studies suggests an implicit regularization regime of many
parameters beyond  overfitting, where SGD generalizes well to test
data,  explaining in part the good results in practice of deep
learning. Nakkiran et al.\ report similar experimental results, termed  
 the {\em double descent\/}
 phenomenon~\cite{nakkiran2019deep,nakkiran2019sgd}. Research into
 the nature of overfitting is  active~\cite{belkin2018overfitting,ma2017power,belkin2018reconciling,belkin2019two,zhang2018dissection}.

\section{Datasets and Software}
We have  discussed in depth background concepts in machine
learning, and important aspects of the theory of neural networks. It
is  time to turn our attention to practical matters. Here we
encounter a rich  field of data, environments, software and blogs on how to use
deep learning in practice.

The field of deep reinforcement learning is an open
 field. Researchers release
their  algorithms and code allowing replication of results. Datasets
and trained networks are shared.  The barrier to
entry is low: high-quality software is available on GitHub for you
to download and start doing research in the field. We point to code
bases and open software suites at GitHub throughout the 
 book, 
Appendix~\ref{ch:env} has pointers to software environments and
open-source code frameworks.

The most popular deep learning packages are PyTorch~\cite{paszke2019pytorch} and TensorFlow~\cite{abadi2016tensorflow}, and its top-level language Keras~\cite{chollet2017deep}. Some machine
learning and mathematical  
packages also offer deep learning tools, such as scikit-learn, MATLAB and
R.
In this section we will start with some easy classification and
behavior examples, using the Keras library. 
We will use Python as our programming language. Python has become the
language of choice for  machine learning packages; not just for PyTorch
and TensorFlow, but also for numpy, scipy, scikit-learn, matplotlib, and
many other mature and high-quality machine learning libraries. 
If Python is not present on your
computer, or if you want to download a newer version, please go to \url{https://www.python.org} to install it.
Note that due to some unfortunate version issues of the Stable
Baselines and TensorFlow, you may have to install different versions
of Python to get these examples to work, and to use virtual
environments to manage your software versions.
\label{sec:pytorch}
\index{TensorFlow}\index{PyTorch}\index{Keras}

Since deep learning is very computationally intensive, these
packages typically support \gls{GPU} parallelism, which can speedup your
training tenfold or more, if you have the right GPU card in your
system. Cloud providers of computing, such as AWS, Google, Azure, and
others, also typically provide modern GPU hardware to support machine
learning, often with student discounts.

\begin{figure}[t]
  \centering{\includegraphics[width=6cm]{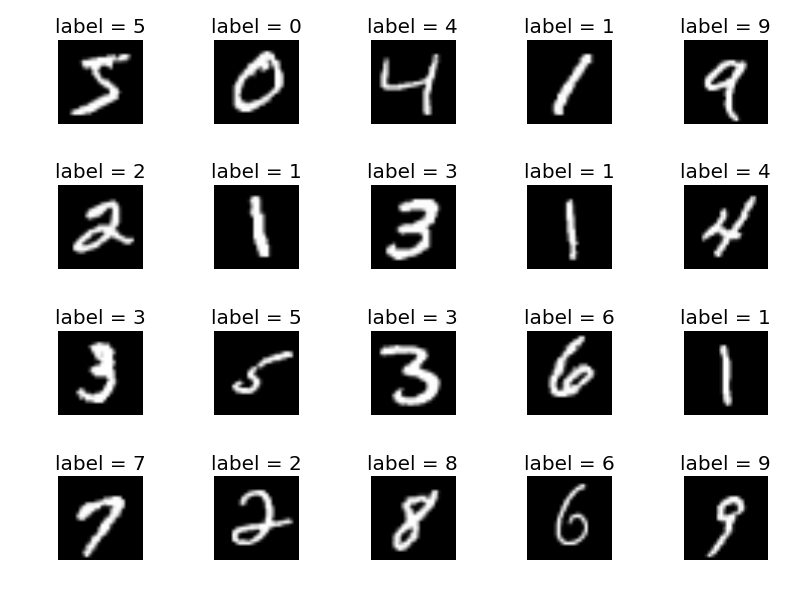}\caption{Some
      MNIST images \cite{lecun1998gradient}}\label{fig:mnist}}
\end{figure}

\subsection{MNIST and ImageNet}\index{MNIST}
One of the most important elements for the success in image
recognition was the availability of good datasets.

In the early days  of deep learning, the field benefited greatly  from efforts in
handwriting recognition. This application was of great value to the
postal service, where accurate recognition of handwritten zip codes or postal codes
allowed great improvements to efficient sorting and delivery of the
mail.  A standard test set for handwriting 
recognition was MNIST (for Modified
National Institute of Standards and Technology)~\cite{lecun1998gradient}.
Standard  MNIST images are low-resolution $32\times 32$ pixel images
of single handwritten digits (Fig.~\ref{fig:mnist}). Of course,
researchers wanted to process more complex scenes than single digits,
and higher-resolution images.
To achieve higher accuracy, and to process more complex scenes,
networks (and datasets) needed to grow in size and complexity.


\subsubsection*{ImageNet}\index{imagenet}
A major dataset in deep learning is
ImageNet~\cite{fei2009imagenet,deng2009imagenet}.
\index{ImageNet}   
It is a collection of more than 14 million URLs of images that have
been hand annotated with the objects that are in the picture. It contains
more than 20,000 categories. A typical category contains several
hundred training images.

The importance of ImageNet for the progress in deep learning is
large. The availability of an accepted standard set of labeled images allowed learning algorithms
to be tested and improved,
and new algorithms to be created. ImageNet was conceived by Fei-Fei Li et al.
in 2006, and in later years she developed it further with her group. Since 2010  an annual software
contest has been organized, the ImageNet Large Scale Visual Recognition Challenge
(ILSVRC)~\cite{deng2009imagenet}. Since 2012 ILSVRC has been won by deep
networks, starting the deep learning boom. The network architecture that won this
challenge in that year has become known as \emph{AlexNet}, after one of its authors~\cite{krizhevsky2012imagenet}.\label{sec:alexnet}

\index{AlexNet}
The 2012 ImageNet database as used by AlexNet has  14 million labeled
images. The network featured a highly optimized 2D two-GPU implementation of 5
convolutional layers and 3 fully connected layers. The filters in the
convolutional layers are $11\times 11$ in size.
The neurons use a ReLU activation function. In AlexNet
images were scaled to $256\times 256$ RGB pixels.
The size of the
network was large, with 60 million parameters. This causes
considerable overfitting. AlexNet used data augmentation and dropouts
to reduce the impact of overfitting.

Krizhevsky et al.\ won the 2012 ImageNet competition with an error rate of
15\%, significantly better than the number two, who achieved 26\%.  Although there were earlier reports of
CNNs that were successful in applications such as bioinformatics and
Chinese handwriting recognition, it was this  win of the 2012 ImageNet
competition for which  AlexNet has become well known.

\subsection{GPU Implementations}\index{GPU}\label{sec:gpu}

The deep learning breakthrough around 2012 was caused by the co-occurrence of
three major developments: (1)
algorithmic advances that solved key problems in deep learning, (2) the
availability of large datasets of labeled training data, and (3)  the
availability of computational power in the form of graphical processing units,
GPUs.

The most expensive operations in  image processing and neural
network training are  operations on matrices. Matrix
operations are some of the most well-studied problems in computer science. Their
algorithmic structure
is well understood, and for  basic linear algebra
operations  high-performance parallel implementations
for \gls{CPU} exist, such as the BLAS~\cite{dongarra1988extended,choi1996pb}.

\begin{figure}[t]
  \centering{\begin{tabular}{cc}\includegraphics[width=7cm]{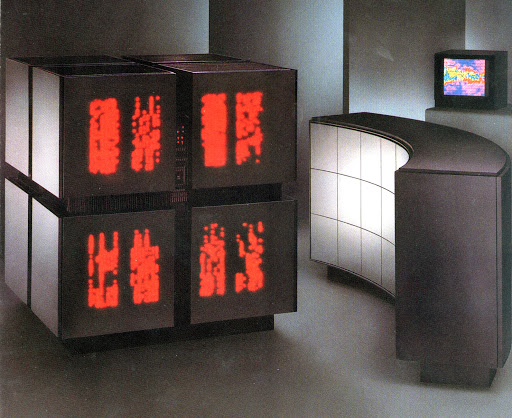}&\includegraphics[width=4cm]{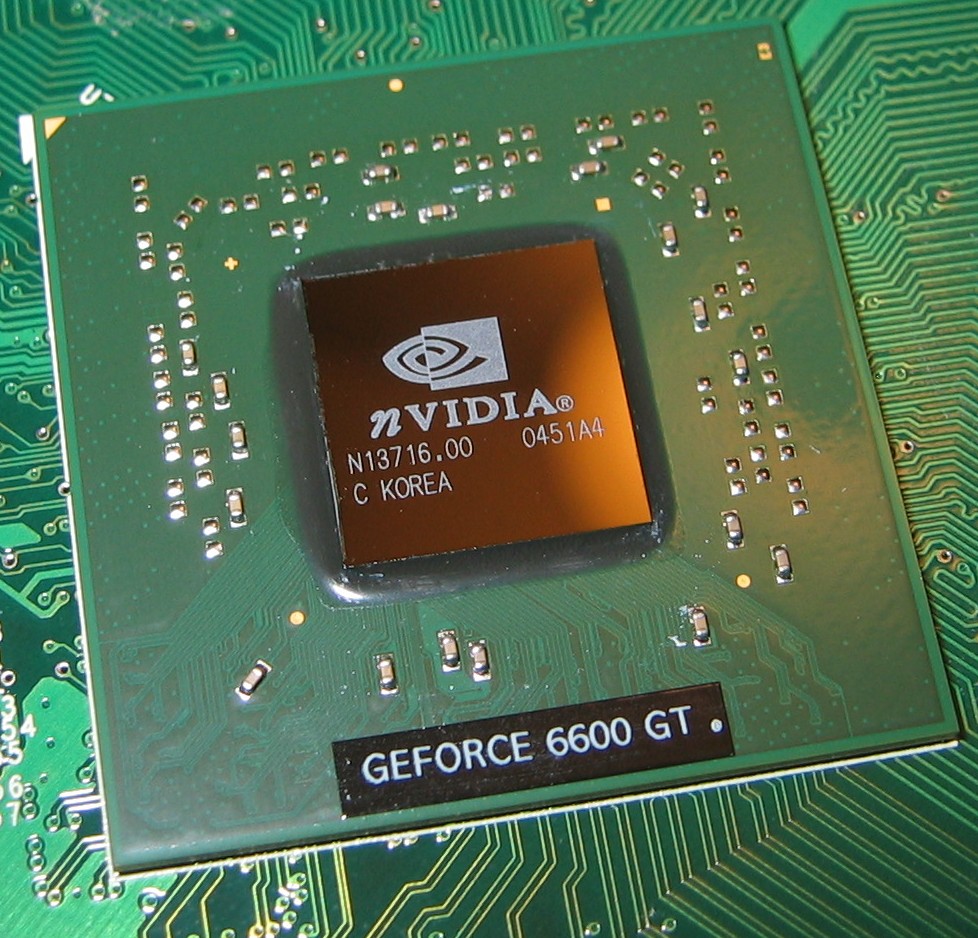}\end{tabular}\caption{SIMD:
      Connection
      Machine 1 and GPU}\label{fig:cm1}}
\end{figure}

GPUs
were originally designed for smooth graphics performance in video games. Graphical processing requires fast linear algebra
computations such as matrix multiply. These are precisely the kind of
operations that are at the core of deep learning training
algorithms. Modern GPUs consist of thousands of small arithmetic units that
are capable of performing linear algebra matrix operations very fast in
parallel. This kind of data parallel processing is based on SIMD
computing, for
single-instruction-multiple-data~\cite{flynn1972some,hennessy2011computer}. SIMD data
parallism goes back to designs
 from  1960s and 1970s vector supercomputers such as the
Connection Machine series  from Thinking
Machines~\cite{hillis1982new,hillis1993cm,leiserson1992network}. Figure~\ref{fig:cm1}
shows a picture of the historic CM-1, and of a modern GPU. 

Modern GPUs consist of 
thousands of processing units optimized to process linear algebra matrix operations
 in
 parallel~\cite{sanders2010cuda,liu2012sparse,song2014scaling},
 offering matrix performance that is orders of magnitude faster than
 CPUs~\cite{oh2004gpu,cirecsan2010deep,sze2017efficient}.

\subsection{\em Hands On: Classification Example} 
It is high time to try out some of the material
in practice. Let us see if we can do some image recognition ourselves.

\subsubsection*{Installing TensorFlow and
  Keras}\index{Keras}\label{sec:keras}

We will first
install TensorFlow.
It is possible to run TensorFlow in
the cloud, in Colab, or in a Docker container. Links to ready 
to run Colab environments are on the TensorFlow website. We will,
however, asssume a traditional local installation on your own
computer. All major operating systems are
supported:  Linux/Ubuntu, macOS, Windows.

The programming model of TensorFlow is
complex, and not very user friendly. Fortunately,  an
easy to use language has been built on top of TensorFlow: Keras. The
Keras language is easy to use, well-documented, and many examples
exist to get you started. When you install TensorFlow, Keras is
installed automatically as well.

To install TensorFlow  and Keras go to the TensorFlow
page on \url{https://www.tensorflow.org}.  It is recommended to make a virtual
environment to isolate the package installation from the rest of your system. This
is achieved by typing
\begin{tcolorbox}
  \verb|python3 -m venv --system-site-packages ./venv|
\end{tcolorbox}
  (or equivalent) to create the virtual environment. Using the
virtual environment requires activation:\footnote{More installation guidance
  can be found on the TensorFlow page at
  \url{https://www.tensorflow.org/install/pip}.}
\begin{tcolorbox}
\verb|source ./venv/bin/activate|
\end{tcolorbox}
  
You will most likely run into version issues when installing packages. Note that for deep 
reinforcement learning we will be making extensive use of
reinforcement learning agent algorithms from the so-called Stable
Baselines.\footnote{\url{https://github.com/hill-a/stable-baselines}} The stable baselines
work with version 1 of TensorFlow and with PyTorch, but, as of this writing, not with
version 2 of TensorFlow. 
TensorFlow version 1.14 has been tested to work.\footnote{You may be surprised  that so many version
  numbers were mentioned. Unfortunately not all versions are
  compatible; some care is necessary to get the software to work: Python 
3.7.9, TensorFlow 1.14.0, Stable Baselines 2, pyglet 1.5.11 worked at
the time of writing. This slightly embarrassing situation is because the field
of deep reinforcement learning is driven by a community of
researchers, who collaborate to make code bases work. When new
insights trigger a rewrite that loses backward compatibility, as
happened with TensorFlow 2.0, frantic rewriting of dependent software
occurs. The field is still a new field, and some instability of
software packages will remain with us for the foreseeable future.}

Installing is easy with Python's pip package
manager: just type
\begin{tcolorbox}
  \verb|pip install tensorflow==1.14|
\end{tcolorbox}
or, for GPU support,
\begin{tcolorbox}
  \verb|pip install tensorflow-gpu==1.14|
\end{tcolorbox}  
This should now download and install TensorFlow and Keras. We will
check if everything is 
working by executing the MNIST training example with the  default training
dataset.

\subsubsection*{Keras MNIST Example}
Keras  is built
on top of TensorFlow, and is installed with TensorFlow. Basic Keras mirrors the familiar Scikit-learn interface~\cite{pedregosa2011scikit}.

Each Keras program specifies a model
that is to be learned. The model is the neural network, that consists
of weights and layers (of neurons). You will specify the architecure
of the model in Keras, and then fit the model on the training
data. When the model is trained, you can evaluate the loss function on
test data,
or perform predictions of the outcome based on some test input
example.

\lstset{label={lst:keras}}
\lstset{caption={Sequential Model in Keras}}
\begin{lstlisting}[language=Python,float]
from tensorflow.keras.models import Sequential
from tensorflow.keras.layers import Dense

model = Sequential()

model.add(Dense(units=64, activation='relu'))
model.add(Dense(units=10, activation='softmax'))

model.compile(loss='categorical_crossentropy',
              optimizer='sgd',
              metrics=['accuracy'])

# x_train and y_train are Numpy arrays --just like in the Scikit-Learn API.
model.fit(x_train, y_train, epochs=5, batch_size=32)

# evaluation on test data is simple
loss_and_metrics = model.evaluate(x_test, y_test, batch_size=128)

# as is predicting output
classes = model.predict(x_test, batch_size=128)
\end{lstlisting}

\lstset{label={lst:mnist}}
\lstset{caption={Functional MNIST Model in Keras}}
\begin{lstlisting}[language=Python,float]
# Get the data as Numpy arrays
(x_train, y_train), (x_test, y_test) = keras.datasets.mnist.load_data()

# Build a simple model
inputs = keras.Input(shape=(28, 28))
x = layers.experimental.preprocessing.Rescaling(1.0 / 255)(inputs)
x = layers.Flatten()(x)
x = layers.Dense(128, activation="relu")(x)
x = layers.Dense(128, activation="relu")(x)
outputs = layers.Dense(10, activation="softmax")(x)
model = keras.Model(inputs, outputs)
model.summary()

# Compile the model
model.compile(optimizer="adam", loss="sparse_categorical_crossentropy")

# Train the model for 1 epoch from Numpy data
batch_size = 64
print("Fit on NumPy data")
history = model.fit(x_train, y_train, batch_size=batch_size, epochs=1)

# Train the model for 1 epoch using a dataset
dataset = tf.data.Dataset.from_tensor_slices((x_train, y_train)).batch(batch_size)
print("Fit on Dataset")
history = model.fit(dataset, epochs=1)
\end{lstlisting}

Keras has two main programming paradigms: sequential and functional.
Listing~\ref{lst:keras} shows the most basic Sequential Keras model,
from the Keras documentation, with a two-layer model, a ReLU layer,
and a softmax 
layer, using simple SGD for backpropagation. The
Sequential model in Keras has an object-oriented syntax.

The Keras documentation is at
\url{https://keras.io/getting_started/intro_to_keras_for_researchers/}. It
is quite accessible, and you are  encouraged to learn Keras by
working through the online tutorials.

A slightly more useful example is fitting a model on MNIST, see
Listing~\ref{lst:mnist}. This example uses a more flexible Keras
syntax, the functional API, in which transformations are chained on
top of the previous layers. This example of Keras code loads MNIST
images in a training set and a test set, and creates a model of dense ReLU
layers using the Functional API. It then creates a Keras model of
these layers, and prints a summary of the model. Then the model is
trained from numpy data, with the fit function, for a single epoch,
and also from a dataset, and the loss
history is printed.



The code of the example shows how close Keras is to the way
in which we think and reason about neural networks. The examples only
show the briefest of glimpses to what is possible in Keras. Keras has
options for performance monitoring, for checkpointing of long training
runs, and for interfacing with TensorBoard, to visualize the training
process. TensorBoard is an indispensible tool, allowing you to debug
your intuition of what should be going on in your network, and what is
going on.

Deep reinforcement learning is still very much a field
with more degrees of freedom in experimentation than established
practices, and being able to plot the progress of training processes
is essential for a better understanding of how the model behaves.
The more you explore Keras, the better you will be able to progress in
deep reinforcement learning~\cite{geron2017hands}.

\section*{Exercises}
\addcontentsline{toc}{section}{\protect\numberline{}Exercises}
Below are some questions to check your understanding of deep
learning. Each question is a closed question where a simple, single 
sentence answer is expected.
\subsubsection*{Questions}\label{sec:dlq}
\begin{enumerate}

\item Datasets are often split into two sub-datasets in machine
  learning. Which are those two, and why are they split?
\item What is generalization?
\item Sometimes a third sub-dataset is used. What is it called and
  what is it used for?
\item If we consider observations and model parameters, when is a
  problem high-dimensional and when is it low-dimensional?
\item How do we measure the capacity of a machine learning model?
\item What is a danger with low capacity models?
\item What is a danger with high capacity models?
\item What is the effect of overfitting on generalization?  
  \item What is the difference between supervised learning and
    reinforcement learning?
  \item What is the difference between a shallow network and a deep
    network?
\item Supervised learning has a model and a dataset, reinforcement
  learning has which two central concepts?
\item What phases does a learning epoch have? What happens in each phase?
  \item Name three factors that were essential for the deep learning
    breakthrough, and why.
  \item What is end-to-end learning? Do you know an alternative?
    What are the advantages of each?
  \item What is underfitting, and what causes it? What is overfitting, and what causes it? How can you see
    if you have overfitting?
  \item Name three ways to prevent overfitting.
\item Which three types of layers does a neural network have?
\item How many hidden layers does a shallow neural network have?
\item Describe how adjusting weights works in a neural
  network. Hint: think of a examples, labels, forward phase, a backward phase, error
  functions, gradients.
    
  \item What is the difference between a fully connected network
    and a convolutional neural network?
  \item What is max pooling?
  \item Why are shared weights advantageous?
\item What is feature learning?
\item What is representation learning?
\item What is deep learning?
\item What is an advantage of convolutional neural networks of
  fully connected neural networks?
\item Name two well-known image recognition data sets.
\end{enumerate}

\subsubsection*{Exercises}
Let us now start with some exercises. 
If you have not done so already, install  PyTorch\footnote{\url{https://pytorch.org}} or TensorFlow and Keras
(see Sect.~\ref{sec:gym} or go to the TensorFlow
page).\footnote{\url{https://www.tensorflow.org}}
Be sure to check  the right versions of Python,  TensorFlow, and the
Stable Baselines to make sure they work well together.
The exercises
below are meant to be done in Keras.

\begin{enumerate}
\item
  {\em Generalization} Install Keras. Go to the Keras MNIST example. Perform a
  classification task. Note how many epochs the training takes, and in
  testing, how well it generalizes. Perform the classification on a
  smaller training set, how does learning rate change, how does
  generalization change. Vary other elements: try a different
  optimizer than adam, try a different learning rate, try a different
  (deeper) architecture, try wider hidden layers. Does it learn
  faster? Does it generalize better?
\item {\em Overfitting} Use Keras again, but this time on ImageNet. Now try
  different overfitting solutions. Does the training speed change?
  Does generalization change? Now try the hold-out  validation set. Do
  training and generalization change?
\item {\em Confidence} How many runs did you do in the previous
  exercises, just a single run to see how long training took and how
  well generalization worked? Try to run it again. Do you get the
  same results? How large is the difference? Can you change the
  random seeds of Keras or TensorFlow? Can you calculate the
  confidence interval, how much does the confidence improve when you
  do 10 randomized runs? How about 100 runs? Make graphs with error
  bars.
\item {\em GPU} It might be that you have access to a GPU machine
  that is capable of running PyTorch or TensorFlow in parallel to speed up the
  training. Install the  GPU version  and check that it
  recognizes the GPU and is indeed using it.
\item {\em Parallelism} It might be that you have access to a multicore CPU
  machine. When you are running multiple runs in order to improve
  confiedence, then an easy way to speed up your experiment is to
  spawn multiple jobs at the shell, assigning the output to different log files,
  and write a script to combine results and draw graphs. Write the
  scripts necessary to achieve this, test them, and do a
  large-confidence experiment.
\end{enumerate}


\chapter{Deep Reinforcement Learning Suites}\label{ch:env}
Deep reinforcement learning is a highly active field of
research. One reason for the progress is the availability of high
quality algorithms and code:
high quality environments, algorithms and deep learning suites are all
being made available by researchers along with their research
papers. This appendix provides pointers to these codes.

  \newpage
\section{Environments}
Progress has benefited greatly from the availability of
high quality environments, on which the algorithms can be tested. We
 provide pointers to some of the environments (Table~\ref{tab:env}).

\begin{table}
\begin{tabular}{l|l|p{6.6cm}|c}
  {\bf Environment} & {\bf Type} & {\bf URL} & {\bf Ref}\\ \hline
  Gym & ALE, MuJoCo & \url{https://gym.openai.com} &\cite{brockman2016openai} \\
ALE & Atari &
              \url{https://github.com/mgbellemare/Arcade-Learning-Environment}
                                             & \cite{bellemare2013arcade}\\
MuJoCo& Simulated Robot&\url{http://www.mujoco.org}& \cite{todorov2012mujoco}\\                                             
  DeepMind Lab & 3D navigation & \url{https://github.com/deepmind/lab}&\cite{beattie2016deepmind}\\
Control Suite& Physics tasks & \url{https://github.com/deepmind/dm_control}&\cite{tassa2018deepmind}\\
Behavior Suite& Core RL &\url{https://github.com/deepmind/bsuite}&\cite{osband2019behaviour}\\
StarCraft& Python interface &\url{https://github.com/deepmind/pysc2}&\cite{vinyals2017starcraft}\\
Omniglot&Images&\url{https://github.com/brendenlake/omniglot}&\cite{lake2011one}\\
Mini ImageNet&Images&\url{https://github.com/yaoyao-liu/mini-imagenet-tools}&\cite{vinyals2016matching}\\
  ProcGen&Procedurally Gen.&\url{https://openai.com/blog/procgen-benchmark/}&\cite{cobbe2020leveraging}\\ 
OpenSpiel&Board Games&\url{https://github.com/deepmind/open_spiel}&\cite{lanctot2019openspiel}\\
RLlib&Scalable RL &\url{https://github.com/ray-project/ray}&\cite{liang2017rllib}\\
Meta Dataset&dataset of datas.&\url{https://github.com/google-research/meta-dataset}&\cite{triantafillou2019meta}\\
Meta World&meta RL&\url{https://meta-world.github.io}&\cite{yu2020meta}\\
Alchemy&meta RL&\url{https://github.com/deepmind/dm_alchemy}&\cite{wang2021alchemy}\\
 Garage&reproducible RL&\url{https://github.com/rlworkgroup/garage}&\cite{garage}\\ 
Football&multi agent&\url{https://github.com/google-research/football}&\cite{kurach2020google}\\
Emergence&Hide and Seek&\url{https://github.com/openai/multi-agent-emergence-environments}&\cite{baker2019emergent}\\
Unplugged&Offline RL&\url{https://github.com/deepmind/deepmind-research/tree/master/rl_unplugged}&\cite{gulcehre2020rl}\\
  Unity&3D&\url{https://github.com/Unity-Technologies/ml-agents}&\cite{juliani2018unity}\\
  PolyGames&board games&\url{https://github.com/teytaud/Polygames}&\cite{cazenave2020polygames}\\
Dopamine&RL framework&\url{https://github.com/google/dopamine}&\cite{castro2018dopamine}\\  
\end{tabular}
\caption{Reinforcement Learning Environments}\label{tab:env}
\end{table}
  
   \newpage   

\section{Agent Algorithms}
For value-based and policy-based methods most mainstream  algorithms
have been collected and are freely available. For two-agent,
multi-agent, hierarchical, and meta learning, the agent algorithms are
also on GitHub, but not always in the same place as the basic
algorithms. Table~\ref{tab:agentalg} provides pointers.

\begin{table}
  \begin{centering}
    \begin{tabular}{l|p{5cm}|l|c}
      {\bf Repo}&{\bf URL}&{\bf Algorithms}&{\bf Ref}\\ \hline
      Spinning Up&\url{https://spinningup.openai.com}&Tutorial on DDPG,
                                                       PPO,
                                                       etc&OpenAI\\
      Baselines&\url{https://github.com/openai/baselines}&DQN, PPO,
                                                           etc&\cite{baselines}\\
      Stable
      Basel. &\url{https://stable-baselines.readthedocs.io/en/master/}&Refactored
                                                                          baselines&\cite{hill2018stable}\\
      PlaNet&\url{https://planetrl.github.io}&Latent Model&\cite{hafner2018learning}\\
      Dreamer&\url{https://github.com/danijar/dreamerv2}&Latent
                                                          Model&\cite{hafner2020mastering}\\
      VPN&\url{https://github.com/junhyukoh/value-prediction-network}&Value
                                                                       Prediction Network&\cite{oh2017value}\\
MuZero&\url{https://github.com/kaesve/muzero}&Reimplementation of MuZero&\cite{vries2021visualizing}\\
      MCTS&pip install mcts&MCTS&\cite{browne2012survey} \\
      AlphaZ.
      Gen&\url{https://github.com/suragnair/alpha-zero-general}&AZ
                                                                     in Python&\cite{nair2017learning}\\
ELF&\url{https://github.com/facebookresearch/ELF}&Framework for Game research&\cite{tian2017elf}\\
PolyGames&\url{https://github.com/teytaud/Polygames}&Zero
                                                      Learning&\cite{cazenave2020polygames}\\
      CFR&\url{https://github.com/bakanaouji/cpp-cfr}&CFR&\cite{zinkevich2008regret}\\
      DeepCFR&\url{https://github.com/EricSteinberger/Deep-CFR}&Deep
                                                                 CFR&\cite{steinberger2019single}\\
      MADDPG&\url{https://github.com/openai/maddpg}&Multi-agent
                                                     DDPG&\cite{lowe2017multi}\\
      PBT&\url{https://github.com/voiler/PopulationBasedTraining}&Population
                                                                   Based Training&\cite{jaderberg2017population}\\
Go-Explore&\url{https://github.com/uber-research/go-explore}&Go Explore&\cite{ecoffet2021first}\\
      MAML&\url{https://github.com/cbfinn/maml}&MAML&\cite{finn2017model}\\
    \end{tabular}
    \caption{Agent Algorithms}\label{tab:agentalg}
  \end{centering}
\end{table}

\newpage
\section{Deep Learning Suites}
The two most well-known deep learning suites are TensorFlow and PyTorch.
Base-TensorFlow has a complicated programming model. Keras has been
developed as an easy to use layer on top of TensorFlow. When you use
TensorFlow, start with Keras. Or use PyTorch.

TensorFlow and Keras are at \url{https://www.tensorflow.org}.

PyTorch is at \url{https://pytorch.org}.



\backmatter

\cleardoublepage

\bibliographystyle{plain}
\bibliography{drl8}

\begin{thebibliography}{100}

\bibitem{abadi2016tensorflow}
Mart{\'{\i}}n Abadi, Paul Barham, Jianmin Chen, Zhifeng Chen, Andy Davis,
  Jeffrey Dean, Matthieu Devin, Sanjay Ghemawat, Geoffrey Irving, Michael
  Isard, Manjunath Kudlur, Josh Levenberg, Rajat Monga, Sherry Moore,
  Derek~Gordon Murray, Benoit Steiner, Paul~A. Tucker, Vijay Vasudevan, Pete
  Warden, Martin Wicke, Yuan Yu, and Xiaoqiang Zheng.
\newblock Tensorflow: A system for large-scale machine learning.
\newblock In {\em 12th {USENIX} Symposium on Operating Systems Design and
  Implementation ({OSDI} 16)}, pages 265--283, 2016.

\bibitem{abbeel2010autonomous}
Pieter Abbeel, Adam Coates, and Andrew~Y Ng.
\newblock Autonomous helicopter aerobatics through apprenticeship learning.
\newblock {\em The International Journal of Robotics Research},
  29(13):1608--1639, 2010.

\bibitem{abbeel2007application}
Pieter Abbeel, Adam Coates, Morgan Quigley, and Andrew~Y Ng.
\newblock An application of reinforcement learning to aerobatic helicopter
  flight.
\newblock In {\em Advances in Neural Information Processing Systems}, pages
  1--8, 2007.

\bibitem{abdolmaleki2018maximum}
Abbas Abdolmaleki, Jost~Tobias Springenberg, Yuval Tassa, Remi Munos, Nicolas
  Heess, and Martin Riedmiller.
\newblock Maximum a posteriori policy optimisation.
\newblock {\em arXiv preprint arXiv:1806.06920}, 2018.

\bibitem{brillio2019}
Abhishek.
\newblock Multi-arm bandits: a potential alternative to a/b tests
  \url{https://medium.com/brillio-data-science/multi-arm-bandits-a-potential-alternative-to-a-b-tests-a647d9bf2a7e},
  2019.

\bibitem{abramson1990expected}
Bruce Abramson.
\newblock Expected-outcome: A general model of static evaluation.
\newblock {\em IEEE Transactions on Pattern Analysis and Machine Intelligence},
  12(2):182--193, 1990.

\bibitem{agarwal2021deep}
Rishabh Agarwal, Max Schwarzer, Pablo~Samuel Castro, Aaron Courville, and
  Marc~G Bellemare.
\newblock Deep reinforcement learning at the edge of the statistical precipice.
\newblock {\em arXiv preprint arXiv:2108.13264}, 2021.

\bibitem{agrawal2014analyzing}
Pulkit Agrawal, Ross Girshick, and Jitendra Malik.
\newblock Analyzing the performance of multilayer neural networks for object
  recognition.
\newblock In {\em European Conference on Computer Vision}, pages 329--344.
  Springer, 2014.

\bibitem{ahilan2019feudal}
Sanjeevan Ahilan and Peter Dayan.
\newblock Feudal multi-agent hierarchies for cooperative reinforcement
  learning.
\newblock {\em arXiv preprint arXiv:1901.08492}, 2019.

\bibitem{akata2013label}
Zeynep Akata, Florent Perronnin, Zaid Harchaoui, and Cordelia Schmid.
\newblock Label-embedding for attribute-based classification.
\newblock In {\em Proceedings of the IEEE Conference on Computer Vision and
  Pattern Recognition}, pages 819--826, 2013.

\bibitem{akiba2019optuna}
Takuya Akiba, Shotaro Sano, Toshihiko Yanase, Takeru Ohta, and Masanori Koyama.
\newblock Optuna: A next-generation hyperparameter optimization framework.
\newblock In {\em Proceedings of the 25th ACM SIGKDD International Conference
  on Knowledge Discovery \& Data Mining}, pages 2623--2631, 2019.

\bibitem{alammer2018}
Jay Alammer.
\newblock The illustrated transformer.
  \url{https://jalammar.github.io/illustrated-transformer/}.

\bibitem{albrecht2017multiagent}
Stefano Albrecht and Peter Stone.
\newblock Multiagent learning: foundations and recent trends.
\newblock In {\em Tutorial at IJCAI-17 conference}, 2017.

\bibitem{albrecht2018autonomous}
Stefano Albrecht and Peter Stone.
\newblock Autonomous agents modelling other agents: A comprehensive survey and
  open problems.
\newblock {\em Artificial Intelligence}, 258:66--95, 2018.

\bibitem{alpaydin2009introduction}
Ethem Alpaydin.
\newblock {\em Introduction to Machine Learning}.
\newblock MIT press, 2009.

\bibitem{alver2018ml}
Safa Alver.
\newblock The option-critic architecture.
\newblock \url{https://alversafa.github.io/blog/2018/11/28/optncrtc.html},
  2018.

\bibitem{amodei2016concrete}
Dario Amodei, Chris Olah, Jacob Steinhardt, Paul Christiano, John Schulman, and
  Dan Man{\'e}.
\newblock Concrete problems in {AI} safety.
\newblock {\em arXiv preprint arXiv:1606.06565}, 2016.

\bibitem{anand2021procedural}
Ankesh Anand, Jacob Walker, Yazhe Li, Eszter V{\'e}rtes, Julian Schrittwieser,
  Sherjil Ozair, Th{\'e}ophane Weber, and Jessica~B Hamrick.
\newblock Procedural generalization by planning with self-supervised world
  models.
\newblock {\em arXiv preprint arXiv:2111.01587}, 2021.

\bibitem{andrychowicz2017hindsight}
Marcin Andrychowicz, Filip Wolski, Alex Ray, Jonas Schneider, Rachel Fong,
  Peter Welinder, Bob McGrew, Josh Tobin, Pieter Abbeel, and Wojciech Zaremba.
\newblock Hindsight experience replay.
\newblock In {\em Advances in Neural Information Processing Systems}, pages
  5048--5058, 2017.

\bibitem{reddit2017}
Anonymous.
\newblock Go {AI} strength vs. time.
\newblock {\em Reddit post}, 2017.

\bibitem{anthony2020learning}
Thomas Anthony, Tom Eccles, Andrea Tacchetti, J{\'{a}}nos Kram{\'{a}}r, Ian~M.
  Gemp, Thomas~C. Hudson, Nicolas Porcel, Marc Lanctot, Julien P{\'{e}}rolat,
  Richard Everett, Satinder Singh, Thore Graepel, and Yoram Bachrach.
\newblock Learning to play no-press diplomacy with best response policy
  iteration.
\newblock In {\em Advances in Neural Information Processing Systems}, 2020.

\bibitem{anthony2017thinking}
Thomas Anthony, Zheng Tian, and David Barber.
\newblock Thinking fast and slow with deep learning and tree search.
\newblock In {\em Advances in Neural Information Processing Systems}, pages
  5360--5370, 2017.

\bibitem{antoniou2018train}
Antreas Antoniou, Harrison Edwards, and Amos Storkey.
\newblock How to train your {MAML}.
\newblock {\em arXiv preprint arXiv:1810.09502}, 2018.

\bibitem{antoniou2004semantic}
Grigoris Antoniou and Frank Van~Harmelen.
\newblock {\em A Semantic Web Primer}.
\newblock MIT press Cambridge, MA, 2008.

\bibitem{arenz2012monte}
Oleg Arenz.
\newblock {Monte Carlo Chess}.
\newblock Master's thesis, Universit\"at Darmstadt, 2012.

\bibitem{argyriou2007multi}
Andreas Argyriou, Theodoros Evgeniou, and Massimiliano Pontil.
\newblock Multi-task feature learning.
\newblock In {\em Advances in Neural Information Processing Systems}, pages
  41--48, 2007.

\bibitem{arulkumaran2017deep}
Kai Arulkumaran, Marc~Peter Deisenroth, Miles Brundage, and Anil~Anthony
  Bharath.
\newblock Deep reinforcement learning: A brief survey.
\newblock {\em IEEE Signal Processing Magazine}, 34(6):26--38, 2017.

\bibitem{asmuth2009bayesian}
John Asmuth, Lihong Li, Michael~L Littman, Ali Nouri, and David Wingate.
\newblock A {Bayesian} sampling approach to exploration in reinforcement
  learning.
\newblock In {\em Proceedings of the Twenty-Fifth Conference on Uncertainty in
  Artificial Intelligence}, pages 19--26. AUAI Press, 2009.

\bibitem{aubret2019survey}
Arthur Aubret, Laetitia Matignon, and Salima Hassas.
\newblock A survey on intrinsic motivation in reinforcement learning.
\newblock {\em arXiv preprint arXiv:1908.06976}, 2019.

\bibitem{auer2002using}
Peter Auer.
\newblock Using confidence bounds for exploitation-exploration trade-offs.
\newblock {\em Journal of Machine Learning Research}, 3(Nov):397--422, 2002.

\bibitem{auer2002finite}
Peter Auer, Nicolo Cesa-Bianchi, and Paul Fischer.
\newblock Finite-time analysis of the multiarmed bandit problem.
\newblock {\em {Machine Learning}}, 47(2-3):235--256, 2002.

\bibitem{auer2010ucb}
Peter Auer and Ronald Ortner.
\newblock {UCB} revisited: Improved regret bounds for the stochastic
  multi-armed bandit problem.
\newblock {\em Periodica Mathematica Hungarica}, 61(1-2):55--65, 2010.

\bibitem{axelrod1986evolutionary}
Robert Axelrod.
\newblock An evolutionary approach to norms.
\newblock {\em The American Political Science Review}, pages 1095--1111, 1986.

\bibitem{axelrod1997complexity}
Robert Axelrod.
\newblock {\em The complexity of cooperation: Agent-based models of competition
  and collaboration}, volume~3.
\newblock Princeton university press, 1997.

\bibitem{axelrod1997dissemination}
Robert Axelrod.
\newblock The dissemination of culture: A model with local convergence and
  global polarization.
\newblock {\em Journal of Conflict Resolution}, 41(2):203--226, 1997.

\bibitem{axelrod1988further}
Robert Axelrod and Douglas Dion.
\newblock The further evolution of cooperation.
\newblock {\em Science}, 242(4884):1385--1390, 1988.

\bibitem{axelrod1981evolution}
Robert Axelrod and William~D Hamilton.
\newblock The evolution of cooperation.
\newblock {\em Science}, 211(4489):1390--1396, 1981.

\bibitem{azizzadenesheli2018surprising}
Kamyar Azizzadenesheli, Brandon Yang, Weitang Liu, Emma Brunskill, Zachary~C
  Lipton, and Animashree Anandkumar.
\newblock Surprising negative results for generative adversarial tree search.
\newblock {\em arXiv preprint arXiv:1806.05780}, 2018.

\bibitem{babaeizadeh2020models}
Mohammad Babaeizadeh, Mohammad~Taghi Saffar, Danijar Hafner, Harini Kannan,
  Chelsea Finn, Sergey Levine, and Dumitru Erhan.
\newblock Models, pixels, and rewards: Evaluating design trade-offs in visual
  model-based reinforcement learning.
\newblock {\em arXiv preprint arXiv:2012.04603}, 2020.

\bibitem{back1996evolutionary}
Thomas B{\"a}ck.
\newblock {\em Evolutionary Algorithms in Theory and Practice: Evolutionary
  Strategies, Evolutionary Programming, Genetic Algorithms}.
\newblock Oxford University Press, 1996.

\bibitem{back1997handbook}
Thomas B{\"a}ck, David~B Fogel, and Zbigniew Michalewicz.
\newblock Handbook of evolutionary computation.
\newblock {\em Release}, 97(1):B1, 1997.

\bibitem{back1991survey}
Thomas B{\"a}ck, Frank Hoffmeister, and Hans-Paul Schwefel.
\newblock A survey of evolution strategies.
\newblock In {\em Proceedings of the fourth International Conference on Genetic
  Algorithms}, 1991.

\bibitem{back1993overview}
Thomas B{\"a}ck and Hans-Paul Schwefel.
\newblock An overview of evolutionary algorithms for parameter optimization.
\newblock {\em Evolutionary Computation}, 1(1):1--23, 1993.

\bibitem{backstrom1995planning}
Christer Backstrom and Peter Jonsson.
\newblock Planning with abstraction hierarchies can be exponentially less
  efficient.
\newblock In {\em Proceedings of the 14th International Joint Conference on
  Artificial Intelligence}, volume~2, pages 1599--1604, 1995.

\bibitem{bacon2017option}
Pierre-Luc Bacon, Jean Harb, and Doina Precup.
\newblock The option-critic architecture.
\newblock In {\em Proceedings of the AAAI Conference on Artificial
  Intelligence}, volume~31, 2017.

\bibitem{badia2020agent57}
Adri{\`a}~Puigdom{\`e}nech Badia, Bilal Piot, Steven Kapturowski, Pablo
  Sprechmann, Alex Vitvitskyi, Daniel Guo, and Charles Blundell.
\newblock Agent57: Outperforming the {Atari} human benchmark.
\newblock {\em arXiv preprint arXiv:2003.13350}, 2020.

\bibitem{bahdanau2014neural}
Dzmitry Bahdanau, Kyunghyun Cho, and Yoshua Bengio.
\newblock Neural machine translation by jointly learning to align and
  translate.
\newblock {\em arXiv preprint arXiv:1409.0473}, 2014.

\bibitem{baird1995residual}
Leemon Baird.
\newblock Residual algorithms: Reinforcement learning with function
  approximation.
\newblock In {\em Machine Learning Proceedings 1995}, pages 30--37. Elsevier,
  1995.

\bibitem{baker2019emergent}
Bowen Baker, Ingmar Kanitscheider, Todor Markov, Yi~Wu, Glenn Powell, Bob
  McGrew, and Igor Mordatch.
\newblock Emergent tool use from multi-agent autocurricula.
\newblock {\em arXiv preprint arXiv:1909.07528}, 2019.

\bibitem{bansal2017emergent}
Trapit Bansal, Jakub Pachocki, Szymon Sidor, Ilya Sutskever, and Igor Mordatch.
\newblock Emergent complexity via multi-agent competition.
\newblock {\em arXiv preprint arXiv:1710.03748}, 2017.

\bibitem{baral2003knowledge}
Chitta Baral.
\newblock {\em Knowledge Representation, Reasoning and Declarative Problem
  Solving}.
\newblock Cambridge university press, 2003.

\bibitem{bard2020hanabi}
Nolan Bard, Jakob~N. Foerster, Sarath Chandar, Neil Burch, Marc Lanctot,
  H.~Francis Song, Emilio Parisotto, Vincent Dumoulin, Subhodeep Moitra, Edward
  Hughes, Iain Dunning, Shibl Mourad, Hugo Larochelle, Marc~G. Bellemare, and
  Michael Bowling.
\newblock The {Hanabi} challenge: A new frontier for {AI} research.
\newblock {\em Artificial Intelligence}, 280:103216, 2020.

\bibitem{bard2013annual}
Nolan Bard, John Hawkin, Jonathan Rubin, and Martin Zinkevich.
\newblock The annual computer poker competition.
\newblock {\em AI Magazine}, 34(2):112, 2013.

\bibitem{baron1985does}
Simon Baron-Cohen, Alan~M Leslie, and Uta Frith.
\newblock Does the autistic child have a ``theory of mind''?
\newblock {\em Cognition}, 21(1):37--46, 1985.

\bibitem{barron1998minimum}
Andrew Barron, Jorma Rissanen, and Bin Yu.
\newblock The minimum description length principle in coding and modeling.
\newblock {\em IEEE Transactions on Information Theory}, 44(6):2743--2760,
  1998.

\bibitem{barto2003recent}
Andrew~G Barto and Sridhar Mahadevan.
\newblock Recent advances in hierarchical reinforcement learning.
\newblock {\em Discrete Event Dynamic Systems}, 13(1-2):41--77, 2003.

\bibitem{barto1983neuronlike}
Andrew~G Barto, Richard~S Sutton, and Charles~W Anderson.
\newblock Neuronlike adaptive elements that can solve difficult learning
  control problems.
\newblock {\em IEEE Transactions on Systems, Man, and Cybernetics},
  (5):834--846, 1983.

\bibitem{openaibaseline2017}
OpenAI Baselines.
\newblock {DQN} \url{https://openai.com/blog/openai-baselines-dqn/}, 2017.

\bibitem{baxter2000model}
Jonathan Baxter.
\newblock A model of inductive bias learning.
\newblock {\em Journal of Artificial Intelligence Research}, 12:149--198, 2000.

\bibitem{baxter1999knightcap}
Jonathan Baxter, Andrew Tridgell, and Lex Weaver.
\newblock Knightcap: a chess program that learns by combining {TD} $(\lambda)$
  with game-tree search.
\newblock {\em arXiv preprint cs/9901002}, 1999.

\bibitem{baxter2000learning}
Jonathan Baxter, Andrew Tridgell, and Lex Weaver.
\newblock Learning to play chess using temporal differences.
\newblock {\em Machine Learning}, 40(3):243--263, 2000.

\bibitem{beal2000temporal}
Don Beal and Martin~C. Smith.
\newblock Temporal difference learning for heuristic search and game playing.
\newblock {\em Information Sciences}, 122(1):3--21, 2000.

\bibitem{bear2007neuroscience}
Mark~F Bear, Barry~W Connors, and Michael~A Paradiso.
\newblock {\em Neuroscience}, volume~2.
\newblock Lippincott Williams \& Wilkins, 2007.

\bibitem{beattie2016deepmind}
Charles Beattie, Joel~Z. Leibo, Denis Teplyashin, Tom Ward, Marcus Wainwright,
  Heinrich K{\"{u}}ttler, Andrew Lefrancq, Simon Green, V{\'{\i}}ctor
  Vald{\'{e}}s, Amir Sadik, Julian Schrittwieser, Keith Anderson, Sarah York,
  Max Cant, Adam Cain, Adrian Bolton, Stephen Gaffney, Helen King, Demis
  Hassabis, Shane Legg, and Stig Petersen.
\newblock Deepmind lab.
\newblock {\em arXiv preprint arXiv:1612.03801}, 2016.

\bibitem{beljaards2017}
Laurens Beljaards.
\newblock Ai agents for the abstract strategy game tak.
\newblock Master's thesis, Leiden University, 2017.

\bibitem{belkin2018reconciling}
Mikhail Belkin, Daniel Hsu, Siyuan Ma, and Soumik Mandal.
\newblock Reconciling modern machine learning and the bias-variance trade-off.
\newblock {\em arXiv preprint arXiv:1812.11118}, 2018.

\bibitem{belkin2019two}
Mikhail Belkin, Daniel Hsu, and Ji~Xu.
\newblock Two models of double descent for weak features.
\newblock {\em arXiv preprint arXiv:1903.07571}, 2019.

\bibitem{belkin2018overfitting}
Mikhail Belkin, Daniel~J Hsu, and Partha Mitra.
\newblock Overfitting or perfect fitting? {Risk} bounds for classification and
  regression rules that interpolate.
\newblock In {\em Advances in Neural Information Processing Systems}, pages
  2300--2311, 2018.

\bibitem{bellemare2013bayesian}
Marc Bellemare, Joel Veness, and Michael Bowling.
\newblock Bayesian learning of recursively factored environments.
\newblock In {\em International Conference on Machine Learning}, pages
  1211--1219, 2013.

\bibitem{bellemare2017distributional}
Marc~G Bellemare, Will Dabney, and R{\'e}mi Munos.
\newblock A distributional perspective on reinforcement learning.
\newblock In {\em International Conference on Machine Learning}, pages
  449--458, 2017.

\bibitem{bellemare2013arcade}
Marc~G Bellemare, Yavar Naddaf, Joel Veness, and Michael Bowling.
\newblock The {Arcade Learning Environment}: An evaluation platform for general
  agents.
\newblock {\em Journal of Artificial Intelligence Research}, 47:253--279, 2013.

\bibitem{bellman1957dynamic}
Richard Bellman.
\newblock {\em Dynamic Programming}.
\newblock Courier Corporation, 1957, 2013.

\bibitem{bellman1965application}
Richard Bellman.
\newblock On the application of dynamic programing to the determination of
  optimal play in chess and checkers.
\newblock {\em Proceedings of the National Academy of Sciences},
  53(2):244--247, 1965.

\bibitem{ben2007analysis}
Shai Ben-David, John Blitzer, Koby Crammer, and Fernando Pereira.
\newblock Analysis of representations for domain adaptation.
\newblock {\em Advances in Neural Information Processing Systems}, 19:137,
  2007.

\bibitem{bengio1990learning}
Yoshua Bengio, Samy Bengio, and Jocelyn Cloutier.
\newblock Learning a synaptic learning rule.
\newblock Technical report, Montreal, 1990.

\bibitem{bengio2013representation}
Yoshua Bengio, Aaron Courville, and Pascal Vincent.
\newblock Representation learning: A review and new perspectives.
\newblock {\em IEEE Transactions on Pattern Analysis and Machine Intelligence},
  35(8):1798--1828, 2013.

\bibitem{bengio2009curriculum}
Yoshua Bengio, J{\'e}r{\^o}me Louradour, Ronan Collobert, and Jason Weston.
\newblock Curriculum learning.
\newblock In {\em {Proceedings of the 26th Annual International Conference on
  Machine Learning}}, pages 41--48, 2009.

\bibitem{beni2020swarm}
Gerardo Beni.
\newblock Swarm intelligence.
\newblock {\em Complex Social and Behavioral Systems: Game Theory and
  Agent-Based Models}, pages 791--818, 2020.

\bibitem{beni1993swarm}
Gerardo Beni and Jing Wang.
\newblock Swarm intelligence in cellular robotic systems.
\newblock In {\em Robots and Biological Systems: Towards a New Bionics?}, pages
  703--712. Springer, 1993.

\bibitem{berner2019dota}
Christopher Berner, Greg Brockman, Brooke Chan, Vicki Cheung, Przemyslaw
  Debiak, Christy Dennison, David Farhi, Quirin Fischer, Shariq Hashme,
  Christopher Hesse, Rafal J{\'{o}}zefowicz, Scott Gray, Catherine Olsson,
  Jakub Pachocki, Michael Petrov, Henrique~Pond{\'{e}} de~Oliveira~Pinto,
  Jonathan Raiman, Tim Salimans, Jeremy Schlatter, Jonas Schneider, Szymon
  Sidor, Ilya Sutskever, Jie Tang, Filip Wolski, and Susan Zhang.
\newblock Dota 2 with large scale deep reinforcement learning.
\newblock {\em arXiv preprint arXiv:1912.06680}, 2019.

\bibitem{berners2001semantic}
Tim Berners-Lee, James Hendler, and Ora Lassila.
\newblock The semantic web.
\newblock {\em Scientific American}, 284(5):28--37, 2001.

\bibitem{bernstein2002complexity}
Daniel~S Bernstein, Robert Givan, Neil Immerman, and Shlomo Zilberstein.
\newblock The complexity of decentralized control of markov decision processes.
\newblock {\em Mathematics of Operations Research}, 27(4):819--840, 2002.

\bibitem{bertinetto2018meta}
Luca Bertinetto, Joao~F Henriques, Philip~HS Torr, and Andrea Vedaldi.
\newblock Meta-learning with differentiable closed-form solvers.
\newblock In {\em International Conference on Learning Representations}, 2018.

\bibitem{bertolami2009novel}
R~Bertolami, H~Bunke, S~Fernandez, A~Graves, M~Liwicki, and J~Schmidhuber.
\newblock A novel connectionist system for improved unconstrained handwriting
  recognition.
\newblock {\em IEEE Transactions on Pattern Analysis and Machine Intelligence},
  31(5), 2009.

\bibitem{bertsekas1995dynamic}
Dimitri~P Bertsekas, Dimitri~P Bertsekas, Dimitri~P Bertsekas, and Dimitri~P
  Bertsekas.
\newblock {\em Dynamic Programming and Optimal Control}, volume~1.
\newblock Athena scientific Belmont, MA, 1995.

\bibitem{bertsekas1996neuro}
Dimitri~P Bertsekas and John Tsitsiklis.
\newblock {\em Neuro-Dynamic Programming}.
\newblock MIT Press Cambridge, 1996.

\bibitem{chichilicious}
Shrisha Bharadwaj.
\newblock Embarrsingly simple zero shot learning.
\newblock
  \url{https://github.com/chichilicious/embarrsingly-simple-zero-shot-learning},
  2018.

\bibitem{bhatnagar2009convergent}
Shalabh Bhatnagar, Doina Precup, David Silver, Richard~S Sutton, Hamid~R Maei,
  and Csaba Szepesv{\'a}ri.
\newblock Convergent temporal-difference learning with arbitrary smooth
  function approximation.
\newblock In {\em Advances in Neural Information Processing Systems}, pages
  1204--1212, 2009.

\bibitem{billings2002challenge}
Darse Billings, Aaron Davidson, Jonathan Schaeffer, and Duane Szafron.
\newblock The challenge of poker.
\newblock {\em Artificial Intelligence}, 134(1-2):201--240, 2002.

\bibitem{billings2004game}
Darse Billings, Aaron Davidson, Terence Schauenberg, Neil Burch, Michael
  Bowling, Robert Holte, Jonathan Schaeffer, and Duane Szafron.
\newblock Game-tree search with adaptation in stochastic imperfect-information
  games.
\newblock In {\em International Conference on Computers and Games}, pages
  21--34. Springer, 2004.

\bibitem{billings1998opponent}
Darse Billings, Denis Papp, Jonathan Schaeffer, and Duane Szafron.
\newblock Opponent modeling in poker.
\newblock {\em {AAAI/IAAI}}, 493:499, 1998.

\bibitem{bird2009natural}
Steven Bird, Ewan Klein, and Edward Loper.
\newblock {\em Natural language processing with Python: analyzing text with the
  natural language toolkit}.
\newblock O'Reilly Media, Inc., 2009.

\bibitem{bishop2006pattern}
Christopher~M Bishop.
\newblock {\em Pattern Recognition and Machine Learning}.
\newblock Information science and statistics. Springer Verlag, Heidelberg,
  2006.

\bibitem{bloem2019}
Peter Bloem.
\newblock Transformers \url{http://peterbloem.nl/blog/transformers}.

\bibitem{blum2008swarm}
Christian Blum and Daniel Merkle.
\newblock {\em Swarm Intelligence: Introduction and Applications}.
\newblock Springer Science \& Business Media, 2008.

\bibitem{bommasani2021opportunities}
Rishi Bommasani, Drew~A. Hudson, Ehsan Adeli, Russ Altman, Simran Arora, Sydney
  von Arx, Michael~S. Bernstein, Jeannette Bohg, Antoine Bosselut, Emma
  Brunskill, Erik Brynjolfsson, Shyamal Buch, Dallas Card, Rodrigo Castellon,
  Niladri Chatterji, Annie~S. Chen, Kathleen Creel, Jared~Quincy Davis,
  Dorottya Demszky, Chris Donahue, Moussa Doumbouya, Esin Durmus, Stefano
  Ermon, John Etchemendy, Kawin Ethayarajh, Li~Fei{-}Fei, Chelsea Finn, Trevor
  Gale, Lauren Gillespie, Karan Goel, Noah~D. Goodman, Shelby Grossman, Neel
  Guha, Tatsunori Hashimoto, Peter Henderson, John Hewitt, Daniel~E. Ho, Jenny
  Hong, Kyle Hsu, Jing Huang, Thomas Icard, Saahil Jain, Dan Jurafsky,
  Pratyusha Kalluri, Siddharth Karamcheti, Geoff Keeling, Fereshte Khani, Omar
  Khattab, Pang~Wei Koh, Mark~S. Krass, Ranjay Krishna, and Rohith Kuditipudi.
\newblock On the opportunities and risks of foundation models.
\newblock {\em arXiv preprint arXiv:2108.07258}, 2021.

\bibitem{bonabeau1999swarm}
Eric Bonabeau, Marco Dorigo, and Guy Theraulaz.
\newblock {\em Swarm Intelligence: From Natural to Artificial Systems}.
\newblock Oxford University Press, 1999.

\bibitem{borealis2019}
Borealis.
\newblock Few shot learning tutorial
  \url{https://www.borealisai.com/en/blog/tutorial-2-few-shot-learning-and-meta-learning-i/}.

\bibitem{botev2013cross}
Zdravko~I Botev, Dirk~P Kroese, Reuven~Y Rubinstein, and Pierre L'Ecuyer.
\newblock The cross-entropy method for optimization.
\newblock In {\em Handbook of Statistics}, volume~31, pages 35--59. Elsevier,
  2013.

\bibitem{botvinick2019reinforcement}
Matthew Botvinick, Sam Ritter, Jane~X Wang, Zeb Kurth-Nelson, Charles Blundell,
  and Demis Hassabis.
\newblock Reinforcement learning, fast and slow.
\newblock {\em Trends in Cognitive Sciences}, 23(5):408--422, 2019.

\bibitem{botvinick2009hierarchically}
Matthew~M Botvinick, Yael Niv, and Andew~G Barto.
\newblock Hierarchically organized behavior and its neural foundations: a
  reinforcement learning perspective.
\newblock {\em Cognition}, 113(3):262--280, 2009.

\bibitem{bouzy2004monte}
Bruno Bouzy and Bernard Helmstetter.
\newblock Monte {Carlo} {Go} developments.
\newblock In {\em Advances in Computer Games}, pages 159--174. Springer, 2004.

\bibitem{bowling2015heads}
Michael Bowling, Neil Burch, Michael Johanson, and Oskari Tammelin.
\newblock Heads-up {Limit Hold'em} poker is solved.
\newblock {\em Science}, 347(6218):145--149, 2015.

\bibitem{bowling2009demonstration}
Michael~H. Bowling, Nicholas~Abou Risk, Nolan Bard, Darse Billings, Neil Burch,
  Joshua Davidson, John~Alexander Hawkin, Robert Holte, Michael Johanson,
  Morgan Kan, Bryce Paradis, Jonathan Schaeffer, David Schnizlein, Duane
  Szafron, Kevin Waugh, and Martin Zinkevich.
\newblock A demonstration of the polaris poker system.
\newblock In {\em Proceedings of The 8th International Conference on Autonomous
  Agents and Multiagent Systems}, volume~2, pages 1391--1392, 2009.

\bibitem{boyd1988culture}
Robert Boyd and Peter~J Richerson.
\newblock {\em Culture and the Evolutionary Process}.
\newblock University of Chicago press, 1988.

\bibitem{brazdil2008metalearning}
Pavel Brazdil, Christophe~Giraud Carrier, Carlos Soares, and Ricardo Vilalta.
\newblock {\em Metalearning: Applications to data mining}.
\newblock Springer Science \& Business Media, 2008.

\bibitem{brochu2010tutorial}
Eric Brochu, Vlad~M Cora, and Nando De~Freitas.
\newblock A tutorial on {Bayesian} optimization of expensive cost functions,
  with application to active user modeling and hierarchical reinforcement
  learning.
\newblock {\em arXiv preprint arXiv:1012.2599}, 2010.

\bibitem{brockman2016openai}
Greg Brockman, Vicki Cheung, Ludwig Pettersson, Jonas Schneider, John Schulman,
  Jie Tang, and Wojciech Zaremba.
\newblock {OpenAI} {G}ym.
\newblock {\em arXiv preprint arXiv:1606.01540}, 2016.

\bibitem{brooks1991intelligence}
Rodney~A Brooks.
\newblock Intelligence without representation.
\newblock {\em Artificial Intelligence}, 47(1-3):139--159, 1991.

\bibitem{brown2015hierarchical}
Noam Brown, Sam Ganzfried, and Tuomas Sandholm.
\newblock Hierarchical abstraction, distributed equilibrium computation, and
  post-processing, with application to a champion {No-Limit Texas Hold'em}
  agent.
\newblock In {\em AAAI Workshop: Computer Poker and Imperfect Information},
  2015.

\bibitem{brown2019deep}
Noam Brown, Adam Lerer, Sam Gross, and Tuomas Sandholm.
\newblock Deep counterfactual regret minimization.
\newblock In {\em International Conference on Machine Learning}, pages
  793--802. PMLR, 2019.

\bibitem{brown2018superhuman}
Noam Brown and Tuomas Sandholm.
\newblock Superhuman {AI} for {Heads-up No-limit} poker: Libratus beats top
  professionals.
\newblock {\em Science}, 359(6374):418--424, 2018.

\bibitem{brown2019superhuman}
Noam Brown and Tuomas Sandholm.
\newblock Superhuman {AI} for multiplayer poker.
\newblock {\em Science}, 365(6456):885--890, 2019.

\bibitem{brown2020language}
Tom~B. Brown, Benjamin Mann, Nick Ryder, Melanie Subbiah, Jared Kaplan,
  Prafulla Dhariwal, Arvind Neelakantan, Pranav Shyam, Girish Sastry, Amanda
  Askell, Sandhini Agarwal, Ariel Herbert{-}Voss, Gretchen Krueger, Tom
  Henighan, Rewon Child, Aditya Ramesh, Daniel~M. Ziegler, Jeffrey Wu, Clemens
  Winter, Christopher Hesse, Mark Chen, Eric Sigler, Mateusz Litwin, Scott
  Gray, Benjamin Chess, Jack Clark, Christopher Berner, Sam McCandlish, Alec
  Radford, Ilya Sutskever, and Dario Amodei.
\newblock Language models are few-shot learners.
\newblock In {\em Advances in Neural Information Processing Systems}, 2020.

\bibitem{browne2000hex}
Cameron Browne.
\newblock {\em Hex Strategy}.
\newblock AK Peters/CRC Press, 2000.

\bibitem{browne2019strategic}
Cameron Browne, Dennis~JNJ Soemers, and Eric Piette.
\newblock Strategic features for general games.
\newblock In {\em KEG@ AAAI}, pages 70--75, 2019.

\bibitem{browne2012survey}
Cameron~B Browne, Edward Powley, Daniel Whitehouse, Simon~M Lucas, Peter~I
  Cowling, Philipp Rohlfshagen, Stephen Tavener, Diego Perez, Spyridon
  Samothrakis, and Simon Colton.
\newblock A survey of {Monte} {Carlo} {Tree} {Search} methods.
\newblock {\em IEEE Transactions on Computational Intelligence and AI in
  Games}, 4(1):1--43, 2012.

\bibitem{brugmann1993monte}
Bernd Br{\"u}gmann.
\newblock Monte {Carlo} {Go}.
\newblock Technical report, Syracuse University, 1993.

\bibitem{buchberger1982computer}
Bruno Buchberger, George~E Collins, R{\"u}diger Loos, and Rudolph Albrecht.
\newblock Computer algebra symbolic and algebraic computation.
\newblock {\em ACM SIGSAM Bulletin}, 16(4):5--5, 1982.

\bibitem{bucilua2006model}
Cristian Bucilu{\v a}, Rich Caruana, and Alexandru Niculescu-Mizil.
\newblock Model compression.
\newblock In {\em Proceedings of the 12th ACM SIGKDD International Conference
  on Knowledge Discovery and Data Mining}, pages 535--541, 2006.

\bibitem{buesing2018learning}
Lars Buesing, Theophane Weber, S{\'e}bastien Racaniere, SM~Eslami, Danilo
  Rezende, David~P Reichert, Fabio Viola, Frederic Besse, Karol Gregor, Demis
  Hassabis, and Daan Wierstra.
\newblock Learning and querying fast generative models for reinforcement
  learning.
\newblock {\em arXiv preprint arXiv:1802.03006}, 2018.

\bibitem{busoniu2008comprehensive}
Lucian Busoniu, Robert Babuska, and Bart De~Schutter.
\newblock A comprehensive survey of multiagent reinforcement learning.
\newblock {\em IEEE Transactions on Systems, Man, and Cybernetics, Part C
  (Applications and Reviews)}, 38(2):156--172, 2008.

\bibitem{cai2021safe}
Zhiyuan Cai, Huanhui Cao, Wenjie Lu, Lin Zhang, and Hao Xiong.
\newblock Safe multi-agent reinforcement learning through decentralized
  multiple control barrier functions.
\newblock {\em arXiv preprint arXiv:2103.12553}, 2021.

\bibitem{campbell2002deep}
Murray Campbell, A~Joseph Hoane~Jr, and Feng-Hsiung Hsu.
\newblock Deep {Blue}.
\newblock {\em Artificial Intelligence}, 134(1-2):57--83, 2002.

\bibitem{campero2020learning}
Andres Campero, Roberta Raileanu, Heinrich K{\"u}ttler, Joshua~B Tenenbaum, Tim
  Rockt{\"a}schel, and Edward Grefenstette.
\newblock Learning with {AMIGo}: Adversarially motivated intrinsic goals.
\newblock In {\em International Conference on Learning Representations}, 2020.

\bibitem{cao2018emergent}
Kris Cao, Angeliki Lazaridou, Marc Lanctot, Joel~Z Leibo, Karl Tuyls, and
  Stephen Clark.
\newblock Emergent communication through negotiation.
\newblock In {\em International Conference on Learning Representations}, 2018.

\bibitem{carr2018domain}
Thomas Carr, Maria Chli, and George Vogiatzis.
\newblock Domain adaptation for reinforcement learning on the atari.
\newblock In {\em Proceedings of the 18th International Conference on
  Autonomous Agents and MultiAgent Systems, {AAMAS} '19, Montreal}, pages
  1859--1861, 2018.

\bibitem{cartwright2018behavioral}
Edward Cartwright.
\newblock {\em Behavioral Economics}.
\newblock Routledge, 2018.

\bibitem{caruana1997multitask}
Rich Caruana.
\newblock Multitask learning.
\newblock {\em Machine Learning}, 28(1):41--75, 1997.

\bibitem{caruana2001overfitting}
Rich Caruana, Steve Lawrence, and C~Lee Giles.
\newblock Overfitting in neural nets: Backpropagation, conjugate gradient, and
  early stopping.
\newblock In {\em Advances in Neural Information Processing Systems}, pages
  402--408, 2001.

\bibitem{castro2018dopamine}
Pablo~Samuel Castro, Subhodeep Moitra, Carles Gelada, Saurabh Kumar, and Marc~G
  Bellemare.
\newblock Dopamine: A research framework for deep reinforcement learning.
\newblock {\em arXiv preprint arXiv:1812.06110}, 2018.

\bibitem{cazenave2018residual}
Tristan Cazenave.
\newblock Residual networks for computer {Go}.
\newblock {\em IEEE Transactions on Games}, 10(1):107--110, 2018.

\bibitem{cazenave2020polygames}
Tristan Cazenave, Yen{-}Chi Chen, Guan{-}Wei Chen, Shi{-}Yu Chen, Xian{-}Dong
  Chiu, Julien Dehos, Maria Elsa, Qucheng Gong, Hengyuan Hu, Vasil Khalidov,
  Cheng{-}Ling Li, Hsin{-}I Lin, Yu{-}Jin Lin, Xavier Martinet, Vegard Mella,
  J{\'{e}}r{\'{e}}my Rapin, Baptiste Rozi{\`{e}}re, Gabriel Synnaeve, Fabien
  Teytaud, Olivier Teytaud, Shi{-}Cheng Ye, Yi{-}Jun Ye, Shi{-}Jim Yen, and
  Sergey Zagoruyko.
\newblock Polygames: Improved zero learning.
\newblock {\em arXiv preprint arXiv:2001.09832}, 2020.

\bibitem{cazenave2005combining}
Tristan Cazenave and Bernard Helmstetter.
\newblock Combining tactical search and {Monte-Carlo} in the game of {Go}.
\newblock In {\em Proceedings of the 2005 {IEEE} Symposium on Computational
  Intelligence and Games (CIG05), Essex University}, volume~5, pages 171--175,
  2005.

\bibitem{chang2005adaptive}
Hyeong~Soo Chang, Michael~C Fu, Jiaqiao Hu, and Steven~I Marcus.
\newblock An adaptive sampling algorithm for solving {Markov} decision
  processes.
\newblock {\em Operations Research}, 53(1):126--139, 2005.

\bibitem{chao2013}
Yang Chao.
\newblock Share and play new sokoban levels. \url{http://Sokoban.org}, 2013.

\bibitem{chaslot2010monte}
Guillaume Chaslot.
\newblock {\em {Monte-Carlo} tree search}.
\newblock PhD thesis, Maastricht University, 2010.

\bibitem{chaslot2008monte}
Guillaume Chaslot, Sander Bakkes, Istvan Szita, and Pieter Spronck.
\newblock {Monte-Carlo} tree search: A new framework for game {AI}.
\newblock In {\em AIIDE}, 2008.

\bibitem{chellapilla1999evolving}
Kumar Chellapilla and David~B Fogel.
\newblock Evolving neural networks to play checkers without relying on expert
  knowledge.
\newblock {\em IEEE Transactions on Neural Networks}, 10(6):1382--1391, 1999.

\bibitem{chen2021decision}
Lili Chen, Kevin Lu, Aravind Rajeswaran, Kimin Lee, Aditya Grover, Michael
  Laskin, Pieter Abbeel, Aravind Srinivas, and Igor Mordatch.
\newblock Decision transformer: Reinforcement learning via sequence modeling.
\newblock {\em arXiv preprint arXiv:2106.01345}, 2021.

\bibitem{chen2019closer}
Wei-Yu Chen, Yen-Cheng Liu, Zsolt Kira, Yu-Chiang~Frank Wang, and Jia-Bin
  Huang.
\newblock A closer look at few-shot classification.
\newblock In {\em International Conference on Learning Representations}, 2019.

\bibitem{cheng2017survey}
Yu~Cheng, Duo Wang, Pan Zhou, and Tao Zhang.
\newblock A survey of model compression and acceleration for deep neural
  networks.
\newblock {\em arXiv preprint arXiv:1710.09282}, 2017.

\bibitem{chevalierminimalistic}
Maxime Chevalier-Boisvert, Lucas Willems, and Sumans Pal.
\newblock Minimalistic gridworld environment for {OpenAI Gym}
  \url{https://github.com/maximecb/gym-minigrid}, 2018.

\bibitem{chiappa2017recurrent}
Silvia Chiappa, S{\'e}bastien Racaniere, Daan Wierstra, and Shakir Mohamed.
\newblock Recurrent environment simulators.
\newblock In {\em International Conference on Learning Representations}, 2017.

\bibitem{choi1996pb}
Jaeyoung Choi, Jack~J Dongarra, and David~W Walker.
\newblock {PB-BLAS}: a set of parallel block basic linear algebra subprograms.
\newblock {\em Concurrency: Practice and Experience}, 8(7):517--535, 1996.

\bibitem{chollet2015keras}
Fran{\c{c}}ois Chollet.
\newblock Keras.
\newblock \url{https://keras.io}, 2015.

\bibitem{chollet2017deep}
Fran{\c{c}}ois Chollet.
\newblock {\em Deep learning with Python}.
\newblock Manning Publications Co., 2017.

\bibitem{chrabaszcz2018back}
Patryk Chrabaszcz, Ilya Loshchilov, and Frank Hutter.
\newblock Back to basics: Benchmarking canonical evolution strategies for
  playing {Atari}.
\newblock In {\em Proceedings of the Twenty-Seventh International Joint
  Conference on Artificial Intelligence, {IJCAI} 2018, July 13-19, 2018,
  Stockholm}, pages 1419--1426, 2018.

\bibitem{chua2018deep}
Kurtland Chua, Roberto Calandra, Rowan McAllister, and Sergey Levine.
\newblock Deep reinforcement learning in a handful of trials using
  probabilistic dynamics models.
\newblock In {\em Advances in Neural Information Processing Systems}, pages
  4754--4765, 2018.

\bibitem{chung2016hierarchical}
Junyoung Chung, Sungjin Ahn, and Yoshua Bengio.
\newblock Hierarchical multiscale recurrent neural networks.
\newblock In {\em International Conference on Learning Representations}, 2016.

\bibitem{ciliberto2015convex}
Carlo Ciliberto, Youssef Mroueh, Tomaso Poggio, and Lorenzo Rosasco.
\newblock Convex learning of multiple tasks and their structure.
\newblock In {\em International Conference on Machine Learning}, pages
  1548--1557. PMLR, 2015.

\bibitem{cirecsan2010deep}
Dan Cire{\c{s}}an, Ueli Meier, Luca~Maria Gambardella, and J{\"u}rgen
  Schmidhuber.
\newblock Deep, big, simple neural nets for handwritten digit recognition.
\newblock {\em Neural Computation}, 22(12):3207--3220, 2010.

\bibitem{cirecsan2012multi}
Dan Cire{\c{s}}an, Ueli Meier, and J{\"u}rgen Schmidhuber.
\newblock Multi-column deep neural networks for image classification.
\newblock In {\em 2012 {IEEE} Conference on Computer Vision and Pattern
  Recognition, Providence, RI, US}, pages 3642--3649, 2012.

\bibitem{clark2014teaching}
Christopher Clark and Amos Storkey.
\newblock Teaching deep convolutional neural networks to play {Go}. arxiv
  preprint.
\newblock {\em arXiv preprint arXiv:1412.3409}, 1, 2014.

\bibitem{clark2015training}
Christopher Clark and Amos Storkey.
\newblock Training deep convolutional neural networks to play {Go}.
\newblock In {\em International Conference on Machine Learning}, pages
  1766--1774, 2015.

\bibitem{clavera2018model}
Ignasi Clavera, Jonas Rothfuss, John Schulman, Yasuhiro Fujita, Tamim Asfour,
  and Pieter Abbeel.
\newblock Model-based reinforcement learning via meta-policy optimization.
\newblock In {\em 2nd Annual Conference on Robot Learning, CoRL 2018,
  Z{\"{u}}rich, Switzerland}, pages 617--629, 2018.

\bibitem{clocksin2012programming}
William~F Clocksin and Christopher~S Mellish.
\newblock {\em Programming in Prolog: Using the ISO standard}.
\newblock Springer Science \& Business Media, 1981.

\bibitem{cobbe2020leveraging}
Karl Cobbe, Chris Hesse, Jacob Hilton, and John Schulman.
\newblock Leveraging procedural generation to benchmark reinforcement learning.
\newblock In {\em International Conference on Machine Learning}, pages
  2048--2056. PMLR, 2020.

\bibitem{cobbe2018quantifying}
Karl Cobbe, Oleg Klimov, Chris Hesse, Taehoon Kim, and John Schulman.
\newblock Quantifying generalization in reinforcement learning.
\newblock In {\em International Conference on Machine Learning}, pages
  1282--1289, 2018.

\bibitem{coelho1986automated}
Helder Coelho and Luis~Moniz Pereira.
\newblock Automated reasoning in geometry theorem proving with prolog.
\newblock {\em Journal of Automated Reasoning}, 2(4):329--390, 1986.

\bibitem{colas2019curious}
C{\'e}dric Colas, Pierre Fournier, Mohamed Chetouani, Olivier Sigaud, and
  Pierre-Yves Oudeyer.
\newblock Curious: intrinsically motivated modular multi-goal reinforcement
  learning.
\newblock In {\em International Conference on Machine Learning}, pages
  1331--1340. PMLR, 2019.

\bibitem{colas2020intrinsically}
C{\'e}dric Colas, Tristan Karch, Olivier Sigaud, and Pierre-Yves Oudeyer.
\newblock Intrinsically motivated goal-conditioned reinforcement learning: a
  short survey.
\newblock {\em arXiv preprint arXiv:2012.09830}, 2020.

\bibitem{conti2017improving}
Edoardo Conti, Vashisht Madhavan, Felipe~Petroski Such, Joel Lehman, Kenneth~O
  Stanley, and Jeff Clune.
\newblock Improving exploration in evolution strategies for deep reinforcement
  learning via a population of novelty-seeking agents.
\newblock In {\em Advances in Neural Information Processing Systems}, pages
  5032--5043, 2018.

\bibitem{coulom2006efficient}
R{\'e}mi Coulom.
\newblock Efficient selectivity and backup operators in {Monte-Carlo} {Tree
  Search}.
\newblock In {\em International Conference on Computers and Games}, pages
  72--83. Springer, 2006.

\bibitem{coulom2007monte}
R{\'e}mi Coulom.
\newblock {Monte-Carlo} tree search in {Crazy Stone}.
\newblock In {\em Proceedings Game Programming Workshop, Tokyo, Japan}, pages
  74--75, 2007.

\bibitem{coulom2009monte}
R{\'e}mi Coulom.
\newblock The {Monte-Carlo} revolution in {Go}.
\newblock In {\em The Japanese-French Frontiers of Science Symposium (JFFoS
  2008), Roscoff, France}, 2009.

\bibitem{coumans2019}
Erwin Coumans and Yunfei Bai.
\newblock Pybullet, a python module for physics simulation for games, robotics
  and machine learning.
\newblock \url{http://pybullet.org}, 2016--2019.

\bibitem{csurka2017domain}
Gabriela Csurka.
\newblock Domain adaptation for visual applications: A comprehensive survey.
\newblock In {\em Domain Adaptation in Computer Vision Applications}, Advances
  in Computer Vision and Pattern Recognition, pages 1--35. Springer, 2017.

\bibitem{culberson1997sokoban}
Joseph Culberson.
\newblock Sokoban is {PSPACE}-complete.
\newblock Technical report, University of Alberta, 1997.

\bibitem{culberson1998pattern}
Joseph~C Culberson and Jonathan Schaeffer.
\newblock Pattern databases.
\newblock {\em Computational Intelligence}, 14(3):318--334, 1998.

\bibitem{currie1991plan}
Ken Currie and Austin Tate.
\newblock O-plan: the open planning architecture.
\newblock {\em Artificial Intelligence}, 52(1):49--86, 1991.

\bibitem{czarnecki2020real}
Wojciech~Marian Czarnecki, Gauthier Gidel, Brendan Tracey, Karl Tuyls, Shayegan
  Omidshafiei, David Balduzzi, and Max Jaderberg.
\newblock Real world games look like spinning tops.
\newblock In {\em Advances in Neural Information Processing Systems}, 2020.

\bibitem{czarnogorski2018monte}
Kamil Czarnog{\'o}rski.
\newblock {Monte Carlo Tree Search} beginners guide
  \url{https://int8.io/monte-carlo-tree-search-beginners-guide/}, 2018.

\bibitem{dabney2020distributional}
Will Dabney, Zeb Kurth-Nelson, Naoshige Uchida, Clara~Kwon Starkweather, Demis
  Hassabis, R{\'e}mi Munos, and Matthew Botvinick.
\newblock A distributional code for value in dopamine-based reinforcement
  learning.
\newblock {\em Nature}, pages 1--5, 2020.

\bibitem{dafoe2020open}
Allan Dafoe, Edward Hughes, Yoram Bachrach, Tantum Collins, Kevin~R McKee,
  Joel~Z Leibo, Kate Larson, and Thore Graepel.
\newblock Open problems in cooperative {AI}.
\newblock {\em arXiv preprint arXiv:2012.08630}, 2020.

\bibitem{dai2020r2}
Zhongxiang Dai, Yizhou Chen, Bryan Kian~Hsiang Low, Patrick Jaillet, and
  Teck-Hua Ho.
\newblock {R2-B2:} recursive reasoning-based {Bayesian} optimization for
  no-regret learning in games.
\newblock In {\em International Conference on Machine Learning}, pages
  2291--2301. PMLR, 2020.

\bibitem{daniel2016probabilistic}
Christian Daniel, Herke Van~Hoof, Jan Peters, and Gerhard Neumann.
\newblock Probabilistic inference for determining options in reinforcement
  learning.
\newblock {\em Machine Learning}, 104(2):337--357, 2016.

\bibitem{das2016incorporating}
Shubhomoy Das, Weng-Keen Wong, Thomas Dietterich, Alan Fern, and Andrew Emmott.
\newblock Incorporating expert feedback into active anomaly discovery.
\newblock In {\em 2016 IEEE 16th International Conference on Data Mining
  (ICDM)}, pages 853--858. IEEE, 2016.

\bibitem{daume2009frustratingly}
Hal Daum{\'e}~III.
\newblock Frustratingly easy domain adaptation.
\newblock In {\em {ACL} 2007, Proceedings of the 45th Annual Meeting of the
  Association for Computational Linguistics, June 23-30, 2007, Prague}, 2007.

\bibitem{davis2012game}
Morton~D Davis.
\newblock {\em Game Theory: a Nontechnical Introduction}.
\newblock Courier Corporation, 2012.

\bibitem{dawkins2017selfish}
Richard Dawkins and Nicola Davis.
\newblock {\em The Selfish Gene}.
\newblock Macat Library, 2017.

\bibitem{dayan1993feudal}
Peter Dayan and Geoffrey~E Hinton.
\newblock Feudal reinforcement learning.
\newblock In {\em Advances in Neural Information Processing Systems}, pages
  271--278, 1993.

\bibitem{de2005tutorial}
Pieter-Tjerk De~Boer, Dirk~P Kroese, Shie Mannor, and Reuven~Y Rubinstein.
\newblock A tutorial on the cross-entropy method.
\newblock {\em Annals of Operations Research}, 134(1):19--67, 2005.

\bibitem{de2002ant}
Luis~M De~Campos, Juan~M Fernandez-Luna, Jos{\'e}~A G{\'a}mez, and Jos{\'e}~M
  Puerta.
\newblock Ant colony optimization for learning {Bayesian} networks.
\newblock {\em International Journal of Approximate Reasoning}, 31(3):291--311,
  2002.

\bibitem{de2018challenge}
Dave De~Jonge, Tim Baarslag, Reyhan Aydo{\u{g}}an, Catholijn Jonker, Katsuhide
  Fujita, and Takayuki Ito.
\newblock The challenge of negotiation in the game of diplomacy.
\newblock In {\em International Conference on Agreement Technologies}, pages
  100--114. Springer, 2018.

\bibitem{vries2021visualizing}
Joery~A. de~Vries, Ken~S. Voskuil, Thomas~M. Moerland, and Aske Plaat.
\newblock Visualizing {MuZero} models.
\newblock {\em arXiv preprint arXiv:2102.12924}, 2021.

\bibitem{dechter1986learning}
Rina Dechter.
\newblock {\em Learning while searching in constraint-satisfaction problems}.
\newblock AAAI, 1986.

\bibitem{deisenroth2011pilco}
Marc Deisenroth and Carl~E Rasmussen.
\newblock {PILCO:} a model-based and data-efficient approach to policy search.
\newblock In {\em Proceedings of the 28th International Conference on Machine
  Learning (ICML-11)}, pages 465--472, 2011.

\bibitem{deisenroth2013gaussian}
Marc~Peter Deisenroth, Dieter Fox, and Carl~Edward Rasmussen.
\newblock Gaussian processes for data-efficient learning in robotics and
  control.
\newblock {\em IEEE Transactions on Pattern Analysis and Machine Intelligence},
  37(2):408--423, 2013.

\bibitem{deisenroth2013survey}
Marc~Peter Deisenroth, Gerhard Neumann, and Jan Peters.
\newblock A survey on policy search for robotics.
\newblock In {\em Foundations and Trends in Robotics 2}, pages 1--142. Now
  publishers, 2013.

\bibitem{deng2009imagenet}
Jia Deng, Wei Dong, Richard Socher, Li-Jia Li, Kai Li, and Li~Fei-Fei.
\newblock Imagenet: A large-scale hierarchical image database.
\newblock In {\em 2009 IEEE Conference on Computer Vision and Pattern
  Recognition}, pages 248--255. Ieee, 2009.

\bibitem{deshpande2019}
Mohit Deshpande.
\newblock Deep {RL} policy methods
  \url{https://mohitd.github.io/2019/01/20/deep-rl-policy-methods.html}.

\bibitem{devlin2018bert}
Jacob Devlin, Ming-Wei Chang, Kenton Lee, and Kristina Toutanova.
\newblock {BERT:} pre-training of deep bidirectional transformers for language
  understanding.
\newblock In {\em Proceedings of the 2019 Conference of the North American
  Chapter of the Association for Computational Linguistics: Human Language
  Technologies, {NAACL-HLT} 2019, Minneapolis}, 2018.

\bibitem{baselines}
Prafulla Dhariwal, Christopher Hesse, Oleg Klimov, Alex Nichol, Matthias
  Plappert, Alec Radford, John Schulman, Szymon Sidor, Yuhuai Wu, and Peter
  Zhokhov.
\newblock {OpenAI} baselines.
\newblock \url{https://github.com/openai/baselines}, 2017.

\bibitem{dietterich1998maxq}
Thomas~G Dietterich.
\newblock The {MAXQ} method for hierarchical reinforcement learning.
\newblock In {\em International Conference on Machine Learning}, volume~98,
  pages 118--126, 1998.

\bibitem{dietterich2000hierarchical}
Thomas~G Dietterich.
\newblock Hierarchical reinforcement learning with the {MAXQ} value function
  decomposition.
\newblock {\em Journal of Artificial Intelligence Research}, 13:227--303, 2000.

\bibitem{do2005transfer}
Chuong~B Do and Andrew~Y Ng.
\newblock Transfer learning for text classification.
\newblock {\em Advances in Neural Information Processing Systems}, 18:299--306,
  2005.

\bibitem{doan2019line}
Thang Doan, Joao Monteiro, Isabela Albuquerque, Bogdan Mazoure, Audrey Durand,
  Joelle Pineau, and R~Devon Hjelm.
\newblock On-line adaptative curriculum learning for {GANs}.
\newblock In {\em Proceedings of the AAAI Conference on Artificial
  Intelligence}, volume~33, pages 3470--3477, 2019.

\bibitem{doerr2018probabilistic}
Andreas Doerr, Christian Daniel, Martin Schiegg, Duy Nguyen-Tuong, Stefan
  Schaal, Marc Toussaint, and Sebastian Trimpe.
\newblock Probabilistic recurrent state-space models.
\newblock {\em arXiv preprint arXiv:1801.10395}, 2018.

\bibitem{donahue2014decaf}
Jeff Donahue, Yangqing Jia, Oriol Vinyals, Judy Hoffman, Ning Zhang, Eric
  Tzeng, and Trevor Darrell.
\newblock Decaf: A deep convolutional activation feature for generic visual
  recognition.
\newblock In {\em International Conference on Machine Learning}, pages
  647--655. PMLR, 2014.

\bibitem{dong2020deep}
Hao Dong, Zihan Ding, and Shanghang Zhang.
\newblock {\em Deep Reinforcement Learning}.
\newblock Springer, 2020.

\bibitem{dongarra1988extended}
Jack~J Dongarra, Jeremy Du~Croz, Sven Hammarling, and Richard~J Hanson.
\newblock An extended set of {FORTRAN} basic linear algebra subprograms.
\newblock {\em ACM Transactions on Mathematical Software}, 14(1):1--17, 1988.

\bibitem{donninger1993null}
Christian Donninger.
\newblock Null move and deep search.
\newblock {\em ICGA Journal}, 16(3):137--143, 1993.

\bibitem{dor1999sokoban}
Dorit Dor and Uri Zwick.
\newblock Sokoban and other motion planning problems.
\newblock {\em Computational Geometry}, 13(4):215--228, 1999.

\bibitem{doran2017does}
Derek Doran, Sarah Schulz, and Tarek~R Besold.
\newblock What does explainable {AI} really mean? {A} new conceptualization of
  perspectives.
\newblock {\em arXiv preprint arXiv:1710.00794}, 2017.

\bibitem{dorigo1992optimization}
Marco Dorigo.
\newblock Optimization, learning and natural algorithms.
\newblock {\em PhD Thesis, Politecnico di Milano}, 1992.

\bibitem{dorigo2007swarm}
Marco Dorigo and Mauro Birattari.
\newblock Swarm intelligence.
\newblock {\em Scholarpedia}, 2(9):1462, 2007.

\bibitem{dorigo2006ant}
Marco Dorigo, Mauro Birattari, and Thomas Stutzle.
\newblock Ant colony optimization.
\newblock {\em IEEE Computational Intelligence Magazine}, 1(4):28--39, 2006.

\bibitem{dorigo1997ant}
Marco Dorigo and Luca~Maria Gambardella.
\newblock Ant colony system: a cooperative learning approach to the traveling
  salesman problem.
\newblock {\em IEEE Transactions on Evolutionary Computation}, 1(1):53--66,
  1997.

\bibitem{draper1998applied}
Norman~R Draper and Harry Smith.
\newblock {\em Applied Regression Analysis}, volume 326.
\newblock John Wiley \& Sons, 1998.

\bibitem{du2020dual}
Yuntao Du, Zhiwen Tan, Qian Chen, Xiaowen Zhang, Yirong Yao, and Chongjun Wang.
\newblock Dual adversarial domain adaptation.
\newblock {\em arXiv preprint arXiv:2001.00153}, 2020.

\bibitem{duan2016benchmarking}
Yan Duan, Xi~Chen, Rein Houthooft, John Schulman, and Pieter Abbeel.
\newblock Benchmarking deep reinforcement learning for continuous control.
\newblock In {\em International Conference on Machine Learning}, pages
  1329--1338, 2016.

\bibitem{duan2016rl}
Yan Duan, John Schulman, Xi~Chen, Peter~L Bartlett, Ilya Sutskever, and Pieter
  Abbeel.
\newblock {RL}${}^{2}$: Fast reinforcement learning via slow reinforcement
  learning.
\newblock {\em arXiv preprint arXiv:1611.02779}, 2016.

\bibitem{durugkar2016deep}
Ishan~P Durugkar, Clemens Rosenbaum, Stefan Dernbach, and Sridhar Mahadevan.
\newblock Deep reinforcement learning with macro-actions.
\newblock {\em arXiv preprint arXiv:1606.04615}, 2016.

\bibitem{muzero-general}
Werner Duvaud and Aur{\`e}le Hainaut.
\newblock {MuZero} general: Open reimplementation of muzero.
  \url{https://github.com/werner-duvaud/muzero-general}, 2019.

\bibitem{dwiel2019hierarchical}
Zach Dwiel, Madhavun Candadai, Mariano Phielipp, and Arjun~K Bansal.
\newblock Hierarchical policy learning is sensitive to goal space design.
\newblock {\em arXiv preprint arXiv:1905.01537}, 2019.

\bibitem{eberhart2001swarm}
Russell~C Eberhart, Yuhui Shi, and James Kennedy.
\newblock {\em Swarm Intelligence}.
\newblock Elsevier, 2001.

\bibitem{ebert2018visual}
Frederik Ebert, Chelsea Finn, Sudeep Dasari, Annie Xie, Alex Lee, and Sergey
  Levine.
\newblock Visual foresight: Model-based deep reinforcement learning for
  vision-based robotic control.
\newblock {\em arXiv preprint arXiv:1812.00568}, 2018.

\bibitem{eccles2019learning}
Tom Eccles, Edward Hughes, J{\'a}nos Kram{\'a}r, Steven Wheelwright, and Joel~Z
  Leibo.
\newblock Learning reciprocity in complex sequential social dilemmas.
\newblock {\em arXiv preprint arXiv:1903.08082}, 2019.

\bibitem{ecofet2018}
Adrien~Lucas Ecofet.
\newblock An intuitive explanation of policy gradient
  \url{https://towardsdatascience.com/an-intuitive-explanation-of-policy-gradient-part-1-reinforce-aa4392cbfd3c}.

\bibitem{ecoffet2019go}
Adrien Ecoffet, Joost Huizinga, Joel Lehman, Kenneth~O Stanley, and Jeff Clune.
\newblock Go-explore: a new approach for hard-exploration problems.
\newblock {\em arXiv preprint arXiv:1901.10995}, 2019.

\bibitem{ecoffet2021first}
Adrien Ecoffet, Joost Huizinga, Joel Lehman, Kenneth~O Stanley, and Jeff Clune.
\newblock First return, then explore.
\newblock {\em Nature}, 590(7847):580--586, 2021.

\bibitem{edwards2016towards}
Harrison Edwards and Amos Storkey.
\newblock Towards a neural statistician.
\newblock In {\em International Conference on Learning Representations}, 2017.

\bibitem{eiben2015evolutionary}
Agoston~E Eiben and Jim~E Smith.
\newblock What is an evolutionary algorithm?
\newblock In {\em Introduction to Evolutionary Computing}, pages 25--48.
  Springer, 2015.

\bibitem{elman1993learning}
Jeffrey~L Elman.
\newblock Learning and development in neural networks: The importance of
  starting small.
\newblock {\em Cognition}, 48(1):71--99, 1993.

\bibitem{elo1978rating}
Arpad~E Elo.
\newblock {\em The Rating of Chessplayers, Past and Present}.
\newblock Arco Pub., 1978.

\bibitem{enzenberger2010fuego}
Markus Enzenberger, Martin Muller, Broderick Arneson, and Richard Segal.
\newblock Fuego---an open-source framework for board games and {Go} engine
  based on {Monte Carlo} tree search.
\newblock {\em IEEE Transactions on Computational Intelligence and AI in
  Games}, 2(4):259--270, 2010.

\bibitem{erez2015simulation}
Tom Erez, Yuval Tassa, and Emanuel Todorov.
\newblock Simulation tools for model-based robotics: Comparison of {Bullet},
  {Havok}, {MuJoCo}, {Ode} and {Physx}.
\newblock In {\em 2015 IEEE International Conference on Robotics and Automation
  (ICRA)}, pages 4397--4404. IEEE, 2015.

\bibitem{ernst2005tree}
Damien Ernst, Pierre Geurts, and Louis Wehenkel.
\newblock Tree-based batch mode reinforcement learning.
\newblock {\em Journal of Machine Learning Research}, 6:503--556, April 2005.

\bibitem{espeholt2018impala}
Lasse Espeholt, Hubert Soyer, Remi Munos, Karen Simonyan, Vlad Mnih, Tom Ward,
  Yotam Doron, Vlad Firoiu, Tim Harley, Iain Dunning, Shane Legg, and Koray
  Kavukcuoglu.
\newblock {IMPALA}: Scalable distributed deep-{RL} with importance weighted
  actor-learner architectures.
\newblock In {\em International Conference on Machine Learning}, pages
  1407--1416. PMLR, 2018.

\bibitem{even2003learning}
Eyal Even-Dar, Yishay Mansour, and Peter Bartlett.
\newblock Learning rates for {Q-learning}.
\newblock {\em Journal of machine learning Research}, 5(1), 2003.

\bibitem{everett2018learning}
Richard Everett and Stephen Roberts.
\newblock Learning against non-stationary agents with opponent modelling and
  deep reinforcement learning.
\newblock In {\em 2018 AAAI Spring Symposium Series}, 2018.

\bibitem{evgeniou2004regularized}
Theodoros Evgeniou and Massimiliano Pontil.
\newblock Regularized multi-task learning.
\newblock In {\em Proceedings of the tenth ACM SIGKDD International Conference
  on Knowledge Discovery and Data Mining}, pages 109--117. ACM, 2004.

\bibitem{fan2018surreal}
Linxi Fan, Yuke Zhu, Jiren Zhu, Zihua Liu, Orien Zeng, Anchit Gupta, Joan
  Creus-Costa, Silvio Savarese, and Li~Fei-Fei.
\newblock Surreal: Open-source reinforcement learning framework and robot
  manipulation benchmark.
\newblock In {\em Conference on Robot Learning}, pages 767--782, 2018.

\bibitem{farebrother2018generalization}
Jesse Farebrother, Marlos~C Machado, and Michael Bowling.
\newblock Generalization and regularization in {DQN}.
\newblock {\em arXiv preprint arXiv:1810.00123}, 2018.

\bibitem{farquhar2018treeqn}
Gregory Farquhar, Tim Rockt{\"a}schel, Maximilian Igl, and SA~Whiteson.
\newblock {TreeQN and ATreeC}: Differentiable tree planning for deep
  reinforcement learning.
\newblock In {\em International Conference on Learning Representations}, 2018.

\bibitem{fei2009imagenet}
Li~Fei-Fei, Jia Deng, and Kai Li.
\newblock Imagenet: Constructing a large-scale image database.
\newblock {\em Journal of Vision}, 9(8):1037--1037, 2009.

\bibitem{feinberg2018model}
Vladimir Feinberg, Alvin Wan, Ion Stoica, Michael~I Jordan, Joseph~E Gonzalez,
  and Sergey Levine.
\newblock Model-based value estimation for efficient model-free reinforcement
  learning.
\newblock {\em arXiv preprint arXiv:1803.00101}, 2018.

\bibitem{feng2020solving}
Dieqiao Feng, Carla~P Gomes, and Bart Selman.
\newblock Solving hard {AI} planning instances using curriculum-driven deep
  reinforcement learning.
\newblock {\em arXiv preprint arXiv:2006.02689}, 2020.

\bibitem{fernandez2007application}
Santiago Fern{\'a}ndez, Alex Graves, and J{\"u}rgen Schmidhuber.
\newblock An application of recurrent neural networks to discriminative keyword
  spotting.
\newblock In {\em International Conference on Artificial Neural Networks},
  pages 220--229. Springer, 2007.

\bibitem{fikes1972learning}
Richard~E Fikes, Peter~E Hart, and Nils~J Nilsson.
\newblock Learning and executing generalized robot plans.
\newblock {\em Artificial Intelligence}, 3:251--288, 1972.

\bibitem{fikes1971strips}
Richard~E Fikes and Nils~J Nilsson.
\newblock {STRIPS: A new approach to the application of theorem proving to
  problem solving}.
\newblock {\em Artificial Intelligence}, 2(3-4):189--208, 1971.

\bibitem{finn2017model}
Chelsea Finn, Pieter Abbeel, and Sergey Levine.
\newblock {Model-Agnostic Meta-Learning} for fast adaptation of deep networks.
\newblock In {\em International Conference on Machine Learning}, pages
  1126--1135. PMLR, 2017.

\bibitem{finn2017deep}
Chelsea Finn and Sergey Levine.
\newblock Deep visual foresight for planning robot motion.
\newblock In {\em 2017 IEEE International Conference on Robotics and Automation
  (ICRA)}, pages 2786--2793. IEEE, 2017.

\bibitem{finn2019online}
Chelsea Finn, Aravind Rajeswaran, Sham Kakade, and Sergey Levine.
\newblock Online meta-learning.
\newblock In {\em International Conference on Machine Learning}, pages
  1920--1930. PMLR, 2019.

\bibitem{finn2018probabilistic}
Chelsea Finn, Kelvin Xu, and Sergey Levine.
\newblock Probabilistic {Model-Agnostic Meta-Learning}.
\newblock In {\em Advances in Neural Information Processing Systems}, pages
  9516--9527, 2018.

\bibitem{flet2019promise}
Yannis Flet-Berliac.
\newblock The promise of hierarchical reinforcement learning.
\newblock
  \url{https://thegradient.pub/the-promise-of-hierarchical-reinforcement-learning/},
  March 2019.

\bibitem{florensa2018automatic}
Carlos Florensa, David Held, Xinyang Geng, and Pieter Abbeel.
\newblock Automatic goal generation for reinforcement learning agents.
\newblock In {\em International Conference on Machine Learning}, pages
  1515--1528. PMLR, 2018.

\bibitem{flynn1972some}
Michael~J Flynn.
\newblock Some computer organizations and their effectiveness.
\newblock {\em IEEE Transactions on Computers}, 100(9):948--960, 1972.

\bibitem{foerster2018counterfactual}
Jakob Foerster, Gregory Farquhar, Triantafyllos Afouras, Nantas Nardelli, and
  Shimon Whiteson.
\newblock Counterfactual multi-agent policy gradients.
\newblock In {\em Proceedings of the AAAI Conference on Artificial
  Intelligence}, volume~32, 2018.

\bibitem{foerster2017learning}
Jakob~N Foerster, Richard~Y Chen, Maruan Al-Shedivat, Shimon Whiteson, Pieter
  Abbeel, and Igor Mordatch.
\newblock Learning with opponent-learning awareness.
\newblock {\em arXiv preprint arXiv:1709.04326}, 2017.

\bibitem{fogel1994introduction}
David~B Fogel.
\newblock An introduction to simulated evolutionary optimization.
\newblock {\em IEEE Transactions on Neural Networks}, 5(1):3--14, 1994.

\bibitem{fogel2005further}
David~B Fogel, Timothy~J Hays, Sarah~L Hahn, and James Quon.
\newblock Further evolution of a self-learning chess program.
\newblock In {\em Computational Intelligence in Games}, 2005.

\bibitem{fortunato2017noisy}
Meire Fortunato, Mohammad~Gheshlaghi Azar, Bilal Piot, Jacob Menick, Ian
  Osband, Alex Graves, Vlad Mnih, Remi Munos, Demis Hassabis, Olivier Pietquin,
  Charles Blundell, and Shane Legg.
\newblock Noisy networks for exploration.
\newblock In {\em International Conference on Learning Representations}, 2018.

\bibitem{franccois2018introduction}
Vincent Fran{\c{c}}ois-Lavet, Peter Henderson, Riashat Islam, Marc~G Bellemare,
  and Joelle Pineau.
\newblock An introduction to deep reinforcement learning.
\newblock {\em Foundations and Trends in Machine Learning}, 11(3-4):219--354,
  2018.

\bibitem{frans2017meta}
Kevin Frans, Jonathan Ho, Xi~Chen, Pieter Abbeel, and John Schulman.
\newblock Meta learning shared hierarchies.
\newblock In {\em International Conference on Learning Representations}, 2018.

\bibitem{frosst2017distilling}
Nicholas Frosst and Geoffrey Hinton.
\newblock Distilling a neural network into a soft decision tree.
\newblock In {\em Proceedings of the First International Workshop on
  Comprehensibility and Explanation in {AI} and {ML}}, 2017.

\bibitem{gallese1998mirror}
Vittorio Gallese and Alvin Goldman.
\newblock Mirror neurons and the simulation theory of mind-reading.
\newblock {\em Trends in Cognitive Sciences}, 2(12):493--501, 1998.

\bibitem{ganzfried2011game}
Sam Ganzfried and Tuomas Sandholm.
\newblock Game theory-based opponent modeling in large imperfect-information
  games.
\newblock In {\em The 10th International Conference on Autonomous Agents and
  Multiagent Systems}, volume~2, pages 533--540, 2011.

\bibitem{ganzfried2015endgame}
Sam Ganzfried and Tuomas Sandholm.
\newblock Endgame solving in large imperfect-information games.
\newblock In {\em Proceedings of the 2015 International Conference on
  Autonomous Agents and Multiagent Systems}, pages 37--45, 2015.

\bibitem{garage}
The garage contributors.
\newblock Garage: A toolkit for reproducible reinforcement learning research.
\newblock \url{https://github.com/rlworkgroup/garage}, 2019.

\bibitem{garcia1989model}
Carlos~E Garcia, David~M Prett, and Manfred Morari.
\newblock Model predictive control: Theory and practice---a survey.
\newblock {\em Automatica}, 25(3):335--348, 1989.

\bibitem{garcia2017few}
Victor Garcia and Joan Bruna.
\newblock Few-shot learning with graph neural networks.
\newblock In {\em International Conference on Learning Representations}, 2017.

\bibitem{garnelo2018conditional}
Marta Garnelo, Dan Rosenbaum, Christopher Maddison, Tiago Ramalho, David
  Saxton, Murray Shanahan, Yee~Whye Teh, Danilo Rezende, and SM~Ali Eslami.
\newblock Conditional neural processes.
\newblock In {\em International Conference on Machine Learning}, pages
  1704--1713. PMLR, 2018.

\bibitem{gasparetto2015path}
Alessandro Gasparetto, Paolo Boscariol, Albano Lanzutti, and Renato Vidoni.
\newblock Path planning and trajectory planning algorithms: A general overview.
\newblock In {\em Motion and Operation Planning of Robotic Systems}, pages
  3--27. Springer, 2015.

\bibitem{gelfond1998action}
Michael Gelfond and Vladimir Lifschitz.
\newblock Action languages.
\newblock {\em Electronic Transactions on Artificial Intelligence},
  2(3--4):193--210, 1998.

\bibitem{gelly2012grand}
Sylvain Gelly, Levente Kocsis, Marc Schoenauer, Michele Sebag, David Silver,
  Csaba Szepesv{\'a}ri, and Olivier Teytaud.
\newblock The grand challenge of computer {Go}: {Monte Carlo} tree search and
  extensions.
\newblock {\em Communications of the ACM}, 55(3):106--113, 2012.

\bibitem{gelly2008achieving}
Sylvain Gelly and David Silver.
\newblock Achieving master level play in $9 \times 9$ computer {Go}.
\newblock In {\em AAAI}, volume~8, pages 1537--1540, 2008.

\bibitem{gelly2006modification}
Sylvain Gelly, Yizao Wang, and Olivier Teytaud.
\newblock Modification of {UCT} with patterns in {Monte-Carlo} {Go}.
\newblock Technical Report RR-6062, INRIA, 2006.

\bibitem{geron2017hands}
Aur{\'e}lien G{\'e}ron.
\newblock {\em Hands-on machine learning with Scikit-Learn and TensorFlow:
  concepts, tools, and techniques to build intelligent systems}.
\newblock {O'Reilly Media, Inc.}, 2019.

\bibitem{gers1999learning}
Felix~A Gers, J{\"u}rgen Schmidhuber, and Fred Cummins.
\newblock Learning to forget: Continual prediction with {LSTM}.
\newblock In {\em Ninth International Conference on Artificial Neural Networks
  ICANN 99}. IET, 1999.

\bibitem{ghallab2004automated}
Malik Ghallab, Dana Nau, and Paolo Traverso.
\newblock {\em Automated Planning: theory and practice}.
\newblock Elsevier, 2004.

\bibitem{ghavamzadeh2006hierarchical}
Mohammad Ghavamzadeh, Sridhar Mahadevan, and Rajbala Makar.
\newblock Hierarchical multi-agent reinforcement learning.
\newblock {\em Autonomous Agents and Multi-Agent Systems}, 13(2):197--229,
  2006.

\bibitem{ghosh2021generalization}
Dibya Ghosh, Jad Rahme, Aviral Kumar, Amy Zhang, Ryan~P Adams, and Sergey
  Levine.
\newblock Why generalization in {RL} is difficult: Epistemic {POMDPs} and
  implicit partial observability.
\newblock {\em Advances in Neural Information Processing Systems}, 34, 2021.

\bibitem{gigerenzer1996reasoning}
Gerd Gigerenzer and Daniel~G Goldstein.
\newblock Reasoning the fast and frugal way: models of bounded rationality.
\newblock {\em Psychological review}, 103(4):650, 1996.

\bibitem{gilovich2002heuristics}
Thomas Gilovich, Dale Griffin, and Daniel Kahneman.
\newblock {\em Heuristics and Biases: The Psychology of Intuitive Judgment}.
\newblock Cambridge university press, 2002.

\bibitem{gilpin2006competitive}
Andrew Gilpin and Tuomas Sandholm.
\newblock A competitive {Texas Hold'em} poker player via automated abstraction
  and real-time equilibrium computation.
\newblock In {\em Proceedings of the National Conference on Artificial
  Intelligence}, volume~21, page 1007, 2006.

\bibitem{girshick2014rich}
Ross Girshick, Jeff Donahue, Trevor Darrell, and Jitendra Malik.
\newblock Rich feature hierarchies for accurate object detection and semantic
  segmentation.
\newblock In {\em Proceedings of the IEEE Conference on Computer Vision and
  Pattern Recognition}, pages 580--587, 2014.

\bibitem{gittins1979bandit}
John~C Gittins.
\newblock Bandit processes and dynamic allocation indices.
\newblock {\em Journal of the Royal Statistical Society: Series B
  (Methodological)}, 41(2):148--164, 1979.

\bibitem{goodfellow2016deep}
Ian Goodfellow, Yoshua Bengio, and Aaron Courville.
\newblock {\em Deep Learning}.
\newblock MIT Press, Cambridge, 2016.

\bibitem{goodfellow2014generative}
Ian Goodfellow, Jean Pouget-Abadie, Mehdi Mirza, Bing Xu, David Warde-Farley,
  Sherjil Ozair, Aaron Courville, and Yoshua Bengio.
\newblock Generative adversarial nets.
\newblock In {\em Advances in Neural Information Processing Systems}, pages
  2672--2680, 2014.

\bibitem{gordon1995stable}
Geoffrey~J Gordon.
\newblock Stable function approximation in dynamic programming.
\newblock In {\em Machine Learning Proceedings 1995}, pages 261--268. Elsevier,
  1995.

\bibitem{gordon1999approximate}
Geoffrey~J Gordon.
\newblock {\em Approximate solutions to Markov decision processes}.
\newblock Carnegie Mellon University, 1999.

\bibitem{graf2015adaptive}
Tobias Graf and Marco Platzner.
\newblock Adaptive playouts in {Monte-Carlo} tree search with policy-gradient
  reinforcement learning.
\newblock In {\em Advances in Computer Games}, pages 1--11. Springer, 2015.

\bibitem{grant2018recasting}
Erin Grant, Chelsea Finn, Sergey Levine, Trevor Darrell, and Thomas Griffiths.
\newblock Recasting gradient-based meta-learning as hierarchical bayes.
\newblock In {\em International Conference on Learning Representations}, 2018.

\bibitem{graves2005bidirectional}
Alex Graves, Santiago Fern{\'a}ndez, and J{\"u}rgen Schmidhuber.
\newblock Bidirectional {LSTM} networks for improved phoneme classification and
  recognition.
\newblock In {\em International Conference on Artificial Neural Networks},
  pages 799--804. Springer, 2005.

\bibitem{graves2013speech}
Alex Graves, Abdel-rahman Mohamed, and Geoffrey Hinton.
\newblock Speech recognition with deep recurrent neural networks.
\newblock In {\em 2013 IEEE International Conference on Acoustics, Speech and
  Signal Processing}, pages 6645--6649. IEEE, 2013.

\bibitem{greff2017lstm}
Klaus Greff, Rupesh~K Srivastava, Jan Koutn{\'\i}k, Bas~R Steunebrink, and
  J{\"u}rgen Schmidhuber.
\newblock {LSTM: A} search space odyssey.
\newblock {\em IEEE Transactions on Neural Networks and Learning Systems},
  28(10):2222--2232, 2017.

\bibitem{grill2020monte}
Jean-Bastien Grill, Florent Altch{\'e}, Yunhao Tang, Thomas Hubert, Michal
  Valko, Ioannis Antonoglou, and R{\'e}mi Munos.
\newblock Monte-carlo tree search as regularized policy optimization.
\newblock In {\em International Conference on Machine Learning}, pages
  3769--3778. PMLR, 2020.

\bibitem{grimm2020value}
Christopher Grimm, Andr{\'e} Barreto, Satinder Singh, and David Silver.
\newblock The value equivalence principle for model-based reinforcement
  learning.
\newblock In {\em Advances in Neural Information Processing Systems}, 2020.

\bibitem{grinsztajn2021there}
Nathan Grinsztajn, Johan Ferret, Olivier Pietquin, Philippe Preux, and Matthieu
  Geist.
\newblock There is no turning back: A self-supervised approach for
  reversibility-aware reinforcement learning.
\newblock {\em arXiv preprint arXiv:2106.04480}, 2021.

\bibitem{gronauer2021multi}
Sven Gronauer and Klaus Diepold.
\newblock Multi-agent deep reinforcement learning: a survey.
\newblock {\em Artificial Intelligence Review}, pages 1--49, 2021.

\bibitem{grondman2012survey}
Ivo Grondman, Lucian Busoniu, Gabriel~AD Lopes, and Robert Babuska.
\newblock A survey of actor-critic reinforcement learning: Standard and natural
  policy gradients.
\newblock {\em IEEE Transactions on Systems, Man, and Cybernetics, Part C
  (Applications and Reviews)}, 42(6):1291--1307, 2012.

\bibitem{grunwald2007minimum}
Peter~D Gr{\"u}nwald.
\newblock {\em The minimum description length principle}.
\newblock MIT press, 2007.

\bibitem{gruslys2017reactor}
Audrunas Gruslys, Will Dabney, Mohammad~Gheshlaghi Azar, Bilal Piot, Marc
  Bellemare, and Remi Munos.
\newblock The reactor: A fast and sample-efficient actor-critic agent for
  reinforcement learning.
\newblock In {\em International Conference on Learning Representations}, 2018.

\bibitem{gu2016continuous}
Shixiang Gu, Timothy Lillicrap, Ilya Sutskever, and Sergey Levine.
\newblock Continuous deep {Q}-learning with model-based acceleration.
\newblock In {\em International Conference on Machine Learning}, pages
  2829--2838, 2016.

\bibitem{guestrin2001multiagent}
Carlos Guestrin, Daphne Koller, and Ronald Parr.
\newblock Multiagent planning with factored {MDPs}.
\newblock In {\em Advances in Neural Information Processing Systems}, volume~1,
  pages 1523--1530, 2001.

\bibitem{guez2019investigation}
Arthur Guez, Mehdi Mirza, Karol Gregor, Rishabh Kabra, S{\'{e}}bastien
  Racani{\`{e}}re, Theophane Weber, David Raposo, Adam Santoro, Laurent Orseau,
  Tom Eccles, Greg Wayne, David Silver, and Timothy~P. Lillicrap.
\newblock An investigation of model-free planning.
\newblock In {\em International Conference on Machine Learning}, pages
  2464--2473, 2019.

\bibitem{gulcehre2020rl}
Caglar Gulcehre, Ziyu Wang, Alexander Novikov, Tom~Le Paine, Sergio~G{\'o}mez
  Colmenarejo, Konrad Zolna, Rishabh Agarwal, Josh Merel, Daniel Mankowitz,
  Cosmin Paduraru, et~al.
\newblock {RL} unplugged: Benchmarks for offline reinforcement learning.
\newblock {\em arXiv preprint arXiv:2006.13888}, 2020.

\bibitem{gunning2017explainable}
David Gunning.
\newblock Explainable artificial intelligence {(XAI)}.
\newblock {\em Defense Advanced Research Projects Agency (DARPA)}, 2, 2017.

\bibitem{guo2016simple}
Xifeng Guo, Wei Chen, and Jianping Yin.
\newblock A simple approach for unsupervised domain adaptation.
\newblock In {\em 2016 23rd International Conference on Pattern Recognition
  (ICPR)}, pages 1566--1570. IEEE, 2016.

\bibitem{gupta2018meta}
Abhishek Gupta, Russell Mendonca, YuXuan Liu, Pieter Abbeel, and Sergey Levine.
\newblock Meta-reinforcement learning of structured exploration strategies.
\newblock In {\em Advances in Neural Information Processing Systems}, pages
  5307--5316, 2018.

\bibitem{ha2018recurrent}
David Ha and J{\"u}rgen Schmidhuber.
\newblock Recurrent world models facilitate policy evolution.
\newblock In {\em Advances in Neural Information Processing Systems}, pages
  2450--2462, 2018.

\bibitem{ha2018world}
David Ha and J{\"u}rgen Schmidhuber.
\newblock World models.
\newblock {\em arXiv preprint arXiv:1803.10122}, 2018.

\bibitem{haarnoja2017reinforcement}
Tuomas Haarnoja, Haoran Tang, Pieter Abbeel, and Sergey Levine.
\newblock Reinforcement learning with deep energy-based policies.
\newblock In {\em International Conference on Machine Learning}, pages
  1352--1361. PMLR, 2017.

\bibitem{haarnoja2018soft}
Tuomas Haarnoja, Aurick Zhou, Pieter Abbeel, and Sergey Levine.
\newblock Soft actor-critic: Off-policy maximum entropy deep reinforcement
  learning with a stochastic actor.
\newblock In {\em International Conference on Machine Learning}, pages
  1861--1870. PMLR, 2018.

\bibitem{haarnoja2019soft}
Tuomas Haarnoja, Aurick Zhou, Kristian Hartikainen, George Tucker, Sehoon Ha,
  Jie Tan, Vikash Kumar, Henry Zhu, Abhishek Gupta, Pieter Abbeel, and Sergey
  Levine.
\newblock Soft actor-critic algorithms and applications.
\newblock {\em arXiv preprint arXiv:1812.05905}, 2018.

\bibitem{hafner2019dream}
Danijar Hafner, Timothy Lillicrap, Jimmy Ba, and Mohammad Norouzi.
\newblock Dream to control: Learning behaviors by latent imagination.
\newblock In {\em International Conference on Learning Representations}, 2020.

\bibitem{hafner2018learning}
Danijar Hafner, Timothy Lillicrap, Ian Fischer, Ruben Villegas, David Ha,
  Honglak Lee, and James Davidson.
\newblock Learning latent dynamics for planning from pixels.
\newblock In {\em International Conference on Machine Learning}, pages
  2555--2565, 2019.

\bibitem{hafner2020mastering}
Danijar Hafner, Timothy Lillicrap, Mohammad Norouzi, and Jimmy Ba.
\newblock Mastering atari with discrete world models.
\newblock In {\em International Conference on Learning Representations}, 2021.

\bibitem{hafner2011reinforcement}
Roland Hafner and Martin Riedmiller.
\newblock Reinforcement learning in feedback control.
\newblock {\em Machine Learning}, 84(1-2):137--169, 2011.

\bibitem{han2019multi}
Dongge Han, Wendelin Boehmer, Michael Wooldridge, and Alex Rogers.
\newblock Multi-agent hierarchical reinforcement learning with dynamic
  termination.
\newblock In {\em Pacific Rim International Conference on Artificial
  Intelligence}, pages 80--92. Springer, 2019.

\bibitem{han2021multiagent}
Dongge Han, Chris~Xiaoxuan Lu, Tomasz Michalak, and Michael Wooldridge.
\newblock Multiagent model-based credit assignment for continuous control,
  2021.

\bibitem{hasselt2010double}
Hado~V Hasselt.
\newblock {Double Q-learning}.
\newblock In {\em Advances in Neural Information Processing Systems}, pages
  2613--2621, 2010.

\bibitem{hausknecht2016cooperation}
Matthew~John Hausknecht.
\newblock {\em Cooperation and Communication in Multiagent Deep Reinforcement
  Learning}.
\newblock PhD thesis, University of Texas at Austin, 2016.

\bibitem{hauskrecht2013hierarchical}
Milos Hauskrecht, Nicolas Meuleau, Leslie~Pack Kaelbling, Thomas~L Dean, and
  Craig Boutilier.
\newblock Hierarchical solution of {Markov} decision processes using
  macro-actions.
\newblock In {\em {UAI} '98: Proceedings of the Fourteenth Conference on
  Uncertainty in Artificial Intelligence, University of Wisconsin Business
  School, Madison, Wisconsin}, 1998.

\bibitem{hayes2021practical}
Conor~F. Hayes, Roxana Radulescu, Eugenio Bargiacchi, Johan
  K{\"{a}}llstr{\"{o}}m, Matthew Macfarlane, Mathieu Reymond, Timothy
  Verstraeten, Luisa~M. Zintgraf, Richard Dazeley, Fredrik Heintz, Enda Howley,
  Athirai~A. Irissappane, Patrick Mannion, Ann Now{\'{e}}, Gabriel
  de~Oliveira~Ramos, Marcello Restelli, Peter Vamplew, and Diederik~M. Roijers.
\newblock A practical guide to multi-objective reinforcement learning and
  planning.
\newblock {\em arXiv preprint arXiv:2103.09568}, 2021.

\bibitem{haykin1994neural}
Simon Haykin.
\newblock {\em Neural Networks: a Comprehensive Foundation}.
\newblock Prentice Hall, 1994.

\bibitem{hayward2019hex}
Ryan~B Hayward and Bjarne Toft.
\newblock {\em Hex: The Full Story}.
\newblock CRC Press, 2019.

\bibitem{he2016opponent}
He~He, Jordan Boyd-Graber, Kevin Kwok, and Hal Daum{\'e}~III.
\newblock Opponent modeling in deep reinforcement learning.
\newblock In {\em International Conference on Machine Learning}, pages
  1804--1813. PMLR, 2016.

\bibitem{he2016deep}
Kaiming He, Xiangyu Zhang, Shaoqing Ren, and Jian Sun.
\newblock Deep residual learning for image recognition.
\newblock In {\em Proceedings of the IEEE Conference on Computer Vision and
  Pattern Recognition}, pages 770--778, 2016.

\bibitem{hearn2009games}
Robert~A Hearn and Erik~D Demaine.
\newblock {\em Games, Puzzles, and Computation}.
\newblock CRC Press, 2009.

\bibitem{heckerman1995learning}
David Heckerman, Dan Geiger, and David~M Chickering.
\newblock Learning {Bayesian} networks: The combination of knowledge and
  statistical data.
\newblock {\em Machine Learning}, 20(3):197--243, 1995.

\bibitem{heess2013actor}
Nicolas Heess, David Silver, and Yee~Whye Teh.
\newblock Actor-critic reinforcement learning with energy-based policies.
\newblock In {\em European Workshop on Reinforcement Learning}, pages 45--58,
  2013.

\bibitem{heess2017emergence}
Nicolas Heess, Dhruva TB, Srinivasan Sriram, Jay Lemmon, Josh Merel, Greg
  Wayne, Yuval Tassa, Tom Erez, Ziyu Wang, SM~Eslami, Martin Riedmiller, and
  David Silver.
\newblock Emergence of locomotion behaviours in rich environments.
\newblock {\em arXiv preprint arXiv:1707.02286}, 2017.

\bibitem{heess2015learning}
Nicolas Heess, Gregory Wayne, David Silver, Timothy Lillicrap, Tom Erez, and
  Yuval Tassa.
\newblock Learning continuous control policies by stochastic value gradients.
\newblock In {\em Advances in Neural Information Processing Systems}, pages
  2944--2952, 2015.

\bibitem{heinz2000new}
Ernst~A Heinz.
\newblock New self-play results in computer chess.
\newblock In {\em International Conference on Computers and Games}, pages
  262--276. Springer, 2000.

\bibitem{henderson2018deep}
Peter Henderson, Riashat Islam, Philip Bachman, Joelle Pineau, Doina Precup,
  and David Meger.
\newblock Deep reinforcement learning that matters.
\newblock In {\em Thirty-Second AAAI Conference on Artificial Intelligence},
  2018.

\bibitem{hendrikx2013procedural}
Mark Hendrikx, Sebastiaan Meijer, Joeri Van Der~Velden, and Alexandru Iosup.
\newblock Procedural content generation for games: A survey.
\newblock {\em ACM Transactions on Multimedia Computing, Communications, and
  Applications}, 9(1):1--22, 2013.

\bibitem{hennessy2011computer}
John~L Hennessy and David~A Patterson.
\newblock {\em Computer Architecture: a Quantitative Approach}.
\newblock Elsevier, 2017.

\bibitem{henrich2008five}
Joseph Henrich, Robert Boyd, and Peter~J Richerson.
\newblock Five misunderstandings about cultural evolution.
\newblock {\em Human Nature}, 19(2):119--137, 2008.

\bibitem{hernandez2017survey}
Pablo Hernandez-Leal, Michael Kaisers, Tim Baarslag, and Enrique~Munoz de~Cote.
\newblock A survey of learning in multiagent environments: Dealing with
  non-stationarity.
\newblock {\em arXiv preprint arXiv:1707.09183}, 2017.

\bibitem{hernandez2019survey}
Pablo Hernandez-Leal, Bilal Kartal, and Matthew~E Taylor.
\newblock A survey and critique of multiagent deep reinforcement learning.
\newblock {\em Autonomous Agents and Multi-Agent Systems}, 33(6):750--797,
  2019.

\bibitem{hessel2021muesli}
Matteo Hessel, Ivo Danihelka, Fabio Viola, Arthur Guez, Simon Schmitt, Laurent
  Sifre, Theophane Weber, David Silver, and Hado van Hasselt.
\newblock Muesli: Combining improvements in policy optimization.
\newblock In {\em International Conference on Machine Learning}, pages
  4214--4226, 2021.

\bibitem{hessel2017rainbow}
Matteo Hessel, Joseph Modayil, Hado Van~Hasselt, Tom Schaul, Georg Ostrovski,
  Will Dabney, Dan Horgan, Bilal Piot, Mohammad Azar, and David Silver.
\newblock Rainbow: Combining improvements in deep reinforcement learning.
\newblock In {\em AAAI}, pages 3215--3222, 2018.

\bibitem{heylighen1998makes}
Francis Heylighen.
\newblock {\em What makes a Meme Successful? Selection Criteria for Cultural
  Evolution}.
\newblock Association Internationale de Cybernetique, 1998.

\bibitem{higgins2017darla}
Irina Higgins, Arka Pal, Andrei Rusu, Loic Matthey, Christopher Burgess,
  Alexander Pritzel, Matthew Botvinick, Charles Blundell, and Alexander
  Lerchner.
\newblock Darla: Improving zero-shot transfer in reinforcement learning.
\newblock In {\em International Conference on Machine Learning}, pages
  1480--1490. PMLR, 2017.

\bibitem{hill2018stable}
Ashley Hill, Antonin Raffin, Maximilian Ernestus, Adam Gleave, Anssi
  Kanervisto, Rene Traore, Prafulla Dhariwal, Christopher Hesse, Oleg Klimov,
  Alex Nichol, Matthias Plappert, Alec Radford, John Schulman, Szymon Sidor,
  and Yuhuai Wu.
\newblock Stable baselines.
\newblock \url{https://github.com/hill-a/stable-baselines}, 2018.

\bibitem{hillis1982new}
W~Daniel Hillis.
\newblock New computer architectures and their relationship to physics or why
  computer science is no good.
\newblock {\em International Journal of Theoretical Physics}, 21(3-4):255--262,
  1982.

\bibitem{hillis1993cm}
W~Daniel Hillis and Lewis~W Tucker.
\newblock The {CM-5} connection machine: A scalable supercomputer.
\newblock {\em Communications of the ACM}, 36(11):30--41, 1993.

\bibitem{hinton2015distilling}
Geoffrey Hinton, Oriol Vinyals, and Jeff Dean.
\newblock Distilling the knowledge in a neural network.
\newblock {\em arXiv preprint arXiv:1503.02531}, 2015.

\bibitem{hinton2006reducing}
Geoffrey~E Hinton and Ruslan~R Salakhutdinov.
\newblock Reducing the dimensionality of data with neural networks.
\newblock {\em Science}, 313(5786):504--507, 2006.

\bibitem{hinton1999unsupervised}
Geoffrey~E Hinton and Terrence~Joseph Sejnowski, editors.
\newblock {\em Unsupervised Learning: Foundations of Neural Computation}.
\newblock MIT press, 1999.

\bibitem{hinton2012improving}
Geoffrey~E Hinton, Nitish Srivastava, Alex Krizhevsky, Ilya Sutskever, and
  Ruslan~R Salakhutdinov.
\newblock Improving neural networks by preventing co-adaptation of feature
  detectors.
\newblock {\em arXiv preprint arXiv:1207.0580}, 2012.

\bibitem{hochreiter1997long}
Sepp Hochreiter and J{\"u}rgen Schmidhuber.
\newblock Long short-term memory.
\newblock {\em Neural Computation}, 9(8):1735--1780, 1997.

\bibitem{holland1975adaptation}
John Holland.
\newblock Adaptation in natural and artificial systems: an introductory
  analysis with application to biology.
\newblock {\em Control and Artificial Intelligence}, 1975.

\bibitem{holland1992genetic}
John~H Holland.
\newblock Genetic algorithms.
\newblock {\em Scientific American}, 267(1):66--73, 1992.

\bibitem{holldobler2009superorganism}
Bert H{\"o}lldobler and Edward~O Wilson.
\newblock {\em The Superorganism: the Beauty, Elegance, and Strangeness of
  Insect Societies}.
\newblock WW Norton \& Company, 2009.

\bibitem{hopfield1982neural}
John~J Hopfield.
\newblock Neural networks and physical systems with emergent collective
  computational abilities.
\newblock {\em Proceedings of the National Academy of Sciences},
  79(8):2554--2558, 1982.

\bibitem{horgan2018distributed}
Dan Horgan, John Quan, David Budden, Gabriel Barth-Maron, Matteo Hessel, Hado
  Van~Hasselt, and David Silver.
\newblock Distributed prioritized experience replay.
\newblock In {\em International Conference on Learning Representations}, 2018.

\bibitem{hospedales2020meta}
Timothy Hospedales, Antreas Antoniou, Paul Micaelli, and Amos Storkey.
\newblock Meta-learning in neural networks: A survey.
\newblock {\em arXiv preprint arXiv:2004.05439}, 2020.

\bibitem{howard1964dynamic}
Ronald~A Howard.
\newblock {\em Dynamic programming and {Markov} processes}.
\newblock New York: John Wiley, 1964.

\bibitem{hsu2020revisiting}
Chloe Ching-Yun Hsu, Celestine Mendler-D{\"u}nner, and Moritz Hardt.
\newblock Revisiting design choices in proximal policy optimization.
\newblock {\em arXiv preprint arXiv:2009.10897}, 2020.

\bibitem{hsu2004behind}
Feng-Hsiung Hsu.
\newblock {\em Behind {Deep Blue}: Building the computer that defeated the
  world chess champion}.
\newblock Princeton University Press, 2004.

\bibitem{hsu1990grandmaster}
Feng-Hsiung Hsu, Thomas Anantharaman, Murray Campbell, and Andreas Nowatzyk.
\newblock A grandmaster chess machine.
\newblock {\em Scientific American}, 263(4):44--51, 1990.

\bibitem{huzero}
R~Lily Hu, Caiming Xiong, and Richard Socher.
\newblock Zero-shot image classification guided by natural language
  descriptions of classes: A meta-learning approach.
\newblock In {\em Advances in Neural Information Processing Systems}, 2018.

\bibitem{hubel1963shape}
David~H Hubel and Torsten~N Wiesel.
\newblock Shape and arrangement of columns in cat's striate cortex.
\newblock {\em The Journal of Physiology}, 165(3):559--568, 1963.

\bibitem{hubel1968receptive}
David~H Hubel and Torsten~N Wiesel.
\newblock Receptive fields and functional architecture of monkey striate
  cortex.
\newblock {\em The Journal of Physiology}, 195(1):215--243, 1968.

\bibitem{hubert2021learning}
Thomas Hubert, Julian Schrittwieser, Ioannis Antonoglou, Mohammadamin
  Barekatain, Simon Schmitt, and David Silver.
\newblock Learning and planning in complex action spaces.
\newblock In {\em International Conference on Machine Learning}, pages
  4476--4486, 2021.

\bibitem{huh2016makes}
Minyoung Huh, Pulkit Agrawal, and Alexei~A Efros.
\newblock What makes {ImageNet} good for transfer learning?
\newblock {\em arXiv preprint arXiv:1608.08614}, 2016.

\bibitem{hui2018rl}
Jonathan Hui.
\newblock {RL---DQN Deep Q-network}
  \url{https://medium.com/@jonathan_hui/rl-dqn-deep-q-network-e207751f7ae4}.
\newblock Medium post.

\bibitem{hui2018model}
Jonathan Hui.
\newblock Model-based reinforcement learning
  \url{https://medium.com/@jonathan_hui/rl-model-based-reinforcement-learning-3c2b6f0aa323}.
\newblock Medium post, 2018.

\bibitem{huisman2020deep}
Mike Huisman, Jan van Rijn, and Aske Plaat.
\newblock Metalearning for deep neural networks.
\newblock In Pavel Brazdil et~al., editors, {\em Metalearning: Applications to
  data mining}. Springer, 2022.

\bibitem{huisman2021survey}
Mike Huisman, Jan~N. van Rijn, and Aske Plaat.
\newblock A survey of deep meta-learning.
\newblock {\em Artificial Intelligence Review}, 2021.

\bibitem{hutson2018artificial}
Matthew Hutson.
\newblock {Artificial Intelligence} faces reproducibility crisis.
\newblock {\em Science}, 359:725--726, 2018.

\bibitem{hutter2011sequential}
Frank Hutter, Holger~H Hoos, and Kevin Leyton-Brown.
\newblock Sequential model-based optimization for general algorithm
  configuration.
\newblock In {\em International Conference on Learning and Intelligent
  Optimization}, pages 507--523. Springer, 2011.

\bibitem{hutter2009paramils}
Frank Hutter, Holger~H Hoos, Kevin Leyton-Brown, and Thomas St{\"u}tzle.
\newblock {ParamILS:} an automatic algorithm configuration framework.
\newblock {\em Journal of Artificial Intelligence Research}, 36:267--306, 2009.

\bibitem{hutter2019automated}
Frank Hutter, Lars Kotthoff, and Joaquin Vanschoren.
\newblock {\em Automated Machine Learning: Methods, Systems, Challenges}.
\newblock Springer Nature, 2019.

\bibitem{ilin2007efficient}
Roman Ilin, Robert Kozma, and Paul~J Werbos.
\newblock Efficient learning in cellular simultaneous recurrent neural
  networks---the case of maze navigation problem.
\newblock In {\em 2007 IEEE International Symposium on Approximate Dynamic
  Programming and Reinforcement Learning}, pages 324--329, 2007.

\bibitem{ioffe2017batch}
Sergey Ioffe.
\newblock Batch renormalization: Towards reducing minibatch dependence in
  batch-normalized models.
\newblock In {\em Advances in Neural Information Processing Systems}, pages
  1945--1953, 2017.

\bibitem{islam2017reproducibility}
Riashat Islam, Peter Henderson, Maziar Gomrokchi, and Doina Precup.
\newblock Reproducibility of benchmarked deep reinforcement learning tasks for
  continuous control.
\newblock {\em arXiv preprint arXiv:1708.04133}, 2017.

\bibitem{jacob2021modeling}
Athul~Paul Jacob, David~J Wu, Gabriele Farina, Adam Lerer, Anton Bakhtin, Jacob
  Andreas, and Noam Brown.
\newblock Modeling strong and human-like gameplay with {KL}-regularized search.
\newblock {\em arXiv preprint arXiv:2112.07544}, 2021.

\bibitem{jaderberg2019human}
Max Jaderberg, Wojciech~M. Czarnecki, Iain Dunning, Luke Marris, Guy Lever,
  Antonio~Garcia Casta{\~n}eda, Charles Beattie, Neil~C. Rabinowitz, Ari~S.
  Morcos, Avraham Ruderman, Nicolas Sonnerat, Tim Green, Louise Deason, Joel~Z.
  Leibo, David Silver, Demis Hassabis, Koray Kavukcuoglu, and Thore Graepel.
\newblock Human-level performance in {3D} multiplayer games with
  population-based reinforcement learning.
\newblock {\em Science}, 364(6443):859--865, 2019.

\bibitem{jaderberg2017population}
Max Jaderberg, Valentin Dalibard, Simon Osindero, Wojciech~M. Czarnecki, Jeff
  Donahue, Ali Razavi, Oriol Vinyals, Tim Green, Iain Dunning, Karen Simonyan,
  Chrisantha Fernando, and Koray Kavukcuoglu.
\newblock Population based training of neural networks.
\newblock {\em arXiv preprint arXiv:1711.09846}, 2017.

\bibitem{james2020rlbench}
Stephen James, Zicong Ma, David~Rovick Arrojo, and Andrew~J Davison.
\newblock {RLbench:} the robot learning benchmark \& learning environment.
\newblock {\em IEEE Robotics and Automation Letters}, 5(2):3019--3026, 2020.

\bibitem{janner2019trust}
Michael Janner, Justin Fu, Marvin Zhang, and Sergey Levine.
\newblock When to trust your model: Model-based policy optimization.
\newblock In {\em Advances in Neural Information Processing Systems}, pages
  12498--12509, 2019.

\bibitem{janner2021reinforcement}
Michael Janner, Qiyang Li, and Sergey Levine.
\newblock Reinforcement learning as one big sequence modeling problem.
\newblock {\em arXiv preprint arXiv:2106.02039}, 2021.

\bibitem{jawlik2016statistics}
Andrew~A Jawlik.
\newblock {\em Statistics from A to Z: Confusing concepts clarified}.
\newblock John Wiley \& Sons, 2016.

\bibitem{johanson2012efficient}
Michael Johanson, Nolan Bard, Marc Lanctot, Richard~G Gibson, and Michael
  Bowling.
\newblock Efficient {Nash} equilibrium approximation through {Monte Carlo}
  counterfactual regret minimization.
\newblock In {\em AAMAS}, pages 837--846, 2012.

\bibitem{jolliffe2016principal}
Ian~T Jolliffe and Jorge Cadima.
\newblock Principal component analysis: a review and recent developments.
\newblock {\em Philosophical Transactions of the Royal Society A: Mathematical,
  Physical and Engineering Sciences}, 374(2065):20150202, 2016.

\bibitem{jordan1998learning}
Michael~Irwin Jordan.
\newblock {\em Learning in Graphical Models}, volume~89.
\newblock Springer Science \& Business Media, 1998.

\bibitem{juliani2016simple}
Arthur Juliani.
\newblock Simple reinforcement learning with tensorflow part 8: Asynchronous
  actor-critic agents {(A3C)}
  \url{https://medium.com/emergent-future/simple-reinforcement-learning-with-tensorflow-part-8-asynchronous-actor-critic-agents-a3c-c88f72a5e9f2},
  2016.

\bibitem{juliani2018unity}
Arthur Juliani, Vincent-Pierre Berges, Ervin Teng, Andrew Cohen, Jonathan
  Harper, Chris Elion, Chris Goy, Yuan Gao, Hunter Henry, Marwan Mattar, and
  Danny Lange.
\newblock Unity: A general platform for intelligent agents.
\newblock {\em arXiv preprint arXiv:1809.02627}, 2018.

\bibitem{juliani2019obstacle}
Arthur Juliani, Ahmed Khalifa, Vincent-Pierre Berges, Jonathan Harper, Ervin
  Teng, Hunter Henry, Adam Crespi, Julian Togelius, and Danny Lange.
\newblock Obstacle tower: A generalization challenge in vision, control, and
  planning.
\newblock {\em arXiv preprint arXiv:1902.01378}, 2019.

\bibitem{jumper2021highly}
John Jumper, Richard Evans, Alexander Pritzel, Tim Green, Michael Figurnov,
  Olaf Ronneberger, Kathryn Tunyasuvunakool, Russ Bates, Augustin {\v
  Z}{\'\i}dek, Anna Potapenko, Alex Bridgland, Clemens Meyer, Simon A.~A. Kohl,
  Andrew~J. Ballard, Andrew Cowie, Bernardino Romera-Paredes, Stanislav
  Nikolov, Rishub Jain, Jonas Adler, Trevor Back, Stig Petersen, David Reiman,
  Ellen Clancy, Michal Zielinski, Martin Steinegger, Michalina Pacholska, Tamas
  Berghammer, Sebastian Bodenstein, David Silver, Oriol Vinyals, Andrew~W.
  Senior, Koray Kavukcuoglu, Pushmeet Kohli, and Demis Hassabis.
\newblock Highly accurate protein structure prediction with {AlphaFold}.
\newblock {\em Nature}, 596(7873):583--589, 2021.

\bibitem{junghanns2001sokoban}
Andreas Junghanns and Jonathan Schaeffer.
\newblock Sokoban: Enhancing general single-agent search methods using domain
  knowledge.
\newblock {\em Artificial Intelligence}, 129(1-2):219--251, 2001.

\bibitem{justesen2019deep}
Niels Justesen, Philip Bontrager, Julian Togelius, and Sebastian Risi.
\newblock Deep learning for video game playing.
\newblock {\em IEEE Transactions on Games}, 12(1):1--20, 2019.

\bibitem{justesen2018illuminating}
Niels Justesen, Ruben~Rodriguez Torrado, Philip Bontrager, Ahmed Khalifa,
  Julian Togelius, and Sebastian Risi.
\newblock Illuminating generalization in deep reinforcement learning through
  procedural level generation.
\newblock {\em arXiv preprint arXiv:1806.10729}, 2018.

\bibitem{kaelbling1996reinforcement}
Leslie~Pack Kaelbling, Michael~L Littman, and Andrew~W Moore.
\newblock Reinforcement learning: A survey.
\newblock {\em Journal of Artificial Intelligence Research}, 4:237--285, 1996.

\bibitem{kahneman2013prospect}
Daniel Kahneman and Amos Tversky.
\newblock Prospect theory: An analysis of decision under risk.
\newblock In {\em Handbook of the Fundamentals of Financial Decision Making:
  Part I}, pages 99--127. World Scientific, 2013.

\bibitem{kaiser2019model}
Lukasz Kaiser, Mohammad Babaeizadeh, Piotr Milos, Blazej Osinski, Roy~H.
  Campbell, Konrad Czechowski, Dumitru Erhan, Chelsea Finn, Piotr Kozakowski,
  Sergey Levine, Ryan Sepassi, George Tucker, and Henryk Michalewski.
\newblock Model-based reinforcement learning for {Atari}.
\newblock {\em arXiv:1903.00374}, 2019.

\bibitem{nakkiran2019sgd}
Dimitris Kalimeris, Gal Kaplun, Preetum Nakkiran, Benjamin~L. Edelman, Tristan
  Yang, Boaz Barak, and Haofeng Zhang.
\newblock {SGD} on neural networks learns functions of increasing complexity.
\newblock In {\em Advances in Neural Information Processing Systems}, pages
  3491--3501, 2019.

\bibitem{kalweit2017uncertainty}
Gabriel Kalweit and Joschka Boedecker.
\newblock Uncertainty-driven imagination for continuous deep reinforcement
  learning.
\newblock In {\em Conference on Robot Learning}, pages 195--206, 2017.

\bibitem{kamyar2014aircraft}
Reza Kamyar and Ehsan Taheri.
\newblock Aircraft optimal terrain/threat-based trajectory planning and
  control.
\newblock {\em Journal of Guidance, Control, and Dynamics}, 37(2):466--483,
  2014.

\bibitem{learn}
Satwik Kansal and Brendan Martin.
\newblock Learn data science webpage., 2018.

\bibitem{kappen2005path}
Hilbert~J Kappen.
\newblock Path integrals and symmetry breaking for optimal control theory.
\newblock {\em Journal of statistical mechanics: theory and experiment},
  2005(11):P11011, 2005.

\bibitem{kapturowski2018recurrent}
Steven Kapturowski, Georg Ostrovski, John Quan, Remi Munos, and Will Dabney.
\newblock Recurrent experience replay in distributed reinforcement learning.
\newblock In {\em International Conference on Learning Representations}, 2018.

\bibitem{karl2016deep}
Maximilian Karl, Maximilian Soelch, Justin Bayer, and Patrick Van~der Smagt.
\newblock Deep variational {Bayes} filters: Unsupervised learning of state
  space models from raw data.
\newblock {\em arXiv preprint arXiv:1605.06432}, 2016.

\bibitem{karpathy2015}
Andrej Karpathy.
\newblock The unreasonable effectiveness of recurrent neural networks.
  \url{http://karpathy.github.io/2015/05/21/rnn-effectiveness/}.
\newblock Andrej Karpathy Blog, 2015.

\bibitem{karpathy2016pong}
Andrej Karpathy.
\newblock Deep reinforcement learning: Pong from pixels.
  \url{http://karpathy.github.io/2016/05/31/rl/}.
\newblock Andrej Karpathy Blog, 2016.

\bibitem{karpathy2015visualizing}
Andrej Karpathy, Justin Johnson, and Li~Fei-Fei.
\newblock Visualizing and understanding recurrent networks.
\newblock {\em arXiv preprint arXiv:1506.02078}, 2015.

\bibitem{karras2017progressive}
Tero Karras, Timo Aila, Samuli Laine, and Jaakko Lehtinen.
\newblock Progressive growing of {GANs} for improved quality, stability, and
  variation.
\newblock In {\em International Conference on Learning Representations}, 2018.

\bibitem{kelley1960gradient}
Henry~J Kelley.
\newblock Gradient theory of optimal flight paths.
\newblock {\em American Rocket Society Journal}, 30(10):947--954, 1960.

\bibitem{kelly2017multi}
Stephen Kelly and Malcolm~I Heywood.
\newblock Multi-task learning in {Atari} video games with emergent tangled
  program graphs.
\newblock In {\em Proceedings of the Genetic and Evolutionary Computation
  Conference}, pages 195--202. ACM, 2017.

\bibitem{kelly2018emergent}
Stephen Kelly and Malcolm~I Heywood.
\newblock Emergent tangled program graphs in multi-task learning.
\newblock In {\em IJCAI}, pages 5294--5298, 2018.

\bibitem{kennedy2006swarm}
James Kennedy.
\newblock Swarm intelligence.
\newblock In {\em Handbook of Nature-Inspired and Innovative Computing}, pages
  187--219. Springer, 2006.

\bibitem{kerschke2019automated}
Pascal Kerschke, Holger~H Hoos, Frank Neumann, and Heike Trautmann.
\newblock Automated algorithm selection: Survey and perspectives.
\newblock {\em Evolutionary Computation}, 27(1):3--45, 2019.

\bibitem{khadka2019collaborative}
Shauharda Khadka, Somdeb Majumdar, Tarek Nassar, Zach Dwiel, Evren Tumer,
  Santiago Miret, Yinyin Liu, and Kagan Tumer.
\newblock Collaborative evolutionary reinforcement learning.
\newblock In {\em International Conference on Machine Learning}, pages
  3341--3350. PMLR, 2019.

\bibitem{khadka2018evolutionary}
Shauharda Khadka and Kagan Tumer.
\newblock Evolutionary reinforcement learning.
\newblock {\em arXiv preprint arXiv:1805.07917}, 2018.

\bibitem{khetarpal2018re}
Khimya Khetarpal, Zafarali Ahmed, Andre Cianflone, Riashat Islam, and Joelle
  Pineau.
\newblock Re-evaluate: Reproducibility in evaluating reinforcement learning
  algorithms.
\newblock In {\em Reproducibility in Machine Learning Workshop, ICML}, 2018.

\bibitem{kingma2013auto}
Diederik~P Kingma and Max Welling.
\newblock Auto-encoding variational {Bayes}.
\newblock In {\em International Conference on Learning Representations}, 2014.

\bibitem{kingma2019introduction}
Diederik~P Kingma and Max Welling.
\newblock An introduction to variational autoencoders.
\newblock {\em Found. Trends Mach. Learn.}, 12(4):307--392, 2019.

\bibitem{kirk2021survey}
Robert Kirk, Amy Zhang, Edward Grefenstette, and Tim Rockt{\"a}schel.
\newblock A survey of generalisation in deep reinforcement learning.
\newblock {\em arXiv preprint arXiv:2111.09794}, 2021.

\bibitem{klijn2021coevolutionairy}
Daan Klijn and AE~Eiben.
\newblock A coevolutionairy approach to deep multi-agent reinforcement
  learning.
\newblock {\em arXiv preprint arXiv:2104.05610}, 2021.

\bibitem{knoblock1990learning}
Craig~A Knoblock.
\newblock Learning abstraction hierarchies for problem solving.
\newblock In {\em AAAI}, pages 923--928, 1990.

\bibitem{knuth1975analysis}
Donald~E Knuth and Ronald~W Moore.
\newblock An analysis of alpha-beta pruning.
\newblock {\em Artificial Intelligence}, 6(4):293--326, 1975.

\bibitem{kober2013reinforcement}
Jens Kober, J~Andrew Bagnell, and Jan Peters.
\newblock Reinforcement learning in robotics: A survey.
\newblock {\em The International Journal of Robotics Research},
  32(11):1238--1274, 2013.

\bibitem{koch2015siamese}
Gregory Koch, Richard Zemel, and Ruslan Salakhutdinov.
\newblock Siamese neural networks for one-shot image recognition.
\newblock In {\em ICML Deep Learning workshop}, volume~2. Lille, 2015.

\bibitem{kocsis2006bandit}
Levente Kocsis and Csaba Szepesv{\'a}ri.
\newblock Bandit based {Monte-Carlo} planning.
\newblock In {\em European Conference on Machine Learning}, pages 282--293.
  Springer, 2006.

\bibitem{konda2000actor}
Vijay~R Konda and John~N Tsitsiklis.
\newblock Actor--critic algorithms.
\newblock In {\em Advances in Neural Information Processing Systems}, pages
  1008--1014, 2000.

\bibitem{konda1999actor}
Vijaymohan~R Konda and Vivek~S Borkar.
\newblock {Actor--Critic}-type learning algorithms for {Markov Decision
  Processes}.
\newblock {\em SIAM Journal on Control and Optimization}, 38(1):94--123, 1999.

\bibitem{korf1985depth}
Richard~E Korf.
\newblock Depth-first iterative-deepening: An optimal admissible tree search.
\newblock {\em Artificial intelligence}, 27(1):97--109, 1985.

\bibitem{kormushev2010robot}
Petar Kormushev, Sylvain Calinon, and Darwin~G Caldwell.
\newblock Robot motor skill coordination with em-based reinforcement learning.
\newblock In {\em 2010 IEEE/RSJ International Conference on Intelligent Robots
  and Systems}, pages 3232--3237. IEEE, 2010.

\bibitem{kottur2017natural}
Satwik Kottur, Jos{\'e}~MF Moura, Stefan Lee, and Dhruv Batra.
\newblock Natural language does not emerge 'naturally' in multi-agent dialog.
\newblock In {\em Proceedings of the 2017 Conference on Empirical Methods in
  Natural Language Processing, {EMNLP} 2017, Copenhagen}, pages 2962--2967,
  2017.

\bibitem{kotz2004continuous}
Samuel Kotz, Narayanaswamy Balakrishnan, and Norman~L Johnson.
\newblock {\em Continuous Multivariate Distributions, Volume 1: Models and
  Applications}.
\newblock John Wiley \& Sons, 2004.

\bibitem{kouvaritakis2016model}
Basil Kouvaritakis and Mark Cannon.
\newblock {\em Model Predictive Control}.
\newblock Springer, 2016.

\bibitem{kraemer2016multi}
Landon Kraemer and Bikramjit Banerjee.
\newblock Multi-agent reinforcement learning as a rehearsal for decentralized
  planning.
\newblock {\em Neurocomputing}, 190:82--94, 2016.

\bibitem{kramer1991nonlinear}
Mark~A Kramer.
\newblock Nonlinear principal component analysis using autoassociative neural
  networks.
\newblock {\em AIChE journal}, 37(2):233--243, 1991.

\bibitem{kraus1994negotiation}
Sarit Kraus, Eithan Ephrati, and Daniel Lehmann.
\newblock Negotiation in a non-cooperative environment.
\newblock {\em Journal of Experimental \& Theoretical Artificial Intelligence},
  3(4):255--281, 1994.

\bibitem{kraus1988diplomat}
Sarit Kraus and Daniel Lehmann.
\newblock Diplomat, an agent in a multi agent environment: An overview.
\newblock In {\em IEEE International Performance Computing and Communications
  Conference}, pages 434--438, 1988.

\bibitem{krizhevsky2012imagenet}
Alex Krizhevsky, Ilya Sutskever, and Geoffrey~E Hinton.
\newblock Imagenet classification with deep convolutional neural networks.
\newblock In {\em Advances in Neural Information Processing Systems}, pages
  1097--1105, 2012.

\bibitem{krueger2009flexible}
Kai~A Krueger and Peter Dayan.
\newblock Flexible shaping: How learning in small steps helps.
\newblock {\em Cognition}, 110(3):380--394, 2009.

\bibitem{kuhn1997prisoner}
Steven Kuhn.
\newblock {\em Prisoner's Dilemma.}
\newblock The Stanford Encyclopedia of Philosophy,
  \url{https://plato.stanford.edu/entries/prisoner-dilemma/}, 1997.

\bibitem{kuipers2013improving}
Jan Kuipers, Aske Plaat, Jos~AM Vermaseren, and H~Jaap van~den Herik.
\newblock Improving multivariate {Horner} schemes with {Monte Carlo} tree
  search.
\newblock {\em Computer Physics Communications}, 184(11):2391--2395, 2013.

\bibitem{kulkarni2016hierarchical}
Tejas~D Kulkarni, Karthik Narasimhan, Ardavan Saeedi, and Josh Tenenbaum.
\newblock Hierarchical deep reinforcement learning: Integrating temporal
  abstraction and intrinsic motivation.
\newblock In {\em Advances in Neural Information Processing Systems}, pages
  3675--3683, 2016.

\bibitem{kullback1951information}
Solomon Kullback and Richard~A Leibler.
\newblock On information and sufficiency.
\newblock {\em The Annals of Mathematical Statistics}, 22(1):79--86, 1951.

\bibitem{kuo2020encoding}
Yen-Ling Kuo, Boris Katz, and Andrei Barbu.
\newblock Encoding formulas as deep networks: Reinforcement learning for
  zero-shot execution of {LTL} formulas.
\newblock {\em arXiv preprint arXiv:2006.01110}, 2020.

\bibitem{kurach2020google}
Karol Kurach, Anton Raichuk, Piotr Sta{\'n}czyk, Micha{\l} Zajac, Olivier
  Bachem, Lasse Espeholt, Carlos Riquelme, Damien Vincent, Marcin Michalski,
  Olivier Bousquet, and Sylvain Gelly.
\newblock Google research football: A novel reinforcement learning environment.
\newblock In {\em Proceedings of the AAAI Conference on Artificial
  Intelligence}, volume~34, pages 4501--4510, 2020.

\bibitem{kurutach2018model}
Thanard Kurutach, Ignasi Clavera, Yan Duan, Aviv Tamar, and Pieter Abbeel.
\newblock Model-ensemble trust-region policy optimization.
\newblock In {\em International Conference on Learning Representations}, 2018.

\bibitem{kwon1983stabilizing}
W~Hi Kwon, AM~Bruckstein, and T~Kailath.
\newblock Stabilizing state-feedback design via the moving horizon method.
\newblock {\em International Journal of Control}, 37(3):631--643, 1983.

\bibitem{lagoudakis2003least}
Michail~G Lagoudakis and Ronald Parr.
\newblock Least-squares policy iteration.
\newblock {\em Journal of Machine Learning Research}, 4:1107--1149, Dec 2003.

\bibitem{lai1987adaptive}
Tze~Leung Lai.
\newblock Adaptive treatment allocation and the multi-armed bandit problem.
\newblock {\em The Annals of Statistics}, pages 1091--1114, 1987.

\bibitem{lai1985asymptotically}
Tze~Leung Lai and Herbert Robbins.
\newblock Asymptotically efficient adaptive allocation rules.
\newblock {\em Advances in Applied Mathematics}, 6(1):4--22, 1985.

\bibitem{laird1986chunking}
John~E Laird, Paul~S Rosenbloom, and Allen Newell.
\newblock Chunking in {Soar:} the anatomy of a general learning mechanism.
\newblock {\em Machine learning}, 1(1):11--46, 1986.

\bibitem{lake2011one}
Brenden Lake, Ruslan Salakhutdinov, Jason Gross, and Joshua Tenenbaum.
\newblock One shot learning of simple visual concepts.
\newblock In {\em Proceedings of the Annual Meeting of the Cognitive Science
  Society}, volume~33, 2011.

\bibitem{lake2015human}
Brenden~M Lake, Ruslan Salakhutdinov, and Joshua~B Tenenbaum.
\newblock Human-level concept learning through probabilistic program induction.
\newblock {\em Science}, 350(6266):1332--1338, 2015.

\bibitem{lake2019omniglot}
Brenden~M Lake, Ruslan Salakhutdinov, and Joshua~B Tenenbaum.
\newblock The {Omniglot} challenge: a 3-year progress report.
\newblock {\em Current Opinion in Behavioral Sciences}, 29:97--104, 2019.

\bibitem{lakshminarayanan2017simple}
Balaji Lakshminarayanan, Alexander Pritzel, and Charles Blundell.
\newblock Simple and scalable predictive uncertainty estimation using deep
  ensembles.
\newblock In {\em Advances in Neural Information Processing Systems}, pages
  6402--6413, 2017.

\bibitem{lampert2009learning}
Christoph~H Lampert, Hannes Nickisch, and Stefan Harmeling.
\newblock Learning to detect unseen object classes by between-class attribute
  transfer.
\newblock In {\em 2009 IEEE Conference on Computer Vision and Pattern
  Recognition}, pages 951--958. IEEE, 2009.

\bibitem{lanctot2019openspiel}
Marc Lanctot, Edward Lockhart, Jean-Baptiste Lespiau, Vin{\'\i}cius~Flores
  Zambaldi, Satyaki Upadhyay, Julien P{\'e}rolat, Sriram Srinivasan, Finbarr
  Timbers, Karl Tuyls, Shayegan Omidshafiei, Daniel Hennes, Dustin Morrill,
  Paul Muller, Timo Ewalds, Ryan Faulkner, J{\'a}nos Kram{\'a}r, Bart~De
  Vylder, Brennan Saeta, James Bradbury, David Ding, Sebastian Borgeaud,
  Matthew Lai, Julian Schrittwieser, Thomas~W. Anthony, Edward Hughes, Ivo
  Danihelka, and Jonah Ryan-Davis.
\newblock Openspiel: A framework for reinforcement learning in games.
\newblock {\em arXiv preprint arXiv:1908.09453}, 2019.

\bibitem{lanctot2009monte}
Marc Lanctot, Kevin Waugh, Martin Zinkevich, and Michael~H Bowling.
\newblock Monte carlo sampling for regret minimization in extensive games.
\newblock In {\em Advances in Neural Information Processing Systems}, pages
  1078--1086, 2009.

\bibitem{lanctot2017unified}
Marc Lanctot, Vinicius Zambaldi, Audrunas Gruslys, Angeliki Lazaridou, Karl
  Tuyls, Julien P{\'e}rolat, David Silver, and Thore Graepel.
\newblock A unified game-theoretic approach to multiagent reinforcement
  learning.
\newblock In {\em Advances in Neural Information Processing Systems}, pages
  4190--4203, 2017.

\bibitem{lange2010deep}
Sascha Lange and Martin Riedmiller.
\newblock Deep auto-encoder neural networks in reinforcement learning.
\newblock In {\em The 2010 International Joint Conference on Neural Networks
  (IJCNN)}, pages 1--8. IEEE, 2010.

\bibitem{larochelle2008zero}
Hugo Larochelle, Dumitru Erhan, and Yoshua Bengio.
\newblock Zero-data learning of new tasks.
\newblock In {\em AAAI}, volume~1, page~3, 2008.

\bibitem{laterre2018ranked}
Alexandre Laterre, Yunguan Fu, Mohamed~Khalil Jabri, Alain-Sam Cohen, David
  Kas, Karl Hajjar, Torbjorn~S Dahl, Amine Kerkeni, and Karim Beguir.
\newblock Ranked reward: Enabling self-play reinforcement learning for
  combinatorial optimization.
\newblock {\em arXiv preprint arXiv:1807.01672}, 2018.

\bibitem{latombe2012robot}
Jean-Claude Latombe.
\newblock {\em Robot Motion Planning}, volume 124.
\newblock Springer Science \& Business Media, 2012.

\bibitem{lauritzen1996graphical}
Steffen~L Lauritzen.
\newblock {\em Graphical Models}, volume~17.
\newblock Clarendon Press, 1996.

\bibitem{lazaridou2016multi}
Angeliki Lazaridou, Alexander Peysakhovich, and Marco Baroni.
\newblock Multi-agent cooperation and the emergence of (natural) language.
\newblock In {\em International Conference on Learning Representations}, 2017.

\bibitem{lecun2015deep}
Yann LeCun, Yoshua Bengio, and Geoffrey Hinton.
\newblock Deep learning.
\newblock {\em Nature}, 521(7553):436, 2015.

\bibitem{lecun1989backpropagation}
Yann LeCun, Bernhard Boser, John~S Denker, Donnie Henderson, Richard~E Howard,
  Wayne Hubbard, and Lawrence~D Jackel.
\newblock Backpropagation applied to handwritten zip code recognition.
\newblock {\em Neural Computation}, 1(4):541--551, 1989.

\bibitem{lecun1998gradient}
Yann LeCun, L{\'e}on Bottou, Yoshua Bengio, and Patrick Haffner.
\newblock Gradient-based learning applied to document recognition.
\newblock {\em Proceedings of the IEEE}, 86(11):2278--2324, 1998.

\bibitem{lee2009convolutional}
Honglak Lee, Roger Grosse, Rajesh Ranganath, and Andrew~Y Ng.
\newblock Convolutional deep belief networks for scalable unsupervised learning
  of hierarchical representations.
\newblock In {\em Proceedings of the 26th Annual International Conference on
  Machine Learning}, pages 609--616. ACM, 2009.

\bibitem{lee2018gradient}
Yoonho Lee and Seungjin Choi.
\newblock Gradient-based meta-learning with learned layerwise metric and
  subspace.
\newblock In {\em International Conference on Machine Learning}, pages
  2927--2936. PMLR, 2018.

\bibitem{leibo2019autocurricula}
Joel~Z Leibo, Edward Hughes, Marc Lanctot, and Thore Graepel.
\newblock Autocurricula and the emergence of innovation from social
  interaction: A manifesto for multi-agent intelligence research.
\newblock {\em arXiv preprint arXiv:1903.00742}, 2019.

\bibitem{leibo2017multi}
Joel~Z Leibo, Vinicius Zambaldi, Marc Lanctot, Janusz Marecki, and Thore
  Graepel.
\newblock Multi-agent reinforcement learning in sequential social dilemmas.
\newblock In {\em Proceedings of the 16th Conference on Autonomous Agents and
  MultiAgent Systems, {AAMAS} 2017, S{\~{a}}o Paulo, Brazil}, pages 464--473,
  2017.

\bibitem{leiserson1992network}
Charles~E. Leiserson, Zahi~S. Abuhamdeh, David~C. Douglas, Carl~R. Feynman,
  Mahesh~N. Ganmukhi, Jeffrey~V. Hill, W.~Daniel Hillis, Bradley~C. Kuszmaul,
  Margaret A.~St. Pierre, David~S. Wells, Monica~C. Wong, Shaw-Wen Yang, and
  Robert~C. Zak.
\newblock The network architecture of the connection machine {CM-5}.
\newblock In {\em Proceedings of the fourth annual ACM Symposium on Parallel
  Algorithms and Architectures}, pages 272--285, 1992.

\bibitem{leonetti2016synthesis}
Matteo Leonetti, Luca Iocchi, and Peter Stone.
\newblock A synthesis of automated planning and reinforcement learning for
  efficient, robust decision-making.
\newblock {\em Artificial Intelligence}, 241:103--130, 2016.

\bibitem{levine2014learning}
Sergey Levine and Pieter Abbeel.
\newblock Learning neural network policies with guided policy search under
  unknown dynamics.
\newblock In {\em Advances in Neural Information Processing Systems}, pages
  1071--1079, 2014.

\bibitem{levine2013guided}
Sergey Levine and Vladlen Koltun.
\newblock Guided policy search.
\newblock In {\em International Conference on Machine Learning}, pages 1--9,
  2013.

\bibitem{levine2020offline}
Sergey Levine, Aviral Kumar, George Tucker, and Justin Fu.
\newblock Offline reinforcement learning: Tutorial, review, and perspectives on
  open problems.
\newblock {\em arXiv preprint arXiv:2005.01643}, 2020.

\bibitem{levy2019learning}
Andrew Levy, George Konidaris, Robert Platt, and Kate Saenko.
\newblock Learning multi-level hierarchies with hindsight.
\newblock In {\em International Conference on Learning Representations}, 2019.

\bibitem{li2017visualizing}
Hao Li, Zheng Xu, Gavin Taylor, Christoph Studer, and Tom Goldstein.
\newblock Visualizing the loss landscape of neural nets.
\newblock In {\em Advances in Neural Information Processing Systems}, pages
  6391--6401, 2018.

\bibitem{li2017learning}
Ke~Li and Jitendra Malik.
\newblock Learning to optimize neural nets.
\newblock {\em arXiv preprint arXiv:1703.00441}, 2017.

\bibitem{li2020deep}
Sheng Li, Jayesh~K Gupta, Peter Morales, Ross Allen, and Mykel~J Kochenderfer.
\newblock Deep implicit coordination graphs for multi-agent reinforcement
  learning.
\newblock In {\em {AAMAS} '21: 20th International Conference on Autonomous
  Agents and Multiagent Systems}, 2021.

\bibitem{li2019hierarchical}
Siyuan Li, Rui Wang, Minxue Tang, and Chongjie Zhang.
\newblock Hierarchical reinforcement learning with advantage-based auxiliary
  rewards.
\newblock In {\em Advances in Neural Information Processing Systems}, pages
  1407--1417, 2019.

\bibitem{li2017meta}
Zhenguo Li, Fengwei Zhou, Fei Chen, and Hang Li.
\newblock {Meta-SGD:} learning to learn quickly for few-shot learning.
\newblock {\em arXiv preprint arXiv:1707.09835}, 2017.

\bibitem{li2017efficient}
Zhuoru Li, Akshay Narayan, and Tze-Yun Leong.
\newblock An efficient approach to model-based hierarchical reinforcement
  learning.
\newblock In {\em Proceedings of the AAAI Conference on Artificial
  Intelligence}, volume~31, 2017.

\bibitem{liang2017rllib}
Eric Liang, Richard Liaw, Philipp Moritz, Robert Nishihara, Roy Fox, Ken
  Goldberg, Joseph~E Gonzalez, Michael~I Jordan, and Ion Stoica.
\newblock {RLlib:} abstractions for distributed reinforcement learning.
\newblock In {\em International Conference on Machine Learning}, pages
  3059--3068, 2018.

\bibitem{liebana2019general}
Diego~P{\'e}rez Li{\'e}bana, Simon~M Lucas, Raluca~D Gaina, Julian Togelius,
  Ahmed Khalifa, and Jialin Liu.
\newblock General video game artificial intelligence.
\newblock {\em Synthesis Lectures on Games and Computational Intelligence},
  3(2):1--191, 2019.

\bibitem{lillicrap2015continuous}
Timothy~P Lillicrap, Jonathan~J Hunt, Alexander Pritzel, Nicolas Heess, Tom
  Erez, Yuval Tassa, David Silver, and Daan Wierstra.
\newblock Continuous control with deep reinforcement learning.
\newblock In {\em International Conference on Learning Representations}, 2016.

\bibitem{lin1992self}
Long-Ji Lin.
\newblock Self-improving reactive agents based on reinforcement learning,
  planning and teaching.
\newblock {\em Machine Learning}, 8(3-4):293--321, 1992.

\bibitem{lin1993reinforcement}
Long-Ji Lin.
\newblock Reinforcement learning for robots using neural networks.
\newblock Technical report, Carnegie-Mellon Univ Pittsburgh PA School of
  Computer Science, 1993.

\bibitem{littman1994markov}
Michael~L Littman.
\newblock {Markov} games as a framework for multi-agent reinforcement learning.
\newblock In {\em Machine Learning Proceedings 1994}, pages 157--163. Elsevier,
  1994.

\bibitem{liu2014multiobjective}
Chunming Liu, Xin Xu, and Dewen Hu.
\newblock Multiobjective reinforcement learning: A comprehensive overview.
\newblock {\em IEEE Transactions on Systems, Man, and Cybernetics: Systems},
  45(3):385--398, 2014.

\bibitem{liu2020hybrid}
Hao Liu and Pieter Abbeel.
\newblock Hybrid discriminative-generative training via contrastive learning.
\newblock {\em arXiv preprint arXiv:2007.09070}, 2020.

\bibitem{liu2012sparse}
Hui Liu, Song Yu, Zhangxin Chen, Ben Hsieh, and Lei Shao.
\newblock Sparse matrix-vector multiplication on {NVIDIA GPU}.
\newblock {\em International Journal of Numerical Analysis \& Modeling, Series
  B}, 3(2):185--191, 2012.

\bibitem{liu2019emergent}
Siqi Liu, Guy Lever, Josh Merel, Saran Tunyasuvunakool, Nicolas Heess, and
  Thore Graepel.
\newblock Emergent coordination through competition.
\newblock In {\em International Conference on Learning Representations}, 2019.

\bibitem{lopez2016irace}
Manuel L{\'o}pez-Ib{\'a}{\~n}ez, J{\'e}r{\'e}mie Dubois-Lacoste,
  Leslie~P{\'e}rez C{\'a}ceres, Mauro Birattari, and Thomas St{\"u}tzle.
\newblock The irace package: Iterated racing for automatic algorithm
  configuration.
\newblock {\em Operations Research Perspectives}, 3:43--58, 2016.

\bibitem{lowe2017multi}
Ryan Lowe, Yi~Wu, Aviv Tamar, Jean Harb, Pieter Abbeel, and Igor Mordatch.
\newblock Multi-agent {Actor-Critic} for mixed cooperative-competitive
  environments.
\newblock In {\em Advances in Neural Information Processing Systems}, pages
  6379--6390, 2017.

\bibitem{loye2019}
Gabriel Loye.
\newblock The attention mechanism.
  \url{https://blog.floydhub.com/attention-mechanism/}.

\bibitem{ma2017power}
Siyuan Ma, Raef Bassily, and Mikhail Belkin.
\newblock The power of interpolation: Understanding the effectiveness of {SGD}
  in modern over-parametrized learning.
\newblock In {\em International Conference on Machine Learning}, pages
  3331--3340, 2018.

\bibitem{ma2018survey}
Xiaoliang Ma, Xiaodong Li, Qingfu Zhang, Ke~Tang, Zhengping Liang, Weixin Xie,
  and Zexuan Zhu.
\newblock A survey on cooperative co-evolutionary algorithms.
\newblock {\em IEEE Transactions on Evolutionary Computation}, 23(3):421--441,
  2018.

\bibitem{maaten2008visualizing}
Laurens van~der Maaten and Geoffrey Hinton.
\newblock Visualizing data using {t-SNE}.
\newblock {\em Journal of Machine Learning Research}, 9:2579--2605, Nov 2008.

\bibitem{machado2018revisiting}
Marlos~C Machado, Marc~G Bellemare, Erik Talvitie, Joel Veness, Matthew
  Hausknecht, and Michael Bowling.
\newblock Revisiting the arcade learning environment: Evaluation protocols and
  open problems for general agents.
\newblock {\em Journal of Artificial Intelligence Research}, 61:523--562, 2018.

\bibitem{maei2010toward}
Hamid~Reza Maei, Csaba Szepesv{\'a}ri, Shalabh Bhatnagar, and Richard~S Sutton.
\newblock Toward off-policy learning control with function approximation.
\newblock In {\em International Conference on Machine Learning}, 2010.

\bibitem{mahajan2019maven}
Anuj Mahajan, Tabish Rashid, Mikayel Samvelyan, and Shimon Whiteson.
\newblock Maven: Multi-agent variational exploration.
\newblock In {\em Advances in Neural Information Processing Systems}, pages
  7611--7622, 2019.

\bibitem{mahajan2018exploring}
Dhruv Mahajan, Ross Girshick, Vignesh Ramanathan, Kaiming He, Manohar Paluri,
  Yixuan Li, Ashwin Bharambe, and Laurens van~der Maaten.
\newblock Exploring the limits of weakly supervised pretraining.
\newblock In {\em Proceedings of the European Conference on Computer Vision
  (ECCV)}, pages 181--196, 2018.

\bibitem{khadka2019evolutionary}
Somdeb Majumdar, Shauharda Khadka, Santiago Miret, Stephen McAleer, and Kagan
  Tumer.
\newblock Evolutionary reinforcement learning for sample-efficient multiagent
  coordination.
\newblock In {\em International Conference on Machine Learning}, 2020.

\bibitem{makar2001hierarchical}
Rajbala Makar, Sridhar Mahadevan, and Mohammad Ghavamzadeh.
\newblock Hierarchical multi-agent reinforcement learning.
\newblock In {\em Proceedings of the Fifth International Conference on
  Autonomous Agents}, pages 246--253. ACM, 2001.

\bibitem{marewski2010good}
Julian~N Marewski, Wolfgang Gaissmaier, and Gerd Gigerenzer.
\newblock Good judgments do not require complex cognition.
\newblock {\em Cognitive Processing}, 11(2):103--121, 2010.

\bibitem{martinelli2019}
Vince Martinelli.
\newblock How robots autonomously see, grasp, and pick.
  \url{https://www.therobotreport.com/grasp-sight-picking-evolve-robots/},
  2019.

\bibitem{matiisen2017teacher}
Tambet Matiisen, Avital Oliver, Taco Cohen, and John Schulman.
\newblock Teacher-student curriculum learning.
\newblock {\em {IEEE} Trans. Neural Networks Learn. Syst.}, 31(9):3732--3740,
  2020.

\bibitem{matsugu2003subject}
Masakazu Matsugu, Katsuhiko Mori, Yusuke Mitari, and Yuji Kaneda.
\newblock Subject independent facial expression recognition with robust face
  detection using a convolutional neural network.
\newblock {\em Neural Networks}, 16(5-6):555--559, 2003.

\bibitem{matsuzaki2018empirical}
Kiminori Matsuzaki.
\newblock Empirical analysis of {PUCT} algorithm with evaluation functions of
  different quality.
\newblock In {\em 2018 Conference on Technologies and Applications of
  Artificial Intelligence (TAAI)}, pages 142--147. IEEE, 2018.

\bibitem{mayne2000constrained}
David~Q Mayne, James~B Rawlings, Christopher~V Rao, and Pierre~OM Scokaert.
\newblock Constrained model predictive control: Stability and optimality.
\newblock {\em Automatica}, 36(6):789--814, 2000.

\bibitem{mcclelland1995there}
James~L McClelland, Bruce~L McNaughton, and Randall~C O'Reilly.
\newblock Why there are complementary learning systems in the hippocampus and
  neocortex: insights from the successes and failures of connectionist models
  of learning and memory.
\newblock {\em Psychological Review}, 102(3):419, 1995.

\bibitem{melo2007convergence}
Francisco~S Melo and M~Isabel Ribeiro.
\newblock Convergence of {Q}-learning with linear function approximation.
\newblock In {\em 2007 European Control Conference (ECC)}, pages 2671--2678.
  IEEE, 2007.

\bibitem{merel2018hierarchical}
Josh Merel, Arun Ahuja, Vu~Pham, Saran Tunyasuvunakool, Siqi Liu, Dhruva
  Tirumala, Nicolas Heess, and Greg Wayne.
\newblock Hierarchical visuomotor control of humanoids.
\newblock In {\em International Conference on Learning Representations}, 2019.

\bibitem{merel2019deep}
Josh Merel, Diego Aldarondo, Jesse Marshall, Yuval Tassa, Greg Wayne, and Bence
  {\"O}lveczky.
\newblock Deep neuroethology of a virtual rodent.
\newblock In {\em International Conference on Learning Representations}, 2020.

\bibitem{merel2018neural}
Josh Merel, Leonard Hasenclever, Alexandre Galashov, Arun Ahuja, Vu~Pham, Greg
  Wayne, Yee~Whye Teh, and Nicolas Heess.
\newblock Neural probabilistic motor primitives for humanoid control.
\newblock In {\em International Conference on Learning Representations}, 2019.

\bibitem{merel2017learning}
Josh Merel, Yuval Tassa, Dhruva TB, Sriram Srinivasan, Jay Lemmon, Ziyu Wang,
  Greg Wayne, and Nicolas Heess.
\newblock Learning human behaviors from motion capture by adversarial
  imitation.
\newblock {\em arXiv preprint arXiv:1707.02201}, 2017.

\bibitem{miikkulainen2019evolving}
Risto Miikkulainen, Jason Liang, Elliot Meyerson, Aditya Rawal, Daniel Fink,
  Olivier Francon, Bala Raju, Hormoz Shahrzad, Arshak Navruzyan, Nigel Duffy,
  and Babak Hodjat.
\newblock Evolving deep neural networks.
\newblock In {\em Artificial Intelligence in the Age of Neural Networks and
  Brain Computing}, pages 293--312. Elsevier, 2019.

\bibitem{mikolov2013efficient}
Tomas Mikolov, Kai Chen, Greg Corrado, and Jeffrey Dean.
\newblock Efficient estimation of word representations in vector space.
\newblock In {\em International Conference on Learning Representations}, 2013.

\bibitem{mikolov2013distributed}
Tomas Mikolov, Ilya Sutskever, Kai Chen, Greg Corrado, and Jeffrey Dean.
\newblock Distributed representations of words and phrases and their
  compositionality.
\newblock In {\em Advances in Neural Information Processing Systems}, 2013.

\bibitem{millen1981programming}
Jonathan~K Millen.
\newblock Programming the game of {Go}.
\newblock {\em Byte Magazine}, 1981.

\bibitem{mirsoleimani2015scaling}
S~Ali Mirsoleimani, Aske Plaat, Jaap Van Den~Herik, and Jos Vermaseren.
\newblock Scaling {Monte Carlo} tree search on {Intel} {Xeon} {Phi}.
\newblock In {\em Parallel and Distributed Systems (ICPADS), 2015 IEEE 21st
  International Conference on}, pages 666--673. IEEE, 2015.

\bibitem{mishra2017simple}
Nikhil Mishra, Mostafa Rohaninejad, Xi~Chen, and Pieter Abbeel.
\newblock A simple neural attentive meta-learner.
\newblock In {\em International Conference on Learning Representations}, 2018.

\bibitem{mitchell1980need}
Tom~M Mitchell.
\newblock The need for biases in learning generalizations.
\newblock Technical Report CBM-TR-117, Department of Computer Science, Rutgers
  University, 1980.

\bibitem{mitchell2006discipline}
Tom~M Mitchell.
\newblock The discipline of machine learning.
\newblock Technical Report CMU-ML-06-108, Carnegie Mellon University, School of
  Computer Science, Machine Learning, 2006.

\bibitem{mittel2019visual}
Akshita Mittel and Purna Sowmya~Munukutla.
\newblock Visual transfer between {Atari} games using competitive reinforcement
  learning.
\newblock In {\em Proceedings of the IEEE/CVF Conference on Computer Vision and
  Pattern Recognition Workshops}, pages 0--0, 2019.

\bibitem{mnih2016asynchronous}
Volodymyr Mnih, Adria~Puigdomenech Badia, Mehdi Mirza, Alex Graves, Timothy
  Lillicrap, Tim Harley, David Silver, and Koray Kavukcuoglu.
\newblock Asynchronous methods for deep reinforcement learning.
\newblock In {\em International Conference on Machine Learning}, pages
  1928--1937, 2016.

\bibitem{mnih2013playing}
Volodymyr Mnih, Koray Kavukcuoglu, David Silver, Alex Graves, Ioannis
  Antonoglou, Daan Wierstra, and Martin Riedmiller.
\newblock Playing {A}tari with deep reinforcement learning.
\newblock {\em arXiv preprint arXiv:1312.5602}, 2013.

\bibitem{mnih2015human}
Volodymyr Mnih, Koray Kavukcuoglu, David Silver, Andrei~A. Rusu, Joel Veness,
  Marc~G. Bellemare, Alex Graves, Martin~A. Riedmiller, Andreas Fidjeland,
  Georg Ostrovski, Stig Petersen, Charles Beattie, Amir Sadik, Ioannis
  Antonoglou, Helen King, Dharshan Kumaran, Daan Wierstra, Shane Legg, and
  Demis Hassabis.
\newblock Human-level control through deep reinforcement learning.
\newblock {\em Nature}, 518(7540):529--533, 2015.

\bibitem{moerland2021lecture}
Thomas Moerland.
\newblock Continuous {Markov} decision process and policy search.
\newblock Lecture notes for the course reinforcement learning, Leiden
  University, 2021.

\bibitem{moerland2021intersection}
Thomas~M Moerland.
\newblock {\em The Intersection of Planning and Learning}.
\newblock PhD thesis, Delft University of Technology, 2021.

\bibitem{moerland2017efficient}
Thomas~M Moerland, Joost Broekens, and Catholijn~M Jonker.
\newblock Efficient exploration with double uncertain value networks.
\newblock {\em arXiv preprint arXiv:1711.10789}, 2017.

\bibitem{moerland2018potential}
Thomas~M Moerland, Joost Broekens, and Catholijn~M Jonker.
\newblock The potential of the return distribution for exploration in {RL}.
\newblock {\em arXiv preprint arXiv:1806.04242}, 2018.

\bibitem{moerland2020framework}
Thomas~M Moerland, Joost Broekens, and Catholijn~M Jonker.
\newblock A framework for reinforcement learning and planning.
\newblock {\em arXiv preprint arXiv:2006.15009}, 2020.

\bibitem{moerland2020model}
Thomas~M Moerland, Joost Broekens, and Catholijn~M Jonker.
\newblock Model-based reinforcement learning: A survey.
\newblock {\em arXiv preprint arXiv:2006.16712}, 2020.

\bibitem{moerland2018a0c}
Thomas~M Moerland, Joost Broekens, Aske Plaat, and Catholijn~M Jonker.
\newblock {A0C}: Alpha zero in continuous action space.
\newblock {\em arXiv preprint arXiv:1805.09613}, 2018.

\bibitem{moerland2018monte}
Thomas~M Moerland, Joost Broekens, Aske Plaat, and Catholijn~M Jonker.
\newblock {Monte Carlo} tree search for asymmetric trees.
\newblock {\em arXiv preprint arXiv:1805.09218}, 2018.

\bibitem{moore1990efficient}
Andrew~William Moore.
\newblock Efficient memory-based learning for robot control.
\newblock Technical Report UCAM-CL-TR-209, University of Cambridge, UK,
  \url{https://www.cl.cam.ac.uk/techreports/UCAM-CL-TR-209.pdf}, 1990.

\bibitem{moravvcik2017deepstack}
Matej Morav{\v{c}}{\'\i}k, Martin Schmid, Neil Burch, Viliam Lis{\`y}, Dustin
  Morrill, Nolan Bard, Trevor Davis, Kevin Waugh, Michael Johanson, and Michael
  Bowling.
\newblock Deepstack: Expert-level artificial intelligence in heads-up no-limit
  poker.
\newblock {\em Science}, 356(6337):508--513, 2017.

\bibitem{mordatch2018emergence}
Igor Mordatch and Pieter Abbeel.
\newblock Emergence of grounded compositional language in multi-agent
  populations.
\newblock In {\em Proceedings of the AAAI Conference on Artificial
  Intelligence}, volume~32, 2018.

\bibitem{moreno2021neural}
Pol Moreno, Edward Hughes, Kevin~R McKee, Bernardo~Avila Pires, and
  Th{\'e}ophane Weber.
\newblock Neural recursive belief states in multi-agent reinforcement learning.
\newblock {\em arXiv preprint arXiv:2102.02274}, 2021.

\bibitem{moriarty1999evolutionary}
David~E Moriarty, Alan~C Schultz, and John~J Grefenstette.
\newblock Evolutionary algorithms for reinforcement learning.
\newblock {\em Journal of Artificial Intelligence Research}, 11:241--276, 1999.

\bibitem{mossalam2016multi}
Hossam Mossalam, Yannis~M Assael, Diederik~M Roijers, and Shimon Whiteson.
\newblock Multi-objective deep reinforcement learning.
\newblock {\em arXiv preprint arXiv:1610.02707}, 2016.

\bibitem{mujtaba2020}
Hussain Mujtaba.
\newblock Introduction to autoencoders.
  \url{https://www.mygreatlearning.com/blog/autoencoder/}, 2020.

\bibitem{mullainathan2000behavioral}
Sendhil Mullainathan and Richard~H Thaler.
\newblock Behavioral economics.
\newblock Technical report, National Bureau of Economic Research, 2000.

\bibitem{muller2002computer}
Martin M{\"u}ller.
\newblock Computer {Go}.
\newblock {\em Artificial Intelligence}, 134(1-2):145--179, 2002.

\bibitem{brockhausen2021procedural}
Matthias M{\"u}ller-Brockhausen, Mike Preuss, and Aske Plaat.
\newblock Procedural content generation: Better benchmarks for transfer
  reinforcement learning.
\newblock In {\em Conference on Games}, 2021.

\bibitem{munkhdalai2017meta}
Tsendsuren Munkhdalai and Hong Yu.
\newblock Meta networks.
\newblock In {\em International Conference on Machine Learning}, pages
  2554--2563. PMLR, 2017.

\bibitem{murase1996automatic}
Yoshio Murase, Hitoshi Matsubara, and Yuzuru Hiraga.
\newblock Automatic making of {S}okoban problems.
\newblock In {\em Pacific Rim International Conference on Artificial
  Intelligence}, pages 592--600. Springer, 1996.

\bibitem{neptune2021}
Derick Mwiti.
\newblock Reinforcement learning applications.
  \url{https://neptune.ai/blog/reinforcement-learning-applications}.

\bibitem{myerson2013game}
Roger~B Myerson.
\newblock {\em Game Theory}.
\newblock Harvard university press, 2013.

\bibitem{nachum2018data}
Ofir Nachum, Shixiang Gu, Honglak Lee, and Sergey Levine.
\newblock Data-efficient hierarchical reinforcement learning.
\newblock In {\em Advances in Neural Information Processing Systems}, pages
  3307--3317, 2018.

\bibitem{nachum2017bridging}
Ofir Nachum, Mohammad Norouzi, Kelvin Xu, and Dale Schuurmans.
\newblock Bridging the gap between value and policy based reinforcement
  learning.
\newblock In {\em Advances in Neural Information Processing Systems}, pages
  2775--2785, 2017.

\bibitem{nadkarni2011natural}
Prakash~M Nadkarni, Lucila Ohno-Machado, and Wendy~W Chapman.
\newblock Natural language processing: an introduction.
\newblock {\em Journal of the American Medical Informatics Association},
  18(5):544--551, 2011.

\bibitem{nagabandi2018neural}
Anusha Nagabandi, Gregory Kahn, Ronald~S Fearing, and Sergey Levine.
\newblock Neural network dynamics for model-based deep reinforcement learning
  with model-free fine-tuning.
\newblock In {\em 2018 IEEE International Conference on Robotics and Automation
  (ICRA)}, pages 7559--7566, 2018.

\bibitem{nakkiran2019deep}
Preetum Nakkiran, Gal Kaplun, Yamini Bansal, Tristan Yang, Boaz Barak, and Ilya
  Sutskever.
\newblock Deep double descent: Where bigger models and more data.
\newblock In {\em 8th International Conference on Learning Representations,
  {ICLR} 2020, Addis Ababa, Ethiopia}, 2020.

\bibitem{nardelli2018value}
Nantas Nardelli, Gabriel Synnaeve, Zeming Lin, Pushmeet Kohli, Philip~HS Torr,
  and Nicolas Usunier.
\newblock Value propagation networks.
\newblock In {\em 7th International Conference on Learning Representations,
  {ICLR} 2019, New Orleans, LA, USA, May 6-9, 2019}, 2018.

\bibitem{narvekar2020curriculum}
Sanmit Narvekar, Bei Peng, Matteo Leonetti, Jivko Sinapov, Matthew~E Taylor,
  and Peter Stone.
\newblock Curriculum learning for reinforcement learning domains: A framework
  and survey.
\newblock {\em Journal Machine Learning Research}, 2020.

\bibitem{nasar2011beautiful}
Sylvia Nasar.
\newblock {\em A Beautiful Mind}.
\newblock Simon and Schuster, 2011.

\bibitem{nash1951non}
John Nash.
\newblock Non-cooperative games.
\newblock {\em Annals of mathematics}, pages 286--295, 1951.

\bibitem{nash1950equilibrium}
John~F Nash.
\newblock Equilibrium points in $n$-person games.
\newblock {\em Proceedings of the National Academy of Sciences}, 36(1):48--49,
  1950.

\bibitem{nash1950bargaining}
John~F Nash~Jr.
\newblock The bargaining problem.
\newblock {\em Econometrica: Journal of the econometric society}, pages
  155--162, 1950.

\bibitem{nasu2018efficiently}
Yu~Nasu.
\newblock Efficiently updatable neural-network-based evaluation functions for
  computer shogi.
\newblock {\em The 28th World Computer Shogi Championship Appeal Document},
  2018.

\bibitem{neapolitan2004learning}
Richard~E Neapolitan.
\newblock {\em Learning {Bayesian} networks}.
\newblock Pearson Prentice Hall, Upper Saddle River, NJ, 2004.

\bibitem{ng2004feature}
Andrew~Y Ng.
\newblock Feature selection, {L1 vs. L2} regularization, and rotational
  invariance.
\newblock In {\em Proceedings of the Twenty-first International Conference on
  Machine Learning}, page~78. ACM, 2004.

\bibitem{ng1999policy}
Andrew~Y Ng, Daishi Harada, and Stuart Russell.
\newblock Policy invariance under reward transformations: Theory and
  application to reward shaping.
\newblock In {\em International Conference on Machine Learning}, volume~99,
  pages 278--287, 1999.

\bibitem{nichol2018first}
Alex Nichol, Joshua Achiam, and John Schulman.
\newblock On first-order meta-learning algorithms.
\newblock {\em arXiv preprint arXiv:1803.02999}, 2018.

\bibitem{niu2018generalized}
Sufeng Niu, Siheng Chen, Hanyu Guo, Colin Targonski, Melissa~C Smith, and
  Jelena Kova{\v{c}}evi{\'c}.
\newblock Generalized value iteration networks: Life beyond lattices.
\newblock In {\em Thirty-Second AAAI Conference on Artificial Intelligence},
  2018.

\bibitem{oh2015action}
Junhyuk Oh, Xiaoxiao Guo, Honglak Lee, Richard~L Lewis, and Satinder Singh.
\newblock Action-conditional video prediction using deep networks in {A}tari
  games.
\newblock In {\em Advances in Neural Information Processing Systems}, pages
  2863--2871, 2015.

\bibitem{oh2017value}
Junhyuk Oh, Satinder Singh, and Honglak Lee.
\newblock Value prediction network.
\newblock In {\em Advances in Neural Information Processing Systems}, pages
  6118--6128, 2017.

\bibitem{oh2017zero}
Junhyuk Oh, Satinder Singh, Honglak Lee, and Pushmeet Kohli.
\newblock Zero-shot task generalization with multi-task deep reinforcement
  learning.
\newblock In {\em International Conference on Machine Learning}, pages
  2661--2670. PMLR, 2017.

\bibitem{oh2004gpu}
Kyoung-Su Oh and Keechul Jung.
\newblock {GPU} implementation of neural networks.
\newblock {\em Pattern Recognition}, 37(6):1311--1314, 2004.

\bibitem{colah2015}
Chris Olah.
\newblock Understanding {LSTM} networks.
  \url{http://colah.github.io/posts/2015-08-Understanding-LSTMs/}, 2015.

\bibitem{oliehoek2012decentralized}
Frans~A Oliehoek.
\newblock Decentralized {POMDPs}.
\newblock In {\em Reinforcement Learning}, pages 471--503. Springer, 2012.

\bibitem{oliehoek2016concise}
Frans~A Oliehoek and Christopher Amato.
\newblock {\em A Concise Introduction to Decentralized POMDPs}.
\newblock Springer, 2016.

\bibitem{oliehoek2013incremental}
Frans~A Oliehoek, Matthijs~TJ Spaan, Christopher Amato, and Shimon Whiteson.
\newblock Incremental clustering and expansion for faster optimal planning in
  {Dec-POMDPs}.
\newblock {\em Journal of Artificial Intelligence Research}, 46:449--509, 2013.

\bibitem{omidshafiei2017deep}
Shayegan Omidshafiei, Jason Pazis, Christopher Amato, Jonathan~P How, and John
  Vian.
\newblock Deep decentralized multi-task multi-agent reinforcement learning
  under partial observability.
\newblock In {\em International Conference on Machine Learning}, pages
  2681--2690. PMLR, 2017.

\bibitem{o2010play}
Joseph O'Neill, Barty Pleydell-Bouverie, David Dupret, and Jozsef Csicsvari.
\newblock Play it again: reactivation of waking experience and memory.
\newblock {\em Trends in Neurosciences}, 33(5):220--229, 2010.

\bibitem{ontanon2013survey}
Santiago Ontan{\'o}n, Gabriel Synnaeve, Alberto Uriarte, Florian Richoux, David
  Churchill, and Mike Preuss.
\newblock A survey of real-time strategy game {AI} research and competition in
  {StarCraft}.
\newblock {\em IEEE Transactions on Computational Intelligence and AI in
  Games}, 5(4):293--311, 2013.

\bibitem{opitz1999popular}
David Opitz and Richard Maclin.
\newblock Popular ensemble methods: An empirical study.
\newblock {\em Journal of Artificial Intelligence Research}, 11:169--198, 1999.

\bibitem{osband2019behaviour}
Ian Osband, Yotam Doron, Matteo Hessel, John Aslanides, Eren Sezener, Andre
  Saraiva, Katrina McKinney, Tor Lattimore, Csaba Szepesv{\'{a}}ri, Satinder
  Singh, Benjamin~Van Roy, Richard~S. Sutton, David Silver, and Hado van
  Hasselt.
\newblock Behaviour suite for reinforcement learning.
\newblock In {\em 8th International Conference on Learning Representations,
  {ICLR} 2020, Addis Ababa, Ethiopia}, April 2020.

\bibitem{oudeyer2008can}
Pierre-Yves Oudeyer and Frederic Kaplan.
\newblock How can we define intrinsic motivation?
\newblock In {\em the 8th International Conference on Epigenetic Robotics:
  Modeling Cognitive Development in Robotic Systems}. Lund University Cognitive
  Studies, Lund: LUCS, Brighton, 2008.

\bibitem{oudeyer2009intrinsic}
Pierre-Yves Oudeyer and Frederic Kaplan.
\newblock What is intrinsic motivation? {A} typology of computational
  approaches.
\newblock {\em Frontiers in Neurorobotics}, 1:6, 2009.

\bibitem{oudeyer2007intrinsic}
Pierre-Yves Oudeyer, Frederic Kaplan, and Verena~V Hafner.
\newblock Intrinsic motivation systems for autonomous mental development.
\newblock {\em IEEE Transactions on Evolutionary Computation}, 11(2):265--286,
  2007.

\bibitem{packer2018assessing}
Charles Packer, Katelyn Gao, Jernej Kos, Philipp Kr{\"a}henb{\"u}hl, Vladlen
  Koltun, and Dawn Song.
\newblock Assessing generalization in deep reinforcement learning.
\newblock {\em arXiv preprint arXiv:1810.12282}, 2018.

\bibitem{palatucci2009zero}
Mark~M Palatucci, Dean~A Pomerleau, Geoffrey~E Hinton, and Tom Mitchell.
\newblock Zero-shot learning with semantic output codes.
\newblock In {\em Advances in Neural Information Processing Systems 22}, 2009.

\bibitem{pan2010survey}
Sinno~Jialin Pan and Qiang Yang.
\newblock A survey on transfer learning.
\newblock {\em IEEE Transactions on Knowledge and Data Engineering},
  22(10):1345--1359, 2010.

\bibitem{panov2018automatic}
Aleksandr~I Panov and Aleksey Skrynnik.
\newblock Automatic formation of the structure of abstract machines in
  hierarchical reinforcement learning with state clustering.
\newblock {\em arXiv preprint arXiv:1806.05292}, 2018.

\bibitem{paparo2014quantum}
Giuseppe~Davide Paparo, Vedran Dunjko, Adi Makmal, Miguel~Angel Martin-Delgado,
  and Hans~J Briegel.
\newblock Quantum speedup for active learning agents.
\newblock {\em Physical Review X}, 4(3):031002, 2014.

\bibitem{paquette2019no}
Philip Paquette, Yuchen Lu, Steven Bocco, Max Smith, O-G Satya, Jonathan~K
  Kummerfeld, Joelle Pineau, Satinder Singh, and Aaron~C Courville.
\newblock No-press diplomacy: Modeling multi-agent gameplay.
\newblock In {\em Advances in Neural Information Processing Systems}, pages
  4476--4487, 2019.

\bibitem{parisi2019continual}
German~I Parisi, Ronald Kemker, Jose~L Part, Christopher Kanan, and Stefan
  Wermter.
\newblock Continual lifelong learning with neural networks: A review.
\newblock {\em Neural Networks}, 113:54--71, 2019.

\bibitem{parisotto2015actor}
Emilio Parisotto, Jimmy~Lei Ba, and Ruslan Salakhutdinov.
\newblock Actor-mimic: Deep multitask and transfer reinforcement learning.
\newblock {\em arXiv preprint arXiv:1511.06342}, 2015.

\bibitem{parisotto2020stabilizing}
Emilio Parisotto, Francis Song, Jack Rae, Razvan Pascanu, Caglar Gulcehre,
  Siddhant Jayakumar, Max Jaderberg, Raphael~Lopez Kaufman, Aidan Clark, Seb
  Noury, et~al.
\newblock Stabilizing transformers for reinforcement learning.
\newblock In {\em International Conference on Machine Learning}, pages
  7487--7498. PMLR, 2020.

\bibitem{parr1998reinforcement}
Ronald Parr and Stuart~J Russell.
\newblock Reinforcement learning with hierarchies of machines.
\newblock In {\em Advances in Neural Information Processing Systems}, pages
  1043--1049, 1998.

\bibitem{pascutto2017leela}
Gian-Carlo Pascutto.
\newblock Leela zero. \url{https://github.com/leela-zero/leela-zero}, 2017.

\bibitem{pashevich2018modulated}
Alexander Pashevich, Danijar Hafner, James Davidson, Rahul Sukthankar, and
  Cordelia Schmid.
\newblock Modulated policy hierarchies.
\newblock {\em arXiv preprint arXiv:1812.00025}, 2018.

\bibitem{paszke2019pytorch}
Adam Paszke, Sam Gross, Francisco Massa, Adam Lerer, James Bradbury, Gregory
  Chanan, Trevor Killeen, Zeming Lin, Natalia Gimelshein, Luca Antiga, Alban
  Desmaison, Andreas K{\"{o}}pf, Edward~Z. Yang, Zachary DeVito, Martin Raison,
  Alykhan Tejani, Sasank Chilamkurthy, Benoit Steiner, Lu~Fang, Junjie Bai, and
  Soumith Chintala.
\newblock Pytorch: An imperative style, high-performance deep learning library.
\newblock In {\em Advances in Neural Information Processing Systems}, pages
  8024--8035, 2019.

\bibitem{pateria2021hierarchical}
Shubham Pateria, Budhitama Subagdja, Ah-hweewee Tan, and Chai Quek.
\newblock Hierarchical reinforcement learning: A comprehensive survey.
\newblock {\em ACM Computing Surveys (CSUR)}, 54(5):1--35, 2021.

\bibitem{pearl1984heuristics}
Judea Pearl.
\newblock {\em Heuristics: Intelligent Search Strategies for Computer Problem
  Solving}.
\newblock Addison-Wesley, Reading, MA, 1984.

\bibitem{pearl2018book}
Judea Pearl and Dana Mackenzie.
\newblock {\em The Book of Why: the New Science of Cause and Effect}.
\newblock Basic Books, 2018.

\bibitem{pedregosa2011scikit}
Fabian Pedregosa, Ga{\"{e}}l Varoquaux, Alexandre Gramfort, Vincent Michel,
  Bertrand Thirion, Olivier Grisel, Mathieu Blondel, Peter Prettenhofer, Ron
  Weiss, Vincent Dubourg, Jake VanderPlas, Alexandre Passos, David Cournapeau,
  Matthieu Brucher, Matthieu Perrot, and Edouard Duchesnay.
\newblock Scikit-learn: Machine learning in {Python}.
\newblock {\em Journal of Machine Learning Research}, 12(Oct):2825--2830, 2011.

\bibitem{pennington2014glove}
Jeffrey Pennington, Richard Socher, and Christopher~D Manning.
\newblock Glove: Global vectors for word representation.
\newblock In {\em Proceedings of the 2014 Conference on Empirical Methods in
  Natural Language Processing (EMNLP)}, pages 1532--1543, 2014.

\bibitem{pere2018unsupervised}
Alexandre P{\'e}r{\'e}, S{\'e}bastien Forestier, Olivier Sigaud, and
  Pierre-Yves Oudeyer.
\newblock Unsupervised learning of goal spaces for intrinsically motivated goal
  exploration.
\newblock In {\em International Conference on Learning Representations}, 2018.

\bibitem{pertsch2020long}
Karl Pertsch, Oleh Rybkin, Frederik Ebert, Shenghao Zhou, Dinesh Jayaraman,
  Chelsea Finn, and Sergey Levine.
\newblock Long-horizon visual planning with goal-conditioned hierarchical
  predictors.
\newblock In {\em Advances in Neural Information Processing Systems}, 2020.

\bibitem{plaat2010vlinder}
Aske Plaat.
\newblock {\em De vlinder en de mier / The butterfly and the ant---on modeling
  behavior in organizations}.
\newblock Inaugural lecture. Tilburg University, 2010.

\bibitem{plaat2020learning}
Aske Plaat.
\newblock {\em Learning to Play: Reinforcement Learning and Games}.
\newblock Springer Verlag, Heidelberg, \url{https://learningtoplay.net}, 2020.

\bibitem{plaat2021high}
Aske Plaat, Walter Kosters, and Mike Preuss.
\newblock High-accuracy model-based reinforcement learning, a survey.
\newblock {\em arXiv preprint arXiv:2107.08241}, 2021.

\bibitem{plaat1996best}
Aske Plaat, Jonathan Schaeffer, Wim Pijls, and Arie De~Bruin.
\newblock Best-first fixed-depth minimax algorithms.
\newblock {\em Artificial Intelligence}, 87(1-2):255--293, 1996.

\bibitem{plappert2016kerasrl}
Matthias Plappert.
\newblock {Keras-RL}.
\newblock \url{https://github.com/keras-rl/keras-rl}, 2016.

\bibitem{pollack1997did}
Jordan~B Pollack and Alan~D Blair.
\newblock Why did {TD-gammon} work?
\newblock In {\em Advances in Neural Information Processing Systems}, pages
  10--16, 1997.

\bibitem{prasaf2018lessons}
Aditya Prasad.
\newblock Lessons from implementing alphazero
  \url{https://medium.com/oracledevs/lessons-from-implementing-alphazero-7e36e9054191},
  2018.

\bibitem{pratt1993discriminability}
Lorien~Y Pratt.
\newblock Discriminability-based transfer between neural networks.
\newblock In {\em Advances in Neural Information Processing Systems}, pages
  204--211, 1993.

\bibitem{prechelt1998automatic}
Lutz Prechelt.
\newblock Automatic early stopping using cross validation: quantifying the
  criteria.
\newblock {\em Neural Networks}, 11(4):761--767, 1998.

\bibitem{prechelt1998early}
Lutz Prechelt.
\newblock Early stopping-but when?
\newblock In {\em Neural Networks: Tricks of the trade}, pages 55--69.
  Springer, 1998.

\bibitem{precup1997planning}
Doina Precup, Richard~S Sutton, and Satinder~P Singh.
\newblock Planning with closed-loop macro actions.
\newblock In {\em Working notes of the 1997 AAAI Fall Symposium on
  Model-directed Autonomous Systems}, pages 70--76, 1997.

\bibitem{premack1978does}
David Premack and Guy Woodruff.
\newblock Does the chimpanzee have a theory of mind?
\newblock {\em Behavioral and Brain Sciences}, 1(4):515--526, 1978.

\bibitem{proencca2020interpretable}
Hugo~M Proen{\c{c}}a and Matthijs van Leeuwen.
\newblock Interpretable multiclass classification by mdl-based rule lists.
\newblock {\em Information Sciences}, 512:1372--1393, 2020.

\bibitem{pumperla2019}
Max Pumperla and Kevin Ferguson.
\newblock {\em Deep Learning and the Game of {Go}}.
\newblock Manning, 2019.

\bibitem{quinlan1983learning}
J~Ross Quinlan.
\newblock Learning efficient classification procedures and their application to
  chess end games.
\newblock In {\em Machine Learning}, pages 463--482. Springer, 1983.

\bibitem{quinlan1986induction}
J~Ross Quinlan.
\newblock Induction of decision trees.
\newblock {\em Machine Learning}, 1(1):81--106, 1986.

\bibitem{racaniere2017imagination}
S{\'{e}}bastien Racani{\`{e}}re, Theophane Weber, David~P. Reichert, Lars
  Buesing, Arthur Guez, Danilo~Jimenez Rezende, Adri{\`{a}}~Puigdom{\`{e}}nech
  Badia, Oriol Vinyals, Nicolas Heess, Yujia Li, Razvan Pascanu, Peter~W.
  Battaglia, Demis Hassabis, David Silver, and Daan Wierstra.
\newblock Imagination-augmented agents for deep reinforcement learning.
\newblock In {\em Advances in Neural Information Processing Systems}, pages
  5690--5701, 2017.

\bibitem{radford2021learning}
Alec Radford, Jong~Wook Kim, Chris Hallacy, Aditya Ramesh, Gabriel Goh,
  Sandhini Agarwal, Girish Sastry, Amanda Askell, Pamela Mishkin, Jack Clark,
  Gretchen Krueger, and Ilya Sutskever.
\newblock Learning transferable visual models from natural language
  supervision.
\newblock In {\em International Conference on Machine Learning}, 2021.

\bibitem{radford2018improving}
Alec Radford, Karthik Narasimhan, Tim Salimans, and Ilya Sutskever.
\newblock Improving language understanding by generative pre-training.
  \url{https://openai.com/blog/language-unsupervised/}, 2018.

\bibitem{ruadulescu2020multi}
Roxana R{\u{a}}dulescu, Patrick Mannion, Diederik~M Roijers, and Ann Now{\'e}.
\newblock Multi-objective multi-agent decision making: a utility-based analysis
  and survey.
\newblock {\em Autonomous Agents and Multi-Agent Systems}, 34(1):1--52, 2020.

\bibitem{rafati2019learning}
Jacob Rafati and David~C Noelle.
\newblock Learning representations in model-free hierarchical reinforcement
  learning.
\newblock In {\em Proceedings of the AAAI Conference on Artificial
  Intelligence}, volume~33, pages 10009--10010, 2019.

\bibitem{raghu2019rapid}
Aniruddh Raghu, Maithra Raghu, Samy Bengio, and Oriol Vinyals.
\newblock Rapid learning or feature reuse? {T}owards understanding the
  effectiveness of maml.
\newblock In {\em International Conference on Learning Representations}, 2020.

\bibitem{raileanu2020ride}
Roberta Raileanu and Tim Rockt{\"a}schel.
\newblock {RIDE:} rewarding impact-driven exploration for
  procedurally-generated environments.
\newblock In {\em International Conference on Learning Representations}, 2020.

\bibitem{raina2006constructing}
Rajat Raina, Andrew~Y Ng, and Daphne Koller.
\newblock Constructing informative priors using transfer learning.
\newblock In {\em Proceedings of the 23rd international conference on Machine
  learning}, pages 713--720, 2006.

\bibitem{rajeswaran2019meta}
Aravind Rajeswaran, Chelsea Finn, Sham Kakade, and Sergey Levine.
\newblock Meta-learning with implicit gradients.
\newblock In {\em Advances in Neural Information Processing Systems}, 2019.

\bibitem{rakelly2019efficient}
Kate Rakelly, Aurick Zhou, Chelsea Finn, Sergey Levine, and Deirdre Quillen.
\newblock Efficient off-policy meta-reinforcement learning via probabilistic
  context variables.
\newblock In {\em International Conference on Machine Learning}, pages
  5331--5340. PMLR, 2019.

\bibitem{ramesh2021zero}
Aditya Ramesh, Mikhail Pavlov, Gabriel Goh, Scott Gray, Chelsea Voss, Alec
  Radford, Mark Chen, and Ilya Sutskever.
\newblock Zero-shot text-to-image generation.
\newblock In {\em International Conference on Machine Learning}, 2021.

\bibitem{randlov1999learning}
Jette Randlov.
\newblock Learning macro-actions in reinforcement learning.
\newblock In {\em Advances in Neural Information Processing Systems}, pages
  1045--1051, 1998.

\bibitem{nevergrad}
J.~Rapin and O.~Teytaud.
\newblock {Nevergrad - A gradient-free optimization platform}.
\newblock \url{https://GitHub.com/FacebookResearch/Nevergrad}, 2018.

\bibitem{rashid2018qmix}
Tabish Rashid, Mikayel Samvelyan, Christian Schroeder, Gregory Farquhar, Jakob
  Foerster, and Shimon Whiteson.
\newblock Qmix: Monotonic value function factorisation for deep multi-agent
  reinforcement learning.
\newblock In {\em International Conference on Machine Learning}, pages
  4295--4304. PMLR, 2018.

\bibitem{ravi2016optimization}
Sachin Ravi and Hugo Larochelle.
\newblock Optimization as a model for few-shot learning.
\newblock In {\em International Conference on Learning Representations}, 2017.

\bibitem{rice1976algorithm}
John~R. Rice.
\newblock The algorithm selection problem.
\newblock {\em Advances in Computers}, 15(65-118):5, 1976.

\bibitem{riedmiller2005neural}
Martin Riedmiller.
\newblock Neural fitted {Q} iteration---first experiences with a data efficient
  neural reinforcement learning method.
\newblock In {\em European Conference on Machine Learning}, pages 317--328.
  Springer, 2005.

\bibitem{robbins1952some}
Herbert Robbins.
\newblock Some aspects of the sequential design of experiments.
\newblock {\em Bulletin of the American Mathematical Society}, 58(5):527--535,
  1952.

\bibitem{roder2020curious}
Frank R{\"o}der, Manfred Eppe, Phuong~DH Nguyen, and Stefan Wermter.
\newblock Curious hierarchical actor-critic reinforcement learning.
\newblock In {\em International Conference on Artificial Neural Networks},
  pages 408--419. Springer, 2020.

\bibitem{roijers2021following}
Diederik~M Roijers, Willem R{\"o}pke, Ann Now{\'e}, and Roxana R{\u{a}}dulescu.
\newblock On following pareto-optimal policies in multi-objective planning and
  reinforcement learning.
\newblock In {\em Multi-Objective Decision Making Workshop}, 2021.

\bibitem{roijers2013survey}
Diederik~M Roijers, Peter Vamplew, Shimon Whiteson, and Richard Dazeley.
\newblock A survey of multi-objective sequential decision-making.
\newblock {\em Journal of Artificial Intelligence Research}, 48:67--113, 2013.

\bibitem{romera2015embarrassingly}
Bernardino Romera-Paredes and Philip Torr.
\newblock An embarrassingly simple approach to zero-shot learning.
\newblock In {\em International Conference on Machine Learning}, pages
  2152--2161, 2015.

\bibitem{ropke2021communication}
Willem R{\"o}pke, Roxana Radulescu, Diederik~M Roijers, and Ann Ann~Now{\'{e}}.
\newblock Communication strategies in multi-objective normal-form games.
\newblock In {\em Adaptive and Learning Agents Workshop 2021}, 2021.

\bibitem{rosin2011multi}
Christopher~D Rosin.
\newblock Multi-armed bandits with episode context.
\newblock {\em Annals of Mathematics and Artificial Intelligence},
  61(3):203--230, 2011.

\bibitem{rothman2021transformers}
Denis Rothman.
\newblock {\em Transformers for Natural Language Processing}.
\newblock Packt Publishing, 2021.

\bibitem{rubens2015active}
Neil Rubens, Mehdi Elahi, Masashi Sugiyama, and Dain Kaplan.
\newblock Active learning in recommender systems.
\newblock In {\em Recommender Systems Handbook}, pages 809--846. Springer,
  2015.

\bibitem{rubin2011computer}
Jonathan Rubin and Ian Watson.
\newblock Computer poker: A review.
\newblock {\em Artificial intelligence}, 175(5-6):958--987, 2011.

\bibitem{ruder2016overview}
Sebastian Ruder.
\newblock An overview of gradient descent optimization algorithms.
\newblock {\em arXiv preprint arXiv:1609.04747}, 2016.

\bibitem{rudin2019stop}
Cynthia Rudin.
\newblock Stop explaining black box machine learning models for high stakes
  decisions and use interpretable models instead.
\newblock {\em Nature Machine Intelligence}, 1(5):206--215, 2019.

\bibitem{ruijl2014hepgame}
Ben Ruijl, Jos Vermaseren, Aske Plaat, and Jaap van~den Herik.
\newblock Hepgame and the simplification of expressions.
\newblock {\em arXiv preprint arXiv:1405.6369}, 2014.

\bibitem{rummery1994line}
Gavin~A Rummery and Mahesan Niranjan.
\newblock On-line {Q}-learning using connectionist systems.
\newblock Technical report, University of Cambridge, Department of Engineering
  Cambridge, UK, 1994.

\bibitem{russakovsky2015imagenet}
Olga Russakovsky, Jia Deng, Hao Su, Jonathan Krause, Sanjeev Satheesh, Sean Ma,
  Zhiheng Huang, Andrej Karpathy, Aditya Khosla, Michael~S. Bernstein,
  Alexander~C. Berg, and Li~Fei{-}Fei.
\newblock Imagenet large scale visual recognition challenge.
\newblock {\em International Journal of Computer Vision}, 115(3):211--252,
  2015.

\bibitem{russell2016artificial}
Stuart~J Russell and Peter Norvig.
\newblock {\em Artificial intelligence: a modern approach}.
\newblock Pearson Education Limited, Malaysia, 2016.

\bibitem{russo2017tutorial}
Daniel Russo, Benjamin Van~Roy, Abbas Kazerouni, Ian Osband, and Zheng Wen.
\newblock A tutorial on {Thompson} sampling.
\newblock {\em Found. Trends Mach. Learn.}, 11(1):1--96, 2018.

\bibitem{rusu2018meta}
Andrei~A Rusu, Dushyant Rao, Jakub Sygnowski, Oriol Vinyals, Razvan Pascanu,
  Simon Osindero, and Raia Hadsell.
\newblock Meta-learning with latent embedding optimization.
\newblock In {\em International Conference on Learning Representations}, 2019.

\bibitem{ryan2000intrinsic}
Richard~M Ryan and Edward~L Deci.
\newblock Intrinsic and extrinsic motivations: Classic definitions and new
  directions.
\newblock {\em Contemporary Educational Psychology}, 25(1):54--67, 2000.

\bibitem{sabater2002reputation}
Jordi Sabater and Carles Sierra.
\newblock Reputation and social network analysis in multi-agent systems.
\newblock In {\em Proceedings of the First International Joint Conference on
  Autonomous Agents and Multiagent Systems: Part 1}, pages 475--482, 2002.

\bibitem{saha2018}
Sumit Saha.
\newblock A comprehensive guide to convolutional neural networks---the {ELI5}
  way.
  \url{https://towardsdatascience.com/a-comprehensive-guide-to-convolutional-neural-networks-the-eli5-way-3bd2b1164a53}.
\newblock Towards Data Science, 2018.

\bibitem{salimans2017evolution}
Tim Salimans, Jonathan Ho, Xi~Chen, Szymon Sidor, and Ilya Sutskever.
\newblock Evolution strategies as a scalable alternative to reinforcement
  learning.
\newblock {\em arXiv:1703.03864}, 2017.

\bibitem{sallans2004reinforcement}
Brian Sallans and Geoffrey~E Hinton.
\newblock Reinforcement learning with factored states and actions.
\newblock {\em Journal of Machine Learning Research}, 5:1063--1088, Aug 2004.

\bibitem{samvelyan2019starcraft}
Mikayel Samvelyan, Tabish Rashid, Christian~Schroeder De~Witt, Gregory
  Farquhar, Nantas Nardelli, Tim~GJ Rudner, Chia-Man Hung, Philip~HS Torr,
  Jakob Foerster, and Shimon Whiteson.
\newblock The starcraft multi-agent challenge.
\newblock In {\em Proceedings of the 18th International Conference on
  Autonomous Agents and MultiAgent Systems, AAMAS '19, Montreal}, 2019.

\bibitem{sanders2010cuda}
Jason Sanders and Edward Kandrot.
\newblock {\em CUDA by example: an introduction to general-purpose GPU
  programming}.
\newblock Addison-Wesley Professional, 2010.

\bibitem{sandholm2010state}
Tuomas Sandholm.
\newblock The state of solving large incomplete-information games, and
  application to poker.
\newblock {\em AI Magazine}, 31(4):13--32, 2010.

\bibitem{sandholm2015abstraction}
Tuomas Sandholm.
\newblock Abstraction for solving large incomplete-information games.
\newblock In {\em Proceedings of the AAAI Conference on Artificial
  Intelligence}, volume~29, 2015.

\bibitem{santoro2016meta}
Adam Santoro, Sergey Bartunov, Matthew Botvinick, Daan Wierstra, and Timothy
  Lillicrap.
\newblock Meta-learning with memory-augmented neural networks.
\newblock In {\em International Conference on Machine Learning}, pages
  1842--1850, 2016.

\bibitem{santucci2020intrinsically}
Vieri~Giuliano Santucci, Pierre-Yves Oudeyer, Andrew Barto, and Gianluca
  Baldassarre.
\newblock Intrinsically motivated open-ended learning in autonomous robots.
\newblock {\em Frontiers in Neurorobotics}, 13:115, 2020.

\bibitem{schaefer2002}
Steve Schaefer.
\newblock Mathematical recreations.
  \url{http://www.mathrec.org/old/2002jan/solutions.html}, 2002.

\bibitem{schaeffer2008one}
Jonathan Schaeffer.
\newblock {\em One Jump Ahead: Computer Perfection at Checkers}.
\newblock Springer Science \& Business Media, 2008.

\bibitem{schaeffer1996chinook}
Jonathan Schaeffer, Robert Lake, Paul Lu, and Martin Bryant.
\newblock Chinook, the world man-machine checkers champion.
\newblock {\em AI Magazine}, 17(1):21, 1996.

\bibitem{schaeffer2001unifying}
Jonathan Schaeffer, Aske Plaat, and Andreas Junghanns.
\newblock Unifying single-agent and two-player search.
\newblock {\em Information Sciences}, 135(3-4):151--175, 2001.

\bibitem{schaul2015universal}
Tom Schaul, Daniel Horgan, Karol Gregor, and David Silver.
\newblock Universal value function approximators.
\newblock In {\em International Conference on Machine Learning}, pages
  1312--1320, 2015.

\bibitem{schaul2015prioritized}
Tom Schaul, John Quan, Ioannis Antonoglou, and David Silver.
\newblock Prioritized experience replay.
\newblock In {\em International Conference on Learning Representations}, 2016.

\bibitem{schaul2010metalearning}
Tom Schaul and J{\"u}rgen Schmidhuber.
\newblock Metalearning.
\newblock {\em Scholarpedia}, 5(6):4650, 2010.

\bibitem{schleich2019value}
Daniel Schleich, Tobias Klamt, and Sven Behnke.
\newblock Value iteration networks on multiple levels of abstraction.
\newblock In {\em Robotics: Science and Systems XV, University of Freiburg,
  Freiburg im Breisgau, Germany}, 2019.

\bibitem{schmidhuber1987evolutionary}
J{\"u}rgen Schmidhuber.
\newblock {\em Evolutionary Principles in Self-Referential Learning, or on
  Learning how to Learn: the Meta-Meta-$\ldots$ Hook}.
\newblock PhD thesis, Technische Universit{\"a}t M{\"u}nchen, 1987.

\bibitem{schmidhuber1990making}
J\"urgen Schmidhuber.
\newblock Making the world differentiable: On using self-supervised fully
  recurrent neural networks for dynamic reinforcement learning and planning in
  non-stationary environments.
\newblock Technical report, Inst. f{\"u}r Informatik, 1990.

\bibitem{schmidhuber1990line}
J{\"u}rgen Schmidhuber.
\newblock An on-line algorithm for dynamic reinforcement learning and planning
  in reactive environments.
\newblock In {\em 1990 IJCNN International Joint Conference on Neural
  Networks}, pages 253--258. IEEE, 1990.

\bibitem{schmidhuber1991curious}
J{\"u}rgen Schmidhuber.
\newblock Curious model-building control systems.
\newblock In {\em Proceedings International Joint Conference on Neural
  Networks}, pages 1458--1463, 1991.

\bibitem{schmidhuber1991learning}
J{\"u}rgen Schmidhuber.
\newblock Learning to generate sub-goals for action sequences.
\newblock In {\em Artificial neural networks}, pages 967--972, 1991.

\bibitem{schmidhuber1991possibility}
J{\"u}rgen Schmidhuber.
\newblock A possibility for implementing curiosity and boredom in
  model-building neural controllers.
\newblock In {\em Proc. of the international conference on simulation of
  adaptive behavior: From animals to animats}, pages 222--227, 1991.

\bibitem{schmidhuber2002learning}
J{\"u}rgen Schmidhuber, F~Gers, and Douglas Eck.
\newblock Learning nonregular languages: A comparison of simple recurrent
  networks and {LSTM}.
\newblock {\em Neural Computation}, 14(9):2039--2041, 2002.

\bibitem{schmidhuber1996simple}
J\"urgen Schmidhuber, Jieyu Zhao, and MA~Wiering.
\newblock Simple principles of metalearning.
\newblock Technical report, IDSIA, 1996.

\bibitem{scholkopf1997kernel}
Bernhard Sch{\"o}lkopf, Alexander Smola, and Klaus-Robert M{\"u}ller.
\newblock Kernel principal component analysis.
\newblock In {\em International Conference on Artificial Neural Networks},
  pages 583--588. Springer, 1997.

\bibitem{schraudolph1994temporal}
Nicol~N Schraudolph, Peter Dayan, and Terrence~J Sejnowski.
\newblock Temporal difference learning of position evaluation in the game of
  {Go}.
\newblock In {\em Advances in Neural Information Processing Systems}, pages
  817--824, 1994.

\bibitem{schrittwieser2020mastering}
Julian Schrittwieser, Ioannis Antonoglou, Thomas Hubert, Karen Simonyan,
  Laurent Sifre, Simon Schmitt, Arthur Guez, Edward Lockhart, Demis Hassabis,
  Thore Graepel, Timothy Lillicrap, and David Silver.
\newblock Mastering {Atari}, go, chess and shogi by planning with a learned
  model.
\newblock {\em Nature}, 588(7839):604--609, 2020.

\bibitem{schrittwieser2021online}
Julian Schrittwieser, Thomas Hubert, Amol Mandhane, Mohammadamin Barekatain,
  Ioannis Antonoglou, and David Silver.
\newblock Online and offline reinforcement learning by planning with a learned
  model.
\newblock {\em arXiv preprint arXiv:2104.06294}, 2021.

\bibitem{schulman2015trust}
John Schulman, Sergey Levine, Pieter Abbeel, Michael Jordan, and Philipp
  Moritz.
\newblock Trust region policy optimization.
\newblock In {\em International Conference on Machine Learning}, pages
  1889--1897, 2015.

\bibitem{schulman2015high}
John Schulman, Philipp Moritz, Sergey Levine, Michael Jordan, and Pieter
  Abbeel.
\newblock High-dimensional continuous control using generalized advantage
  estimation.
\newblock In {\em International Conference on Learning Representations}, 2016.

\bibitem{schulman2017proximal}
John Schulman, Filip Wolski, Prafulla Dhariwal, Alec Radford, and Oleg Klimov.
\newblock Proximal policy optimization algorithms.
\newblock {\em arXiv preprint arXiv:1707.06347}, 2017.

\bibitem{schweighofer2003meta}
Nicolas Schweighofer and Kenji Doya.
\newblock Meta-learning in reinforcement learning.
\newblock {\em Neural Networks}, 16(1):5--9, 2003.

\bibitem{scutari2009learning}
Marco Scutari.
\newblock Learning {Bayesian} networks with the bnlearn {R} package.
\newblock {\em Journal of Statistical Software}, 35(i03), 2010.

\bibitem{seeley1989honey}
Thomas~D Seeley.
\newblock The honey bee colony as a superorganism.
\newblock {\em American Scientist}, 77(6):546--553, 1989.

\bibitem{segler2018planning}
Marwin~HS Segler, Mike Preuss, and Mark~P Waller.
\newblock Planning chemical syntheses with deep neural networks and symbolic
  {AI}.
\newblock {\em Nature}, 555(7698):604, 2018.

\bibitem{sekar2020planning}
Ramanan Sekar, Oleh Rybkin, Kostas Daniilidis, Pieter Abbeel, Danijar Hafner,
  and Deepak Pathak.
\newblock Planning to explore via self-supervised world models.
\newblock In {\em International Conference on Machine Learning}, 2020.

\bibitem{selfridge1985training}
Oliver~G Selfridge, Richard~S Sutton, and Andrew~G Barto.
\newblock Training and tracking in robotics.
\newblock In {\em International Joint Conference on Artificial Intelligence},
  pages 670--672, 1985.

\bibitem{senior2020improved}
Andrew~W. Senior, Richard Evans, John Jumper, James Kirkpatrick, Laurent Sifre,
  Tim Green, Chongli Qin, Augustin Z{\'{\i}}dek, Alexander W.~R. Nelson, Alex
  Bridgland, Hugo Penedones, Stig Petersen, Karen Simonyan, Steve Crossan,
  Pushmeet Kohli, David~T. Jones, David Silver, Koray Kavukcuoglu, and Demis
  Hassabis.
\newblock Improved protein structure prediction using potentials from deep
  learning.
\newblock {\em Nature}, 577(7792):706--710, 2020.

\bibitem{settles2009active}
Burr Settles.
\newblock Active learning literature survey.
\newblock Technical report, University of Wisconsin-Madison Department of
  Computer Sciences, 2009.

\bibitem{shaker2016procedural}
Noor Shaker, Julian Togelius, and Mark~J Nelson.
\newblock {\em Procedural Content Generation in Games}.
\newblock Springer, 2016.

\bibitem{shani2013survey}
Guy Shani, Joelle Pineau, and Robert Kaplow.
\newblock A survey of point-based {POMDP} solvers.
\newblock {\em Autonomous Agents and Multi-Agent Systems}, 27(1):1--51, 2013.

\bibitem{shannon1988programming}
Claude~E Shannon.
\newblock Programming a computer for playing chess.
\newblock In {\em Computer Chess Compendium}, pages 2--13. Springer, 1988.

\bibitem{shapley1953stochastic}
Lloyd~S Shapley.
\newblock Stochastic games.
\newblock In {\em Proceedings of the National Academy of Sciences}, volume~39,
  pages 1095--1100, 1953.

\bibitem{shoham2021solving}
Yaron Shoham and Gal Elidan.
\newblock Solving {Sokoban} with forward-backward reinforcement learning.
\newblock In {\em Proceedings of the International Symposium on Combinatorial
  Search}, volume~12, pages 191--193, 2021.

\bibitem{shoham2008multiagent}
Yoav Shoham and Kevin Leyton-Brown.
\newblock {\em Multiagent Systems: Algorithmic, Game-Theoretic, and Logical
  Foundations}.
\newblock Cambridge University Press, 2008.

\bibitem{shoham2003multi}
Yoav Shoham, Rob Powers, and Trond Grenager.
\newblock Multi-agent reinforcement learning: a critical survey.
\newblock Technical report, Stanford University, 2003.

\bibitem{shyam2017attentive}
Pranav Shyam, Shubham Gupta, and Ambedkar Dukkipati.
\newblock Attentive recurrent comparators.
\newblock In {\em International Conference on Machine Learning}, pages
  3173--3181. PMLR, 2017.

\bibitem{sickles2019measurement}
Robin~C Sickles and Valentin Zelenyuk.
\newblock {\em Measurement of productivity and efficiency}.
\newblock Cambridge University Press, 2019.

\bibitem{silver2013lifelong}
Daniel~L Silver, Qiang Yang, and Lianghao Li.
\newblock Lifelong machine learning systems: Beyond learning algorithms.
\newblock In {\em 2013 AAAI Spring Symposium Series}, 2013.

\bibitem{silver2009reinforcement}
David Silver.
\newblock {\em Reinforcement learning and simulation based search in the game
  of {Go}}.
\newblock PhD thesis, University of Alberta, 2009.

\bibitem{silver2016mastering}
David Silver, Aja Huang, Chris~J. Maddison, Arthur Guez, Laurent Sifre, George
  van~den Driessche, Julian Schrittwieser, Ioannis Antonoglou, Veda
  Panneershelvam, Marc Lanctot, Sander Dieleman, Dominik Grewe, John Nham, Nal
  Kalchbrenner, Ilya Sutskever, Timothy Lillicrap, Madeleine Leach, Koray
  Kavukcuoglu, Thore Graepel, and Demis Hassabis.
\newblock Mastering the game of {Go} with deep neural networks and tree search.
\newblock {\em Nature}, 529(7587):484, 2016.

\bibitem{silver2018general}
David Silver, Thomas Hubert, Julian Schrittwieser, Ioannis Antonoglou, Matthew
  Lai, Arthur Guez, Marc Lanctot, Laurent Sifre, Dharshan Kumaran, Thore
  Graepel, Timothy Lillicrap, Karen Simonyan, and Demis Hassabis.
\newblock A general reinforcement learning algorithm that masters chess, shogi,
  and {Go} through self-play.
\newblock {\em Science}, 362(6419):1140--1144, 2018.

\bibitem{silver2014deterministic}
David Silver, Guy Lever, Nicolas Heess, Thomas Degris, Daan Wierstra, and
  Martin Riedmiller.
\newblock Deterministic policy gradient algorithms.
\newblock In {\em International Conference on Machine Learning}, pages
  387--395, 2014.

\bibitem{silver2017mastering}
David Silver, Julian Schrittwieser, Karen Simonyan, Ioannis Antonoglou, Aja
  Huang, Arthur Guez, Thomas Hubert, Lucas Baker, Matthew Lai, Adrian Bolton,
  Yutian Chen, Timothy Lillicrap, Fan Hui, Laurent Sifre, George van~den
  Driessche, Thore Graepel, and Demis Hassabis.
\newblock Mastering the game of {Go} without human knowledge.
\newblock {\em Nature}, 550(7676):354, 2017.

\bibitem{silver2021reward}
David Silver, Satinder Singh, Doina Precup, and Richard~S Sutton.
\newblock Reward is enough.
\newblock {\em Artificial Intelligence}, page 103535, 2021.

\bibitem{silver2007reinforcement}
David Silver, Richard~S Sutton, and Martin M{\"u}ller.
\newblock Reinforcement learning of local shape in the game of {Go}.
\newblock In {\em International Joint Conference on Artificial Intelligence},
  volume~7, pages 1053--1058, 2007.

\bibitem{silver2012temporal}
David Silver, Richard~S Sutton, and Martin M{\"u}ller.
\newblock Temporal-difference search in computer {Go}.
\newblock {\em Machine Learning}, 87(2):183--219, 2012.

\bibitem{silver2017predictron}
David Silver, Hado van Hasselt, Matteo Hessel, Tom Schaul, Arthur Guez, Tim
  Harley, Gabriel Dulac-Arnold, David Reichert, Neil Rabinowitz, Andre Barreto,
  and Thomas Degris.
\newblock The predictron: End-to-end learning and planning.
\newblock In {\em Proceedings of the 34th International Conference on Machine
  Learning}, pages 3191--3199, 2017.

\bibitem{simoes2020multi}
David Sim{\~o}es, Nuno Lau, and Lu{\'\i}s~Paulo Reis.
\newblock Multi agent deep learning with cooperative communication.
\newblock {\em Journal of Artificial Intelligence and Soft Computing Research},
  10, 2020.

\bibitem{singh2005intrinsically}
Satinder Singh, Andrew~G Barto, and Nuttapong Chentanez.
\newblock Intrinsically motivated reinforcement learning.
\newblock Technical report, University of Amherst, Mass, Department of Computer
  Science, 2005.

\bibitem{singh2010intrinsically}
Satinder Singh, Richard~L Lewis, Andrew~G Barto, and Jonathan Sorg.
\newblock Intrinsically motivated reinforcement learning: An evolutionary
  perspective.
\newblock {\em IEEE Transactions on Autonomous Mental Development},
  2(2):70--82, 2010.

\bibitem{slate1983chess}
David~J Slate and Lawrence~R Atkin.
\newblock Chess 4.5---the northwestern university chess program.
\newblock In {\em Chess skill in Man and Machine}, pages 82--118. Springer,
  1983.

\bibitem{smith2015analog}
Gillian Smith.
\newblock An analog history of procedural content generation.
\newblock In {\em Foundations of Digital Games}, 2015.

\bibitem{smith1998computer}
Stephen~J Smith, Dana Nau, and Tom Throop.
\newblock Computer bridge: A big win for {AI} planning.
\newblock {\em AI magazine}, 19(2):93--93, 1998.

\bibitem{snell2017prototypical}
Jake Snell, Kevin Swersky, and Richard Zemel.
\newblock Prototypical networks for few-shot learning.
\newblock In {\em Advances in Neural Information Processing Systems}, pages
  4077--4087, 2017.

\bibitem{sobol2018visual}
Doron Sobol, Lior Wolf, and Yaniv Taigman.
\newblock Visual analogies between {Atari} games for studying transfer learning
  in {RL}.
\newblock {\em arXiv preprint arXiv:1807.11074}, 2018.

\bibitem{sohn2018hierarchical}
Sungryull Sohn, Junhyuk Oh, and Honglak Lee.
\newblock Hierarchical reinforcement learning for zero-shot generalization with
  subtask dependencies.
\newblock In {\em Advances in Neural Information Processing Systems}, pages
  7156--7166, 2018.

\bibitem{son2019qtran}
Kyunghwan Son, Daewoo Kim, Wan~Ju Kang, David~Earl Hostallero, and Yung Yi.
\newblock Qtran: Learning to factorize with transformation for cooperative
  multi-agent reinforcement learning.
\newblock In {\em International Conference on Machine Learning}, pages
  5887--5896. PMLR, 2019.

\bibitem{song2014scaling}
Fengguang Song and Jack Dongarra.
\newblock Scaling up matrix computations on shared-memory manycore systems with
  1000 {CPU} cores.
\newblock In {\em Proceedings of the 28th ACM International Conference on
  Supercomputing}, pages 333--342. ACM, 2014.

\bibitem{song2019v}
H.~Francis Song, Abbas Abdolmaleki, Jost~Tobias Springenberg, Aidan Clark,
  Hubert Soyer, Jack~W. Rae, Seb Noury, Arun Ahuja, Siqi Liu, Dhruva Tirumala,
  Nicolas Heess, Dan Belov, Martin~A. Riedmiller, and Matthew~M. Botvinick.
\newblock {V-MPO:} on-policy maximum a posteriori policy optimization for
  discrete and continuous control.
\newblock In {\em International Conference on Learning Representations}, 2019.

\bibitem{song2018mean}
Mei Song, A~Montanari, and P~Nguyen.
\newblock A mean field view of the landscape of two-layers neural networks.
\newblock In {\em Proceedings of the National Academy of Sciences}, volume 115,
  pages E7665--E7671, 2018.

\bibitem{srinivas2018universal}
Aravind Srinivas, Allan Jabri, Pieter Abbeel, Sergey Levine, and Chelsea Finn.
\newblock Universal planning networks.
\newblock In {\em International Conference on Machine Learning}, pages
  4739--4748, 2018.

\bibitem{srivastava2014dropout}
Nitish Srivastava, Geoffrey Hinton, Alex Krizhevsky, Ilya Sutskever, and Ruslan
  Salakhutdinov.
\newblock Dropout: a simple way to prevent neural networks from overfitting.
\newblock {\em The Journal of Machine Learning Research}, 15(1):1929--1958,
  2014.

\bibitem{steinberger2019single}
Eric Steinberger.
\newblock Single deep counterfactual regret minimization.
\newblock {\em arXiv preprint arXiv:1901.07621}, 2019.

\bibitem{stolle2002learning}
Martin Stolle and Doina Precup.
\newblock Learning options in reinforcement learning.
\newblock In {\em International Symposium on Abstraction, Reformulation, and
  Approximation}, pages 212--223. Springer, 2002.

\bibitem{stork2021large}
Lise Stork, Andreas Weber, Jaap van~den Herik, Aske Plaat, Fons Verbeek, and
  Katherine Wolstencroft.
\newblock Large-scale zero-shot learning in the wild: Classifying zoological
  illustrations.
\newblock {\em Ecological Informatics}, 62:101222, 2021.

\bibitem{straus2018alphazero}
Darin Straus.
\newblock Alphazero implementation and tutorial.
  \url{https://towardsdatascience.com/alphazero-implementation-and-tutorial-f4324d65fdfc},
  2018.

\bibitem{su2019one}
Jiawei Su, Danilo~Vasconcellos Vargas, and Kouichi Sakurai.
\newblock One pixel attack for fooling deep neural networks.
\newblock {\em IEEE Transactions on Evolutionary Computation}, 2019.

\bibitem{such2017deep}
Felipe~Petroski Such, Vashisht Madhavan, Edoardo Conti, Joel Lehman, Kenneth~O
  Stanley, and Jeff Clune.
\newblock Deep neuroevolution: Genetic algorithms are a competitive alternative
  for training deep neural networks for reinforcement learning.
\newblock {\em arXiv preprint arXiv:1712.06567}, 2017.

\bibitem{sukhbaatar2018learning}
Sainbayar Sukhbaatar, Emily Denton, Arthur Szlam, and Rob Fergus.
\newblock Learning goal embeddings via self-play for hierarchical reinforcement
  learning.
\newblock {\em arXiv preprint arXiv:1811.09083}, 2018.

\bibitem{sun2016return}
Baochen Sun, Jiashi Feng, and Kate Saenko.
\newblock Return of frustratingly easy domain adaptation.
\newblock In {\em Proceedings of the AAAI Conference on Artificial
  Intelligence}, volume~30, 2016.

\bibitem{sun2006optimization}
Wenyu Sun and Ya-Xiang Yuan.
\newblock {\em Optimization Theory and Methods: Nonlinear Programming},
  volume~1.
\newblock Springer Science \& Business Media, 2006.

\bibitem{sunehag2017value}
Peter Sunehag, Guy Lever, Audrunas Gruslys, Wojciech~Marian Czarnecki, Vinicius
  Zambaldi, Max Jaderberg, Marc Lanctot, Nicolas Sonnerat, Joel~Z Leibo, Karl
  Tuyls, and Thore Graepel.
\newblock Value-decomposition networks for cooperative multi-agent learning.
\newblock In {\em Proceedings of the 17th International Conference on
  Autonomous Agents and MultiAgent Systems, {AAMAS} 2018, Stockholm, Sweden},
  2017.

\bibitem{sunehag2019reinforcement}
Peter Sunehag, Guy Lever, Siqi Liu, Josh Merel, Nicolas Heess, Joel~Z Leibo,
  Edward Hughes, Tom Eccles, and Thore Graepel.
\newblock Reinforcement learning agents acquire flocking and symbiotic
  behaviour in simulated ecosystems.
\newblock In {\em Artificial Life Conference Proceedings}, pages 103--110. MIT
  Press, 2019.

\bibitem{sung2018learning}
Flood Sung, Yongxin Yang, Li~Zhang, Tao Xiang, Philip~HS Torr, and Timothy~M
  Hospedales.
\newblock Learning to compare: Relation network for few-shot learning.
\newblock In {\em Proceedings of the IEEE conference on computer vision and
  pattern recognition}, pages 1199--1208, 2018.

\bibitem{sutskever2008mimicking}
Ilya Sutskever and Vinod Nair.
\newblock Mimicking {Go} experts with convolutional neural networks.
\newblock In {\em International Conf. on Artificial Neural Networks}, pages
  101--110. Springer, 2008.

\bibitem{sutskever2014sequence}
Ilya Sutskever, Oriol Vinyals, and Quoc~V Le.
\newblock Sequence to sequence learning with neural networks.
\newblock In {\em Advances in Neural Information Processing Systems}, pages
  3104--3112, 2014.

\bibitem{sutton1988learning}
Richard~S Sutton.
\newblock Learning to predict by the methods of temporal differences.
\newblock {\em Machine Learning}, 3(1):9--44, 1988.

\bibitem{sutton1990integrated}
Richard~S Sutton.
\newblock Integrated architectures for learning, planning, and reacting based
  on approximating dynamic programming.
\newblock In {\em Machine Learning Proceedings 1990}, pages 216--224. Elsevier,
  1990.

\bibitem{sutton1991dyna}
Richard~S Sutton.
\newblock Dyna, an integrated architecture for learning, planning, and
  reacting.
\newblock {\em ACM Sigart Bulletin}, 2(4):160--163, 1991.

\bibitem{sutton2018introduction}
Richard~S Sutton and Andrew~G Barto.
\newblock {\em Reinforcement learning, An Introduction, Second Edition}.
\newblock MIT Press, 2018.

\bibitem{sutton2000policy}
Richard~S Sutton, David~A McAllester, Satinder~P Singh, and Yishay Mansour.
\newblock Policy gradient methods for reinforcement learning with function
  approximation.
\newblock In {\em Advances in Neural Information Processing Systems}, pages
  1057--1063, 2000.

\bibitem{sutton1999between}
Richard~S Sutton, Doina Precup, and Satinder Singh.
\newblock Between {MDPs} and {semi-MDPs:} a framework for temporal abstraction
  in reinforcement learning.
\newblock {\em Artificial Intelligence}, 112(1-2):181--211, 1999.

\bibitem{sze2017efficient}
Vivienne Sze, Yu-Hsin Chen, Tien-Ju Yang, and Joel~S Emer.
\newblock Efficient processing of deep neural networks: A tutorial and survey.
\newblock {\em Proceedings of the IEEE}, 105(12):2295--2329, 2017.

\bibitem{szegedy2013intriguing}
Christian Szegedy, Wojciech Zaremba, Ilya Sutskever, Joan Bruna, Dumitru Erhan,
  Ian Goodfellow, and Rob Fergus.
\newblock Intriguing properties of neural networks.
\newblock In {\em International Conference on Learning Representations}, 2013.

\bibitem{tamar2016value}
Aviv Tamar, Yi~Wu, Garrett Thomas, Sergey Levine, and Pieter Abbeel.
\newblock Value iteration networks.
\newblock In {\em Advances in Neural Information Processing Systems}, pages
  2154--2162, 2016.

\bibitem{tammelin2014solving}
Oskari Tammelin.
\newblock Solving large imperfect information games using {CFR+}.
\newblock {\em arXiv preprint arXiv:1407.5042}, 2014.

\bibitem{tampuu2017multiagent}
Ardi Tampuu, Tambet Matiisen, Dorian Kodelja, Ilya Kuzovkin, Kristjan Korjus,
  Juhan Aru, Jaan Aru, and Raul Vicente.
\newblock Multiagent cooperation and competition with deep reinforcement
  learning.
\newblock {\em PloS one}, 12(4):e0172395, 2017.

\bibitem{tan1993multi}
Ming Tan.
\newblock Multi-agent reinforcement learning: Independent vs. cooperative
  agents.
\newblock In {\em International Conference on Machine Learning}, pages
  330--337, 1993.

\bibitem{tan2016introduction}
Pang-Ning Tan, Michael Steinbach, and Vipin Kumar.
\newblock {\em Introduction to Data Mining}.
\newblock Pearson Education India, 2016.

\bibitem{tang2018hierarchical}
Hongyao Tang, Jianye Hao, Tangjie Lv, Yingfeng Chen, Zongzhang Zhang, Hangtian
  Jia, Chunxu Ren, Yan Zheng, Zhaopeng Meng, Changjie Fan, and Li~Wang.
\newblock Hierarchical deep multiagent reinforcement learning with temporal
  abstraction.
\newblock {\em arXiv preprint arXiv:1809.09332}, 2018.

\bibitem{tanno2018adaptive}
Ryutaro Tanno, Kai Arulkumaran, Daniel~C Alexander, Antonio Criminisi, and
  Aditya Nori.
\newblock Adaptive neural trees.
\newblock In {\em International Conference on Machine Learning}, pages
  6166--6175, 2019.

\bibitem{tassa2018deepmind}
Yuval Tassa, Yotam Doron, Alistair Muldal, Tom Erez, Yazhe Li, Diego de~Las
  Casas, David Budden, Abbas Abdolmaleki, Josh Merel, Andrew Lefrancq, Timothy
  Lillicrap, and Martin Riedmiller.
\newblock Deepmind control suite.
\newblock {\em arXiv preprint arXiv:1801.00690}, 2018.

\bibitem{tassa2012synthesis}
Yuval Tassa, Tom Erez, and Emanuel Todorov.
\newblock Synthesis and stabilization of complex behaviors through online
  trajectory optimization.
\newblock In {\em 2012 IEEE/RSJ International Conference on Intelligent Robots
  and Systems}, pages 4906--4913, 2012.

\bibitem{tassa2020dm_control}
Yuval Tassa, Saran Tunyasuvunakool, Alistair Muldal, Yotam Doron, Siqi Liu,
  Steven Bohez, Josh Merel, Tom Erez, Timothy Lillicrap, and Nicolas Heess.
\newblock dm\_control: Software and tasks for continuous control.
\newblock {\em arXiv preprint arXiv:2006.12983}, 2020.

\bibitem{taylor2009transfer}
Matthew~E Taylor and Peter Stone.
\newblock Transfer learning for reinforcement learning domains: A survey.
\newblock {\em Journal of Machine Learning Research}, 10(Jul):1633--1685, 2009.

\bibitem{tekofsky2015past}
Shoshannah Tekofsky, Pieter Spronck, Martijn Goudbeek, Aske Plaat, and Jaap
  van~den Herik.
\newblock Past our prime: A study of age and play style development in
  {Battlefield 3}.
\newblock {\em IEEE Transactions on Computational Intelligence and AI in
  Games}, 7(3):292--303, 2015.

\bibitem{terry2020multiplayer}
Justin~K Terry and Benjamin Black.
\newblock Multiplayer support for the arcade learning environment.
\newblock {\em arXiv preprint arXiv:2009.09341}, 2020.

\bibitem{terry2020pettingzoo}
Justin~K Terry, Benjamin Black, Ananth Hari, Luis Santos, Clemens Dieffendahl,
  Niall~L Williams, Yashas Lokesh, Caroline Horsch, and Praveen Ravi.
\newblock Pettingzoo: Gym for multi-agent reinforcement learning.
\newblock {\em arXiv preprint arXiv:2009.14471}, 2020.

\bibitem{tesauro1989neurogammon}
Gerald Tesauro.
\newblock Neurogammon wins {Computer Olympiad}.
\newblock {\em Neural Computation}, 1(3):321--323, 1989.

\bibitem{tesauro1995td}
Gerald Tesauro.
\newblock {TD}-gammon: A self-teaching backgammon program.
\newblock In {\em Applications of Neural Networks}, pages 267--285. Springer,
  1995.

\bibitem{tesauro1995temporal}
Gerald Tesauro.
\newblock Temporal difference learning and {TD}-{G}ammon.
\newblock {\em Communications of the ACM}, 38(3):58--68, 1995.

\bibitem{tesauro2002programming}
Gerald Tesauro.
\newblock Programming backgammon using self-teaching neural nets.
\newblock {\em Artificial Intelligence}, 134(1-2):181--199, 2002.

\bibitem{tessler2017deep}
Chen Tessler, Shahar Givony, Tom Zahavy, Daniel Mankowitz, and Shie Mannor.
\newblock A deep hierarchical approach to lifelong learning in minecraft.
\newblock In {\em Proceedings of the AAAI Conference on Artificial
  Intelligence}, volume~31, 2017.

\bibitem{teyssier2012ordering}
Marc Teyssier and Daphne Koller.
\newblock Ordering-based search: A simple and effective algorithm for learning
  {Bayesian} networks.
\newblock {\em arXiv preprint arXiv:1207.1429}, 2012.

\bibitem{nair2017learning}
Shantanu Thakoor, Surag Nair, and Megha Jhunjhunwala.
\newblock Learning to play othello without human knowledge.
\newblock Stanford University CS238 Final Project Report, 2017.

\bibitem{theodoridis2009pattern}
Sergios Theodoridis and Konstantinos Koutroumbas.
\newblock {\em Pattern recognition}.
\newblock Academic Press, 1999.

\bibitem{thompson1933likelihood}
William~R Thompson.
\newblock On the likelihood that one unknown probability exceeds another in
  view of the evidence of two samples.
\newblock {\em Biometrika}, 25(3/4):285--294, 1933.

\bibitem{thrun1995learning}
Sebastian Thrun.
\newblock Learning to play the game of chess.
\newblock In {\em Advances in Neural Information Processing Systems}, pages
  1069--1076, 1995.

\bibitem{thrun1996learning}
Sebastian Thrun.
\newblock Is learning the $n$-th thing any easier than learning the first?
\newblock In {\em Advances in Neural Information Processing Systems}, pages
  640--646. Morgan Kaufman, 1996.

\bibitem{thrun2012explanation}
Sebastian Thrun.
\newblock {\em Explanation-based neural network learning: A lifelong learning
  approach}, volume 357.
\newblock Springer, 2012.

\bibitem{thrun2012learning}
Sebastian Thrun and Lorien Pratt.
\newblock {\em Learning to Learn}.
\newblock Springer, 2012.

\bibitem{tian2020rethinking}
Yonglong Tian, Yue Wang, Dilip Krishnan, Joshua~B Tenenbaum, and Phillip Isola.
\newblock Rethinking few-shot image classification: a good embedding is all you
  need?
\newblock In {\em European Conference on Computer Vision}, 2020.

\bibitem{tian2017elf}
Yuandong Tian, Qucheng Gong, Wenling Shang, Yuxin Wu, and C~Lawrence Zitnick.
\newblock {ELF}: An extensive, lightweight and flexible research platform for
  real-time strategy games.
\newblock In {\em Advances in Neural Information Processing Systems}, pages
  2659--2669, 2017.

\bibitem{ELFOpenGo2018}
Yuandong Tian, {Jerry Ma}, {Qucheng Gong}, Shubho Sengupta, Zhuoyuan Chen, and
  C.~Lawrence Zitnick.
\newblock {ELF OpenGo}.
\newblock \url{https://github.com/pytorch/ELF}, 2018.

\bibitem{tian2015better}
Yuandong Tian and Yan Zhu.
\newblock Better computer {Go} player with neural network and long-term
  prediction.
\newblock In {\em International Conference on Learning Representations}, 2016.

\bibitem{todorov2007linearly}
Emanuel Todorov.
\newblock Linearly-solvable markov decision problems.
\newblock In {\em Advances in Neural Information Processing Systems}, pages
  1369--1376, 2007.

\bibitem{todorov2012mujoco}
Emanuel Todorov, Tom Erez, and Yuval Tassa.
\newblock {MuJoCo}: A physics engine for model-based control.
\newblock In {\em IEEE/RSJ International Conference on Intelligent Robots and
  Systems (IROS)}, pages 5026--5033, 2012.

\bibitem{togelius2013procedural}
Julian Togelius, Alex~J Champandard, Pier~Luca Lanzi, Michael Mateas, Ana
  Paiva, Mike Preuss, and Kenneth~O Stanley.
\newblock Procedural content generation: Goals, challenges and actionable
  steps.
\newblock In {\em Artificial and Computational Intelligence in Games}. Schloss
  Dagstuhl-Leibniz-Zentrum f\"ur Informatik, 2013.

\bibitem{tommasi2016learning}
Tatiana Tommasi, Martina Lanzi, Paolo Russo, and Barbara Caputo.
\newblock Learning the roots of visual domain shift.
\newblock In {\em European Conference on Computer Vision}, pages 475--482.
  Springer, 2016.

\bibitem{toubman2014dynamic}
Armon Toubman, Jan~Joris Roessingh, Pieter Spronck, Aske Plaat, and Jaap Van
  Den~Herik.
\newblock Dynamic scripting with team coordination in air combat simulation.
\newblock In {\em International Conference on Industrial, Engineering and other
  Applications of Applied Intelligent Systems}, pages 440--449. Springer, 2014.

\bibitem{trenner2020cfr}
Thomas Trenner.
\newblock Beating kuhn poker with {CFR} using python.
  \url{https://ai.plainenglish.io/building-a-poker-ai-part-6-beating-kuhn-poker-with-cfr-using-python-1b4172a6ab2d}.

\bibitem{triantafillou2019meta}
Eleni Triantafillou, Tyler Zhu, Vincent Dumoulin, Pascal Lamblin, Utku Evci,
  Kelvin Xu, Ross Goroshin, Carles Gelada, Kevin Swersky, Pierre-Antoine
  Manzagol, and Hugo Larochelle.
\newblock Meta-dataset: A dataset of datasets for learning to learn from few
  examples.
\newblock In {\em International Conference on Learning Representations}, 2020.

\bibitem{tromp2016}
John Tromp.
\newblock Number of legal {Go} states.
  \url{http://tromp.github.io/go/legal.html}, 2016.

\bibitem{tsitsiklis1997analysis}
John~N Tsitsiklis and Benjamin Van~Roy.
\newblock Analysis of temporal-diffference learning with function
  approximation.
\newblock In {\em Advances in Neural Information Processing Systems}, pages
  1075--1081, 1997.

\bibitem{turing1953digital}
Alan~M Turing.
\newblock {\em Digital Computers Applied to Games}.
\newblock Pitman \& Sons, 1953.

\bibitem{tuyls2018generalised}
Karl Tuyls, Julien Perolat, Marc Lanctot, Joel~Z Leibo, and Thore Graepel.
\newblock A generalised method for empirical game theoretic analysis.
\newblock In {\em Proceedings of the 17th International Conference on
  Autonomous Agents and MultiAgent Systems, {AAMAS} 2018, Stockholm, Sweden},
  2018.

\bibitem{tuyls2012multiagent}
Karl Tuyls and Gerhard Weiss.
\newblock Multiagent learning: Basics, challenges, and prospects.
\newblock {\em AI Magazine}, 33(3):41--41, 2012.

\bibitem{tylkin2021learning}
Paul Tylkin, Goran Radanovic, and David~C Parkes.
\newblock Learning robust helpful behaviors in two-player cooperative atari
  environments.
\newblock In {\em Proceedings of the 20th International Conference on
  Autonomous Agents and MultiAgent Systems}, pages 1686--1688, 2021.

\bibitem{tzeng2017adversarial}
Eric Tzeng, Judy Hoffman, Kate Saenko, and Trevor Darrell.
\newblock Adversarial discriminative domain adaptation.
\newblock In {\em Proceedings of the IEEE Conference on Computer Vision and
  Pattern Recognition}, pages 7167--7176, 2017.

\bibitem{van2008multi}
Wiebe Van~der Hoek and Michael Wooldridge.
\newblock Multi-agent systems.
\newblock {\em Foundations of Artificial Intelligence}, 3:887--928, 2008.

\bibitem{van2013reinforcement}
Michiel Van Der~Ree and Marco Wiering.
\newblock Reinforcement learning in the game of {Othello}: learning against a
  fixed opponent and learning from self-play.
\newblock In {\em IEEE Adaptive Dynamic Programming and Reinforcement
  Learning}, pages 108--115. IEEE, 2013.

\bibitem{van2016lazy}
Max~J van Duijn.
\newblock {\em The Lazy Mindreader: a Humanities Perspective on Mindreading and
  Multiple-Order Intentionality}.
\newblock PhD thesis, Leiden University, 2016.

\bibitem{van2015narrative}
Max~J Van~Duijn, Ineke Sluiter, and Arie Verhagen.
\newblock When narrative takes over: The representation of embedded mindstates
  in {Shakespeare's Othello}.
\newblock {\em Language and Literature}, 24(2):148--166, 2015.

\bibitem{van2019recursive}
Max~J Van~Duijn and Arie Verhagen.
\newblock Recursive embedding of viewpoints, irregularity, and the role for a
  flexible framework.
\newblock {\em Pragmatics}, 29(2):198--225, 2019.

\bibitem{van2008handbook}
Frank Van~Harmelen, Vladimir Lifschitz, and Bruce Porter.
\newblock {\em Handbook of Knowledge Representation}.
\newblock Elsevier, 2008.

\bibitem{van2018deep}
Hado Van~Hasselt, Yotam Doron, Florian Strub, Matteo Hessel, Nicolas Sonnerat,
  and Joseph Modayil.
\newblock Deep reinforcement learning and the deadly triad.
\newblock {\em arXiv:1812.02648}, 2018.

\bibitem{van2016deep}
Hado Van~Hasselt, Arthur Guez, and David Silver.
\newblock Deep reinforcement learning with {Double Q-Learning}.
\newblock In {\em AAAI}, volume~2, page~5. Phoenix, AZ, 2016.

\bibitem{van2012diverse}
Matthijs Van~Leeuwen and Arno Knobbe.
\newblock Diverse subgroup set discovery.
\newblock {\em Data Mining and Knowledge Discovery}, 25(2):208--242, 2012.

\bibitem{van2014multi}
Kristof Van~Moffaert and Ann Now{\'e}.
\newblock Multi-objective reinforcement learning using sets of pareto
  dominating policies.
\newblock {\em Journal of Machine Learning Research}, 15(1):3483--3512, 2014.

\bibitem{van2011compound}
Gerard~JP Van~Westen, J{\"o}rg~K Wegner, Peggy Geluykens, Leen Kwanten, Inge
  Vereycken, Anik Peeters, Adriaan~P IJzerman, Herman~WT van Vlijmen, and
  Andreas Bender.
\newblock Which compound to select in lead optimization? {Prospectively}
  validated proteochemometric models guide preclinical development.
\newblock {\em PloS One}, 6(11):e27518, 2011.

\bibitem{vanschoren2018meta}
Joaquin Vanschoren.
\newblock Meta-learning: A survey.
\newblock {\em arXiv preprint arXiv:1810.03548}, 2018.

\bibitem{vaswani2017attention}
Ashish Vaswani, Noam Shazeer, Niki Parmar, Jakob Uszkoreit, Llion Jones,
  Aidan~N Gomez, Lukasz Kaiser, and Illia Polosukhin.
\newblock Attention is all you need.
\newblock In {\em Advances in Neural Information Processing Systems}, pages
  5998--6008, 2017.

\bibitem{veeriah2021discovery}
Vivek Veeriah, Tom Zahavy, Matteo Hessel, Zhongwen Xu, Junhyuk Oh, Iurii
  Kemaev, Hado van Hasselt, David Silver, and Satinder Singh.
\newblock Discovery of options via meta-learned subgoals.
\newblock {\em arXiv preprint arXiv:2102.06741}, 2021.

\bibitem{vellido2012making}
Alfredo Vellido, Jos{\'e}~David Mart{\'\i}n-Guerrero, and Paulo~JG Lisboa.
\newblock Making machine learning models interpretable.
\newblock In {\em ESANN}, volume~12, pages 163--172, 2012.

\bibitem{vermaseren2000new}
Jos~AM Vermaseren.
\newblock New features of form.
\newblock {\em arXiv preprint math-ph/0010025}, 2000.

\bibitem{vert2004primer}
Jean-Philippe Vert, Koji Tsuda, and Bernhard Sch{\"o}lkopf.
\newblock A primer on kernel methods.
\newblock In {\em Kernel Methods in Computational Biology}, volume~47, pages
  35--70. MIT press Cambridge, MA, 2004.

\bibitem{vezhnevets2016strategic}
Alexander Vezhnevets, Volodymyr Mnih, Simon Osindero, Alex Graves, Oriol
  Vinyals, John Agapiou, and Koray Kavukcuoglu.
\newblock Strategic attentive writer for learning macro-actions.
\newblock In {\em Advances in Neural Information Processing Systems}, pages
  3486--3494, 2016.

\bibitem{vezhnevets2017feudal}
Alexander Vezhnevets, Simon Osindero, Tom Schaul, Nicolas Heess, Max Jaderberg,
  David Silver, and Koray Kavukcuoglu.
\newblock Feudal networks for hierarchical reinforcement learning.
\newblock In {\em Intl Conf on Machine Learning}, pages 3540--3549. PMLR, 2017.

\bibitem{vilalta2002perspective}
Ricardo Vilalta and Youssef Drissi.
\newblock A perspective view and survey of meta-learning.
\newblock {\em Artificial Intelligence Review}, 18(2):77--95, 2002.

\bibitem{vinyals2019grandmaster}
Oriol Vinyals, Igor Babuschkin, Wojciech~M. Czarnecki, Micha{\"{e}}l Mathieu,
  Andrew Dudzik, Junyoung Chung, David~H. Choi, Richard Powell, Timo Ewalds,
  Petko Georgiev, Junhyuk Oh, Dan Horgan, Manuel Kroiss, Ivo Danihelka, Aja
  Huang, Laurent Sifre, Trevor Cai, John~P. Agapiou, Max Jaderberg,
  Alexander~Sasha Vezhnevets, R{\'{e}}mi Leblond, Tobias Pohlen, Valentin
  Dalibard, David Budden, Yury Sulsky, James Molloy, Tom~Le Paine, {\c{C}}aglar
  G{\"{u}}l{\c{c}}ehre, Ziyu Wang, Tobias Pfaff, Yuhuai Wu, Roman Ring, Dani
  Yogatama, Dario W{\"{u}}nsch, Katrina McKinney, Oliver Smith, Tom Schaul,
  Timothy~P. Lillicrap, Koray Kavukcuoglu, Demis Hassabis, Chris Apps, and
  David Silver.
\newblock Grandmaster level in starcraft {II} using multi-agent reinforcement
  learning.
\newblock {\em Nature}, 575(7782):350--354, 2019.

\bibitem{vinyals2016matching}
Oriol Vinyals, Charles Blundell, Timothy Lillicrap, Daan Wierstra, et~al.
\newblock Matching networks for one shot learning.
\newblock In {\em Advances in Neural Information Processing Systems}, pages
  3630--3638, 2016.

\bibitem{vinyals2017starcraft}
Oriol Vinyals, Timo Ewalds, Sergey Bartunov, Petko Georgiev, Alexander~Sasha
  Vezhnevets, Michelle Yeo, Alireza Makhzani, Heinrich K{\"{u}}ttler, John~P.
  Agapiou, Julian Schrittwieser, John Quan, Stephen Gaffney, Stig Petersen,
  Karen Simonyan, Tom Schaul, Hado van Hasselt, David Silver, Timothy~P.
  Lillicrap, Kevin Calderone, Paul Keet, Anthony Brunasso, David Lawrence,
  Anders Ekermo, Jacob Repp, and Rodney Tsing.
\newblock {Starcraft II}: A new challenge for reinforcement learning.
\newblock {\em arXiv:1708.04782}, 2017.

\bibitem{vinyals2015show}
Oriol Vinyals, Alexander Toshev, Samy Bengio, and Dumitru Erhan.
\newblock Show and tell: A neural image caption generator.
\newblock In {\em Proceedings of the IEEE Conference on Computer Vision and
  Pattern Recognition}, pages 3156--3164, 2015.

\bibitem{volz2018evolving}
Vanessa Volz, Jacob Schrum, Jialin Liu, Simon~M Lucas, Adam Smith, and
  Sebastian Risi.
\newblock Evolving mario levels in the latent space of a deep convolutional
  generative adversarial network.
\newblock In {\em Proceedings of the Genetic and Evolutionary Computation
  Conference}, pages 221--228, 2018.

\bibitem{von1944theory}
John Von~Neumann and Oskar Morgenstern.
\newblock {\em Theory of Games and Economic Behavior}.
\newblock Princeton University Press, 1944.

\bibitem{von2007theory}
John Von~Neumann, Oskar Morgenstern, and Harold~William Kuhn.
\newblock {\em Theory of Games and Economic Behavior (commemorative edition)}.
\newblock Princeton University Press, 2007.

\bibitem{vreeken2011krimp}
Jilles Vreeken, Matthijs Van~Leeuwen, and Arno Siebes.
\newblock Krimp: mining itemsets that compress.
\newblock {\em Data Mining and Knowledge Discovery}, 23(1):169--214, 2011.

\bibitem{quoc2018}
Loc Vu-Quoc.
\newblock Neuron and myelinated axon.
  \url{https://commons.wikimedia.org/w/index.php?curid=72816083}, 2018.

\bibitem{walker2004official}
Douglas Walker and Graham Walker.
\newblock {\em The Official Rock Paper Scissors Strategy Guide}.
\newblock Simon and Schuster, 2004.

\bibitem{wang2019alternative}
Hui Wang, Michael Emmerich, Mike Preuss, and Aske Plaat.
\newblock Alternative loss functions in {AlphaZero}-like self-play.
\newblock In {\em 2019 IEEE Symposium Series on Computational Intelligence
  (SSCI)}, pages 155--162, 2019.

\bibitem{wang2020tackling}
Hui Wang, Mike Preuss, Michael Emmerich, and Aske Plaat.
\newblock Tackling {Morpion Solitaire} with {AlphaZero-like Ranked Reward}
  reinforcement learning.
\newblock In {\em 22nd International Symposium on Symbolic and Numeric
  Algorithms for Scientific Computing, {SYNASC} 2020, Timisoara, Romania},
  2020.

\bibitem{wang2021alchemy}
Jane~X. Wang, Michael King, Nicolas Porcel, Zeb Kurth{-}Nelson, Tina Zhu,
  Charlie Deck, Peter Choy, Mary Cassin, Malcolm Reynolds, H.~Francis Song,
  Gavin Buttimore, David~P. Reichert, Neil~C. Rabinowitz, Loic Matthey, Demis
  Hassabis, Alexander Lerchner, and Matthew Botvinick.
\newblock Alchemy: A structured task distribution for meta-reinforcement
  learning.
\newblock {\em arXiv:2102.02926}, 2021.

\bibitem{wang2016learning}
Jane~X. Wang, Zeb Kurth{-}Nelson, Dhruva Tirumala, Hubert Soyer, Joel~Z. Leibo,
  R{\'{e}}mi Munos, Charles Blundell, Dharshan Kumaran, and Matthew Botvinick.
\newblock Learning to reinforcement learn.
\newblock {\em arXiv preprint arXiv:1611.05763}, 2016.

\bibitem{wang2015basic}
Panqu Wang and Garrison~W Cottrell.
\newblock Basic level categorization facilitates visual object recognition.
\newblock {\em arXiv preprint arXiv:1511.04103}, 2015.

\bibitem{wang2019benchmarking}
Tingwu Wang, Xuchan Bao, Ignasi Clavera, Jerrick Hoang, Yeming Wen, Eric
  Langlois, Shunshi Zhang, Guodong Zhang, Pieter Abbeel, and Jimmy Ba.
\newblock Benchmarking model-based reinforcement learning.
\newblock {\em arXiv:1907.02057}, 2019.

\bibitem{wang2020generalizing}
Yaqing Wang, Quanming Yao, James~T Kwok, and Lionel~M Ni.
\newblock Generalizing from a few examples: A survey on few-shot learning.
\newblock {\em ACM Computing Surveys}, 53(3):1--34, 2020.

\bibitem{wang2016dueling}
Ziyu Wang, Tom Schaul, Matteo Hessel, Hado Hasselt, Marc Lanctot, and Nando
  Freitas.
\newblock Dueling network architectures for deep reinforcement learning.
\newblock In {\em International Conference on Machine Learning}, pages
  1995--2003, 2016.

\bibitem{watkins1989learning}
Christopher~JCH Watkins.
\newblock {\em Learning from Delayed Rewards}.
\newblock PhD thesis, King's College, Cambridge, 1989.

\bibitem{weill}
Eddie Weill.
\newblock {LeNet} in {Keras} on {Github}.
  \url{https://github.com/eweill/keras-deepcv/tree/master/models/classification}.

\bibitem{weinshall2018curriculum}
Daphna Weinshall, Gad Cohen, and Dan Amir.
\newblock Curriculum learning by transfer learning: Theory and experiments with
  deep networks.
\newblock In {\em International Conference on Machine Learning}, pages
  5235--5243, 2018.

\bibitem{weiss2016survey}
Karl Weiss, Taghi~M Khoshgoftaar, and DingDing Wang.
\newblock A survey of transfer learning.
\newblock {\em Journal of Big data}, 3(1):1--40, 2016.

\bibitem{wen2019probabilistic}
Ying Wen, Yaodong Yang, Rui Luo, Jun Wang, and Wei Pan.
\newblock Probabilistic recursive reasoning for multi-agent reinforcement
  learning.
\newblock In {\em International Conference on Learning Representations}, 2019.

\bibitem{weng2020meta}
Lilian Weng.
\newblock Meta-learning: Learning to learn fast.
\newblock Lil'Log
  \url{https://lilianweng.github.io/lil-log/2018/11/30/meta-learning.html},
  November 2018.

\bibitem{weng2020curriculum}
Lilian Weng.
\newblock Curriculum for reinforcement learning
  \url{https://lilianweng.github.io/lil-log/2020/01/29/curriculum-for-reinforcement-learning.html}.
\newblock Lil'Log, January 2020.

\bibitem{whiteson2012evolutionary}
Shimon Whiteson.
\newblock Evolutionary computation for reinforcement learning.
\newblock In Marco~A. Wiering and Martijn van Otterlo, editors, {\em
  Reinforcement Learning}, volume~12 of {\em Adaptation, Learning, and
  Optimization}, pages 325--355. Springer, 2012.

\bibitem{whiteson2011protecting}
Shimon Whiteson, Brian Tanner, Matthew~E Taylor, and Peter Stone.
\newblock Protecting against evaluation overfitting in empirical reinforcement
  learning.
\newblock In {\em 2011 IEEE Symposium on Adaptive Dynamic Programming and
  Reinforcement Learning (ADPRL)}, pages 120--127. IEEE, 2011.

\bibitem{wiering2010self}
Marco~A Wiering.
\newblock Self-play and using an expert to learn to play backgammon with
  temporal difference learning.
\newblock {\em JILSA}, 2(2):57--68, 2010.

\bibitem{wiering2014model}
Marco~A Wiering, Maikel Withagen, and M{\u{a}}d{\u{a}}lina~M Drugan.
\newblock Model-based multi-objective reinforcement learning.
\newblock In {\em 2014 IEEE Symposium on Adaptive Dynamic Programming and
  Reinforcement Learning (ADPRL)}, pages 1--6. IEEE, 2014.

\bibitem{wierstra2008natural}
Daan Wierstra, Tom Schaul, Jan Peters, and J\"urgen Schmidhuber.
\newblock Natural evolution strategies.
\newblock In {\em IEEE Congress on Evolutionary Computation}, pages 3381--3387,
  2008.

\bibitem{wilkinson2017introduction}
Nick Wilkinson and Matthias Klaes.
\newblock {\em An Introduction to Behavioral Economics}.
\newblock Macmillan International Higher Education, 2017.

\bibitem{williams1992simple}
Ronald~J Williams.
\newblock Simple statistical gradient-following algorithms for connectionist
  reinforcement learning.
\newblock {\em Machine Learning}, 8(3-4):229--256, 1992.

\bibitem{witten1976apparent}
Ian~H Witten.
\newblock The apparent conflict between estimation and control---a survey of
  the two-armed bandit problem.
\newblock {\em Journal of the Franklin Institute}, 301(1-2):161--189, 1976.

\bibitem{wong2021survey}
Annie Wong, Thomas B\"ack, Anna~V. Kononova, and Aske Plaat.
\newblock Deep multiagent reinforcement learning: Challenges and directions.
\newblock {\em Artificial Intelligence Review}, 2022.

\bibitem{wooldridge2009introduction}
Michael Wooldridge.
\newblock {\em An Introduction to Multiagent Systems}.
\newblock Wiley, 2009.

\bibitem{woolley2010evidence}
Anita~Williams Woolley, Christopher~F Chabris, Alex Pentland, Nada Hashmi, and
  Thomas~W Malone.
\newblock Evidence for a collective intelligence factor in the performance of
  human groups.
\newblock {\em science}, 330(6004):686--688, 2010.

\bibitem{wu2017scalable}
Yuhuai Wu, Elman Mansimov, Roger~B Grosse, Shun Liao, and Jimmy Ba.
\newblock Scalable trust-region method for deep reinforcement learning using
  kronecker-factored approximation.
\newblock In {\em Advances in Neural Information Processing Systems}, pages
  5279--5288, 2017.

\bibitem{wulfmeier2017addressing}
Markus Wulfmeier, Alex Bewley, and Ingmar Posner.
\newblock Addressing appearance change in outdoor robotics with adversarial
  domain adaptation.
\newblock In {\em 2017 IEEE/RSJ International Conference on Intelligent Robots
  and Systems (IROS)}, pages 1551--1558. IEEE, 2017.

\bibitem{xian2018zero}
Yongqin Xian, Christoph~H Lampert, Bernt Schiele, and Zeynep Akata.
\newblock Zero-shot learning---a comprehensive evaluation of the good, the bad
  and the ugly.
\newblock {\em IEEE Transactions on Pattern Analysis and Machine Intelligence},
  41(9):2251--2265, 2018.

\bibitem{xiao2020macro}
Yuchen Xiao, Joshua Hoffman, and Christopher Amato.
\newblock Macro-action-based deep multi-agent reinforcement learning.
\newblock In {\em Conference on Robot Learning}, pages 1146--1161. PMLR, 2020.

\bibitem{xiong2018microsoft}
Wayne Xiong, Lingfeng Wu, Fil Alleva, Jasha Droppo, Xuedong Huang, and Andreas
  Stolcke.
\newblock The microsoft 2017 conversational speech recognition system 2017
  conversational speech recognition system.
\newblock In {\em IEEE International Conference on Acoustics, Speech and Signal
  Processing (ICASSP)}, pages 5934--5938. IEEE, 2018.

\bibitem{xu2008satzilla}
Lin Xu, Frank Hutter, Holger~H Hoos, and Kevin Leyton-Brown.
\newblock Satzilla: portfolio-based algorithm selection for sat.
\newblock {\em Journal of Artificial Intelligence Research}, 32:565--606, 2008.

\bibitem{xu2019macro}
Sijia Xu, Hongyu Kuang, Zhuang Zhi, Renjie Hu, Yang Liu, and Huyang Sun.
\newblock Macro action selection with deep reinforcement learning in starcraft.
\newblock In {\em {AAAI} Artificial Intelligence and Interactive Digital
  Entertainment}, volume~15, pages 94--99, 2019.

\bibitem{xu2020transfer}
Wen Xu, Jing He, and Yanfeng Shu.
\newblock Transfer learning and deep domain adaptation.
\newblock In {\em Advances in Deep Learning}. IntechOpen, 2020.

\bibitem{yang2019exposing}
Xin Yang, Yuezun Li, and Siwei Lyu.
\newblock Exposing deep fakes using inconsistent head poses.
\newblock In {\em ICASSP 2019-2019 IEEE International Conference on Acoustics,
  Speech and Signal Processing}, pages 8261--8265. IEEE, 2019.

\bibitem{yang2020overview}
Yaodong Yang and Jun Wang.
\newblock An overview of multi-agent reinforcement learning from game
  theoretical perspective.
\newblock {\em arXiv preprint arXiv:2011.00583}, 2020.

\bibitem{yang2021transfer}
Zhao Yang, Mike Preuss, and Aske Plaat.
\newblock Transfer learning and curriculum learning in sokoban.
\newblock {\em arXiv preprint arXiv:2105.11702}, 2021.

\bibitem{ye2021mastering}
Weirui Ye, Shaohuai Liu, Thanard Kurutach, Pieter Abbeel, and Yang Gao.
\newblock Mastering {Atari} games with limited data.
\newblock {\em arXiv preprint arXiv:2111.00210}, 2021.

\bibitem{yoon2018bayesian}
Jaesik Yoon, Taesup Kim, Ousmane Dia, Sungwoong Kim, Yoshua Bengio, and Sungjin
  Ahn.
\newblock Bayesian model-agnostic meta-learning.
\newblock In {\em Proceedings of the 32nd International Conference on Neural
  Information Processing Systems}, pages 7343--7353, 2018.

\bibitem{yosinski2014transferable}
Jason Yosinski, Jeff Clune, Yoshua Bengio, and Hod Lipson.
\newblock How transferable are features in deep neural networks?
\newblock In {\em Neural Information Processing Systems}, pages 3320--3328,
  2014.

\bibitem{chao2021surprising}
Chao Yu, Akash Velu, Eugene Vinitsky, Yu~Wang, Alexandre Bayen, and Yi~Wu.
\newblock The surprising effectiveness of {PPO} in cooperative, multi-agent
  games.
\newblock {\em arXiv preprint arXiv:2103.01955}, 2021.

\bibitem{yu2020meta}
Tianhe Yu, Deirdre Quillen, Zhanpeng He, Ryan Julian, Karol Hausman, Chelsea
  Finn, and Sergey Levine.
\newblock Meta-world: A benchmark and evaluation for multi-task and meta
  reinforcement learning.
\newblock In {\em Conference on Robot Learning}, pages 1094--1100. PMLR, 2020.

\bibitem{zeiler2014visualizing}
Matthew~D Zeiler and Rob Fergus.
\newblock Visualizing and understanding convolutional networks.
\newblock In {\em European Conference on Computer Vision}, pages 818--833.
  Springer, 2014.

\bibitem{PhoenixGo2018}
Qinsong Zeng, Jianchang Zhang, Zhanpeng Zeng, Yongsheng Li, Ming Chen, and
  Sifan Liu.
\newblock {PhoenixGo}.
\newblock \url{https://github.com/Tencent/PhoenixGo}, 2018.

\bibitem{zhang2018dissection}
Amy Zhang, Nicolas Ballas, and Joelle Pineau.
\newblock A dissection of overfitting and generalization in continuous
  reinforcement learning.
\newblock {\em arXiv preprint arXiv:1806.07937}, 2018.

\bibitem{zhang2021understanding}
Chiyuan Zhang, Samy Bengio, Moritz Hardt, Benjamin Recht, and Oriol Vinyals.
\newblock Understanding deep learning (still) requires rethinking
  generalization.
\newblock {\em Communications of the ACM}, 64(3):107--115, 2021.

\bibitem{zhang2018study}
Chiyuan Zhang, Oriol Vinyals, Remi Munos, and Samy Bengio.
\newblock A study on overfitting in deep reinforcement learning.
\newblock {\em arXiv preprint arXiv:1804.06893}, 2018.

\bibitem{zhang2019multi}
Kaiqing Zhang, Zhuoran Yang, and Tamer Ba{\c{s}}ar.
\newblock Multi-agent reinforcement learning: A selective overview of theories
  and algorithms.
\newblock {\em arXiv preprint arXiv:1911.10635}, 2019.

\bibitem{zhang2018fully}
Kaiqing Zhang, Zhuoran Yang, Han Liu, Tong Zhang, and Tamer Basar.
\newblock Fully decentralized multi-agent reinforcement learning with networked
  agents.
\newblock In {\em International Conference on Machine Learning}, pages
  5872--5881. PMLR, 2018.

\bibitem{zhang2019transfer}
Lei Zhang.
\newblock Transfer adaptation learning: A decade survey.
\newblock {\em arXiv:1903.04687}, 2019.

\bibitem{zhang2021world}
Lunjun Zhang, Ge~Yang, and Bradly~C Stadie.
\newblock World model as a graph: Learning latent landmarks for planning.
\newblock In {\em International Conference on Machine Learning}, pages
  12611--12620. PMLR, 2021.

\bibitem{zhang2018solar}
Marvin Zhang, Sharad Vikram, Laura Smith, Pieter Abbeel, Matthew~J Johnson, and
  Sergey Levine.
\newblock Solar: {D}eep structured representations for model-based
  reinforcement learning.
\newblock In {\em International Conference on Machine Learning}, pages
  7444--7453, 2019.

\bibitem{zhang2017deeper}
Shangtong Zhang and Richard~S Sutton.
\newblock A deeper look at experience replay.
\newblock {\em arXiv preprint arXiv:1712.01275}, 2017.

\bibitem{zhao2020sim}
Wenshuai Zhao, Jorge~Pe{\~n}a Queralta, and Tomi Westerlund.
\newblock Sim-to-real transfer in deep reinforcement learning for robotics: a
  survey.
\newblock In {\em 2020 IEEE Symposium Series on Computational Intelligence
  (SSCI)}, pages 737--744. IEEE, 2020.

\bibitem{zheng2018deep}
Yan Zheng, Zhaopeng Meng, Jianye Hao, Zongzhang Zhang, Tianpei Yang, and
  Changjie Fan.
\newblock A deep bayesian policy reuse approach against non-stationary agents.
\newblock In {\em 32nd Neural Information Processing Systems}, pages 962--972,
  2018.

\bibitem{zhou2013tabled}
Neng-Fa Zhou and Agostino Dovier.
\newblock A tabled {P}rolog program for solving {S}okoban.
\newblock {\em Fundamenta Informaticae}, 124(4):561--575, 2013.

\bibitem{zhuang2020comprehensive}
Fuzhen Zhuang, Zhiyuan Qi, Keyu Duan, Dongbo Xi, Yongchun Zhu, Hengshu Zhu, Hui
  Xiong, and Qing He.
\newblock A comprehensive survey on transfer learning.
\newblock {\em Proceedings of the IEEE}, 109(1):43--76, 2020.

\bibitem{ziebart2008maximum}
Brian~D Ziebart, Andrew~L Maas, J~Andrew Bagnell, and Anind~K Dey.
\newblock Maximum entropy inverse reinforcement learning.
\newblock In {\em AAAI}, volume~8, pages 1433--1438. Chicago, IL, USA, 2008.

\bibitem{zinkevich2008regret}
Martin Zinkevich, Michael Johanson, Michael Bowling, and Carmelo Piccione.
\newblock Regret minimization in games with incomplete information.
\newblock In {\em Advances in Neural Information Processing Systems}, pages
  1729--1736, 2008.

\bibitem{zintgraf2019fast}
Luisa Zintgraf, Kyriacos Shiarli, Vitaly Kurin, Katja Hofmann, and Shimon
  Whiteson.
\newblock Fast context adaptation via meta-learning.
\newblock In {\em International Conference on Machine Learning}, pages
  7693--7702. PMLR, 2019.

\end{thebibliography}

\cleardoublepage

\addcontentsline{toc}{chapter}{Tables}
\listoftables



\setglossarystyle{altlist}

{\small
\printglossary
}

\printindex

\end{document}